\documentclass[10pt,twocolumn,letterpaper]{article}
\usepackage{silence}
\newif\ifappendix
\newif\ifforceanon\forceanonfalse
\newif\ifincludeemails\includeemailsfalse
\newif\ifacks\acksfalse

\usepackage[pagenumbers]{cvpr}\appendixtrue\ackstrue

\usepackage[utf8]{inputenc}
\usepackage[T1]{fontenc}
\usepackage{setspace}
\usepackage{url}
\usepackage{booktabs}
\usepackage{amsfonts}
\usepackage{nicefrac}
\usepackage{xfrac}
\usepackage{microtype}
\usepackage{color}
\usepackage{graphicx}
\usepackage{tikz}
\usepackage{amsmath,amssymb}
\usepackage{mathtools}
\usepackage[normalem]{ulem}
\usepackage{tabu}
\usepackage{multirow}
\usepackage{pgfplots,pgfplotstable}
\usepgfplotslibrary{fillbetween}
\usepackage{calc}
\usepackage{pbox}
\usepackage{algorithm}
\usepackage{caption}
\usepackage{afterpage}
\usepackage{tcolorbox}
\usepackage{etoolbox}
\usepackage{xkeyval}
\usepackage{listings}

\pgfplotsset{compat=1.14}
\pgfplotsset{compat/show suggested version=false}

\definecolor{cvprblue}{rgb}{0.21,0.49,0.74}
\usepackage[pagebackref,breaklinks,colorlinks,allcolors=.,citecolor=cvprblue]{hyperref}

\makeatletter
\patchcmd{\hyper@makecurrent}{\ifx\Hy@param\Hy@chapterstring\let\Hy@param\Hy@chapapp\fi}{\iftoggle{inappendix}{\@checkappendixparam{section}\@checkappendixparam{subsection}\@checkappendixparam{subsubsection}}{}}{}{\errmessage{failed to patch}}
\newcommand*{\@checkappendixparam}[1]{\def\@checkappendixparamtmp{#1}\ifx\Hy@param\@checkappendixparamtmp\let\Hy@param\Hy@appendixstring\fi}
\newtoggle{inappendix}\togglefalse{inappendix}\apptocmd{\appendix}{\toggletrue{inappendix}}{}{\errmessage{failed to patch}}
\makeatletter

\def\cvprsect#1{\cvprsection{\texorpdfstring{\hskip -1em.~#1}{#1}}}
\def\cvprsubsect#1{\cvprsubsection{\texorpdfstring{\hskip -1em.~#1}{#1}}}

\graphicspath{{./figures/}}
\definecolor{olive}{rgb}{0.5, 0.5, 0.0}
\definecolor{maroon}{rgb}{0.69, 0.19, 0.38}
\definecolor{celestialblue}{rgb}{0.29, 0.59, 0.82}
\definecolor{darkgreen}{rgb}{0.0, 0.6, 0.0}
\definecolor{grey}{rgb}{0.5,0.5,0.5}
\definecolor{darkblue}{rgb}{0.19, 0.19, 0.62}
\definecolor{silver}{rgb}{0.7,0.7,0.7}
\definecolor{darkcyan}{rgb}{0.0, 0.55, 0.55}

\newcommand{\TBD}[1]{{\color{red}#1}}

\newcommand{\OURS}{EDM2}

\def\clap#1{\hbox to 0pt{\hss #1\hss}}%

\newcommand{\norm}[1]{\left\lVert#1\right\rVert}

\newcommand{\tstrut}{\vphantom{$\hat{A}$}}
\newcommand{\unc}{\rlap{${}^{*}$}}

\newcommand{\RR}{\mathbb{R}}

\newcommand{\EE}{\mathbb{E}}
\newcommand{\LL}{\mathcal{L}}
\newcommand{\NN}{\mathcal{N}}
\newcommand{\UU}{\mathcal{U}}
\newcommand{\MM}{\mathcal{M}}
\newcommand{\OO}{\mathcal{O}}

\newcommand{\xx}{\boldsymbol{x}}

\newcommand{\yy}{\boldsymbol{y}}

\newcommand{\cc}{\boldsymbol{c}}

\newcommand{\boldzero}{\mathbf{0}}
\newcommand{\boldI}{\mathbf{I}}
\newcommand{\boldx}{\mathbf{x}}

\newcommand{\boldw}{\mathbf{w}}
\newcommand{\boldW}{\mathbf{W}}
\newcommand{\boldv}{\mathbf{v}}
\newcommand{\boldV}{\mathbf{V}}
\newcommand{\bolda}{\mathbf{a}}
\newcommand{\boldA}{\mathbf{A}}
\newcommand{\boldb}{\mathbf{b}}
\newcommand{\boldp}{\mathbf{p}}
\newcommand{\boldh}{\mathbf{h}}

\newcommand{\boldq}{\mathbf{q}}
\newcommand{\boldQ}{\mathbf{Q}}
\newcommand{\boldk}{\mathbf{k}}
\newcommand{\boldK}{\mathbf{K}}
\newcommand{\boldc}{\mathbf{c}}
\newcommand{\boldqhat}{\boldsymbol{\hat{\mathbf{q}}}}
\newcommand{\boldkhat}{\boldsymbol{\hat{\mathbf{k}}}}
\newcommand{\boldwhat}{\boldsymbol{\hat{\mathbf{w}}}}
\newcommand{\boldchat}{\boldsymbol{\hat{\mathbf{c}}}}

\newcommand{\nnabla}{\nabla\hspace*{-.1em}}
\newcommand{\boldwi}{\boldw_{\hspace*{-.05em}i}}
\newcommand{\boldwhati}{\boldwhat_{\hspace*{-.05em}i}}

\newcommand{\concat}{\oplus}

\newcommand{\config}[1]{{\mdseries\textsc{Config~\small\uppercase{#1}}}}
\newcommand{\configs}[2]{{\mdseries\textsc{Configs~\small\uppercase{#1}--\uppercase{#2}}}}
\newcommand{\vconfig}[1]{{\mdseries\textsc{Config~\scalebox{0.93}{\uppercase{#1}}}}}
\newcommand{\bfconfig}[1]{\textbf{C{\footnotesize{}O\hspace*{0.025em}N\hspace*{0.025em}F\hspace*{0.025em}I\hspace*{0.025em}G}~\small\uppercase{#1}}}

\newcommand\undefcolumntype[1]{\expandafter\let\csname NC@find@#1\endcsname\relax}

\newcommand{\s}{\hphantom{0}}

\newcommand{\signal}{\boldsymbol{y}}

\newcommand{\noise}{\boldsymbol{n}}

\newcommand{\pdata}{p_\text{data}}

\newcommand{\diff}{\mathrm{d}}
\newcommand{\srel}{\sigma_\text{rel}}

\DeclareMathOperator{\Var}{Var}

\DeclareMathOperator{\silu}{silu}
\DeclareMathOperator{\softmax}{softmax}

\newcommand{\stext}[1]{\text{\raisebox{0pt}[0pt][0pt]{#1}}}

\newcommand{\sdata}{\sigma_\text{data}}
\newcommand{\cskip}{c_\text{skip}}
\newcommand{\cout}{c_\text{out}}
\newcommand{\cin}{c_\text{in}}
\newcommand{\cnoise}{c_\text{noise}}
\newcommand{\tcur}{t_c}
\newcommand{\tnext}{t_n}
\newcommand{\ptc}{p_{\tcur\hspace*{-.1em}}}
\newcommand{\ptn}{p_{\tnext\hspace*{-.15em}}}

\newcommand{\vparagraph}[1]{\vspace*{-1.5mm}\paragraph{#1}}

\newcommand{\textlabel}[1]{{\scriptsize #1}}
\newcommand{\datalabel}[1]{\scalebox{0.9}{\scriptsize #1}}
\pgfplotsset{tick label style={font=\scriptsize}}
\pgfplotsset{legend style={font=\scriptsize}}
\pgfplotsset{tick style={draw=none}}
\pgfplotsset{major grid style={gray!40}}
\pgfplotsset{every axis plot/.append style={line width=0.6pt, mark size=1.1pt}}
\pgfplotsset{legend image code/.code={\draw[mark repeat=2, mark phase=2] plot coordinates {(0mm, 0.2mm) (1.5mm, 0.2mm) (3mm, 0.2mm)};}}

\definecolor{C0}{rgb}{0.121569, 0.466667, 0.705882}
\definecolor{C1}{rgb}{1.000000, 0.498039, 0.054902}
\definecolor{C2}{rgb}{0.172549, 0.627451, 0.172549}
\definecolor{C3}{rgb}{0.839216, 0.152941, 0.156863}
\definecolor{C4}{rgb}{0.580392, 0.403922, 0.741176}
\definecolor{C5}{rgb}{0.549020, 0.337255, 0.294118}
\definecolor{C6}{rgb}{0.890196, 0.466667, 0.760784}
\definecolor{C7}{rgb}{0.498039, 0.498039, 0.498039}
\definecolor{C8}{rgb}{0.737255, 0.741176, 0.133333}
\definecolor{C9}{rgb}{0.090196, 0.745098, 0.811765}
\definecolor{C10}{rgb}{1,1,1}
\definecolor{C11}{rgb}{0,0,0}

\newcommand{\hs}[1]{\hspace{#1mm}}
\newcommand{\fillbetween}[3][]{\addplot+[name path=A, draw=none, mark=none, forget plot] #2; \addplot+[name path=B, draw=none, mark=none, forget plot] #3; \addplot[#1] fill between[of=A and B]}

\newcommand{\data}[2][\TBD{???}]{\ldata{#1}{#2}}
\newcommand{\ldata}[2]{\ifdata{#2}\rawdata{#2}\else#1\fi}
\newcommand{\ifdata}[1]{\ifcsname data-#1\endcsname}
\newcommand{\rawdata}[1]{\csname data-#1\endcsname}

\newcommand{\defdata}[2]{\expandafter\newcommand\csname data-#1\endcsname{#2}}

\newcommand{\eval}[2][2]{\pgfkeys{/pgf/fpu}\pgfmathparse{#2}\pgfmathprintnumber[fixed, zerofill, precision=#1, set thousands separator={}]{\pgfmathresult}}

\newcommand{\cidA}{b1}
\newcommand{\cidB}{cfgB-kl}
\newcommand{\cidC}{cfgB-kl-rb-rmb-rgsb-rgm-aa-ff-ca-zi}
\newcommand{\cidD}{cfgB-kl-rb-rmb-rgsb-rgm-aa-ff-ca-zi-eq-nls-wn-lr100}
\newcommand{\cidE}{cfgB-kl-rb-rmb-rgsb-rgm-aa-ff-ca-zi-eq-nls-wn-fwn-lrd-ref70kb}
\newcommand{\cidF}{cfgB-kl-rb-rmb-rgsb-rgm-aa-ff-ca-zi-eq-nls-wn-fwn-lrd-ref70kb-rc-rr-res-pne-nnd}
\newcommand{\cidG}{cfgB-kl-rb-rmb-rgsb-rgm-aa-ff-ca-zi-eq-nls-wn-fwn-lrd-ref70kb-rc-rr-res-pne-nnd-np-rb0.3-lag-log-bc-rd}

\newcommand{\cidXS}{b9-mc128-lr120-lim120}
\newcommand{\cidS}{\cidG}
\newcommand{\cidM}{b9-mc256-lr90-lim90}
\newcommand{\cidL}{b9-mc320-lr80-lim80}
\newcommand{\cidXL}{b9-mc384-lr70-lim70}
\newcommand{\cidXXL}{b9-mc448-lr65-lim65}

\newcommand{\cidXSD}{b9-mc128-lr120-lim120-s256M-drop0.10}
\newcommand{\cidSD}{cfgB-kl-rb-rmb-rgsb-rgm-aa-ff-ca-zi-eq-nls-wn-fwn-lrd-ref70kb-rc-rr-res-pne-nnd-np-rb0.3-lag-log-bc}
\newcommand{\cidMD}{b9-mc256-lr90-lim90-drop0.10}
\newcommand{\cidLD}{b9-mc320-lr80-lim80-drop0.10}
\newcommand{\cidXLD}{b9-mc384-lr70-lim70-drop0.10}
\newcommand{\cidXXLD}{b9-mc448-lr65-lim65-drop0.10}

\newcommand{\cidSfS}{b9-mu-0.8-sigma1.6-ref35k}
\newcommand{\cidSfM}{b9-mu-0.8-sigma1.6-ref35k-mc256-lr90-lim90}
\newcommand{\cidSfL}{b9-mu-0.8-sigma1.6-ref35k-mc320-lr80-lim80}
\newcommand{\cidSfXL}{b9-mu-0.8-sigma1.6-ref35k-mc384-lr70-lim70}

\newcommand{\cidSfSD}{b9-mu-0.8-sigma1.6-ref35k-drop0.10}
\newcommand{\cidSfMD}{b9-mu-0.8-sigma1.6-ref35k-mc256-lr90-lim90-drop0.10}
\newcommand{\cidSfLD}{b9-mu-0.8-sigma1.6-ref35k-mc320-lr80-lim80-drop0.10}
\newcommand{\cidSfXLD}{b9-mu-0.8-sigma1.6-ref35k-mc384-lr70-lim70-drop0.10}

\newcommand{\figQualityComputeScatter}{%
\begin{figure}[t]%
\centering\footnotesize%
\begin{tikzpicture}%
\begin{axis}[
  width={1.12\linewidth}, height={70mm}, grid={major},
  xmin={33}, xmax={3200}, x coord trafo/.code=\pgfmathparse{##1^0.2}, xtick={50, 100, 200, 500, 1000, 2000}, xticklabels={$50$, $100$, $200$, $500$, $1000$, $2000$},
  ymin={1.7}, ymax={30}, y coord trafo/.code=\pgfmathparse{ln(ln(##1))}, ytick={2, 3, 5, 10, 20, 30}, yticklabels={$2$, $3$, $5$, $10$, $20$, \raisebox{-1.5ex}[0ex][0ex]{FID}},
  xlabel={Model complexity (gigaflops per evaluation), ImageNet-512}, x label style={at={(axis description cs:0.5,-0.06)}, anchor=north},
  legend pos={north west}, legend cell align={left},
]
\gdef\did{img512}
\gdef\pdotSize{1.5pt}

\gdef\pmx##1{\addplot[black, opacity=0.08, thin, forget plot] coordinates {(##1,1.7) (##1,30)};}
\gdef\pmy##1{\addplot[black, opacity=0.08, thin, forget plot] coordinates {(33,##1) (3200,##1)};}
\pmx{40}\pmx{60}\pmx{70}\pmx{80}\pmx{90}\pmx{300}\pmx{400}\pmx{600}\pmx{700}\pmx{800}\pmx{900}\pmx{3000}
\pmy{1.8}\pmy{1.9}\pmy{2.1}\pmy{2.2}\pmy{2.3}\pmy{2.4}\pmy{2.5}\pmy{2.6}\pmy{2.7}\pmy{2.8}\pmy{2.9}\pmy{4}\pmy{6}\pmy{7}\pmy{8}\pmy{9}

\gdef\pcNoGuid##1{(\rawdata{\did-##1-gflops0}, \rawdata{\did-##1-fid2})}
\gdef\pcPrevGuid##1{(\rawdata{\did-##1-gflops0}, \rawdata{\did-##1-cfg2})}
\gdef\pcOurGuid##1{(\rawdata{\did-##1-gflops0}, \rawdata{\did-##1-cfg2})}

\addplot[C2, forget plot] coordinates {\pcNoGuid{\cidXS}\pcNoGuid{\cidS}\pcNoGuid{\cidMD}\pcNoGuid{\cidLD}\pcNoGuid{\cidXLD}\pcNoGuid{\cidXXLD}};
\addplot[C4, forget plot] coordinates {\pcOurGuid{\cidXS}\pcOurGuid{\cidS}\pcOurGuid{\cidMD}\pcOurGuid{\cidLD}\pcOurGuid{\cidXLD}\pcOurGuid{\cidXXLD}};

\gdef\pdot##1##2##3##4{\addplot[##1, mark=*, mark size=\pdotSize, forget plot, nodes near coords align={##2}, nodes near coords=\textlabel{##3}] coordinates {##4};}
\pdot{C0}{east}{ADM}{\pcNoGuid{ADM}}
\pdot{C1}{east}{ADM}{\pcPrevGuid{ADM}}
\pdot{C0}{east}{ADM-U}{\pcNoGuid{ADM-U}}
\pdot{C1}{east}{ADM-U}{\pcPrevGuid{ADM-U}}
\pdot{C0}{west}{DiT-XL/2}{\pcNoGuid{DiT-XL}}
\pdot{C1}{east}{DiT-XL/2}{\pcPrevGuid{DiT-XL}}
\pdot{C0}{west}{RIN}{\pcNoGuid{RIN}}
\pdot{C0}{west}{U-ViT, L}{\pcNoGuid{U-ViT-L}}
\pdot{C0}{west}{VDM++}{\pcNoGuid{VDMpp}}
\pdot{C1}{west}{VDM++}{\pcPrevGuid{VDMpp}}
\pdot{C1}{east}{StyleGAN-XL}{\pcPrevGuid{StyleGAN-XL}}
\pdot{C2}{south}{XS}{\pcNoGuid{\cidXS}}
\pdot{C2}{south}{S}{\pcNoGuid{\cidS}}
\pdot{C2}{south}{M}{\pcNoGuid{\cidMD}}
\pdot{C2}{south}{L}{\pcNoGuid{\cidLD}}
\pdot{C2}{south}{XL}{\pcNoGuid{\cidXLD}}
\pdot{C2}{south}{XXL}{\pcNoGuid{\cidXXLD}}
\pdot{C4}{north}{XS}{\pcOurGuid{\cidXS}}
\pdot{C4}{north}{S}{\pcOurGuid{\cidS}}
\pdot{C4}{north}{M}{\pcOurGuid{\cidMD}}
\pdot{C4}{north}{L}{\pcOurGuid{\cidLD}}
\pdot{C4}{north}{XL}{\pcOurGuid{\cidXLD}}
\pdot{C4}{north}{XXL}{\pcOurGuid{\cidXXLD}}

\addlegendimage{C0, only marks, mark size=\pdotSize}\addlegendentry{Previous, no guidance}
\addlegendimage{C1, only marks, mark size=\pdotSize}\addlegendentry{Previous, with guidance}
\addlegendimage{C2, mark=*, mark size=\pdotSize}\addlegendentry{Ours, no guidance}
\addlegendimage{C4, mark=*, mark size=\pdotSize}\addlegendentry{Ours, with guidance}

\end{axis}
\end{tikzpicture}%
\caption{\label{figQualityComputeScatter}%
Our contributions significantly improve the quality of results w.r.t.~model complexity, surpassing the previous state-of-the-art with a 5$\times$ smaller model.
In this plot,
we use gigaflops per single model evaluation as a measure of a model's intrinsic computational complexity;
a similar advantage holds in terms of parameter count, as well as training and sampling cost (see \refappResults{}).
}%
\vspace{-0.5\baselineskip}%
\end{figure}
}%

\newcommand{\figArchitecture}{%
\begin{figure*}[t]%
\centering\footnotesize%
\includegraphics[width=\linewidth, trim={16.5 99 16.5 104.5}, clip, page=1]{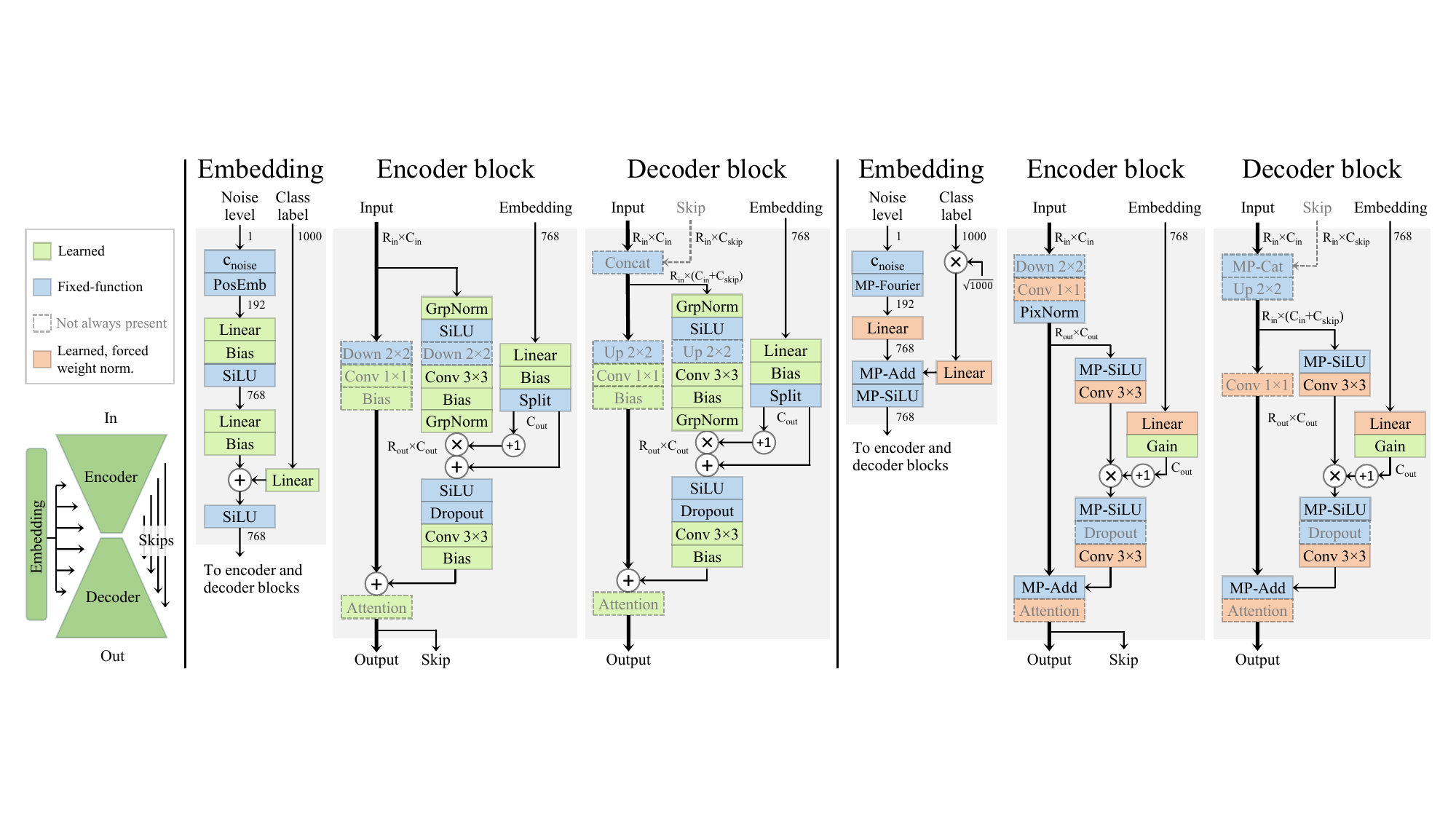}\\%
\makebox[0.114\linewidth]{(a) Overall view}\hfill%
\makebox[0.464\linewidth]{(b) ADM architecture blocks by \citet{Dhariwal2021} (\vconfig{b})}\hfill%
\makebox[0.422\linewidth]{(c) Our magnitude-preserving (MP) variant (\vconfig{g})}\hfill%
\vspace{-1mm}\caption{\label{figArchitecture}%
The widely used ADM architecture~\cite{Dhariwal2021} for image denoising is structured as a U-Net~\cite{Ronneberger2015}.
\textbf{(a)}
The encoder blocks are connected to decoder blocks using skip connections, and an auxiliary embedding network conditions the U-Net with noise level and class label.
\textbf{(b)}
The original building blocks follow the pre-activation design of ResNets~\cite{He2016b}. 
Residual blocks accumulate contributions to the main path (bold). 
Explicit normalizations in the residual paths try to keep magnitudes under control, but nothing prevents them from growing in the main path.
\textbf{(c)}
We update all of the operations (e.g., convolutions, activations, concatenation, summation) to maintain magnitudes on expectation.
}%
\vspace{-1.5mm}%
\end{figure*}
}%

\newcommand{\tabBridgeMain}{%
\begin{table}[t]%
\centering\footnotesize%
\resizebox{\linewidth}{!}{%
\begin{tabu}{|c@{\hs{2}}l|r@{\hs{3.2}}r@{\hs{4.1}}r@{\hs{2.6}}|}
\tabucline{-}
\multicolumn{2}{|l|}{\tstrut\bf Training configurations, ImageNet-512} & \multicolumn{1}{c}{FID $\downarrow$} & \multicolumn{1}{@{}c}{Mparams} & \multicolumn{1}{@{}c|}{Gflops} \\
\tabucline{-}\tstrut%
\gdef\did{img512}%
\gdef\cid{\cidA}{\scalebox{.9}{A}} & {EDM baseline}                          & {\data{\did-\cid-fid2}}    & {\data{\did-\cid-mparams1}} & {\data{\did-\cid-gflops1}} \\
\gdef\cid{\cidB}{\scalebox{.9}{B}} & {+ Minor improvements}                  & {\data{\did-\cid-fid2}}    & {\data{\did-\cid-mparams1}} & {\data{\did-\cid-gflops1}} \\
\gdef\cid{\cidC}{\scalebox{.9}{C}} & {+ Architectural streamlining}          & {\data{\did-\cid-fid2}}    & {\data{\did-\cid-mparams1}} & {\data{\did-\cid-gflops1}} \\
\gdef\cid{\cidD}{\scalebox{.9}{D}} & {+ Magnitude-preserving learned layers} & {\data{\did-\cid-fid2}}    & {\data{\did-\cid-mparams1}} & {\data{\did-\cid-gflops1}} \\
\gdef\cid{\cidE}{\scalebox{.9}{E}} & {+ Control effective learning rate}     & {\data{\did-\cid-fid2}}    & {\data{\did-\cid-mparams1}} & {\data{\did-\cid-gflops1}} \\
\gdef\cid{\cidF}{\scalebox{.9}{F}} & {+ Remove group normalizations}        & {\data{\did-\cid-fid2}}    & {\data{\did-\cid-mparams1}} & {\data{\did-\cid-gflops1}} \\
\gdef\cid{\cidG}{\scalebox{.9}{G}} & {+ Magnitude-preserving fixed-function layers} & {\bf\data{\did-\cid-fid2}} & {\data{\did-\cid-mparams1}} & {\data{\did-\cid-gflops1}} \\
\tabucline{-}
\end{tabu}%
}%
\caption{\label{tabBridgeMain}%
Effect of our changes evaluated on ImageNet-512.
We report Fr\'echet inception distance (FID, lower is better)~\cite{Heusel2017} without guidance, computed between 50,000 randomly generated images and the entire training set.
Each number represents the minimum of three independent evaluations using the same model.
}%
\end{table}
}%

\newcommand{\figMagnitudes}{%
\begin{figure}[t]%
\centering\footnotesize%
\gdef\figMagnitudeAxisWidth{0.40\columnwidth}
\gdef\figMagnitudeAxisHeight{0.20\columnwidth}
\gdef\figMagnitudesTicksAxisMin{0.07}
\gdef\figMagnitudesTicksAxisMax{2.0}
\gdef\figMagnitudesTicks{0.5,1.0,1.5}
\gdef\colorEncA{blue}\gdef\colorEncB{white}
\gdef\colorDecA{red}\gdef\colorDecB{white}
\gdef\colorEncOne{\colorEncA!100!\colorEncB}\gdef\colorEncTwo{\colorEncA!80!\colorEncB}\gdef\colorEncThree{\colorEncA!60!\colorEncB}\gdef\colorEncFour{\colorEncA!40!\colorEncB}
\gdef\colorDecOne{\colorDecA!100!\colorDecB}\gdef\colorDecTwo{\colorDecA!80!\colorDecB}\gdef\colorDecThree{\colorDecA!60!\colorDecB}\gdef\colorDecFour{\colorDecA!40!\colorDecB}
\begin{tabular}{@{}c@{\hs{0.1}}r@{\hs{0.6}}r@{}}%
\rotatebox{90}{\makebox[\figMagnitudeAxisHeight]{Config~\textsc{c}}}&
\begin{tikzpicture}[baseline]
\begin{axis}[
  title={Activations}, title style={at={(0.5,1.15)}, anchor=north},
  xmin=\figMagnitudesTicksAxisMin,xmax=\figMagnitudesTicksAxisMax,ymin=-80, ymax=680,
  xtick={\figMagnitudesTicks}, xticklabels=\empty, ytick={0,200,400,600}, yticklabels={$0$,$200$,$400$,$600$},
  width=\figMagnitudeAxisWidth,height=\figMagnitudeAxisHeight,scale only axis,grid=major,
]
\addplot[\colorEncOne] coordinates {\rawdata{magnitudes-Config-C-1533-41-2147483-activation-norms-enc-64-max}};
\addplot[\colorDecOne] coordinates {\rawdata{magnitudes-Config-C-1533-41-2147483-activation-norms-dec-64-max}};
\addplot[\colorEncTwo] coordinates {\rawdata{magnitudes-Config-C-1533-41-2147483-activation-norms-enc-32-max}};
\addplot[\colorDecTwo] coordinates {\rawdata{magnitudes-Config-C-1533-41-2147483-activation-norms-dec-32-max}};
\addplot[\colorEncThree] coordinates {\rawdata{magnitudes-Config-C-1533-41-2147483-activation-norms-enc-16-max}};
\addplot[\colorDecThree] coordinates {\rawdata{magnitudes-Config-C-1533-41-2147483-activation-norms-dec-16-max}};
\addplot[\colorEncFour] coordinates {\rawdata{magnitudes-Config-C-1533-41-2147483-activation-norms-enc-8-max}};
\addplot[\colorDecFour] coordinates {\rawdata{magnitudes-Config-C-1533-41-2147483-activation-norms-dec-8-max}};
\end{axis}
\end{tikzpicture}&
\begin{tikzpicture}[baseline]
\begin{axis}[
  title={Weights}, title style={at={(0.5,1.15)}, anchor=north},
  xmin=\figMagnitudesTicksAxisMin,xmax=\figMagnitudesTicksAxisMax,ymin=-0.2, ymax=1.7,
  xtick={\figMagnitudesTicks}, xticklabels=\empty, ytick={0.0,0.5,1.0,1.5}, yticklabels={$0$,$0.5$,$1.0$,$1.5$},
  width=\figMagnitudeAxisWidth,height=\figMagnitudeAxisHeight,scale only axis,grid=major,
]
\addplot[\colorEncOne] coordinates {\rawdata{magnitudes-Config-C-1533-41-2147483-weight-norms-enc-64-max}};
\addplot[\colorDecOne] coordinates {\rawdata{magnitudes-Config-C-1533-41-2147483-weight-norms-dec-64-max}};
\addplot[\colorEncTwo] coordinates {\rawdata{magnitudes-Config-C-1533-41-2147483-weight-norms-enc-32-max}};
\addplot[\colorDecTwo] coordinates {\rawdata{magnitudes-Config-C-1533-41-2147483-weight-norms-dec-32-max}};
\addplot[\colorEncThree] coordinates {\rawdata{magnitudes-Config-C-1533-41-2147483-weight-norms-enc-16-max}};
\addplot[\colorDecThree] coordinates {\rawdata{magnitudes-Config-C-1533-41-2147483-weight-norms-dec-16-max}};
\addplot[\colorEncFour] coordinates {\rawdata{magnitudes-Config-C-1533-41-2147483-weight-norms-enc-8-max}};
\addplot[\colorDecFour] coordinates {\rawdata{magnitudes-Config-C-1533-41-2147483-weight-norms-dec-8-max}};
\end{axis}
\end{tikzpicture}\\
\rotatebox{90}{\makebox[\figMagnitudeAxisHeight]{Config~\textsc{d}}}&
\begin{tikzpicture}[baseline]
\begin{axis}[
  xmin=\figMagnitudesTicksAxisMin,xmax=\figMagnitudesTicksAxisMax,ymin=-2, ymax=17,
  xtick={\figMagnitudesTicks}, xticklabels=\empty, ytick={0,5,10,15}, yticklabels={$0$,$5$,$10$,$15$},
  width=\figMagnitudeAxisWidth,height=\figMagnitudeAxisHeight,scale only axis,grid=major,
]
\addplot[\colorEncOne] coordinates {\rawdata{magnitudes-Config-D-1533-45-2147483-activation-norms-enc-64-max}};
\addplot[\colorDecOne] coordinates {\rawdata{magnitudes-Config-D-1533-45-2147483-activation-norms-dec-64-max}};
\addplot[\colorEncTwo] coordinates {\rawdata{magnitudes-Config-D-1533-45-2147483-activation-norms-enc-32-max}};
\addplot[\colorDecTwo] coordinates {\rawdata{magnitudes-Config-D-1533-45-2147483-activation-norms-dec-32-max}};
\addplot[\colorEncThree] coordinates {\rawdata{magnitudes-Config-D-1533-45-2147483-activation-norms-enc-16-max}};
\addplot[\colorDecThree] coordinates {\rawdata{magnitudes-Config-D-1533-45-2147483-activation-norms-dec-16-max}};
\addplot[\colorEncFour] coordinates {\rawdata{magnitudes-Config-D-1533-45-2147483-activation-norms-enc-8-max}};
\addplot[\colorDecFour] coordinates {\rawdata{magnitudes-Config-D-1533-45-2147483-activation-norms-dec-8-max}};
\end{axis}
\end{tikzpicture}&
\begin{tikzpicture}[baseline]
\begin{axis}[
  xmin=\figMagnitudesTicksAxisMin,xmax=\figMagnitudesTicksAxisMax,ymin=-4, ymax=34,
  xtick={\figMagnitudesTicks}, xticklabels=\empty, ytick={0,10,20,30}, yticklabels={$0$,$10$,$20$,$30$},
  width=\figMagnitudeAxisWidth,height=\figMagnitudeAxisHeight,scale only axis,grid=major,
]
\addplot[\colorEncOne] coordinates {\rawdata{magnitudes-Config-D-1533-45-2147483-weight-norms-enc-64-max}};
\addplot[\colorDecOne] coordinates {\rawdata{magnitudes-Config-D-1533-45-2147483-weight-norms-dec-64-max}};
\addplot[\colorEncTwo] coordinates {\rawdata{magnitudes-Config-D-1533-45-2147483-weight-norms-enc-32-max}};
\addplot[\colorDecTwo] coordinates {\rawdata{magnitudes-Config-D-1533-45-2147483-weight-norms-dec-32-max}};
\addplot[\colorEncThree] coordinates {\rawdata{magnitudes-Config-D-1533-45-2147483-weight-norms-enc-16-max}};
\addplot[\colorDecThree] coordinates {\rawdata{magnitudes-Config-D-1533-45-2147483-weight-norms-dec-16-max}};
\addplot[\colorEncFour] coordinates {\rawdata{magnitudes-Config-D-1533-45-2147483-weight-norms-enc-8-max}};
\addplot[\colorDecFour] coordinates {\rawdata{magnitudes-Config-D-1533-45-2147483-weight-norms-dec-8-max}};
\end{axis}
\end{tikzpicture}\\
\rotatebox{90}{\makebox[\figMagnitudeAxisHeight]{Config~\textsc{e}}}&
\begin{tikzpicture}[baseline]
\begin{axis}[
  xmin=\figMagnitudesTicksAxisMin,xmax=\figMagnitudesTicksAxisMax,ymin=-2, ymax=17,
  xtick={0.5,1.0,1.5}, xticklabels={\makebox[0mm][r]{Gimg $=$ }$0.5$,$1.0$,$1.5$}, ytick={0,5,10,15}, yticklabels={$0$,$5$,$10$,$15$},
  width=\figMagnitudeAxisWidth,height=\figMagnitudeAxisHeight,scale only axis,grid=major,
]
\addplot[\colorEncOne] coordinates {\rawdata{magnitudes-Config-E-1533-46-2147483-activation-norms-enc-64-max}};
\addplot[\colorDecOne] coordinates {\rawdata{magnitudes-Config-E-1533-46-2147483-activation-norms-dec-64-max}};
\addplot[\colorEncTwo] coordinates {\rawdata{magnitudes-Config-E-1533-46-2147483-activation-norms-enc-32-max}};
\addplot[\colorDecTwo] coordinates {\rawdata{magnitudes-Config-E-1533-46-2147483-activation-norms-dec-32-max}};
\addplot[\colorEncThree] coordinates {\rawdata{magnitudes-Config-E-1533-46-2147483-activation-norms-enc-16-max}};
\addplot[\colorDecThree] coordinates {\rawdata{magnitudes-Config-E-1533-46-2147483-activation-norms-dec-16-max}};
\addplot[\colorEncFour] coordinates {\rawdata{magnitudes-Config-E-1533-46-2147483-activation-norms-enc-8-max}};
\addplot[\colorDecFour] coordinates {\rawdata{magnitudes-Config-E-1533-46-2147483-activation-norms-dec-8-max}};
\end{axis}
\end{tikzpicture}&
\begin{tikzpicture}[baseline]
\begin{axis}[
  xmin=\figMagnitudesTicksAxisMin,xmax=\figMagnitudesTicksAxisMax,ymin=-0.2, ymax=1.7,
  xtick={\figMagnitudesTicks}, xticklabels={\makebox[0mm][r]{Gimg $=$ }$0.5$,$1.0$,$1.5$}, ytick={0.0,0.5,1.0,1.5}, yticklabels={$0$,$0.5$,$1.0$,$1.5$},
  width=\figMagnitudeAxisWidth,height=\figMagnitudeAxisHeight,scale only axis,grid=major,
  legend pos={south east}, legend cell align={center}, legend columns={2}, legend cell align={left},
  legend style={column sep=1mm,font=\tiny,inner xsep=1pt,inner ysep=1pt,nodes={inner sep=0.5pt,text depth=0.15em,scale=0.75}}
]
\addplot[\colorEncOne] coordinates {\rawdata{magnitudes-Config-E-1533-46-2147483-weight-norms-enc-64-max}};
\addplot[\colorDecOne] coordinates {\rawdata{magnitudes-Config-E-1533-46-2147483-weight-norms-dec-64-max}};
\addplot[\colorEncTwo] coordinates {\rawdata{magnitudes-Config-E-1533-46-2147483-weight-norms-enc-32-max}};
\addplot[\colorDecTwo] coordinates {\rawdata{magnitudes-Config-E-1533-46-2147483-weight-norms-dec-32-max}};
\addplot[\colorEncThree] coordinates {\rawdata{magnitudes-Config-E-1533-46-2147483-weight-norms-enc-16-max}};
\addplot[\colorDecThree] coordinates {\rawdata{magnitudes-Config-E-1533-46-2147483-weight-norms-dec-16-max}};
\addplot[\colorEncFour] coordinates {\rawdata{magnitudes-Config-E-1533-46-2147483-weight-norms-enc-8-max}};
\addplot[\colorDecFour] coordinates {\rawdata{magnitudes-Config-E-1533-46-2147483-weight-norms-dec-8-max}};
\legend{Enc-64x64,Dec-64x64,Enc-32x32,Dec-32x32,Enc-16x16,Dec-16x16,Enc-8x8,Dec-8x8}
\end{axis}
\end{tikzpicture}\\
\end{tabular}%
\caption{\label{figMagnitudes}%
Training-time evolution of activation and weight magnitudes over different depths of the network;
  see \refappResults{} for further details.
\textbf{Top:}
In \vconfig{c}, the magnitudes of both activations and weights grow without bound over training.
\textbf{Middle:}
The magnitude-preserving design introduced in \vconfig{d} curbs activation magnitude growth,
  but leads to even starker growth in weights.
\textbf{Bottom:}
The forced weight normalization in \vconfig{e} ensures that both activations and weights remain bounded.
}%
\vspace{-1mm}%
\end{figure}
}%

\newcommand{\figEmaSnapshots}{%
\begin{figure}[t]%
\centering\footnotesize%
\begin{tikzpicture}%
\begin{axis}[
  width={1.07\linewidth}, height={35mm}, grid={major},
  xmin={0}, xmax={1}, xmode={linear}, xtick={0.0, 1.0}, xticklabels={\hphantom{0},\hphantom{$t_\text{max}$}},
  ymin={0}, ymax={1}, ymode={linear}, ytick={0.0, 0.25, 0.5, 0.75, 1.0}, yticklabels={},
  y label style={at={(axis description cs:-0.075,1.0)},anchor=north},
  ylabel={\vphantom{Relativeweightof$\theta(t)$}},
]
\gdef\gplot##1##2##3{\fillbetween[##1,opacity=0.2,forget plot] {coordinates {(0,0) (##2,0)}} {coordinates {\rawdata{emaprofile-##3}}};\addplot[##1,opacity=0.8,thin] coordinates {\rawdata{emaprofile-##3}};}
\gplot{C0}{0.125}{1-long}\gplot{C9}{0.125}{1-short}
\gplot{C6}{0.250}{2-long}\gplot{C1}{0.250}{2-short}
\gplot{C7}{0.375}{3-long}\gplot{C2}{0.375}{3-short}
\gplot{C8}{0.500}{4-long}\gplot{C3}{0.500}{4-short}
\gplot{C0}{0.625}{5-long}\gplot{C5}{0.625}{5-short}
\gplot{C1}{0.750}{6-long}\gplot{C6}{0.750}{6-short}
\gplot{C5}{0.875}{7-long}\gplot{C7}{0.875}{7-short}
\gplot{C3}{1.000}{8-long}\gplot{C8}{1.000}{8-short}
\addplot[forget plot] coordinates {(0.7,0.98)} node[pos=0.7,below]{\scalebox{0.85}{Snapshots}};
\addplot[black,forget plot] coordinates {(0.0,0.0) (1.0,0.0)};
\end{axis}
\end{tikzpicture}\vspace*{-2.5mm}
\begin{tikzpicture}%
\begin{axis}[
  width={1.07\linewidth}, height={35mm}, grid={major},
  xmin={0}, xmax={1}, xmode={linear}, xtick={0.0, 1.0}, xticklabels={0,$t_\text{max}$},
  ymin={0}, ymax={1}, ymode={linear}, ytick={0.0, 0.25, 0.5, 0.75, 1.0}, yticklabels={},
  x label style={at={(axis description cs:0.5,-0.02)},anchor=north},
  y label style={at={(axis description cs:-0.075,1.0)},anchor=north},
  xlabel={Training time $t$},
  ylabel={\makebox[0mm][c]{\ Rel. weight of $\theta(t)$ in average}},
]
\fillbetween[C4, opacity=0.36, forget plot] {coordinates {(0,0) (1.00,0)}} {coordinates {\rawdata{emaprofile-fit}}};
\addplot[black,dashed] coordinates {\rawdata{emaprofile-target}};
\addplot[C4,thick,opacity=0.8]  coordinates {\rawdata{emaprofile-fit}};
\addplot[forget plot] coordinates {(0.7,0.98)} node[pos=0.7,below]{\scalebox{0.85}{Reconstruction}};
\end{axis}
\end{tikzpicture}%
\caption{\label{figEmaSnapshots}%
\textbf{Top:}
To simulate EMA with arbitrary length after training, we store a number of averaged network parameter snapshots during training.
Each shaded area corresponds to a weighted average of network parameters.
Here, two averages with different power function EMA profiles (\protect\autoref{sec:powerEMA})
are maintained during training and stored at~8 snapshots.
\textbf{Bottom:}
The dashed line shows an example post-hoc EMA to be synthesized,
  and the purple area shows the least-squares optimal approximation based on the stored snapshots.
With two averaged parameter vectors stored per snapshot,
  the mean squared error of the reconstructed weighting profile decreases extremely rapidly as the
  number of snapshots $n$ increases, experimentally in the order of $\OO(1/n^4)$.
In practice, a few dozen snapshots is more than sufficient for a \mbox{virtually perfect EMA reconstruction.}
}%
\end{figure}
}%

\newcommand{\plotEmaBridge}{%
\centering\footnotesize%
\begin{tikzpicture}%
\def\pct{\makebox[0mm][l]{\raisebox{0.15mm}{\scalebox{0.75}{\%}}}}
\begin{axis}[
  width={1.13\columnwidth}, height={60mm}, grid={major},
  xmin={0.02}, xmax={0.25}, xmode={linear}, xtick={0.05, 0.10, 0.15, 0.20, 0.25}, xticklabels={\makebox[0mm][r]{EMA $=$ }$5\pct$, $10\pct$, $15\pct$, $20\pct$, $25\pct$},
  ymin={0.6}, ymax={9}, ymode={linear}, ytick={1, 2, 3, 4, 5, 6, 7, 8, 9}, yticklabels={$1$, $2$, $3$, $4$, $5$, $6$, $7$, $8$, \raisebox{-1.5ex}[0ex][0ex]{FID}},
  legend pos={south east}, legend cell align={left}, legend columns={7}, legend style={/tikz/every even column/.append style={column sep=0.47mm}},
]
\gdef\did{img512}\gdef\rid{phema-2147}

\gdef\pshade##1##2{\fillbetween[##1, opacity=0.2, forget plot]{coordinates {\rawdata{\did-##2-\rid-lo}}}{coordinates {\rawdata{\did-##2-\rid-hi}}};}
\pshade{C1}{\cidB}\pshade{C3}{\cidC}\pshade{C4}{\cidD}\pshade{C5}{\cidE}\pshade{C6}{\cidF}\pshade{C2}{\cidG}

\gdef\pcurve##1##2{\addplot[##1, forget plot] coordinates {\rawdata{\did-##2-\rid-lo}}; \addlegendimage{##1, mark=*}}
\addlegendimage{C0, only marks}\pcurve{C1}{\cidB}\pcurve{C3}{\cidC}\pcurve{C4}{\cidD}\pcurve{C5}{\cidE}\pcurve{C6}{\cidF}\pcurve{C2}{\cidG}

\gdef\pdot##1##2##3##4{\addplot[##1, mark=*, forget plot, nodes near coords align={##3}, nodes near coords=\datalabel{##4}] coordinates {(\rawdata{\did-##2-\rid-phema3}, \rawdata{\did-##2-\rid-fid4})};}
\pdot{C0}{\cidA}{south}{\hs{2.5}$\data{\did-\cidA-\rid-fid2}$}
\pdot{C1}{\cidB}{south}{\raisebox{1mm}{$\data{\did-\cidB-\rid-fid2}$}}
\pdot{C3}{\cidC}{west}{\raisebox{-2.5mm}{\hs{1}$\data{\did-\cidC-\rid-fid2}$}}
\pdot{C4}{\cidD}{north}{$\data{\did-\cidD-\rid-fid2}$}
\pdot{C5}{\cidE}{east}{\raisebox{5mm}{$\data{\did-\cidE-\rid-fid2}$}}
\pdot{C6}{\cidF}{north}{\hs{-1}$\data{\did-\cidF-\rid-fid2}$}
\pdot{C2}{\cidG}{north}{\hs{1}$\data{\did-\cidG-\rid-fid2}$}

\legend{A, B, C, D, E, F, G}
\end{axis}
\end{tikzpicture}%
}%

\newcommand{\plotEmaPerLayer}{%
\centering\footnotesize%
\begin{tikzpicture}%
\def\pct{\makebox[0mm][l]{\raisebox{0.15mm}{\scalebox{0.75}{\%}}}}
\begin{axis}[
    width={1.13\columnwidth}, height={60mm}, grid={major},
    xmin={0.02}, xmax={0.25}, xmode={linear}, xtick={0.05, 0.10, 0.15, 0.20, 0.25}, xticklabels={\makebox[0mm][r]{EMA $=$ }$5\pct$, $10\pct$, $15\pct$, $20\pct$, $25\pct$},
    ymin={0.6}, ymax={9}, ymode={linear}, ytick={1, 2, 3, 4, 5, 6, 7, 8, 9}, yticklabels={$1$, $2$, $3$, $4$, $5$, $6$, $7$, $8$, \raisebox{-1.5ex}[0ex][0ex]{FID}},
    legend pos={south east}, legend cell align={center}, legend columns={3}, legend style={column sep=2.37mm},
  ]
\figPerTensorPlotCommands
\addlegendimage{C1!80!black, thick}
\addlegendimage{C2!80!black, thick}
\addlegendimage{opacity=0.5, legend image code/.code={\draw[C1] plot coordinates {(0mm, 0.7mm) (3.3mm, 0.7mm)}; \draw[C2] plot coordinates {(0mm, -0.3mm) (3.3mm, -0.3mm)};}}
\legend{\hspace*{-1mm}B, \hspace*{-1mm}G, \hspace*{-1mm}Individual tensors}
\end{axis}
\end{tikzpicture}%
}%

\newcommand{\plotEmaEvolution}{%
\centering\footnotesize%
\begin{tikzpicture}%
\def\pct{\makebox[0mm][l]{\raisebox{0.15mm}{\scalebox{0.75}{\%}}}}
\def\CA{C2!40!C10}\def\CB{C2!55!C10}\def\CC{C2!75!C10}\def\CD{C2!99!C11}\def\CE{C2!75!C11}\def\CF{C2!60!C11}
\begin{axis}[
  width={1.13\columnwidth}, height={60mm}, grid={major},
  xmin={0.04}, xmax={0.18}, xmode={linear}, xtick={0.06, 0.08, 0.10, 0.12, 0.14, 0.16}, xticklabels={\makebox[0mm][r]{EMA $=$ }$6\pct$, $8\pct$, $10\pct$, $12\pct$, $14\pct$, $16\pct$},
  ymin={2.2}, ymax={5.5}, ymode={linear}, ytick={2.5, 3.0, 3.5, 4.0, 4.5, 5.0, 5.5}, yticklabels={$2.5$, $3.0$, $3.5$, $4.0$, $4.5$, $5.0$, \raisebox{-1.5ex}[0ex][0ex]{FID}},
  legend pos={north west}, legend cell align={left},
]
\gdef\did{img512}\gdef\cid{\cidG}

\gdef\pshade##1##2{\fillbetween[##1, opacity=0.2, forget plot]{coordinates {\rawdata{\did-\cid-phema-##2-lo}}}{coordinates {\rawdata{\did-\cid-phema-##2-hi}}};}
\pshade{\CA}{268}\pshade{\CB}{537}\pshade{\CC}{1074}\pshade{\CD}{1611}\pshade{\CE}{2147}\pshade{\CF}{2684}

\gdef\pcurve##1##2{\addplot[##1, forget plot] coordinates {\rawdata{\did-\cid-phema-##2-lo}}; \addlegendimage{##1, mark=*}}
\gdef\pcdash##1##2{\addplot[##1, dashed, forget plot] coordinates {\rawdata{\did-\cid-phema-##2-lo}}; \addlegendimage{##1, dashed}}
\pcurve{\CA}{268}\pcurve{\CB}{537}\pcurve{\CC}{1074}\pcurve{\CD}{1611}\pcurve{\CE}{2147}\pcurve{\CF}{2684}

\gdef\pdot##1##2##3##4{\addplot[##1, mark=*, forget plot, nodes near coords align={##3}, nodes near coords=\datalabel{##4}] coordinates {(\rawdata{\did-\cid-phema-##2-phema3}, \rawdata{\did-\cid-phema-##2-fid4})};}
\pdot{\CA}{268}{north}{$\data{\did-\cid-phema-268-fid2}$}
\pdot{\CB}{537}{north}{\hs{2}$\data{\did-\cid-phema-537-fid2}$}
\pdot{\CC}{1074}{north}{\hs{-7}$\data{\did-\cid-phema-1074-fid2}$}
\pdot{\CD}{1611}{north}{$\data{\did-\cid-phema-1611-fid2}$}
\pdot{\CE}{2147}{north}{$\data{\did-\cid-phema-2147-fid2}$}
\pdot{\CF}{2684}{north}{$\data{\did-\cid-phema-2684-fid2}$}

\legend{268M, 537M, 1074M, 1611M, 2147M, 2684M}
\end{axis}
\end{tikzpicture}%
}%

\newcommand{\figEmaTriplet}{%
\begin{figure*}[t]%
\begin{subfigure}{0.33\linewidth}\plotEmaBridge\vspace{1mm}\caption{\small FID vs.~EMA for each training config}\end{subfigure}\hfill%
\begin{subfigure}{0.33\linewidth}\plotEmaPerLayer\vspace{1mm}\caption{\small Per-layer sensitivity to EMA length\label{figPerTensor}}\end{subfigure}\hfill%
\begin{subfigure}{0.33\linewidth}\plotEmaEvolution\vspace{1mm}\caption{\small Evolution of \config{g} over training}\end{subfigure}%
\caption{\label{figEmaTriplet}%
\textbf{(a)}
FID vs.~EMA length for our training configs on ImageNet-512.
\config{a} uses traditional EMA, and thus only a single point is shown.
The shaded regions indicate the min/max FID over 3 evaluations.
\textbf{(b)}
The orange \config{B} is fairly insensitive to the exact EMA length (x-axis) because the network's weight tensors disagree about the optimal EMA length.
We elucidate this by letting the EMA length vary for \emph{one tensor at a time}  (faint lines), while using the globally optimal EMA length of 9\% for the others.
This has a strong effect on FID and, remarkably, sometimes improves it.
In the green \config{G}, the situation is different;
  per-tensor sweeping has a much smaller effect, 
  and deviating from the common optimum of 13\% is detrimental.
\textbf{(c)}
Evolution of the EMA curve for \config{g} over the course of training.
}%
\end{figure*}
}%

\newcommand{\tabResultsFiveTwelve}{%
\begin{table}[t]%
\centering\footnotesize%
\resizebox{\linewidth}{!}{%
\begin{tabu}{|l@{\hs{1.5}}l|r@{\hs{3.5}}c|r@{\hs{4.5}}r@{\hs{3.1}}r@{\hs{2.4}}|}
\tabucline{-}
\multicolumn{2}{|l|}{\multirow{2}{*}{\bf ImageNet-512}} & \multicolumn{2}{c|}{\tstrut FID $\downarrow$} & \multicolumn{3}{c|}{Model size} \\
&& \multicolumn{1}{c}{no CFG} & \multicolumn{1}{@{}c|}{w/CFG} & \multicolumn{1}{c}{Mparams} & \multicolumn{1}{@{}c}{Gflops} & \multicolumn{1}{@{}c|}{NFE} \\
\tabucline{-}\tstrut%
\gdef\did{img512}%
\gdef\cid{ADM}{ADM}                 & {\cite{Dhariwal2021}}   & {\data{\did-\cid-fid2}}       & {\data{\did-\cid-cfg2}}    & {\data{\did-\cid-mparams0}} & {\data{\did-\cid-gflops0}} & {\data{\did-\cid-nfe}} \\
\gdef\cid{DiT-XL}{DiT-XL/2}         & {\cite{Peebles2022}}    & {\data{\did-\cid-fid2}}       & {\data{\did-\cid-cfg2}}    & {\data{\did-\cid-mparams0}} & {\data{\did-\cid-gflops0}} & {\data{\did-\cid-nfe}} \\
\gdef\cid{ADM-U}{ADM-U}             & {\cite{Dhariwal2021}}   & {\data{\did-\cid-fid2}}       & {\data{\did-\cid-cfg2}}    & {\data{\did-\cid-mparams0}} & {\data{\did-\cid-gflops0}} & {\data{\did-\cid-nfe}} \\
\gdef\cid{RIN}{RIN}                 & {\cite{Jabri2023}}      & {\data{\did-\cid-fid2}}       & {\data{\did-\cid-cfg2}}    & {\data{\did-\cid-mparams0}} & {\data{\did-\cid-gflops0}} & {\data{\did-\cid-nfe}} \\
\gdef\cid{U-ViT-L}{U-ViT, L}        & {\cite{Hoogeboom2023b}} & {\data{\did-\cid-fid2}}       & {\data{\did-\cid-cfg2}}    & {\data{\did-\cid-mparams0}} & {\data{\did-\cid-gflops0}}\unc{} & {\data{\did-\cid-nfe}} \\
\gdef\cid{VDMpp}{VDM++}             & {\cite{Kingma2023}}     & {\data{\did-\cid-fid2}}       & {\data{\did-\cid-cfg2}}    & {\data{\did-\cid-mparams0}} & {\data{\did-\cid-gflops0}}\unc{} & {\data{\did-\cid-nfe}} \\
\gdef\cid{StyleGAN-XL}{StyleGAN-XL} & {\cite{Sauer2022}}      & {\data{\did-\cid-fid2}\hs{2}} & {\data{\did-\cid-cfg2}}    & {\data{\did-\cid-mparams0}}\unc{} & {\data{\did-\cid-gflops0}}\unc{} & {\data{\did-\cid-nfe}} \\
\tabucline{-}\tstrut%
\gdef\cid{\cidXS}{\OURS-XS}                             & {}  & {\data{\did-\cid-fid2}}       & {\data{\did-\cid-cfg2}}    & {\data{\did-\cid-mparams0}} & {\data{\did-\cid-gflops0}} & {\data{\did-\cid-nfe}} \\
\gdef\cid{\cidS}{\OURS-S}                               & {}  & {\data{\did-\cid-fid2}}       & {\data{\did-\cid-cfg2}}    & {\data{\did-\cid-mparams0}} & {\data{\did-\cid-gflops0}} & {\data{\did-\cid-nfe}} \\
\gdef\cid{\cidMD}{\OURS-M}                              & {}  & {\data{\did-\cid-fid2}}       & {\data{\did-\cid-cfg2}}    & {\data{\did-\cid-mparams0}} & {\data{\did-\cid-gflops0}} & {\data{\did-\cid-nfe}} \\
\gdef\cid{\cidLD}{\OURS-L}                              & {}  & {\data{\did-\cid-fid2}}       & {\data{\did-\cid-cfg2}}    & {\data{\did-\cid-mparams0}} & {\data{\did-\cid-gflops0}} & {\data{\did-\cid-nfe}} \\
\gdef\cid{\cidXLD}{\OURS-XL}                            & {}  & {\data{\did-\cid-fid2}}       & {\data{\did-\cid-cfg2}}    & {\data{\did-\cid-mparams0}} & {\data{\did-\cid-gflops0}} & {\data{\did-\cid-nfe}} \\
\gdef\cid{\cidXXLD}{\OURS-XXL}                          & {}  & {\bf\data{\did-\cid-fid2}}    & {\bf\data{\did-\cid-cfg2}} & {\data{\did-\cid-mparams0}} & {\data{\did-\cid-gflops0}} & {\data{\did-\cid-nfe}} \\
\tabucline{-}
\end{tabu}%
}%
\caption{\label{tabResultsFiveTwelve}%
Results on ImageNet-512.
``\OURS-S'' is the same as \vconfig{g} in \autoref{tabBridgeMain}.
The ``w/CFG'' and ``no CFG'' columns show the lowest FID obtained with and without classifier-free guidance, respectively.
NFE tells how many times the score function is evaluated when generating an image.
All diffusion models above the horizontal line use stochastic sampling,
  whereas our models below the line use deterministic sampling.
Whether stochastic sampling would improve our results further is left for future work.
Asterisks (${*}$) indicate values that could not be determined from primary sources,
  and have been approximated to within $\sim$10\% accuracy.
}\vspace*{-1mm}%
\end{table}
}%

\newcommand{\figEmaGuidance}{%
\begin{figure}[t]%
\centering\footnotesize%
\begin{tikzpicture}%
\def\pct{\makebox[0mm][l]{\raisebox{0.15mm}{\scalebox{0.75}{\%}}}}
\def\CA{C2}\def\CB{C0}\def\CC{C1}\def\CD{C3}\def\CE{C4}\def\CF{C5}
\begin{axis}[
  width={1.12\columnwidth}, height={60mm}, grid={major},
  xmin={0.01}, xmax={0.15}, xmode={linear}, xtick={0.02, 0.04, 0.06, 0.08, 0.10, 0.12, 0.14}, xticklabels={\makebox[0mm][r]{EMA $=$ }$2\pct$, $4\pct$, $6\pct$, $8\pct$, $10\pct$, $12\pct$, $14\pct$},
  ymin={1.9}, ymax={5}, ymode={linear}, ytick={2.0, 2.5, 3.0, 3.5, 4.0, 4.5, 5.0}, yticklabels={$2.0$, $2.5$, $3.0$, $3.5$, $4.0$, $4.5$, \raisebox{-1.5ex}[0ex][0ex]{FID}},
  legend pos={north west}, legend cell align={left}, legend columns={6},
]
\gdef\did{img512}\gdef\cid{\cidG}\gdef\Mimg{2147}

\gdef\pshade##1##2{\fillbetween[##1, opacity=0.2, forget plot]{coordinates {\rawdata{\did-\cid-guid##2-\Mimg-lo}}}{coordinates {\rawdata{\did-\cid-guid##2-\Mimg-hi}}};}
\pshade{\CA}{1.0}\pshade{\CB}{1.1}\pshade{\CC}{1.2}\pshade{\CD}{1.3}\pshade{\CE}{1.4}\pshade{\CF}{1.5}

\gdef\pcurve##1##2{\addplot[##1, forget plot] coordinates {\rawdata{\did-\cid-guid##2-\Mimg-lo}}; \addlegendimage{##1, mark=*}}
\gdef\pcdash##1##2{\addplot[##1, dashed, forget plot] coordinates {\rawdata{\did-\cid-guid##2-\Mimg-lo}}; \addlegendimage{##1, dashed}}
\pcurve{\CA}{1.0}\pcurve{\CB}{1.1}\pcurve{\CC}{1.2}\pcurve{\CD}{1.3}\pcurve{\CE}{1.4}\pcurve{\CF}{1.5}

\gdef\pdot##1##2##3##4{\addplot[##1, mark=*, forget plot, nodes near coords align={##3}, nodes near coords=\datalabel{##4}] coordinates {(\rawdata{\did-\cid-guid##2-\Mimg-phema3}, \rawdata{\did-\cid-guid##2-\Mimg-fid4})};}
\pdot{\CA}{1.0}{north}{$\data{\did-\cid-guid1.0-\Mimg-fid2}$}
\pdot{\CB}{1.1}{north}{$\data{\did-\cid-guid1.1-\Mimg-fid2}$}
\pdot{\CC}{1.2}{north}{$\data{\did-\cid-guid1.2-\Mimg-fid2}$}
\pdot{\CD}{1.3}{north}{$\data{\did-\cid-guid1.3-\Mimg-fid2}$}
\pdot{\CE}{1.4}{north}{$\data{\did-\cid-guid1.4-\Mimg-fid2}$}
\pdot{\CF}{1.5}{north}{\hs{-3}$\data{\did-\cid-guid1.5-\Mimg-fid2}$}

\legend{1.0 (no guidance), 1.1, 1.2, 1.3, 1.4, 1.5}
\end{axis}
\end{tikzpicture}%
\vspace{-1mm}%
\caption{\label{figEmaGuidance}%
Interaction between EMA length and guidance strength using \OURS-S on ImageNet-512.
}%
\vspace{-1mm}%
\end{figure}
}%

\newcommand{\tabResultsSixtyFour}{%
\begin{table}[t]%
\centering\footnotesize%
\resizebox{\linewidth}{!}{%
\begin{tabu}{|l@{\hs{1.5}}l|c@{\hs{2.5}}c|c@{\hs{2.5}}r@{\hs{2.5}}|r@{\hs{5}}c|}
\tabucline{-}
\multicolumn{2}{|l|}{\multirow{2}{*}{\bf ImageNet-64}} & \multicolumn{2}{c|}{\tstrut Deterministic} & \multicolumn{2}{c|}{Stochastic} & \multicolumn{2}{c|}{Model size} \\
&& \multicolumn{1}{c@{\hs{2.5}}}{FID $\downarrow$} & \multicolumn{1}{@{}c|}{NFE} & \multicolumn{1}{c@{\hs{2.5}}}{FID $\downarrow$} & \multicolumn{1}{@{}c|}{NFE} & \multicolumn{1}{c@{\hs{2.5}}}{Mparams} & \multicolumn{1}{@{}c|}{Gflops} \\
\tabucline{-}\tstrut%
\gdef\did{img64}%
\gdef\cid{ADM}{ADM}                 & {\cite{Dhariwal2021}}           & {\data{\did-\cid-det-fid2}} & {\data{\did-\cid-det-nfe}} & {\data{\did-\cid-stoch-fid2}}    & {\data{\did-\cid-stoch-nfe}} & {\data{\did-\cid-mparams0}} & {\data{\did-\cid-gflops0}} \\
\gdef\cid{EDM-ADM}{+ EDM sampling}  & {\cite{Karras2022elucidating}}  & {\data{\did-\cid-det-fid2}} & {\data{\did-\cid-det-nfe}} & {\data{\did-\cid-stoch-fid2}}    & {\data{\did-\cid-stoch-nfe}} & {\data{\did-\cid-mparams0}} & {\data{\did-\cid-gflops0}} \\
\gdef\cid{EDM-EDM}{+ EDM training}  & {\cite{Karras2022elucidating}}  & {\data{\did-\cid-det-fid2}} & {\data{\did-\cid-det-nfe}} & {\data{\did-\cid-stoch-fid2}}    & {\data{\did-\cid-stoch-nfe}} & {\data{\did-\cid-mparams0}} & {\data{\did-\cid-gflops0}} \\
\gdef\cid{VDMpp}{VDM++}             & {\cite{Kingma2023}}             & {\data{\did-\cid-det-fid2}} & {\data{\did-\cid-det-nfe}} & {\data{\did-\cid-stoch-fid2}}    & {\data{\did-\cid-stoch-nfe}} & {\data{\did-\cid-mparams0}} & {\data{\did-\cid-gflops0}} \\
\gdef\cid{RIN}{RIN}                 & {\cite{Jabri2023}}              & {\data{\did-\cid-det-fid2}} & {\data{\did-\cid-det-nfe}} & {\bf\data{\did-\cid-stoch-fid2}} & {\data{\did-\cid-stoch-nfe}} & {\data{\did-\cid-mparams0}} & {\data{\did-\cid-gflops0}} \\
\tabucline{-}\tstrut%
\gdef\cid{\cidSfS}{\OURS-S}                                     & {}  & {\data{\did-\cid-fid2}}      & {\data{\did-\cid-nfe}}     & {--}                             & {--\hs{2}}                   & {\data{\did-\cid-mparams0}} & {\data{\did-\cid-gflops0}} \\
\gdef\cid{\cidSfMD}{\OURS-M}                                    & {}  & {\data{\did-\cid-fid2}}      & {\data{\did-\cid-nfe}}     & {--}                             & {--\hs{2}}                   & {\data{\did-\cid-mparams0}} & {\data{\did-\cid-gflops0}} \\
\gdef\cid{\cidSfLD}{\OURS-L}                                    & {}  & {\bf\data{\did-\cid-fid2}}   & {\data{\did-\cid-nfe}}     & {--}                             & {--\hs{2}}                   & {\data{\did-\cid-mparams0}} & {\data{\did-\cid-gflops0}} \\
\gdef\cid{\cidSfXLD}{\OURS-XL}                                  & {}  & {\bf\data{\did-\cid-fid2}}   & {\data{\did-\cid-nfe}}     & {--}                             & {--\hs{2}}                   & {\data{\did-\cid-mparams0}} & {\data{\did-\cid-gflops0}} \\
\tabucline{-}
\end{tabu}%
}%
\vspace{-1mm}%
\caption{\label{tabResultsSixtyFour}%
Results on ImageNet-64.
}%
\vspace{-1mm}%
\end{table}
}%

\defdata{img512-b1-fid2}{8.00}
\defdata{img512-b1-cfg2}{\TBD{TBD}}
\defdata{img512-b1-mparams0}{296}
\defdata{img512-b1-mparams1}{295.9}
\defdata{img512-b1-gflops0}{110}
\defdata{img512-b1-gflops1}{110.4}
\defdata{img512-b1-nfe}{63}
\defdata{img512-b1-mimg0}{2147}
\defdata{img512-b1-mimg1}{2147.5}
\defdata{img512-b1-spkimg3}{0.305}
\defdata{img512-b1-pkl}{1532/0/2147483.pkl}

\defdata{img512-cfgB-kl-fid2}{7.24}
\defdata{img512-cfgB-kl-cfg2}{\TBD{TBD}}
\defdata{img512-cfgB-kl-mparams0}{292}
\defdata{img512-cfgB-kl-mparams1}{291.8}
\defdata{img512-cfgB-kl-gflops0}{100}
\defdata{img512-cfgB-kl-gflops1}{100.4}
\defdata{img512-cfgB-kl-nfe}{63}
\defdata{img512-cfgB-kl-mimg0}{2147}
\defdata{img512-cfgB-kl-mimg1}{2147.5}
\defdata{img512-cfgB-kl-spkimg3}{0.212}
\defdata{img512-cfgB-kl-pkl}{1532/29/2147483.pkl}

\defdata{img512-cfgB-kl-rb-rmb-rgsb-rgm-aa-ff-ca-zi-fid2}{6.96}
\defdata{img512-cfgB-kl-rb-rmb-rgsb-rgm-aa-ff-ca-zi-cfg2}{\TBD{TBD}}
\defdata{img512-cfgB-kl-rb-rmb-rgsb-rgm-aa-ff-ca-zi-mparams0}{278}
\defdata{img512-cfgB-kl-rb-rmb-rgsb-rgm-aa-ff-ca-zi-mparams1}{277.8}
\defdata{img512-cfgB-kl-rb-rmb-rgsb-rgm-aa-ff-ca-zi-gflops0}{100}
\defdata{img512-cfgB-kl-rb-rmb-rgsb-rgm-aa-ff-ca-zi-gflops1}{100.3}
\defdata{img512-cfgB-kl-rb-rmb-rgsb-rgm-aa-ff-ca-zi-nfe}{63}
\defdata{img512-cfgB-kl-rb-rmb-rgsb-rgm-aa-ff-ca-zi-mimg0}{2147}
\defdata{img512-cfgB-kl-rb-rmb-rgsb-rgm-aa-ff-ca-zi-mimg1}{2147.5}
\defdata{img512-cfgB-kl-rb-rmb-rgsb-rgm-aa-ff-ca-zi-spkimg3}{0.226}
\defdata{img512-cfgB-kl-rb-rmb-rgsb-rgm-aa-ff-ca-zi-pkl}{1533/41/2147483.pkl}

\defdata{img512-cfgB-kl-rb-rmb-rgsb-rgm-aa-ff-ca-zi-eq-nls-wn-lr100-fid2}{3.75}
\defdata{img512-cfgB-kl-rb-rmb-rgsb-rgm-aa-ff-ca-zi-eq-nls-wn-lr100-cfg2}{\TBD{TBD}}
\defdata{img512-cfgB-kl-rb-rmb-rgsb-rgm-aa-ff-ca-zi-eq-nls-wn-lr100-mparams0}{278}
\defdata{img512-cfgB-kl-rb-rmb-rgsb-rgm-aa-ff-ca-zi-eq-nls-wn-lr100-mparams1}{277.8}
\defdata{img512-cfgB-kl-rb-rmb-rgsb-rgm-aa-ff-ca-zi-eq-nls-wn-lr100-gflops0}{101}
\defdata{img512-cfgB-kl-rb-rmb-rgsb-rgm-aa-ff-ca-zi-eq-nls-wn-lr100-gflops1}{101.2}
\defdata{img512-cfgB-kl-rb-rmb-rgsb-rgm-aa-ff-ca-zi-eq-nls-wn-lr100-nfe}{63}
\defdata{img512-cfgB-kl-rb-rmb-rgsb-rgm-aa-ff-ca-zi-eq-nls-wn-lr100-mimg0}{2147}
\defdata{img512-cfgB-kl-rb-rmb-rgsb-rgm-aa-ff-ca-zi-eq-nls-wn-lr100-mimg1}{2147.5}
\defdata{img512-cfgB-kl-rb-rmb-rgsb-rgm-aa-ff-ca-zi-eq-nls-wn-lr100-spkimg3}{0.239}
\defdata{img512-cfgB-kl-rb-rmb-rgsb-rgm-aa-ff-ca-zi-eq-nls-wn-lr100-pkl}{1533/45/2147483.pkl}

\defdata{img512-cfgB-kl-rb-rmb-rgsb-rgm-aa-ff-ca-zi-eq-nls-wn-fwn-lrd-ref70kb-fid2}{3.02}
\defdata{img512-cfgB-kl-rb-rmb-rgsb-rgm-aa-ff-ca-zi-eq-nls-wn-fwn-lrd-ref70kb-cfg2}{\TBD{TBD}}
\defdata{img512-cfgB-kl-rb-rmb-rgsb-rgm-aa-ff-ca-zi-eq-nls-wn-fwn-lrd-ref70kb-mparams0}{278}
\defdata{img512-cfgB-kl-rb-rmb-rgsb-rgm-aa-ff-ca-zi-eq-nls-wn-fwn-lrd-ref70kb-mparams1}{277.8}
\defdata{img512-cfgB-kl-rb-rmb-rgsb-rgm-aa-ff-ca-zi-eq-nls-wn-fwn-lrd-ref70kb-gflops0}{101}
\defdata{img512-cfgB-kl-rb-rmb-rgsb-rgm-aa-ff-ca-zi-eq-nls-wn-fwn-lrd-ref70kb-gflops1}{101.2}
\defdata{img512-cfgB-kl-rb-rmb-rgsb-rgm-aa-ff-ca-zi-eq-nls-wn-fwn-lrd-ref70kb-nfe}{63}
\defdata{img512-cfgB-kl-rb-rmb-rgsb-rgm-aa-ff-ca-zi-eq-nls-wn-fwn-lrd-ref70kb-mimg0}{2147}
\defdata{img512-cfgB-kl-rb-rmb-rgsb-rgm-aa-ff-ca-zi-eq-nls-wn-fwn-lrd-ref70kb-mimg1}{2147.5}
\defdata{img512-cfgB-kl-rb-rmb-rgsb-rgm-aa-ff-ca-zi-eq-nls-wn-fwn-lrd-ref70kb-spkimg3}{0.243}
\defdata{img512-cfgB-kl-rb-rmb-rgsb-rgm-aa-ff-ca-zi-eq-nls-wn-fwn-lrd-ref70kb-pkl}{1533/46/2147483.pkl}

\defdata{img512-cfgB-kl-rb-rmb-rgsb-rgm-aa-ff-ca-zi-eq-nls-wn-fwn-lrd-ref70kb-rc-rr-res-pne-nnd-fid2}{2.71}
\defdata{img512-cfgB-kl-rb-rmb-rgsb-rgm-aa-ff-ca-zi-eq-nls-wn-fwn-lrd-ref70kb-rc-rr-res-pne-nnd-cfg2}{\TBD{TBD}}
\defdata{img512-cfgB-kl-rb-rmb-rgsb-rgm-aa-ff-ca-zi-eq-nls-wn-fwn-lrd-ref70kb-rc-rr-res-pne-nnd-mparams0}{280}
\defdata{img512-cfgB-kl-rb-rmb-rgsb-rgm-aa-ff-ca-zi-eq-nls-wn-fwn-lrd-ref70kb-rc-rr-res-pne-nnd-mparams1}{280.2}
\defdata{img512-cfgB-kl-rb-rmb-rgsb-rgm-aa-ff-ca-zi-eq-nls-wn-fwn-lrd-ref70kb-rc-rr-res-pne-nnd-gflops0}{102}
\defdata{img512-cfgB-kl-rb-rmb-rgsb-rgm-aa-ff-ca-zi-eq-nls-wn-fwn-lrd-ref70kb-rc-rr-res-pne-nnd-gflops1}{102.1}
\defdata{img512-cfgB-kl-rb-rmb-rgsb-rgm-aa-ff-ca-zi-eq-nls-wn-fwn-lrd-ref70kb-rc-rr-res-pne-nnd-nfe}{63}
\defdata{img512-cfgB-kl-rb-rmb-rgsb-rgm-aa-ff-ca-zi-eq-nls-wn-fwn-lrd-ref70kb-rc-rr-res-pne-nnd-mimg0}{2147}
\defdata{img512-cfgB-kl-rb-rmb-rgsb-rgm-aa-ff-ca-zi-eq-nls-wn-fwn-lrd-ref70kb-rc-rr-res-pne-nnd-mimg1}{2147.5}
\defdata{img512-cfgB-kl-rb-rmb-rgsb-rgm-aa-ff-ca-zi-eq-nls-wn-fwn-lrd-ref70kb-rc-rr-res-pne-nnd-spkimg3}{0.199}
\defdata{img512-cfgB-kl-rb-rmb-rgsb-rgm-aa-ff-ca-zi-eq-nls-wn-fwn-lrd-ref70kb-rc-rr-res-pne-nnd-pkl}{1533/72/2147483.pkl}

\defdata{img512-cfgB-kl-rb-rmb-rgsb-rgm-aa-ff-ca-zi-eq-nls-wn-fwn-lrd-ref70kb-rc-rr-res-pne-nnd-np-rb0.3-lag-log-bc-rd-fid2}{2.56}
\defdata{img512-cfgB-kl-rb-rmb-rgsb-rgm-aa-ff-ca-zi-eq-nls-wn-fwn-lrd-ref70kb-rc-rr-res-pne-nnd-np-rb0.3-lag-log-bc-rd-cfg2}{2.23}
\defdata{img512-cfgB-kl-rb-rmb-rgsb-rgm-aa-ff-ca-zi-eq-nls-wn-fwn-lrd-ref70kb-rc-rr-res-pne-nnd-np-rb0.3-lag-log-bc-rd-mparams0}{280}
\defdata{img512-cfgB-kl-rb-rmb-rgsb-rgm-aa-ff-ca-zi-eq-nls-wn-fwn-lrd-ref70kb-rc-rr-res-pne-nnd-np-rb0.3-lag-log-bc-rd-mparams1}{280.2}
\defdata{img512-cfgB-kl-rb-rmb-rgsb-rgm-aa-ff-ca-zi-eq-nls-wn-fwn-lrd-ref70kb-rc-rr-res-pne-nnd-np-rb0.3-lag-log-bc-rd-gflops0}{102}
\defdata{img512-cfgB-kl-rb-rmb-rgsb-rgm-aa-ff-ca-zi-eq-nls-wn-fwn-lrd-ref70kb-rc-rr-res-pne-nnd-np-rb0.3-lag-log-bc-rd-gflops1}{102.2}
\defdata{img512-cfgB-kl-rb-rmb-rgsb-rgm-aa-ff-ca-zi-eq-nls-wn-fwn-lrd-ref70kb-rc-rr-res-pne-nnd-np-rb0.3-lag-log-bc-rd-nfe}{63}
\defdata{img512-cfgB-kl-rb-rmb-rgsb-rgm-aa-ff-ca-zi-eq-nls-wn-fwn-lrd-ref70kb-rc-rr-res-pne-nnd-np-rb0.3-lag-log-bc-rd-mimg0}{2147}
\defdata{img512-cfgB-kl-rb-rmb-rgsb-rgm-aa-ff-ca-zi-eq-nls-wn-fwn-lrd-ref70kb-rc-rr-res-pne-nnd-np-rb0.3-lag-log-bc-rd-mimg1}{2147.5}
\defdata{img512-cfgB-kl-rb-rmb-rgsb-rgm-aa-ff-ca-zi-eq-nls-wn-fwn-lrd-ref70kb-rc-rr-res-pne-nnd-np-rb0.3-lag-log-bc-rd-spkimg3}{0.212}
\defdata{img512-cfgB-kl-rb-rmb-rgsb-rgm-aa-ff-ca-zi-eq-nls-wn-fwn-lrd-ref70kb-rc-rr-res-pne-nnd-np-rb0.3-lag-log-bc-rd-pkl}{1533/52/2147483.pkl}

\defdata{img512-b9-mc128-lr120-lim120-fid2}{3.53}
\defdata{img512-b9-mc128-lr120-lim120-cfg2}{2.91}
\defdata{img512-b9-mc128-lr120-lim120-mparams0}{125}
\defdata{img512-b9-mc128-lr120-lim120-mparams1}{124.7}
\defdata{img512-b9-mc128-lr120-lim120-gflops0}{46}
\defdata{img512-b9-mc128-lr120-lim120-gflops1}{45.5}
\defdata{img512-b9-mc128-lr120-lim120-nfe}{63}
\defdata{img512-b9-mc128-lr120-lim120-mimg0}{2147}
\defdata{img512-b9-mc128-lr120-lim120-mimg1}{2147.5}
\defdata{img512-b9-mc128-lr120-lim120-spkimg3}{0.121}
\defdata{img512-b9-mc128-lr120-lim120-pkl}{1538/23/2147483.pkl}

\defdata{img512-b9-mc256-lr90-lim90-fid2}{2.37}
\defdata{img512-b9-mc256-lr90-lim90-cfg2}{2.03}
\defdata{img512-b9-mc256-lr90-lim90-mparams0}{498}
\defdata{img512-b9-mc256-lr90-lim90-mparams1}{497.8}
\defdata{img512-b9-mc256-lr90-lim90-gflops0}{181}
\defdata{img512-b9-mc256-lr90-lim90-gflops1}{180.8}
\defdata{img512-b9-mc256-lr90-lim90-nfe}{63}
\defdata{img512-b9-mc256-lr90-lim90-mimg0}{1879}
\defdata{img512-b9-mc256-lr90-lim90-mimg1}{1879.0}
\defdata{img512-b9-mc256-lr90-lim90-spkimg3}{0.312}
\defdata{img512-b9-mc256-lr90-lim90-pkl}{1538/13/1879048.pkl}

\defdata{img512-b9-mc320-lr80-lim80-fid2}{2.23}
\defdata{img512-b9-mc320-lr80-lim80-cfg2}{1.95}
\defdata{img512-b9-mc320-lr80-lim80-mparams0}{777}
\defdata{img512-b9-mc320-lr80-lim80-mparams1}{777.5}
\defdata{img512-b9-mc320-lr80-lim80-gflops0}{282}
\defdata{img512-b9-mc320-lr80-lim80-gflops1}{282.2}
\defdata{img512-b9-mc320-lr80-lim80-nfe}{63}
\defdata{img512-b9-mc320-lr80-lim80-mimg0}{1074}
\defdata{img512-b9-mc320-lr80-lim80-mimg1}{1073.7}
\defdata{img512-b9-mc320-lr80-lim80-spkimg3}{0.469}
\defdata{img512-b9-mc320-lr80-lim80-pkl}{1538/6/1073741.pkl}

\defdata{img512-b9-mc384-lr70-lim70-fid2}{2.15}
\defdata{img512-b9-mc384-lr70-lim70-cfg2}{1.90}
\defdata{img512-b9-mc384-lr70-lim70-mparams0}{1119}
\defdata{img512-b9-mc384-lr70-lim70-mparams1}{1119.3}
\defdata{img512-b9-mc384-lr70-lim70-gflops0}{406}
\defdata{img512-b9-mc384-lr70-lim70-gflops1}{405.9}
\defdata{img512-b9-mc384-lr70-lim70-nfe}{63}
\defdata{img512-b9-mc384-lr70-lim70-mimg0}{805}
\defdata{img512-b9-mc384-lr70-lim70-mimg1}{805.3}
\defdata{img512-b9-mc384-lr70-lim70-spkimg3}{0.622}
\defdata{img512-b9-mc384-lr70-lim70-pkl}{1538/12/805306.pkl}

\defdata{img512-b9-mc448-lr65-lim65-fid2}{2.12}
\defdata{img512-b9-mc448-lr65-lim65-cfg2}{1.89}
\defdata{img512-b9-mc448-lr65-lim65-mparams0}{1523}
\defdata{img512-b9-mc448-lr65-lim65-mparams1}{1523.2}
\defdata{img512-b9-mc448-lr65-lim65-gflops0}{552}
\defdata{img512-b9-mc448-lr65-lim65-gflops1}{552.1}
\defdata{img512-b9-mc448-lr65-lim65-nfe}{63}
\defdata{img512-b9-mc448-lr65-lim65-mimg0}{537}
\defdata{img512-b9-mc448-lr65-lim65-mimg1}{536.9}
\defdata{img512-b9-mc448-lr65-lim65-spkimg3}{0.837}
\defdata{img512-b9-mc448-lr65-lim65-pkl}{1538/11/536870.pkl}

\defdata{img512-b9-mc128-lr120-lim120-s256M-drop0.10-fid2}{4.62}
\defdata{img512-b9-mc128-lr120-lim120-s256M-drop0.10-cfg2}{\TBD{TBD}}
\defdata{img512-b9-mc128-lr120-lim120-s256M-drop0.10-mparams0}{125}
\defdata{img512-b9-mc128-lr120-lim120-s256M-drop0.10-mparams1}{124.7}
\defdata{img512-b9-mc128-lr120-lim120-s256M-drop0.10-gflops0}{46}
\defdata{img512-b9-mc128-lr120-lim120-s256M-drop0.10-gflops1}{45.5}
\defdata{img512-b9-mc128-lr120-lim120-s256M-drop0.10-nfe}{63}
\defdata{img512-b9-mc128-lr120-lim120-s256M-drop0.10-mimg0}{2147}
\defdata{img512-b9-mc128-lr120-lim120-s256M-drop0.10-mimg1}{2147.5}
\defdata{img512-b9-mc128-lr120-lim120-s256M-drop0.10-spkimg3}{0.123}
\defdata{img512-b9-mc128-lr120-lim120-s256M-drop0.10-pkl}{1538/31/2147483.pkl}

\defdata{img512-cfgB-kl-rb-rmb-rgsb-rgm-aa-ff-ca-zi-eq-nls-wn-fwn-lrd-ref70kb-rc-rr-res-pne-nnd-np-rb0.3-lag-log-bc-fid2}{2.73}
\defdata{img512-cfgB-kl-rb-rmb-rgsb-rgm-aa-ff-ca-zi-eq-nls-wn-fwn-lrd-ref70kb-rc-rr-res-pne-nnd-np-rb0.3-lag-log-bc-cfg2}{\TBD{TBD}}
\defdata{img512-cfgB-kl-rb-rmb-rgsb-rgm-aa-ff-ca-zi-eq-nls-wn-fwn-lrd-ref70kb-rc-rr-res-pne-nnd-np-rb0.3-lag-log-bc-mparams0}{280}
\defdata{img512-cfgB-kl-rb-rmb-rgsb-rgm-aa-ff-ca-zi-eq-nls-wn-fwn-lrd-ref70kb-rc-rr-res-pne-nnd-np-rb0.3-lag-log-bc-mparams1}{280.2}
\defdata{img512-cfgB-kl-rb-rmb-rgsb-rgm-aa-ff-ca-zi-eq-nls-wn-fwn-lrd-ref70kb-rc-rr-res-pne-nnd-np-rb0.3-lag-log-bc-gflops0}{102}
\defdata{img512-cfgB-kl-rb-rmb-rgsb-rgm-aa-ff-ca-zi-eq-nls-wn-fwn-lrd-ref70kb-rc-rr-res-pne-nnd-np-rb0.3-lag-log-bc-gflops1}{102.2}
\defdata{img512-cfgB-kl-rb-rmb-rgsb-rgm-aa-ff-ca-zi-eq-nls-wn-fwn-lrd-ref70kb-rc-rr-res-pne-nnd-np-rb0.3-lag-log-bc-nfe}{63}
\defdata{img512-cfgB-kl-rb-rmb-rgsb-rgm-aa-ff-ca-zi-eq-nls-wn-fwn-lrd-ref70kb-rc-rr-res-pne-nnd-np-rb0.3-lag-log-bc-mimg0}{2147}
\defdata{img512-cfgB-kl-rb-rmb-rgsb-rgm-aa-ff-ca-zi-eq-nls-wn-fwn-lrd-ref70kb-rc-rr-res-pne-nnd-np-rb0.3-lag-log-bc-mimg1}{2147.5}
\defdata{img512-cfgB-kl-rb-rmb-rgsb-rgm-aa-ff-ca-zi-eq-nls-wn-fwn-lrd-ref70kb-rc-rr-res-pne-nnd-np-rb0.3-lag-log-bc-spkimg3}{0.215}
\defdata{img512-cfgB-kl-rb-rmb-rgsb-rgm-aa-ff-ca-zi-eq-nls-wn-fwn-lrd-ref70kb-rc-rr-res-pne-nnd-np-rb0.3-lag-log-bc-pkl}{1533/70/2147483.pkl}

\defdata{img512-b9-mc256-lr90-lim90-drop0.10-fid2}{2.25}
\defdata{img512-b9-mc256-lr90-lim90-drop0.10-cfg2}{2.01}
\defdata{img512-b9-mc256-lr90-lim90-drop0.10-mparams0}{498}
\defdata{img512-b9-mc256-lr90-lim90-drop0.10-mparams1}{497.8}
\defdata{img512-b9-mc256-lr90-lim90-drop0.10-gflops0}{181}
\defdata{img512-b9-mc256-lr90-lim90-drop0.10-gflops1}{180.8}
\defdata{img512-b9-mc256-lr90-lim90-drop0.10-nfe}{63}
\defdata{img512-b9-mc256-lr90-lim90-drop0.10-mimg0}{2147}
\defdata{img512-b9-mc256-lr90-lim90-drop0.10-mimg1}{2147.5}
\defdata{img512-b9-mc256-lr90-lim90-drop0.10-spkimg3}{0.312}
\defdata{img512-b9-mc256-lr90-lim90-drop0.10-pkl}{1538/29/2147483.pkl}

\defdata{img512-b9-mc320-lr80-lim80-drop0.10-fid2}{2.06}
\defdata{img512-b9-mc320-lr80-lim80-drop0.10-cfg2}{1.88}
\defdata{img512-b9-mc320-lr80-lim80-drop0.10-mparams0}{777}
\defdata{img512-b9-mc320-lr80-lim80-drop0.10-mparams1}{777.5}
\defdata{img512-b9-mc320-lr80-lim80-drop0.10-gflops0}{282}
\defdata{img512-b9-mc320-lr80-lim80-drop0.10-gflops1}{282.2}
\defdata{img512-b9-mc320-lr80-lim80-drop0.10-nfe}{63}
\defdata{img512-b9-mc320-lr80-lim80-drop0.10-mimg0}{1879}
\defdata{img512-b9-mc320-lr80-lim80-drop0.10-mimg1}{1879.0}
\defdata{img512-b9-mc320-lr80-lim80-drop0.10-spkimg3}{0.468}
\defdata{img512-b9-mc320-lr80-lim80-drop0.10-pkl}{1538/30/1879048.pkl}

\defdata{img512-b9-mc384-lr70-lim70-drop0.10-fid2}{1.96}
\defdata{img512-b9-mc384-lr70-lim70-drop0.10-cfg2}{1.85}
\defdata{img512-b9-mc384-lr70-lim70-drop0.10-mparams0}{1119}
\defdata{img512-b9-mc384-lr70-lim70-drop0.10-mparams1}{1119.3}
\defdata{img512-b9-mc384-lr70-lim70-drop0.10-gflops0}{406}
\defdata{img512-b9-mc384-lr70-lim70-drop0.10-gflops1}{405.9}
\defdata{img512-b9-mc384-lr70-lim70-drop0.10-nfe}{63}
\defdata{img512-b9-mc384-lr70-lim70-drop0.10-mimg0}{1342}
\defdata{img512-b9-mc384-lr70-lim70-drop0.10-mimg1}{1342.2}
\defdata{img512-b9-mc384-lr70-lim70-drop0.10-spkimg3}{0.626}
\defdata{img512-b9-mc384-lr70-lim70-drop0.10-pkl}{1538/28/1342177.pkl}

\defdata{img512-b9-mc448-lr65-lim65-drop0.10-fid2}{1.91}
\defdata{img512-b9-mc448-lr65-lim65-drop0.10-cfg2}{1.81}
\defdata{img512-b9-mc448-lr65-lim65-drop0.10-mparams0}{1523}
\defdata{img512-b9-mc448-lr65-lim65-drop0.10-mparams1}{1523.2}
\defdata{img512-b9-mc448-lr65-lim65-drop0.10-gflops0}{552}
\defdata{img512-b9-mc448-lr65-lim65-drop0.10-gflops1}{552.1}
\defdata{img512-b9-mc448-lr65-lim65-drop0.10-nfe}{63}
\defdata{img512-b9-mc448-lr65-lim65-drop0.10-mimg0}{940}
\defdata{img512-b9-mc448-lr65-lim65-drop0.10-mimg1}{939.5}
\defdata{img512-b9-mc448-lr65-lim65-drop0.10-spkimg3}{0.841}
\defdata{img512-b9-mc448-lr65-lim65-drop0.10-pkl}{1538/21/939524.pkl}

\defdata{img512-b9-ref30k-fid2}{2.82}
\defdata{img512-b9-ref30k-cfg2}{\TBD{TBD}}
\defdata{img512-b9-ref30k-mparams0}{280}
\defdata{img512-b9-ref30k-mparams1}{280.2}
\defdata{img512-b9-ref30k-gflops0}{102}
\defdata{img512-b9-ref30k-gflops1}{101.9}
\defdata{img512-b9-ref30k-nfe}{63}
\defdata{img512-b9-ref30k-mimg0}{2147}
\defdata{img512-b9-ref30k-mimg1}{2147.5}
\defdata{img512-b9-ref30k-spkimg3}{0.209}
\defdata{img512-b9-ref30k-pkl}{1536/8/2147483.pkl}

\defdata{img512-b9-ref40k-fid2}{2.69}
\defdata{img512-b9-ref40k-cfg2}{\TBD{TBD}}
\defdata{img512-b9-ref40k-mparams0}{280}
\defdata{img512-b9-ref40k-mparams1}{280.2}
\defdata{img512-b9-ref40k-gflops0}{102}
\defdata{img512-b9-ref40k-gflops1}{101.9}
\defdata{img512-b9-ref40k-nfe}{63}
\defdata{img512-b9-ref40k-mimg0}{2147}
\defdata{img512-b9-ref40k-mimg1}{2147.5}
\defdata{img512-b9-ref40k-spkimg3}{0.210}
\defdata{img512-b9-ref40k-pkl}{1536/7/2147483.pkl}

\defdata{img512-b9-ref50k-fid2}{2.61}
\defdata{img512-b9-ref50k-cfg2}{\TBD{TBD}}
\defdata{img512-b9-ref50k-mparams0}{280}
\defdata{img512-b9-ref50k-mparams1}{280.2}
\defdata{img512-b9-ref50k-gflops0}{102}
\defdata{img512-b9-ref50k-gflops1}{101.9}
\defdata{img512-b9-ref50k-nfe}{63}
\defdata{img512-b9-ref50k-mimg0}{2147}
\defdata{img512-b9-ref50k-mimg1}{2147.5}
\defdata{img512-b9-ref50k-spkimg3}{0.211}
\defdata{img512-b9-ref50k-pkl}{1536/5/2147483.pkl}

\defdata{img512-b9-ref100k-fid2}{2.68}
\defdata{img512-b9-ref100k-cfg2}{\TBD{TBD}}
\defdata{img512-b9-ref100k-mparams0}{280}
\defdata{img512-b9-ref100k-mparams1}{280.2}
\defdata{img512-b9-ref100k-gflops0}{102}
\defdata{img512-b9-ref100k-gflops1}{101.9}
\defdata{img512-b9-ref100k-nfe}{63}
\defdata{img512-b9-ref100k-mimg0}{2147}
\defdata{img512-b9-ref100k-mimg1}{2147.5}
\defdata{img512-b9-ref100k-spkimg3}{0.213}
\defdata{img512-b9-ref100k-pkl}{1536/3/2147483.pkl}

\defdata{img512-b9-ref120k-fid2}{2.68}
\defdata{img512-b9-ref120k-cfg2}{\TBD{TBD}}
\defdata{img512-b9-ref120k-mparams0}{280}
\defdata{img512-b9-ref120k-mparams1}{280.2}
\defdata{img512-b9-ref120k-gflops0}{102}
\defdata{img512-b9-ref120k-gflops1}{101.9}
\defdata{img512-b9-ref120k-nfe}{63}
\defdata{img512-b9-ref120k-mimg0}{2147}
\defdata{img512-b9-ref120k-mimg1}{2147.5}
\defdata{img512-b9-ref120k-spkimg3}{0.210}
\defdata{img512-b9-ref120k-pkl}{1536/6/2147483.pkl}

\defdata{img512-b9-ref160k-fid2}{2.75}
\defdata{img512-b9-ref160k-cfg2}{\TBD{TBD}}
\defdata{img512-b9-ref160k-mparams0}{280}
\defdata{img512-b9-ref160k-mparams1}{280.2}
\defdata{img512-b9-ref160k-gflops0}{102}
\defdata{img512-b9-ref160k-gflops1}{101.9}
\defdata{img512-b9-ref160k-nfe}{63}
\defdata{img512-b9-ref160k-mimg0}{2147}
\defdata{img512-b9-ref160k-mimg1}{2147.5}
\defdata{img512-b9-ref160k-spkimg3}{0.210}
\defdata{img512-b9-ref160k-pkl}{1536/9/2147483.pkl}

\defdata{img64-b9-mu-0.8-sigma1.6-ref35k-fid2}{1.58}
\defdata{img64-b9-mu-0.8-sigma1.6-ref35k-cfg2}{\TBD{TBD}}
\defdata{img64-b9-mu-0.8-sigma1.6-ref35k-mparams0}{280}
\defdata{img64-b9-mu-0.8-sigma1.6-ref35k-mparams1}{280.2}
\defdata{img64-b9-mu-0.8-sigma1.6-ref35k-gflops0}{102}
\defdata{img64-b9-mu-0.8-sigma1.6-ref35k-gflops1}{101.9}
\defdata{img64-b9-mu-0.8-sigma1.6-ref35k-nfe}{63}
\defdata{img64-b9-mu-0.8-sigma1.6-ref35k-mimg0}{1074}
\defdata{img64-b9-mu-0.8-sigma1.6-ref35k-mimg1}{1073.7}
\defdata{img64-b9-mu-0.8-sigma1.6-ref35k-spkimg3}{0.208}
\defdata{img64-b9-mu-0.8-sigma1.6-ref35k-pkl}{1539/12/1073741.pkl}

\defdata{img64-b9-mu-0.8-sigma1.6-ref35k-mc256-lr90-lim90-fid2}{1.46}
\defdata{img64-b9-mu-0.8-sigma1.6-ref35k-mc256-lr90-lim90-cfg2}{\TBD{TBD}}
\defdata{img64-b9-mu-0.8-sigma1.6-ref35k-mc256-lr90-lim90-mparams0}{498}
\defdata{img64-b9-mu-0.8-sigma1.6-ref35k-mc256-lr90-lim90-mparams1}{497.8}
\defdata{img64-b9-mu-0.8-sigma1.6-ref35k-mc256-lr90-lim90-gflops0}{181}
\defdata{img64-b9-mu-0.8-sigma1.6-ref35k-mc256-lr90-lim90-gflops1}{180.8}
\defdata{img64-b9-mu-0.8-sigma1.6-ref35k-mc256-lr90-lim90-nfe}{63}
\defdata{img64-b9-mu-0.8-sigma1.6-ref35k-mc256-lr90-lim90-mimg0}{805}
\defdata{img64-b9-mu-0.8-sigma1.6-ref35k-mc256-lr90-lim90-mimg1}{805.3}
\defdata{img64-b9-mu-0.8-sigma1.6-ref35k-mc256-lr90-lim90-spkimg3}{0.313}
\defdata{img64-b9-mu-0.8-sigma1.6-ref35k-mc256-lr90-lim90-pkl}{1539/8/805306.pkl}

\defdata{img64-b9-mu-0.8-sigma1.6-ref35k-mc320-lr80-lim80-fid2}{1.41}
\defdata{img64-b9-mu-0.8-sigma1.6-ref35k-mc320-lr80-lim80-cfg2}{\TBD{TBD}}
\defdata{img64-b9-mu-0.8-sigma1.6-ref35k-mc320-lr80-lim80-mparams0}{777}
\defdata{img64-b9-mu-0.8-sigma1.6-ref35k-mc320-lr80-lim80-mparams1}{777.5}
\defdata{img64-b9-mu-0.8-sigma1.6-ref35k-mc320-lr80-lim80-gflops0}{282}
\defdata{img64-b9-mu-0.8-sigma1.6-ref35k-mc320-lr80-lim80-gflops1}{282.1}
\defdata{img64-b9-mu-0.8-sigma1.6-ref35k-mc320-lr80-lim80-nfe}{63}
\defdata{img64-b9-mu-0.8-sigma1.6-ref35k-mc320-lr80-lim80-mimg0}{537}
\defdata{img64-b9-mu-0.8-sigma1.6-ref35k-mc320-lr80-lim80-mimg1}{536.9}
\defdata{img64-b9-mu-0.8-sigma1.6-ref35k-mc320-lr80-lim80-spkimg3}{0.456}
\defdata{img64-b9-mu-0.8-sigma1.6-ref35k-mc320-lr80-lim80-pkl}{1539/10/536870.pkl}

\defdata{img64-b9-mu-0.8-sigma1.6-ref35k-mc384-lr70-lim70-fid2}{1.37}
\defdata{img64-b9-mu-0.8-sigma1.6-ref35k-mc384-lr70-lim70-cfg2}{\TBD{TBD}}
\defdata{img64-b9-mu-0.8-sigma1.6-ref35k-mc384-lr70-lim70-mparams0}{1119}
\defdata{img64-b9-mu-0.8-sigma1.6-ref35k-mc384-lr70-lim70-mparams1}{1119.3}
\defdata{img64-b9-mu-0.8-sigma1.6-ref35k-mc384-lr70-lim70-gflops0}{406}
\defdata{img64-b9-mu-0.8-sigma1.6-ref35k-mc384-lr70-lim70-gflops1}{405.9}
\defdata{img64-b9-mu-0.8-sigma1.6-ref35k-mc384-lr70-lim70-nfe}{63}
\defdata{img64-b9-mu-0.8-sigma1.6-ref35k-mc384-lr70-lim70-mimg0}{403}
\defdata{img64-b9-mu-0.8-sigma1.6-ref35k-mc384-lr70-lim70-mimg1}{402.7}
\defdata{img64-b9-mu-0.8-sigma1.6-ref35k-mc384-lr70-lim70-spkimg3}{0.603}
\defdata{img64-b9-mu-0.8-sigma1.6-ref35k-mc384-lr70-lim70-pkl}{1539/13/402653.pkl}

\defdata{img64-b9-mu-0.8-sigma1.6-ref35k-drop0.10-fid2}{1.60}
\defdata{img64-b9-mu-0.8-sigma1.6-ref35k-drop0.10-cfg2}{\TBD{TBD}}
\defdata{img64-b9-mu-0.8-sigma1.6-ref35k-drop0.10-mparams0}{280}
\defdata{img64-b9-mu-0.8-sigma1.6-ref35k-drop0.10-mparams1}{280.2}
\defdata{img64-b9-mu-0.8-sigma1.6-ref35k-drop0.10-gflops0}{102}
\defdata{img64-b9-mu-0.8-sigma1.6-ref35k-drop0.10-gflops1}{101.9}
\defdata{img64-b9-mu-0.8-sigma1.6-ref35k-drop0.10-nfe}{63}
\defdata{img64-b9-mu-0.8-sigma1.6-ref35k-drop0.10-mimg0}{2147}
\defdata{img64-b9-mu-0.8-sigma1.6-ref35k-drop0.10-mimg1}{2147.5}
\defdata{img64-b9-mu-0.8-sigma1.6-ref35k-drop0.10-spkimg3}{0.212}
\defdata{img64-b9-mu-0.8-sigma1.6-ref35k-drop0.10-pkl}{1539/14/2147483.pkl}

\defdata{img64-b9-mu-0.8-sigma1.6-ref35k-mc256-lr90-lim90-drop0.10-fid2}{1.43}
\defdata{img64-b9-mu-0.8-sigma1.6-ref35k-mc256-lr90-lim90-drop0.10-cfg2}{\TBD{TBD}}
\defdata{img64-b9-mu-0.8-sigma1.6-ref35k-mc256-lr90-lim90-drop0.10-mparams0}{498}
\defdata{img64-b9-mu-0.8-sigma1.6-ref35k-mc256-lr90-lim90-drop0.10-mparams1}{497.8}
\defdata{img64-b9-mu-0.8-sigma1.6-ref35k-mc256-lr90-lim90-drop0.10-gflops0}{181}
\defdata{img64-b9-mu-0.8-sigma1.6-ref35k-mc256-lr90-lim90-drop0.10-gflops1}{180.8}
\defdata{img64-b9-mu-0.8-sigma1.6-ref35k-mc256-lr90-lim90-drop0.10-nfe}{63}
\defdata{img64-b9-mu-0.8-sigma1.6-ref35k-mc256-lr90-lim90-drop0.10-mimg0}{2147}
\defdata{img64-b9-mu-0.8-sigma1.6-ref35k-mc256-lr90-lim90-drop0.10-mimg1}{2147.5}
\defdata{img64-b9-mu-0.8-sigma1.6-ref35k-mc256-lr90-lim90-drop0.10-spkimg3}{0.314}
\defdata{img64-b9-mu-0.8-sigma1.6-ref35k-mc256-lr90-lim90-drop0.10-pkl}{1539/19/2147483.pkl}

\defdata{img64-b9-mu-0.8-sigma1.6-ref35k-mc320-lr80-lim80-drop0.10-fid2}{1.33}
\defdata{img64-b9-mu-0.8-sigma1.6-ref35k-mc320-lr80-lim80-drop0.10-cfg2}{\TBD{TBD}}
\defdata{img64-b9-mu-0.8-sigma1.6-ref35k-mc320-lr80-lim80-drop0.10-mparams0}{777}
\defdata{img64-b9-mu-0.8-sigma1.6-ref35k-mc320-lr80-lim80-drop0.10-mparams1}{777.5}
\defdata{img64-b9-mu-0.8-sigma1.6-ref35k-mc320-lr80-lim80-drop0.10-gflops0}{282}
\defdata{img64-b9-mu-0.8-sigma1.6-ref35k-mc320-lr80-lim80-drop0.10-gflops1}{282.1}
\defdata{img64-b9-mu-0.8-sigma1.6-ref35k-mc320-lr80-lim80-drop0.10-nfe}{63}
\defdata{img64-b9-mu-0.8-sigma1.6-ref35k-mc320-lr80-lim80-drop0.10-mimg0}{1074}
\defdata{img64-b9-mu-0.8-sigma1.6-ref35k-mc320-lr80-lim80-drop0.10-mimg1}{1073.7}
\defdata{img64-b9-mu-0.8-sigma1.6-ref35k-mc320-lr80-lim80-drop0.10-spkimg3}{0.464}
\defdata{img64-b9-mu-0.8-sigma1.6-ref35k-mc320-lr80-lim80-drop0.10-pkl}{1539/15/1073741.pkl}

\defdata{img64-b9-mu-0.8-sigma1.6-ref35k-mc384-lr70-lim70-drop0.10-fid2}{1.33}
\defdata{img64-b9-mu-0.8-sigma1.6-ref35k-mc384-lr70-lim70-drop0.10-cfg2}{\TBD{TBD}}
\defdata{img64-b9-mu-0.8-sigma1.6-ref35k-mc384-lr70-lim70-drop0.10-mparams0}{1119}
\defdata{img64-b9-mu-0.8-sigma1.6-ref35k-mc384-lr70-lim70-drop0.10-mparams1}{1119.3}
\defdata{img64-b9-mu-0.8-sigma1.6-ref35k-mc384-lr70-lim70-drop0.10-gflops0}{406}
\defdata{img64-b9-mu-0.8-sigma1.6-ref35k-mc384-lr70-lim70-drop0.10-gflops1}{405.9}
\defdata{img64-b9-mu-0.8-sigma1.6-ref35k-mc384-lr70-lim70-drop0.10-nfe}{63}
\defdata{img64-b9-mu-0.8-sigma1.6-ref35k-mc384-lr70-lim70-drop0.10-mimg0}{671}
\defdata{img64-b9-mu-0.8-sigma1.6-ref35k-mc384-lr70-lim70-drop0.10-mimg1}{671.1}
\defdata{img64-b9-mu-0.8-sigma1.6-ref35k-mc384-lr70-lim70-drop0.10-spkimg3}{0.626}
\defdata{img64-b9-mu-0.8-sigma1.6-ref35k-mc384-lr70-lim70-drop0.10-pkl}{1539/18/671088.pkl}

\input{data/data-phema}
\defdata{emaprofile-exp0}{
(0.000, 0.0000)
(0.001, 0.0069)
(0.002, 0.0114)
(0.003, 0.0153)
(0.004, 0.0188)
(0.005, 0.0221)
(0.006, 0.0252)
(0.007, 0.0281)
(0.008, 0.0309)
(0.009, 0.0337)
(0.010, 0.0363)
(0.011, 0.0389)
(0.012, 0.0414)
(0.013, 0.0439)
(0.014, 0.0463)
(0.015, 0.0487)
(0.016, 0.0510)
(0.017, 0.0532)
(0.018, 0.0555)
(0.019, 0.0577)
(0.020, 0.0598)
(0.021, 0.0620)
(0.022, 0.0641)
(0.023, 0.0662)
(0.024, 0.0682)
(0.025, 0.0703)
(0.026, 0.0723)
(0.027, 0.0743)
(0.028, 0.0762)
(0.029, 0.0782)
(0.030, 0.0801)
(0.031, 0.0820)
(0.032, 0.0839)
(0.033, 0.0858)
(0.034, 0.0877)
(0.035, 0.0895)
(0.036, 0.0914)
(0.037, 0.0932)
(0.038, 0.0950)
(0.039, 0.0968)
(0.040, 0.0986)
(0.041, 0.1003)
(0.042, 0.1021)
(0.043, 0.1038)
(0.044, 0.1056)
(0.045, 0.1073)
(0.046, 0.1090)
(0.047, 0.1107)
(0.048, 0.1124)
(0.049, 0.1141)
(0.050, 0.1157)
(0.051, 0.1174)
(0.052, 0.1191)
(0.053, 0.1207)
(0.054, 0.1223)
(0.055, 0.1240)
(0.056, 0.1256)
(0.057, 0.1272)
(0.058, 0.1288)
(0.059, 0.1304)
(0.060, 0.1320)
(0.061, 0.1336)
(0.062, 0.1351)
(0.063, 0.1367)
(0.064, 0.1382)
(0.065, 0.1398)
(0.066, 0.1413)
(0.067, 0.1429)
(0.068, 0.1444)
(0.069, 0.1459)
(0.070, 0.1475)
(0.071, 0.1490)
(0.072, 0.1505)
(0.073, 0.1520)
(0.074, 0.1535)
(0.075, 0.1550)
(0.076, 0.1565)
(0.077, 0.1579)
(0.078, 0.1594)
(0.079, 0.1609)
(0.080, 0.1623)
(0.081, 0.1638)
(0.082, 0.1653)
(0.083, 0.1667)
(0.084, 0.1681)
(0.085, 0.1696)
(0.086, 0.1710)
(0.087, 0.1724)
(0.088, 0.1739)
(0.089, 0.1753)
(0.090, 0.1767)
(0.091, 0.1781)
(0.092, 0.1795)
(0.093, 0.1809)
(0.094, 0.1823)
(0.095, 0.1837)
(0.096, 0.1851)
(0.097, 0.1865)
(0.098, 0.1879)
(0.099, 0.1893)
(0.100, 0.1906)
(0.101, 0.1920)
(0.102, 0.1934)
(0.103, 0.1947)
(0.104, 0.1961)
(0.105, 0.1974)
(0.106, 0.1988)
(0.107, 0.2001)
(0.108, 0.2015)
(0.109, 0.2028)
(0.110, 0.2042)
(0.111, 0.2055)
(0.112, 0.2068)
(0.113, 0.2082)
(0.114, 0.2095)
(0.115, 0.2108)
(0.116, 0.2121)
(0.117, 0.2134)
(0.118, 0.2147)
(0.119, 0.2161)
(0.120, 0.2174)
(0.121, 0.2187)
(0.122, 0.2200)
(0.123, 0.2213)
(0.124, 0.2225)
(0.125, 0.2238)
(0.126, 0.2251)
(0.127, 0.2264)
(0.128, 0.2277)
(0.129, 0.2290)
(0.130, 0.2302)
(0.131, 0.2315)
(0.132, 0.2328)
(0.133, 0.2341)
(0.134, 0.2353)
(0.135, 0.2366)
(0.136, 0.2379)
(0.137, 0.2391)
(0.138, 0.2404)
(0.139, 0.2416)
(0.140, 0.2429)
(0.141, 0.2441)
(0.142, 0.2454)
(0.143, 0.2466)
(0.144, 0.2478)
(0.145, 0.2491)
(0.146, 0.2503)
(0.147, 0.2515)
(0.148, 0.2528)
(0.149, 0.2540)
(0.150, 0.2552)
(0.151, 0.2565)
(0.152, 0.2577)
(0.153, 0.2589)
(0.154, 0.2601)
(0.155, 0.2613)
(0.156, 0.2625)
(0.157, 0.2638)
(0.158, 0.2650)
(0.159, 0.2662)
(0.160, 0.2674)
(0.161, 0.2686)
(0.162, 0.2698)
(0.163, 0.2710)
(0.164, 0.2722)
(0.165, 0.2734)
(0.166, 0.2745)
(0.167, 0.2757)
(0.168, 0.2769)
(0.169, 0.2781)
(0.170, 0.2793)
(0.171, 0.2805)
(0.172, 0.2817)
(0.173, 0.2828)
(0.174, 0.2840)
(0.175, 0.2852)
(0.176, 0.2864)
(0.177, 0.2875)
(0.178, 0.2887)
(0.179, 0.2899)
(0.180, 0.2910)
(0.181, 0.2922)
(0.182, 0.2933)
(0.183, 0.2945)
(0.184, 0.2957)
(0.185, 0.2968)
(0.186, 0.2980)
(0.187, 0.2991)
(0.188, 0.3003)
(0.189, 0.3014)
(0.190, 0.3026)
(0.191, 0.3037)
(0.192, 0.3049)
(0.193, 0.3060)
(0.194, 0.3071)
(0.195, 0.3083)
(0.196, 0.3094)
(0.197, 0.3106)
(0.198, 0.3117)
(0.199, 0.3128)
(0.200, 0.3140)
(0.201, 0.3151)
(0.202, 0.3162)
(0.203, 0.3173)
(0.204, 0.3185)
(0.205, 0.3196)
(0.206, 0.3207)
(0.207, 0.3218)
(0.208, 0.3229)
(0.209, 0.3241)
(0.210, 0.3252)
(0.211, 0.3263)
(0.212, 0.3274)
(0.213, 0.3285)
(0.214, 0.3296)
(0.215, 0.3307)
(0.216, 0.3318)
(0.217, 0.3329)
(0.218, 0.3340)
(0.219, 0.3351)
(0.220, 0.3363)
(0.221, 0.3373)
(0.222, 0.3384)
(0.223, 0.3395)
(0.224, 0.3406)
(0.225, 0.3417)
(0.226, 0.3428)
(0.227, 0.3439)
(0.228, 0.3450)
(0.229, 0.3461)
(0.230, 0.3472)
(0.231, 0.3483)
(0.232, 0.3494)
(0.233, 0.3504)
(0.234, 0.3515)
(0.235, 0.3526)
(0.236, 0.3537)
(0.237, 0.3548)
(0.238, 0.3558)
(0.239, 0.3569)
(0.240, 0.3580)
(0.241, 0.3591)
(0.242, 0.3601)
(0.243, 0.3612)
(0.244, 0.3623)
(0.245, 0.3633)
(0.246, 0.3644)
(0.247, 0.3655)
(0.248, 0.3665)
(0.249, 0.3676)
(0.250, 0.3687)
(0.251, 0.3697)
(0.252, 0.3708)
(0.253, 0.3718)
(0.254, 0.3729)
(0.255, 0.3740)
(0.256, 0.3750)
(0.257, 0.3761)
(0.258, 0.3771)
(0.259, 0.3782)
(0.260, 0.3792)
(0.261, 0.3803)
(0.262, 0.3813)
(0.263, 0.3824)
(0.264, 0.3834)
(0.265, 0.3845)
(0.266, 0.3855)
(0.267, 0.3865)
(0.268, 0.3876)
(0.269, 0.3886)
(0.270, 0.3897)
(0.271, 0.3907)
(0.272, 0.3917)
(0.273, 0.3928)
(0.274, 0.3938)
(0.275, 0.3948)
(0.276, 0.3959)
(0.277, 0.3969)
(0.278, 0.3979)
(0.279, 0.3990)
(0.280, 0.4000)
(0.281, 0.4010)
(0.282, 0.4020)
(0.283, 0.4031)
(0.284, 0.4041)
(0.285, 0.4051)
(0.286, 0.4061)
(0.287, 0.4072)
(0.288, 0.4082)
(0.289, 0.4092)
(0.290, 0.4102)
(0.291, 0.4112)
(0.292, 0.4123)
(0.293, 0.4133)
(0.294, 0.4143)
(0.295, 0.4153)
(0.296, 0.4163)
(0.297, 0.4173)
(0.298, 0.4183)
(0.299, 0.4194)
(0.300, 0.4204)
(0.301, 0.4214)
(0.302, 0.4224)
(0.303, 0.4234)
(0.304, 0.4244)
(0.305, 0.4254)
(0.306, 0.4264)
(0.307, 0.4274)
(0.308, 0.4284)
(0.309, 0.4294)
(0.310, 0.4304)
(0.311, 0.4314)
(0.312, 0.4324)
(0.313, 0.4334)
(0.314, 0.4344)
(0.315, 0.4354)
(0.316, 0.4364)
(0.317, 0.4374)
(0.318, 0.4384)
(0.319, 0.4394)
(0.320, 0.4403)
(0.321, 0.4413)
(0.322, 0.4423)
(0.323, 0.4433)
(0.324, 0.4443)
(0.325, 0.4453)
(0.326, 0.4463)
(0.327, 0.4473)
(0.328, 0.4482)
(0.329, 0.4492)
(0.330, 0.4502)
(0.331, 0.4512)
(0.332, 0.4522)
(0.333, 0.4532)
(0.334, 0.4541)
(0.335, 0.4551)
(0.336, 0.4561)
(0.337, 0.4571)
(0.338, 0.4580)
(0.339, 0.4590)
(0.340, 0.4600)
(0.341, 0.4610)
(0.342, 0.4619)
(0.343, 0.4629)
(0.344, 0.4639)
(0.345, 0.4649)
(0.346, 0.4658)
(0.347, 0.4668)
(0.348, 0.4678)
(0.349, 0.4687)
(0.350, 0.4697)
(0.351, 0.4707)
(0.352, 0.4716)
(0.353, 0.4726)
(0.354, 0.4735)
(0.355, 0.4745)
(0.356, 0.4755)
(0.357, 0.4764)
(0.358, 0.4774)
(0.359, 0.4784)
(0.360, 0.4793)
(0.361, 0.4803)
(0.362, 0.4812)
(0.363, 0.4822)
(0.364, 0.4831)
(0.365, 0.4841)
(0.366, 0.4850)
(0.367, 0.4860)
(0.368, 0.4870)
(0.369, 0.4879)
(0.370, 0.4889)
(0.371, 0.4898)
(0.372, 0.4908)
(0.373, 0.4917)
(0.374, 0.4927)
(0.375, 0.4936)
(0.376, 0.4946)
(0.377, 0.4955)
(0.378, 0.4964)
(0.379, 0.4974)
(0.380, 0.4983)
(0.381, 0.4993)
(0.382, 0.5002)
(0.383, 0.5012)
(0.384, 0.5021)
(0.385, 0.5030)
(0.386, 0.5040)
(0.387, 0.5049)
(0.388, 0.5059)
(0.389, 0.5068)
(0.390, 0.5077)
(0.391, 0.5087)
(0.392, 0.5096)
(0.393, 0.5105)
(0.394, 0.5115)
(0.395, 0.5124)
(0.396, 0.5133)
(0.397, 0.5143)
(0.398, 0.5152)
(0.399, 0.5161)
(0.400, 0.5171)
(0.401, 0.5180)
(0.402, 0.5189)
(0.403, 0.5199)
(0.404, 0.5208)
(0.405, 0.5217)
(0.406, 0.5226)
(0.407, 0.5236)
(0.408, 0.5245)
(0.409, 0.5254)
(0.410, 0.5263)
(0.411, 0.5273)
(0.412, 0.5282)
(0.413, 0.5291)
(0.414, 0.5300)
(0.415, 0.5310)
(0.416, 0.5319)
(0.417, 0.5328)
(0.418, 0.5337)
(0.419, 0.5346)
(0.420, 0.5356)
(0.421, 0.5365)
(0.422, 0.5374)
(0.423, 0.5383)
(0.424, 0.5392)
(0.425, 0.5401)
(0.426, 0.5411)
(0.427, 0.5420)
(0.428, 0.5429)
(0.429, 0.5438)
(0.430, 0.5447)
(0.431, 0.5456)
(0.432, 0.5465)
(0.433, 0.5474)
(0.434, 0.5484)
(0.435, 0.5493)
(0.436, 0.5502)
(0.437, 0.5511)
(0.438, 0.5520)
(0.439, 0.5529)
(0.440, 0.5538)
(0.441, 0.5547)
(0.442, 0.5556)
(0.443, 0.5565)
(0.444, 0.5574)
(0.445, 0.5583)
(0.446, 0.5592)
(0.447, 0.5601)
(0.448, 0.5610)
(0.449, 0.5619)
(0.450, 0.5628)
(0.451, 0.5637)
(0.452, 0.5646)
(0.453, 0.5655)
(0.454, 0.5664)
(0.455, 0.5673)
(0.456, 0.5682)
(0.457, 0.5691)
(0.458, 0.5700)
(0.459, 0.5709)
(0.460, 0.5718)
(0.461, 0.5727)
(0.462, 0.5736)
(0.463, 0.5745)
(0.464, 0.5754)
(0.465, 0.5763)
(0.466, 0.5772)
(0.467, 0.5781)
(0.468, 0.5789)
(0.469, 0.5798)
(0.470, 0.5807)
(0.471, 0.5816)
(0.472, 0.5825)
(0.473, 0.5834)
(0.474, 0.5843)
(0.475, 0.5852)
(0.476, 0.5861)
(0.477, 0.5869)
(0.478, 0.5878)
(0.479, 0.5887)
(0.480, 0.5896)
(0.481, 0.5905)
(0.482, 0.5914)
(0.483, 0.5922)
(0.484, 0.5931)
(0.485, 0.5940)
(0.486, 0.5949)
(0.487, 0.5958)
(0.488, 0.5966)
(0.489, 0.5975)
(0.490, 0.5984)
(0.491, 0.5993)
(0.492, 0.6002)
(0.493, 0.6010)
(0.494, 0.6019)
(0.495, 0.6028)
(0.496, 0.6037)
(0.497, 0.6045)
(0.498, 0.6054)
(0.499, 0.6063)
(0.500, 0.6072)
(0.501, 0.6080)
(0.502, 0.6089)
(0.503, 0.6098)
(0.504, 0.6107)
(0.505, 0.6115)
(0.506, 0.6124)
(0.507, 0.6133)
(0.508, 0.6142)
(0.509, 0.6150)
(0.510, 0.6159)
(0.511, 0.6168)
(0.512, 0.6176)
(0.513, 0.6185)
(0.514, 0.6194)
(0.515, 0.6202)
(0.516, 0.6211)
(0.517, 0.6220)
(0.518, 0.6228)
(0.519, 0.6237)
(0.520, 0.6246)
(0.521, 0.6254)
(0.522, 0.6263)
(0.523, 0.6272)
(0.524, 0.6280)
(0.525, 0.6289)
(0.526, 0.6297)
(0.527, 0.6306)
(0.528, 0.6315)
(0.529, 0.6323)
(0.530, 0.6332)
(0.531, 0.6340)
(0.532, 0.6349)
(0.533, 0.6358)
(0.534, 0.6366)
(0.535, 0.6375)
(0.536, 0.6383)
(0.537, 0.6392)
(0.538, 0.6400)
(0.539, 0.6409)
(0.540, 0.6418)
(0.541, 0.6426)
(0.542, 0.6435)
(0.543, 0.6443)
(0.544, 0.6452)
(0.545, 0.6460)
(0.546, 0.6469)
(0.547, 0.6477)
(0.548, 0.6486)
(0.549, 0.6494)
(0.550, 0.6503)
(0.551, 0.6511)
(0.552, 0.6520)
(0.553, 0.6528)
(0.554, 0.6537)
(0.555, 0.6545)
(0.556, 0.6554)
(0.557, 0.6562)
(0.558, 0.6571)
(0.559, 0.6579)
(0.560, 0.6588)
(0.561, 0.6596)
(0.562, 0.6605)
(0.563, 0.6613)
(0.564, 0.6622)
(0.565, 0.6630)
(0.566, 0.6639)
(0.567, 0.6647)
(0.568, 0.6655)
(0.569, 0.6664)
(0.570, 0.6672)
(0.571, 0.6681)
(0.572, 0.6689)
(0.573, 0.6698)
(0.574, 0.6706)
(0.575, 0.6714)
(0.576, 0.6723)
(0.577, 0.6731)
(0.578, 0.6740)
(0.579, 0.6748)
(0.580, 0.6756)
(0.581, 0.6765)
(0.582, 0.6773)
(0.583, 0.6781)
(0.584, 0.6790)
(0.585, 0.6798)
(0.586, 0.6807)
(0.587, 0.6815)
(0.588, 0.6823)
(0.589, 0.6832)
(0.590, 0.6840)
(0.591, 0.6848)
(0.592, 0.6857)
(0.593, 0.6865)
(0.594, 0.6873)
(0.595, 0.6882)
(0.596, 0.6890)
(0.597, 0.6898)
(0.598, 0.6907)
(0.599, 0.6915)
(0.600, 0.6923)
(0.601, 0.6932)
(0.602, 0.6940)
(0.603, 0.6948)
(0.604, 0.6956)
(0.605, 0.6965)
(0.606, 0.6973)
(0.607, 0.6981)
(0.608, 0.6990)
(0.609, 0.6998)
(0.610, 0.7006)
(0.611, 0.7014)
(0.612, 0.7023)
(0.613, 0.7031)
(0.614, 0.7039)
(0.615, 0.7047)
(0.616, 0.7056)
(0.617, 0.7064)
(0.618, 0.7072)
(0.619, 0.7080)
(0.620, 0.7089)
(0.621, 0.7097)
(0.622, 0.7105)
(0.623, 0.7113)
(0.624, 0.7121)
(0.625, 0.7130)
(0.626, 0.7138)
(0.627, 0.7146)
(0.628, 0.7154)
(0.629, 0.7163)
(0.630, 0.7171)
(0.631, 0.7179)
(0.632, 0.7187)
(0.633, 0.7195)
(0.634, 0.7203)
(0.635, 0.7212)
(0.636, 0.7220)
(0.637, 0.7228)
(0.638, 0.7236)
(0.639, 0.7244)
(0.640, 0.7252)
(0.641, 0.7261)
(0.642, 0.7269)
(0.643, 0.7277)
(0.644, 0.7285)
(0.645, 0.7293)
(0.646, 0.7301)
(0.647, 0.7309)
(0.648, 0.7318)
(0.649, 0.7326)
(0.650, 0.7334)
(0.651, 0.7342)
(0.652, 0.7350)
(0.653, 0.7358)
(0.654, 0.7366)
(0.655, 0.7374)
(0.656, 0.7383)
(0.657, 0.7391)
(0.658, 0.7399)
(0.659, 0.7407)
(0.660, 0.7415)
(0.661, 0.7423)
(0.662, 0.7431)
(0.663, 0.7439)
(0.664, 0.7447)
(0.665, 0.7455)
(0.666, 0.7463)
(0.667, 0.7471)
(0.668, 0.7479)
(0.669, 0.7488)
(0.670, 0.7496)
(0.671, 0.7504)
(0.672, 0.7512)
(0.673, 0.7520)
(0.674, 0.7528)
(0.675, 0.7536)
(0.676, 0.7544)
(0.677, 0.7552)
(0.678, 0.7560)
(0.679, 0.7568)
(0.680, 0.7576)
(0.681, 0.7584)
(0.682, 0.7592)
(0.683, 0.7600)
(0.684, 0.7608)
(0.685, 0.7616)
(0.686, 0.7624)
(0.687, 0.7632)
(0.688, 0.7640)
(0.689, 0.7648)
(0.690, 0.7656)
(0.691, 0.7664)
(0.692, 0.7672)
(0.693, 0.7680)
(0.694, 0.7688)
(0.695, 0.7696)
(0.696, 0.7704)
(0.697, 0.7712)
(0.698, 0.7720)
(0.699, 0.7728)
(0.700, 0.7736)
(0.701, 0.7744)
(0.702, 0.7752)
(0.703, 0.7760)
(0.704, 0.7767)
(0.705, 0.7775)
(0.706, 0.7783)
(0.707, 0.7791)
(0.708, 0.7799)
(0.709, 0.7807)
(0.710, 0.7815)
(0.711, 0.7823)
(0.712, 0.7831)
(0.713, 0.7839)
(0.714, 0.7847)
(0.715, 0.7855)
(0.716, 0.7863)
(0.717, 0.7870)
(0.718, 0.7878)
(0.719, 0.7886)
(0.720, 0.7894)
(0.721, 0.7902)
(0.722, 0.7910)
(0.723, 0.7918)
(0.724, 0.7926)
(0.725, 0.7934)
(0.726, 0.7941)
(0.727, 0.7949)
(0.728, 0.7957)
(0.729, 0.7965)
(0.730, 0.7973)
(0.731, 0.7981)
(0.732, 0.7989)
(0.733, 0.7996)
(0.734, 0.8004)
(0.735, 0.8012)
(0.736, 0.8020)
(0.737, 0.8028)
(0.738, 0.8036)
(0.739, 0.8044)
(0.740, 0.8051)
(0.741, 0.8059)
(0.742, 0.8067)
(0.743, 0.8075)
(0.744, 0.8083)
(0.745, 0.8091)
(0.746, 0.8098)
(0.747, 0.8106)
(0.748, 0.8114)
(0.749, 0.8122)
(0.750, 0.8130)
(0.751, 0.8137)
(0.752, 0.8145)
(0.753, 0.8153)
(0.754, 0.8161)
(0.755, 0.8169)
(0.756, 0.8176)
(0.757, 0.8184)
(0.758, 0.8192)
(0.759, 0.8200)
(0.760, 0.8207)
(0.761, 0.8215)
(0.762, 0.8223)
(0.763, 0.8231)
(0.764, 0.8239)
(0.765, 0.8246)
(0.766, 0.8254)
(0.767, 0.8262)
(0.768, 0.8270)
(0.769, 0.8277)
(0.770, 0.8285)
(0.771, 0.8293)
(0.772, 0.8301)
(0.773, 0.8308)
(0.774, 0.8316)
(0.775, 0.8324)
(0.776, 0.8331)
(0.777, 0.8339)
(0.778, 0.8347)
(0.779, 0.8355)
(0.780, 0.8362)
(0.781, 0.8370)
(0.782, 0.8378)
(0.783, 0.8385)
(0.784, 0.8393)
(0.785, 0.8401)
(0.786, 0.8409)
(0.787, 0.8416)
(0.788, 0.8424)
(0.789, 0.8432)
(0.790, 0.8439)
(0.791, 0.8447)
(0.792, 0.8455)
(0.793, 0.8462)
(0.794, 0.8470)
(0.795, 0.8478)
(0.796, 0.8485)
(0.797, 0.8493)
(0.798, 0.8501)
(0.799, 0.8508)
(0.800, 0.8516)
(0.801, 0.8524)
(0.802, 0.8531)
(0.803, 0.8539)
(0.804, 0.8547)
(0.805, 0.8554)
(0.806, 0.8562)
(0.807, 0.8570)
(0.808, 0.8577)
(0.809, 0.8585)
(0.810, 0.8593)
(0.811, 0.8600)
(0.812, 0.8608)
(0.813, 0.8616)
(0.814, 0.8623)
(0.815, 0.8631)
(0.816, 0.8638)
(0.817, 0.8646)
(0.818, 0.8654)
(0.819, 0.8661)
(0.820, 0.8669)
(0.821, 0.8676)
(0.822, 0.8684)
(0.823, 0.8692)
(0.824, 0.8699)
(0.825, 0.8707)
(0.826, 0.8714)
(0.827, 0.8722)
(0.828, 0.8730)
(0.829, 0.8737)
(0.830, 0.8745)
(0.831, 0.8752)
(0.832, 0.8760)
(0.833, 0.8768)
(0.834, 0.8775)
(0.835, 0.8783)
(0.836, 0.8790)
(0.837, 0.8798)
(0.838, 0.8805)
(0.839, 0.8813)
(0.840, 0.8821)
(0.841, 0.8828)
(0.842, 0.8836)
(0.843, 0.8843)
(0.844, 0.8851)
(0.845, 0.8858)
(0.846, 0.8866)
(0.847, 0.8873)
(0.848, 0.8881)
(0.849, 0.8888)
(0.850, 0.8896)
(0.851, 0.8904)
(0.852, 0.8911)
(0.853, 0.8919)
(0.854, 0.8926)
(0.855, 0.8934)
(0.856, 0.8941)
(0.857, 0.8949)
(0.858, 0.8956)
(0.859, 0.8964)
(0.860, 0.8971)
(0.861, 0.8979)
(0.862, 0.8986)
(0.863, 0.8994)
(0.864, 0.9001)
(0.865, 0.9009)
(0.866, 0.9016)
(0.867, 0.9024)
(0.868, 0.9031)
(0.869, 0.9039)
(0.870, 0.9046)
(0.871, 0.9054)
(0.872, 0.9061)
(0.873, 0.9069)
(0.874, 0.9076)
(0.875, 0.9084)
(0.876, 0.9091)
(0.877, 0.9099)
(0.878, 0.9106)
(0.879, 0.9113)
(0.880, 0.9121)
(0.881, 0.9128)
(0.882, 0.9136)
(0.883, 0.9143)
(0.884, 0.9151)
(0.885, 0.9158)
(0.886, 0.9166)
(0.887, 0.9173)
(0.888, 0.9181)
(0.889, 0.9188)
(0.890, 0.9195)
(0.891, 0.9203)
(0.892, 0.9210)
(0.893, 0.9218)
(0.894, 0.9225)
(0.895, 0.9233)
(0.896, 0.9240)
(0.897, 0.9247)
(0.898, 0.9255)
(0.899, 0.9262)
(0.900, 0.9270)
(0.901, 0.9277)
(0.902, 0.9284)
(0.903, 0.9292)
(0.904, 0.9299)
(0.905, 0.9307)
(0.906, 0.9314)
(0.907, 0.9321)
(0.908, 0.9329)
(0.909, 0.9336)
(0.910, 0.9344)
(0.911, 0.9351)
(0.912, 0.9358)
(0.913, 0.9366)
(0.914, 0.9373)
(0.915, 0.9381)
(0.916, 0.9388)
(0.917, 0.9395)
(0.918, 0.9403)
(0.919, 0.9410)
(0.920, 0.9417)
(0.921, 0.9425)
(0.922, 0.9432)
(0.923, 0.9440)
(0.924, 0.9447)
(0.925, 0.9454)
(0.926, 0.9462)
(0.927, 0.9469)
(0.928, 0.9476)
(0.929, 0.9484)
(0.930, 0.9491)
(0.931, 0.9498)
(0.932, 0.9506)
(0.933, 0.9513)
(0.934, 0.9520)
(0.935, 0.9528)
(0.936, 0.9535)
(0.937, 0.9542)
(0.938, 0.9550)
(0.939, 0.9557)
(0.940, 0.9564)
(0.941, 0.9572)
(0.942, 0.9579)
(0.943, 0.9586)
(0.944, 0.9594)
(0.945, 0.9601)
(0.946, 0.9608)
(0.947, 0.9616)
(0.948, 0.9623)
(0.949, 0.9630)
(0.950, 0.9638)
(0.951, 0.9645)
(0.952, 0.9652)
(0.953, 0.9659)
(0.954, 0.9667)
(0.955, 0.9674)
(0.956, 0.9681)
(0.957, 0.9689)
(0.958, 0.9696)
(0.959, 0.9703)
(0.960, 0.9710)
(0.961, 0.9718)
(0.962, 0.9725)
(0.963, 0.9732)
(0.964, 0.9740)
(0.965, 0.9747)
(0.966, 0.9754)
(0.967, 0.9761)
(0.968, 0.9769)
(0.969, 0.9776)
(0.970, 0.9783)
(0.971, 0.9790)
(0.972, 0.9798)
(0.973, 0.9805)
(0.974, 0.9812)
(0.975, 0.9819)
(0.976, 0.9827)
(0.977, 0.9834)
(0.978, 0.9841)
(0.979, 0.9848)
(0.980, 0.9856)
(0.981, 0.9863)
(0.982, 0.9870)
(0.983, 0.9877)
(0.984, 0.9885)
(0.985, 0.9892)
(0.986, 0.9899)
(0.987, 0.9906)
(0.988, 0.9913)
(0.989, 0.9921)
(0.990, 0.9928)
(0.991, 0.9935)
(0.992, 0.9942)
(0.993, 0.9950)
(0.994, 0.9957)
(0.995, 0.9964)
(0.996, 0.9971)
(0.997, 0.9978)
(0.998, 0.9986)
(0.999, 0.9993)
(1.000, 1.0000)
}
\defdata{emaprofile-exp1}{
(0.000, 0.0000)
(0.004, 0.0000)
(0.005, 0.0001)
(0.006, 0.0001)
(0.007, 0.0001)
(0.008, 0.0001)
(0.009, 0.0002)
(0.010, 0.0002)
(0.011, 0.0003)
(0.012, 0.0003)
(0.013, 0.0004)
(0.014, 0.0004)
(0.015, 0.0005)
(0.016, 0.0005)
(0.017, 0.0006)
(0.018, 0.0007)
(0.019, 0.0007)
(0.020, 0.0008)
(0.021, 0.0009)
(0.022, 0.0009)
(0.023, 0.0010)
(0.024, 0.0011)
(0.025, 0.0012)
(0.026, 0.0013)
(0.027, 0.0014)
(0.028, 0.0015)
(0.029, 0.0016)
(0.030, 0.0017)
(0.031, 0.0018)
(0.032, 0.0019)
(0.033, 0.0020)
(0.034, 0.0021)
(0.035, 0.0022)
(0.036, 0.0023)
(0.037, 0.0024)
(0.038, 0.0025)
(0.039, 0.0027)
(0.040, 0.0028)
(0.041, 0.0029)
(0.042, 0.0031)
(0.043, 0.0032)
(0.044, 0.0033)
(0.045, 0.0035)
(0.046, 0.0036)
(0.047, 0.0038)
(0.048, 0.0039)
(0.049, 0.0041)
(0.050, 0.0042)
(0.051, 0.0044)
(0.052, 0.0045)
(0.053, 0.0047)
(0.054, 0.0048)
(0.055, 0.0050)
(0.056, 0.0052)
(0.057, 0.0053)
(0.058, 0.0055)
(0.059, 0.0057)
(0.060, 0.0059)
(0.061, 0.0061)
(0.062, 0.0062)
(0.063, 0.0064)
(0.064, 0.0066)
(0.065, 0.0068)
(0.066, 0.0070)
(0.067, 0.0072)
(0.068, 0.0074)
(0.069, 0.0076)
(0.070, 0.0078)
(0.071, 0.0080)
(0.072, 0.0082)
(0.073, 0.0084)
(0.074, 0.0086)
(0.075, 0.0088)
(0.076, 0.0090)
(0.077, 0.0093)
(0.078, 0.0095)
(0.079, 0.0097)
(0.080, 0.0099)
(0.081, 0.0102)
(0.082, 0.0104)
(0.083, 0.0106)
(0.084, 0.0109)
(0.085, 0.0111)
(0.086, 0.0113)
(0.087, 0.0116)
(0.088, 0.0118)
(0.089, 0.0121)
(0.090, 0.0123)
(0.091, 0.0126)
(0.092, 0.0128)
(0.093, 0.0131)
(0.094, 0.0133)
(0.095, 0.0136)
(0.096, 0.0138)
(0.097, 0.0141)
(0.098, 0.0144)
(0.099, 0.0146)
(0.100, 0.0149)
(0.101, 0.0152)
(0.102, 0.0155)
(0.103, 0.0157)
(0.104, 0.0160)
(0.105, 0.0163)
(0.106, 0.0166)
(0.107, 0.0169)
(0.108, 0.0172)
(0.109, 0.0175)
(0.110, 0.0178)
(0.111, 0.0181)
(0.112, 0.0184)
(0.113, 0.0187)
(0.114, 0.0190)
(0.115, 0.0193)
(0.116, 0.0196)
(0.117, 0.0199)
(0.118, 0.0202)
(0.119, 0.0205)
(0.120, 0.0208)
(0.121, 0.0211)
(0.122, 0.0215)
(0.123, 0.0218)
(0.124, 0.0221)
(0.125, 0.0224)
(0.126, 0.0228)
(0.127, 0.0231)
(0.128, 0.0234)
(0.129, 0.0238)
(0.130, 0.0241)
(0.131, 0.0244)
(0.132, 0.0248)
(0.133, 0.0251)
(0.134, 0.0255)
(0.135, 0.0258)
(0.136, 0.0262)
(0.137, 0.0265)
(0.138, 0.0269)
(0.139, 0.0272)
(0.140, 0.0276)
(0.141, 0.0279)
(0.142, 0.0283)
(0.143, 0.0287)
(0.144, 0.0290)
(0.145, 0.0294)
(0.146, 0.0298)
(0.147, 0.0302)
(0.148, 0.0305)
(0.149, 0.0309)
(0.150, 0.0313)
(0.151, 0.0317)
(0.152, 0.0321)
(0.153, 0.0324)
(0.154, 0.0328)
(0.155, 0.0332)
(0.156, 0.0336)
(0.157, 0.0340)
(0.158, 0.0344)
(0.159, 0.0348)
(0.160, 0.0352)
(0.161, 0.0356)
(0.162, 0.0360)
(0.163, 0.0364)
(0.164, 0.0368)
(0.165, 0.0372)
(0.166, 0.0376)
(0.167, 0.0381)
(0.168, 0.0385)
(0.169, 0.0389)
(0.170, 0.0393)
(0.171, 0.0397)
(0.172, 0.0402)
(0.173, 0.0406)
(0.174, 0.0410)
(0.175, 0.0415)
(0.176, 0.0419)
(0.177, 0.0423)
(0.178, 0.0428)
(0.179, 0.0432)
(0.180, 0.0436)
(0.181, 0.0441)
(0.182, 0.0445)
(0.183, 0.0450)
(0.184, 0.0454)
(0.185, 0.0459)
(0.186, 0.0463)
(0.187, 0.0468)
(0.188, 0.0473)
(0.189, 0.0477)
(0.190, 0.0482)
(0.191, 0.0486)
(0.192, 0.0491)
(0.193, 0.0496)
(0.194, 0.0500)
(0.195, 0.0505)
(0.196, 0.0510)
(0.197, 0.0515)
(0.198, 0.0519)
(0.199, 0.0524)
(0.200, 0.0529)
(0.201, 0.0534)
(0.202, 0.0539)
(0.203, 0.0544)
(0.204, 0.0549)
(0.205, 0.0554)
(0.206, 0.0558)
(0.207, 0.0563)
(0.208, 0.0568)
(0.209, 0.0573)
(0.210, 0.0578)
(0.211, 0.0583)
(0.212, 0.0589)
(0.213, 0.0594)
(0.214, 0.0599)
(0.215, 0.0604)
(0.216, 0.0609)
(0.217, 0.0614)
(0.218, 0.0619)
(0.219, 0.0624)
(0.220, 0.0630)
(0.221, 0.0635)
(0.222, 0.0640)
(0.223, 0.0645)
(0.224, 0.0651)
(0.225, 0.0656)
(0.226, 0.0661)
(0.227, 0.0667)
(0.228, 0.0672)
(0.229, 0.0678)
(0.230, 0.0683)
(0.231, 0.0688)
(0.232, 0.0694)
(0.233, 0.0699)
(0.234, 0.0705)
(0.235, 0.0710)
(0.236, 0.0716)
(0.237, 0.0721)
(0.238, 0.0727)
(0.239, 0.0733)
(0.240, 0.0738)
(0.241, 0.0744)
(0.242, 0.0749)
(0.243, 0.0755)
(0.244, 0.0761)
(0.245, 0.0766)
(0.246, 0.0772)
(0.247, 0.0778)
(0.248, 0.0784)
(0.249, 0.0789)
(0.250, 0.0795)
(0.251, 0.0801)
(0.252, 0.0807)
(0.253, 0.0813)
(0.254, 0.0819)
(0.255, 0.0825)
(0.256, 0.0830)
(0.257, 0.0836)
(0.258, 0.0842)
(0.259, 0.0848)
(0.260, 0.0854)
(0.261, 0.0860)
(0.262, 0.0866)
(0.263, 0.0872)
(0.264, 0.0878)
(0.265, 0.0885)
(0.266, 0.0891)
(0.267, 0.0897)
(0.268, 0.0903)
(0.269, 0.0909)
(0.270, 0.0915)
(0.271, 0.0921)
(0.272, 0.0928)
(0.273, 0.0934)
(0.274, 0.0940)
(0.275, 0.0946)
(0.276, 0.0953)
(0.277, 0.0959)
(0.278, 0.0965)
(0.279, 0.0972)
(0.280, 0.0978)
(0.281, 0.0985)
(0.282, 0.0991)
(0.283, 0.0997)
(0.284, 0.1004)
(0.285, 0.1010)
(0.286, 0.1017)
(0.287, 0.1023)
(0.288, 0.1030)
(0.289, 0.1036)
(0.290, 0.1043)
(0.291, 0.1049)
(0.292, 0.1056)
(0.293, 0.1063)
(0.294, 0.1069)
(0.295, 0.1076)
(0.296, 0.1083)
(0.297, 0.1089)
(0.298, 0.1096)
(0.299, 0.1103)
(0.300, 0.1109)
(0.301, 0.1116)
(0.302, 0.1123)
(0.303, 0.1130)
(0.304, 0.1137)
(0.305, 0.1143)
(0.306, 0.1150)
(0.307, 0.1157)
(0.308, 0.1164)
(0.309, 0.1171)
(0.310, 0.1178)
(0.311, 0.1185)
(0.312, 0.1192)
(0.313, 0.1199)
(0.314, 0.1206)
(0.315, 0.1213)
(0.316, 0.1220)
(0.317, 0.1227)
(0.318, 0.1234)
(0.319, 0.1241)
(0.320, 0.1248)
(0.321, 0.1255)
(0.322, 0.1263)
(0.323, 0.1270)
(0.324, 0.1277)
(0.325, 0.1284)
(0.326, 0.1291)
(0.327, 0.1299)
(0.328, 0.1306)
(0.329, 0.1313)
(0.330, 0.1320)
(0.331, 0.1328)
(0.332, 0.1335)
(0.333, 0.1342)
(0.334, 0.1350)
(0.335, 0.1357)
(0.336, 0.1365)
(0.337, 0.1372)
(0.338, 0.1379)
(0.339, 0.1387)
(0.340, 0.1394)
(0.341, 0.1402)
(0.342, 0.1409)
(0.343, 0.1417)
(0.344, 0.1424)
(0.345, 0.1432)
(0.346, 0.1440)
(0.347, 0.1447)
(0.348, 0.1455)
(0.349, 0.1463)
(0.350, 0.1470)
(0.351, 0.1478)
(0.352, 0.1486)
(0.353, 0.1493)
(0.354, 0.1501)
(0.355, 0.1509)
(0.356, 0.1517)
(0.357, 0.1524)
(0.358, 0.1532)
(0.359, 0.1540)
(0.360, 0.1548)
(0.361, 0.1556)
(0.362, 0.1564)
(0.363, 0.1571)
(0.364, 0.1579)
(0.365, 0.1587)
(0.366, 0.1595)
(0.367, 0.1603)
(0.368, 0.1611)
(0.369, 0.1619)
(0.370, 0.1627)
(0.371, 0.1635)
(0.372, 0.1643)
(0.373, 0.1651)
(0.374, 0.1659)
(0.375, 0.1668)
(0.376, 0.1676)
(0.377, 0.1684)
(0.378, 0.1692)
(0.379, 0.1700)
(0.380, 0.1708)
(0.381, 0.1717)
(0.382, 0.1725)
(0.383, 0.1733)
(0.384, 0.1741)
(0.385, 0.1750)
(0.386, 0.1758)
(0.387, 0.1766)
(0.388, 0.1775)
(0.389, 0.1783)
(0.390, 0.1791)
(0.391, 0.1800)
(0.392, 0.1808)
(0.393, 0.1817)
(0.394, 0.1825)
(0.395, 0.1834)
(0.396, 0.1842)
(0.397, 0.1851)
(0.398, 0.1859)
(0.399, 0.1868)
(0.400, 0.1876)
(0.401, 0.1885)
(0.402, 0.1893)
(0.403, 0.1902)
(0.404, 0.1911)
(0.405, 0.1919)
(0.406, 0.1928)
(0.407, 0.1937)
(0.408, 0.1945)
(0.409, 0.1954)
(0.410, 0.1963)
(0.411, 0.1971)
(0.412, 0.1980)
(0.413, 0.1989)
(0.414, 0.1998)
(0.415, 0.2007)
(0.416, 0.2016)
(0.417, 0.2024)
(0.418, 0.2033)
(0.419, 0.2042)
(0.420, 0.2051)
(0.421, 0.2060)
(0.422, 0.2069)
(0.423, 0.2078)
(0.424, 0.2087)
(0.425, 0.2096)
(0.426, 0.2105)
(0.427, 0.2114)
(0.428, 0.2123)
(0.429, 0.2132)
(0.430, 0.2141)
(0.431, 0.2150)
(0.432, 0.2159)
(0.433, 0.2168)
(0.434, 0.2178)
(0.435, 0.2187)
(0.436, 0.2196)
(0.437, 0.2205)
(0.438, 0.2214)
(0.439, 0.2224)
(0.440, 0.2233)
(0.441, 0.2242)
(0.442, 0.2251)
(0.443, 0.2261)
(0.444, 0.2270)
(0.445, 0.2279)
(0.446, 0.2289)
(0.447, 0.2298)
(0.448, 0.2308)
(0.449, 0.2317)
(0.450, 0.2326)
(0.451, 0.2336)
(0.452, 0.2345)
(0.453, 0.2355)
(0.454, 0.2364)
(0.455, 0.2374)
(0.456, 0.2383)
(0.457, 0.2393)
(0.458, 0.2403)
(0.459, 0.2412)
(0.460, 0.2422)
(0.461, 0.2431)
(0.462, 0.2441)
(0.463, 0.2451)
(0.464, 0.2460)
(0.465, 0.2470)
(0.466, 0.2480)
(0.467, 0.2489)
(0.468, 0.2499)
(0.469, 0.2509)
(0.470, 0.2519)
(0.471, 0.2529)
(0.472, 0.2538)
(0.473, 0.2548)
(0.474, 0.2558)
(0.475, 0.2568)
(0.476, 0.2578)
(0.477, 0.2588)
(0.478, 0.2598)
(0.479, 0.2608)
(0.480, 0.2617)
(0.481, 0.2627)
(0.482, 0.2637)
(0.483, 0.2647)
(0.484, 0.2657)
(0.485, 0.2667)
(0.486, 0.2678)
(0.487, 0.2688)
(0.488, 0.2698)
(0.489, 0.2708)
(0.490, 0.2718)
(0.491, 0.2728)
(0.492, 0.2738)
(0.493, 0.2748)
(0.494, 0.2759)
(0.495, 0.2769)
(0.496, 0.2779)
(0.497, 0.2789)
(0.498, 0.2799)
(0.499, 0.2810)
(0.500, 0.2820)
(0.501, 0.2830)
(0.502, 0.2841)
(0.503, 0.2851)
(0.504, 0.2861)
(0.505, 0.2872)
(0.506, 0.2882)
(0.507, 0.2893)
(0.508, 0.2903)
(0.509, 0.2913)
(0.510, 0.2924)
(0.511, 0.2934)
(0.512, 0.2945)
(0.513, 0.2955)
(0.514, 0.2966)
(0.515, 0.2976)
(0.516, 0.2987)
(0.517, 0.2998)
(0.518, 0.3008)
(0.519, 0.3019)
(0.520, 0.3029)
(0.521, 0.3040)
(0.522, 0.3051)
(0.523, 0.3061)
(0.524, 0.3072)
(0.525, 0.3083)
(0.526, 0.3094)
(0.527, 0.3104)
(0.528, 0.3115)
(0.529, 0.3126)
(0.530, 0.3137)
(0.531, 0.3147)
(0.532, 0.3158)
(0.533, 0.3169)
(0.534, 0.3180)
(0.535, 0.3191)
(0.536, 0.3202)
(0.537, 0.3213)
(0.538, 0.3224)
(0.539, 0.3235)
(0.540, 0.3246)
(0.541, 0.3257)
(0.542, 0.3268)
(0.543, 0.3279)
(0.544, 0.3290)
(0.545, 0.3301)
(0.546, 0.3312)
(0.547, 0.3323)
(0.548, 0.3334)
(0.549, 0.3345)
(0.550, 0.3356)
(0.551, 0.3367)
(0.552, 0.3379)
(0.553, 0.3390)
(0.554, 0.3401)
(0.555, 0.3412)
(0.556, 0.3423)
(0.557, 0.3435)
(0.558, 0.3446)
(0.559, 0.3457)
(0.560, 0.3468)
(0.561, 0.3480)
(0.562, 0.3491)
(0.563, 0.3502)
(0.564, 0.3514)
(0.565, 0.3525)
(0.566, 0.3537)
(0.567, 0.3548)
(0.568, 0.3559)
(0.569, 0.3571)
(0.570, 0.3582)
(0.571, 0.3594)
(0.572, 0.3605)
(0.573, 0.3617)
(0.574, 0.3628)
(0.575, 0.3640)
(0.576, 0.3652)
(0.577, 0.3663)
(0.578, 0.3675)
(0.579, 0.3686)
(0.580, 0.3698)
(0.581, 0.3710)
(0.582, 0.3721)
(0.583, 0.3733)
(0.584, 0.3745)
(0.585, 0.3756)
(0.586, 0.3768)
(0.587, 0.3780)
(0.588, 0.3792)
(0.589, 0.3803)
(0.590, 0.3815)
(0.591, 0.3827)
(0.592, 0.3839)
(0.593, 0.3851)
(0.594, 0.3863)
(0.595, 0.3875)
(0.596, 0.3886)
(0.597, 0.3898)
(0.598, 0.3910)
(0.599, 0.3922)
(0.600, 0.3934)
(0.601, 0.3946)
(0.602, 0.3958)
(0.603, 0.3970)
(0.604, 0.3982)
(0.605, 0.3994)
(0.606, 0.4006)
(0.607, 0.4018)
(0.608, 0.4031)
(0.609, 0.4043)
(0.610, 0.4055)
(0.611, 0.4067)
(0.612, 0.4079)
(0.613, 0.4091)
(0.614, 0.4103)
(0.615, 0.4116)
(0.616, 0.4128)
(0.617, 0.4140)
(0.618, 0.4152)
(0.619, 0.4165)
(0.620, 0.4177)
(0.621, 0.4189)
(0.622, 0.4202)
(0.623, 0.4214)
(0.624, 0.4226)
(0.625, 0.4239)
(0.626, 0.4251)
(0.627, 0.4264)
(0.628, 0.4276)
(0.629, 0.4288)
(0.630, 0.4301)
(0.631, 0.4313)
(0.632, 0.4326)
(0.633, 0.4338)
(0.634, 0.4351)
(0.635, 0.4363)
(0.636, 0.4376)
(0.637, 0.4389)
(0.638, 0.4401)
(0.639, 0.4414)
(0.640, 0.4426)
(0.641, 0.4439)
(0.642, 0.4452)
(0.643, 0.4464)
(0.644, 0.4477)
(0.645, 0.4490)
(0.646, 0.4502)
(0.647, 0.4515)
(0.648, 0.4528)
(0.649, 0.4541)
(0.650, 0.4553)
(0.651, 0.4566)
(0.652, 0.4579)
(0.653, 0.4592)
(0.654, 0.4605)
(0.655, 0.4618)
(0.656, 0.4631)
(0.657, 0.4643)
(0.658, 0.4656)
(0.659, 0.4669)
(0.660, 0.4682)
(0.661, 0.4695)
(0.662, 0.4708)
(0.663, 0.4721)
(0.664, 0.4734)
(0.665, 0.4747)
(0.666, 0.4760)
(0.667, 0.4773)
(0.668, 0.4786)
(0.669, 0.4799)
(0.670, 0.4813)
(0.671, 0.4826)
(0.672, 0.4839)
(0.673, 0.4852)
(0.674, 0.4865)
(0.675, 0.4878)
(0.676, 0.4892)
(0.677, 0.4905)
(0.678, 0.4918)
(0.679, 0.4931)
(0.680, 0.4945)
(0.681, 0.4958)
(0.682, 0.4971)
(0.683, 0.4984)
(0.684, 0.4998)
(0.685, 0.5011)
(0.686, 0.5025)
(0.687, 0.5038)
(0.688, 0.5051)
(0.689, 0.5065)
(0.690, 0.5078)
(0.691, 0.5092)
(0.692, 0.5105)
(0.693, 0.5119)
(0.694, 0.5132)
(0.695, 0.5146)
(0.696, 0.5159)
(0.697, 0.5173)
(0.698, 0.5186)
(0.699, 0.5200)
(0.700, 0.5213)
(0.701, 0.5227)
(0.702, 0.5241)
(0.703, 0.5254)
(0.704, 0.5268)
(0.705, 0.5282)
(0.706, 0.5295)
(0.707, 0.5309)
(0.708, 0.5323)
(0.709, 0.5336)
(0.710, 0.5350)
(0.711, 0.5364)
(0.712, 0.5378)
(0.713, 0.5392)
(0.714, 0.5405)
(0.715, 0.5419)
(0.716, 0.5433)
(0.717, 0.5447)
(0.718, 0.5461)
(0.719, 0.5475)
(0.720, 0.5489)
(0.721, 0.5502)
(0.722, 0.5516)
(0.723, 0.5530)
(0.724, 0.5544)
(0.725, 0.5558)
(0.726, 0.5572)
(0.727, 0.5586)
(0.728, 0.5600)
(0.729, 0.5615)
(0.730, 0.5629)
(0.731, 0.5643)
(0.732, 0.5657)
(0.733, 0.5671)
(0.734, 0.5685)
(0.735, 0.5699)
(0.736, 0.5713)
(0.737, 0.5728)
(0.738, 0.5742)
(0.739, 0.5756)
(0.740, 0.5770)
(0.741, 0.5784)
(0.742, 0.5799)
(0.743, 0.5813)
(0.744, 0.5827)
(0.745, 0.5842)
(0.746, 0.5856)
(0.747, 0.5870)
(0.748, 0.5885)
(0.749, 0.5899)
(0.750, 0.5913)
(0.751, 0.5928)
(0.752, 0.5942)
(0.753, 0.5957)
(0.754, 0.5971)
(0.755, 0.5986)
(0.756, 0.6000)
(0.757, 0.6015)
(0.758, 0.6029)
(0.759, 0.6044)
(0.760, 0.6058)
(0.761, 0.6073)
(0.762, 0.6087)
(0.763, 0.6102)
(0.764, 0.6117)
(0.765, 0.6131)
(0.766, 0.6146)
(0.767, 0.6160)
(0.768, 0.6175)
(0.769, 0.6190)
(0.770, 0.6205)
(0.771, 0.6219)
(0.772, 0.6234)
(0.773, 0.6249)
(0.774, 0.6264)
(0.775, 0.6278)
(0.776, 0.6293)
(0.777, 0.6308)
(0.778, 0.6323)
(0.779, 0.6338)
(0.780, 0.6352)
(0.781, 0.6367)
(0.782, 0.6382)
(0.783, 0.6397)
(0.784, 0.6412)
(0.785, 0.6427)
(0.786, 0.6442)
(0.787, 0.6457)
(0.788, 0.6472)
(0.789, 0.6487)
(0.790, 0.6502)
(0.791, 0.6517)
(0.792, 0.6532)
(0.793, 0.6547)
(0.794, 0.6562)
(0.795, 0.6577)
(0.796, 0.6592)
(0.797, 0.6608)
(0.798, 0.6623)
(0.799, 0.6638)
(0.800, 0.6653)
(0.801, 0.6668)
(0.802, 0.6683)
(0.803, 0.6699)
(0.804, 0.6714)
(0.805, 0.6729)
(0.806, 0.6744)
(0.807, 0.6760)
(0.808, 0.6775)
(0.809, 0.6790)
(0.810, 0.6806)
(0.811, 0.6821)
(0.812, 0.6836)
(0.813, 0.6852)
(0.814, 0.6867)
(0.815, 0.6883)
(0.816, 0.6898)
(0.817, 0.6914)
(0.818, 0.6929)
(0.819, 0.6944)
(0.820, 0.6960)
(0.821, 0.6975)
(0.822, 0.6991)
(0.823, 0.7007)
(0.824, 0.7022)
(0.825, 0.7038)
(0.826, 0.7053)
(0.827, 0.7069)
(0.828, 0.7084)
(0.829, 0.7100)
(0.830, 0.7116)
(0.831, 0.7131)
(0.832, 0.7147)
(0.833, 0.7163)
(0.834, 0.7178)
(0.835, 0.7194)
(0.836, 0.7210)
(0.837, 0.7226)
(0.838, 0.7241)
(0.839, 0.7257)
(0.840, 0.7273)
(0.841, 0.7289)
(0.842, 0.7305)
(0.843, 0.7321)
(0.844, 0.7336)
(0.845, 0.7352)
(0.846, 0.7368)
(0.847, 0.7384)
(0.848, 0.7400)
(0.849, 0.7416)
(0.850, 0.7432)
(0.851, 0.7448)
(0.852, 0.7464)
(0.853, 0.7480)
(0.854, 0.7496)
(0.855, 0.7512)
(0.856, 0.7528)
(0.857, 0.7544)
(0.858, 0.7560)
(0.859, 0.7576)
(0.860, 0.7592)
(0.861, 0.7609)
(0.862, 0.7625)
(0.863, 0.7641)
(0.864, 0.7657)
(0.865, 0.7673)
(0.866, 0.7689)
(0.867, 0.7706)
(0.868, 0.7722)
(0.869, 0.7738)
(0.870, 0.7754)
(0.871, 0.7771)
(0.872, 0.7787)
(0.873, 0.7803)
(0.874, 0.7820)
(0.875, 0.7836)
(0.876, 0.7852)
(0.877, 0.7869)
(0.878, 0.7885)
(0.879, 0.7902)
(0.880, 0.7918)
(0.881, 0.7934)
(0.882, 0.7951)
(0.883, 0.7967)
(0.884, 0.7984)
(0.885, 0.8000)
(0.886, 0.8017)
(0.887, 0.8033)
(0.888, 0.8050)
(0.889, 0.8066)
(0.890, 0.8083)
(0.891, 0.8100)
(0.892, 0.8116)
(0.893, 0.8133)
(0.894, 0.8150)
(0.895, 0.8166)
(0.896, 0.8183)
(0.897, 0.8200)
(0.898, 0.8216)
(0.899, 0.8233)
(0.900, 0.8250)
(0.901, 0.8266)
(0.902, 0.8283)
(0.903, 0.8300)
(0.904, 0.8317)
(0.905, 0.8334)
(0.906, 0.8350)
(0.907, 0.8367)
(0.908, 0.8384)
(0.909, 0.8401)
(0.910, 0.8418)
(0.911, 0.8435)
(0.912, 0.8452)
(0.913, 0.8469)
(0.914, 0.8486)
(0.915, 0.8503)
(0.916, 0.8519)
(0.917, 0.8536)
(0.918, 0.8553)
(0.919, 0.8571)
(0.920, 0.8588)
(0.921, 0.8605)
(0.922, 0.8622)
(0.923, 0.8639)
(0.924, 0.8656)
(0.925, 0.8673)
(0.926, 0.8690)
(0.927, 0.8707)
(0.928, 0.8724)
(0.929, 0.8742)
(0.930, 0.8759)
(0.931, 0.8776)
(0.932, 0.8793)
(0.933, 0.8810)
(0.934, 0.8828)
(0.935, 0.8845)
(0.936, 0.8862)
(0.937, 0.8880)
(0.938, 0.8897)
(0.939, 0.8914)
(0.940, 0.8932)
(0.941, 0.8949)
(0.942, 0.8966)
(0.943, 0.8984)
(0.944, 0.9001)
(0.945, 0.9018)
(0.946, 0.9036)
(0.947, 0.9053)
(0.948, 0.9071)
(0.949, 0.9088)
(0.950, 0.9106)
(0.951, 0.9123)
(0.952, 0.9141)
(0.953, 0.9158)
(0.954, 0.9176)
(0.955, 0.9194)
(0.956, 0.9211)
(0.957, 0.9229)
(0.958, 0.9246)
(0.959, 0.9264)
(0.960, 0.9282)
(0.961, 0.9299)
(0.962, 0.9317)
(0.963, 0.9335)
(0.964, 0.9352)
(0.965, 0.9370)
(0.966, 0.9388)
(0.967, 0.9406)
(0.968, 0.9423)
(0.969, 0.9441)
(0.970, 0.9459)
(0.971, 0.9477)
(0.972, 0.9495)
(0.973, 0.9512)
(0.974, 0.9530)
(0.975, 0.9548)
(0.976, 0.9566)
(0.977, 0.9584)
(0.978, 0.9602)
(0.979, 0.9620)
(0.980, 0.9638)
(0.981, 0.9656)
(0.982, 0.9674)
(0.983, 0.9692)
(0.984, 0.9710)
(0.985, 0.9728)
(0.986, 0.9746)
(0.987, 0.9764)
(0.988, 0.9782)
(0.989, 0.9800)
(0.990, 0.9818)
(0.991, 0.9836)
(0.992, 0.9854)
(0.993, 0.9873)
(0.994, 0.9891)
(0.995, 0.9909)
(0.996, 0.9927)
(0.997, 0.9945)
(0.998, 0.9964)
(0.999, 0.9982)
(1.000, 1.0000)
}
\defdata{emaprofile-exp2}{
(0.000, 0.0000)
(0.059, 0.0000)
(0.060, 0.0000)
(0.061, 0.0000)
(0.062, 0.0001)
(0.063, 0.0001)
(0.064, 0.0001)
(0.065, 0.0001)
(0.066, 0.0001)
(0.067, 0.0001)
(0.068, 0.0001)
(0.069, 0.0001)
(0.070, 0.0001)
(0.071, 0.0001)
(0.072, 0.0001)
(0.073, 0.0001)
(0.074, 0.0001)
(0.075, 0.0001)
(0.076, 0.0001)
(0.077, 0.0001)
(0.078, 0.0001)
(0.079, 0.0001)
(0.080, 0.0001)
(0.081, 0.0001)
(0.082, 0.0001)
(0.083, 0.0001)
(0.084, 0.0001)
(0.085, 0.0002)
(0.086, 0.0002)
(0.087, 0.0002)
(0.088, 0.0002)
(0.089, 0.0002)
(0.090, 0.0002)
(0.091, 0.0002)
(0.092, 0.0002)
(0.093, 0.0002)
(0.094, 0.0002)
(0.095, 0.0002)
(0.096, 0.0002)
(0.097, 0.0002)
(0.098, 0.0003)
(0.099, 0.0003)
(0.100, 0.0003)
(0.101, 0.0003)
(0.102, 0.0003)
(0.103, 0.0003)
(0.104, 0.0003)
(0.105, 0.0003)
(0.106, 0.0003)
(0.107, 0.0004)
(0.108, 0.0004)
(0.109, 0.0004)
(0.110, 0.0004)
(0.111, 0.0004)
(0.112, 0.0004)
(0.113, 0.0004)
(0.114, 0.0004)
(0.115, 0.0005)
(0.116, 0.0005)
(0.117, 0.0005)
(0.118, 0.0005)
(0.119, 0.0005)
(0.120, 0.0005)
(0.121, 0.0005)
(0.122, 0.0006)
(0.123, 0.0006)
(0.124, 0.0006)
(0.125, 0.0006)
(0.126, 0.0006)
(0.127, 0.0006)
(0.128, 0.0007)
(0.129, 0.0007)
(0.130, 0.0007)
(0.131, 0.0007)
(0.132, 0.0007)
(0.133, 0.0008)
(0.134, 0.0008)
(0.135, 0.0008)
(0.136, 0.0008)
(0.137, 0.0008)
(0.138, 0.0009)
(0.139, 0.0009)
(0.140, 0.0009)
(0.141, 0.0009)
(0.142, 0.0010)
(0.143, 0.0010)
(0.144, 0.0010)
(0.145, 0.0010)
(0.146, 0.0011)
(0.147, 0.0011)
(0.148, 0.0011)
(0.149, 0.0011)
(0.150, 0.0012)
(0.151, 0.0012)
(0.152, 0.0012)
(0.153, 0.0013)
(0.154, 0.0013)
(0.155, 0.0013)
(0.156, 0.0013)
(0.157, 0.0014)
(0.158, 0.0014)
(0.159, 0.0014)
(0.160, 0.0015)
(0.161, 0.0015)
(0.162, 0.0015)
(0.163, 0.0016)
(0.164, 0.0016)
(0.165, 0.0016)
(0.166, 0.0017)
(0.167, 0.0017)
(0.168, 0.0018)
(0.169, 0.0018)
(0.170, 0.0018)
(0.171, 0.0019)
(0.172, 0.0019)
(0.173, 0.0019)
(0.174, 0.0020)
(0.175, 0.0020)
(0.176, 0.0021)
(0.177, 0.0021)
(0.178, 0.0022)
(0.179, 0.0022)
(0.180, 0.0022)
(0.181, 0.0023)
(0.182, 0.0023)
(0.183, 0.0024)
(0.184, 0.0024)
(0.185, 0.0025)
(0.186, 0.0025)
(0.187, 0.0026)
(0.188, 0.0026)
(0.189, 0.0027)
(0.190, 0.0027)
(0.191, 0.0028)
(0.192, 0.0028)
(0.193, 0.0029)
(0.194, 0.0029)
(0.195, 0.0030)
(0.196, 0.0030)
(0.197, 0.0031)
(0.198, 0.0031)
(0.199, 0.0032)
(0.200, 0.0033)
(0.201, 0.0033)
(0.202, 0.0034)
(0.203, 0.0034)
(0.204, 0.0035)
(0.205, 0.0036)
(0.206, 0.0036)
(0.207, 0.0037)
(0.208, 0.0037)
(0.209, 0.0038)
(0.210, 0.0039)
(0.211, 0.0039)
(0.212, 0.0040)
(0.213, 0.0041)
(0.214, 0.0041)
(0.215, 0.0042)
(0.216, 0.0043)
(0.217, 0.0044)
(0.218, 0.0044)
(0.219, 0.0045)
(0.220, 0.0046)
(0.221, 0.0046)
(0.222, 0.0047)
(0.223, 0.0048)
(0.224, 0.0049)
(0.225, 0.0050)
(0.226, 0.0050)
(0.227, 0.0051)
(0.228, 0.0052)
(0.229, 0.0053)
(0.230, 0.0054)
(0.231, 0.0054)
(0.232, 0.0055)
(0.233, 0.0056)
(0.234, 0.0057)
(0.235, 0.0058)
(0.236, 0.0059)
(0.237, 0.0060)
(0.238, 0.0061)
(0.239, 0.0061)
(0.240, 0.0062)
(0.241, 0.0063)
(0.242, 0.0064)
(0.243, 0.0065)
(0.244, 0.0066)
(0.245, 0.0067)
(0.246, 0.0068)
(0.247, 0.0069)
(0.248, 0.0070)
(0.249, 0.0071)
(0.250, 0.0072)
(0.251, 0.0073)
(0.252, 0.0074)
(0.253, 0.0075)
(0.254, 0.0076)
(0.255, 0.0077)
(0.256, 0.0078)
(0.257, 0.0080)
(0.258, 0.0081)
(0.259, 0.0082)
(0.260, 0.0083)
(0.261, 0.0084)
(0.262, 0.0085)
(0.263, 0.0086)
(0.264, 0.0088)
(0.265, 0.0089)
(0.266, 0.0090)
(0.267, 0.0091)
(0.268, 0.0092)
(0.269, 0.0094)
(0.270, 0.0095)
(0.271, 0.0096)
(0.272, 0.0097)
(0.273, 0.0099)
(0.274, 0.0100)
(0.275, 0.0101)
(0.276, 0.0103)
(0.277, 0.0104)
(0.278, 0.0105)
(0.279, 0.0107)
(0.280, 0.0108)
(0.281, 0.0109)
(0.282, 0.0111)
(0.283, 0.0112)
(0.284, 0.0113)
(0.285, 0.0115)
(0.286, 0.0116)
(0.287, 0.0118)
(0.288, 0.0119)
(0.289, 0.0121)
(0.290, 0.0122)
(0.291, 0.0124)
(0.292, 0.0125)
(0.293, 0.0127)
(0.294, 0.0128)
(0.295, 0.0130)
(0.296, 0.0131)
(0.297, 0.0133)
(0.298, 0.0135)
(0.299, 0.0136)
(0.300, 0.0138)
(0.301, 0.0140)
(0.302, 0.0141)
(0.303, 0.0143)
(0.304, 0.0145)
(0.305, 0.0146)
(0.306, 0.0148)
(0.307, 0.0150)
(0.308, 0.0151)
(0.309, 0.0153)
(0.310, 0.0155)
(0.311, 0.0157)
(0.312, 0.0159)
(0.313, 0.0160)
(0.314, 0.0162)
(0.315, 0.0164)
(0.316, 0.0166)
(0.317, 0.0168)
(0.318, 0.0170)
(0.319, 0.0172)
(0.320, 0.0174)
(0.321, 0.0175)
(0.322, 0.0177)
(0.323, 0.0179)
(0.324, 0.0181)
(0.325, 0.0183)
(0.326, 0.0185)
(0.327, 0.0187)
(0.328, 0.0189)
(0.329, 0.0192)
(0.330, 0.0194)
(0.331, 0.0196)
(0.332, 0.0198)
(0.333, 0.0200)
(0.334, 0.0202)
(0.335, 0.0204)
(0.336, 0.0206)
(0.337, 0.0209)
(0.338, 0.0211)
(0.339, 0.0213)
(0.340, 0.0215)
(0.341, 0.0218)
(0.342, 0.0220)
(0.343, 0.0222)
(0.344, 0.0224)
(0.345, 0.0227)
(0.346, 0.0229)
(0.347, 0.0232)
(0.348, 0.0234)
(0.349, 0.0236)
(0.350, 0.0239)
(0.351, 0.0241)
(0.352, 0.0244)
(0.353, 0.0246)
(0.354, 0.0249)
(0.355, 0.0251)
(0.356, 0.0254)
(0.357, 0.0256)
(0.358, 0.0259)
(0.359, 0.0261)
(0.360, 0.0264)
(0.361, 0.0266)
(0.362, 0.0269)
(0.363, 0.0272)
(0.364, 0.0274)
(0.365, 0.0277)
(0.366, 0.0280)
(0.367, 0.0283)
(0.368, 0.0285)
(0.369, 0.0288)
(0.370, 0.0291)
(0.371, 0.0294)
(0.372, 0.0297)
(0.373, 0.0299)
(0.374, 0.0302)
(0.375, 0.0305)
(0.376, 0.0308)
(0.377, 0.0311)
(0.378, 0.0314)
(0.379, 0.0317)
(0.380, 0.0320)
(0.381, 0.0323)
(0.382, 0.0326)
(0.383, 0.0329)
(0.384, 0.0332)
(0.385, 0.0335)
(0.386, 0.0338)
(0.387, 0.0341)
(0.388, 0.0344)
(0.389, 0.0348)
(0.390, 0.0351)
(0.391, 0.0354)
(0.392, 0.0357)
(0.393, 0.0360)
(0.394, 0.0364)
(0.395, 0.0367)
(0.396, 0.0370)
(0.397, 0.0374)
(0.398, 0.0377)
(0.399, 0.0380)
(0.400, 0.0384)
(0.401, 0.0387)
(0.402, 0.0391)
(0.403, 0.0394)
(0.404, 0.0398)
(0.405, 0.0401)
(0.406, 0.0405)
(0.407, 0.0408)
(0.408, 0.0412)
(0.409, 0.0415)
(0.410, 0.0419)
(0.411, 0.0423)
(0.412, 0.0426)
(0.413, 0.0430)
(0.414, 0.0434)
(0.415, 0.0438)
(0.416, 0.0441)
(0.417, 0.0445)
(0.418, 0.0449)
(0.419, 0.0453)
(0.420, 0.0457)
(0.421, 0.0461)
(0.422, 0.0464)
(0.423, 0.0468)
(0.424, 0.0472)
(0.425, 0.0476)
(0.426, 0.0480)
(0.427, 0.0484)
(0.428, 0.0488)
(0.429, 0.0492)
(0.430, 0.0497)
(0.431, 0.0501)
(0.432, 0.0505)
(0.433, 0.0509)
(0.434, 0.0513)
(0.435, 0.0517)
(0.436, 0.0522)
(0.437, 0.0526)
(0.438, 0.0530)
(0.439, 0.0534)
(0.440, 0.0539)
(0.441, 0.0543)
(0.442, 0.0548)
(0.443, 0.0552)
(0.444, 0.0556)
(0.445, 0.0561)
(0.446, 0.0565)
(0.447, 0.0570)
(0.448, 0.0575)
(0.449, 0.0579)
(0.450, 0.0584)
(0.451, 0.0588)
(0.452, 0.0593)
(0.453, 0.0598)
(0.454, 0.0602)
(0.455, 0.0607)
(0.456, 0.0612)
(0.457, 0.0617)
(0.458, 0.0621)
(0.459, 0.0626)
(0.460, 0.0631)
(0.461, 0.0636)
(0.462, 0.0641)
(0.463, 0.0646)
(0.464, 0.0651)
(0.465, 0.0656)
(0.466, 0.0661)
(0.467, 0.0666)
(0.468, 0.0671)
(0.469, 0.0676)
(0.470, 0.0681)
(0.471, 0.0687)
(0.472, 0.0692)
(0.473, 0.0697)
(0.474, 0.0702)
(0.475, 0.0707)
(0.476, 0.0713)
(0.477, 0.0718)
(0.478, 0.0724)
(0.479, 0.0729)
(0.480, 0.0734)
(0.481, 0.0740)
(0.482, 0.0745)
(0.483, 0.0751)
(0.484, 0.0756)
(0.485, 0.0762)
(0.486, 0.0768)
(0.487, 0.0773)
(0.488, 0.0779)
(0.489, 0.0785)
(0.490, 0.0790)
(0.491, 0.0796)
(0.492, 0.0802)
(0.493, 0.0808)
(0.494, 0.0813)
(0.495, 0.0819)
(0.496, 0.0825)
(0.497, 0.0831)
(0.498, 0.0837)
(0.499, 0.0843)
(0.500, 0.0849)
(0.501, 0.0855)
(0.502, 0.0861)
(0.503, 0.0867)
(0.504, 0.0874)
(0.505, 0.0880)
(0.506, 0.0886)
(0.507, 0.0892)
(0.508, 0.0898)
(0.509, 0.0905)
(0.510, 0.0911)
(0.511, 0.0917)
(0.512, 0.0924)
(0.513, 0.0930)
(0.514, 0.0937)
(0.515, 0.0943)
(0.516, 0.0950)
(0.517, 0.0956)
(0.518, 0.0963)
(0.519, 0.0970)
(0.520, 0.0976)
(0.521, 0.0983)
(0.522, 0.0990)
(0.523, 0.0996)
(0.524, 0.1003)
(0.525, 0.1010)
(0.526, 0.1017)
(0.527, 0.1024)
(0.528, 0.1031)
(0.529, 0.1038)
(0.530, 0.1045)
(0.531, 0.1052)
(0.532, 0.1059)
(0.533, 0.1066)
(0.534, 0.1073)
(0.535, 0.1080)
(0.536, 0.1087)
(0.537, 0.1095)
(0.538, 0.1102)
(0.539, 0.1109)
(0.540, 0.1117)
(0.541, 0.1124)
(0.542, 0.1131)
(0.543, 0.1139)
(0.544, 0.1146)
(0.545, 0.1154)
(0.546, 0.1161)
(0.547, 0.1169)
(0.548, 0.1177)
(0.549, 0.1184)
(0.550, 0.1192)
(0.551, 0.1200)
(0.552, 0.1207)
(0.553, 0.1215)
(0.554, 0.1223)
(0.555, 0.1231)
(0.556, 0.1239)
(0.557, 0.1247)
(0.558, 0.1255)
(0.559, 0.1263)
(0.560, 0.1271)
(0.561, 0.1279)
(0.562, 0.1287)
(0.563, 0.1295)
(0.564, 0.1303)
(0.565, 0.1312)
(0.566, 0.1320)
(0.567, 0.1328)
(0.568, 0.1337)
(0.569, 0.1345)
(0.570, 0.1353)
(0.571, 0.1362)
(0.572, 0.1370)
(0.573, 0.1379)
(0.574, 0.1388)
(0.575, 0.1396)
(0.576, 0.1405)
(0.577, 0.1413)
(0.578, 0.1422)
(0.579, 0.1431)
(0.580, 0.1440)
(0.581, 0.1449)
(0.582, 0.1458)
(0.583, 0.1466)
(0.584, 0.1475)
(0.585, 0.1484)
(0.586, 0.1494)
(0.587, 0.1503)
(0.588, 0.1512)
(0.589, 0.1521)
(0.590, 0.1530)
(0.591, 0.1539)
(0.592, 0.1549)
(0.593, 0.1558)
(0.594, 0.1567)
(0.595, 0.1577)
(0.596, 0.1586)
(0.597, 0.1596)
(0.598, 0.1605)
(0.599, 0.1615)
(0.600, 0.1624)
(0.601, 0.1634)
(0.602, 0.1644)
(0.603, 0.1653)
(0.604, 0.1663)
(0.605, 0.1673)
(0.606, 0.1683)
(0.607, 0.1693)
(0.608, 0.1703)
(0.609, 0.1713)
(0.610, 0.1723)
(0.611, 0.1733)
(0.612, 0.1743)
(0.613, 0.1753)
(0.614, 0.1763)
(0.615, 0.1774)
(0.616, 0.1784)
(0.617, 0.1794)
(0.618, 0.1805)
(0.619, 0.1815)
(0.620, 0.1825)
(0.621, 0.1836)
(0.622, 0.1846)
(0.623, 0.1857)
(0.624, 0.1868)
(0.625, 0.1878)
(0.626, 0.1889)
(0.627, 0.1900)
(0.628, 0.1911)
(0.629, 0.1921)
(0.630, 0.1932)
(0.631, 0.1943)
(0.632, 0.1954)
(0.633, 0.1965)
(0.634, 0.1976)
(0.635, 0.1987)
(0.636, 0.1999)
(0.637, 0.2010)
(0.638, 0.2021)
(0.639, 0.2032)
(0.640, 0.2044)
(0.641, 0.2055)
(0.642, 0.2066)
(0.643, 0.2078)
(0.644, 0.2089)
(0.645, 0.2101)
(0.646, 0.2113)
(0.647, 0.2124)
(0.648, 0.2136)
(0.649, 0.2148)
(0.650, 0.2160)
(0.651, 0.2171)
(0.652, 0.2183)
(0.653, 0.2195)
(0.654, 0.2207)
(0.655, 0.2219)
(0.656, 0.2231)
(0.657, 0.2243)
(0.658, 0.2256)
(0.659, 0.2268)
(0.660, 0.2280)
(0.661, 0.2292)
(0.662, 0.2305)
(0.663, 0.2317)
(0.664, 0.2330)
(0.665, 0.2342)
(0.666, 0.2355)
(0.667, 0.2367)
(0.668, 0.2380)
(0.669, 0.2393)
(0.670, 0.2405)
(0.671, 0.2418)
(0.672, 0.2431)
(0.673, 0.2444)
(0.674, 0.2457)
(0.675, 0.2470)
(0.676, 0.2483)
(0.677, 0.2496)
(0.678, 0.2509)
(0.679, 0.2522)
(0.680, 0.2536)
(0.681, 0.2549)
(0.682, 0.2562)
(0.683, 0.2576)
(0.684, 0.2589)
(0.685, 0.2603)
(0.686, 0.2616)
(0.687, 0.2630)
(0.688, 0.2643)
(0.689, 0.2657)
(0.690, 0.2671)
(0.691, 0.2685)
(0.692, 0.2698)
(0.693, 0.2712)
(0.694, 0.2726)
(0.695, 0.2740)
(0.696, 0.2754)
(0.697, 0.2768)
(0.698, 0.2783)
(0.699, 0.2797)
(0.700, 0.2811)
(0.701, 0.2825)
(0.702, 0.2840)
(0.703, 0.2854)
(0.704, 0.2869)
(0.705, 0.2883)
(0.706, 0.2898)
(0.707, 0.2912)
(0.708, 0.2927)
(0.709, 0.2942)
(0.710, 0.2957)
(0.711, 0.2971)
(0.712, 0.2986)
(0.713, 0.3001)
(0.714, 0.3016)
(0.715, 0.3031)
(0.716, 0.3046)
(0.717, 0.3062)
(0.718, 0.3077)
(0.719, 0.3092)
(0.720, 0.3107)
(0.721, 0.3123)
(0.722, 0.3138)
(0.723, 0.3154)
(0.724, 0.3169)
(0.725, 0.3185)
(0.726, 0.3201)
(0.727, 0.3216)
(0.728, 0.3232)
(0.729, 0.3248)
(0.730, 0.3264)
(0.731, 0.3280)
(0.732, 0.3296)
(0.733, 0.3312)
(0.734, 0.3328)
(0.735, 0.3344)
(0.736, 0.3360)
(0.737, 0.3376)
(0.738, 0.3393)
(0.739, 0.3409)
(0.740, 0.3426)
(0.741, 0.3442)
(0.742, 0.3459)
(0.743, 0.3475)
(0.744, 0.3492)
(0.745, 0.3509)
(0.746, 0.3526)
(0.747, 0.3542)
(0.748, 0.3559)
(0.749, 0.3576)
(0.750, 0.3593)
(0.751, 0.3610)
(0.752, 0.3627)
(0.753, 0.3645)
(0.754, 0.3662)
(0.755, 0.3679)
(0.756, 0.3697)
(0.757, 0.3714)
(0.758, 0.3731)
(0.759, 0.3749)
(0.760, 0.3767)
(0.761, 0.3784)
(0.762, 0.3802)
(0.763, 0.3820)
(0.764, 0.3838)
(0.765, 0.3856)
(0.766, 0.3873)
(0.767, 0.3891)
(0.768, 0.3910)
(0.769, 0.3928)
(0.770, 0.3946)
(0.771, 0.3964)
(0.772, 0.3983)
(0.773, 0.4001)
(0.774, 0.4019)
(0.775, 0.4038)
(0.776, 0.4056)
(0.777, 0.4075)
(0.778, 0.4094)
(0.779, 0.4112)
(0.780, 0.4131)
(0.781, 0.4150)
(0.782, 0.4169)
(0.783, 0.4188)
(0.784, 0.4207)
(0.785, 0.4226)
(0.786, 0.4245)
(0.787, 0.4265)
(0.788, 0.4284)
(0.789, 0.4303)
(0.790, 0.4323)
(0.791, 0.4342)
(0.792, 0.4362)
(0.793, 0.4382)
(0.794, 0.4401)
(0.795, 0.4421)
(0.796, 0.4441)
(0.797, 0.4461)
(0.798, 0.4481)
(0.799, 0.4501)
(0.800, 0.4521)
(0.801, 0.4541)
(0.802, 0.4561)
(0.803, 0.4581)
(0.804, 0.4602)
(0.805, 0.4622)
(0.806, 0.4643)
(0.807, 0.4663)
(0.808, 0.4684)
(0.809, 0.4704)
(0.810, 0.4725)
(0.811, 0.4746)
(0.812, 0.4767)
(0.813, 0.4788)
(0.814, 0.4809)
(0.815, 0.4830)
(0.816, 0.4851)
(0.817, 0.4872)
(0.818, 0.4893)
(0.819, 0.4914)
(0.820, 0.4936)
(0.821, 0.4957)
(0.822, 0.4979)
(0.823, 0.5000)
(0.824, 0.5022)
(0.825, 0.5044)
(0.826, 0.5066)
(0.827, 0.5087)
(0.828, 0.5109)
(0.829, 0.5131)
(0.830, 0.5153)
(0.831, 0.5175)
(0.832, 0.5198)
(0.833, 0.5220)
(0.834, 0.5242)
(0.835, 0.5265)
(0.836, 0.5287)
(0.837, 0.5310)
(0.838, 0.5332)
(0.839, 0.5355)
(0.840, 0.5378)
(0.841, 0.5400)
(0.842, 0.5423)
(0.843, 0.5446)
(0.844, 0.5469)
(0.845, 0.5492)
(0.846, 0.5516)
(0.847, 0.5539)
(0.848, 0.5562)
(0.849, 0.5586)
(0.850, 0.5609)
(0.851, 0.5632)
(0.852, 0.5656)
(0.853, 0.5680)
(0.854, 0.5703)
(0.855, 0.5727)
(0.856, 0.5751)
(0.857, 0.5775)
(0.858, 0.5799)
(0.859, 0.5823)
(0.860, 0.5847)
(0.861, 0.5872)
(0.862, 0.5896)
(0.863, 0.5920)
(0.864, 0.5945)
(0.865, 0.5969)
(0.866, 0.5994)
(0.867, 0.6018)
(0.868, 0.6043)
(0.869, 0.6068)
(0.870, 0.6093)
(0.871, 0.6118)
(0.872, 0.6143)
(0.873, 0.6168)
(0.874, 0.6193)
(0.875, 0.6218)
(0.876, 0.6244)
(0.877, 0.6269)
(0.878, 0.6294)
(0.879, 0.6320)
(0.880, 0.6346)
(0.881, 0.6371)
(0.882, 0.6397)
(0.883, 0.6423)
(0.884, 0.6449)
(0.885, 0.6475)
(0.886, 0.6501)
(0.887, 0.6527)
(0.888, 0.6553)
(0.889, 0.6580)
(0.890, 0.6606)
(0.891, 0.6632)
(0.892, 0.6659)
(0.893, 0.6686)
(0.894, 0.6712)
(0.895, 0.6739)
(0.896, 0.6766)
(0.897, 0.6793)
(0.898, 0.6820)
(0.899, 0.6847)
(0.900, 0.6874)
(0.901, 0.6901)
(0.902, 0.6928)
(0.903, 0.6956)
(0.904, 0.6983)
(0.905, 0.7011)
(0.906, 0.7038)
(0.907, 0.7066)
(0.908, 0.7094)
(0.909, 0.7122)
(0.910, 0.7149)
(0.911, 0.7177)
(0.912, 0.7206)
(0.913, 0.7234)
(0.914, 0.7262)
(0.915, 0.7290)
(0.916, 0.7319)
(0.917, 0.7347)
(0.918, 0.7376)
(0.919, 0.7404)
(0.920, 0.7433)
(0.921, 0.7462)
(0.922, 0.7491)
(0.923, 0.7520)
(0.924, 0.7549)
(0.925, 0.7578)
(0.926, 0.7607)
(0.927, 0.7636)
(0.928, 0.7665)
(0.929, 0.7695)
(0.930, 0.7724)
(0.931, 0.7754)
(0.932, 0.7784)
(0.933, 0.7813)
(0.934, 0.7843)
(0.935, 0.7873)
(0.936, 0.7903)
(0.937, 0.7933)
(0.938, 0.7963)
(0.939, 0.7994)
(0.940, 0.8024)
(0.941, 0.8054)
(0.942, 0.8085)
(0.943, 0.8116)
(0.944, 0.8146)
(0.945, 0.8177)
(0.946, 0.8208)
(0.947, 0.8239)
(0.948, 0.8270)
(0.949, 0.8301)
(0.950, 0.8332)
(0.951, 0.8363)
(0.952, 0.8394)
(0.953, 0.8426)
(0.954, 0.8457)
(0.955, 0.8489)
(0.956, 0.8521)
(0.957, 0.8552)
(0.958, 0.8584)
(0.959, 0.8616)
(0.960, 0.8648)
(0.961, 0.8680)
(0.962, 0.8712)
(0.963, 0.8745)
(0.964, 0.8777)
(0.965, 0.8809)
(0.966, 0.8842)
(0.967, 0.8875)
(0.968, 0.8907)
(0.969, 0.8940)
(0.970, 0.8973)
(0.971, 0.9006)
(0.972, 0.9039)
(0.973, 0.9072)
(0.974, 0.9105)
(0.975, 0.9139)
(0.976, 0.9172)
(0.977, 0.9205)
(0.978, 0.9239)
(0.979, 0.9273)
(0.980, 0.9306)
(0.981, 0.9340)
(0.982, 0.9374)
(0.983, 0.9408)
(0.984, 0.9442)
(0.985, 0.9476)
(0.986, 0.9511)
(0.987, 0.9545)
(0.988, 0.9580)
(0.989, 0.9614)
(0.990, 0.9649)
(0.991, 0.9683)
(0.992, 0.9718)
(0.993, 0.9753)
(0.994, 0.9788)
(0.995, 0.9823)
(0.996, 0.9858)
(0.997, 0.9894)
(0.998, 0.9929)
(0.999, 0.9964)
(1.000, 1.0000)
}
\defdata{emaprofile-exp3}{
(0.000, 0.0000)
(0.233, 0.0000)
(0.234, 0.0000)
(0.235, 0.0000)
(0.236, 0.0000)
(0.237, 0.0000)
(0.238, 0.0000)
(0.239, 0.0000)
(0.240, 0.0001)
(0.241, 0.0001)
(0.242, 0.0001)
(0.243, 0.0001)
(0.244, 0.0001)
(0.245, 0.0001)
(0.246, 0.0001)
(0.247, 0.0001)
(0.248, 0.0001)
(0.249, 0.0001)
(0.250, 0.0001)
(0.251, 0.0001)
(0.252, 0.0001)
(0.253, 0.0001)
(0.254, 0.0001)
(0.255, 0.0001)
(0.256, 0.0001)
(0.257, 0.0001)
(0.258, 0.0001)
(0.259, 0.0001)
(0.260, 0.0001)
(0.261, 0.0001)
(0.262, 0.0001)
(0.263, 0.0001)
(0.264, 0.0001)
(0.265, 0.0001)
(0.266, 0.0001)
(0.267, 0.0001)
(0.268, 0.0001)
(0.269, 0.0001)
(0.270, 0.0001)
(0.271, 0.0001)
(0.272, 0.0001)
(0.273, 0.0001)
(0.274, 0.0001)
(0.275, 0.0001)
(0.276, 0.0001)
(0.277, 0.0001)
(0.278, 0.0001)
(0.279, 0.0001)
(0.280, 0.0001)
(0.281, 0.0001)
(0.282, 0.0002)
(0.283, 0.0002)
(0.284, 0.0002)
(0.285, 0.0002)
(0.286, 0.0002)
(0.287, 0.0002)
(0.288, 0.0002)
(0.289, 0.0002)
(0.290, 0.0002)
(0.291, 0.0002)
(0.292, 0.0002)
(0.293, 0.0002)
(0.294, 0.0002)
(0.295, 0.0002)
(0.296, 0.0002)
(0.297, 0.0002)
(0.298, 0.0002)
(0.299, 0.0002)
(0.300, 0.0002)
(0.301, 0.0002)
(0.302, 0.0002)
(0.303, 0.0003)
(0.304, 0.0003)
(0.305, 0.0003)
(0.306, 0.0003)
(0.307, 0.0003)
(0.308, 0.0003)
(0.309, 0.0003)
(0.310, 0.0003)
(0.311, 0.0003)
(0.312, 0.0003)
(0.313, 0.0003)
(0.314, 0.0003)
(0.315, 0.0003)
(0.316, 0.0003)
(0.317, 0.0003)
(0.318, 0.0004)
(0.319, 0.0004)
(0.320, 0.0004)
(0.321, 0.0004)
(0.322, 0.0004)
(0.323, 0.0004)
(0.324, 0.0004)
(0.325, 0.0004)
(0.326, 0.0004)
(0.327, 0.0004)
(0.328, 0.0004)
(0.329, 0.0004)
(0.330, 0.0005)
(0.331, 0.0005)
(0.332, 0.0005)
(0.333, 0.0005)
(0.334, 0.0005)
(0.335, 0.0005)
(0.336, 0.0005)
(0.337, 0.0005)
(0.338, 0.0005)
(0.339, 0.0006)
(0.340, 0.0006)
(0.341, 0.0006)
(0.342, 0.0006)
(0.343, 0.0006)
(0.344, 0.0006)
(0.345, 0.0006)
(0.346, 0.0006)
(0.347, 0.0006)
(0.348, 0.0007)
(0.349, 0.0007)
(0.350, 0.0007)
(0.351, 0.0007)
(0.352, 0.0007)
(0.353, 0.0007)
(0.354, 0.0007)
(0.355, 0.0008)
(0.356, 0.0008)
(0.357, 0.0008)
(0.358, 0.0008)
(0.359, 0.0008)
(0.360, 0.0008)
(0.361, 0.0009)
(0.362, 0.0009)
(0.363, 0.0009)
(0.364, 0.0009)
(0.365, 0.0009)
(0.366, 0.0009)
(0.367, 0.0010)
(0.368, 0.0010)
(0.369, 0.0010)
(0.370, 0.0010)
(0.371, 0.0010)
(0.372, 0.0010)
(0.373, 0.0011)
(0.374, 0.0011)
(0.375, 0.0011)
(0.376, 0.0011)
(0.377, 0.0012)
(0.378, 0.0012)
(0.379, 0.0012)
(0.380, 0.0012)
(0.381, 0.0012)
(0.382, 0.0013)
(0.383, 0.0013)
(0.384, 0.0013)
(0.385, 0.0013)
(0.386, 0.0014)
(0.387, 0.0014)
(0.388, 0.0014)
(0.389, 0.0014)
(0.390, 0.0015)
(0.391, 0.0015)
(0.392, 0.0015)
(0.393, 0.0015)
(0.394, 0.0016)
(0.395, 0.0016)
(0.396, 0.0016)
(0.397, 0.0016)
(0.398, 0.0017)
(0.399, 0.0017)
(0.400, 0.0017)
(0.401, 0.0018)
(0.402, 0.0018)
(0.403, 0.0018)
(0.404, 0.0019)
(0.405, 0.0019)
(0.406, 0.0019)
(0.407, 0.0020)
(0.408, 0.0020)
(0.409, 0.0020)
(0.410, 0.0021)
(0.411, 0.0021)
(0.412, 0.0021)
(0.413, 0.0022)
(0.414, 0.0022)
(0.415, 0.0022)
(0.416, 0.0023)
(0.417, 0.0023)
(0.418, 0.0024)
(0.419, 0.0024)
(0.420, 0.0024)
(0.421, 0.0025)
(0.422, 0.0025)
(0.423, 0.0026)
(0.424, 0.0026)
(0.425, 0.0026)
(0.426, 0.0027)
(0.427, 0.0027)
(0.428, 0.0028)
(0.429, 0.0028)
(0.430, 0.0029)
(0.431, 0.0029)
(0.432, 0.0030)
(0.433, 0.0030)
(0.434, 0.0031)
(0.435, 0.0031)
(0.436, 0.0032)
(0.437, 0.0032)
(0.438, 0.0033)
(0.439, 0.0033)
(0.440, 0.0034)
(0.441, 0.0034)
(0.442, 0.0035)
(0.443, 0.0035)
(0.444, 0.0036)
(0.445, 0.0036)
(0.446, 0.0037)
(0.447, 0.0038)
(0.448, 0.0038)
(0.449, 0.0039)
(0.450, 0.0039)
(0.451, 0.0040)
(0.452, 0.0041)
(0.453, 0.0041)
(0.454, 0.0042)
(0.455, 0.0042)
(0.456, 0.0043)
(0.457, 0.0044)
(0.458, 0.0044)
(0.459, 0.0045)
(0.460, 0.0046)
(0.461, 0.0046)
(0.462, 0.0047)
(0.463, 0.0048)
(0.464, 0.0049)
(0.465, 0.0049)
(0.466, 0.0050)
(0.467, 0.0051)
(0.468, 0.0052)
(0.469, 0.0052)
(0.470, 0.0053)
(0.471, 0.0054)
(0.472, 0.0055)
(0.473, 0.0056)
(0.474, 0.0056)
(0.475, 0.0057)
(0.476, 0.0058)
(0.477, 0.0059)
(0.478, 0.0060)
(0.479, 0.0061)
(0.480, 0.0061)
(0.481, 0.0062)
(0.482, 0.0063)
(0.483, 0.0064)
(0.484, 0.0065)
(0.485, 0.0066)
(0.486, 0.0067)
(0.487, 0.0068)
(0.488, 0.0069)
(0.489, 0.0070)
(0.490, 0.0071)
(0.491, 0.0072)
(0.492, 0.0073)
(0.493, 0.0074)
(0.494, 0.0075)
(0.495, 0.0076)
(0.496, 0.0077)
(0.497, 0.0078)
(0.498, 0.0079)
(0.499, 0.0080)
(0.500, 0.0082)
(0.501, 0.0083)
(0.502, 0.0084)
(0.503, 0.0085)
(0.504, 0.0086)
(0.505, 0.0087)
(0.506, 0.0089)
(0.507, 0.0090)
(0.508, 0.0091)
(0.509, 0.0092)
(0.510, 0.0094)
(0.511, 0.0095)
(0.512, 0.0096)
(0.513, 0.0098)
(0.514, 0.0099)
(0.515, 0.0100)
(0.516, 0.0102)
(0.517, 0.0103)
(0.518, 0.0104)
(0.519, 0.0106)
(0.520, 0.0107)
(0.521, 0.0109)
(0.522, 0.0110)
(0.523, 0.0111)
(0.524, 0.0113)
(0.525, 0.0114)
(0.526, 0.0116)
(0.527, 0.0118)
(0.528, 0.0119)
(0.529, 0.0121)
(0.530, 0.0122)
(0.531, 0.0124)
(0.532, 0.0125)
(0.533, 0.0127)
(0.534, 0.0129)
(0.535, 0.0130)
(0.536, 0.0132)
(0.537, 0.0134)
(0.538, 0.0136)
(0.539, 0.0137)
(0.540, 0.0139)
(0.541, 0.0141)
(0.542, 0.0143)
(0.543, 0.0145)
(0.544, 0.0146)
(0.545, 0.0148)
(0.546, 0.0150)
(0.547, 0.0152)
(0.548, 0.0154)
(0.549, 0.0156)
(0.550, 0.0158)
(0.551, 0.0160)
(0.552, 0.0162)
(0.553, 0.0164)
(0.554, 0.0166)
(0.555, 0.0168)
(0.556, 0.0170)
(0.557, 0.0173)
(0.558, 0.0175)
(0.559, 0.0177)
(0.560, 0.0179)
(0.561, 0.0181)
(0.562, 0.0184)
(0.563, 0.0186)
(0.564, 0.0188)
(0.565, 0.0191)
(0.566, 0.0193)
(0.567, 0.0195)
(0.568, 0.0198)
(0.569, 0.0200)
(0.570, 0.0203)
(0.571, 0.0205)
(0.572, 0.0207)
(0.573, 0.0210)
(0.574, 0.0213)
(0.575, 0.0215)
(0.576, 0.0218)
(0.577, 0.0220)
(0.578, 0.0223)
(0.579, 0.0226)
(0.580, 0.0228)
(0.581, 0.0231)
(0.582, 0.0234)
(0.583, 0.0237)
(0.584, 0.0240)
(0.585, 0.0243)
(0.586, 0.0245)
(0.587, 0.0248)
(0.588, 0.0251)
(0.589, 0.0254)
(0.590, 0.0257)
(0.591, 0.0260)
(0.592, 0.0263)
(0.593, 0.0266)
(0.594, 0.0270)
(0.595, 0.0273)
(0.596, 0.0276)
(0.597, 0.0279)
(0.598, 0.0282)
(0.599, 0.0286)
(0.600, 0.0289)
(0.601, 0.0292)
(0.602, 0.0296)
(0.603, 0.0299)
(0.604, 0.0303)
(0.605, 0.0306)
(0.606, 0.0310)
(0.607, 0.0313)
(0.608, 0.0317)
(0.609, 0.0321)
(0.610, 0.0324)
(0.611, 0.0328)
(0.612, 0.0332)
(0.613, 0.0335)
(0.614, 0.0339)
(0.615, 0.0343)
(0.616, 0.0347)
(0.617, 0.0351)
(0.618, 0.0355)
(0.619, 0.0359)
(0.620, 0.0363)
(0.621, 0.0367)
(0.622, 0.0371)
(0.623, 0.0375)
(0.624, 0.0379)
(0.625, 0.0384)
(0.626, 0.0388)
(0.627, 0.0392)
(0.628, 0.0397)
(0.629, 0.0401)
(0.630, 0.0405)
(0.631, 0.0410)
(0.632, 0.0415)
(0.633, 0.0419)
(0.634, 0.0424)
(0.635, 0.0428)
(0.636, 0.0433)
(0.637, 0.0438)
(0.638, 0.0443)
(0.639, 0.0447)
(0.640, 0.0452)
(0.641, 0.0457)
(0.642, 0.0462)
(0.643, 0.0467)
(0.644, 0.0472)
(0.645, 0.0477)
(0.646, 0.0483)
(0.647, 0.0488)
(0.648, 0.0493)
(0.649, 0.0498)
(0.650, 0.0504)
(0.651, 0.0509)
(0.652, 0.0515)
(0.653, 0.0520)
(0.654, 0.0526)
(0.655, 0.0531)
(0.656, 0.0537)
(0.657, 0.0543)
(0.658, 0.0548)
(0.659, 0.0554)
(0.660, 0.0560)
(0.661, 0.0566)
(0.662, 0.0572)
(0.663, 0.0578)
(0.664, 0.0584)
(0.665, 0.0590)
(0.666, 0.0596)
(0.667, 0.0602)
(0.668, 0.0609)
(0.669, 0.0615)
(0.670, 0.0622)
(0.671, 0.0628)
(0.672, 0.0634)
(0.673, 0.0641)
(0.674, 0.0648)
(0.675, 0.0654)
(0.676, 0.0661)
(0.677, 0.0668)
(0.678, 0.0675)
(0.679, 0.0682)
(0.680, 0.0689)
(0.681, 0.0696)
(0.682, 0.0703)
(0.683, 0.0710)
(0.684, 0.0717)
(0.685, 0.0725)
(0.686, 0.0732)
(0.687, 0.0739)
(0.688, 0.0747)
(0.689, 0.0755)
(0.690, 0.0762)
(0.691, 0.0770)
(0.692, 0.0778)
(0.693, 0.0785)
(0.694, 0.0793)
(0.695, 0.0801)
(0.696, 0.0809)
(0.697, 0.0817)
(0.698, 0.0826)
(0.699, 0.0834)
(0.700, 0.0842)
(0.701, 0.0851)
(0.702, 0.0859)
(0.703, 0.0868)
(0.704, 0.0876)
(0.705, 0.0885)
(0.706, 0.0894)
(0.707, 0.0902)
(0.708, 0.0911)
(0.709, 0.0920)
(0.710, 0.0929)
(0.711, 0.0938)
(0.712, 0.0948)
(0.713, 0.0957)
(0.714, 0.0966)
(0.715, 0.0976)
(0.716, 0.0985)
(0.717, 0.0995)
(0.718, 0.1004)
(0.719, 0.1014)
(0.720, 0.1024)
(0.721, 0.1034)
(0.722, 0.1044)
(0.723, 0.1054)
(0.724, 0.1064)
(0.725, 0.1074)
(0.726, 0.1085)
(0.727, 0.1095)
(0.728, 0.1106)
(0.729, 0.1116)
(0.730, 0.1127)
(0.731, 0.1138)
(0.732, 0.1148)
(0.733, 0.1159)
(0.734, 0.1170)
(0.735, 0.1181)
(0.736, 0.1193)
(0.737, 0.1204)
(0.738, 0.1215)
(0.739, 0.1227)
(0.740, 0.1238)
(0.741, 0.1250)
(0.742, 0.1262)
(0.743, 0.1274)
(0.744, 0.1286)
(0.745, 0.1298)
(0.746, 0.1310)
(0.747, 0.1322)
(0.748, 0.1334)
(0.749, 0.1347)
(0.750, 0.1359)
(0.751, 0.1372)
(0.752, 0.1385)
(0.753, 0.1397)
(0.754, 0.1410)
(0.755, 0.1423)
(0.756, 0.1436)
(0.757, 0.1450)
(0.758, 0.1463)
(0.759, 0.1476)
(0.760, 0.1490)
(0.761, 0.1504)
(0.762, 0.1517)
(0.763, 0.1531)
(0.764, 0.1545)
(0.765, 0.1559)
(0.766, 0.1574)
(0.767, 0.1588)
(0.768, 0.1602)
(0.769, 0.1617)
(0.770, 0.1631)
(0.771, 0.1646)
(0.772, 0.1661)
(0.773, 0.1676)
(0.774, 0.1691)
(0.775, 0.1706)
(0.776, 0.1722)
(0.777, 0.1737)
(0.778, 0.1753)
(0.779, 0.1768)
(0.780, 0.1784)
(0.781, 0.1800)
(0.782, 0.1816)
(0.783, 0.1832)
(0.784, 0.1849)
(0.785, 0.1865)
(0.786, 0.1882)
(0.787, 0.1898)
(0.788, 0.1915)
(0.789, 0.1932)
(0.790, 0.1949)
(0.791, 0.1966)
(0.792, 0.1984)
(0.793, 0.2001)
(0.794, 0.2019)
(0.795, 0.2036)
(0.796, 0.2054)
(0.797, 0.2072)
(0.798, 0.2090)
(0.799, 0.2108)
(0.800, 0.2127)
(0.801, 0.2145)
(0.802, 0.2164)
(0.803, 0.2183)
(0.804, 0.2202)
(0.805, 0.2221)
(0.806, 0.2240)
(0.807, 0.2259)
(0.808, 0.2279)
(0.809, 0.2298)
(0.810, 0.2318)
(0.811, 0.2338)
(0.812, 0.2358)
(0.813, 0.2378)
(0.814, 0.2399)
(0.815, 0.2419)
(0.816, 0.2440)
(0.817, 0.2461)
(0.818, 0.2482)
(0.819, 0.2503)
(0.820, 0.2524)
(0.821, 0.2546)
(0.822, 0.2567)
(0.823, 0.2589)
(0.824, 0.2611)
(0.825, 0.2633)
(0.826, 0.2655)
(0.827, 0.2677)
(0.828, 0.2700)
(0.829, 0.2723)
(0.830, 0.2746)
(0.831, 0.2769)
(0.832, 0.2792)
(0.833, 0.2815)
(0.834, 0.2839)
(0.835, 0.2862)
(0.836, 0.2886)
(0.837, 0.2910)
(0.838, 0.2934)
(0.839, 0.2959)
(0.840, 0.2983)
(0.841, 0.3008)
(0.842, 0.3033)
(0.843, 0.3058)
(0.844, 0.3083)
(0.845, 0.3109)
(0.846, 0.3134)
(0.847, 0.3160)
(0.848, 0.3186)
(0.849, 0.3212)
(0.850, 0.3239)
(0.851, 0.3265)
(0.852, 0.3292)
(0.853, 0.3319)
(0.854, 0.3346)
(0.855, 0.3373)
(0.856, 0.3401)
(0.857, 0.3428)
(0.858, 0.3456)
(0.859, 0.3484)
(0.860, 0.3512)
(0.861, 0.3541)
(0.862, 0.3569)
(0.863, 0.3598)
(0.864, 0.3627)
(0.865, 0.3657)
(0.866, 0.3686)
(0.867, 0.3716)
(0.868, 0.3745)
(0.869, 0.3775)
(0.870, 0.3806)
(0.871, 0.3836)
(0.872, 0.3867)
(0.873, 0.3898)
(0.874, 0.3929)
(0.875, 0.3960)
(0.876, 0.3992)
(0.877, 0.4023)
(0.878, 0.4055)
(0.879, 0.4087)
(0.880, 0.4120)
(0.881, 0.4152)
(0.882, 0.4185)
(0.883, 0.4218)
(0.884, 0.4251)
(0.885, 0.4285)
(0.886, 0.4319)
(0.887, 0.4352)
(0.888, 0.4387)
(0.889, 0.4421)
(0.890, 0.4456)
(0.891, 0.4490)
(0.892, 0.4526)
(0.893, 0.4561)
(0.894, 0.4596)
(0.895, 0.4632)
(0.896, 0.4668)
(0.897, 0.4704)
(0.898, 0.4741)
(0.899, 0.4778)
(0.900, 0.4815)
(0.901, 0.4852)
(0.902, 0.4889)
(0.903, 0.4927)
(0.904, 0.4965)
(0.905, 0.5003)
(0.906, 0.5042)
(0.907, 0.5081)
(0.908, 0.5120)
(0.909, 0.5159)
(0.910, 0.5198)
(0.911, 0.5238)
(0.912, 0.5278)
(0.913, 0.5318)
(0.914, 0.5359)
(0.915, 0.5400)
(0.916, 0.5441)
(0.917, 0.5482)
(0.918, 0.5524)
(0.919, 0.5566)
(0.920, 0.5608)
(0.921, 0.5650)
(0.922, 0.5693)
(0.923, 0.5736)
(0.924, 0.5779)
(0.925, 0.5823)
(0.926, 0.5866)
(0.927, 0.5911)
(0.928, 0.5955)
(0.929, 0.6000)
(0.930, 0.6044)
(0.931, 0.6090)
(0.932, 0.6135)
(0.933, 0.6181)
(0.934, 0.6227)
(0.935, 0.6274)
(0.936, 0.6320)
(0.937, 0.6367)
(0.938, 0.6415)
(0.939, 0.6462)
(0.940, 0.6510)
(0.941, 0.6558)
(0.942, 0.6607)
(0.943, 0.6656)
(0.944, 0.6705)
(0.945, 0.6754)
(0.946, 0.6804)
(0.947, 0.6854)
(0.948, 0.6904)
(0.949, 0.6955)
(0.950, 0.7006)
(0.951, 0.7057)
(0.952, 0.7109)
(0.953, 0.7161)
(0.954, 0.7213)
(0.955, 0.7266)
(0.956, 0.7319)
(0.957, 0.7372)
(0.958, 0.7426)
(0.959, 0.7479)
(0.960, 0.7534)
(0.961, 0.7588)
(0.962, 0.7643)
(0.963, 0.7699)
(0.964, 0.7754)
(0.965, 0.7810)
(0.966, 0.7867)
(0.967, 0.7923)
(0.968, 0.7980)
(0.969, 0.8038)
(0.970, 0.8095)
(0.971, 0.8153)
(0.972, 0.8212)
(0.973, 0.8271)
(0.974, 0.8330)
(0.975, 0.8389)
(0.976, 0.8449)
(0.977, 0.8509)
(0.978, 0.8570)
(0.979, 0.8631)
(0.980, 0.8692)
(0.981, 0.8754)
(0.982, 0.8816)
(0.983, 0.8879)
(0.984, 0.8941)
(0.985, 0.9005)
(0.986, 0.9068)
(0.987, 0.9132)
(0.988, 0.9197)
(0.989, 0.9261)
(0.990, 0.9327)
(0.991, 0.9392)
(0.992, 0.9458)
(0.993, 0.9524)
(0.994, 0.9591)
(0.995, 0.9658)
(0.996, 0.9726)
(0.997, 0.9794)
(0.998, 0.9862)
(0.999, 0.9931)
(1.000, 1.0000)
}
\defdata{emaprofile-exp4}{
(0.000, 0.0000)
(0.551, 0.0000)
(0.552, 0.0000)
(0.553, 0.0000)
(0.554, 0.0000)
(0.555, 0.0000)
(0.556, 0.0000)
(0.557, 0.0000)
(0.558, 0.0001)
(0.559, 0.0001)
(0.560, 0.0001)
(0.561, 0.0001)
(0.562, 0.0001)
(0.563, 0.0001)
(0.564, 0.0001)
(0.565, 0.0001)
(0.566, 0.0001)
(0.567, 0.0001)
(0.568, 0.0001)
(0.569, 0.0001)
(0.570, 0.0001)
(0.571, 0.0001)
(0.572, 0.0001)
(0.573, 0.0001)
(0.574, 0.0001)
(0.575, 0.0001)
(0.576, 0.0001)
(0.577, 0.0001)
(0.578, 0.0001)
(0.579, 0.0001)
(0.580, 0.0001)
(0.581, 0.0001)
(0.582, 0.0001)
(0.583, 0.0001)
(0.584, 0.0001)
(0.585, 0.0001)
(0.586, 0.0001)
(0.587, 0.0001)
(0.588, 0.0001)
(0.589, 0.0001)
(0.590, 0.0001)
(0.591, 0.0001)
(0.592, 0.0001)
(0.593, 0.0001)
(0.594, 0.0001)
(0.595, 0.0001)
(0.596, 0.0002)
(0.597, 0.0002)
(0.598, 0.0002)
(0.599, 0.0002)
(0.600, 0.0002)
(0.601, 0.0002)
(0.602, 0.0002)
(0.603, 0.0002)
(0.604, 0.0002)
(0.605, 0.0002)
(0.606, 0.0002)
(0.607, 0.0002)
(0.608, 0.0002)
(0.609, 0.0002)
(0.610, 0.0002)
(0.611, 0.0002)
(0.612, 0.0002)
(0.613, 0.0002)
(0.614, 0.0003)
(0.615, 0.0003)
(0.616, 0.0003)
(0.617, 0.0003)
(0.618, 0.0003)
(0.619, 0.0003)
(0.620, 0.0003)
(0.621, 0.0003)
(0.622, 0.0003)
(0.623, 0.0003)
(0.624, 0.0003)
(0.625, 0.0003)
(0.626, 0.0004)
(0.627, 0.0004)
(0.628, 0.0004)
(0.629, 0.0004)
(0.630, 0.0004)
(0.631, 0.0004)
(0.632, 0.0004)
(0.633, 0.0004)
(0.634, 0.0004)
(0.635, 0.0004)
(0.636, 0.0005)
(0.637, 0.0005)
(0.638, 0.0005)
(0.639, 0.0005)
(0.640, 0.0005)
(0.641, 0.0005)
(0.642, 0.0005)
(0.643, 0.0006)
(0.644, 0.0006)
(0.645, 0.0006)
(0.646, 0.0006)
(0.647, 0.0006)
(0.648, 0.0006)
(0.649, 0.0007)
(0.650, 0.0007)
(0.651, 0.0007)
(0.652, 0.0007)
(0.653, 0.0007)
(0.654, 0.0007)
(0.655, 0.0008)
(0.656, 0.0008)
(0.657, 0.0008)
(0.658, 0.0008)
(0.659, 0.0008)
(0.660, 0.0009)
(0.661, 0.0009)
(0.662, 0.0009)
(0.663, 0.0009)
(0.664, 0.0010)
(0.665, 0.0010)
(0.666, 0.0010)
(0.667, 0.0010)
(0.668, 0.0011)
(0.669, 0.0011)
(0.670, 0.0011)
(0.671, 0.0011)
(0.672, 0.0012)
(0.673, 0.0012)
(0.674, 0.0012)
(0.675, 0.0013)
(0.676, 0.0013)
(0.677, 0.0013)
(0.678, 0.0014)
(0.679, 0.0014)
(0.680, 0.0014)
(0.681, 0.0015)
(0.682, 0.0015)
(0.683, 0.0015)
(0.684, 0.0016)
(0.685, 0.0016)
(0.686, 0.0017)
(0.687, 0.0017)
(0.688, 0.0018)
(0.689, 0.0018)
(0.690, 0.0018)
(0.691, 0.0019)
(0.692, 0.0019)
(0.693, 0.0020)
(0.694, 0.0020)
(0.695, 0.0021)
(0.696, 0.0021)
(0.697, 0.0022)
(0.698, 0.0022)
(0.699, 0.0023)
(0.700, 0.0023)
(0.701, 0.0024)
(0.702, 0.0025)
(0.703, 0.0025)
(0.704, 0.0026)
(0.705, 0.0027)
(0.706, 0.0027)
(0.707, 0.0028)
(0.708, 0.0028)
(0.709, 0.0029)
(0.710, 0.0030)
(0.711, 0.0031)
(0.712, 0.0031)
(0.713, 0.0032)
(0.714, 0.0033)
(0.715, 0.0034)
(0.716, 0.0034)
(0.717, 0.0035)
(0.718, 0.0036)
(0.719, 0.0037)
(0.720, 0.0038)
(0.721, 0.0039)
(0.722, 0.0040)
(0.723, 0.0041)
(0.724, 0.0042)
(0.725, 0.0043)
(0.726, 0.0044)
(0.727, 0.0045)
(0.728, 0.0046)
(0.729, 0.0047)
(0.730, 0.0048)
(0.731, 0.0049)
(0.732, 0.0050)
(0.733, 0.0051)
(0.734, 0.0053)
(0.735, 0.0054)
(0.736, 0.0055)
(0.737, 0.0056)
(0.738, 0.0058)
(0.739, 0.0059)
(0.740, 0.0060)
(0.741, 0.0062)
(0.742, 0.0063)
(0.743, 0.0065)
(0.744, 0.0066)
(0.745, 0.0068)
(0.746, 0.0069)
(0.747, 0.0071)
(0.748, 0.0072)
(0.749, 0.0074)
(0.750, 0.0076)
(0.751, 0.0078)
(0.752, 0.0079)
(0.753, 0.0081)
(0.754, 0.0083)
(0.755, 0.0085)
(0.756, 0.0087)
(0.757, 0.0089)
(0.758, 0.0091)
(0.759, 0.0093)
(0.760, 0.0095)
(0.761, 0.0097)
(0.762, 0.0099)
(0.763, 0.0101)
(0.764, 0.0104)
(0.765, 0.0106)
(0.766, 0.0108)
(0.767, 0.0111)
(0.768, 0.0113)
(0.769, 0.0116)
(0.770, 0.0118)
(0.771, 0.0121)
(0.772, 0.0124)
(0.773, 0.0127)
(0.774, 0.0129)
(0.775, 0.0132)
(0.776, 0.0135)
(0.777, 0.0138)
(0.778, 0.0141)
(0.779, 0.0144)
(0.780, 0.0147)
(0.781, 0.0151)
(0.782, 0.0154)
(0.783, 0.0157)
(0.784, 0.0161)
(0.785, 0.0164)
(0.786, 0.0168)
(0.787, 0.0172)
(0.788, 0.0175)
(0.789, 0.0179)
(0.790, 0.0183)
(0.791, 0.0187)
(0.792, 0.0191)
(0.793, 0.0195)
(0.794, 0.0199)
(0.795, 0.0204)
(0.796, 0.0208)
(0.797, 0.0213)
(0.798, 0.0217)
(0.799, 0.0222)
(0.800, 0.0227)
(0.801, 0.0231)
(0.802, 0.0236)
(0.803, 0.0241)
(0.804, 0.0247)
(0.805, 0.0252)
(0.806, 0.0257)
(0.807, 0.0263)
(0.808, 0.0268)
(0.809, 0.0274)
(0.810, 0.0280)
(0.811, 0.0286)
(0.812, 0.0292)
(0.813, 0.0298)
(0.814, 0.0304)
(0.815, 0.0311)
(0.816, 0.0317)
(0.817, 0.0324)
(0.818, 0.0331)
(0.819, 0.0337)
(0.820, 0.0345)
(0.821, 0.0352)
(0.822, 0.0359)
(0.823, 0.0367)
(0.824, 0.0374)
(0.825, 0.0382)
(0.826, 0.0390)
(0.827, 0.0398)
(0.828, 0.0406)
(0.829, 0.0415)
(0.830, 0.0423)
(0.831, 0.0432)
(0.832, 0.0441)
(0.833, 0.0450)
(0.834, 0.0459)
(0.835, 0.0469)
(0.836, 0.0478)
(0.837, 0.0488)
(0.838, 0.0498)
(0.839, 0.0508)
(0.840, 0.0519)
(0.841, 0.0529)
(0.842, 0.0540)
(0.843, 0.0551)
(0.844, 0.0562)
(0.845, 0.0574)
(0.846, 0.0585)
(0.847, 0.0597)
(0.848, 0.0609)
(0.849, 0.0621)
(0.850, 0.0634)
(0.851, 0.0647)
(0.852, 0.0660)
(0.853, 0.0673)
(0.854, 0.0687)
(0.855, 0.0700)
(0.856, 0.0714)
(0.857, 0.0729)
(0.858, 0.0743)
(0.859, 0.0758)
(0.860, 0.0773)
(0.861, 0.0789)
(0.862, 0.0804)
(0.863, 0.0820)
(0.864, 0.0837)
(0.865, 0.0853)
(0.866, 0.0870)
(0.867, 0.0887)
(0.868, 0.0905)
(0.869, 0.0923)
(0.870, 0.0941)
(0.871, 0.0959)
(0.872, 0.0978)
(0.873, 0.0997)
(0.874, 0.1017)
(0.875, 0.1037)
(0.876, 0.1057)
(0.877, 0.1078)
(0.878, 0.1099)
(0.879, 0.1120)
(0.880, 0.1142)
(0.881, 0.1164)
(0.882, 0.1187)
(0.883, 0.1210)
(0.884, 0.1234)
(0.885, 0.1258)
(0.886, 0.1282)
(0.887, 0.1307)
(0.888, 0.1332)
(0.889, 0.1358)
(0.890, 0.1384)
(0.891, 0.1410)
(0.892, 0.1437)
(0.893, 0.1465)
(0.894, 0.1493)
(0.895, 0.1522)
(0.896, 0.1551)
(0.897, 0.1580)
(0.898, 0.1611)
(0.899, 0.1641)
(0.900, 0.1673)
(0.901, 0.1704)
(0.902, 0.1737)
(0.903, 0.1770)
(0.904, 0.1803)
(0.905, 0.1838)
(0.906, 0.1872)
(0.907, 0.1908)
(0.908, 0.1944)
(0.909, 0.1980)
(0.910, 0.2018)
(0.911, 0.2056)
(0.912, 0.2094)
(0.913, 0.2134)
(0.914, 0.2174)
(0.915, 0.2214)
(0.916, 0.2256)
(0.917, 0.2298)
(0.918, 0.2341)
(0.919, 0.2384)
(0.920, 0.2429)
(0.921, 0.2474)
(0.922, 0.2520)
(0.923, 0.2567)
(0.924, 0.2614)
(0.925, 0.2663)
(0.926, 0.2712)
(0.927, 0.2762)
(0.928, 0.2813)
(0.929, 0.2865)
(0.930, 0.2918)
(0.931, 0.2972)
(0.932, 0.3026)
(0.933, 0.3082)
(0.934, 0.3138)
(0.935, 0.3196)
(0.936, 0.3255)
(0.937, 0.3314)
(0.938, 0.3375)
(0.939, 0.3436)
(0.940, 0.3499)
(0.941, 0.3563)
(0.942, 0.3627)
(0.943, 0.3693)
(0.944, 0.3760)
(0.945, 0.3828)
(0.946, 0.3898)
(0.947, 0.3968)
(0.948, 0.4040)
(0.949, 0.4113)
(0.950, 0.4187)
(0.951, 0.4263)
(0.952, 0.4339)
(0.953, 0.4417)
(0.954, 0.4497)
(0.955, 0.4577)
(0.956, 0.4659)
(0.957, 0.4743)
(0.958, 0.4828)
(0.959, 0.4914)
(0.960, 0.5002)
(0.961, 0.5091)
(0.962, 0.5181)
(0.963, 0.5274)
(0.964, 0.5367)
(0.965, 0.5463)
(0.966, 0.5559)
(0.967, 0.5658)
(0.968, 0.5758)
(0.969, 0.5860)
(0.970, 0.5963)
(0.971, 0.6069)
(0.972, 0.6175)
(0.973, 0.6284)
(0.974, 0.6395)
(0.975, 0.6507)
(0.976, 0.6621)
(0.977, 0.6737)
(0.978, 0.6855)
(0.979, 0.6975)
(0.980, 0.7097)
(0.981, 0.7221)
(0.982, 0.7347)
(0.983, 0.7475)
(0.984, 0.7605)
(0.985, 0.7737)
(0.986, 0.7872)
(0.987, 0.8008)
(0.988, 0.8147)
(0.989, 0.8288)
(0.990, 0.8432)
(0.991, 0.8578)
(0.992, 0.8726)
(0.993, 0.8876)
(0.994, 0.9029)
(0.995, 0.9184)
(0.996, 0.9342)
(0.997, 0.9503)
(0.998, 0.9666)
(0.999, 0.9832)
(1.000, 1.0000)
}
\defdata{emaprofile-1-short}{
(0.000, 0.0000)
(0.065, 0.0000)
(0.066, 0.0001)
(0.067, 0.0001)
(0.068, 0.0001)
(0.069, 0.0001)
(0.070, 0.0002)
(0.071, 0.0002)
(0.072, 0.0002)
(0.073, 0.0003)
(0.074, 0.0004)
(0.075, 0.0005)
(0.076, 0.0006)
(0.077, 0.0008)
(0.078, 0.0010)
(0.079, 0.0012)
(0.080, 0.0015)
(0.081, 0.0018)
(0.082, 0.0023)
(0.083, 0.0028)
(0.084, 0.0034)
(0.085, 0.0042)
(0.086, 0.0051)
(0.087, 0.0062)
(0.088, 0.0075)
(0.089, 0.0091)
(0.090, 0.0110)
(0.091, 0.0132)
(0.092, 0.0159)
(0.093, 0.0191)
(0.094, 0.0230)
(0.095, 0.0275)
(0.096, 0.0328)
(0.097, 0.0391)
(0.098, 0.0466)
(0.099, 0.0553)
(0.100, 0.0656)
(0.101, 0.0777)
(0.102, 0.0918)
(0.103, 0.1084)
(0.104, 0.1277)
(0.105, 0.1502)
(0.106, 0.1764)
(0.107, 0.2068)
(0.108, 0.2422)
(0.109, 0.2832)
(0.110, 0.3307)
(0.111, 0.3856)
(0.112, 0.4490)
(0.113, 0.5221)
(0.114, 0.6063)
(0.115, 0.7031)
(0.116, 0.8144)
(0.117, 0.9421)
(0.118, 1.0885)
(0.119, 1.2561)
(0.120, 1.4478)
(0.121, 1.6667)
(0.122, 1.9166)
(0.123, 2.2014)
(0.124, 2.5256)
(0.125, 2.8944)
(0.125, 0.0000)}
\defdata{emaprofile-1-long}{
(0.000, 0.0000)
(0.028, 0.0000)
(0.029, 0.0001)
(0.030, 0.0001)
(0.031, 0.0001)
(0.032, 0.0001)
(0.033, 0.0001)
(0.034, 0.0002)
(0.035, 0.0002)
(0.036, 0.0002)
(0.037, 0.0003)
(0.038, 0.0003)
(0.039, 0.0004)
(0.040, 0.0005)
(0.041, 0.0006)
(0.042, 0.0007)
(0.043, 0.0008)
(0.044, 0.0009)
(0.045, 0.0011)
(0.046, 0.0013)
(0.047, 0.0015)
(0.048, 0.0017)
(0.049, 0.0020)
(0.050, 0.0023)
(0.051, 0.0026)
(0.052, 0.0030)
(0.053, 0.0035)
(0.054, 0.0039)
(0.055, 0.0045)
(0.056, 0.0051)
(0.057, 0.0057)
(0.058, 0.0065)
(0.059, 0.0073)
(0.060, 0.0082)
(0.061, 0.0092)
(0.062, 0.0102)
(0.063, 0.0115)
(0.064, 0.0128)
(0.065, 0.0142)
(0.066, 0.0158)
(0.067, 0.0176)
(0.068, 0.0195)
(0.069, 0.0215)
(0.070, 0.0238)
(0.071, 0.0262)
(0.072, 0.0289)
(0.073, 0.0318)
(0.074, 0.0350)
(0.075, 0.0384)
(0.076, 0.0421)
(0.077, 0.0461)
(0.078, 0.0504)
(0.079, 0.0551)
(0.080, 0.0601)
(0.081, 0.0655)
(0.082, 0.0713)
(0.083, 0.0776)
(0.084, 0.0843)
(0.085, 0.0915)
(0.086, 0.0993)
(0.087, 0.1076)
(0.088, 0.1164)
(0.089, 0.1259)
(0.090, 0.1361)
(0.091, 0.1469)
(0.092, 0.1585)
(0.093, 0.1709)
(0.094, 0.1840)
(0.095, 0.1981)
(0.096, 0.2130)
(0.097, 0.2289)
(0.098, 0.2457)
(0.099, 0.2637)
(0.100, 0.2827)
(0.101, 0.3030)
(0.102, 0.3244)
(0.103, 0.3471)
(0.104, 0.3712)
(0.105, 0.3967)
(0.106, 0.4236)
(0.107, 0.4522)
(0.108, 0.4823)
(0.109, 0.5142)
(0.110, 0.5478)
(0.111, 0.5833)
(0.112, 0.6208)
(0.113, 0.6603)
(0.114, 0.7019)
(0.115, 0.7458)
(0.116, 0.7920)
(0.117, 0.8406)
(0.118, 0.8918)
(0.119, 0.9455)
(0.120, 1.0021)
(0.121, 1.0615)
(0.122, 1.1239)
(0.123, 1.1894)
(0.124, 1.2581)
(0.125, 1.3303)
(0.125, 0.0000)}
\defdata{emaprofile-2-short}{
(0.000, 0.0000)
(0.135, 0.0000)
(0.136, 0.0000)
(0.137, 0.0001)
(0.138, 0.0001)
(0.139, 0.0001)
(0.140, 0.0001)
(0.141, 0.0001)
(0.142, 0.0001)
(0.143, 0.0001)
(0.144, 0.0001)
(0.145, 0.0001)
(0.146, 0.0002)
(0.147, 0.0002)
(0.148, 0.0002)
(0.149, 0.0002)
(0.150, 0.0003)
(0.151, 0.0003)
(0.152, 0.0003)
(0.153, 0.0004)
(0.154, 0.0004)
(0.155, 0.0004)
(0.156, 0.0005)
(0.157, 0.0006)
(0.158, 0.0006)
(0.159, 0.0007)
(0.160, 0.0008)
(0.161, 0.0009)
(0.162, 0.0010)
(0.163, 0.0011)
(0.164, 0.0012)
(0.165, 0.0013)
(0.166, 0.0014)
(0.167, 0.0016)
(0.168, 0.0018)
(0.169, 0.0019)
(0.170, 0.0022)
(0.171, 0.0024)
(0.172, 0.0026)
(0.173, 0.0029)
(0.174, 0.0032)
(0.175, 0.0035)
(0.176, 0.0039)
(0.177, 0.0043)
(0.178, 0.0047)
(0.179, 0.0052)
(0.180, 0.0057)
(0.181, 0.0062)
(0.182, 0.0069)
(0.183, 0.0075)
(0.184, 0.0083)
(0.185, 0.0091)
(0.186, 0.0099)
(0.187, 0.0109)
(0.188, 0.0119)
(0.189, 0.0130)
(0.190, 0.0142)
(0.191, 0.0156)
(0.192, 0.0170)
(0.193, 0.0186)
(0.194, 0.0203)
(0.195, 0.0221)
(0.196, 0.0241)
(0.197, 0.0263)
(0.198, 0.0287)
(0.199, 0.0312)
(0.200, 0.0340)
(0.201, 0.0370)
(0.202, 0.0402)
(0.203, 0.0438)
(0.204, 0.0476)
(0.205, 0.0517)
(0.206, 0.0561)
(0.207, 0.0609)
(0.208, 0.0661)
(0.209, 0.0717)
(0.210, 0.0778)
(0.211, 0.0843)
(0.212, 0.0914)
(0.213, 0.0989)
(0.214, 0.1071)
(0.215, 0.1160)
(0.216, 0.1255)
(0.217, 0.1357)
(0.218, 0.1467)
(0.219, 0.1585)
(0.220, 0.1713)
(0.221, 0.1850)
(0.222, 0.1997)
(0.223, 0.2155)
(0.224, 0.2325)
(0.225, 0.2508)
(0.226, 0.2704)
(0.227, 0.2914)
(0.228, 0.3140)
(0.229, 0.3382)
(0.230, 0.3642)
(0.231, 0.3920)
(0.232, 0.4218)
(0.233, 0.4538)
(0.234, 0.4880)
(0.235, 0.5246)
(0.236, 0.5638)
(0.237, 0.6057)
(0.238, 0.6506)
(0.239, 0.6986)
(0.240, 0.7499)
(0.241, 0.8047)
(0.242, 0.8633)
(0.243, 0.9258)
(0.244, 0.9927)
(0.245, 1.0640)
(0.246, 1.1401)
(0.247, 1.2214)
(0.248, 1.3081)
(0.249, 1.4005)
(0.250, 1.4991)
(0.250, 0.0000)}
\defdata{emaprofile-2-long}{
(0.000, 0.0000)
(0.062, 0.0000)
(0.063, 0.0000)
(0.064, 0.0001)
(0.065, 0.0001)
(0.066, 0.0001)
(0.067, 0.0001)
(0.068, 0.0001)
(0.069, 0.0001)
(0.070, 0.0001)
(0.071, 0.0001)
(0.072, 0.0001)
(0.073, 0.0001)
(0.074, 0.0001)
(0.075, 0.0002)
(0.076, 0.0002)
(0.077, 0.0002)
(0.078, 0.0002)
(0.079, 0.0002)
(0.080, 0.0002)
(0.081, 0.0003)
(0.082, 0.0003)
(0.083, 0.0003)
(0.084, 0.0003)
(0.085, 0.0004)
(0.086, 0.0004)
(0.087, 0.0004)
(0.088, 0.0005)
(0.089, 0.0005)
(0.090, 0.0006)
(0.091, 0.0006)
(0.092, 0.0007)
(0.093, 0.0007)
(0.094, 0.0008)
(0.095, 0.0008)
(0.096, 0.0009)
(0.097, 0.0009)
(0.098, 0.0010)
(0.099, 0.0011)
(0.100, 0.0012)
(0.101, 0.0013)
(0.102, 0.0013)
(0.103, 0.0014)
(0.104, 0.0015)
(0.105, 0.0016)
(0.106, 0.0018)
(0.107, 0.0019)
(0.108, 0.0020)
(0.109, 0.0021)
(0.110, 0.0023)
(0.111, 0.0024)
(0.112, 0.0026)
(0.113, 0.0027)
(0.114, 0.0029)
(0.115, 0.0031)
(0.116, 0.0033)
(0.117, 0.0035)
(0.118, 0.0037)
(0.119, 0.0039)
(0.120, 0.0041)
(0.121, 0.0044)
(0.122, 0.0046)
(0.123, 0.0049)
(0.124, 0.0052)
(0.125, 0.0055)
(0.126, 0.0058)
(0.127, 0.0061)
(0.128, 0.0065)
(0.129, 0.0068)
(0.130, 0.0072)
(0.131, 0.0076)
(0.132, 0.0080)
(0.133, 0.0085)
(0.134, 0.0089)
(0.135, 0.0094)
(0.136, 0.0099)
(0.137, 0.0104)
(0.138, 0.0109)
(0.139, 0.0115)
(0.140, 0.0121)
(0.141, 0.0127)
(0.142, 0.0133)
(0.143, 0.0140)
(0.144, 0.0147)
(0.145, 0.0154)
(0.146, 0.0162)
(0.147, 0.0170)
(0.148, 0.0178)
(0.149, 0.0186)
(0.150, 0.0195)
(0.151, 0.0204)
(0.152, 0.0214)
(0.153, 0.0224)
(0.154, 0.0234)
(0.155, 0.0245)
(0.156, 0.0256)
(0.157, 0.0268)
(0.158, 0.0280)
(0.159, 0.0292)
(0.160, 0.0305)
(0.161, 0.0319)
(0.162, 0.0333)
(0.163, 0.0347)
(0.164, 0.0362)
(0.165, 0.0378)
(0.166, 0.0394)
(0.167, 0.0411)
(0.168, 0.0428)
(0.169, 0.0446)
(0.170, 0.0465)
(0.171, 0.0484)
(0.172, 0.0504)
(0.173, 0.0525)
(0.174, 0.0546)
(0.175, 0.0568)
(0.176, 0.0591)
(0.177, 0.0615)
(0.178, 0.0640)
(0.179, 0.0665)
(0.180, 0.0691)
(0.181, 0.0718)
(0.182, 0.0746)
(0.183, 0.0775)
(0.184, 0.0805)
(0.185, 0.0836)
(0.186, 0.0868)
(0.187, 0.0901)
(0.188, 0.0935)
(0.189, 0.0970)
(0.190, 0.1006)
(0.191, 0.1043)
(0.192, 0.1082)
(0.193, 0.1122)
(0.194, 0.1162)
(0.195, 0.1205)
(0.196, 0.1248)
(0.197, 0.1293)
(0.198, 0.1339)
(0.199, 0.1387)
(0.200, 0.1436)
(0.201, 0.1487)
(0.202, 0.1539)
(0.203, 0.1592)
(0.204, 0.1648)
(0.205, 0.1705)
(0.206, 0.1763)
(0.207, 0.1823)
(0.208, 0.1885)
(0.209, 0.1949)
(0.210, 0.2015)
(0.211, 0.2082)
(0.212, 0.2152)
(0.213, 0.2223)
(0.214, 0.2297)
(0.215, 0.2372)
(0.216, 0.2450)
(0.217, 0.2530)
(0.218, 0.2612)
(0.219, 0.2696)
(0.220, 0.2783)
(0.221, 0.2872)
(0.222, 0.2963)
(0.223, 0.3057)
(0.224, 0.3153)
(0.225, 0.3252)
(0.226, 0.3354)
(0.227, 0.3458)
(0.228, 0.3565)
(0.229, 0.3675)
(0.230, 0.3788)
(0.231, 0.3904)
(0.232, 0.4023)
(0.233, 0.4145)
(0.234, 0.4270)
(0.235, 0.4398)
(0.236, 0.4529)
(0.237, 0.4664)
(0.238, 0.4803)
(0.239, 0.4944)
(0.240, 0.5090)
(0.241, 0.5239)
(0.242, 0.5392)
(0.243, 0.5548)
(0.244, 0.5708)
(0.245, 0.5873)
(0.246, 0.6041)
(0.247, 0.6214)
(0.248, 0.6390)
(0.249, 0.6571)
(0.250, 0.6757)
(0.250, 0.0000)}
\defdata{emaprofile-3-short}{
(0.000, 0.0000)
(0.207, 0.0000)
(0.208, 0.0000)
(0.209, 0.0000)
(0.210, 0.0001)
(0.211, 0.0001)
(0.212, 0.0001)
(0.213, 0.0001)
(0.214, 0.0001)
(0.215, 0.0001)
(0.216, 0.0001)
(0.217, 0.0001)
(0.218, 0.0001)
(0.219, 0.0001)
(0.220, 0.0001)
(0.221, 0.0001)
(0.222, 0.0001)
(0.223, 0.0001)
(0.224, 0.0002)
(0.225, 0.0002)
(0.226, 0.0002)
(0.227, 0.0002)
(0.228, 0.0002)
(0.229, 0.0002)
(0.230, 0.0003)
(0.231, 0.0003)
(0.232, 0.0003)
(0.233, 0.0003)
(0.234, 0.0003)
(0.235, 0.0004)
(0.236, 0.0004)
(0.237, 0.0004)
(0.238, 0.0005)
(0.239, 0.0005)
(0.240, 0.0005)
(0.241, 0.0006)
(0.242, 0.0006)
(0.243, 0.0006)
(0.244, 0.0007)
(0.245, 0.0007)
(0.246, 0.0008)
(0.247, 0.0008)
(0.248, 0.0009)
(0.249, 0.0010)
(0.250, 0.0010)
(0.251, 0.0011)
(0.252, 0.0012)
(0.253, 0.0013)
(0.254, 0.0014)
(0.255, 0.0015)
(0.256, 0.0016)
(0.257, 0.0017)
(0.258, 0.0018)
(0.259, 0.0019)
(0.260, 0.0020)
(0.261, 0.0022)
(0.262, 0.0023)
(0.263, 0.0025)
(0.264, 0.0026)
(0.265, 0.0028)
(0.266, 0.0030)
(0.267, 0.0032)
(0.268, 0.0034)
(0.269, 0.0036)
(0.270, 0.0038)
(0.271, 0.0041)
(0.272, 0.0043)
(0.273, 0.0046)
(0.274, 0.0049)
(0.275, 0.0052)
(0.276, 0.0056)
(0.277, 0.0059)
(0.278, 0.0063)
(0.279, 0.0067)
(0.280, 0.0071)
(0.281, 0.0076)
(0.282, 0.0080)
(0.283, 0.0085)
(0.284, 0.0090)
(0.285, 0.0096)
(0.286, 0.0102)
(0.287, 0.0108)
(0.288, 0.0115)
(0.289, 0.0122)
(0.290, 0.0129)
(0.291, 0.0137)
(0.292, 0.0145)
(0.293, 0.0154)
(0.294, 0.0163)
(0.295, 0.0172)
(0.296, 0.0183)
(0.297, 0.0193)
(0.298, 0.0205)
(0.299, 0.0217)
(0.300, 0.0229)
(0.301, 0.0243)
(0.302, 0.0257)
(0.303, 0.0271)
(0.304, 0.0287)
(0.305, 0.0303)
(0.306, 0.0321)
(0.307, 0.0339)
(0.308, 0.0358)
(0.309, 0.0379)
(0.310, 0.0400)
(0.311, 0.0422)
(0.312, 0.0446)
(0.313, 0.0471)
(0.314, 0.0497)
(0.315, 0.0525)
(0.316, 0.0554)
(0.317, 0.0584)
(0.318, 0.0616)
(0.319, 0.0650)
(0.320, 0.0685)
(0.321, 0.0723)
(0.322, 0.0762)
(0.323, 0.0803)
(0.324, 0.0846)
(0.325, 0.0892)
(0.326, 0.0939)
(0.327, 0.0990)
(0.328, 0.1042)
(0.329, 0.1097)
(0.330, 0.1155)
(0.331, 0.1216)
(0.332, 0.1280)
(0.333, 0.1347)
(0.334, 0.1418)
(0.335, 0.1491)
(0.336, 0.1569)
(0.337, 0.1650)
(0.338, 0.1735)
(0.339, 0.1824)
(0.340, 0.1918)
(0.341, 0.2016)
(0.342, 0.2118)
(0.343, 0.2226)
(0.344, 0.2339)
(0.345, 0.2457)
(0.346, 0.2580)
(0.347, 0.2710)
(0.348, 0.2846)
(0.349, 0.2988)
(0.350, 0.3136)
(0.351, 0.3292)
(0.352, 0.3455)
(0.353, 0.3625)
(0.354, 0.3803)
(0.355, 0.3990)
(0.356, 0.4185)
(0.357, 0.4389)
(0.358, 0.4602)
(0.359, 0.4825)
(0.360, 0.5059)
(0.361, 0.5302)
(0.362, 0.5557)
(0.363, 0.5824)
(0.364, 0.6102)
(0.365, 0.6393)
(0.366, 0.6697)
(0.367, 0.7014)
(0.368, 0.7345)
(0.369, 0.7692)
(0.370, 0.8053)
(0.371, 0.8431)
(0.372, 0.8825)
(0.373, 0.9236)
(0.374, 0.9665)
(0.375, 1.0113)
(0.375, 0.0000)}
\defdata{emaprofile-3-long}{
(0.000, 0.0000)
(0.098, 0.0000)
(0.099, 0.0000)
(0.100, 0.0000)
(0.101, 0.0001)
(0.102, 0.0001)
(0.103, 0.0001)
(0.104, 0.0001)
(0.105, 0.0001)
(0.106, 0.0001)
(0.107, 0.0001)
(0.108, 0.0001)
(0.109, 0.0001)
(0.110, 0.0001)
(0.111, 0.0001)
(0.112, 0.0001)
(0.113, 0.0001)
(0.114, 0.0001)
(0.115, 0.0001)
(0.116, 0.0001)
(0.117, 0.0001)
(0.118, 0.0001)
(0.119, 0.0002)
(0.120, 0.0002)
(0.121, 0.0002)
(0.122, 0.0002)
(0.123, 0.0002)
(0.124, 0.0002)
(0.125, 0.0002)
(0.126, 0.0002)
(0.127, 0.0002)
(0.128, 0.0003)
(0.129, 0.0003)
(0.130, 0.0003)
(0.131, 0.0003)
(0.132, 0.0003)
(0.133, 0.0003)
(0.134, 0.0004)
(0.135, 0.0004)
(0.136, 0.0004)
(0.137, 0.0004)
(0.138, 0.0004)
(0.139, 0.0005)
(0.140, 0.0005)
(0.141, 0.0005)
(0.142, 0.0005)
(0.143, 0.0006)
(0.144, 0.0006)
(0.145, 0.0006)
(0.146, 0.0006)
(0.147, 0.0007)
(0.148, 0.0007)
(0.149, 0.0007)
(0.150, 0.0008)
(0.151, 0.0008)
(0.152, 0.0009)
(0.153, 0.0009)
(0.154, 0.0009)
(0.155, 0.0010)
(0.156, 0.0010)
(0.157, 0.0011)
(0.158, 0.0011)
(0.159, 0.0012)
(0.160, 0.0012)
(0.161, 0.0013)
(0.162, 0.0013)
(0.163, 0.0014)
(0.164, 0.0015)
(0.165, 0.0015)
(0.166, 0.0016)
(0.167, 0.0017)
(0.168, 0.0017)
(0.169, 0.0018)
(0.170, 0.0019)
(0.171, 0.0019)
(0.172, 0.0020)
(0.173, 0.0021)
(0.174, 0.0022)
(0.175, 0.0023)
(0.176, 0.0024)
(0.177, 0.0025)
(0.178, 0.0026)
(0.179, 0.0027)
(0.180, 0.0028)
(0.181, 0.0029)
(0.182, 0.0030)
(0.183, 0.0031)
(0.184, 0.0032)
(0.185, 0.0034)
(0.186, 0.0035)
(0.187, 0.0036)
(0.188, 0.0038)
(0.189, 0.0039)
(0.190, 0.0040)
(0.191, 0.0042)
(0.192, 0.0043)
(0.193, 0.0045)
(0.194, 0.0047)
(0.195, 0.0048)
(0.196, 0.0050)
(0.197, 0.0052)
(0.198, 0.0054)
(0.199, 0.0056)
(0.200, 0.0058)
(0.201, 0.0060)
(0.202, 0.0062)
(0.203, 0.0064)
(0.204, 0.0066)
(0.205, 0.0069)
(0.206, 0.0071)
(0.207, 0.0073)
(0.208, 0.0076)
(0.209, 0.0078)
(0.210, 0.0081)
(0.211, 0.0084)
(0.212, 0.0086)
(0.213, 0.0089)
(0.214, 0.0092)
(0.215, 0.0095)
(0.216, 0.0098)
(0.217, 0.0102)
(0.218, 0.0105)
(0.219, 0.0108)
(0.220, 0.0112)
(0.221, 0.0115)
(0.222, 0.0119)
(0.223, 0.0123)
(0.224, 0.0127)
(0.225, 0.0131)
(0.226, 0.0135)
(0.227, 0.0139)
(0.228, 0.0143)
(0.229, 0.0148)
(0.230, 0.0152)
(0.231, 0.0157)
(0.232, 0.0162)
(0.233, 0.0167)
(0.234, 0.0172)
(0.235, 0.0177)
(0.236, 0.0182)
(0.237, 0.0187)
(0.238, 0.0193)
(0.239, 0.0199)
(0.240, 0.0205)
(0.241, 0.0211)
(0.242, 0.0217)
(0.243, 0.0223)
(0.244, 0.0229)
(0.245, 0.0236)
(0.246, 0.0243)
(0.247, 0.0250)
(0.248, 0.0257)
(0.249, 0.0264)
(0.250, 0.0272)
(0.251, 0.0279)
(0.252, 0.0287)
(0.253, 0.0295)
(0.254, 0.0303)
(0.255, 0.0312)
(0.256, 0.0320)
(0.257, 0.0329)
(0.258, 0.0338)
(0.259, 0.0347)
(0.260, 0.0357)
(0.261, 0.0366)
(0.262, 0.0376)
(0.263, 0.0386)
(0.264, 0.0396)
(0.265, 0.0407)
(0.266, 0.0418)
(0.267, 0.0429)
(0.268, 0.0440)
(0.269, 0.0451)
(0.270, 0.0463)
(0.271, 0.0475)
(0.272, 0.0488)
(0.273, 0.0500)
(0.274, 0.0513)
(0.275, 0.0526)
(0.276, 0.0540)
(0.277, 0.0553)
(0.278, 0.0567)
(0.279, 0.0582)
(0.280, 0.0596)
(0.281, 0.0611)
(0.282, 0.0626)
(0.283, 0.0642)
(0.284, 0.0658)
(0.285, 0.0674)
(0.286, 0.0691)
(0.287, 0.0708)
(0.288, 0.0725)
(0.289, 0.0743)
(0.290, 0.0761)
(0.291, 0.0779)
(0.292, 0.0798)
(0.293, 0.0817)
(0.294, 0.0837)
(0.295, 0.0856)
(0.296, 0.0877)
(0.297, 0.0898)
(0.298, 0.0919)
(0.299, 0.0940)
(0.300, 0.0962)
(0.301, 0.0985)
(0.302, 0.1008)
(0.303, 0.1031)
(0.304, 0.1055)
(0.305, 0.1079)
(0.306, 0.1104)
(0.307, 0.1130)
(0.308, 0.1155)
(0.309, 0.1182)
(0.310, 0.1208)
(0.311, 0.1236)
(0.312, 0.1264)
(0.313, 0.1292)
(0.314, 0.1321)
(0.315, 0.1350)
(0.316, 0.1380)
(0.317, 0.1411)
(0.318, 0.1442)
(0.319, 0.1474)
(0.320, 0.1506)
(0.321, 0.1539)
(0.322, 0.1573)
(0.323, 0.1607)
(0.324, 0.1642)
(0.325, 0.1677)
(0.326, 0.1714)
(0.327, 0.1750)
(0.328, 0.1788)
(0.329, 0.1826)
(0.330, 0.1865)
(0.331, 0.1904)
(0.332, 0.1945)
(0.333, 0.1986)
(0.334, 0.2027)
(0.335, 0.2070)
(0.336, 0.2113)
(0.337, 0.2157)
(0.338, 0.2202)
(0.339, 0.2248)
(0.340, 0.2294)
(0.341, 0.2341)
(0.342, 0.2389)
(0.343, 0.2438)
(0.344, 0.2488)
(0.345, 0.2539)
(0.346, 0.2590)
(0.347, 0.2643)
(0.348, 0.2696)
(0.349, 0.2750)
(0.350, 0.2805)
(0.351, 0.2861)
(0.352, 0.2919)
(0.353, 0.2977)
(0.354, 0.3036)
(0.355, 0.3096)
(0.356, 0.3157)
(0.357, 0.3219)
(0.358, 0.3282)
(0.359, 0.3346)
(0.360, 0.3411)
(0.361, 0.3477)
(0.362, 0.3545)
(0.363, 0.3613)
(0.364, 0.3683)
(0.365, 0.3754)
(0.366, 0.3826)
(0.367, 0.3899)
(0.368, 0.3973)
(0.369, 0.4049)
(0.370, 0.4125)
(0.371, 0.4204)
(0.372, 0.4283)
(0.373, 0.4363)
(0.374, 0.4445)
(0.375, 0.4528)
(0.375, 0.0000)}
\defdata{emaprofile-4-short}{
(0.000, 0.0000)
(0.280, 0.0000)
(0.281, 0.0000)
(0.282, 0.0000)
(0.283, 0.0000)
(0.284, 0.0001)
(0.285, 0.0001)
(0.286, 0.0001)
(0.287, 0.0001)
(0.288, 0.0001)
(0.289, 0.0001)
(0.290, 0.0001)
(0.291, 0.0001)
(0.292, 0.0001)
(0.293, 0.0001)
(0.294, 0.0001)
(0.295, 0.0001)
(0.296, 0.0001)
(0.297, 0.0001)
(0.298, 0.0001)
(0.299, 0.0001)
(0.300, 0.0001)
(0.301, 0.0001)
(0.302, 0.0001)
(0.303, 0.0002)
(0.304, 0.0002)
(0.305, 0.0002)
(0.306, 0.0002)
(0.307, 0.0002)
(0.308, 0.0002)
(0.309, 0.0002)
(0.310, 0.0002)
(0.311, 0.0002)
(0.312, 0.0003)
(0.313, 0.0003)
(0.314, 0.0003)
(0.315, 0.0003)
(0.316, 0.0003)
(0.317, 0.0003)
(0.318, 0.0004)
(0.319, 0.0004)
(0.320, 0.0004)
(0.321, 0.0004)
(0.322, 0.0004)
(0.323, 0.0005)
(0.324, 0.0005)
(0.325, 0.0005)
(0.326, 0.0005)
(0.327, 0.0006)
(0.328, 0.0006)
(0.329, 0.0006)
(0.330, 0.0007)
(0.331, 0.0007)
(0.332, 0.0007)
(0.333, 0.0008)
(0.334, 0.0008)
(0.335, 0.0009)
(0.336, 0.0009)
(0.337, 0.0009)
(0.338, 0.0010)
(0.339, 0.0010)
(0.340, 0.0011)
(0.341, 0.0012)
(0.342, 0.0012)
(0.343, 0.0013)
(0.344, 0.0013)
(0.345, 0.0014)
(0.346, 0.0015)
(0.347, 0.0016)
(0.348, 0.0016)
(0.349, 0.0017)
(0.350, 0.0018)
(0.351, 0.0019)
(0.352, 0.0020)
(0.353, 0.0021)
(0.354, 0.0022)
(0.355, 0.0023)
(0.356, 0.0024)
(0.357, 0.0025)
(0.358, 0.0026)
(0.359, 0.0028)
(0.360, 0.0029)
(0.361, 0.0030)
(0.362, 0.0032)
(0.363, 0.0033)
(0.364, 0.0035)
(0.365, 0.0037)
(0.366, 0.0038)
(0.367, 0.0040)
(0.368, 0.0042)
(0.369, 0.0044)
(0.370, 0.0046)
(0.371, 0.0048)
(0.372, 0.0050)
(0.373, 0.0053)
(0.374, 0.0055)
(0.375, 0.0058)
(0.376, 0.0061)
(0.377, 0.0063)
(0.378, 0.0066)
(0.379, 0.0069)
(0.380, 0.0072)
(0.381, 0.0076)
(0.382, 0.0079)
(0.383, 0.0083)
(0.384, 0.0087)
(0.385, 0.0090)
(0.386, 0.0094)
(0.387, 0.0099)
(0.388, 0.0103)
(0.389, 0.0108)
(0.390, 0.0113)
(0.391, 0.0118)
(0.392, 0.0123)
(0.393, 0.0128)
(0.394, 0.0134)
(0.395, 0.0140)
(0.396, 0.0146)
(0.397, 0.0152)
(0.398, 0.0159)
(0.399, 0.0166)
(0.400, 0.0173)
(0.401, 0.0180)
(0.402, 0.0188)
(0.403, 0.0196)
(0.404, 0.0205)
(0.405, 0.0214)
(0.406, 0.0223)
(0.407, 0.0232)
(0.408, 0.0242)
(0.409, 0.0252)
(0.410, 0.0263)
(0.411, 0.0274)
(0.412, 0.0286)
(0.413, 0.0298)
(0.414, 0.0310)
(0.415, 0.0323)
(0.416, 0.0337)
(0.417, 0.0351)
(0.418, 0.0365)
(0.419, 0.0380)
(0.420, 0.0396)
(0.421, 0.0412)
(0.422, 0.0429)
(0.423, 0.0447)
(0.424, 0.0465)
(0.425, 0.0484)
(0.426, 0.0504)
(0.427, 0.0524)
(0.428, 0.0545)
(0.429, 0.0567)
(0.430, 0.0590)
(0.431, 0.0614)
(0.432, 0.0639)
(0.433, 0.0664)
(0.434, 0.0691)
(0.435, 0.0718)
(0.436, 0.0747)
(0.437, 0.0776)
(0.438, 0.0807)
(0.439, 0.0839)
(0.440, 0.0872)
(0.441, 0.0906)
(0.442, 0.0942)
(0.443, 0.0978)
(0.444, 0.1016)
(0.445, 0.1056)
(0.446, 0.1097)
(0.447, 0.1140)
(0.448, 0.1184)
(0.449, 0.1229)
(0.450, 0.1277)
(0.451, 0.1326)
(0.452, 0.1376)
(0.453, 0.1429)
(0.454, 0.1483)
(0.455, 0.1540)
(0.456, 0.1598)
(0.457, 0.1659)
(0.458, 0.1721)
(0.459, 0.1786)
(0.460, 0.1854)
(0.461, 0.1923)
(0.462, 0.1995)
(0.463, 0.2070)
(0.464, 0.2147)
(0.465, 0.2227)
(0.466, 0.2310)
(0.467, 0.2395)
(0.468, 0.2484)
(0.469, 0.2575)
(0.470, 0.2670)
(0.471, 0.2768)
(0.472, 0.2870)
(0.473, 0.2974)
(0.474, 0.3083)
(0.475, 0.3195)
(0.476, 0.3311)
(0.477, 0.3431)
(0.478, 0.3556)
(0.479, 0.3684)
(0.480, 0.3817)
(0.481, 0.3954)
(0.482, 0.4096)
(0.483, 0.4242)
(0.484, 0.4394)
(0.485, 0.4550)
(0.486, 0.4712)
(0.487, 0.4880)
(0.488, 0.5052)
(0.489, 0.5231)
(0.490, 0.5415)
(0.491, 0.5606)
(0.492, 0.5803)
(0.493, 0.6007)
(0.494, 0.6217)
(0.495, 0.6434)
(0.496, 0.6658)
(0.497, 0.6889)
(0.498, 0.7128)
(0.499, 0.7375)
(0.500, 0.7630)
(0.500, 0.0000)}
\defdata{emaprofile-4-long}{
(0.000, 0.0000)
(0.136, 0.0000)
(0.137, 0.0000)
(0.138, 0.0000)
(0.139, 0.0000)
(0.140, 0.0000)
(0.141, 0.0001)
(0.142, 0.0001)
(0.143, 0.0001)
(0.144, 0.0001)
(0.145, 0.0001)
(0.146, 0.0001)
(0.147, 0.0001)
(0.148, 0.0001)
(0.149, 0.0001)
(0.150, 0.0001)
(0.151, 0.0001)
(0.152, 0.0001)
(0.153, 0.0001)
(0.154, 0.0001)
(0.155, 0.0001)
(0.156, 0.0001)
(0.157, 0.0001)
(0.158, 0.0001)
(0.159, 0.0001)
(0.160, 0.0001)
(0.161, 0.0001)
(0.162, 0.0001)
(0.163, 0.0001)
(0.164, 0.0001)
(0.165, 0.0002)
(0.166, 0.0002)
(0.167, 0.0002)
(0.168, 0.0002)
(0.169, 0.0002)
(0.170, 0.0002)
(0.171, 0.0002)
(0.172, 0.0002)
(0.173, 0.0002)
(0.174, 0.0002)
(0.175, 0.0002)
(0.176, 0.0002)
(0.177, 0.0003)
(0.178, 0.0003)
(0.179, 0.0003)
(0.180, 0.0003)
(0.181, 0.0003)
(0.182, 0.0003)
(0.183, 0.0003)
(0.184, 0.0003)
(0.185, 0.0003)
(0.186, 0.0004)
(0.187, 0.0004)
(0.188, 0.0004)
(0.189, 0.0004)
(0.190, 0.0004)
(0.191, 0.0004)
(0.192, 0.0004)
(0.193, 0.0005)
(0.194, 0.0005)
(0.195, 0.0005)
(0.196, 0.0005)
(0.197, 0.0005)
(0.198, 0.0005)
(0.199, 0.0006)
(0.200, 0.0006)
(0.201, 0.0006)
(0.202, 0.0006)
(0.203, 0.0007)
(0.204, 0.0007)
(0.205, 0.0007)
(0.206, 0.0007)
(0.207, 0.0007)
(0.208, 0.0008)
(0.209, 0.0008)
(0.210, 0.0008)
(0.211, 0.0009)
(0.212, 0.0009)
(0.213, 0.0009)
(0.214, 0.0009)
(0.215, 0.0010)
(0.216, 0.0010)
(0.217, 0.0010)
(0.218, 0.0011)
(0.219, 0.0011)
(0.220, 0.0011)
(0.221, 0.0012)
(0.222, 0.0012)
(0.223, 0.0013)
(0.224, 0.0013)
(0.225, 0.0013)
(0.226, 0.0014)
(0.227, 0.0014)
(0.228, 0.0015)
(0.229, 0.0015)
(0.230, 0.0016)
(0.231, 0.0016)
(0.232, 0.0017)
(0.233, 0.0017)
(0.234, 0.0018)
(0.235, 0.0018)
(0.236, 0.0019)
(0.237, 0.0019)
(0.238, 0.0020)
(0.239, 0.0020)
(0.240, 0.0021)
(0.241, 0.0022)
(0.242, 0.0022)
(0.243, 0.0023)
(0.244, 0.0023)
(0.245, 0.0024)
(0.246, 0.0025)
(0.247, 0.0026)
(0.248, 0.0026)
(0.249, 0.0027)
(0.250, 0.0028)
(0.251, 0.0029)
(0.252, 0.0029)
(0.253, 0.0030)
(0.254, 0.0031)
(0.255, 0.0032)
(0.256, 0.0033)
(0.257, 0.0034)
(0.258, 0.0035)
(0.259, 0.0035)
(0.260, 0.0036)
(0.261, 0.0037)
(0.262, 0.0038)
(0.263, 0.0039)
(0.264, 0.0040)
(0.265, 0.0042)
(0.266, 0.0043)
(0.267, 0.0044)
(0.268, 0.0045)
(0.269, 0.0046)
(0.270, 0.0047)
(0.271, 0.0049)
(0.272, 0.0050)
(0.273, 0.0051)
(0.274, 0.0052)
(0.275, 0.0054)
(0.276, 0.0055)
(0.277, 0.0057)
(0.278, 0.0058)
(0.279, 0.0059)
(0.280, 0.0061)
(0.281, 0.0062)
(0.282, 0.0064)
(0.283, 0.0066)
(0.284, 0.0067)
(0.285, 0.0069)
(0.286, 0.0071)
(0.287, 0.0072)
(0.288, 0.0074)
(0.289, 0.0076)
(0.290, 0.0078)
(0.291, 0.0080)
(0.292, 0.0081)
(0.293, 0.0083)
(0.294, 0.0085)
(0.295, 0.0087)
(0.296, 0.0090)
(0.297, 0.0092)
(0.298, 0.0094)
(0.299, 0.0096)
(0.300, 0.0098)
(0.301, 0.0101)
(0.302, 0.0103)
(0.303, 0.0105)
(0.304, 0.0108)
(0.305, 0.0110)
(0.306, 0.0113)
(0.307, 0.0115)
(0.308, 0.0118)
(0.309, 0.0121)
(0.310, 0.0123)
(0.311, 0.0126)
(0.312, 0.0129)
(0.313, 0.0132)
(0.314, 0.0135)
(0.315, 0.0138)
(0.316, 0.0141)
(0.317, 0.0144)
(0.318, 0.0147)
(0.319, 0.0151)
(0.320, 0.0154)
(0.321, 0.0157)
(0.322, 0.0161)
(0.323, 0.0164)
(0.324, 0.0168)
(0.325, 0.0171)
(0.326, 0.0175)
(0.327, 0.0179)
(0.328, 0.0183)
(0.329, 0.0186)
(0.330, 0.0190)
(0.331, 0.0194)
(0.332, 0.0199)
(0.333, 0.0203)
(0.334, 0.0207)
(0.335, 0.0211)
(0.336, 0.0216)
(0.337, 0.0220)
(0.338, 0.0225)
(0.339, 0.0230)
(0.340, 0.0234)
(0.341, 0.0239)
(0.342, 0.0244)
(0.343, 0.0249)
(0.344, 0.0254)
(0.345, 0.0259)
(0.346, 0.0265)
(0.347, 0.0270)
(0.348, 0.0275)
(0.349, 0.0281)
(0.350, 0.0286)
(0.351, 0.0292)
(0.352, 0.0298)
(0.353, 0.0304)
(0.354, 0.0310)
(0.355, 0.0316)
(0.356, 0.0322)
(0.357, 0.0329)
(0.358, 0.0335)
(0.359, 0.0342)
(0.360, 0.0348)
(0.361, 0.0355)
(0.362, 0.0362)
(0.363, 0.0369)
(0.364, 0.0376)
(0.365, 0.0383)
(0.366, 0.0391)
(0.367, 0.0398)
(0.368, 0.0406)
(0.369, 0.0413)
(0.370, 0.0421)
(0.371, 0.0429)
(0.372, 0.0437)
(0.373, 0.0446)
(0.374, 0.0454)
(0.375, 0.0462)
(0.376, 0.0471)
(0.377, 0.0480)
(0.378, 0.0489)
(0.379, 0.0498)
(0.380, 0.0507)
(0.381, 0.0516)
(0.382, 0.0526)
(0.383, 0.0535)
(0.384, 0.0545)
(0.385, 0.0555)
(0.386, 0.0565)
(0.387, 0.0575)
(0.388, 0.0586)
(0.389, 0.0596)
(0.390, 0.0607)
(0.391, 0.0618)
(0.392, 0.0629)
(0.393, 0.0640)
(0.394, 0.0652)
(0.395, 0.0663)
(0.396, 0.0675)
(0.397, 0.0687)
(0.398, 0.0699)
(0.399, 0.0711)
(0.400, 0.0724)
(0.401, 0.0736)
(0.402, 0.0749)
(0.403, 0.0762)
(0.404, 0.0775)
(0.405, 0.0789)
(0.406, 0.0803)
(0.407, 0.0816)
(0.408, 0.0830)
(0.409, 0.0845)
(0.410, 0.0859)
(0.411, 0.0874)
(0.412, 0.0889)
(0.413, 0.0904)
(0.414, 0.0919)
(0.415, 0.0934)
(0.416, 0.0950)
(0.417, 0.0966)
(0.418, 0.0982)
(0.419, 0.0999)
(0.420, 0.1015)
(0.421, 0.1032)
(0.422, 0.1049)
(0.423, 0.1067)
(0.424, 0.1084)
(0.425, 0.1102)
(0.426, 0.1120)
(0.427, 0.1139)
(0.428, 0.1157)
(0.429, 0.1176)
(0.430, 0.1196)
(0.431, 0.1215)
(0.432, 0.1235)
(0.433, 0.1255)
(0.434, 0.1275)
(0.435, 0.1295)
(0.436, 0.1316)
(0.437, 0.1337)
(0.438, 0.1359)
(0.439, 0.1380)
(0.440, 0.1402)
(0.441, 0.1425)
(0.442, 0.1447)
(0.443, 0.1470)
(0.444, 0.1493)
(0.445, 0.1517)
(0.446, 0.1541)
(0.447, 0.1565)
(0.448, 0.1589)
(0.449, 0.1614)
(0.450, 0.1639)
(0.451, 0.1664)
(0.452, 0.1690)
(0.453, 0.1716)
(0.454, 0.1743)
(0.455, 0.1770)
(0.456, 0.1797)
(0.457, 0.1824)
(0.458, 0.1852)
(0.459, 0.1880)
(0.460, 0.1909)
(0.461, 0.1938)
(0.462, 0.1967)
(0.463, 0.1997)
(0.464, 0.2027)
(0.465, 0.2058)
(0.466, 0.2089)
(0.467, 0.2120)
(0.468, 0.2152)
(0.469, 0.2184)
(0.470, 0.2216)
(0.471, 0.2249)
(0.472, 0.2283)
(0.473, 0.2316)
(0.474, 0.2351)
(0.475, 0.2385)
(0.476, 0.2420)
(0.477, 0.2456)
(0.478, 0.2492)
(0.479, 0.2528)
(0.480, 0.2565)
(0.481, 0.2602)
(0.482, 0.2640)
(0.483, 0.2678)
(0.484, 0.2717)
(0.485, 0.2756)
(0.486, 0.2796)
(0.487, 0.2836)
(0.488, 0.2877)
(0.489, 0.2918)
(0.490, 0.2960)
(0.491, 0.3002)
(0.492, 0.3045)
(0.493, 0.3088)
(0.494, 0.3132)
(0.495, 0.3176)
(0.496, 0.3221)
(0.497, 0.3266)
(0.498, 0.3312)
(0.499, 0.3358)
(0.500, 0.3405)
(0.500, 0.0000)}
\defdata{emaprofile-5-short}{
(0.000, 0.0000)
(0.355, 0.0000)
(0.356, 0.0000)
(0.357, 0.0000)
(0.358, 0.0000)
(0.359, 0.0001)
(0.360, 0.0001)
(0.361, 0.0001)
(0.362, 0.0001)
(0.363, 0.0001)
(0.364, 0.0001)
(0.365, 0.0001)
(0.366, 0.0001)
(0.367, 0.0001)
(0.368, 0.0001)
(0.369, 0.0001)
(0.370, 0.0001)
(0.371, 0.0001)
(0.372, 0.0001)
(0.373, 0.0001)
(0.374, 0.0001)
(0.375, 0.0001)
(0.376, 0.0001)
(0.377, 0.0001)
(0.378, 0.0001)
(0.379, 0.0001)
(0.380, 0.0001)
(0.381, 0.0001)
(0.382, 0.0001)
(0.383, 0.0002)
(0.384, 0.0002)
(0.385, 0.0002)
(0.386, 0.0002)
(0.387, 0.0002)
(0.388, 0.0002)
(0.389, 0.0002)
(0.390, 0.0002)
(0.391, 0.0002)
(0.392, 0.0002)
(0.393, 0.0002)
(0.394, 0.0002)
(0.395, 0.0003)
(0.396, 0.0003)
(0.397, 0.0003)
(0.398, 0.0003)
(0.399, 0.0003)
(0.400, 0.0003)
(0.401, 0.0003)
(0.402, 0.0003)
(0.403, 0.0004)
(0.404, 0.0004)
(0.405, 0.0004)
(0.406, 0.0004)
(0.407, 0.0004)
(0.408, 0.0004)
(0.409, 0.0005)
(0.410, 0.0005)
(0.411, 0.0005)
(0.412, 0.0005)
(0.413, 0.0005)
(0.414, 0.0006)
(0.415, 0.0006)
(0.416, 0.0006)
(0.417, 0.0006)
(0.418, 0.0007)
(0.419, 0.0007)
(0.420, 0.0007)
(0.421, 0.0008)
(0.422, 0.0008)
(0.423, 0.0008)
(0.424, 0.0008)
(0.425, 0.0009)
(0.426, 0.0009)
(0.427, 0.0010)
(0.428, 0.0010)
(0.429, 0.0010)
(0.430, 0.0011)
(0.431, 0.0011)
(0.432, 0.0012)
(0.433, 0.0012)
(0.434, 0.0013)
(0.435, 0.0013)
(0.436, 0.0014)
(0.437, 0.0014)
(0.438, 0.0015)
(0.439, 0.0015)
(0.440, 0.0016)
(0.441, 0.0016)
(0.442, 0.0017)
(0.443, 0.0018)
(0.444, 0.0019)
(0.445, 0.0019)
(0.446, 0.0020)
(0.447, 0.0021)
(0.448, 0.0022)
(0.449, 0.0022)
(0.450, 0.0023)
(0.451, 0.0024)
(0.452, 0.0025)
(0.453, 0.0026)
(0.454, 0.0027)
(0.455, 0.0028)
(0.456, 0.0029)
(0.457, 0.0030)
(0.458, 0.0031)
(0.459, 0.0033)
(0.460, 0.0034)
(0.461, 0.0035)
(0.462, 0.0036)
(0.463, 0.0038)
(0.464, 0.0039)
(0.465, 0.0041)
(0.466, 0.0042)
(0.467, 0.0044)
(0.468, 0.0045)
(0.469, 0.0047)
(0.470, 0.0049)
(0.471, 0.0050)
(0.472, 0.0052)
(0.473, 0.0054)
(0.474, 0.0056)
(0.475, 0.0058)
(0.476, 0.0060)
(0.477, 0.0062)
(0.478, 0.0065)
(0.479, 0.0067)
(0.480, 0.0069)
(0.481, 0.0072)
(0.482, 0.0075)
(0.483, 0.0077)
(0.484, 0.0080)
(0.485, 0.0083)
(0.486, 0.0086)
(0.487, 0.0089)
(0.488, 0.0092)
(0.489, 0.0095)
(0.490, 0.0099)
(0.491, 0.0102)
(0.492, 0.0106)
(0.493, 0.0109)
(0.494, 0.0113)
(0.495, 0.0117)
(0.496, 0.0121)
(0.497, 0.0125)
(0.498, 0.0130)
(0.499, 0.0134)
(0.500, 0.0139)
(0.501, 0.0144)
(0.502, 0.0149)
(0.503, 0.0154)
(0.504, 0.0159)
(0.505, 0.0164)
(0.506, 0.0170)
(0.507, 0.0176)
(0.508, 0.0182)
(0.509, 0.0188)
(0.510, 0.0194)
(0.511, 0.0201)
(0.512, 0.0208)
(0.513, 0.0215)
(0.514, 0.0222)
(0.515, 0.0229)
(0.516, 0.0237)
(0.517, 0.0245)
(0.518, 0.0253)
(0.519, 0.0262)
(0.520, 0.0270)
(0.521, 0.0279)
(0.522, 0.0288)
(0.523, 0.0298)
(0.524, 0.0308)
(0.525, 0.0318)
(0.526, 0.0328)
(0.527, 0.0339)
(0.528, 0.0350)
(0.529, 0.0362)
(0.530, 0.0373)
(0.531, 0.0385)
(0.532, 0.0398)
(0.533, 0.0411)
(0.534, 0.0424)
(0.535, 0.0438)
(0.536, 0.0452)
(0.537, 0.0466)
(0.538, 0.0481)
(0.539, 0.0497)
(0.540, 0.0513)
(0.541, 0.0529)
(0.542, 0.0546)
(0.543, 0.0563)
(0.544, 0.0581)
(0.545, 0.0599)
(0.546, 0.0618)
(0.547, 0.0638)
(0.548, 0.0658)
(0.549, 0.0679)
(0.550, 0.0700)
(0.551, 0.0722)
(0.552, 0.0744)
(0.553, 0.0768)
(0.554, 0.0791)
(0.555, 0.0816)
(0.556, 0.0841)
(0.557, 0.0867)
(0.558, 0.0894)
(0.559, 0.0922)
(0.560, 0.0950)
(0.561, 0.0979)
(0.562, 0.1010)
(0.563, 0.1040)
(0.564, 0.1072)
(0.565, 0.1105)
(0.566, 0.1139)
(0.567, 0.1173)
(0.568, 0.1209)
(0.569, 0.1246)
(0.570, 0.1283)
(0.571, 0.1322)
(0.572, 0.1362)
(0.573, 0.1403)
(0.574, 0.1445)
(0.575, 0.1488)
(0.576, 0.1533)
(0.577, 0.1578)
(0.578, 0.1626)
(0.579, 0.1674)
(0.580, 0.1724)
(0.581, 0.1775)
(0.582, 0.1827)
(0.583, 0.1881)
(0.584, 0.1937)
(0.585, 0.1994)
(0.586, 0.2053)
(0.587, 0.2113)
(0.588, 0.2175)
(0.589, 0.2238)
(0.590, 0.2304)
(0.591, 0.2371)
(0.592, 0.2440)
(0.593, 0.2511)
(0.594, 0.2584)
(0.595, 0.2659)
(0.596, 0.2735)
(0.597, 0.2814)
(0.598, 0.2895)
(0.599, 0.2979)
(0.600, 0.3064)
(0.601, 0.3152)
(0.602, 0.3242)
(0.603, 0.3335)
(0.604, 0.3430)
(0.605, 0.3528)
(0.606, 0.3628)
(0.607, 0.3731)
(0.608, 0.3836)
(0.609, 0.3945)
(0.610, 0.4056)
(0.611, 0.4171)
(0.612, 0.4288)
(0.613, 0.4409)
(0.614, 0.4532)
(0.615, 0.4659)
(0.616, 0.4789)
(0.617, 0.4923)
(0.618, 0.5060)
(0.619, 0.5201)
(0.620, 0.5345)
(0.621, 0.5494)
(0.622, 0.5646)
(0.623, 0.5802)
(0.624, 0.5962)
(0.625, 0.6126)
(0.625, 0.0000)}
\defdata{emaprofile-5-long}{
(0.000, 0.0000)
(0.176, 0.0000)
(0.177, 0.0000)
(0.178, 0.0000)
(0.179, 0.0000)
(0.180, 0.0000)
(0.181, 0.0001)
(0.182, 0.0001)
(0.183, 0.0001)
(0.184, 0.0001)
(0.185, 0.0001)
(0.186, 0.0001)
(0.187, 0.0001)
(0.188, 0.0001)
(0.189, 0.0001)
(0.190, 0.0001)
(0.191, 0.0001)
(0.192, 0.0001)
(0.193, 0.0001)
(0.194, 0.0001)
(0.195, 0.0001)
(0.196, 0.0001)
(0.197, 0.0001)
(0.198, 0.0001)
(0.199, 0.0001)
(0.200, 0.0001)
(0.201, 0.0001)
(0.202, 0.0001)
(0.203, 0.0001)
(0.204, 0.0001)
(0.205, 0.0001)
(0.206, 0.0001)
(0.207, 0.0001)
(0.208, 0.0001)
(0.209, 0.0001)
(0.210, 0.0001)
(0.211, 0.0001)
(0.212, 0.0002)
(0.213, 0.0002)
(0.214, 0.0002)
(0.215, 0.0002)
(0.216, 0.0002)
(0.217, 0.0002)
(0.218, 0.0002)
(0.219, 0.0002)
(0.220, 0.0002)
(0.221, 0.0002)
(0.222, 0.0002)
(0.223, 0.0002)
(0.224, 0.0002)
(0.225, 0.0002)
(0.226, 0.0002)
(0.227, 0.0002)
(0.228, 0.0002)
(0.229, 0.0003)
(0.230, 0.0003)
(0.231, 0.0003)
(0.232, 0.0003)
(0.233, 0.0003)
(0.234, 0.0003)
(0.235, 0.0003)
(0.236, 0.0003)
(0.237, 0.0003)
(0.238, 0.0003)
(0.239, 0.0003)
(0.240, 0.0004)
(0.241, 0.0004)
(0.242, 0.0004)
(0.243, 0.0004)
(0.244, 0.0004)
(0.245, 0.0004)
(0.246, 0.0004)
(0.247, 0.0004)
(0.248, 0.0004)
(0.249, 0.0005)
(0.250, 0.0005)
(0.251, 0.0005)
(0.252, 0.0005)
(0.253, 0.0005)
(0.254, 0.0005)
(0.255, 0.0005)
(0.256, 0.0006)
(0.257, 0.0006)
(0.258, 0.0006)
(0.259, 0.0006)
(0.260, 0.0006)
(0.261, 0.0006)
(0.262, 0.0007)
(0.263, 0.0007)
(0.264, 0.0007)
(0.265, 0.0007)
(0.266, 0.0007)
(0.267, 0.0007)
(0.268, 0.0008)
(0.269, 0.0008)
(0.270, 0.0008)
(0.271, 0.0008)
(0.272, 0.0008)
(0.273, 0.0009)
(0.274, 0.0009)
(0.275, 0.0009)
(0.276, 0.0009)
(0.277, 0.0010)
(0.278, 0.0010)
(0.279, 0.0010)
(0.280, 0.0010)
(0.281, 0.0011)
(0.282, 0.0011)
(0.283, 0.0011)
(0.284, 0.0011)
(0.285, 0.0012)
(0.286, 0.0012)
(0.287, 0.0012)
(0.288, 0.0013)
(0.289, 0.0013)
(0.290, 0.0013)
(0.291, 0.0014)
(0.292, 0.0014)
(0.293, 0.0014)
(0.294, 0.0015)
(0.295, 0.0015)
(0.296, 0.0015)
(0.297, 0.0016)
(0.298, 0.0016)
(0.299, 0.0016)
(0.300, 0.0017)
(0.301, 0.0017)
(0.302, 0.0018)
(0.303, 0.0018)
(0.304, 0.0018)
(0.305, 0.0019)
(0.306, 0.0019)
(0.307, 0.0020)
(0.308, 0.0020)
(0.309, 0.0021)
(0.310, 0.0021)
(0.311, 0.0021)
(0.312, 0.0022)
(0.313, 0.0022)
(0.314, 0.0023)
(0.315, 0.0023)
(0.316, 0.0024)
(0.317, 0.0025)
(0.318, 0.0025)
(0.319, 0.0026)
(0.320, 0.0026)
(0.321, 0.0027)
(0.322, 0.0027)
(0.323, 0.0028)
(0.324, 0.0029)
(0.325, 0.0029)
(0.326, 0.0030)
(0.327, 0.0030)
(0.328, 0.0031)
(0.329, 0.0032)
(0.330, 0.0032)
(0.331, 0.0033)
(0.332, 0.0034)
(0.333, 0.0035)
(0.334, 0.0035)
(0.335, 0.0036)
(0.336, 0.0037)
(0.337, 0.0038)
(0.338, 0.0038)
(0.339, 0.0039)
(0.340, 0.0040)
(0.341, 0.0041)
(0.342, 0.0042)
(0.343, 0.0042)
(0.344, 0.0043)
(0.345, 0.0044)
(0.346, 0.0045)
(0.347, 0.0046)
(0.348, 0.0047)
(0.349, 0.0048)
(0.350, 0.0049)
(0.351, 0.0050)
(0.352, 0.0051)
(0.353, 0.0052)
(0.354, 0.0053)
(0.355, 0.0054)
(0.356, 0.0055)
(0.357, 0.0056)
(0.358, 0.0057)
(0.359, 0.0058)
(0.360, 0.0059)
(0.361, 0.0060)
(0.362, 0.0062)
(0.363, 0.0063)
(0.364, 0.0064)
(0.365, 0.0065)
(0.366, 0.0067)
(0.367, 0.0068)
(0.368, 0.0069)
(0.369, 0.0070)
(0.370, 0.0072)
(0.371, 0.0073)
(0.372, 0.0074)
(0.373, 0.0076)
(0.374, 0.0077)
(0.375, 0.0079)
(0.376, 0.0080)
(0.377, 0.0082)
(0.378, 0.0083)
(0.379, 0.0085)
(0.380, 0.0086)
(0.381, 0.0088)
(0.382, 0.0090)
(0.383, 0.0091)
(0.384, 0.0093)
(0.385, 0.0095)
(0.386, 0.0096)
(0.387, 0.0098)
(0.388, 0.0100)
(0.389, 0.0102)
(0.390, 0.0103)
(0.391, 0.0105)
(0.392, 0.0107)
(0.393, 0.0109)
(0.394, 0.0111)
(0.395, 0.0113)
(0.396, 0.0115)
(0.397, 0.0117)
(0.398, 0.0119)
(0.399, 0.0121)
(0.400, 0.0123)
(0.401, 0.0125)
(0.402, 0.0128)
(0.403, 0.0130)
(0.404, 0.0132)
(0.405, 0.0134)
(0.406, 0.0137)
(0.407, 0.0139)
(0.408, 0.0141)
(0.409, 0.0144)
(0.410, 0.0146)
(0.411, 0.0149)
(0.412, 0.0151)
(0.413, 0.0154)
(0.414, 0.0156)
(0.415, 0.0159)
(0.416, 0.0162)
(0.417, 0.0165)
(0.418, 0.0167)
(0.419, 0.0170)
(0.420, 0.0173)
(0.421, 0.0176)
(0.422, 0.0179)
(0.423, 0.0182)
(0.424, 0.0185)
(0.425, 0.0188)
(0.426, 0.0191)
(0.427, 0.0194)
(0.428, 0.0197)
(0.429, 0.0200)
(0.430, 0.0204)
(0.431, 0.0207)
(0.432, 0.0210)
(0.433, 0.0214)
(0.434, 0.0217)
(0.435, 0.0221)
(0.436, 0.0224)
(0.437, 0.0228)
(0.438, 0.0231)
(0.439, 0.0235)
(0.440, 0.0239)
(0.441, 0.0243)
(0.442, 0.0246)
(0.443, 0.0250)
(0.444, 0.0254)
(0.445, 0.0258)
(0.446, 0.0262)
(0.447, 0.0266)
(0.448, 0.0271)
(0.449, 0.0275)
(0.450, 0.0279)
(0.451, 0.0283)
(0.452, 0.0288)
(0.453, 0.0292)
(0.454, 0.0297)
(0.455, 0.0301)
(0.456, 0.0306)
(0.457, 0.0311)
(0.458, 0.0315)
(0.459, 0.0320)
(0.460, 0.0325)
(0.461, 0.0330)
(0.462, 0.0335)
(0.463, 0.0340)
(0.464, 0.0345)
(0.465, 0.0350)
(0.466, 0.0356)
(0.467, 0.0361)
(0.468, 0.0366)
(0.469, 0.0372)
(0.470, 0.0377)
(0.471, 0.0383)
(0.472, 0.0389)
(0.473, 0.0394)
(0.474, 0.0400)
(0.475, 0.0406)
(0.476, 0.0412)
(0.477, 0.0418)
(0.478, 0.0424)
(0.479, 0.0431)
(0.480, 0.0437)
(0.481, 0.0443)
(0.482, 0.0450)
(0.483, 0.0456)
(0.484, 0.0463)
(0.485, 0.0469)
(0.486, 0.0476)
(0.487, 0.0483)
(0.488, 0.0490)
(0.489, 0.0497)
(0.490, 0.0504)
(0.491, 0.0511)
(0.492, 0.0518)
(0.493, 0.0526)
(0.494, 0.0533)
(0.495, 0.0541)
(0.496, 0.0548)
(0.497, 0.0556)
(0.498, 0.0564)
(0.499, 0.0572)
(0.500, 0.0580)
(0.501, 0.0588)
(0.502, 0.0596)
(0.503, 0.0604)
(0.504, 0.0613)
(0.505, 0.0621)
(0.506, 0.0630)
(0.507, 0.0639)
(0.508, 0.0647)
(0.509, 0.0656)
(0.510, 0.0665)
(0.511, 0.0674)
(0.512, 0.0684)
(0.513, 0.0693)
(0.514, 0.0702)
(0.515, 0.0712)
(0.516, 0.0722)
(0.517, 0.0731)
(0.518, 0.0741)
(0.519, 0.0751)
(0.520, 0.0761)
(0.521, 0.0772)
(0.522, 0.0782)
(0.523, 0.0792)
(0.524, 0.0803)
(0.525, 0.0814)
(0.526, 0.0824)
(0.527, 0.0835)
(0.528, 0.0846)
(0.529, 0.0858)
(0.530, 0.0869)
(0.531, 0.0880)
(0.532, 0.0892)
(0.533, 0.0904)
(0.534, 0.0915)
(0.535, 0.0927)
(0.536, 0.0940)
(0.537, 0.0952)
(0.538, 0.0964)
(0.539, 0.0977)
(0.540, 0.0989)
(0.541, 0.1002)
(0.542, 0.1015)
(0.543, 0.1028)
(0.544, 0.1041)
(0.545, 0.1055)
(0.546, 0.1068)
(0.547, 0.1082)
(0.548, 0.1096)
(0.549, 0.1110)
(0.550, 0.1124)
(0.551, 0.1138)
(0.552, 0.1152)
(0.553, 0.1167)
(0.554, 0.1182)
(0.555, 0.1196)
(0.556, 0.1212)
(0.557, 0.1227)
(0.558, 0.1242)
(0.559, 0.1258)
(0.560, 0.1273)
(0.561, 0.1289)
(0.562, 0.1305)
(0.563, 0.1321)
(0.564, 0.1338)
(0.565, 0.1354)
(0.566, 0.1371)
(0.567, 0.1388)
(0.568, 0.1405)
(0.569, 0.1422)
(0.570, 0.1440)
(0.571, 0.1457)
(0.572, 0.1475)
(0.573, 0.1493)
(0.574, 0.1511)
(0.575, 0.1530)
(0.576, 0.1548)
(0.577, 0.1567)
(0.578, 0.1586)
(0.579, 0.1605)
(0.580, 0.1624)
(0.581, 0.1644)
(0.582, 0.1664)
(0.583, 0.1684)
(0.584, 0.1704)
(0.585, 0.1724)
(0.586, 0.1745)
(0.587, 0.1765)
(0.588, 0.1786)
(0.589, 0.1808)
(0.590, 0.1829)
(0.591, 0.1851)
(0.592, 0.1873)
(0.593, 0.1895)
(0.594, 0.1917)
(0.595, 0.1939)
(0.596, 0.1962)
(0.597, 0.1985)
(0.598, 0.2008)
(0.599, 0.2032)
(0.600, 0.2055)
(0.601, 0.2079)
(0.602, 0.2103)
(0.603, 0.2128)
(0.604, 0.2152)
(0.605, 0.2177)
(0.606, 0.2202)
(0.607, 0.2228)
(0.608, 0.2253)
(0.609, 0.2279)
(0.610, 0.2305)
(0.611, 0.2332)
(0.612, 0.2358)
(0.613, 0.2385)
(0.614, 0.2412)
(0.615, 0.2440)
(0.616, 0.2467)
(0.617, 0.2495)
(0.618, 0.2523)
(0.619, 0.2552)
(0.620, 0.2581)
(0.621, 0.2610)
(0.622, 0.2639)
(0.623, 0.2668)
(0.624, 0.2698)
(0.625, 0.2728)
(0.625, 0.0000)}
\defdata{emaprofile-6-short}{
(0.000, 0.0000)
(0.430, 0.0000)
(0.431, 0.0000)
(0.432, 0.0000)
(0.433, 0.0000)
(0.434, 0.0000)
(0.435, 0.0000)
(0.436, 0.0001)
(0.437, 0.0001)
(0.438, 0.0001)
(0.439, 0.0001)
(0.440, 0.0001)
(0.441, 0.0001)
(0.442, 0.0001)
(0.443, 0.0001)
(0.444, 0.0001)
(0.445, 0.0001)
(0.446, 0.0001)
(0.447, 0.0001)
(0.448, 0.0001)
(0.449, 0.0001)
(0.450, 0.0001)
(0.451, 0.0001)
(0.452, 0.0001)
(0.453, 0.0001)
(0.454, 0.0001)
(0.455, 0.0001)
(0.456, 0.0001)
(0.457, 0.0001)
(0.458, 0.0001)
(0.459, 0.0001)
(0.460, 0.0001)
(0.461, 0.0001)
(0.462, 0.0001)
(0.463, 0.0001)
(0.464, 0.0001)
(0.465, 0.0002)
(0.466, 0.0002)
(0.467, 0.0002)
(0.468, 0.0002)
(0.469, 0.0002)
(0.470, 0.0002)
(0.471, 0.0002)
(0.472, 0.0002)
(0.473, 0.0002)
(0.474, 0.0002)
(0.475, 0.0002)
(0.476, 0.0002)
(0.477, 0.0002)
(0.478, 0.0002)
(0.479, 0.0003)
(0.480, 0.0003)
(0.481, 0.0003)
(0.482, 0.0003)
(0.483, 0.0003)
(0.484, 0.0003)
(0.485, 0.0003)
(0.486, 0.0003)
(0.487, 0.0003)
(0.488, 0.0003)
(0.489, 0.0004)
(0.490, 0.0004)
(0.491, 0.0004)
(0.492, 0.0004)
(0.493, 0.0004)
(0.494, 0.0004)
(0.495, 0.0004)
(0.496, 0.0005)
(0.497, 0.0005)
(0.498, 0.0005)
(0.499, 0.0005)
(0.500, 0.0005)
(0.501, 0.0005)
(0.502, 0.0006)
(0.503, 0.0006)
(0.504, 0.0006)
(0.505, 0.0006)
(0.506, 0.0006)
(0.507, 0.0007)
(0.508, 0.0007)
(0.509, 0.0007)
(0.510, 0.0007)
(0.511, 0.0008)
(0.512, 0.0008)
(0.513, 0.0008)
(0.514, 0.0008)
(0.515, 0.0009)
(0.516, 0.0009)
(0.517, 0.0009)
(0.518, 0.0010)
(0.519, 0.0010)
(0.520, 0.0010)
(0.521, 0.0011)
(0.522, 0.0011)
(0.523, 0.0011)
(0.524, 0.0012)
(0.525, 0.0012)
(0.526, 0.0012)
(0.527, 0.0013)
(0.528, 0.0013)
(0.529, 0.0014)
(0.530, 0.0014)
(0.531, 0.0015)
(0.532, 0.0015)
(0.533, 0.0016)
(0.534, 0.0016)
(0.535, 0.0017)
(0.536, 0.0017)
(0.537, 0.0018)
(0.538, 0.0018)
(0.539, 0.0019)
(0.540, 0.0019)
(0.541, 0.0020)
(0.542, 0.0021)
(0.543, 0.0021)
(0.544, 0.0022)
(0.545, 0.0023)
(0.546, 0.0023)
(0.547, 0.0024)
(0.548, 0.0025)
(0.549, 0.0026)
(0.550, 0.0026)
(0.551, 0.0027)
(0.552, 0.0028)
(0.553, 0.0029)
(0.554, 0.0030)
(0.555, 0.0031)
(0.556, 0.0032)
(0.557, 0.0033)
(0.558, 0.0034)
(0.559, 0.0035)
(0.560, 0.0036)
(0.561, 0.0037)
(0.562, 0.0038)
(0.563, 0.0039)
(0.564, 0.0041)
(0.565, 0.0042)
(0.566, 0.0043)
(0.567, 0.0044)
(0.568, 0.0046)
(0.569, 0.0047)
(0.570, 0.0049)
(0.571, 0.0050)
(0.572, 0.0052)
(0.573, 0.0053)
(0.574, 0.0055)
(0.575, 0.0056)
(0.576, 0.0058)
(0.577, 0.0060)
(0.578, 0.0062)
(0.579, 0.0063)
(0.580, 0.0065)
(0.581, 0.0067)
(0.582, 0.0069)
(0.583, 0.0071)
(0.584, 0.0073)
(0.585, 0.0075)
(0.586, 0.0078)
(0.587, 0.0080)
(0.588, 0.0082)
(0.589, 0.0085)
(0.590, 0.0087)
(0.591, 0.0090)
(0.592, 0.0092)
(0.593, 0.0095)
(0.594, 0.0098)
(0.595, 0.0101)
(0.596, 0.0104)
(0.597, 0.0107)
(0.598, 0.0110)
(0.599, 0.0113)
(0.600, 0.0116)
(0.601, 0.0119)
(0.602, 0.0123)
(0.603, 0.0126)
(0.604, 0.0130)
(0.605, 0.0134)
(0.606, 0.0137)
(0.607, 0.0141)
(0.608, 0.0145)
(0.609, 0.0149)
(0.610, 0.0154)
(0.611, 0.0158)
(0.612, 0.0162)
(0.613, 0.0167)
(0.614, 0.0172)
(0.615, 0.0176)
(0.616, 0.0181)
(0.617, 0.0186)
(0.618, 0.0192)
(0.619, 0.0197)
(0.620, 0.0202)
(0.621, 0.0208)
(0.622, 0.0214)
(0.623, 0.0220)
(0.624, 0.0226)
(0.625, 0.0232)
(0.626, 0.0238)
(0.627, 0.0245)
(0.628, 0.0252)
(0.629, 0.0258)
(0.630, 0.0265)
(0.631, 0.0273)
(0.632, 0.0280)
(0.633, 0.0288)
(0.634, 0.0296)
(0.635, 0.0304)
(0.636, 0.0312)
(0.637, 0.0320)
(0.638, 0.0329)
(0.639, 0.0338)
(0.640, 0.0347)
(0.641, 0.0356)
(0.642, 0.0366)
(0.643, 0.0375)
(0.644, 0.0386)
(0.645, 0.0396)
(0.646, 0.0406)
(0.647, 0.0417)
(0.648, 0.0428)
(0.649, 0.0440)
(0.650, 0.0451)
(0.651, 0.0463)
(0.652, 0.0475)
(0.653, 0.0488)
(0.654, 0.0501)
(0.655, 0.0514)
(0.656, 0.0527)
(0.657, 0.0541)
(0.658, 0.0555)
(0.659, 0.0570)
(0.660, 0.0585)
(0.661, 0.0600)
(0.662, 0.0615)
(0.663, 0.0631)
(0.664, 0.0648)
(0.665, 0.0665)
(0.666, 0.0682)
(0.667, 0.0699)
(0.668, 0.0717)
(0.669, 0.0736)
(0.670, 0.0755)
(0.671, 0.0774)
(0.672, 0.0794)
(0.673, 0.0814)
(0.674, 0.0835)
(0.675, 0.0856)
(0.676, 0.0878)
(0.677, 0.0900)
(0.678, 0.0923)
(0.679, 0.0946)
(0.680, 0.0970)
(0.681, 0.0995)
(0.682, 0.1020)
(0.683, 0.1046)
(0.684, 0.1072)
(0.685, 0.1099)
(0.686, 0.1126)
(0.687, 0.1154)
(0.688, 0.1183)
(0.689, 0.1213)
(0.690, 0.1243)
(0.691, 0.1274)
(0.692, 0.1306)
(0.693, 0.1338)
(0.694, 0.1371)
(0.695, 0.1405)
(0.696, 0.1440)
(0.697, 0.1475)
(0.698, 0.1512)
(0.699, 0.1549)
(0.700, 0.1587)
(0.701, 0.1626)
(0.702, 0.1666)
(0.703, 0.1706)
(0.704, 0.1748)
(0.705, 0.1791)
(0.706, 0.1834)
(0.707, 0.1879)
(0.708, 0.1924)
(0.709, 0.1971)
(0.710, 0.2019)
(0.711, 0.2068)
(0.712, 0.2117)
(0.713, 0.2169)
(0.714, 0.2221)
(0.715, 0.2274)
(0.716, 0.2329)
(0.717, 0.2385)
(0.718, 0.2442)
(0.719, 0.2500)
(0.720, 0.2560)
(0.721, 0.2621)
(0.722, 0.2683)
(0.723, 0.2747)
(0.724, 0.2812)
(0.725, 0.2879)
(0.726, 0.2947)
(0.727, 0.3016)
(0.728, 0.3088)
(0.729, 0.3160)
(0.730, 0.3235)
(0.731, 0.3311)
(0.732, 0.3388)
(0.733, 0.3468)
(0.734, 0.3549)
(0.735, 0.3632)
(0.736, 0.3717)
(0.737, 0.3803)
(0.738, 0.3892)
(0.739, 0.3982)
(0.740, 0.4075)
(0.741, 0.4169)
(0.742, 0.4266)
(0.743, 0.4364)
(0.744, 0.4465)
(0.745, 0.4568)
(0.746, 0.4673)
(0.747, 0.4781)
(0.748, 0.4890)
(0.749, 0.5003)
(0.750, 0.5117)
(0.750, 0.0000)}
\defdata{emaprofile-6-long}{
(0.000, 0.0000)
(0.216, 0.0000)
(0.217, 0.0000)
(0.218, 0.0000)
(0.219, 0.0000)
(0.220, 0.0000)
(0.221, 0.0000)
(0.222, 0.0000)
(0.223, 0.0001)
(0.224, 0.0001)
(0.225, 0.0001)
(0.226, 0.0001)
(0.227, 0.0001)
(0.228, 0.0001)
(0.229, 0.0001)
(0.230, 0.0001)
(0.231, 0.0001)
(0.232, 0.0001)
(0.233, 0.0001)
(0.234, 0.0001)
(0.235, 0.0001)
(0.236, 0.0001)
(0.237, 0.0001)
(0.238, 0.0001)
(0.239, 0.0001)
(0.240, 0.0001)
(0.241, 0.0001)
(0.242, 0.0001)
(0.243, 0.0001)
(0.244, 0.0001)
(0.245, 0.0001)
(0.246, 0.0001)
(0.247, 0.0001)
(0.248, 0.0001)
(0.249, 0.0001)
(0.250, 0.0001)
(0.251, 0.0001)
(0.252, 0.0001)
(0.253, 0.0001)
(0.254, 0.0001)
(0.255, 0.0001)
(0.256, 0.0001)
(0.257, 0.0001)
(0.258, 0.0001)
(0.259, 0.0001)
(0.260, 0.0001)
(0.261, 0.0001)
(0.262, 0.0002)
(0.263, 0.0002)
(0.264, 0.0002)
(0.265, 0.0002)
(0.266, 0.0002)
(0.267, 0.0002)
(0.268, 0.0002)
(0.269, 0.0002)
(0.270, 0.0002)
(0.271, 0.0002)
(0.272, 0.0002)
(0.273, 0.0002)
(0.274, 0.0002)
(0.275, 0.0002)
(0.276, 0.0002)
(0.277, 0.0002)
(0.278, 0.0002)
(0.279, 0.0002)
(0.280, 0.0002)
(0.281, 0.0003)
(0.282, 0.0003)
(0.283, 0.0003)
(0.284, 0.0003)
(0.285, 0.0003)
(0.286, 0.0003)
(0.287, 0.0003)
(0.288, 0.0003)
(0.289, 0.0003)
(0.290, 0.0003)
(0.291, 0.0003)
(0.292, 0.0003)
(0.293, 0.0003)
(0.294, 0.0003)
(0.295, 0.0004)
(0.296, 0.0004)
(0.297, 0.0004)
(0.298, 0.0004)
(0.299, 0.0004)
(0.300, 0.0004)
(0.301, 0.0004)
(0.302, 0.0004)
(0.303, 0.0004)
(0.304, 0.0004)
(0.305, 0.0004)
(0.306, 0.0005)
(0.307, 0.0005)
(0.308, 0.0005)
(0.309, 0.0005)
(0.310, 0.0005)
(0.311, 0.0005)
(0.312, 0.0005)
(0.313, 0.0005)
(0.314, 0.0005)
(0.315, 0.0006)
(0.316, 0.0006)
(0.317, 0.0006)
(0.318, 0.0006)
(0.319, 0.0006)
(0.320, 0.0006)
(0.321, 0.0006)
(0.322, 0.0006)
(0.323, 0.0007)
(0.324, 0.0007)
(0.325, 0.0007)
(0.326, 0.0007)
(0.327, 0.0007)
(0.328, 0.0007)
(0.329, 0.0007)
(0.330, 0.0008)
(0.331, 0.0008)
(0.332, 0.0008)
(0.333, 0.0008)
(0.334, 0.0008)
(0.335, 0.0008)
(0.336, 0.0009)
(0.337, 0.0009)
(0.338, 0.0009)
(0.339, 0.0009)
(0.340, 0.0009)
(0.341, 0.0010)
(0.342, 0.0010)
(0.343, 0.0010)
(0.344, 0.0010)
(0.345, 0.0010)
(0.346, 0.0011)
(0.347, 0.0011)
(0.348, 0.0011)
(0.349, 0.0011)
(0.350, 0.0011)
(0.351, 0.0012)
(0.352, 0.0012)
(0.353, 0.0012)
(0.354, 0.0012)
(0.355, 0.0013)
(0.356, 0.0013)
(0.357, 0.0013)
(0.358, 0.0013)
(0.359, 0.0014)
(0.360, 0.0014)
(0.361, 0.0014)
(0.362, 0.0015)
(0.363, 0.0015)
(0.364, 0.0015)
(0.365, 0.0015)
(0.366, 0.0016)
(0.367, 0.0016)
(0.368, 0.0016)
(0.369, 0.0017)
(0.370, 0.0017)
(0.371, 0.0017)
(0.372, 0.0018)
(0.373, 0.0018)
(0.374, 0.0018)
(0.375, 0.0019)
(0.376, 0.0019)
(0.377, 0.0019)
(0.378, 0.0020)
(0.379, 0.0020)
(0.380, 0.0020)
(0.381, 0.0021)
(0.382, 0.0021)
(0.383, 0.0021)
(0.384, 0.0022)
(0.385, 0.0022)
(0.386, 0.0023)
(0.387, 0.0023)
(0.388, 0.0023)
(0.389, 0.0024)
(0.390, 0.0024)
(0.391, 0.0025)
(0.392, 0.0025)
(0.393, 0.0026)
(0.394, 0.0026)
(0.395, 0.0027)
(0.396, 0.0027)
(0.397, 0.0028)
(0.398, 0.0028)
(0.399, 0.0029)
(0.400, 0.0029)
(0.401, 0.0030)
(0.402, 0.0030)
(0.403, 0.0031)
(0.404, 0.0031)
(0.405, 0.0032)
(0.406, 0.0032)
(0.407, 0.0033)
(0.408, 0.0033)
(0.409, 0.0034)
(0.410, 0.0034)
(0.411, 0.0035)
(0.412, 0.0036)
(0.413, 0.0036)
(0.414, 0.0037)
(0.415, 0.0037)
(0.416, 0.0038)
(0.417, 0.0039)
(0.418, 0.0039)
(0.419, 0.0040)
(0.420, 0.0041)
(0.421, 0.0041)
(0.422, 0.0042)
(0.423, 0.0043)
(0.424, 0.0043)
(0.425, 0.0044)
(0.426, 0.0045)
(0.427, 0.0046)
(0.428, 0.0046)
(0.429, 0.0047)
(0.430, 0.0048)
(0.431, 0.0049)
(0.432, 0.0049)
(0.433, 0.0050)
(0.434, 0.0051)
(0.435, 0.0052)
(0.436, 0.0053)
(0.437, 0.0054)
(0.438, 0.0054)
(0.439, 0.0055)
(0.440, 0.0056)
(0.441, 0.0057)
(0.442, 0.0058)
(0.443, 0.0059)
(0.444, 0.0060)
(0.445, 0.0061)
(0.446, 0.0062)
(0.447, 0.0063)
(0.448, 0.0064)
(0.449, 0.0065)
(0.450, 0.0066)
(0.451, 0.0067)
(0.452, 0.0068)
(0.453, 0.0069)
(0.454, 0.0070)
(0.455, 0.0071)
(0.456, 0.0072)
(0.457, 0.0073)
(0.458, 0.0074)
(0.459, 0.0075)
(0.460, 0.0077)
(0.461, 0.0078)
(0.462, 0.0079)
(0.463, 0.0080)
(0.464, 0.0081)
(0.465, 0.0082)
(0.466, 0.0084)
(0.467, 0.0085)
(0.468, 0.0086)
(0.469, 0.0088)
(0.470, 0.0089)
(0.471, 0.0090)
(0.472, 0.0092)
(0.473, 0.0093)
(0.474, 0.0094)
(0.475, 0.0096)
(0.476, 0.0097)
(0.477, 0.0098)
(0.478, 0.0100)
(0.479, 0.0101)
(0.480, 0.0103)
(0.481, 0.0104)
(0.482, 0.0106)
(0.483, 0.0107)
(0.484, 0.0109)
(0.485, 0.0110)
(0.486, 0.0112)
(0.487, 0.0114)
(0.488, 0.0115)
(0.489, 0.0117)
(0.490, 0.0119)
(0.491, 0.0120)
(0.492, 0.0122)
(0.493, 0.0124)
(0.494, 0.0126)
(0.495, 0.0127)
(0.496, 0.0129)
(0.497, 0.0131)
(0.498, 0.0133)
(0.499, 0.0135)
(0.500, 0.0136)
(0.501, 0.0138)
(0.502, 0.0140)
(0.503, 0.0142)
(0.504, 0.0144)
(0.505, 0.0146)
(0.506, 0.0148)
(0.507, 0.0150)
(0.508, 0.0152)
(0.509, 0.0154)
(0.510, 0.0157)
(0.511, 0.0159)
(0.512, 0.0161)
(0.513, 0.0163)
(0.514, 0.0165)
(0.515, 0.0168)
(0.516, 0.0170)
(0.517, 0.0172)
(0.518, 0.0174)
(0.519, 0.0177)
(0.520, 0.0179)
(0.521, 0.0182)
(0.522, 0.0184)
(0.523, 0.0186)
(0.524, 0.0189)
(0.525, 0.0192)
(0.526, 0.0194)
(0.527, 0.0197)
(0.528, 0.0199)
(0.529, 0.0202)
(0.530, 0.0205)
(0.531, 0.0207)
(0.532, 0.0210)
(0.533, 0.0213)
(0.534, 0.0215)
(0.535, 0.0218)
(0.536, 0.0221)
(0.537, 0.0224)
(0.538, 0.0227)
(0.539, 0.0230)
(0.540, 0.0233)
(0.541, 0.0236)
(0.542, 0.0239)
(0.543, 0.0242)
(0.544, 0.0245)
(0.545, 0.0248)
(0.546, 0.0251)
(0.547, 0.0255)
(0.548, 0.0258)
(0.549, 0.0261)
(0.550, 0.0264)
(0.551, 0.0268)
(0.552, 0.0271)
(0.553, 0.0275)
(0.554, 0.0278)
(0.555, 0.0282)
(0.556, 0.0285)
(0.557, 0.0289)
(0.558, 0.0292)
(0.559, 0.0296)
(0.560, 0.0300)
(0.561, 0.0303)
(0.562, 0.0307)
(0.563, 0.0311)
(0.564, 0.0315)
(0.565, 0.0319)
(0.566, 0.0323)
(0.567, 0.0327)
(0.568, 0.0331)
(0.569, 0.0335)
(0.570, 0.0339)
(0.571, 0.0343)
(0.572, 0.0347)
(0.573, 0.0351)
(0.574, 0.0356)
(0.575, 0.0360)
(0.576, 0.0364)
(0.577, 0.0369)
(0.578, 0.0373)
(0.579, 0.0378)
(0.580, 0.0382)
(0.581, 0.0387)
(0.582, 0.0392)
(0.583, 0.0396)
(0.584, 0.0401)
(0.585, 0.0406)
(0.586, 0.0411)
(0.587, 0.0416)
(0.588, 0.0420)
(0.589, 0.0425)
(0.590, 0.0431)
(0.591, 0.0436)
(0.592, 0.0441)
(0.593, 0.0446)
(0.594, 0.0451)
(0.595, 0.0456)
(0.596, 0.0462)
(0.597, 0.0467)
(0.598, 0.0473)
(0.599, 0.0478)
(0.600, 0.0484)
(0.601, 0.0489)
(0.602, 0.0495)
(0.603, 0.0501)
(0.604, 0.0507)
(0.605, 0.0512)
(0.606, 0.0518)
(0.607, 0.0524)
(0.608, 0.0530)
(0.609, 0.0536)
(0.610, 0.0543)
(0.611, 0.0549)
(0.612, 0.0555)
(0.613, 0.0561)
(0.614, 0.0568)
(0.615, 0.0574)
(0.616, 0.0581)
(0.617, 0.0587)
(0.618, 0.0594)
(0.619, 0.0601)
(0.620, 0.0607)
(0.621, 0.0614)
(0.622, 0.0621)
(0.623, 0.0628)
(0.624, 0.0635)
(0.625, 0.0642)
(0.626, 0.0649)
(0.627, 0.0657)
(0.628, 0.0664)
(0.629, 0.0671)
(0.630, 0.0679)
(0.631, 0.0686)
(0.632, 0.0694)
(0.633, 0.0701)
(0.634, 0.0709)
(0.635, 0.0717)
(0.636, 0.0725)
(0.637, 0.0733)
(0.638, 0.0741)
(0.639, 0.0749)
(0.640, 0.0757)
(0.641, 0.0765)
(0.642, 0.0774)
(0.643, 0.0782)
(0.644, 0.0791)
(0.645, 0.0799)
(0.646, 0.0808)
(0.647, 0.0816)
(0.648, 0.0825)
(0.649, 0.0834)
(0.650, 0.0843)
(0.651, 0.0852)
(0.652, 0.0861)
(0.653, 0.0870)
(0.654, 0.0880)
(0.655, 0.0889)
(0.656, 0.0899)
(0.657, 0.0908)
(0.658, 0.0918)
(0.659, 0.0928)
(0.660, 0.0937)
(0.661, 0.0947)
(0.662, 0.0957)
(0.663, 0.0967)
(0.664, 0.0977)
(0.665, 0.0988)
(0.666, 0.0998)
(0.667, 0.1009)
(0.668, 0.1019)
(0.669, 0.1030)
(0.670, 0.1040)
(0.671, 0.1051)
(0.672, 0.1062)
(0.673, 0.1073)
(0.674, 0.1084)
(0.675, 0.1096)
(0.676, 0.1107)
(0.677, 0.1118)
(0.678, 0.1130)
(0.679, 0.1141)
(0.680, 0.1153)
(0.681, 0.1165)
(0.682, 0.1177)
(0.683, 0.1189)
(0.684, 0.1201)
(0.685, 0.1213)
(0.686, 0.1226)
(0.687, 0.1238)
(0.688, 0.1251)
(0.689, 0.1263)
(0.690, 0.1276)
(0.691, 0.1289)
(0.692, 0.1302)
(0.693, 0.1315)
(0.694, 0.1328)
(0.695, 0.1342)
(0.696, 0.1355)
(0.697, 0.1369)
(0.698, 0.1382)
(0.699, 0.1396)
(0.700, 0.1410)
(0.701, 0.1424)
(0.702, 0.1438)
(0.703, 0.1453)
(0.704, 0.1467)
(0.705, 0.1482)
(0.706, 0.1496)
(0.707, 0.1511)
(0.708, 0.1526)
(0.709, 0.1541)
(0.710, 0.1556)
(0.711, 0.1571)
(0.712, 0.1587)
(0.713, 0.1602)
(0.714, 0.1618)
(0.715, 0.1634)
(0.716, 0.1650)
(0.717, 0.1666)
(0.718, 0.1682)
(0.719, 0.1698)
(0.720, 0.1715)
(0.721, 0.1731)
(0.722, 0.1748)
(0.723, 0.1765)
(0.724, 0.1782)
(0.725, 0.1799)
(0.726, 0.1816)
(0.727, 0.1834)
(0.728, 0.1851)
(0.729, 0.1869)
(0.730, 0.1887)
(0.731, 0.1905)
(0.732, 0.1923)
(0.733, 0.1941)
(0.734, 0.1960)
(0.735, 0.1978)
(0.736, 0.1997)
(0.737, 0.2016)
(0.738, 0.2035)
(0.739, 0.2054)
(0.740, 0.2074)
(0.741, 0.2093)
(0.742, 0.2113)
(0.743, 0.2133)
(0.744, 0.2153)
(0.745, 0.2173)
(0.746, 0.2193)
(0.747, 0.2214)
(0.748, 0.2234)
(0.749, 0.2255)
(0.750, 0.2276)
(0.750, 0.0000)}
\defdata{emaprofile-7-short}{
(0.000, 0.0000)
(0.506, 0.0000)
(0.507, 0.0000)
(0.508, 0.0000)
(0.509, 0.0000)
(0.510, 0.0000)
(0.511, 0.0000)
(0.512, 0.0000)
(0.513, 0.0001)
(0.514, 0.0001)
(0.515, 0.0001)
(0.516, 0.0001)
(0.517, 0.0001)
(0.518, 0.0001)
(0.519, 0.0001)
(0.520, 0.0001)
(0.521, 0.0001)
(0.522, 0.0001)
(0.523, 0.0001)
(0.524, 0.0001)
(0.525, 0.0001)
(0.526, 0.0001)
(0.527, 0.0001)
(0.528, 0.0001)
(0.529, 0.0001)
(0.530, 0.0001)
(0.531, 0.0001)
(0.532, 0.0001)
(0.533, 0.0001)
(0.534, 0.0001)
(0.535, 0.0001)
(0.536, 0.0001)
(0.537, 0.0001)
(0.538, 0.0001)
(0.539, 0.0001)
(0.540, 0.0001)
(0.541, 0.0001)
(0.542, 0.0001)
(0.543, 0.0001)
(0.544, 0.0001)
(0.545, 0.0001)
(0.546, 0.0001)
(0.547, 0.0002)
(0.548, 0.0002)
(0.549, 0.0002)
(0.550, 0.0002)
(0.551, 0.0002)
(0.552, 0.0002)
(0.553, 0.0002)
(0.554, 0.0002)
(0.555, 0.0002)
(0.556, 0.0002)
(0.557, 0.0002)
(0.558, 0.0002)
(0.559, 0.0002)
(0.560, 0.0002)
(0.561, 0.0002)
(0.562, 0.0002)
(0.563, 0.0002)
(0.564, 0.0003)
(0.565, 0.0003)
(0.566, 0.0003)
(0.567, 0.0003)
(0.568, 0.0003)
(0.569, 0.0003)
(0.570, 0.0003)
(0.571, 0.0003)
(0.572, 0.0003)
(0.573, 0.0003)
(0.574, 0.0003)
(0.575, 0.0004)
(0.576, 0.0004)
(0.577, 0.0004)
(0.578, 0.0004)
(0.579, 0.0004)
(0.580, 0.0004)
(0.581, 0.0004)
(0.582, 0.0004)
(0.583, 0.0004)
(0.584, 0.0005)
(0.585, 0.0005)
(0.586, 0.0005)
(0.587, 0.0005)
(0.588, 0.0005)
(0.589, 0.0005)
(0.590, 0.0005)
(0.591, 0.0006)
(0.592, 0.0006)
(0.593, 0.0006)
(0.594, 0.0006)
(0.595, 0.0006)
(0.596, 0.0006)
(0.597, 0.0007)
(0.598, 0.0007)
(0.599, 0.0007)
(0.600, 0.0007)
(0.601, 0.0007)
(0.602, 0.0008)
(0.603, 0.0008)
(0.604, 0.0008)
(0.605, 0.0008)
(0.606, 0.0009)
(0.607, 0.0009)
(0.608, 0.0009)
(0.609, 0.0009)
(0.610, 0.0010)
(0.611, 0.0010)
(0.612, 0.0010)
(0.613, 0.0010)
(0.614, 0.0011)
(0.615, 0.0011)
(0.616, 0.0011)
(0.617, 0.0012)
(0.618, 0.0012)
(0.619, 0.0012)
(0.620, 0.0013)
(0.621, 0.0013)
(0.622, 0.0013)
(0.623, 0.0014)
(0.624, 0.0014)
(0.625, 0.0015)
(0.626, 0.0015)
(0.627, 0.0015)
(0.628, 0.0016)
(0.629, 0.0016)
(0.630, 0.0017)
(0.631, 0.0017)
(0.632, 0.0018)
(0.633, 0.0018)
(0.634, 0.0019)
(0.635, 0.0019)
(0.636, 0.0020)
(0.637, 0.0020)
(0.638, 0.0021)
(0.639, 0.0021)
(0.640, 0.0022)
(0.641, 0.0022)
(0.642, 0.0023)
(0.643, 0.0024)
(0.644, 0.0024)
(0.645, 0.0025)
(0.646, 0.0026)
(0.647, 0.0026)
(0.648, 0.0027)
(0.649, 0.0028)
(0.650, 0.0028)
(0.651, 0.0029)
(0.652, 0.0030)
(0.653, 0.0031)
(0.654, 0.0031)
(0.655, 0.0032)
(0.656, 0.0033)
(0.657, 0.0034)
(0.658, 0.0035)
(0.659, 0.0036)
(0.660, 0.0037)
(0.661, 0.0038)
(0.662, 0.0039)
(0.663, 0.0040)
(0.664, 0.0041)
(0.665, 0.0042)
(0.666, 0.0043)
(0.667, 0.0044)
(0.668, 0.0045)
(0.669, 0.0046)
(0.670, 0.0047)
(0.671, 0.0049)
(0.672, 0.0050)
(0.673, 0.0051)
(0.674, 0.0052)
(0.675, 0.0054)
(0.676, 0.0055)
(0.677, 0.0056)
(0.678, 0.0058)
(0.679, 0.0059)
(0.680, 0.0061)
(0.681, 0.0062)
(0.682, 0.0064)
(0.683, 0.0066)
(0.684, 0.0067)
(0.685, 0.0069)
(0.686, 0.0071)
(0.687, 0.0072)
(0.688, 0.0074)
(0.689, 0.0076)
(0.690, 0.0078)
(0.691, 0.0080)
(0.692, 0.0082)
(0.693, 0.0084)
(0.694, 0.0086)
(0.695, 0.0088)
(0.696, 0.0090)
(0.697, 0.0093)
(0.698, 0.0095)
(0.699, 0.0097)
(0.700, 0.0100)
(0.701, 0.0102)
(0.702, 0.0105)
(0.703, 0.0107)
(0.704, 0.0110)
(0.705, 0.0112)
(0.706, 0.0115)
(0.707, 0.0118)
(0.708, 0.0121)
(0.709, 0.0124)
(0.710, 0.0127)
(0.711, 0.0130)
(0.712, 0.0133)
(0.713, 0.0136)
(0.714, 0.0139)
(0.715, 0.0143)
(0.716, 0.0146)
(0.717, 0.0150)
(0.718, 0.0153)
(0.719, 0.0157)
(0.720, 0.0161)
(0.721, 0.0164)
(0.722, 0.0168)
(0.723, 0.0172)
(0.724, 0.0176)
(0.725, 0.0181)
(0.726, 0.0185)
(0.727, 0.0189)
(0.728, 0.0194)
(0.729, 0.0198)
(0.730, 0.0203)
(0.731, 0.0208)
(0.732, 0.0213)
(0.733, 0.0218)
(0.734, 0.0223)
(0.735, 0.0228)
(0.736, 0.0233)
(0.737, 0.0239)
(0.738, 0.0244)
(0.739, 0.0250)
(0.740, 0.0256)
(0.741, 0.0262)
(0.742, 0.0268)
(0.743, 0.0274)
(0.744, 0.0280)
(0.745, 0.0287)
(0.746, 0.0293)
(0.747, 0.0300)
(0.748, 0.0307)
(0.749, 0.0314)
(0.750, 0.0321)
(0.751, 0.0329)
(0.752, 0.0336)
(0.753, 0.0344)
(0.754, 0.0352)
(0.755, 0.0360)
(0.756, 0.0368)
(0.757, 0.0376)
(0.758, 0.0385)
(0.759, 0.0393)
(0.760, 0.0402)
(0.761, 0.0411)
(0.762, 0.0420)
(0.763, 0.0430)
(0.764, 0.0440)
(0.765, 0.0449)
(0.766, 0.0460)
(0.767, 0.0470)
(0.768, 0.0480)
(0.769, 0.0491)
(0.770, 0.0502)
(0.771, 0.0513)
(0.772, 0.0525)
(0.773, 0.0536)
(0.774, 0.0548)
(0.775, 0.0560)
(0.776, 0.0573)
(0.777, 0.0585)
(0.778, 0.0598)
(0.779, 0.0611)
(0.780, 0.0625)
(0.781, 0.0639)
(0.782, 0.0653)
(0.783, 0.0667)
(0.784, 0.0682)
(0.785, 0.0696)
(0.786, 0.0712)
(0.787, 0.0727)
(0.788, 0.0743)
(0.789, 0.0759)
(0.790, 0.0776)
(0.791, 0.0792)
(0.792, 0.0810)
(0.793, 0.0827)
(0.794, 0.0845)
(0.795, 0.0863)
(0.796, 0.0882)
(0.797, 0.0901)
(0.798, 0.0920)
(0.799, 0.0940)
(0.800, 0.0960)
(0.801, 0.0981)
(0.802, 0.1002)
(0.803, 0.1023)
(0.804, 0.1045)
(0.805, 0.1067)
(0.806, 0.1090)
(0.807, 0.1113)
(0.808, 0.1137)
(0.809, 0.1161)
(0.810, 0.1186)
(0.811, 0.1211)
(0.812, 0.1236)
(0.813, 0.1262)
(0.814, 0.1289)
(0.815, 0.1316)
(0.816, 0.1344)
(0.817, 0.1372)
(0.818, 0.1401)
(0.819, 0.1430)
(0.820, 0.1460)
(0.821, 0.1491)
(0.822, 0.1522)
(0.823, 0.1553)
(0.824, 0.1586)
(0.825, 0.1619)
(0.826, 0.1652)
(0.827, 0.1687)
(0.828, 0.1722)
(0.829, 0.1757)
(0.830, 0.1793)
(0.831, 0.1831)
(0.832, 0.1868)
(0.833, 0.1907)
(0.834, 0.1946)
(0.835, 0.1986)
(0.836, 0.2027)
(0.837, 0.2068)
(0.838, 0.2111)
(0.839, 0.2154)
(0.840, 0.2198)
(0.841, 0.2242)
(0.842, 0.2288)
(0.843, 0.2335)
(0.844, 0.2382)
(0.845, 0.2431)
(0.846, 0.2480)
(0.847, 0.2530)
(0.848, 0.2581)
(0.849, 0.2633)
(0.850, 0.2686)
(0.851, 0.2741)
(0.852, 0.2796)
(0.853, 0.2852)
(0.854, 0.2909)
(0.855, 0.2968)
(0.856, 0.3027)
(0.857, 0.3088)
(0.858, 0.3149)
(0.859, 0.3212)
(0.860, 0.3276)
(0.861, 0.3342)
(0.862, 0.3408)
(0.863, 0.3476)
(0.864, 0.3545)
(0.865, 0.3615)
(0.866, 0.3687)
(0.867, 0.3759)
(0.868, 0.3834)
(0.869, 0.3909)
(0.870, 0.3986)
(0.871, 0.4065)
(0.872, 0.4145)
(0.873, 0.4226)
(0.874, 0.4309)
(0.875, 0.4394)
(0.875, 0.0000)}
\defdata{emaprofile-7-long}{
(0.000, 0.0000)
(0.258, 0.0000)
(0.259, 0.0000)
(0.260, 0.0000)
(0.261, 0.0000)
(0.262, 0.0000)
(0.263, 0.0000)
(0.264, 0.0000)
(0.265, 0.0000)
(0.266, 0.0001)
(0.267, 0.0001)
(0.268, 0.0001)
(0.269, 0.0001)
(0.270, 0.0001)
(0.271, 0.0001)
(0.272, 0.0001)
(0.273, 0.0001)
(0.274, 0.0001)
(0.275, 0.0001)
(0.276, 0.0001)
(0.277, 0.0001)
(0.278, 0.0001)
(0.279, 0.0001)
(0.280, 0.0001)
(0.281, 0.0001)
(0.282, 0.0001)
(0.283, 0.0001)
(0.284, 0.0001)
(0.285, 0.0001)
(0.286, 0.0001)
(0.287, 0.0001)
(0.288, 0.0001)
(0.289, 0.0001)
(0.290, 0.0001)
(0.291, 0.0001)
(0.292, 0.0001)
(0.293, 0.0001)
(0.294, 0.0001)
(0.295, 0.0001)
(0.296, 0.0001)
(0.297, 0.0001)
(0.298, 0.0001)
(0.299, 0.0001)
(0.300, 0.0001)
(0.301, 0.0001)
(0.302, 0.0001)
(0.303, 0.0001)
(0.304, 0.0001)
(0.305, 0.0001)
(0.306, 0.0001)
(0.307, 0.0001)
(0.308, 0.0001)
(0.309, 0.0001)
(0.310, 0.0001)
(0.311, 0.0001)
(0.312, 0.0002)
(0.313, 0.0002)
(0.314, 0.0002)
(0.315, 0.0002)
(0.316, 0.0002)
(0.317, 0.0002)
(0.318, 0.0002)
(0.319, 0.0002)
(0.320, 0.0002)
(0.321, 0.0002)
(0.322, 0.0002)
(0.323, 0.0002)
(0.324, 0.0002)
(0.325, 0.0002)
(0.326, 0.0002)
(0.327, 0.0002)
(0.328, 0.0002)
(0.329, 0.0002)
(0.330, 0.0002)
(0.331, 0.0002)
(0.332, 0.0002)
(0.333, 0.0002)
(0.334, 0.0002)
(0.335, 0.0002)
(0.336, 0.0003)
(0.337, 0.0003)
(0.338, 0.0003)
(0.339, 0.0003)
(0.340, 0.0003)
(0.341, 0.0003)
(0.342, 0.0003)
(0.343, 0.0003)
(0.344, 0.0003)
(0.345, 0.0003)
(0.346, 0.0003)
(0.347, 0.0003)
(0.348, 0.0003)
(0.349, 0.0003)
(0.350, 0.0003)
(0.351, 0.0003)
(0.352, 0.0004)
(0.353, 0.0004)
(0.354, 0.0004)
(0.355, 0.0004)
(0.356, 0.0004)
(0.357, 0.0004)
(0.358, 0.0004)
(0.359, 0.0004)
(0.360, 0.0004)
(0.361, 0.0004)
(0.362, 0.0004)
(0.363, 0.0004)
(0.364, 0.0004)
(0.365, 0.0005)
(0.366, 0.0005)
(0.367, 0.0005)
(0.368, 0.0005)
(0.369, 0.0005)
(0.370, 0.0005)
(0.371, 0.0005)
(0.372, 0.0005)
(0.373, 0.0005)
(0.374, 0.0005)
(0.375, 0.0005)
(0.376, 0.0006)
(0.377, 0.0006)
(0.378, 0.0006)
(0.379, 0.0006)
(0.380, 0.0006)
(0.381, 0.0006)
(0.382, 0.0006)
(0.383, 0.0006)
(0.384, 0.0006)
(0.385, 0.0007)
(0.386, 0.0007)
(0.387, 0.0007)
(0.388, 0.0007)
(0.389, 0.0007)
(0.390, 0.0007)
(0.391, 0.0007)
(0.392, 0.0007)
(0.393, 0.0008)
(0.394, 0.0008)
(0.395, 0.0008)
(0.396, 0.0008)
(0.397, 0.0008)
(0.398, 0.0008)
(0.399, 0.0008)
(0.400, 0.0009)
(0.401, 0.0009)
(0.402, 0.0009)
(0.403, 0.0009)
(0.404, 0.0009)
(0.405, 0.0009)
(0.406, 0.0009)
(0.407, 0.0010)
(0.408, 0.0010)
(0.409, 0.0010)
(0.410, 0.0010)
(0.411, 0.0010)
(0.412, 0.0010)
(0.413, 0.0011)
(0.414, 0.0011)
(0.415, 0.0011)
(0.416, 0.0011)
(0.417, 0.0011)
(0.418, 0.0012)
(0.419, 0.0012)
(0.420, 0.0012)
(0.421, 0.0012)
(0.422, 0.0012)
(0.423, 0.0013)
(0.424, 0.0013)
(0.425, 0.0013)
(0.426, 0.0013)
(0.427, 0.0013)
(0.428, 0.0014)
(0.429, 0.0014)
(0.430, 0.0014)
(0.431, 0.0014)
(0.432, 0.0015)
(0.433, 0.0015)
(0.434, 0.0015)
(0.435, 0.0015)
(0.436, 0.0016)
(0.437, 0.0016)
(0.438, 0.0016)
(0.439, 0.0016)
(0.440, 0.0017)
(0.441, 0.0017)
(0.442, 0.0017)
(0.443, 0.0017)
(0.444, 0.0018)
(0.445, 0.0018)
(0.446, 0.0018)
(0.447, 0.0018)
(0.448, 0.0019)
(0.449, 0.0019)
(0.450, 0.0019)
(0.451, 0.0020)
(0.452, 0.0020)
(0.453, 0.0020)
(0.454, 0.0021)
(0.455, 0.0021)
(0.456, 0.0021)
(0.457, 0.0022)
(0.458, 0.0022)
(0.459, 0.0022)
(0.460, 0.0023)
(0.461, 0.0023)
(0.462, 0.0023)
(0.463, 0.0024)
(0.464, 0.0024)
(0.465, 0.0024)
(0.466, 0.0025)
(0.467, 0.0025)
(0.468, 0.0025)
(0.469, 0.0026)
(0.470, 0.0026)
(0.471, 0.0027)
(0.472, 0.0027)
(0.473, 0.0027)
(0.474, 0.0028)
(0.475, 0.0028)
(0.476, 0.0029)
(0.477, 0.0029)
(0.478, 0.0029)
(0.479, 0.0030)
(0.480, 0.0030)
(0.481, 0.0031)
(0.482, 0.0031)
(0.483, 0.0032)
(0.484, 0.0032)
(0.485, 0.0033)
(0.486, 0.0033)
(0.487, 0.0033)
(0.488, 0.0034)
(0.489, 0.0034)
(0.490, 0.0035)
(0.491, 0.0035)
(0.492, 0.0036)
(0.493, 0.0036)
(0.494, 0.0037)
(0.495, 0.0037)
(0.496, 0.0038)
(0.497, 0.0039)
(0.498, 0.0039)
(0.499, 0.0040)
(0.500, 0.0040)
(0.501, 0.0041)
(0.502, 0.0041)
(0.503, 0.0042)
(0.504, 0.0042)
(0.505, 0.0043)
(0.506, 0.0044)
(0.507, 0.0044)
(0.508, 0.0045)
(0.509, 0.0045)
(0.510, 0.0046)
(0.511, 0.0047)
(0.512, 0.0047)
(0.513, 0.0048)
(0.514, 0.0049)
(0.515, 0.0049)
(0.516, 0.0050)
(0.517, 0.0051)
(0.518, 0.0051)
(0.519, 0.0052)
(0.520, 0.0053)
(0.521, 0.0053)
(0.522, 0.0054)
(0.523, 0.0055)
(0.524, 0.0056)
(0.525, 0.0056)
(0.526, 0.0057)
(0.527, 0.0058)
(0.528, 0.0059)
(0.529, 0.0059)
(0.530, 0.0060)
(0.531, 0.0061)
(0.532, 0.0062)
(0.533, 0.0063)
(0.534, 0.0063)
(0.535, 0.0064)
(0.536, 0.0065)
(0.537, 0.0066)
(0.538, 0.0067)
(0.539, 0.0068)
(0.540, 0.0069)
(0.541, 0.0069)
(0.542, 0.0070)
(0.543, 0.0071)
(0.544, 0.0072)
(0.545, 0.0073)
(0.546, 0.0074)
(0.547, 0.0075)
(0.548, 0.0076)
(0.549, 0.0077)
(0.550, 0.0078)
(0.551, 0.0079)
(0.552, 0.0080)
(0.553, 0.0081)
(0.554, 0.0082)
(0.555, 0.0083)
(0.556, 0.0084)
(0.557, 0.0085)
(0.558, 0.0086)
(0.559, 0.0087)
(0.560, 0.0088)
(0.561, 0.0089)
(0.562, 0.0090)
(0.563, 0.0092)
(0.564, 0.0093)
(0.565, 0.0094)
(0.566, 0.0095)
(0.567, 0.0096)
(0.568, 0.0097)
(0.569, 0.0099)
(0.570, 0.0100)
(0.571, 0.0101)
(0.572, 0.0102)
(0.573, 0.0103)
(0.574, 0.0105)
(0.575, 0.0106)
(0.576, 0.0107)
(0.577, 0.0109)
(0.578, 0.0110)
(0.579, 0.0111)
(0.580, 0.0113)
(0.581, 0.0114)
(0.582, 0.0115)
(0.583, 0.0117)
(0.584, 0.0118)
(0.585, 0.0119)
(0.586, 0.0121)
(0.587, 0.0122)
(0.588, 0.0124)
(0.589, 0.0125)
(0.590, 0.0127)
(0.591, 0.0128)
(0.592, 0.0130)
(0.593, 0.0131)
(0.594, 0.0133)
(0.595, 0.0134)
(0.596, 0.0136)
(0.597, 0.0137)
(0.598, 0.0139)
(0.599, 0.0141)
(0.600, 0.0142)
(0.601, 0.0144)
(0.602, 0.0146)
(0.603, 0.0147)
(0.604, 0.0149)
(0.605, 0.0151)
(0.606, 0.0153)
(0.607, 0.0154)
(0.608, 0.0156)
(0.609, 0.0158)
(0.610, 0.0160)
(0.611, 0.0161)
(0.612, 0.0163)
(0.613, 0.0165)
(0.614, 0.0167)
(0.615, 0.0169)
(0.616, 0.0171)
(0.617, 0.0173)
(0.618, 0.0175)
(0.619, 0.0177)
(0.620, 0.0179)
(0.621, 0.0181)
(0.622, 0.0183)
(0.623, 0.0185)
(0.624, 0.0187)
(0.625, 0.0189)
(0.626, 0.0191)
(0.627, 0.0193)
(0.628, 0.0195)
(0.629, 0.0198)
(0.630, 0.0200)
(0.631, 0.0202)
(0.632, 0.0204)
(0.633, 0.0206)
(0.634, 0.0209)
(0.635, 0.0211)
(0.636, 0.0213)
(0.637, 0.0216)
(0.638, 0.0218)
(0.639, 0.0220)
(0.640, 0.0223)
(0.641, 0.0225)
(0.642, 0.0228)
(0.643, 0.0230)
(0.644, 0.0233)
(0.645, 0.0235)
(0.646, 0.0238)
(0.647, 0.0240)
(0.648, 0.0243)
(0.649, 0.0245)
(0.650, 0.0248)
(0.651, 0.0251)
(0.652, 0.0253)
(0.653, 0.0256)
(0.654, 0.0259)
(0.655, 0.0262)
(0.656, 0.0264)
(0.657, 0.0267)
(0.658, 0.0270)
(0.659, 0.0273)
(0.660, 0.0276)
(0.661, 0.0279)
(0.662, 0.0282)
(0.663, 0.0285)
(0.664, 0.0288)
(0.665, 0.0291)
(0.666, 0.0294)
(0.667, 0.0297)
(0.668, 0.0300)
(0.669, 0.0303)
(0.670, 0.0306)
(0.671, 0.0309)
(0.672, 0.0313)
(0.673, 0.0316)
(0.674, 0.0319)
(0.675, 0.0322)
(0.676, 0.0326)
(0.677, 0.0329)
(0.678, 0.0332)
(0.679, 0.0336)
(0.680, 0.0339)
(0.681, 0.0343)
(0.682, 0.0346)
(0.683, 0.0350)
(0.684, 0.0353)
(0.685, 0.0357)
(0.686, 0.0361)
(0.687, 0.0364)
(0.688, 0.0368)
(0.689, 0.0372)
(0.690, 0.0376)
(0.691, 0.0379)
(0.692, 0.0383)
(0.693, 0.0387)
(0.694, 0.0391)
(0.695, 0.0395)
(0.696, 0.0399)
(0.697, 0.0403)
(0.698, 0.0407)
(0.699, 0.0411)
(0.700, 0.0415)
(0.701, 0.0419)
(0.702, 0.0423)
(0.703, 0.0427)
(0.704, 0.0432)
(0.705, 0.0436)
(0.706, 0.0440)
(0.707, 0.0445)
(0.708, 0.0449)
(0.709, 0.0453)
(0.710, 0.0458)
(0.711, 0.0462)
(0.712, 0.0467)
(0.713, 0.0471)
(0.714, 0.0476)
(0.715, 0.0481)
(0.716, 0.0485)
(0.717, 0.0490)
(0.718, 0.0495)
(0.719, 0.0500)
(0.720, 0.0505)
(0.721, 0.0509)
(0.722, 0.0514)
(0.723, 0.0519)
(0.724, 0.0524)
(0.725, 0.0529)
(0.726, 0.0534)
(0.727, 0.0540)
(0.728, 0.0545)
(0.729, 0.0550)
(0.730, 0.0555)
(0.731, 0.0561)
(0.732, 0.0566)
(0.733, 0.0571)
(0.734, 0.0577)
(0.735, 0.0582)
(0.736, 0.0588)
(0.737, 0.0593)
(0.738, 0.0599)
(0.739, 0.0605)
(0.740, 0.0610)
(0.741, 0.0616)
(0.742, 0.0622)
(0.743, 0.0628)
(0.744, 0.0634)
(0.745, 0.0639)
(0.746, 0.0645)
(0.747, 0.0651)
(0.748, 0.0658)
(0.749, 0.0664)
(0.750, 0.0670)
(0.751, 0.0676)
(0.752, 0.0682)
(0.753, 0.0689)
(0.754, 0.0695)
(0.755, 0.0701)
(0.756, 0.0708)
(0.757, 0.0714)
(0.758, 0.0721)
(0.759, 0.0728)
(0.760, 0.0734)
(0.761, 0.0741)
(0.762, 0.0748)
(0.763, 0.0755)
(0.764, 0.0762)
(0.765, 0.0769)
(0.766, 0.0776)
(0.767, 0.0783)
(0.768, 0.0790)
(0.769, 0.0797)
(0.770, 0.0804)
(0.771, 0.0811)
(0.772, 0.0819)
(0.773, 0.0826)
(0.774, 0.0833)
(0.775, 0.0841)
(0.776, 0.0849)
(0.777, 0.0856)
(0.778, 0.0864)
(0.779, 0.0872)
(0.780, 0.0879)
(0.781, 0.0887)
(0.782, 0.0895)
(0.783, 0.0903)
(0.784, 0.0911)
(0.785, 0.0919)
(0.786, 0.0927)
(0.787, 0.0936)
(0.788, 0.0944)
(0.789, 0.0952)
(0.790, 0.0961)
(0.791, 0.0969)
(0.792, 0.0978)
(0.793, 0.0986)
(0.794, 0.0995)
(0.795, 0.1004)
(0.796, 0.1012)
(0.797, 0.1021)
(0.798, 0.1030)
(0.799, 0.1039)
(0.800, 0.1048)
(0.801, 0.1057)
(0.802, 0.1067)
(0.803, 0.1076)
(0.804, 0.1085)
(0.805, 0.1095)
(0.806, 0.1104)
(0.807, 0.1114)
(0.808, 0.1123)
(0.809, 0.1133)
(0.810, 0.1143)
(0.811, 0.1153)
(0.812, 0.1162)
(0.813, 0.1172)
(0.814, 0.1182)
(0.815, 0.1193)
(0.816, 0.1203)
(0.817, 0.1213)
(0.818, 0.1223)
(0.819, 0.1234)
(0.820, 0.1244)
(0.821, 0.1255)
(0.822, 0.1265)
(0.823, 0.1276)
(0.824, 0.1287)
(0.825, 0.1298)
(0.826, 0.1309)
(0.827, 0.1320)
(0.828, 0.1331)
(0.829, 0.1342)
(0.830, 0.1353)
(0.831, 0.1365)
(0.832, 0.1376)
(0.833, 0.1388)
(0.834, 0.1399)
(0.835, 0.1411)
(0.836, 0.1423)
(0.837, 0.1435)
(0.838, 0.1447)
(0.839, 0.1459)
(0.840, 0.1471)
(0.841, 0.1483)
(0.842, 0.1495)
(0.843, 0.1508)
(0.844, 0.1520)
(0.845, 0.1533)
(0.846, 0.1545)
(0.847, 0.1558)
(0.848, 0.1571)
(0.849, 0.1584)
(0.850, 0.1597)
(0.851, 0.1610)
(0.852, 0.1623)
(0.853, 0.1636)
(0.854, 0.1650)
(0.855, 0.1663)
(0.856, 0.1677)
(0.857, 0.1690)
(0.858, 0.1704)
(0.859, 0.1718)
(0.860, 0.1732)
(0.861, 0.1746)
(0.862, 0.1760)
(0.863, 0.1774)
(0.864, 0.1788)
(0.865, 0.1803)
(0.866, 0.1817)
(0.867, 0.1832)
(0.868, 0.1847)
(0.869, 0.1861)
(0.870, 0.1876)
(0.871, 0.1891)
(0.872, 0.1906)
(0.873, 0.1922)
(0.874, 0.1937)
(0.875, 0.1952)
(0.875, 0.0000)}
\defdata{emaprofile-8-short}{
(0.000, 0.0000)
(0.583, 0.0000)
(0.584, 0.0000)
(0.585, 0.0000)
(0.586, 0.0000)
(0.587, 0.0000)
(0.588, 0.0000)
(0.589, 0.0000)
(0.590, 0.0000)
(0.591, 0.0001)
(0.592, 0.0001)
(0.593, 0.0001)
(0.594, 0.0001)
(0.595, 0.0001)
(0.596, 0.0001)
(0.597, 0.0001)
(0.598, 0.0001)
(0.599, 0.0001)
(0.600, 0.0001)
(0.601, 0.0001)
(0.602, 0.0001)
(0.603, 0.0001)
(0.604, 0.0001)
(0.605, 0.0001)
(0.606, 0.0001)
(0.607, 0.0001)
(0.608, 0.0001)
(0.609, 0.0001)
(0.610, 0.0001)
(0.611, 0.0001)
(0.612, 0.0001)
(0.613, 0.0001)
(0.614, 0.0001)
(0.615, 0.0001)
(0.616, 0.0001)
(0.617, 0.0001)
(0.618, 0.0001)
(0.619, 0.0001)
(0.620, 0.0001)
(0.621, 0.0001)
(0.622, 0.0001)
(0.623, 0.0001)
(0.624, 0.0001)
(0.625, 0.0001)
(0.626, 0.0001)
(0.627, 0.0001)
(0.628, 0.0001)
(0.629, 0.0001)
(0.630, 0.0002)
(0.631, 0.0002)
(0.632, 0.0002)
(0.633, 0.0002)
(0.634, 0.0002)
(0.635, 0.0002)
(0.636, 0.0002)
(0.637, 0.0002)
(0.638, 0.0002)
(0.639, 0.0002)
(0.640, 0.0002)
(0.641, 0.0002)
(0.642, 0.0002)
(0.643, 0.0002)
(0.644, 0.0002)
(0.645, 0.0002)
(0.646, 0.0002)
(0.647, 0.0002)
(0.648, 0.0002)
(0.649, 0.0003)
(0.650, 0.0003)
(0.651, 0.0003)
(0.652, 0.0003)
(0.653, 0.0003)
(0.654, 0.0003)
(0.655, 0.0003)
(0.656, 0.0003)
(0.657, 0.0003)
(0.658, 0.0003)
(0.659, 0.0003)
(0.660, 0.0003)
(0.661, 0.0003)
(0.662, 0.0004)
(0.663, 0.0004)
(0.664, 0.0004)
(0.665, 0.0004)
(0.666, 0.0004)
(0.667, 0.0004)
(0.668, 0.0004)
(0.669, 0.0004)
(0.670, 0.0004)
(0.671, 0.0004)
(0.672, 0.0005)
(0.673, 0.0005)
(0.674, 0.0005)
(0.675, 0.0005)
(0.676, 0.0005)
(0.677, 0.0005)
(0.678, 0.0005)
(0.679, 0.0005)
(0.680, 0.0006)
(0.681, 0.0006)
(0.682, 0.0006)
(0.683, 0.0006)
(0.684, 0.0006)
(0.685, 0.0006)
(0.686, 0.0006)
(0.687, 0.0007)
(0.688, 0.0007)
(0.689, 0.0007)
(0.690, 0.0007)
(0.691, 0.0007)
(0.692, 0.0007)
(0.693, 0.0008)
(0.694, 0.0008)
(0.695, 0.0008)
(0.696, 0.0008)
(0.697, 0.0008)
(0.698, 0.0009)
(0.699, 0.0009)
(0.700, 0.0009)
(0.701, 0.0009)
(0.702, 0.0010)
(0.703, 0.0010)
(0.704, 0.0010)
(0.705, 0.0010)
(0.706, 0.0010)
(0.707, 0.0011)
(0.708, 0.0011)
(0.709, 0.0011)
(0.710, 0.0012)
(0.711, 0.0012)
(0.712, 0.0012)
(0.713, 0.0012)
(0.714, 0.0013)
(0.715, 0.0013)
(0.716, 0.0013)
(0.717, 0.0014)
(0.718, 0.0014)
(0.719, 0.0014)
(0.720, 0.0015)
(0.721, 0.0015)
(0.722, 0.0015)
(0.723, 0.0016)
(0.724, 0.0016)
(0.725, 0.0016)
(0.726, 0.0017)
(0.727, 0.0017)
(0.728, 0.0018)
(0.729, 0.0018)
(0.730, 0.0018)
(0.731, 0.0019)
(0.732, 0.0019)
(0.733, 0.0020)
(0.734, 0.0020)
(0.735, 0.0021)
(0.736, 0.0021)
(0.737, 0.0022)
(0.738, 0.0022)
(0.739, 0.0023)
(0.740, 0.0023)
(0.741, 0.0024)
(0.742, 0.0024)
(0.743, 0.0025)
(0.744, 0.0025)
(0.745, 0.0026)
(0.746, 0.0027)
(0.747, 0.0027)
(0.748, 0.0028)
(0.749, 0.0029)
(0.750, 0.0029)
(0.751, 0.0030)
(0.752, 0.0031)
(0.753, 0.0031)
(0.754, 0.0032)
(0.755, 0.0033)
(0.756, 0.0033)
(0.757, 0.0034)
(0.758, 0.0035)
(0.759, 0.0036)
(0.760, 0.0037)
(0.761, 0.0037)
(0.762, 0.0038)
(0.763, 0.0039)
(0.764, 0.0040)
(0.765, 0.0041)
(0.766, 0.0042)
(0.767, 0.0043)
(0.768, 0.0044)
(0.769, 0.0045)
(0.770, 0.0046)
(0.771, 0.0047)
(0.772, 0.0048)
(0.773, 0.0049)
(0.774, 0.0050)
(0.775, 0.0051)
(0.776, 0.0052)
(0.777, 0.0053)
(0.778, 0.0054)
(0.779, 0.0056)
(0.780, 0.0057)
(0.781, 0.0058)
(0.782, 0.0059)
(0.783, 0.0061)
(0.784, 0.0062)
(0.785, 0.0063)
(0.786, 0.0065)
(0.787, 0.0066)
(0.788, 0.0068)
(0.789, 0.0069)
(0.790, 0.0070)
(0.791, 0.0072)
(0.792, 0.0074)
(0.793, 0.0075)
(0.794, 0.0077)
(0.795, 0.0078)
(0.796, 0.0080)
(0.797, 0.0082)
(0.798, 0.0084)
(0.799, 0.0085)
(0.800, 0.0087)
(0.801, 0.0089)
(0.802, 0.0091)
(0.803, 0.0093)
(0.804, 0.0095)
(0.805, 0.0097)
(0.806, 0.0099)
(0.807, 0.0101)
(0.808, 0.0103)
(0.809, 0.0106)
(0.810, 0.0108)
(0.811, 0.0110)
(0.812, 0.0112)
(0.813, 0.0115)
(0.814, 0.0117)
(0.815, 0.0120)
(0.816, 0.0122)
(0.817, 0.0125)
(0.818, 0.0127)
(0.819, 0.0130)
(0.820, 0.0133)
(0.821, 0.0135)
(0.822, 0.0138)
(0.823, 0.0141)
(0.824, 0.0144)
(0.825, 0.0147)
(0.826, 0.0150)
(0.827, 0.0153)
(0.828, 0.0156)
(0.829, 0.0160)
(0.830, 0.0163)
(0.831, 0.0166)
(0.832, 0.0170)
(0.833, 0.0173)
(0.834, 0.0177)
(0.835, 0.0180)
(0.836, 0.0184)
(0.837, 0.0188)
(0.838, 0.0192)
(0.839, 0.0196)
(0.840, 0.0200)
(0.841, 0.0204)
(0.842, 0.0208)
(0.843, 0.0212)
(0.844, 0.0216)
(0.845, 0.0221)
(0.846, 0.0225)
(0.847, 0.0230)
(0.848, 0.0235)
(0.849, 0.0239)
(0.850, 0.0244)
(0.851, 0.0249)
(0.852, 0.0254)
(0.853, 0.0259)
(0.854, 0.0264)
(0.855, 0.0270)
(0.856, 0.0275)
(0.857, 0.0281)
(0.858, 0.0286)
(0.859, 0.0292)
(0.860, 0.0298)
(0.861, 0.0304)
(0.862, 0.0310)
(0.863, 0.0316)
(0.864, 0.0322)
(0.865, 0.0329)
(0.866, 0.0335)
(0.867, 0.0342)
(0.868, 0.0348)
(0.869, 0.0355)
(0.870, 0.0362)
(0.871, 0.0369)
(0.872, 0.0377)
(0.873, 0.0384)
(0.874, 0.0392)
(0.875, 0.0399)
(0.876, 0.0407)
(0.877, 0.0415)
(0.878, 0.0423)
(0.879, 0.0431)
(0.880, 0.0440)
(0.881, 0.0448)
(0.882, 0.0457)
(0.883, 0.0466)
(0.884, 0.0475)
(0.885, 0.0484)
(0.886, 0.0494)
(0.887, 0.0503)
(0.888, 0.0513)
(0.889, 0.0523)
(0.890, 0.0533)
(0.891, 0.0543)
(0.892, 0.0553)
(0.893, 0.0564)
(0.894, 0.0575)
(0.895, 0.0586)
(0.896, 0.0597)
(0.897, 0.0609)
(0.898, 0.0620)
(0.899, 0.0632)
(0.900, 0.0644)
(0.901, 0.0656)
(0.902, 0.0669)
(0.903, 0.0681)
(0.904, 0.0694)
(0.905, 0.0707)
(0.906, 0.0721)
(0.907, 0.0734)
(0.908, 0.0748)
(0.909, 0.0762)
(0.910, 0.0777)
(0.911, 0.0791)
(0.912, 0.0806)
(0.913, 0.0821)
(0.914, 0.0837)
(0.915, 0.0853)
(0.916, 0.0868)
(0.917, 0.0885)
(0.918, 0.0901)
(0.919, 0.0918)
(0.920, 0.0935)
(0.921, 0.0953)
(0.922, 0.0970)
(0.923, 0.0988)
(0.924, 0.1007)
(0.925, 0.1025)
(0.926, 0.1044)
(0.927, 0.1063)
(0.928, 0.1083)
(0.929, 0.1103)
(0.930, 0.1123)
(0.931, 0.1144)
(0.932, 0.1165)
(0.933, 0.1187)
(0.934, 0.1208)
(0.935, 0.1230)
(0.936, 0.1253)
(0.937, 0.1276)
(0.938, 0.1299)
(0.939, 0.1323)
(0.940, 0.1347)
(0.941, 0.1372)
(0.942, 0.1396)
(0.943, 0.1422)
(0.944, 0.1448)
(0.945, 0.1474)
(0.946, 0.1501)
(0.947, 0.1528)
(0.948, 0.1555)
(0.949, 0.1583)
(0.950, 0.1612)
(0.951, 0.1641)
(0.952, 0.1671)
(0.953, 0.1701)
(0.954, 0.1731)
(0.955, 0.1762)
(0.956, 0.1794)
(0.957, 0.1826)
(0.958, 0.1858)
(0.959, 0.1892)
(0.960, 0.1925)
(0.961, 0.1960)
(0.962, 0.1995)
(0.963, 0.2030)
(0.964, 0.2066)
(0.965, 0.2103)
(0.966, 0.2140)
(0.967, 0.2178)
(0.968, 0.2217)
(0.969, 0.2256)
(0.970, 0.2296)
(0.971, 0.2336)
(0.972, 0.2377)
(0.973, 0.2419)
(0.974, 0.2462)
(0.975, 0.2505)
(0.976, 0.2549)
(0.977, 0.2594)
(0.978, 0.2639)
(0.979, 0.2685)
(0.980, 0.2732)
(0.981, 0.2780)
(0.982, 0.2828)
(0.983, 0.2878)
(0.984, 0.2928)
(0.985, 0.2978)
(0.986, 0.3030)
(0.987, 0.3083)
(0.988, 0.3136)
(0.989, 0.3191)
(0.990, 0.3246)
(0.991, 0.3302)
(0.992, 0.3359)
(0.993, 0.3417)
(0.994, 0.3476)
(0.995, 0.3535)
(0.996, 0.3596)
(0.997, 0.3658)
(0.998, 0.3721)
(0.999, 0.3785)
(1.000, 0.3849)
(1.000, 0.0000)}
\defdata{emaprofile-8-long}{
(0.000, 0.0000)
(0.300, 0.0000)
(0.301, 0.0000)
(0.302, 0.0000)
(0.303, 0.0000)
(0.304, 0.0000)
(0.305, 0.0000)
(0.306, 0.0000)
(0.307, 0.0000)
(0.308, 0.0000)
(0.309, 0.0000)
(0.310, 0.0001)
(0.311, 0.0001)
(0.312, 0.0001)
(0.313, 0.0001)
(0.314, 0.0001)
(0.315, 0.0001)
(0.316, 0.0001)
(0.317, 0.0001)
(0.318, 0.0001)
(0.319, 0.0001)
(0.320, 0.0001)
(0.321, 0.0001)
(0.322, 0.0001)
(0.323, 0.0001)
(0.324, 0.0001)
(0.325, 0.0001)
(0.326, 0.0001)
(0.327, 0.0001)
(0.328, 0.0001)
(0.329, 0.0001)
(0.330, 0.0001)
(0.331, 0.0001)
(0.332, 0.0001)
(0.333, 0.0001)
(0.334, 0.0001)
(0.335, 0.0001)
(0.336, 0.0001)
(0.337, 0.0001)
(0.338, 0.0001)
(0.339, 0.0001)
(0.340, 0.0001)
(0.341, 0.0001)
(0.342, 0.0001)
(0.343, 0.0001)
(0.344, 0.0001)
(0.345, 0.0001)
(0.346, 0.0001)
(0.347, 0.0001)
(0.348, 0.0001)
(0.349, 0.0001)
(0.350, 0.0001)
(0.351, 0.0001)
(0.352, 0.0001)
(0.353, 0.0001)
(0.354, 0.0001)
(0.355, 0.0001)
(0.356, 0.0001)
(0.357, 0.0001)
(0.358, 0.0001)
(0.359, 0.0001)
(0.360, 0.0001)
(0.361, 0.0001)
(0.362, 0.0001)
(0.363, 0.0002)
(0.364, 0.0002)
(0.365, 0.0002)
(0.366, 0.0002)
(0.367, 0.0002)
(0.368, 0.0002)
(0.369, 0.0002)
(0.370, 0.0002)
(0.371, 0.0002)
(0.372, 0.0002)
(0.373, 0.0002)
(0.374, 0.0002)
(0.375, 0.0002)
(0.376, 0.0002)
(0.377, 0.0002)
(0.378, 0.0002)
(0.379, 0.0002)
(0.380, 0.0002)
(0.381, 0.0002)
(0.382, 0.0002)
(0.383, 0.0002)
(0.384, 0.0002)
(0.385, 0.0002)
(0.386, 0.0002)
(0.387, 0.0002)
(0.388, 0.0002)
(0.389, 0.0002)
(0.390, 0.0002)
(0.391, 0.0003)
(0.392, 0.0003)
(0.393, 0.0003)
(0.394, 0.0003)
(0.395, 0.0003)
(0.396, 0.0003)
(0.397, 0.0003)
(0.398, 0.0003)
(0.399, 0.0003)
(0.400, 0.0003)
(0.401, 0.0003)
(0.402, 0.0003)
(0.403, 0.0003)
(0.404, 0.0003)
(0.405, 0.0003)
(0.406, 0.0003)
(0.407, 0.0003)
(0.408, 0.0003)
(0.409, 0.0003)
(0.410, 0.0004)
(0.411, 0.0004)
(0.412, 0.0004)
(0.413, 0.0004)
(0.414, 0.0004)
(0.415, 0.0004)
(0.416, 0.0004)
(0.417, 0.0004)
(0.418, 0.0004)
(0.419, 0.0004)
(0.420, 0.0004)
(0.421, 0.0004)
(0.422, 0.0004)
(0.423, 0.0004)
(0.424, 0.0004)
(0.425, 0.0005)
(0.426, 0.0005)
(0.427, 0.0005)
(0.428, 0.0005)
(0.429, 0.0005)
(0.430, 0.0005)
(0.431, 0.0005)
(0.432, 0.0005)
(0.433, 0.0005)
(0.434, 0.0005)
(0.435, 0.0005)
(0.436, 0.0005)
(0.437, 0.0005)
(0.438, 0.0006)
(0.439, 0.0006)
(0.440, 0.0006)
(0.441, 0.0006)
(0.442, 0.0006)
(0.443, 0.0006)
(0.444, 0.0006)
(0.445, 0.0006)
(0.446, 0.0006)
(0.447, 0.0006)
(0.448, 0.0006)
(0.449, 0.0007)
(0.450, 0.0007)
(0.451, 0.0007)
(0.452, 0.0007)
(0.453, 0.0007)
(0.454, 0.0007)
(0.455, 0.0007)
(0.456, 0.0007)
(0.457, 0.0007)
(0.458, 0.0008)
(0.459, 0.0008)
(0.460, 0.0008)
(0.461, 0.0008)
(0.462, 0.0008)
(0.463, 0.0008)
(0.464, 0.0008)
(0.465, 0.0008)
(0.466, 0.0009)
(0.467, 0.0009)
(0.468, 0.0009)
(0.469, 0.0009)
(0.470, 0.0009)
(0.471, 0.0009)
(0.472, 0.0009)
(0.473, 0.0009)
(0.474, 0.0010)
(0.475, 0.0010)
(0.476, 0.0010)
(0.477, 0.0010)
(0.478, 0.0010)
(0.479, 0.0010)
(0.480, 0.0010)
(0.481, 0.0011)
(0.482, 0.0011)
(0.483, 0.0011)
(0.484, 0.0011)
(0.485, 0.0011)
(0.486, 0.0011)
(0.487, 0.0012)
(0.488, 0.0012)
(0.489, 0.0012)
(0.490, 0.0012)
(0.491, 0.0012)
(0.492, 0.0012)
(0.493, 0.0013)
(0.494, 0.0013)
(0.495, 0.0013)
(0.496, 0.0013)
(0.497, 0.0013)
(0.498, 0.0014)
(0.499, 0.0014)
(0.500, 0.0014)
(0.501, 0.0014)
(0.502, 0.0014)
(0.503, 0.0015)
(0.504, 0.0015)
(0.505, 0.0015)
(0.506, 0.0015)
(0.507, 0.0015)
(0.508, 0.0016)
(0.509, 0.0016)
(0.510, 0.0016)
(0.511, 0.0016)
(0.512, 0.0016)
(0.513, 0.0017)
(0.514, 0.0017)
(0.515, 0.0017)
(0.516, 0.0017)
(0.517, 0.0018)
(0.518, 0.0018)
(0.519, 0.0018)
(0.520, 0.0018)
(0.521, 0.0019)
(0.522, 0.0019)
(0.523, 0.0019)
(0.524, 0.0019)
(0.525, 0.0020)
(0.526, 0.0020)
(0.527, 0.0020)
(0.528, 0.0020)
(0.529, 0.0021)
(0.530, 0.0021)
(0.531, 0.0021)
(0.532, 0.0021)
(0.533, 0.0022)
(0.534, 0.0022)
(0.535, 0.0022)
(0.536, 0.0023)
(0.537, 0.0023)
(0.538, 0.0023)
(0.539, 0.0023)
(0.540, 0.0024)
(0.541, 0.0024)
(0.542, 0.0024)
(0.543, 0.0025)
(0.544, 0.0025)
(0.545, 0.0025)
(0.546, 0.0026)
(0.547, 0.0026)
(0.548, 0.0026)
(0.549, 0.0027)
(0.550, 0.0027)
(0.551, 0.0027)
(0.552, 0.0028)
(0.553, 0.0028)
(0.554, 0.0028)
(0.555, 0.0029)
(0.556, 0.0029)
(0.557, 0.0029)
(0.558, 0.0030)
(0.559, 0.0030)
(0.560, 0.0031)
(0.561, 0.0031)
(0.562, 0.0031)
(0.563, 0.0032)
(0.564, 0.0032)
(0.565, 0.0033)
(0.566, 0.0033)
(0.567, 0.0033)
(0.568, 0.0034)
(0.569, 0.0034)
(0.570, 0.0035)
(0.571, 0.0035)
(0.572, 0.0035)
(0.573, 0.0036)
(0.574, 0.0036)
(0.575, 0.0037)
(0.576, 0.0037)
(0.577, 0.0038)
(0.578, 0.0038)
(0.579, 0.0039)
(0.580, 0.0039)
(0.581, 0.0039)
(0.582, 0.0040)
(0.583, 0.0040)
(0.584, 0.0041)
(0.585, 0.0041)
(0.586, 0.0042)
(0.587, 0.0042)
(0.588, 0.0043)
(0.589, 0.0043)
(0.590, 0.0044)
(0.591, 0.0044)
(0.592, 0.0045)
(0.593, 0.0045)
(0.594, 0.0046)
(0.595, 0.0047)
(0.596, 0.0047)
(0.597, 0.0048)
(0.598, 0.0048)
(0.599, 0.0049)
(0.600, 0.0049)
(0.601, 0.0050)
(0.602, 0.0050)
(0.603, 0.0051)
(0.604, 0.0052)
(0.605, 0.0052)
(0.606, 0.0053)
(0.607, 0.0053)
(0.608, 0.0054)
(0.609, 0.0055)
(0.610, 0.0055)
(0.611, 0.0056)
(0.612, 0.0057)
(0.613, 0.0057)
(0.614, 0.0058)
(0.615, 0.0059)
(0.616, 0.0059)
(0.617, 0.0060)
(0.618, 0.0061)
(0.619, 0.0061)
(0.620, 0.0062)
(0.621, 0.0063)
(0.622, 0.0063)
(0.623, 0.0064)
(0.624, 0.0065)
(0.625, 0.0065)
(0.626, 0.0066)
(0.627, 0.0067)
(0.628, 0.0068)
(0.629, 0.0068)
(0.630, 0.0069)
(0.631, 0.0070)
(0.632, 0.0071)
(0.633, 0.0072)
(0.634, 0.0072)
(0.635, 0.0073)
(0.636, 0.0074)
(0.637, 0.0075)
(0.638, 0.0076)
(0.639, 0.0076)
(0.640, 0.0077)
(0.641, 0.0078)
(0.642, 0.0079)
(0.643, 0.0080)
(0.644, 0.0081)
(0.645, 0.0082)
(0.646, 0.0082)
(0.647, 0.0083)
(0.648, 0.0084)
(0.649, 0.0085)
(0.650, 0.0086)
(0.651, 0.0087)
(0.652, 0.0088)
(0.653, 0.0089)
(0.654, 0.0090)
(0.655, 0.0091)
(0.656, 0.0092)
(0.657, 0.0093)
(0.658, 0.0094)
(0.659, 0.0095)
(0.660, 0.0096)
(0.661, 0.0097)
(0.662, 0.0098)
(0.663, 0.0099)
(0.664, 0.0100)
(0.665, 0.0101)
(0.666, 0.0102)
(0.667, 0.0103)
(0.668, 0.0104)
(0.669, 0.0105)
(0.670, 0.0106)
(0.671, 0.0107)
(0.672, 0.0108)
(0.673, 0.0109)
(0.674, 0.0111)
(0.675, 0.0112)
(0.676, 0.0113)
(0.677, 0.0114)
(0.678, 0.0115)
(0.679, 0.0116)
(0.680, 0.0118)
(0.681, 0.0119)
(0.682, 0.0120)
(0.683, 0.0121)
(0.684, 0.0122)
(0.685, 0.0124)
(0.686, 0.0125)
(0.687, 0.0126)
(0.688, 0.0128)
(0.689, 0.0129)
(0.690, 0.0130)
(0.691, 0.0131)
(0.692, 0.0133)
(0.693, 0.0134)
(0.694, 0.0135)
(0.695, 0.0137)
(0.696, 0.0138)
(0.697, 0.0140)
(0.698, 0.0141)
(0.699, 0.0142)
(0.700, 0.0144)
(0.701, 0.0145)
(0.702, 0.0147)
(0.703, 0.0148)
(0.704, 0.0150)
(0.705, 0.0151)
(0.706, 0.0153)
(0.707, 0.0154)
(0.708, 0.0156)
(0.709, 0.0157)
(0.710, 0.0159)
(0.711, 0.0160)
(0.712, 0.0162)
(0.713, 0.0163)
(0.714, 0.0165)
(0.715, 0.0167)
(0.716, 0.0168)
(0.717, 0.0170)
(0.718, 0.0172)
(0.719, 0.0173)
(0.720, 0.0175)
(0.721, 0.0177)
(0.722, 0.0178)
(0.723, 0.0180)
(0.724, 0.0182)
(0.725, 0.0183)
(0.726, 0.0185)
(0.727, 0.0187)
(0.728, 0.0189)
(0.729, 0.0191)
(0.730, 0.0192)
(0.731, 0.0194)
(0.732, 0.0196)
(0.733, 0.0198)
(0.734, 0.0200)
(0.735, 0.0202)
(0.736, 0.0204)
(0.737, 0.0206)
(0.738, 0.0208)
(0.739, 0.0210)
(0.740, 0.0211)
(0.741, 0.0213)
(0.742, 0.0215)
(0.743, 0.0218)
(0.744, 0.0220)
(0.745, 0.0222)
(0.746, 0.0224)
(0.747, 0.0226)
(0.748, 0.0228)
(0.749, 0.0230)
(0.750, 0.0232)
(0.751, 0.0234)
(0.752, 0.0236)
(0.753, 0.0239)
(0.754, 0.0241)
(0.755, 0.0243)
(0.756, 0.0245)
(0.757, 0.0248)
(0.758, 0.0250)
(0.759, 0.0252)
(0.760, 0.0254)
(0.761, 0.0257)
(0.762, 0.0259)
(0.763, 0.0262)
(0.764, 0.0264)
(0.765, 0.0266)
(0.766, 0.0269)
(0.767, 0.0271)
(0.768, 0.0274)
(0.769, 0.0276)
(0.770, 0.0279)
(0.771, 0.0281)
(0.772, 0.0284)
(0.773, 0.0286)
(0.774, 0.0289)
(0.775, 0.0291)
(0.776, 0.0294)
(0.777, 0.0297)
(0.778, 0.0299)
(0.779, 0.0302)
(0.780, 0.0305)
(0.781, 0.0307)
(0.782, 0.0310)
(0.783, 0.0313)
(0.784, 0.0316)
(0.785, 0.0319)
(0.786, 0.0321)
(0.787, 0.0324)
(0.788, 0.0327)
(0.789, 0.0330)
(0.790, 0.0333)
(0.791, 0.0336)
(0.792, 0.0339)
(0.793, 0.0342)
(0.794, 0.0345)
(0.795, 0.0348)
(0.796, 0.0351)
(0.797, 0.0354)
(0.798, 0.0357)
(0.799, 0.0360)
(0.800, 0.0363)
(0.801, 0.0366)
(0.802, 0.0370)
(0.803, 0.0373)
(0.804, 0.0376)
(0.805, 0.0379)
(0.806, 0.0383)
(0.807, 0.0386)
(0.808, 0.0389)
(0.809, 0.0393)
(0.810, 0.0396)
(0.811, 0.0399)
(0.812, 0.0403)
(0.813, 0.0406)
(0.814, 0.0410)
(0.815, 0.0413)
(0.816, 0.0417)
(0.817, 0.0420)
(0.818, 0.0424)
(0.819, 0.0428)
(0.820, 0.0431)
(0.821, 0.0435)
(0.822, 0.0439)
(0.823, 0.0442)
(0.824, 0.0446)
(0.825, 0.0450)
(0.826, 0.0454)
(0.827, 0.0457)
(0.828, 0.0461)
(0.829, 0.0465)
(0.830, 0.0469)
(0.831, 0.0473)
(0.832, 0.0477)
(0.833, 0.0481)
(0.834, 0.0485)
(0.835, 0.0489)
(0.836, 0.0493)
(0.837, 0.0497)
(0.838, 0.0501)
(0.839, 0.0506)
(0.840, 0.0510)
(0.841, 0.0514)
(0.842, 0.0518)
(0.843, 0.0522)
(0.844, 0.0527)
(0.845, 0.0531)
(0.846, 0.0536)
(0.847, 0.0540)
(0.848, 0.0544)
(0.849, 0.0549)
(0.850, 0.0553)
(0.851, 0.0558)
(0.852, 0.0562)
(0.853, 0.0567)
(0.854, 0.0572)
(0.855, 0.0576)
(0.856, 0.0581)
(0.857, 0.0586)
(0.858, 0.0591)
(0.859, 0.0595)
(0.860, 0.0600)
(0.861, 0.0605)
(0.862, 0.0610)
(0.863, 0.0615)
(0.864, 0.0620)
(0.865, 0.0625)
(0.866, 0.0630)
(0.867, 0.0635)
(0.868, 0.0640)
(0.869, 0.0645)
(0.870, 0.0650)
(0.871, 0.0655)
(0.872, 0.0661)
(0.873, 0.0666)
(0.874, 0.0671)
(0.875, 0.0677)
(0.876, 0.0682)
(0.877, 0.0687)
(0.878, 0.0693)
(0.879, 0.0698)
(0.880, 0.0704)
(0.881, 0.0710)
(0.882, 0.0715)
(0.883, 0.0721)
(0.884, 0.0726)
(0.885, 0.0732)
(0.886, 0.0738)
(0.887, 0.0744)
(0.888, 0.0750)
(0.889, 0.0755)
(0.890, 0.0761)
(0.891, 0.0767)
(0.892, 0.0773)
(0.893, 0.0779)
(0.894, 0.0785)
(0.895, 0.0792)
(0.896, 0.0798)
(0.897, 0.0804)
(0.898, 0.0810)
(0.899, 0.0816)
(0.900, 0.0823)
(0.901, 0.0829)
(0.902, 0.0836)
(0.903, 0.0842)
(0.904, 0.0848)
(0.905, 0.0855)
(0.906, 0.0862)
(0.907, 0.0868)
(0.908, 0.0875)
(0.909, 0.0882)
(0.910, 0.0888)
(0.911, 0.0895)
(0.912, 0.0902)
(0.913, 0.0909)
(0.914, 0.0916)
(0.915, 0.0923)
(0.916, 0.0930)
(0.917, 0.0937)
(0.918, 0.0944)
(0.919, 0.0951)
(0.920, 0.0958)
(0.921, 0.0966)
(0.922, 0.0973)
(0.923, 0.0980)
(0.924, 0.0988)
(0.925, 0.0995)
(0.926, 0.1003)
(0.927, 0.1010)
(0.928, 0.1018)
(0.929, 0.1025)
(0.930, 0.1033)
(0.931, 0.1041)
(0.932, 0.1049)
(0.933, 0.1056)
(0.934, 0.1064)
(0.935, 0.1072)
(0.936, 0.1080)
(0.937, 0.1088)
(0.938, 0.1096)
(0.939, 0.1104)
(0.940, 0.1113)
(0.941, 0.1121)
(0.942, 0.1129)
(0.943, 0.1137)
(0.944, 0.1146)
(0.945, 0.1154)
(0.946, 0.1163)
(0.947, 0.1171)
(0.948, 0.1180)
(0.949, 0.1189)
(0.950, 0.1197)
(0.951, 0.1206)
(0.952, 0.1215)
(0.953, 0.1224)
(0.954, 0.1233)
(0.955, 0.1242)
(0.956, 0.1251)
(0.957, 0.1260)
(0.958, 0.1269)
(0.959, 0.1278)
(0.960, 0.1288)
(0.961, 0.1297)
(0.962, 0.1306)
(0.963, 0.1316)
(0.964, 0.1325)
(0.965, 0.1335)
(0.966, 0.1345)
(0.967, 0.1354)
(0.968, 0.1364)
(0.969, 0.1374)
(0.970, 0.1384)
(0.971, 0.1394)
(0.972, 0.1404)
(0.973, 0.1414)
(0.974, 0.1424)
(0.975, 0.1434)
(0.976, 0.1444)
(0.977, 0.1454)
(0.978, 0.1465)
(0.979, 0.1475)
(0.980, 0.1486)
(0.981, 0.1496)
(0.982, 0.1507)
(0.983, 0.1518)
(0.984, 0.1528)
(0.985, 0.1539)
(0.986, 0.1550)
(0.987, 0.1561)
(0.988, 0.1572)
(0.989, 0.1583)
(0.990, 0.1594)
(0.991, 0.1605)
(0.992, 0.1617)
(0.993, 0.1628)
(0.994, 0.1639)
(0.995, 0.1651)
(0.996, 0.1662)
(0.997, 0.1674)
(0.998, 0.1686)
(0.999, 0.1698)
(1.000, 0.1709)
(1.000, 0.0000)}
\defdata{emaprofile-target}{
(0.000, 0.0000)
(0.059, 0.0000)
(0.060, 0.0000)
(0.061, 0.0000)
(0.062, 0.0001)
(0.063, 0.0001)
(0.064, 0.0001)
(0.065, 0.0001)
(0.066, 0.0001)
(0.067, 0.0001)
(0.068, 0.0001)
(0.069, 0.0001)
(0.070, 0.0001)
(0.071, 0.0001)
(0.072, 0.0001)
(0.073, 0.0001)
(0.074, 0.0001)
(0.075, 0.0001)
(0.076, 0.0001)
(0.077, 0.0001)
(0.078, 0.0001)
(0.079, 0.0001)
(0.080, 0.0001)
(0.081, 0.0001)
(0.082, 0.0001)
(0.083, 0.0001)
(0.084, 0.0001)
(0.085, 0.0002)
(0.086, 0.0002)
(0.087, 0.0002)
(0.088, 0.0002)
(0.089, 0.0002)
(0.090, 0.0002)
(0.091, 0.0002)
(0.092, 0.0002)
(0.093, 0.0002)
(0.094, 0.0002)
(0.095, 0.0002)
(0.096, 0.0002)
(0.097, 0.0002)
(0.098, 0.0003)
(0.099, 0.0003)
(0.100, 0.0003)
(0.101, 0.0003)
(0.102, 0.0003)
(0.103, 0.0003)
(0.104, 0.0003)
(0.105, 0.0003)
(0.106, 0.0003)
(0.107, 0.0004)
(0.108, 0.0004)
(0.109, 0.0004)
(0.110, 0.0004)
(0.111, 0.0004)
(0.112, 0.0004)
(0.113, 0.0004)
(0.114, 0.0004)
(0.115, 0.0005)
(0.116, 0.0005)
(0.117, 0.0005)
(0.118, 0.0005)
(0.119, 0.0005)
(0.120, 0.0005)
(0.121, 0.0005)
(0.122, 0.0006)
(0.123, 0.0006)
(0.124, 0.0006)
(0.125, 0.0006)
(0.126, 0.0006)
(0.127, 0.0006)
(0.128, 0.0007)
(0.129, 0.0007)
(0.130, 0.0007)
(0.131, 0.0007)
(0.132, 0.0007)
(0.133, 0.0008)
(0.134, 0.0008)
(0.135, 0.0008)
(0.136, 0.0008)
(0.137, 0.0008)
(0.138, 0.0009)
(0.139, 0.0009)
(0.140, 0.0009)
(0.141, 0.0009)
(0.142, 0.0010)
(0.143, 0.0010)
(0.144, 0.0010)
(0.145, 0.0010)
(0.146, 0.0011)
(0.147, 0.0011)
(0.148, 0.0011)
(0.149, 0.0011)
(0.150, 0.0012)
(0.151, 0.0012)
(0.152, 0.0012)
(0.153, 0.0013)
(0.154, 0.0013)
(0.155, 0.0013)
(0.156, 0.0013)
(0.157, 0.0014)
(0.158, 0.0014)
(0.159, 0.0014)
(0.160, 0.0015)
(0.161, 0.0015)
(0.162, 0.0015)
(0.163, 0.0016)
(0.164, 0.0016)
(0.165, 0.0016)
(0.166, 0.0017)
(0.167, 0.0017)
(0.168, 0.0017)
(0.169, 0.0018)
(0.170, 0.0018)
(0.171, 0.0019)
(0.172, 0.0019)
(0.173, 0.0019)
(0.174, 0.0020)
(0.175, 0.0020)
(0.176, 0.0021)
(0.177, 0.0021)
(0.178, 0.0021)
(0.179, 0.0022)
(0.180, 0.0022)
(0.181, 0.0023)
(0.182, 0.0023)
(0.183, 0.0024)
(0.184, 0.0024)
(0.185, 0.0025)
(0.186, 0.0025)
(0.187, 0.0026)
(0.188, 0.0026)
(0.189, 0.0027)
(0.190, 0.0027)
(0.191, 0.0028)
(0.192, 0.0028)
(0.193, 0.0029)
(0.194, 0.0029)
(0.195, 0.0030)
(0.196, 0.0030)
(0.197, 0.0031)
(0.198, 0.0031)
(0.199, 0.0032)
(0.200, 0.0032)
(0.201, 0.0033)
(0.202, 0.0034)
(0.203, 0.0034)
(0.204, 0.0035)
(0.205, 0.0035)
(0.206, 0.0036)
(0.207, 0.0037)
(0.208, 0.0037)
(0.209, 0.0038)
(0.210, 0.0039)
(0.211, 0.0039)
(0.212, 0.0040)
(0.213, 0.0041)
(0.214, 0.0041)
(0.215, 0.0042)
(0.216, 0.0043)
(0.217, 0.0043)
(0.218, 0.0044)
(0.219, 0.0045)
(0.220, 0.0046)
(0.221, 0.0046)
(0.222, 0.0047)
(0.223, 0.0048)
(0.224, 0.0049)
(0.225, 0.0049)
(0.226, 0.0050)
(0.227, 0.0051)
(0.228, 0.0052)
(0.229, 0.0053)
(0.230, 0.0053)
(0.231, 0.0054)
(0.232, 0.0055)
(0.233, 0.0056)
(0.234, 0.0057)
(0.235, 0.0058)
(0.236, 0.0059)
(0.237, 0.0059)
(0.238, 0.0060)
(0.239, 0.0061)
(0.240, 0.0062)
(0.241, 0.0063)
(0.242, 0.0064)
(0.243, 0.0065)
(0.244, 0.0066)
(0.245, 0.0067)
(0.246, 0.0068)
(0.247, 0.0069)
(0.248, 0.0070)
(0.249, 0.0071)
(0.250, 0.0072)
(0.251, 0.0073)
(0.252, 0.0074)
(0.253, 0.0075)
(0.254, 0.0076)
(0.255, 0.0077)
(0.256, 0.0078)
(0.257, 0.0079)
(0.258, 0.0080)
(0.259, 0.0082)
(0.260, 0.0083)
(0.261, 0.0084)
(0.262, 0.0085)
(0.263, 0.0086)
(0.264, 0.0087)
(0.265, 0.0088)
(0.266, 0.0090)
(0.267, 0.0091)
(0.268, 0.0092)
(0.269, 0.0093)
(0.270, 0.0095)
(0.271, 0.0096)
(0.272, 0.0097)
(0.273, 0.0098)
(0.274, 0.0100)
(0.275, 0.0101)
(0.276, 0.0102)
(0.277, 0.0104)
(0.278, 0.0105)
(0.279, 0.0106)
(0.280, 0.0108)
(0.281, 0.0109)
(0.282, 0.0110)
(0.283, 0.0112)
(0.284, 0.0113)
(0.285, 0.0115)
(0.286, 0.0116)
(0.287, 0.0118)
(0.288, 0.0119)
(0.289, 0.0120)
(0.290, 0.0122)
(0.291, 0.0123)
(0.292, 0.0125)
(0.293, 0.0126)
(0.294, 0.0128)
(0.295, 0.0130)
(0.296, 0.0131)
(0.297, 0.0133)
(0.298, 0.0134)
(0.299, 0.0136)
(0.300, 0.0138)
(0.301, 0.0139)
(0.302, 0.0141)
(0.303, 0.0143)
(0.304, 0.0144)
(0.305, 0.0146)
(0.306, 0.0148)
(0.307, 0.0149)
(0.308, 0.0151)
(0.309, 0.0153)
(0.310, 0.0155)
(0.311, 0.0156)
(0.312, 0.0158)
(0.313, 0.0160)
(0.314, 0.0162)
(0.315, 0.0164)
(0.316, 0.0166)
(0.317, 0.0167)
(0.318, 0.0169)
(0.319, 0.0171)
(0.320, 0.0173)
(0.321, 0.0175)
(0.322, 0.0177)
(0.323, 0.0179)
(0.324, 0.0181)
(0.325, 0.0183)
(0.326, 0.0185)
(0.327, 0.0187)
(0.328, 0.0189)
(0.329, 0.0191)
(0.330, 0.0193)
(0.331, 0.0195)
(0.332, 0.0197)
(0.333, 0.0199)
(0.334, 0.0202)
(0.335, 0.0204)
(0.336, 0.0206)
(0.337, 0.0208)
(0.338, 0.0210)
(0.339, 0.0213)
(0.340, 0.0215)
(0.341, 0.0217)
(0.342, 0.0219)
(0.343, 0.0222)
(0.344, 0.0224)
(0.345, 0.0226)
(0.346, 0.0229)
(0.347, 0.0231)
(0.348, 0.0233)
(0.349, 0.0236)
(0.350, 0.0238)
(0.351, 0.0241)
(0.352, 0.0243)
(0.353, 0.0246)
(0.354, 0.0248)
(0.355, 0.0251)
(0.356, 0.0253)
(0.357, 0.0256)
(0.358, 0.0258)
(0.359, 0.0261)
(0.360, 0.0263)
(0.361, 0.0266)
(0.362, 0.0269)
(0.363, 0.0271)
(0.364, 0.0274)
(0.365, 0.0277)
(0.366, 0.0279)
(0.367, 0.0282)
(0.368, 0.0285)
(0.369, 0.0287)
(0.370, 0.0290)
(0.371, 0.0293)
(0.372, 0.0296)
(0.373, 0.0299)
(0.374, 0.0302)
(0.375, 0.0304)
(0.376, 0.0307)
(0.377, 0.0310)
(0.378, 0.0313)
(0.379, 0.0316)
(0.380, 0.0319)
(0.381, 0.0322)
(0.382, 0.0325)
(0.383, 0.0328)
(0.384, 0.0331)
(0.385, 0.0334)
(0.386, 0.0337)
(0.387, 0.0341)
(0.388, 0.0344)
(0.389, 0.0347)
(0.390, 0.0350)
(0.391, 0.0353)
(0.392, 0.0357)
(0.393, 0.0360)
(0.394, 0.0363)
(0.395, 0.0366)
(0.396, 0.0370)
(0.397, 0.0373)
(0.398, 0.0376)
(0.399, 0.0380)
(0.400, 0.0383)
(0.401, 0.0387)
(0.402, 0.0390)
(0.403, 0.0393)
(0.404, 0.0397)
(0.405, 0.0400)
(0.406, 0.0404)
(0.407, 0.0408)
(0.408, 0.0411)
(0.409, 0.0415)
(0.410, 0.0418)
(0.411, 0.0422)
(0.412, 0.0426)
(0.413, 0.0429)
(0.414, 0.0433)
(0.415, 0.0437)
(0.416, 0.0441)
(0.417, 0.0444)
(0.418, 0.0448)
(0.419, 0.0452)
(0.420, 0.0456)
(0.421, 0.0460)
(0.422, 0.0464)
(0.423, 0.0467)
(0.424, 0.0471)
(0.425, 0.0475)
(0.426, 0.0479)
(0.427, 0.0483)
(0.428, 0.0487)
(0.429, 0.0492)
(0.430, 0.0496)
(0.431, 0.0500)
(0.432, 0.0504)
(0.433, 0.0508)
(0.434, 0.0512)
(0.435, 0.0516)
(0.436, 0.0521)
(0.437, 0.0525)
(0.438, 0.0529)
(0.439, 0.0534)
(0.440, 0.0538)
(0.441, 0.0542)
(0.442, 0.0547)
(0.443, 0.0551)
(0.444, 0.0555)
(0.445, 0.0560)
(0.446, 0.0564)
(0.447, 0.0569)
(0.448, 0.0574)
(0.449, 0.0578)
(0.450, 0.0583)
(0.451, 0.0587)
(0.452, 0.0592)
(0.453, 0.0597)
(0.454, 0.0601)
(0.455, 0.0606)
(0.456, 0.0611)
(0.457, 0.0616)
(0.458, 0.0620)
(0.459, 0.0625)
(0.460, 0.0630)
(0.461, 0.0635)
(0.462, 0.0640)
(0.463, 0.0645)
(0.464, 0.0650)
(0.465, 0.0655)
(0.466, 0.0660)
(0.467, 0.0665)
(0.468, 0.0670)
(0.469, 0.0675)
(0.470, 0.0680)
(0.471, 0.0685)
(0.472, 0.0691)
(0.473, 0.0696)
(0.474, 0.0701)
(0.475, 0.0706)
(0.476, 0.0712)
(0.477, 0.0717)
(0.478, 0.0722)
(0.479, 0.0728)
(0.480, 0.0733)
(0.481, 0.0739)
(0.482, 0.0744)
(0.483, 0.0750)
(0.484, 0.0755)
(0.485, 0.0761)
(0.486, 0.0766)
(0.487, 0.0772)
(0.488, 0.0778)
(0.489, 0.0783)
(0.490, 0.0789)
(0.491, 0.0795)
(0.492, 0.0801)
(0.493, 0.0806)
(0.494, 0.0812)
(0.495, 0.0818)
(0.496, 0.0824)
(0.497, 0.0830)
(0.498, 0.0836)
(0.499, 0.0842)
(0.500, 0.0848)
(0.501, 0.0854)
(0.502, 0.0860)
(0.503, 0.0866)
(0.504, 0.0872)
(0.505, 0.0878)
(0.506, 0.0885)
(0.507, 0.0891)
(0.508, 0.0897)
(0.509, 0.0903)
(0.510, 0.0910)
(0.511, 0.0916)
(0.512, 0.0923)
(0.513, 0.0929)
(0.514, 0.0935)
(0.515, 0.0942)
(0.516, 0.0948)
(0.517, 0.0955)
(0.518, 0.0962)
(0.519, 0.0968)
(0.520, 0.0975)
(0.521, 0.0982)
(0.522, 0.0988)
(0.523, 0.0995)
(0.524, 0.1002)
(0.525, 0.1009)
(0.526, 0.1016)
(0.527, 0.1022)
(0.528, 0.1029)
(0.529, 0.1036)
(0.530, 0.1043)
(0.531, 0.1050)
(0.532, 0.1057)
(0.533, 0.1065)
(0.534, 0.1072)
(0.535, 0.1079)
(0.536, 0.1086)
(0.537, 0.1093)
(0.538, 0.1100)
(0.539, 0.1108)
(0.540, 0.1115)
(0.541, 0.1122)
(0.542, 0.1130)
(0.543, 0.1137)
(0.544, 0.1145)
(0.545, 0.1152)
(0.546, 0.1160)
(0.547, 0.1167)
(0.548, 0.1175)
(0.549, 0.1183)
(0.550, 0.1190)
(0.551, 0.1198)
(0.552, 0.1206)
(0.553, 0.1214)
(0.554, 0.1222)
(0.555, 0.1229)
(0.556, 0.1237)
(0.557, 0.1245)
(0.558, 0.1253)
(0.559, 0.1261)
(0.560, 0.1269)
(0.561, 0.1277)
(0.562, 0.1285)
(0.563, 0.1294)
(0.564, 0.1302)
(0.565, 0.1310)
(0.566, 0.1318)
(0.567, 0.1327)
(0.568, 0.1335)
(0.569, 0.1343)
(0.570, 0.1352)
(0.571, 0.1360)
(0.572, 0.1369)
(0.573, 0.1377)
(0.574, 0.1386)
(0.575, 0.1394)
(0.576, 0.1403)
(0.577, 0.1412)
(0.578, 0.1421)
(0.579, 0.1429)
(0.580, 0.1438)
(0.581, 0.1447)
(0.582, 0.1456)
(0.583, 0.1465)
(0.584, 0.1474)
(0.585, 0.1483)
(0.586, 0.1492)
(0.587, 0.1501)
(0.588, 0.1510)
(0.589, 0.1519)
(0.590, 0.1528)
(0.591, 0.1538)
(0.592, 0.1547)
(0.593, 0.1556)
(0.594, 0.1566)
(0.595, 0.1575)
(0.596, 0.1584)
(0.597, 0.1594)
(0.598, 0.1603)
(0.599, 0.1613)
(0.600, 0.1623)
(0.601, 0.1632)
(0.602, 0.1642)
(0.603, 0.1652)
(0.604, 0.1661)
(0.605, 0.1671)
(0.606, 0.1681)
(0.607, 0.1691)
(0.608, 0.1701)
(0.609, 0.1711)
(0.610, 0.1721)
(0.611, 0.1731)
(0.612, 0.1741)
(0.613, 0.1751)
(0.614, 0.1761)
(0.615, 0.1772)
(0.616, 0.1782)
(0.617, 0.1792)
(0.618, 0.1803)
(0.619, 0.1813)
(0.620, 0.1824)
(0.621, 0.1834)
(0.622, 0.1845)
(0.623, 0.1855)
(0.624, 0.1866)
(0.625, 0.1876)
(0.626, 0.1887)
(0.627, 0.1898)
(0.628, 0.1909)
(0.629, 0.1920)
(0.630, 0.1930)
(0.631, 0.1941)
(0.632, 0.1952)
(0.633, 0.1963)
(0.634, 0.1974)
(0.635, 0.1986)
(0.636, 0.1997)
(0.637, 0.2008)
(0.638, 0.2019)
(0.639, 0.2030)
(0.640, 0.2042)
(0.641, 0.2053)
(0.642, 0.2065)
(0.643, 0.2076)
(0.644, 0.2088)
(0.645, 0.2099)
(0.646, 0.2111)
(0.647, 0.2122)
(0.648, 0.2134)
(0.649, 0.2146)
(0.650, 0.2158)
(0.651, 0.2169)
(0.652, 0.2181)
(0.653, 0.2193)
(0.654, 0.2205)
(0.655, 0.2217)
(0.656, 0.2229)
(0.657, 0.2241)
(0.658, 0.2254)
(0.659, 0.2266)
(0.660, 0.2278)
(0.661, 0.2290)
(0.662, 0.2303)
(0.663, 0.2315)
(0.664, 0.2328)
(0.665, 0.2340)
(0.666, 0.2353)
(0.667, 0.2365)
(0.668, 0.2378)
(0.669, 0.2391)
(0.670, 0.2403)
(0.671, 0.2416)
(0.672, 0.2429)
(0.673, 0.2442)
(0.674, 0.2455)
(0.675, 0.2468)
(0.676, 0.2481)
(0.677, 0.2494)
(0.678, 0.2507)
(0.679, 0.2520)
(0.680, 0.2534)
(0.681, 0.2547)
(0.682, 0.2560)
(0.683, 0.2574)
(0.684, 0.2587)
(0.685, 0.2601)
(0.686, 0.2614)
(0.687, 0.2628)
(0.688, 0.2641)
(0.689, 0.2655)
(0.690, 0.2669)
(0.691, 0.2683)
(0.692, 0.2696)
(0.693, 0.2710)
(0.694, 0.2724)
(0.695, 0.2738)
(0.696, 0.2752)
(0.697, 0.2766)
(0.698, 0.2781)
(0.699, 0.2795)
(0.700, 0.2809)
(0.701, 0.2823)
(0.702, 0.2838)
(0.703, 0.2852)
(0.704, 0.2867)
(0.705, 0.2881)
(0.706, 0.2896)
(0.707, 0.2910)
(0.708, 0.2925)
(0.709, 0.2940)
(0.710, 0.2954)
(0.711, 0.2969)
(0.712, 0.2984)
(0.713, 0.2999)
(0.714, 0.3014)
(0.715, 0.3029)
(0.716, 0.3044)
(0.717, 0.3059)
(0.718, 0.3075)
(0.719, 0.3090)
(0.720, 0.3105)
(0.721, 0.3121)
(0.722, 0.3136)
(0.723, 0.3152)
(0.724, 0.3167)
(0.725, 0.3183)
(0.726, 0.3198)
(0.727, 0.3214)
(0.728, 0.3230)
(0.729, 0.3246)
(0.730, 0.3262)
(0.731, 0.3278)
(0.732, 0.3294)
(0.733, 0.3310)
(0.734, 0.3326)
(0.735, 0.3342)
(0.736, 0.3358)
(0.737, 0.3374)
(0.738, 0.3391)
(0.739, 0.3407)
(0.740, 0.3423)
(0.741, 0.3440)
(0.742, 0.3457)
(0.743, 0.3473)
(0.744, 0.3490)
(0.745, 0.3507)
(0.746, 0.3523)
(0.747, 0.3540)
(0.748, 0.3557)
(0.749, 0.3574)
(0.750, 0.3591)
(0.751, 0.3608)
(0.752, 0.3625)
(0.753, 0.3642)
(0.754, 0.3660)
(0.755, 0.3677)
(0.756, 0.3694)
(0.757, 0.3712)
(0.758, 0.3729)
(0.759, 0.3747)
(0.760, 0.3764)
(0.761, 0.3782)
(0.762, 0.3800)
(0.763, 0.3818)
(0.764, 0.3835)
(0.765, 0.3853)
(0.766, 0.3871)
(0.767, 0.3889)
(0.768, 0.3907)
(0.769, 0.3926)
(0.770, 0.3944)
(0.771, 0.3962)
(0.772, 0.3980)
(0.773, 0.3999)
(0.774, 0.4017)
(0.775, 0.4036)
(0.776, 0.4054)
(0.777, 0.4073)
(0.778, 0.4092)
(0.779, 0.4110)
(0.780, 0.4129)
(0.781, 0.4148)
(0.782, 0.4167)
(0.783, 0.4186)
(0.784, 0.4205)
(0.785, 0.4224)
(0.786, 0.4243)
(0.787, 0.4263)
(0.788, 0.4282)
(0.789, 0.4301)
(0.790, 0.4321)
(0.791, 0.4340)
(0.792, 0.4360)
(0.793, 0.4379)
(0.794, 0.4399)
(0.795, 0.4419)
(0.796, 0.4439)
(0.797, 0.4459)
(0.798, 0.4478)
(0.799, 0.4498)
(0.800, 0.4519)
(0.801, 0.4539)
(0.802, 0.4559)
(0.803, 0.4579)
(0.804, 0.4600)
(0.805, 0.4620)
(0.806, 0.4640)
(0.807, 0.4661)
(0.808, 0.4681)
(0.809, 0.4702)
(0.810, 0.4723)
(0.811, 0.4744)
(0.812, 0.4765)
(0.813, 0.4785)
(0.814, 0.4806)
(0.815, 0.4827)
(0.816, 0.4849)
(0.817, 0.4870)
(0.818, 0.4891)
(0.819, 0.4912)
(0.820, 0.4934)
(0.821, 0.4955)
(0.822, 0.4977)
(0.823, 0.4998)
(0.824, 0.5020)
(0.825, 0.5042)
(0.826, 0.5063)
(0.827, 0.5085)
(0.828, 0.5107)
(0.829, 0.5129)
(0.830, 0.5151)
(0.831, 0.5173)
(0.832, 0.5196)
(0.833, 0.5218)
(0.834, 0.5240)
(0.835, 0.5263)
(0.836, 0.5285)
(0.837, 0.5308)
(0.838, 0.5330)
(0.839, 0.5353)
(0.840, 0.5376)
(0.841, 0.5399)
(0.842, 0.5421)
(0.843, 0.5444)
(0.844, 0.5467)
(0.845, 0.5490)
(0.846, 0.5514)
(0.847, 0.5537)
(0.848, 0.5560)
(0.849, 0.5584)
(0.850, 0.5607)
(0.851, 0.5631)
(0.852, 0.5654)
(0.853, 0.5678)
(0.854, 0.5702)
(0.855, 0.5725)
(0.856, 0.5749)
(0.857, 0.5773)
(0.858, 0.5797)
(0.859, 0.5821)
(0.860, 0.5845)
(0.861, 0.5870)
(0.862, 0.5894)
(0.863, 0.5918)
(0.864, 0.5943)
(0.865, 0.5967)
(0.866, 0.5992)
(0.867, 0.6017)
(0.868, 0.6041)
(0.869, 0.6066)
(0.870, 0.6091)
(0.871, 0.6116)
(0.872, 0.6141)
(0.873, 0.6166)
(0.874, 0.6191)
(0.875, 0.6217)
(0.876, 0.6242)
(0.877, 0.6267)
(0.878, 0.6293)
(0.879, 0.6318)
(0.880, 0.6344)
(0.881, 0.6370)
(0.882, 0.6395)
(0.883, 0.6421)
(0.884, 0.6447)
(0.885, 0.6473)
(0.886, 0.6499)
(0.887, 0.6525)
(0.888, 0.6552)
(0.889, 0.6578)
(0.890, 0.6604)
(0.891, 0.6631)
(0.892, 0.6657)
(0.893, 0.6684)
(0.894, 0.6711)
(0.895, 0.6737)
(0.896, 0.6764)
(0.897, 0.6791)
(0.898, 0.6818)
(0.899, 0.6845)
(0.900, 0.6872)
(0.901, 0.6900)
(0.902, 0.6927)
(0.903, 0.6954)
(0.904, 0.6982)
(0.905, 0.7009)
(0.906, 0.7037)
(0.907, 0.7065)
(0.908, 0.7092)
(0.909, 0.7120)
(0.910, 0.7148)
(0.911, 0.7176)
(0.912, 0.7204)
(0.913, 0.7232)
(0.914, 0.7261)
(0.915, 0.7289)
(0.916, 0.7317)
(0.917, 0.7346)
(0.918, 0.7374)
(0.919, 0.7403)
(0.920, 0.7432)
(0.921, 0.7460)
(0.922, 0.7489)
(0.923, 0.7518)
(0.924, 0.7547)
(0.925, 0.7576)
(0.926, 0.7606)
(0.927, 0.7635)
(0.928, 0.7664)
(0.929, 0.7694)
(0.930, 0.7723)
(0.931, 0.7753)
(0.932, 0.7783)
(0.933, 0.7812)
(0.934, 0.7842)
(0.935, 0.7872)
(0.936, 0.7902)
(0.937, 0.7932)
(0.938, 0.7962)
(0.939, 0.7993)
(0.940, 0.8023)
(0.941, 0.8053)
(0.942, 0.8084)
(0.943, 0.8114)
(0.944, 0.8145)
(0.945, 0.8176)
(0.946, 0.8207)
(0.947, 0.8238)
(0.948, 0.8269)
(0.949, 0.8300)
(0.950, 0.8331)
(0.951, 0.8362)
(0.952, 0.8394)
(0.953, 0.8425)
(0.954, 0.8457)
(0.955, 0.8488)
(0.956, 0.8520)
(0.957, 0.8552)
(0.958, 0.8583)
(0.959, 0.8615)
(0.960, 0.8647)
(0.961, 0.8680)
(0.962, 0.8712)
(0.963, 0.8744)
(0.964, 0.8776)
(0.965, 0.8809)
(0.966, 0.8841)
(0.967, 0.8874)
(0.968, 0.8907)
(0.969, 0.8939)
(0.970, 0.8972)
(0.971, 0.9005)
(0.972, 0.9038)
(0.973, 0.9072)
(0.974, 0.9105)
(0.975, 0.9138)
(0.976, 0.9172)
(0.977, 0.9205)
(0.978, 0.9239)
(0.979, 0.9272)
(0.980, 0.9306)
(0.981, 0.9340)
(0.982, 0.9374)
(0.983, 0.9408)
(0.984, 0.9442)
(0.985, 0.9476)
(0.986, 0.9510)
(0.987, 0.9545)
(0.988, 0.9579)
(0.989, 0.9614)
(0.990, 0.9649)
(0.991, 0.9683)
(0.992, 0.9718)
(0.993, 0.9753)
(0.994, 0.9788)
(0.995, 0.9823)
(0.996, 0.9858)
(0.997, 0.9894)
(0.998, 0.9929)
(0.999, 0.9964)
(1.000, 1.0000)
(1.000, 0.0000)}
\defdata{emaprofile-fit}{
(0.000, 0.0000)
(0.077, 0.0000)
(0.078, 0.0000)
(0.079, 0.0000)
(0.080, 0.0001)
(0.081, 0.0001)
(0.082, 0.0001)
(0.083, 0.0001)
(0.084, 0.0001)
(0.085, 0.0001)
(0.086, 0.0001)
(0.087, 0.0001)
(0.088, 0.0001)
(0.089, 0.0001)
(0.090, 0.0001)
(0.091, 0.0001)
(0.092, 0.0001)
(0.093, 0.0002)
(0.094, 0.0002)
(0.095, 0.0002)
(0.096, 0.0002)
(0.097, 0.0002)
(0.098, 0.0002)
(0.099, 0.0002)
(0.100, 0.0002)
(0.101, 0.0003)
(0.102, 0.0003)
(0.103, 0.0003)
(0.104, 0.0003)
(0.105, 0.0003)
(0.106, 0.0003)
(0.107, 0.0004)
(0.108, 0.0004)
(0.109, 0.0004)
(0.110, 0.0004)
(0.111, 0.0004)
(0.112, 0.0005)
(0.113, 0.0005)
(0.114, 0.0005)
(0.115, 0.0005)
(0.116, 0.0005)
(0.117, 0.0005)
(0.118, 0.0005)
(0.119, 0.0006)
(0.120, 0.0006)
(0.121, 0.0006)
(0.122, 0.0006)
(0.123, 0.0006)
(0.124, 0.0005)
(0.125, 0.0005)
(0.126, 0.0001)
(0.127, 0.0001)
(0.128, 0.0001)
(0.129, 0.0001)
(0.130, 0.0002)
(0.131, 0.0002)
(0.132, 0.0002)
(0.133, 0.0002)
(0.134, 0.0002)
(0.135, 0.0002)
(0.136, 0.0002)
(0.137, 0.0002)
(0.138, 0.0002)
(0.139, 0.0002)
(0.140, 0.0003)
(0.141, 0.0003)
(0.142, 0.0003)
(0.143, 0.0003)
(0.144, 0.0003)
(0.145, 0.0003)
(0.146, 0.0003)
(0.147, 0.0004)
(0.148, 0.0004)
(0.149, 0.0004)
(0.150, 0.0004)
(0.151, 0.0004)
(0.152, 0.0005)
(0.153, 0.0005)
(0.154, 0.0005)
(0.155, 0.0005)
(0.156, 0.0005)
(0.157, 0.0006)
(0.158, 0.0006)
(0.159, 0.0006)
(0.160, 0.0006)
(0.161, 0.0007)
(0.162, 0.0007)
(0.163, 0.0007)
(0.164, 0.0008)
(0.165, 0.0008)
(0.166, 0.0008)
(0.167, 0.0009)
(0.168, 0.0009)
(0.169, 0.0009)
(0.170, 0.0010)
(0.171, 0.0010)
(0.172, 0.0011)
(0.173, 0.0011)
(0.174, 0.0011)
(0.175, 0.0012)
(0.176, 0.0012)
(0.177, 0.0013)
(0.178, 0.0013)
(0.179, 0.0014)
(0.180, 0.0014)
(0.181, 0.0015)
(0.182, 0.0015)
(0.183, 0.0016)
(0.184, 0.0017)
(0.185, 0.0017)
(0.186, 0.0018)
(0.187, 0.0018)
(0.188, 0.0019)
(0.189, 0.0020)
(0.190, 0.0020)
(0.191, 0.0021)
(0.192, 0.0022)
(0.193, 0.0023)
(0.194, 0.0023)
(0.195, 0.0024)
(0.196, 0.0025)
(0.197, 0.0026)
(0.198, 0.0027)
(0.199, 0.0028)
(0.200, 0.0028)
(0.201, 0.0029)
(0.202, 0.0030)
(0.203, 0.0031)
(0.204, 0.0032)
(0.205, 0.0033)
(0.206, 0.0034)
(0.207, 0.0035)
(0.208, 0.0036)
(0.209, 0.0037)
(0.210, 0.0038)
(0.211, 0.0039)
(0.212, 0.0040)
(0.213, 0.0042)
(0.214, 0.0043)
(0.215, 0.0044)
(0.216, 0.0045)
(0.217, 0.0046)
(0.218, 0.0047)
(0.219, 0.0048)
(0.220, 0.0049)
(0.221, 0.0050)
(0.222, 0.0052)
(0.223, 0.0053)
(0.224, 0.0054)
(0.225, 0.0055)
(0.226, 0.0056)
(0.227, 0.0057)
(0.228, 0.0058)
(0.229, 0.0059)
(0.230, 0.0060)
(0.231, 0.0061)
(0.232, 0.0062)
(0.233, 0.0063)
(0.234, 0.0063)
(0.235, 0.0064)
(0.236, 0.0065)
(0.237, 0.0065)
(0.238, 0.0066)
(0.239, 0.0066)
(0.240, 0.0066)
(0.241, 0.0066)
(0.242, 0.0066)
(0.243, 0.0066)
(0.244, 0.0066)
(0.245, 0.0065)
(0.246, 0.0065)
(0.247, 0.0064)
(0.248, 0.0063)
(0.249, 0.0062)
(0.250, 0.0060)
(0.251, 0.0036)
(0.252, 0.0037)
(0.253, 0.0038)
(0.254, 0.0039)
(0.255, 0.0040)
(0.256, 0.0041)
(0.257, 0.0043)
(0.258, 0.0044)
(0.259, 0.0045)
(0.260, 0.0046)
(0.261, 0.0047)
(0.262, 0.0048)
(0.263, 0.0050)
(0.264, 0.0051)
(0.265, 0.0052)
(0.266, 0.0054)
(0.267, 0.0055)
(0.268, 0.0056)
(0.269, 0.0058)
(0.270, 0.0059)
(0.271, 0.0061)
(0.272, 0.0062)
(0.273, 0.0064)
(0.274, 0.0066)
(0.275, 0.0067)
(0.276, 0.0069)
(0.277, 0.0070)
(0.278, 0.0072)
(0.279, 0.0074)
(0.280, 0.0076)
(0.281, 0.0077)
(0.282, 0.0079)
(0.283, 0.0081)
(0.284, 0.0083)
(0.285, 0.0085)
(0.286, 0.0087)
(0.287, 0.0089)
(0.288, 0.0091)
(0.289, 0.0093)
(0.290, 0.0095)
(0.291, 0.0097)
(0.292, 0.0100)
(0.293, 0.0102)
(0.294, 0.0104)
(0.295, 0.0106)
(0.296, 0.0109)
(0.297, 0.0111)
(0.298, 0.0113)
(0.299, 0.0116)
(0.300, 0.0118)
(0.301, 0.0121)
(0.302, 0.0123)
(0.303, 0.0126)
(0.304, 0.0129)
(0.305, 0.0131)
(0.306, 0.0134)
(0.307, 0.0137)
(0.308, 0.0139)
(0.309, 0.0142)
(0.310, 0.0145)
(0.311, 0.0148)
(0.312, 0.0151)
(0.313, 0.0153)
(0.314, 0.0156)
(0.315, 0.0159)
(0.316, 0.0162)
(0.317, 0.0165)
(0.318, 0.0168)
(0.319, 0.0171)
(0.320, 0.0174)
(0.321, 0.0178)
(0.322, 0.0181)
(0.323, 0.0184)
(0.324, 0.0187)
(0.325, 0.0190)
(0.326, 0.0193)
(0.327, 0.0196)
(0.328, 0.0199)
(0.329, 0.0203)
(0.330, 0.0206)
(0.331, 0.0209)
(0.332, 0.0212)
(0.333, 0.0215)
(0.334, 0.0218)
(0.335, 0.0222)
(0.336, 0.0225)
(0.337, 0.0228)
(0.338, 0.0231)
(0.339, 0.0234)
(0.340, 0.0237)
(0.341, 0.0240)
(0.342, 0.0243)
(0.343, 0.0245)
(0.344, 0.0248)
(0.345, 0.0251)
(0.346, 0.0254)
(0.347, 0.0256)
(0.348, 0.0259)
(0.349, 0.0261)
(0.350, 0.0263)
(0.351, 0.0265)
(0.352, 0.0268)
(0.353, 0.0270)
(0.354, 0.0271)
(0.355, 0.0273)
(0.356, 0.0275)
(0.357, 0.0276)
(0.358, 0.0277)
(0.359, 0.0278)
(0.360, 0.0279)
(0.361, 0.0279)
(0.362, 0.0280)
(0.363, 0.0280)
(0.364, 0.0280)
(0.365, 0.0280)
(0.366, 0.0279)
(0.367, 0.0278)
(0.368, 0.0277)
(0.369, 0.0275)
(0.370, 0.0273)
(0.371, 0.0271)
(0.372, 0.0268)
(0.373, 0.0265)
(0.374, 0.0262)
(0.375, 0.0258)
(0.376, 0.0212)
(0.377, 0.0216)
(0.378, 0.0220)
(0.379, 0.0224)
(0.380, 0.0227)
(0.381, 0.0231)
(0.382, 0.0235)
(0.383, 0.0239)
(0.384, 0.0244)
(0.385, 0.0248)
(0.386, 0.0252)
(0.387, 0.0256)
(0.388, 0.0261)
(0.389, 0.0265)
(0.390, 0.0270)
(0.391, 0.0274)
(0.392, 0.0279)
(0.393, 0.0283)
(0.394, 0.0288)
(0.395, 0.0293)
(0.396, 0.0297)
(0.397, 0.0302)
(0.398, 0.0307)
(0.399, 0.0312)
(0.400, 0.0317)
(0.401, 0.0322)
(0.402, 0.0327)
(0.403, 0.0332)
(0.404, 0.0338)
(0.405, 0.0343)
(0.406, 0.0348)
(0.407, 0.0354)
(0.408, 0.0359)
(0.409, 0.0364)
(0.410, 0.0370)
(0.411, 0.0376)
(0.412, 0.0381)
(0.413, 0.0387)
(0.414, 0.0393)
(0.415, 0.0398)
(0.416, 0.0404)
(0.417, 0.0410)
(0.418, 0.0416)
(0.419, 0.0422)
(0.420, 0.0428)
(0.421, 0.0434)
(0.422, 0.0440)
(0.423, 0.0446)
(0.424, 0.0452)
(0.425, 0.0459)
(0.426, 0.0465)
(0.427, 0.0471)
(0.428, 0.0477)
(0.429, 0.0484)
(0.430, 0.0490)
(0.431, 0.0497)
(0.432, 0.0503)
(0.433, 0.0510)
(0.434, 0.0516)
(0.435, 0.0523)
(0.436, 0.0529)
(0.437, 0.0536)
(0.438, 0.0542)
(0.439, 0.0549)
(0.440, 0.0555)
(0.441, 0.0562)
(0.442, 0.0568)
(0.443, 0.0575)
(0.444, 0.0582)
(0.445, 0.0588)
(0.446, 0.0595)
(0.447, 0.0601)
(0.448, 0.0608)
(0.449, 0.0614)
(0.450, 0.0621)
(0.451, 0.0627)
(0.452, 0.0633)
(0.453, 0.0640)
(0.454, 0.0646)
(0.455, 0.0652)
(0.456, 0.0659)
(0.457, 0.0665)
(0.458, 0.0671)
(0.459, 0.0677)
(0.460, 0.0683)
(0.461, 0.0688)
(0.462, 0.0694)
(0.463, 0.0700)
(0.464, 0.0705)
(0.465, 0.0711)
(0.466, 0.0716)
(0.467, 0.0721)
(0.468, 0.0726)
(0.469, 0.0731)
(0.470, 0.0735)
(0.471, 0.0740)
(0.472, 0.0744)
(0.473, 0.0748)
(0.474, 0.0752)
(0.475, 0.0756)
(0.476, 0.0760)
(0.477, 0.0763)
(0.478, 0.0766)
(0.479, 0.0769)
(0.480, 0.0771)
(0.481, 0.0774)
(0.482, 0.0775)
(0.483, 0.0777)
(0.484, 0.0778)
(0.485, 0.0779)
(0.486, 0.0780)
(0.487, 0.0780)
(0.488, 0.0780)
(0.489, 0.0780)
(0.490, 0.0779)
(0.491, 0.0777)
(0.492, 0.0775)
(0.493, 0.0773)
(0.494, 0.0770)
(0.495, 0.0767)
(0.496, 0.0763)
(0.497, 0.0759)
(0.498, 0.0754)
(0.499, 0.0748)
(0.500, 0.0742)
(0.501, 0.0683)
(0.502, 0.0691)
(0.503, 0.0700)
(0.504, 0.0709)
(0.505, 0.0718)
(0.506, 0.0727)
(0.507, 0.0736)
(0.508, 0.0745)
(0.509, 0.0755)
(0.510, 0.0764)
(0.511, 0.0773)
(0.512, 0.0783)
(0.513, 0.0792)
(0.514, 0.0802)
(0.515, 0.0812)
(0.516, 0.0821)
(0.517, 0.0831)
(0.518, 0.0841)
(0.519, 0.0851)
(0.520, 0.0861)
(0.521, 0.0871)
(0.522, 0.0882)
(0.523, 0.0892)
(0.524, 0.0902)
(0.525, 0.0913)
(0.526, 0.0923)
(0.527, 0.0934)
(0.528, 0.0944)
(0.529, 0.0955)
(0.530, 0.0965)
(0.531, 0.0976)
(0.532, 0.0987)
(0.533, 0.0998)
(0.534, 0.1009)
(0.535, 0.1020)
(0.536, 0.1031)
(0.537, 0.1042)
(0.538, 0.1053)
(0.539, 0.1064)
(0.540, 0.1076)
(0.541, 0.1087)
(0.542, 0.1098)
(0.543, 0.1109)
(0.544, 0.1121)
(0.545, 0.1132)
(0.546, 0.1144)
(0.547, 0.1155)
(0.548, 0.1167)
(0.549, 0.1178)
(0.550, 0.1190)
(0.551, 0.1201)
(0.552, 0.1213)
(0.553, 0.1225)
(0.554, 0.1236)
(0.555, 0.1248)
(0.556, 0.1259)
(0.557, 0.1271)
(0.558, 0.1283)
(0.559, 0.1294)
(0.560, 0.1306)
(0.561, 0.1318)
(0.562, 0.1329)
(0.563, 0.1341)
(0.564, 0.1352)
(0.565, 0.1364)
(0.566, 0.1375)
(0.567, 0.1386)
(0.568, 0.1398)
(0.569, 0.1409)
(0.570, 0.1420)
(0.571, 0.1431)
(0.572, 0.1443)
(0.573, 0.1454)
(0.574, 0.1465)
(0.575, 0.1475)
(0.576, 0.1486)
(0.577, 0.1497)
(0.578, 0.1507)
(0.579, 0.1518)
(0.580, 0.1528)
(0.581, 0.1538)
(0.582, 0.1548)
(0.583, 0.1558)
(0.584, 0.1568)
(0.585, 0.1577)
(0.586, 0.1587)
(0.587, 0.1596)
(0.588, 0.1605)
(0.589, 0.1614)
(0.590, 0.1623)
(0.591, 0.1631)
(0.592, 0.1639)
(0.593, 0.1647)
(0.594, 0.1655)
(0.595, 0.1662)
(0.596, 0.1669)
(0.597, 0.1676)
(0.598, 0.1683)
(0.599, 0.1689)
(0.600, 0.1695)
(0.601, 0.1701)
(0.602, 0.1706)
(0.603, 0.1711)
(0.604, 0.1715)
(0.605, 0.1719)
(0.606, 0.1723)
(0.607, 0.1727)
(0.608, 0.1729)
(0.609, 0.1732)
(0.610, 0.1734)
(0.611, 0.1735)
(0.612, 0.1736)
(0.613, 0.1737)
(0.614, 0.1737)
(0.615, 0.1736)
(0.616, 0.1735)
(0.617, 0.1733)
(0.618, 0.1731)
(0.619, 0.1728)
(0.620, 0.1724)
(0.621, 0.1719)
(0.622, 0.1714)
(0.623, 0.1709)
(0.624, 0.1702)
(0.625, 0.1695)
(0.626, 0.1628)
(0.627, 0.1644)
(0.628, 0.1660)
(0.629, 0.1676)
(0.630, 0.1692)
(0.631, 0.1709)
(0.632, 0.1725)
(0.633, 0.1742)
(0.634, 0.1758)
(0.635, 0.1775)
(0.636, 0.1792)
(0.637, 0.1809)
(0.638, 0.1825)
(0.639, 0.1843)
(0.640, 0.1860)
(0.641, 0.1877)
(0.642, 0.1894)
(0.643, 0.1911)
(0.644, 0.1929)
(0.645, 0.1946)
(0.646, 0.1964)
(0.647, 0.1981)
(0.648, 0.1999)
(0.649, 0.2017)
(0.650, 0.2035)
(0.651, 0.2053)
(0.652, 0.2070)
(0.653, 0.2088)
(0.654, 0.2107)
(0.655, 0.2125)
(0.656, 0.2143)
(0.657, 0.2161)
(0.658, 0.2179)
(0.659, 0.2198)
(0.660, 0.2216)
(0.661, 0.2234)
(0.662, 0.2253)
(0.663, 0.2271)
(0.664, 0.2290)
(0.665, 0.2308)
(0.666, 0.2327)
(0.667, 0.2345)
(0.668, 0.2364)
(0.669, 0.2382)
(0.670, 0.2401)
(0.671, 0.2419)
(0.672, 0.2438)
(0.673, 0.2456)
(0.674, 0.2475)
(0.675, 0.2494)
(0.676, 0.2512)
(0.677, 0.2531)
(0.678, 0.2549)
(0.679, 0.2568)
(0.680, 0.2586)
(0.681, 0.2605)
(0.682, 0.2623)
(0.683, 0.2641)
(0.684, 0.2659)
(0.685, 0.2678)
(0.686, 0.2696)
(0.687, 0.2714)
(0.688, 0.2732)
(0.689, 0.2750)
(0.690, 0.2768)
(0.691, 0.2785)
(0.692, 0.2803)
(0.693, 0.2820)
(0.694, 0.2838)
(0.695, 0.2855)
(0.696, 0.2872)
(0.697, 0.2889)
(0.698, 0.2906)
(0.699, 0.2923)
(0.700, 0.2939)
(0.701, 0.2956)
(0.702, 0.2972)
(0.703, 0.2988)
(0.704, 0.3004)
(0.705, 0.3020)
(0.706, 0.3035)
(0.707, 0.3050)
(0.708, 0.3065)
(0.709, 0.3080)
(0.710, 0.3094)
(0.711, 0.3109)
(0.712, 0.3123)
(0.713, 0.3136)
(0.714, 0.3150)
(0.715, 0.3163)
(0.716, 0.3176)
(0.717, 0.3188)
(0.718, 0.3200)
(0.719, 0.3212)
(0.720, 0.3223)
(0.721, 0.3234)
(0.722, 0.3245)
(0.723, 0.3255)
(0.724, 0.3265)
(0.725, 0.3274)
(0.726, 0.3283)
(0.727, 0.3292)
(0.728, 0.3300)
(0.729, 0.3307)
(0.730, 0.3314)
(0.731, 0.3321)
(0.732, 0.3327)
(0.733, 0.3332)
(0.734, 0.3337)
(0.735, 0.3341)
(0.736, 0.3345)
(0.737, 0.3348)
(0.738, 0.3350)
(0.739, 0.3352)
(0.740, 0.3353)
(0.741, 0.3353)
(0.742, 0.3353)
(0.743, 0.3352)
(0.744, 0.3350)
(0.745, 0.3347)
(0.746, 0.3344)
(0.747, 0.3339)
(0.748, 0.3334)
(0.749, 0.3328)
(0.750, 0.3321)
(0.751, 0.3251)
(0.752, 0.3277)
(0.753, 0.3302)
(0.754, 0.3328)
(0.755, 0.3354)
(0.756, 0.3380)
(0.757, 0.3406)
(0.758, 0.3432)
(0.759, 0.3458)
(0.760, 0.3485)
(0.761, 0.3511)
(0.762, 0.3538)
(0.763, 0.3564)
(0.764, 0.3591)
(0.765, 0.3618)
(0.766, 0.3644)
(0.767, 0.3671)
(0.768, 0.3698)
(0.769, 0.3725)
(0.770, 0.3752)
(0.771, 0.3779)
(0.772, 0.3807)
(0.773, 0.3834)
(0.774, 0.3861)
(0.775, 0.3889)
(0.776, 0.3916)
(0.777, 0.3943)
(0.778, 0.3971)
(0.779, 0.3998)
(0.780, 0.4026)
(0.781, 0.4053)
(0.782, 0.4081)
(0.783, 0.4109)
(0.784, 0.4136)
(0.785, 0.4164)
(0.786, 0.4192)
(0.787, 0.4219)
(0.788, 0.4247)
(0.789, 0.4275)
(0.790, 0.4302)
(0.791, 0.4330)
(0.792, 0.4358)
(0.793, 0.4385)
(0.794, 0.4413)
(0.795, 0.4441)
(0.796, 0.4468)
(0.797, 0.4496)
(0.798, 0.4523)
(0.799, 0.4551)
(0.800, 0.4578)
(0.801, 0.4605)
(0.802, 0.4632)
(0.803, 0.4660)
(0.804, 0.4687)
(0.805, 0.4714)
(0.806, 0.4741)
(0.807, 0.4768)
(0.808, 0.4794)
(0.809, 0.4821)
(0.810, 0.4847)
(0.811, 0.4874)
(0.812, 0.4900)
(0.813, 0.4926)
(0.814, 0.4952)
(0.815, 0.4978)
(0.816, 0.5004)
(0.817, 0.5029)
(0.818, 0.5054)
(0.819, 0.5080)
(0.820, 0.5104)
(0.821, 0.5129)
(0.822, 0.5154)
(0.823, 0.5178)
(0.824, 0.5202)
(0.825, 0.5226)
(0.826, 0.5250)
(0.827, 0.5273)
(0.828, 0.5296)
(0.829, 0.5319)
(0.830, 0.5341)
(0.831, 0.5364)
(0.832, 0.5386)
(0.833, 0.5407)
(0.834, 0.5429)
(0.835, 0.5449)
(0.836, 0.5470)
(0.837, 0.5490)
(0.838, 0.5510)
(0.839, 0.5530)
(0.840, 0.5549)
(0.841, 0.5567)
(0.842, 0.5586)
(0.843, 0.5603)
(0.844, 0.5621)
(0.845, 0.5638)
(0.846, 0.5654)
(0.847, 0.5670)
(0.848, 0.5685)
(0.849, 0.5700)
(0.850, 0.5715)
(0.851, 0.5728)
(0.852, 0.5742)
(0.853, 0.5754)
(0.854, 0.5766)
(0.855, 0.5778)
(0.856, 0.5789)
(0.857, 0.5799)
(0.858, 0.5808)
(0.859, 0.5817)
(0.860, 0.5825)
(0.861, 0.5833)
(0.862, 0.5839)
(0.863, 0.5845)
(0.864, 0.5850)
(0.865, 0.5854)
(0.866, 0.5858)
(0.867, 0.5861)
(0.868, 0.5862)
(0.869, 0.5863)
(0.870, 0.5863)
(0.871, 0.5862)
(0.872, 0.5860)
(0.873, 0.5858)
(0.874, 0.5854)
(0.875, 0.5849)
(0.876, 0.5779)
(0.877, 0.5816)
(0.878, 0.5854)
(0.879, 0.5892)
(0.880, 0.5930)
(0.881, 0.5968)
(0.882, 0.6006)
(0.883, 0.6044)
(0.884, 0.6082)
(0.885, 0.6121)
(0.886, 0.6159)
(0.887, 0.6198)
(0.888, 0.6236)
(0.889, 0.6275)
(0.890, 0.6313)
(0.891, 0.6352)
(0.892, 0.6391)
(0.893, 0.6430)
(0.894, 0.6469)
(0.895, 0.6508)
(0.896, 0.6547)
(0.897, 0.6586)
(0.898, 0.6625)
(0.899, 0.6664)
(0.900, 0.6703)
(0.901, 0.6742)
(0.902, 0.6781)
(0.903, 0.6820)
(0.904, 0.6859)
(0.905, 0.6899)
(0.906, 0.6938)
(0.907, 0.6977)
(0.908, 0.7016)
(0.909, 0.7055)
(0.910, 0.7094)
(0.911, 0.7133)
(0.912, 0.7172)
(0.913, 0.7212)
(0.914, 0.7250)
(0.915, 0.7289)
(0.916, 0.7328)
(0.917, 0.7367)
(0.918, 0.7406)
(0.919, 0.7445)
(0.920, 0.7483)
(0.921, 0.7522)
(0.922, 0.7560)
(0.923, 0.7599)
(0.924, 0.7637)
(0.925, 0.7675)
(0.926, 0.7713)
(0.927, 0.7751)
(0.928, 0.7789)
(0.929, 0.7827)
(0.930, 0.7864)
(0.931, 0.7901)
(0.932, 0.7939)
(0.933, 0.7976)
(0.934, 0.8013)
(0.935, 0.8049)
(0.936, 0.8086)
(0.937, 0.8122)
(0.938, 0.8158)
(0.939, 0.8194)
(0.940, 0.8230)
(0.941, 0.8265)
(0.942, 0.8301)
(0.943, 0.8336)
(0.944, 0.8370)
(0.945, 0.8405)
(0.946, 0.8439)
(0.947, 0.8473)
(0.948, 0.8506)
(0.949, 0.8540)
(0.950, 0.8573)
(0.951, 0.8605)
(0.952, 0.8638)
(0.953, 0.8670)
(0.954, 0.8701)
(0.955, 0.8732)
(0.956, 0.8763)
(0.957, 0.8794)
(0.958, 0.8824)
(0.959, 0.8853)
(0.960, 0.8883)
(0.961, 0.8911)
(0.962, 0.8940)
(0.963, 0.8967)
(0.964, 0.8995)
(0.965, 0.9022)
(0.966, 0.9048)
(0.967, 0.9074)
(0.968, 0.9099)
(0.969, 0.9124)
(0.970, 0.9148)
(0.971, 0.9171)
(0.972, 0.9194)
(0.973, 0.9217)
(0.974, 0.9238)
(0.975, 0.9260)
(0.976, 0.9280)
(0.977, 0.9300)
(0.978, 0.9319)
(0.979, 0.9337)
(0.980, 0.9355)
(0.981, 0.9372)
(0.982, 0.9388)
(0.983, 0.9403)
(0.984, 0.9418)
(0.985, 0.9431)
(0.986, 0.9444)
(0.987, 0.9456)
(0.988, 0.9468)
(0.989, 0.9478)
(0.990, 0.9487)
(0.991, 0.9496)
(0.992, 0.9503)
(0.993, 0.9510)
(0.994, 0.9515)
(0.995, 0.9520)
(0.996, 0.9523)
(0.997, 0.9525)
(0.998, 0.9527)
(0.999, 0.9527)
(1.000, 0.9526)
(1.000, 0.0000)}

\defdata{img512-ADM-fid2}{23.24}
\defdata{img512-ADM-cfg2}{7.72}
\defdata{img512-ADM-mparams0}{559}
\defdata{img512-ADM-gflops0}{1983}
\defdata{img512-ADM-nfe}{250}
\defdata{img512-ADM-mimg0}{497}
\defdata{img512-ADM-U-fid2}{9.96}
\defdata{img512-ADM-U-cfg2}{3.85}
\defdata{img512-ADM-U-mparams0}{730}
\defdata{img512-ADM-U-gflops0}{2813}
\defdata{img512-ADM-U-nfe}{250}
\defdata{img512-ADM-U-mimg0}{361}
\defdata{img64-ADM-det-fid2}{--}
\defdata{img64-ADM-det-nfe}{--}
\defdata{img64-ADM-stoch-fid2}{2.07}
\defdata{img64-ADM-stoch-nfe}{250}
\defdata{img64-ADM-mparams0}{296}
\defdata{img64-ADM-gflops0}{110}
\defdata{img64-ADM-mimg0}{1106}

\defdata{img512-DiT-XL-fid2}{12.03}
\defdata{img512-DiT-XL-cfg2}{3.04}
\defdata{img512-DiT-XL-mparams0}{675}
\defdata{img512-DiT-XL-gflops0}{525}
\defdata{img512-DiT-XL-nfe}{250}
\defdata{img512-DiT-XL-mimg0}{768}

\defdata{img512-RIN-fid2}{3.95}
\defdata{img512-RIN-cfg2}{--}
\defdata{img512-RIN-mparams0}{320}
\defdata{img512-RIN-gflops0}{415}
\defdata{img512-RIN-nfe}{1000}
\defdata{img512-RIN-mimg0}{1024}
\defdata{img64-RIN-det-fid2}{--}
\defdata{img64-RIN-det-nfe}{--}
\defdata{img64-RIN-stoch-fid2}{1.23}
\defdata{img64-RIN-stoch-nfe}{1000}
\defdata{img64-RIN-mparams0}{281}
\defdata{img64-RIN-gflops0}{106}
\defdata{img64-RIN-mimg0}{307}

\defdata{img512-U-ViT-L-fid2}{3.54}
\defdata{img512-U-ViT-L-cfg2}{3.02}
\defdata{img512-U-ViT-L-mparams0}{2455}
\defdata{img512-U-ViT-L-gflops0}{555}
\defdata{img512-U-ViT-L-nfe}{256}
\defdata{img512-U-ViT-L-mimg0}{1024}

\defdata{img512-VDMpp-fid2}{2.99}
\defdata{img512-VDMpp-cfg2}{2.65}
\defdata{img512-VDMpp-mparams0}{2455}
\defdata{img512-VDMpp-gflops0}{555}
\defdata{img512-VDMpp-nfe}{256}
\defdata{img512-VDMpp-mimg0}{1434}
\defdata{img64-VDMpp-det-fid2}{--}
\defdata{img64-VDMpp-det-nfe}{--}
\defdata{img64-VDMpp-stoch-fid2}{1.43}
\defdata{img64-VDMpp-stoch-nfe}{511}
\defdata{img64-VDMpp-mparams0}{296}
\defdata{img64-VDMpp-gflops0}{110}
\defdata{img64-VDMpp-mimg0}{2867}

\defdata{img512-StyleGAN-XL-fid2}{--}
\defdata{img512-StyleGAN-XL-cfg2}{2.41}
\defdata{img512-StyleGAN-XL-mparams0}{168}
\defdata{img512-StyleGAN-XL-gflops0}{2067}
\defdata{img512-StyleGAN-XL-nfe}{1}
\defdata{img512-StyleGAN-XL-mimg0}{640}
\defdata{img64-StyleGAN-XL-det-fid2}{1.52}
\defdata{img64-StyleGAN-XL-det-nfe}{1}
\defdata{img64-StyleGAN-XL-stoch-fid2}{--}
\defdata{img64-StyleGAN-XL-stoch-nfe}{--}
\defdata{img64-StyleGAN-XL-mparams0}{134}
\defdata{img64-StyleGAN-XL-gflops0}{549}
\defdata{img64-StyleGAN-XL-mimg0}{570}

\defdata{img64-EDM-ADM-det-fid2}{2.66}
\defdata{img64-EDM-ADM-det-nfe}{79}
\defdata{img64-EDM-ADM-stoch-fid2}{1.57}
\defdata{img64-EDM-ADM-stoch-nfe}{511}
\defdata{img64-EDM-ADM-mparams0}{\rawdata{img64-ADM-mparams0}}
\defdata{img64-EDM-ADM-gflops0}{\rawdata{img64-ADM-gflops0}}
\defdata{img64-EDM-ADM-mimg0}{\rawdata{img64-ADM-mimg0}}
\defdata{img64-EDM-EDM-det-fid2}{2.22}
\defdata{img64-EDM-EDM-det-nfe}{79}
\defdata{img64-EDM-EDM-stoch-fid2}{1.36}
\defdata{img64-EDM-EDM-stoch-nfe}{511}
\defdata{img64-EDM-EDM-mparams0}{296}
\defdata{img64-EDM-EDM-gflops0}{110}
\defdata{img64-EDM-EDM-mimg0}{2500}

\defdata{img512-VAE-gflops0}{1261}
\defdata{img512-VAE-gflops1}{1260.9}

\defdata{img512-cfgB-kl-rb-rmb-rgsb-rgm-aa-ff-ca-zi-eq-nls-wn-fwn-lrd-ref70kb-rc-rr-res-pne-nnd-np-rb0.3-lag-log-bc-rd-guid1.0-2147-fid2}{2.56}
\defdata{img512-cfgB-kl-rb-rmb-rgsb-rgm-aa-ff-ca-zi-eq-nls-wn-fwn-lrd-ref70kb-rc-rr-res-pne-nnd-np-rb0.3-lag-log-bc-rd-guid1.0-2147-fid4}{2.5555}
\defdata{img512-cfgB-kl-rb-rmb-rgsb-rgm-aa-ff-ca-zi-eq-nls-wn-fwn-lrd-ref70kb-rc-rr-res-pne-nnd-np-rb0.3-lag-log-bc-rd-guid1.0-2147-phema3}{0.130}
\defdata{img512-cfgB-kl-rb-rmb-rgsb-rgm-aa-ff-ca-zi-eq-nls-wn-fwn-lrd-ref70kb-rc-rr-res-pne-nnd-np-rb0.3-lag-log-bc-rd-guid1.0-2147-lo}{(0.010, 11.8664) (0.015, 8.6684) (0.020, 7.3421) (0.025, 6.7381) (0.030, 6.3240) (0.035, 5.9067) (0.040, 5.6074) (0.045, 5.1538) (0.050, 4.9897) (0.055, 4.6545) (0.060, 4.4475) (0.065, 4.0713) (0.070, 3.8972) (0.075, 3.6710) (0.080, 3.4648) (0.085, 3.2699) (0.090, 3.1578) (0.095, 3.0557) (0.100, 2.9153) (0.105, 2.7542) (0.110, 2.7053) (0.115, 2.6453) (0.120, 2.5668) (0.125, 2.5853) (0.130, 2.5555) (0.135, 2.6162) (0.140, 2.6274) (0.145, 2.7232) (0.150, 2.8026) (0.155, 2.9088) (0.160, 3.0992) (0.165, 3.2793) (0.170, 3.5150) (0.175, 3.7516) (0.180, 3.9318) (0.185, 4.3639) (0.190, 4.6058) (0.195, 4.9291) (0.200, 5.3385) (0.205, 5.6643) (0.210, 6.0781) (0.215, 6.5125) (0.220, 6.8325) (0.225, 7.3471) (0.230, 7.9054) (0.235, 8.2642) (0.240, 8.7922) (0.245, 9.2545) (0.250, 9.7974) (0.255, 10.4093) (0.260, 10.9113) (0.265, 11.5845) (0.270, 12.1902) (0.275, 12.9134) (0.280, 13.8327) (0.285, 14.7976) (0.290, 16.2687) (0.295, 18.3425) (0.300, 23.9241)}
\defdata{img512-cfgB-kl-rb-rmb-rgsb-rgm-aa-ff-ca-zi-eq-nls-wn-fwn-lrd-ref70kb-rc-rr-res-pne-nnd-np-rb0.3-lag-log-bc-rd-guid1.0-2147-hi}{(0.010, 12.0703) (0.015, 8.7676) (0.020, 7.6972) (0.025, 6.9144) (0.030, 6.5107) (0.035, 6.0458) (0.040, 5.6635) (0.045, 5.1886) (0.050, 5.0325) (0.055, 4.8046) (0.060, 4.5005) (0.065, 4.2480) (0.070, 4.0449) (0.075, 3.7944) (0.080, 3.5311) (0.085, 3.3526) (0.090, 3.2134) (0.095, 3.1051) (0.100, 2.9315) (0.105, 2.8247) (0.110, 2.8020) (0.115, 2.6651) (0.120, 2.6193) (0.125, 2.6305) (0.130, 2.6133) (0.135, 2.7012) (0.140, 2.6509) (0.145, 2.7567) (0.150, 2.8820) (0.155, 2.9698) (0.160, 3.1974) (0.165, 3.3372) (0.170, 3.6294) (0.175, 3.8559) (0.180, 4.1098) (0.185, 4.4465) (0.190, 4.6545) (0.195, 5.0302) (0.200, 5.3694) (0.205, 5.9006) (0.210, 6.1870) (0.215, 6.5938) (0.220, 6.9454) (0.225, 7.4353) (0.230, 7.9491) (0.235, 8.4100) (0.240, 8.8655) (0.245, 9.3413) (0.250, 9.8377) (0.255, 10.4202) (0.260, 11.0504) (0.265, 11.6030) (0.270, 12.2804) (0.275, 12.9893) (0.280, 13.9517) (0.285, 15.0209) (0.290, 16.3807) (0.295, 18.5726) (0.300, 24.0702)}
\defdata{img512-cfgB-kl-rb-rmb-rgsb-rgm-aa-ff-ca-zi-eq-nls-wn-fwn-lrd-ref70kb-rc-rr-res-pne-nnd-np-rb0.3-lag-log-bc-rd-guid1.0-2147-pkl}{1533/52/phema/2147483-0.130.pkl}
\defdata{img512-cfgB-kl-rb-rmb-rgsb-rgm-aa-ff-ca-zi-eq-nls-wn-fwn-lrd-ref70kb-rc-rr-res-pne-nnd-np-rb0.3-lag-log-bc-rd-guid1.0-2147-is1}{309.8}
\defdata{img512-cfgB-kl-rb-rmb-rgsb-rgm-aa-ff-ca-zi-eq-nls-wn-fwn-lrd-ref70kb-rc-rr-res-pne-nnd-np-rb0.3-lag-log-bc-rd-guid1.0-2147-is-phema3}{0.230}
\defdata{img512-cfgB-kl-rb-rmb-rgsb-rgm-aa-ff-ca-zi-eq-nls-wn-fwn-lrd-ref70kb-rc-rr-res-pne-nnd-np-rb0.3-lag-log-bc-rd-guid1.0-2147-is-lo}{(0.010, 118.28) (0.015, 136.62) (0.020, 145.86) (0.025, 153.37) (0.030, 158.26) (0.035, 161.86) (0.040, 166.93) (0.045, 173.14) (0.050, 174.94) (0.055, 177.94) (0.060, 182.41) (0.065, 188.09) (0.070, 189.90) (0.075, 194.71) (0.080, 201.24) (0.085, 204.04) (0.090, 209.19) (0.095, 211.56) (0.100, 216.42) (0.105, 220.78) (0.110, 225.99) (0.115, 228.50) (0.120, 234.49) (0.125, 239.24) (0.130, 244.74) (0.135, 249.85) (0.140, 251.50) (0.145, 259.20) (0.150, 261.27) (0.155, 267.02) (0.160, 271.18) (0.165, 275.56) (0.170, 281.05) (0.175, 285.99) (0.180, 284.96) (0.185, 291.66) (0.190, 292.75) (0.195, 295.85) (0.200, 301.21) (0.205, 300.90) (0.210, 303.50) (0.215, 302.15) (0.220, 303.91) (0.225, 303.00) (0.230, 306.97) (0.235, 303.46) (0.240, 301.82) (0.245, 297.48) (0.250, 291.89) (0.255, 288.04) (0.260, 282.35) (0.265, 272.73) (0.270, 263.23) (0.275, 253.84) (0.280, 239.92) (0.285, 222.93) (0.290, 201.40) (0.295, 174.52) (0.300, 126.63)}
\defdata{img512-cfgB-kl-rb-rmb-rgsb-rgm-aa-ff-ca-zi-eq-nls-wn-fwn-lrd-ref70kb-rc-rr-res-pne-nnd-np-rb0.3-lag-log-bc-rd-guid1.0-2147-is-hi}{(0.010, 119.14) (0.015, 138.21) (0.020, 148.22) (0.025, 153.46) (0.030, 158.81) (0.035, 164.99) (0.040, 168.21) (0.045, 175.10) (0.050, 177.55) (0.055, 180.16) (0.060, 183.79) (0.065, 190.25) (0.070, 191.34) (0.075, 199.45) (0.080, 201.71) (0.085, 204.96) (0.090, 210.20) (0.095, 215.02) (0.100, 219.59) (0.105, 223.14) (0.110, 227.05) (0.115, 234.57) (0.120, 237.28) (0.125, 241.01) (0.130, 246.95) (0.135, 253.04) (0.140, 255.54) (0.145, 260.43) (0.150, 263.28) (0.155, 270.26) (0.160, 272.30) (0.165, 277.44) (0.170, 283.61) (0.175, 286.91) (0.180, 290.58) (0.185, 295.21) (0.190, 296.06) (0.195, 297.51) (0.200, 301.88) (0.205, 303.55) (0.210, 305.12) (0.215, 305.76) (0.220, 307.30) (0.225, 307.02) (0.230, 309.76) (0.235, 304.57) (0.240, 302.69) (0.245, 298.92) (0.250, 295.74) (0.255, 290.70) (0.260, 285.09) (0.265, 275.97) (0.270, 266.35) (0.275, 255.22) (0.280, 241.40) (0.285, 226.37) (0.290, 203.88) (0.295, 177.38) (0.300, 127.38)}
\defdata{img512-cfgB-kl-rb-rmb-rgsb-rgm-aa-ff-ca-zi-eq-nls-wn-fwn-lrd-ref70kb-rc-rr-res-pne-nnd-np-rb0.3-lag-log-bc-rd-guid1.0-2147-recall3}{0.622}
\defdata{img512-cfgB-kl-rb-rmb-rgsb-rgm-aa-ff-ca-zi-eq-nls-wn-fwn-lrd-ref70kb-rc-rr-res-pne-nnd-np-rb0.3-lag-log-bc-rd-guid1.0-2147-recall-phema3}{0.180}
\defdata{img512-cfgB-kl-rb-rmb-rgsb-rgm-aa-ff-ca-zi-eq-nls-wn-fwn-lrd-ref70kb-rc-rr-res-pne-nnd-np-rb0.3-lag-log-bc-rd-guid1.0-2147-recall-lo}{(0.010, 0.4942) (0.015, 0.5205) (0.020, 0.5245) (0.025, 0.5329) (0.030, 0.5347) (0.035, 0.5394) (0.040, 0.5408) (0.045, 0.5479) (0.050, 0.5470) (0.055, 0.5529) (0.060, 0.5564) (0.065, 0.5605) (0.070, 0.5594) (0.075, 0.5619) (0.080, 0.5660) (0.085, 0.5712) (0.090, 0.5750) (0.095, 0.5784) (0.100, 0.5782) (0.105, 0.5838) (0.110, 0.5862) (0.115, 0.5914) (0.120, 0.5930) (0.125, 0.5976) (0.130, 0.5993) (0.135, 0.6038) (0.140, 0.6057) (0.145, 0.6090) (0.150, 0.6117) (0.155, 0.6140) (0.160, 0.6139) (0.165, 0.6181) (0.170, 0.6156) (0.175, 0.6157) (0.180, 0.6172) (0.185, 0.6143) (0.190, 0.6173) (0.195, 0.6108) (0.200, 0.6081) (0.205, 0.6022) (0.210, 0.5961) (0.215, 0.5895) (0.220, 0.5797) (0.225, 0.5668) (0.230, 0.5512) (0.235, 0.5340) (0.240, 0.5170) (0.245, 0.4989) (0.250, 0.4697) (0.255, 0.4458) (0.260, 0.4203) (0.265, 0.3854) (0.270, 0.3538) (0.275, 0.3160) (0.280, 0.2771) (0.285, 0.2376) (0.290, 0.2026) (0.295, 0.1585) (0.300, 0.0947)}
\defdata{img512-cfgB-kl-rb-rmb-rgsb-rgm-aa-ff-ca-zi-eq-nls-wn-fwn-lrd-ref70kb-rc-rr-res-pne-nnd-np-rb0.3-lag-log-bc-rd-guid1.0-2147-recall-hi}{(0.010, 0.4972) (0.015, 0.5220) (0.020, 0.5327) (0.025, 0.5366) (0.030, 0.5402) (0.035, 0.5425) (0.040, 0.5423) (0.045, 0.5509) (0.050, 0.5491) (0.055, 0.5570) (0.060, 0.5578) (0.065, 0.5616) (0.070, 0.5632) (0.075, 0.5679) (0.080, 0.5723) (0.085, 0.5719) (0.090, 0.5773) (0.095, 0.5801) (0.100, 0.5829) (0.105, 0.5855) (0.110, 0.5887) (0.115, 0.5936) (0.120, 0.5988) (0.125, 0.6011) (0.130, 0.6028) (0.135, 0.6055) (0.140, 0.6113) (0.145, 0.6126) (0.150, 0.6155) (0.155, 0.6152) (0.160, 0.6186) (0.165, 0.6199) (0.170, 0.6209) (0.175, 0.6201) (0.180, 0.6220) (0.185, 0.6180) (0.190, 0.6197) (0.195, 0.6160) (0.200, 0.6152) (0.205, 0.6060) (0.210, 0.6004) (0.215, 0.5921) (0.220, 0.5827) (0.225, 0.5758) (0.230, 0.5561) (0.235, 0.5365) (0.240, 0.5208) (0.245, 0.5021) (0.250, 0.4743) (0.255, 0.4573) (0.260, 0.4234) (0.265, 0.3920) (0.270, 0.3568) (0.275, 0.3192) (0.280, 0.2855) (0.285, 0.2444) (0.290, 0.2047) (0.295, 0.1617) (0.300, 0.0953)}

\defdata{img512-cfgB-kl-rb-rmb-rgsb-rgm-aa-ff-ca-zi-eq-nls-wn-fwn-lrd-ref70kb-rc-rr-res-pne-nnd-np-rb0.3-lag-log-bc-rd-guid1.1-2147-fid2}{2.39}
\defdata{img512-cfgB-kl-rb-rmb-rgsb-rgm-aa-ff-ca-zi-eq-nls-wn-fwn-lrd-ref70kb-rc-rr-res-pne-nnd-np-rb0.3-lag-log-bc-rd-guid1.1-2147-fid4}{2.3925}
\defdata{img512-cfgB-kl-rb-rmb-rgsb-rgm-aa-ff-ca-zi-eq-nls-wn-fwn-lrd-ref70kb-rc-rr-res-pne-nnd-np-rb0.3-lag-log-bc-rd-guid1.1-2147-phema3}{0.095}
\defdata{img512-cfgB-kl-rb-rmb-rgsb-rgm-aa-ff-ca-zi-eq-nls-wn-fwn-lrd-ref70kb-rc-rr-res-pne-nnd-np-rb0.3-lag-log-bc-rd-guid1.1-2147-lo}{(0.010, 8.4091) (0.015, 5.5317) (0.020, 4.8130) (0.025, 4.2613) (0.030, 4.0164) (0.035, 3.7866) (0.040, 3.5009) (0.045, 3.4433) (0.050, 3.1747) (0.055, 3.0713) (0.060, 2.8807) (0.065, 2.8051) (0.070, 2.7361) (0.075, 2.6141) (0.080, 2.5589) (0.085, 2.4174) (0.090, 2.4824) (0.095, 2.3925) (0.100, 2.4292) (0.105, 2.4173) (0.110, 2.4582) (0.115, 2.5054) (0.120, 2.5783) (0.125, 2.6034) (0.130, 2.7448) (0.135, 2.8641) (0.140, 2.9853) (0.145, 3.2009) (0.150, 3.3474) (0.155, 3.5953) (0.160, 3.8455) (0.165, 4.0435) (0.170, 4.3560) (0.175, 4.6891) (0.180, 4.9057) (0.185, 5.3008) (0.190, 5.5738) (0.195, 5.8713) (0.200, 6.3654)}
\defdata{img512-cfgB-kl-rb-rmb-rgsb-rgm-aa-ff-ca-zi-eq-nls-wn-fwn-lrd-ref70kb-rc-rr-res-pne-nnd-np-rb0.3-lag-log-bc-rd-guid1.1-2147-hi}{(0.010, 8.5892) (0.015, 5.6399) (0.020, 4.8285) (0.025, 4.3974) (0.030, 4.1290) (0.035, 3.8913) (0.040, 3.6520) (0.045, 3.4629) (0.050, 3.2491) (0.055, 3.1396) (0.060, 2.9413) (0.065, 2.9022) (0.070, 2.7958) (0.075, 2.6453) (0.080, 2.6133) (0.085, 2.5628) (0.090, 2.4848) (0.095, 2.4396) (0.100, 2.5045) (0.105, 2.4584) (0.110, 2.5326) (0.115, 2.5230) (0.120, 2.6635) (0.125, 2.6960) (0.130, 2.8475) (0.135, 2.9133) (0.140, 3.0645) (0.145, 3.2391) (0.150, 3.4152) (0.155, 3.7637) (0.160, 3.8998) (0.165, 4.1612) (0.170, 4.4504) (0.175, 4.8144) (0.180, 4.9889) (0.185, 5.3523) (0.190, 5.6797) (0.195, 6.0768) (0.200, 6.5426)}
\defdata{img512-cfgB-kl-rb-rmb-rgsb-rgm-aa-ff-ca-zi-eq-nls-wn-fwn-lrd-ref70kb-rc-rr-res-pne-nnd-np-rb0.3-lag-log-bc-rd-guid1.1-2147-pkl}{1533/52/phema/2147483-0.095.pkl}
\defdata{img512-cfgB-kl-rb-rmb-rgsb-rgm-aa-ff-ca-zi-eq-nls-wn-fwn-lrd-ref70kb-rc-rr-res-pne-nnd-np-rb0.3-lag-log-bc-rd-guid1.1-2147-is1}{328.4}
\defdata{img512-cfgB-kl-rb-rmb-rgsb-rgm-aa-ff-ca-zi-eq-nls-wn-fwn-lrd-ref70kb-rc-rr-res-pne-nnd-np-rb0.3-lag-log-bc-rd-guid1.1-2147-is-phema3}{0.200}
\defdata{img512-cfgB-kl-rb-rmb-rgsb-rgm-aa-ff-ca-zi-eq-nls-wn-fwn-lrd-ref70kb-rc-rr-res-pne-nnd-np-rb0.3-lag-log-bc-rd-guid1.1-2147-is-lo}{(0.010, 146.18) (0.015, 170.60) (0.020, 179.28) (0.025, 186.65) (0.030, 190.88) (0.035, 195.90) (0.040, 201.73) (0.045, 206.12) (0.050, 211.53) (0.055, 214.28) (0.060, 219.14) (0.065, 222.44) (0.070, 225.41) (0.075, 231.64) (0.080, 236.10) (0.085, 239.81) (0.090, 243.96) (0.095, 246.90) (0.100, 252.10) (0.105, 256.35) (0.110, 260.19) (0.115, 267.05) (0.120, 269.70) (0.125, 274.65) (0.130, 279.11) (0.135, 283.33) (0.140, 287.80) (0.145, 292.01) (0.150, 296.67) (0.155, 301.40) (0.160, 303.97) (0.165, 305.84) (0.170, 312.19) (0.175, 314.67) (0.180, 317.13) (0.185, 321.72) (0.190, 322.90) (0.195, 325.27) (0.200, 328.27)}
\defdata{img512-cfgB-kl-rb-rmb-rgsb-rgm-aa-ff-ca-zi-eq-nls-wn-fwn-lrd-ref70kb-rc-rr-res-pne-nnd-np-rb0.3-lag-log-bc-rd-guid1.1-2147-is-hi}{(0.010, 148.60) (0.015, 172.23) (0.020, 181.96) (0.025, 189.19) (0.030, 194.26) (0.035, 197.41) (0.040, 203.16) (0.045, 207.60) (0.050, 212.09) (0.055, 216.13) (0.060, 219.99) (0.065, 226.48) (0.070, 229.46) (0.075, 233.24) (0.080, 236.45) (0.085, 242.30) (0.090, 246.24) (0.095, 251.85) (0.100, 255.10) (0.105, 259.82) (0.110, 264.17) (0.115, 268.08) (0.120, 271.57) (0.125, 278.30) (0.130, 282.68) (0.135, 287.59) (0.140, 289.41) (0.145, 294.08) (0.150, 299.84) (0.155, 302.65) (0.160, 305.74) (0.165, 310.92) (0.170, 313.90) (0.175, 317.29) (0.180, 320.15) (0.185, 322.37) (0.190, 325.64) (0.195, 327.65) (0.200, 328.37)}
\defdata{img512-cfgB-kl-rb-rmb-rgsb-rgm-aa-ff-ca-zi-eq-nls-wn-fwn-lrd-ref70kb-rc-rr-res-pne-nnd-np-rb0.3-lag-log-bc-rd-guid1.1-2147-recall3}{0.622}
\defdata{img512-cfgB-kl-rb-rmb-rgsb-rgm-aa-ff-ca-zi-eq-nls-wn-fwn-lrd-ref70kb-rc-rr-res-pne-nnd-np-rb0.3-lag-log-bc-rd-guid1.1-2147-recall-phema3}{0.175}
\defdata{img512-cfgB-kl-rb-rmb-rgsb-rgm-aa-ff-ca-zi-eq-nls-wn-fwn-lrd-ref70kb-rc-rr-res-pne-nnd-np-rb0.3-lag-log-bc-rd-guid1.1-2147-recall-lo}{(0.010, 0.5099) (0.015, 0.5357) (0.020, 0.5433) (0.025, 0.5504) (0.030, 0.5508) (0.035, 0.5545) (0.040, 0.5580) (0.045, 0.5618) (0.050, 0.5629) (0.055, 0.5670) (0.060, 0.5667) (0.065, 0.5726) (0.070, 0.5754) (0.075, 0.5740) (0.080, 0.5800) (0.085, 0.5823) (0.090, 0.5825) (0.095, 0.5879) (0.100, 0.5860) (0.105, 0.5912) (0.110, 0.5943) (0.115, 0.5977) (0.120, 0.6008) (0.125, 0.6027) (0.130, 0.6065) (0.135, 0.6056) (0.140, 0.6107) (0.145, 0.6125) (0.150, 0.6152) (0.155, 0.6166) (0.160, 0.6185) (0.165, 0.6170) (0.170, 0.6152) (0.175, 0.6154) (0.180, 0.6114) (0.185, 0.6129) (0.190, 0.6104) (0.195, 0.6070) (0.200, 0.6035)}
\defdata{img512-cfgB-kl-rb-rmb-rgsb-rgm-aa-ff-ca-zi-eq-nls-wn-fwn-lrd-ref70kb-rc-rr-res-pne-nnd-np-rb0.3-lag-log-bc-rd-guid1.1-2147-recall-hi}{(0.010, 0.5126) (0.015, 0.5410) (0.020, 0.5446) (0.025, 0.5519) (0.030, 0.5544) (0.035, 0.5557) (0.040, 0.5608) (0.045, 0.5667) (0.050, 0.5648) (0.055, 0.5705) (0.060, 0.5732) (0.065, 0.5738) (0.070, 0.5773) (0.075, 0.5789) (0.080, 0.5826) (0.085, 0.5873) (0.090, 0.5860) (0.095, 0.5915) (0.100, 0.5898) (0.105, 0.5957) (0.110, 0.5973) (0.115, 0.6056) (0.120, 0.6023) (0.125, 0.6042) (0.130, 0.6089) (0.135, 0.6106) (0.140, 0.6126) (0.145, 0.6129) (0.150, 0.6190) (0.155, 0.6189) (0.160, 0.6188) (0.165, 0.6191) (0.170, 0.6180) (0.175, 0.6218) (0.180, 0.6178) (0.185, 0.6142) (0.190, 0.6121) (0.195, 0.6122) (0.200, 0.6047)}

\defdata{img512-cfgB-kl-rb-rmb-rgsb-rgm-aa-ff-ca-zi-eq-nls-wn-fwn-lrd-ref70kb-rc-rr-res-pne-nnd-np-rb0.3-lag-log-bc-rd-guid1.2-2147-fid2}{2.32}
\defdata{img512-cfgB-kl-rb-rmb-rgsb-rgm-aa-ff-ca-zi-eq-nls-wn-fwn-lrd-ref70kb-rc-rr-res-pne-nnd-np-rb0.3-lag-log-bc-rd-guid1.2-2147-fid4}{2.3199}
\defdata{img512-cfgB-kl-rb-rmb-rgsb-rgm-aa-ff-ca-zi-eq-nls-wn-fwn-lrd-ref70kb-rc-rr-res-pne-nnd-np-rb0.3-lag-log-bc-rd-guid1.2-2147-phema3}{0.080}
\defdata{img512-cfgB-kl-rb-rmb-rgsb-rgm-aa-ff-ca-zi-eq-nls-wn-fwn-lrd-ref70kb-rc-rr-res-pne-nnd-np-rb0.3-lag-log-bc-rd-guid1.2-2147-lo}{(0.010, 6.2335) (0.015, 3.7517) (0.020, 3.3108) (0.025, 3.0152) (0.030, 2.8832) (0.035, 2.7241) (0.040, 2.6753) (0.045, 2.5332) (0.050, 2.4537) (0.055, 2.4552) (0.060, 2.3443) (0.065, 2.3854) (0.070, 2.3324) (0.075, 2.3294) (0.080, 2.3199) (0.085, 2.4080) (0.090, 2.4381) (0.095, 2.5124) (0.100, 2.5141) (0.105, 2.6755) (0.110, 2.7755) (0.115, 2.9147) (0.120, 3.0136) (0.125, 3.1480) (0.130, 3.3155) (0.135, 3.6068) (0.140, 3.7415) (0.145, 3.8914) (0.150, 4.1684) (0.155, 4.4541) (0.160, 4.6982) (0.165, 4.9521) (0.170, 5.2382) (0.175, 5.5649) (0.180, 5.9283) (0.185, 6.2359) (0.190, 6.5966) (0.195, 6.8433) (0.200, 7.2379)}
\defdata{img512-cfgB-kl-rb-rmb-rgsb-rgm-aa-ff-ca-zi-eq-nls-wn-fwn-lrd-ref70kb-rc-rr-res-pne-nnd-np-rb0.3-lag-log-bc-rd-guid1.2-2147-hi}{(0.010, 6.3380) (0.015, 3.9434) (0.020, 3.3591) (0.025, 3.0562) (0.030, 2.9069) (0.035, 2.7447) (0.040, 2.6869) (0.045, 2.6049) (0.050, 2.4827) (0.055, 2.4899) (0.060, 2.4059) (0.065, 2.4703) (0.070, 2.4049) (0.075, 2.3816) (0.080, 2.4073) (0.085, 2.4502) (0.090, 2.4879) (0.095, 2.5214) (0.100, 2.6143) (0.105, 2.7239) (0.110, 2.8144) (0.115, 2.9988) (0.120, 3.0434) (0.125, 3.2213) (0.130, 3.4110) (0.135, 3.6323) (0.140, 3.8441) (0.145, 3.9969) (0.150, 4.3119) (0.155, 4.5222) (0.160, 4.8005) (0.165, 5.1775) (0.170, 5.3685) (0.175, 5.7125) (0.180, 5.9487) (0.185, 6.3022) (0.190, 6.6790) (0.195, 7.0427) (0.200, 7.4049)}
\defdata{img512-cfgB-kl-rb-rmb-rgsb-rgm-aa-ff-ca-zi-eq-nls-wn-fwn-lrd-ref70kb-rc-rr-res-pne-nnd-np-rb0.3-lag-log-bc-rd-guid1.2-2147-pkl}{1533/52/phema/2147483-0.080.pkl}
\defdata{img512-cfgB-kl-rb-rmb-rgsb-rgm-aa-ff-ca-zi-eq-nls-wn-fwn-lrd-ref70kb-rc-rr-res-pne-nnd-np-rb0.3-lag-log-bc-rd-guid1.2-2147-is1}{350.0}
\defdata{img512-cfgB-kl-rb-rmb-rgsb-rgm-aa-ff-ca-zi-eq-nls-wn-fwn-lrd-ref70kb-rc-rr-res-pne-nnd-np-rb0.3-lag-log-bc-rd-guid1.2-2147-is-phema3}{0.195}
\defdata{img512-cfgB-kl-rb-rmb-rgsb-rgm-aa-ff-ca-zi-eq-nls-wn-fwn-lrd-ref70kb-rc-rr-res-pne-nnd-np-rb0.3-lag-log-bc-rd-guid1.2-2147-is-lo}{(0.010, 174.20) (0.015, 200.58) (0.020, 209.93) (0.025, 216.69) (0.030, 222.84) (0.035, 228.11) (0.040, 232.93) (0.045, 236.82) (0.050, 240.02) (0.055, 245.81) (0.060, 249.64) (0.065, 254.78) (0.070, 258.31) (0.075, 261.30) (0.080, 264.73) (0.085, 270.79) (0.090, 274.58) (0.095, 278.97) (0.100, 284.57) (0.105, 287.81) (0.110, 293.17) (0.115, 296.44) (0.120, 298.32) (0.125, 302.78) (0.130, 305.24) (0.135, 312.07) (0.140, 313.37) (0.145, 316.38) (0.150, 323.55) (0.155, 325.94) (0.160, 331.50) (0.165, 329.85) (0.170, 336.63) (0.175, 339.27) (0.180, 339.93) (0.185, 342.81) (0.190, 344.43) (0.195, 344.43) (0.200, 348.02)}
\defdata{img512-cfgB-kl-rb-rmb-rgsb-rgm-aa-ff-ca-zi-eq-nls-wn-fwn-lrd-ref70kb-rc-rr-res-pne-nnd-np-rb0.3-lag-log-bc-rd-guid1.2-2147-is-hi}{(0.010, 175.46) (0.015, 202.72) (0.020, 211.94) (0.025, 219.76) (0.030, 223.18) (0.035, 229.04) (0.040, 234.56) (0.045, 239.63) (0.050, 243.62) (0.055, 246.64) (0.060, 250.29) (0.065, 257.89) (0.070, 259.41) (0.075, 263.97) (0.080, 269.81) (0.085, 273.88) (0.090, 275.92) (0.095, 280.74) (0.100, 286.69) (0.105, 290.61) (0.110, 294.15) (0.115, 297.70) (0.120, 301.98) (0.125, 305.12) (0.130, 310.31) (0.135, 312.30) (0.140, 316.80) (0.145, 318.04) (0.150, 326.28) (0.155, 327.92) (0.160, 331.92) (0.165, 333.96) (0.170, 337.33) (0.175, 340.80) (0.180, 341.87) (0.185, 345.68) (0.190, 346.07) (0.195, 350.02) (0.200, 349.95)}
\defdata{img512-cfgB-kl-rb-rmb-rgsb-rgm-aa-ff-ca-zi-eq-nls-wn-fwn-lrd-ref70kb-rc-rr-res-pne-nnd-np-rb0.3-lag-log-bc-rd-guid1.2-2147-recall3}{0.620}
\defdata{img512-cfgB-kl-rb-rmb-rgsb-rgm-aa-ff-ca-zi-eq-nls-wn-fwn-lrd-ref70kb-rc-rr-res-pne-nnd-np-rb0.3-lag-log-bc-rd-guid1.2-2147-recall-phema3}{0.150}
\defdata{img512-cfgB-kl-rb-rmb-rgsb-rgm-aa-ff-ca-zi-eq-nls-wn-fwn-lrd-ref70kb-rc-rr-res-pne-nnd-np-rb0.3-lag-log-bc-rd-guid1.2-2147-recall-lo}{(0.010, 0.5195) (0.015, 0.5495) (0.020, 0.5555) (0.025, 0.5604) (0.030, 0.5618) (0.035, 0.5648) (0.040, 0.5683) (0.045, 0.5716) (0.050, 0.5732) (0.055, 0.5748) (0.060, 0.5754) (0.065, 0.5791) (0.070, 0.5809) (0.075, 0.5808) (0.080, 0.5888) (0.085, 0.5861) (0.090, 0.5909) (0.095, 0.5942) (0.100, 0.5971) (0.105, 0.5993) (0.110, 0.5989) (0.115, 0.6020) (0.120, 0.6037) (0.125, 0.6044) (0.130, 0.6053) (0.135, 0.6101) (0.140, 0.6106) (0.145, 0.6135) (0.150, 0.6141) (0.155, 0.6122) (0.160, 0.6137) (0.165, 0.6129) (0.170, 0.6116) (0.175, 0.6136) (0.180, 0.6081) (0.185, 0.6067) (0.190, 0.6016) (0.195, 0.6018) (0.200, 0.5935)}
\defdata{img512-cfgB-kl-rb-rmb-rgsb-rgm-aa-ff-ca-zi-eq-nls-wn-fwn-lrd-ref70kb-rc-rr-res-pne-nnd-np-rb0.3-lag-log-bc-rd-guid1.2-2147-recall-hi}{(0.010, 0.5233) (0.015, 0.5512) (0.020, 0.5572) (0.025, 0.5638) (0.030, 0.5630) (0.035, 0.5700) (0.040, 0.5716) (0.045, 0.5736) (0.050, 0.5746) (0.055, 0.5774) (0.060, 0.5820) (0.065, 0.5807) (0.070, 0.5842) (0.075, 0.5867) (0.080, 0.5911) (0.085, 0.5912) (0.090, 0.5963) (0.095, 0.5962) (0.100, 0.5986) (0.105, 0.6015) (0.110, 0.6022) (0.115, 0.6051) (0.120, 0.6058) (0.125, 0.6099) (0.130, 0.6084) (0.135, 0.6106) (0.140, 0.6152) (0.145, 0.6158) (0.150, 0.6197) (0.155, 0.6151) (0.160, 0.6163) (0.165, 0.6161) (0.170, 0.6158) (0.175, 0.6158) (0.180, 0.6113) (0.185, 0.6111) (0.190, 0.6108) (0.195, 0.6041) (0.200, 0.6007)}

\defdata{img512-cfgB-kl-rb-rmb-rgsb-rgm-aa-ff-ca-zi-eq-nls-wn-fwn-lrd-ref70kb-rc-rr-res-pne-nnd-np-rb0.3-lag-log-bc-rd-guid1.3-2147-fid2}{2.25}
\defdata{img512-cfgB-kl-rb-rmb-rgsb-rgm-aa-ff-ca-zi-eq-nls-wn-fwn-lrd-ref70kb-rc-rr-res-pne-nnd-np-rb0.3-lag-log-bc-rd-guid1.3-2147-fid4}{2.2456}
\defdata{img512-cfgB-kl-rb-rmb-rgsb-rgm-aa-ff-ca-zi-eq-nls-wn-fwn-lrd-ref70kb-rc-rr-res-pne-nnd-np-rb0.3-lag-log-bc-rd-guid1.3-2147-phema3}{0.035}
\defdata{img512-cfgB-kl-rb-rmb-rgsb-rgm-aa-ff-ca-zi-eq-nls-wn-fwn-lrd-ref70kb-rc-rr-res-pne-nnd-np-rb0.3-lag-log-bc-rd-guid1.3-2147-lo}{(0.010, 4.7509) (0.015, 2.9029) (0.020, 2.5414) (0.025, 2.4521) (0.030, 2.3278) (0.035, 2.2456) (0.040, 2.2679) (0.045, 2.2683) (0.050, 2.2666) (0.055, 2.3120) (0.060, 2.3946) (0.065, 2.3516) (0.070, 2.4147) (0.075, 2.4777) (0.080, 2.6159) (0.085, 2.6742) (0.090, 2.7637) (0.095, 2.8325) (0.100, 2.9278) (0.105, 3.2085) (0.110, 3.3541) (0.115, 3.4483) (0.120, 3.7058) (0.125, 3.8441) (0.130, 4.0666) (0.135, 4.2728) (0.140, 4.5139) (0.145, 4.7961) (0.150, 4.9562) (0.155, 5.2207) (0.160, 5.5781) (0.165, 5.8809) (0.170, 6.2034) (0.175, 6.5855) (0.180, 6.8941) (0.185, 7.1611) (0.190, 7.5031) (0.195, 7.8717) (0.200, 8.0640)}
\defdata{img512-cfgB-kl-rb-rmb-rgsb-rgm-aa-ff-ca-zi-eq-nls-wn-fwn-lrd-ref70kb-rc-rr-res-pne-nnd-np-rb0.3-lag-log-bc-rd-guid1.3-2147-hi}{(0.010, 4.8466) (0.015, 3.0013) (0.020, 2.5898) (0.025, 2.4750) (0.030, 2.3678) (0.035, 2.3197) (0.040, 2.2983) (0.045, 2.3378) (0.050, 2.3092) (0.055, 2.3849) (0.060, 2.4065) (0.065, 2.4331) (0.070, 2.4399) (0.075, 2.5439) (0.080, 2.6392) (0.085, 2.7404) (0.090, 2.8375) (0.095, 2.9395) (0.100, 3.1465) (0.105, 3.3304) (0.110, 3.4344) (0.115, 3.4889) (0.120, 3.8189) (0.125, 3.9393) (0.130, 4.2240) (0.135, 4.3886) (0.140, 4.6513) (0.145, 4.9113) (0.150, 5.0237) (0.155, 5.3312) (0.160, 5.7685) (0.165, 6.1125) (0.170, 6.3172) (0.175, 6.7259) (0.180, 7.0214) (0.185, 7.3483) (0.190, 7.5616) (0.195, 7.9252) (0.200, 8.2515)}
\defdata{img512-cfgB-kl-rb-rmb-rgsb-rgm-aa-ff-ca-zi-eq-nls-wn-fwn-lrd-ref70kb-rc-rr-res-pne-nnd-np-rb0.3-lag-log-bc-rd-guid1.3-2147-pkl}{1533/52/phema/2147483-0.035.pkl}
\defdata{img512-cfgB-kl-rb-rmb-rgsb-rgm-aa-ff-ca-zi-eq-nls-wn-fwn-lrd-ref70kb-rc-rr-res-pne-nnd-np-rb0.3-lag-log-bc-rd-guid1.3-2147-is1}{366.8}
\defdata{img512-cfgB-kl-rb-rmb-rgsb-rgm-aa-ff-ca-zi-eq-nls-wn-fwn-lrd-ref70kb-rc-rr-res-pne-nnd-np-rb0.3-lag-log-bc-rd-guid1.3-2147-is-phema3}{0.200}
\defdata{img512-cfgB-kl-rb-rmb-rgsb-rgm-aa-ff-ca-zi-eq-nls-wn-fwn-lrd-ref70kb-rc-rr-res-pne-nnd-np-rb0.3-lag-log-bc-rd-guid1.3-2147-is-lo}{(0.010, 200.64) (0.015, 228.77) (0.020, 236.53) (0.025, 243.94) (0.030, 249.95) (0.035, 256.09) (0.040, 259.72) (0.045, 265.00) (0.050, 269.20) (0.055, 273.04) (0.060, 275.69) (0.065, 280.92) (0.070, 282.97) (0.075, 287.79) (0.080, 293.35) (0.085, 296.56) (0.090, 300.72) (0.095, 303.89) (0.100, 308.23) (0.105, 311.48) (0.110, 314.27) (0.115, 318.09) (0.120, 324.34) (0.125, 325.32) (0.130, 330.04) (0.135, 331.77) (0.140, 336.84) (0.145, 338.61) (0.150, 343.44) (0.155, 346.20) (0.160, 350.62) (0.165, 352.42) (0.170, 355.51) (0.175, 355.86) (0.180, 358.37) (0.185, 360.42) (0.190, 363.18) (0.195, 363.17) (0.200, 362.36)}
\defdata{img512-cfgB-kl-rb-rmb-rgsb-rgm-aa-ff-ca-zi-eq-nls-wn-fwn-lrd-ref70kb-rc-rr-res-pne-nnd-np-rb0.3-lag-log-bc-rd-guid1.3-2147-is-hi}{(0.010, 202.90) (0.015, 230.05) (0.020, 239.73) (0.025, 245.27) (0.030, 252.38) (0.035, 257.70) (0.040, 260.78) (0.045, 267.08) (0.050, 270.78) (0.055, 274.91) (0.060, 277.55) (0.065, 282.47) (0.070, 285.75) (0.075, 289.82) (0.080, 295.50) (0.085, 300.53) (0.090, 302.88) (0.095, 306.03) (0.100, 310.77) (0.105, 314.34) (0.110, 318.15) (0.115, 323.07) (0.120, 326.79) (0.125, 326.36) (0.130, 331.70) (0.135, 336.46) (0.140, 337.87) (0.145, 341.85) (0.150, 343.98) (0.155, 348.88) (0.160, 352.93) (0.165, 355.29) (0.170, 358.93) (0.175, 358.62) (0.180, 361.34) (0.185, 361.01) (0.190, 363.70) (0.195, 366.33) (0.200, 366.81)}
\defdata{img512-cfgB-kl-rb-rmb-rgsb-rgm-aa-ff-ca-zi-eq-nls-wn-fwn-lrd-ref70kb-rc-rr-res-pne-nnd-np-rb0.3-lag-log-bc-rd-guid1.3-2147-recall3}{0.615}
\defdata{img512-cfgB-kl-rb-rmb-rgsb-rgm-aa-ff-ca-zi-eq-nls-wn-fwn-lrd-ref70kb-rc-rr-res-pne-nnd-np-rb0.3-lag-log-bc-rd-guid1.3-2147-recall-phema3}{0.170}
\defdata{img512-cfgB-kl-rb-rmb-rgsb-rgm-aa-ff-ca-zi-eq-nls-wn-fwn-lrd-ref70kb-rc-rr-res-pne-nnd-np-rb0.3-lag-log-bc-rd-guid1.3-2147-recall-lo}{(0.010, 0.5275) (0.015, 0.5572) (0.020, 0.5637) (0.025, 0.5682) (0.030, 0.5702) (0.035, 0.5730) (0.040, 0.5759) (0.045, 0.5799) (0.050, 0.5821) (0.055, 0.5804) (0.060, 0.5866) (0.065, 0.5898) (0.070, 0.5863) (0.075, 0.5896) (0.080, 0.5920) (0.085, 0.5964) (0.090, 0.5943) (0.095, 0.5963) (0.100, 0.6003) (0.105, 0.6000) (0.110, 0.6036) (0.115, 0.6036) (0.120, 0.6029) (0.125, 0.6058) (0.130, 0.6068) (0.135, 0.6077) (0.140, 0.6078) (0.145, 0.6097) (0.150, 0.6131) (0.155, 0.6091) (0.160, 0.6097) (0.165, 0.6093) (0.170, 0.6064) (0.175, 0.6065) (0.180, 0.6014) (0.185, 0.5970) (0.190, 0.5950) (0.195, 0.5912) (0.200, 0.5831)}
\defdata{img512-cfgB-kl-rb-rmb-rgsb-rgm-aa-ff-ca-zi-eq-nls-wn-fwn-lrd-ref70kb-rc-rr-res-pne-nnd-np-rb0.3-lag-log-bc-rd-guid1.3-2147-recall-hi}{(0.010, 0.5316) (0.015, 0.5594) (0.020, 0.5689) (0.025, 0.5717) (0.030, 0.5748) (0.035, 0.5761) (0.040, 0.5792) (0.045, 0.5829) (0.050, 0.5850) (0.055, 0.5862) (0.060, 0.5877) (0.065, 0.5900) (0.070, 0.5893) (0.075, 0.5934) (0.080, 0.5969) (0.085, 0.5971) (0.090, 0.5956) (0.095, 0.6008) (0.100, 0.6037) (0.105, 0.6041) (0.110, 0.6065) (0.115, 0.6077) (0.120, 0.6078) (0.125, 0.6081) (0.130, 0.6102) (0.135, 0.6132) (0.140, 0.6148) (0.145, 0.6128) (0.150, 0.6138) (0.155, 0.6139) (0.160, 0.6132) (0.165, 0.6139) (0.170, 0.6151) (0.175, 0.6118) (0.180, 0.6052) (0.185, 0.6021) (0.190, 0.5986) (0.195, 0.5980) (0.200, 0.5914)}

\defdata{img512-cfgB-kl-rb-rmb-rgsb-rgm-aa-ff-ca-zi-eq-nls-wn-fwn-lrd-ref70kb-rc-rr-res-pne-nnd-np-rb0.3-lag-log-bc-rd-guid1.4-2147-fid2}{2.23}
\defdata{img512-cfgB-kl-rb-rmb-rgsb-rgm-aa-ff-ca-zi-eq-nls-wn-fwn-lrd-ref70kb-rc-rr-res-pne-nnd-np-rb0.3-lag-log-bc-rd-guid1.4-2147-fid4}{2.2268}
\defdata{img512-cfgB-kl-rb-rmb-rgsb-rgm-aa-ff-ca-zi-eq-nls-wn-fwn-lrd-ref70kb-rc-rr-res-pne-nnd-np-rb0.3-lag-log-bc-rd-guid1.4-2147-phema3}{0.025}
\defdata{img512-cfgB-kl-rb-rmb-rgsb-rgm-aa-ff-ca-zi-eq-nls-wn-fwn-lrd-ref70kb-rc-rr-res-pne-nnd-np-rb0.3-lag-log-bc-rd-guid1.4-2147-lo}{(0.010, 4.0678) (0.015, 2.5194) (0.020, 2.3238) (0.025, 2.2268) (0.030, 2.2277) (0.035, 2.2686) (0.040, 2.2831) (0.045, 2.3769) (0.050, 2.4464) (0.055, 2.4894) (0.060, 2.5890) (0.065, 2.6553) (0.070, 2.7773) (0.075, 2.8589) (0.080, 2.9804) (0.085, 3.1722) (0.090, 3.3119) (0.095, 3.4560) (0.100, 3.5585) (0.105, 3.8369) (0.110, 3.9309) (0.115, 4.2251) (0.120, 4.3953) (0.125, 4.6629) (0.130, 4.8508) (0.135, 5.0662) (0.140, 5.3164) (0.145, 5.4616) (0.150, 5.9126) (0.155, 6.1844) (0.160, 6.4738) (0.165, 6.7967) (0.170, 7.0697) (0.175, 7.3234) (0.180, 7.5766) (0.185, 7.9786) (0.190, 8.2449) (0.195, 8.5869) (0.200, 8.9659)}
\defdata{img512-cfgB-kl-rb-rmb-rgsb-rgm-aa-ff-ca-zi-eq-nls-wn-fwn-lrd-ref70kb-rc-rr-res-pne-nnd-np-rb0.3-lag-log-bc-rd-guid1.4-2147-hi}{(0.010, 4.1573) (0.015, 2.6101) (0.020, 2.3496) (0.025, 2.2716) (0.030, 2.2576) (0.035, 2.2986) (0.040, 2.3217) (0.045, 2.3881) (0.050, 2.4602) (0.055, 2.5630) (0.060, 2.6775) (0.065, 2.7742) (0.070, 2.8715) (0.075, 2.9360) (0.080, 3.1288) (0.085, 3.2356) (0.090, 3.3404) (0.095, 3.5483) (0.100, 3.6485) (0.105, 3.8932) (0.110, 4.0191) (0.115, 4.2620) (0.120, 4.4636) (0.125, 4.7436) (0.130, 4.8676) (0.135, 5.1591) (0.140, 5.3468) (0.145, 5.6437) (0.150, 6.0267) (0.155, 6.2451) (0.160, 6.5314) (0.165, 6.8876) (0.170, 7.1872) (0.175, 7.5905) (0.180, 7.6711) (0.185, 8.0823) (0.190, 8.3687) (0.195, 8.6954) (0.200, 9.0270)}
\defdata{img512-cfgB-kl-rb-rmb-rgsb-rgm-aa-ff-ca-zi-eq-nls-wn-fwn-lrd-ref70kb-rc-rr-res-pne-nnd-np-rb0.3-lag-log-bc-rd-guid1.4-2147-pkl}{1533/52/phema/2147483-0.025.pkl}
\defdata{img512-cfgB-kl-rb-rmb-rgsb-rgm-aa-ff-ca-zi-eq-nls-wn-fwn-lrd-ref70kb-rc-rr-res-pne-nnd-np-rb0.3-lag-log-bc-rd-guid1.4-2147-is1}{378.9}
\defdata{img512-cfgB-kl-rb-rmb-rgsb-rgm-aa-ff-ca-zi-eq-nls-wn-fwn-lrd-ref70kb-rc-rr-res-pne-nnd-np-rb0.3-lag-log-bc-rd-guid1.4-2147-is-phema3}{0.200}
\defdata{img512-cfgB-kl-rb-rmb-rgsb-rgm-aa-ff-ca-zi-eq-nls-wn-fwn-lrd-ref70kb-rc-rr-res-pne-nnd-np-rb0.3-lag-log-bc-rd-guid1.4-2147-is-lo}{(0.010, 222.78) (0.015, 251.45) (0.020, 262.86) (0.025, 269.17) (0.030, 275.23) (0.035, 279.44) (0.040, 284.02) (0.045, 287.97) (0.050, 291.92) (0.055, 294.69) (0.060, 298.56) (0.065, 302.82) (0.070, 309.60) (0.075, 310.26) (0.080, 313.76) (0.085, 315.26) (0.090, 320.23) (0.095, 323.31) (0.100, 329.12) (0.105, 333.11) (0.110, 333.90) (0.115, 338.30) (0.120, 342.56) (0.125, 342.36) (0.130, 348.06) (0.135, 352.40) (0.140, 352.98) (0.145, 355.51) (0.150, 362.45) (0.155, 361.78) (0.160, 365.82) (0.165, 368.58) (0.170, 370.21) (0.175, 372.32) (0.180, 372.72) (0.185, 374.49) (0.190, 373.90) (0.195, 375.10) (0.200, 377.88)}
\defdata{img512-cfgB-kl-rb-rmb-rgsb-rgm-aa-ff-ca-zi-eq-nls-wn-fwn-lrd-ref70kb-rc-rr-res-pne-nnd-np-rb0.3-lag-log-bc-rd-guid1.4-2147-is-hi}{(0.010, 225.38) (0.015, 253.30) (0.020, 263.86) (0.025, 270.10) (0.030, 277.13) (0.035, 281.34) (0.040, 286.18) (0.045, 289.04) (0.050, 294.53) (0.055, 296.68) (0.060, 301.08) (0.065, 307.89) (0.070, 310.47) (0.075, 311.58) (0.080, 316.22) (0.085, 320.78) (0.090, 322.73) (0.095, 326.85) (0.100, 329.99) (0.105, 335.33) (0.110, 335.41) (0.115, 342.15) (0.120, 343.93) (0.125, 347.02) (0.130, 350.22) (0.135, 355.75) (0.140, 355.81) (0.145, 359.51) (0.150, 364.71) (0.155, 366.11) (0.160, 367.50) (0.165, 370.00) (0.170, 371.88) (0.175, 374.75) (0.180, 375.31) (0.185, 376.67) (0.190, 378.62) (0.195, 377.11) (0.200, 378.90)}
\defdata{img512-cfgB-kl-rb-rmb-rgsb-rgm-aa-ff-ca-zi-eq-nls-wn-fwn-lrd-ref70kb-rc-rr-res-pne-nnd-np-rb0.3-lag-log-bc-rd-guid1.4-2147-recall3}{0.610}
\defdata{img512-cfgB-kl-rb-rmb-rgsb-rgm-aa-ff-ca-zi-eq-nls-wn-fwn-lrd-ref70kb-rc-rr-res-pne-nnd-np-rb0.3-lag-log-bc-rd-guid1.4-2147-recall-phema3}{0.140}
\defdata{img512-cfgB-kl-rb-rmb-rgsb-rgm-aa-ff-ca-zi-eq-nls-wn-fwn-lrd-ref70kb-rc-rr-res-pne-nnd-np-rb0.3-lag-log-bc-rd-guid1.4-2147-recall-lo}{(0.010, 0.5328) (0.015, 0.5647) (0.020, 0.5707) (0.025, 0.5732) (0.030, 0.5781) (0.035, 0.5789) (0.040, 0.5797) (0.045, 0.5843) (0.050, 0.5850) (0.055, 0.5902) (0.060, 0.5878) (0.065, 0.5916) (0.070, 0.5920) (0.075, 0.5922) (0.080, 0.5965) (0.085, 0.5968) (0.090, 0.5962) (0.095, 0.5983) (0.100, 0.6017) (0.105, 0.6001) (0.110, 0.6022) (0.115, 0.6024) (0.120, 0.6009) (0.125, 0.6038) (0.130, 0.6017) (0.135, 0.6074) (0.140, 0.6064) (0.145, 0.6065) (0.150, 0.6076) (0.155, 0.6067) (0.160, 0.6060) (0.165, 0.6019) (0.170, 0.6009) (0.175, 0.5973) (0.180, 0.5919) (0.185, 0.5874) (0.190, 0.5859) (0.195, 0.5820) (0.200, 0.5771)}
\defdata{img512-cfgB-kl-rb-rmb-rgsb-rgm-aa-ff-ca-zi-eq-nls-wn-fwn-lrd-ref70kb-rc-rr-res-pne-nnd-np-rb0.3-lag-log-bc-rd-guid1.4-2147-recall-hi}{(0.010, 0.5383) (0.015, 0.5687) (0.020, 0.5740) (0.025, 0.5774) (0.030, 0.5811) (0.035, 0.5843) (0.040, 0.5847) (0.045, 0.5871) (0.050, 0.5884) (0.055, 0.5946) (0.060, 0.5909) (0.065, 0.5926) (0.070, 0.5958) (0.075, 0.5995) (0.080, 0.5993) (0.085, 0.5989) (0.090, 0.5974) (0.095, 0.6000) (0.100, 0.6031) (0.105, 0.6054) (0.110, 0.6055) (0.115, 0.6079) (0.120, 0.6069) (0.125, 0.6089) (0.130, 0.6080) (0.135, 0.6099) (0.140, 0.6100) (0.145, 0.6095) (0.150, 0.6098) (0.155, 0.6099) (0.160, 0.6087) (0.165, 0.6086) (0.170, 0.6075) (0.175, 0.6032) (0.180, 0.5998) (0.185, 0.5919) (0.190, 0.5914) (0.195, 0.5859) (0.200, 0.5789)}

\defdata{img512-cfgB-kl-rb-rmb-rgsb-rgm-aa-ff-ca-zi-eq-nls-wn-fwn-lrd-ref70kb-rc-rr-res-pne-nnd-np-rb0.3-lag-log-bc-rd-guid1.5-2147-fid2}{2.36}
\defdata{img512-cfgB-kl-rb-rmb-rgsb-rgm-aa-ff-ca-zi-eq-nls-wn-fwn-lrd-ref70kb-rc-rr-res-pne-nnd-np-rb0.3-lag-log-bc-rd-guid1.5-2147-fid4}{2.3558}
\defdata{img512-cfgB-kl-rb-rmb-rgsb-rgm-aa-ff-ca-zi-eq-nls-wn-fwn-lrd-ref70kb-rc-rr-res-pne-nnd-np-rb0.3-lag-log-bc-rd-guid1.5-2147-phema3}{0.020}
\defdata{img512-cfgB-kl-rb-rmb-rgsb-rgm-aa-ff-ca-zi-eq-nls-wn-fwn-lrd-ref70kb-rc-rr-res-pne-nnd-np-rb0.3-lag-log-bc-rd-guid1.5-2147-lo}{(0.010, 3.7879) (0.015, 2.4876) (0.020, 2.3558) (0.025, 2.3591) (0.030, 2.3674) (0.035, 2.4683) (0.040, 2.5543) (0.045, 2.6710) (0.050, 2.7997) (0.055, 2.8680) (0.060, 3.0271) (0.065, 3.0804) (0.070, 3.2306) (0.075, 3.3501) (0.080, 3.5055) (0.085, 3.6902) (0.090, 3.8026) (0.095, 4.0348) (0.100, 4.2157) (0.105, 4.4132) (0.110, 4.6526) (0.115, 4.7768) (0.120, 5.1633) (0.125, 5.3716) (0.130, 5.6078) (0.135, 5.8814) (0.140, 6.1030) (0.145, 6.4282) (0.150, 6.7633) (0.155, 6.8675) (0.160, 7.1057) (0.165, 7.4821) (0.170, 7.8045) (0.175, 8.1242) (0.180, 8.4242) (0.185, 8.7337) (0.190, 8.9066) (0.195, 9.2065) (0.200, 9.4771)}
\defdata{img512-cfgB-kl-rb-rmb-rgsb-rgm-aa-ff-ca-zi-eq-nls-wn-fwn-lrd-ref70kb-rc-rr-res-pne-nnd-np-rb0.3-lag-log-bc-rd-guid1.5-2147-hi}{(0.010, 3.8082) (0.015, 2.5738) (0.020, 2.4065) (0.025, 2.4086) (0.030, 2.5208) (0.035, 2.5937) (0.040, 2.6373) (0.045, 2.7028) (0.050, 2.8327) (0.055, 2.9019) (0.060, 3.0455) (0.065, 3.2555) (0.070, 3.3199) (0.075, 3.4436) (0.080, 3.6838) (0.085, 3.8336) (0.090, 4.0143) (0.095, 4.1226) (0.100, 4.3248) (0.105, 4.5232) (0.110, 4.7279) (0.115, 4.8722) (0.120, 5.2128) (0.125, 5.5113) (0.130, 5.6768) (0.135, 5.9350) (0.140, 6.2189) (0.145, 6.4968) (0.150, 6.8371) (0.155, 6.9827) (0.160, 7.2050) (0.165, 7.5892) (0.170, 7.8762) (0.175, 8.1381) (0.180, 8.5831) (0.185, 8.7926) (0.190, 9.0656) (0.195, 9.3860) (0.200, 9.7490)}
\defdata{img512-cfgB-kl-rb-rmb-rgsb-rgm-aa-ff-ca-zi-eq-nls-wn-fwn-lrd-ref70kb-rc-rr-res-pne-nnd-np-rb0.3-lag-log-bc-rd-guid1.5-2147-pkl}{1533/52/phema/2147483-0.020.pkl}
\defdata{img512-cfgB-kl-rb-rmb-rgsb-rgm-aa-ff-ca-zi-eq-nls-wn-fwn-lrd-ref70kb-rc-rr-res-pne-nnd-np-rb0.3-lag-log-bc-rd-guid1.5-2147-is1}{390.2}
\defdata{img512-cfgB-kl-rb-rmb-rgsb-rgm-aa-ff-ca-zi-eq-nls-wn-fwn-lrd-ref70kb-rc-rr-res-pne-nnd-np-rb0.3-lag-log-bc-rd-guid1.5-2147-is-phema3}{0.200}
\defdata{img512-cfgB-kl-rb-rmb-rgsb-rgm-aa-ff-ca-zi-eq-nls-wn-fwn-lrd-ref70kb-rc-rr-res-pne-nnd-np-rb0.3-lag-log-bc-rd-guid1.5-2147-is-lo}{(0.010, 244.14) (0.015, 272.59) (0.020, 283.06) (0.025, 290.93) (0.030, 293.94) (0.035, 300.37) (0.040, 303.23) (0.045, 308.67) (0.050, 309.23) (0.055, 314.80) (0.060, 317.71) (0.065, 322.10) (0.070, 326.57) (0.075, 329.29) (0.080, 332.20) (0.085, 335.79) (0.090, 339.34) (0.095, 341.11) (0.100, 347.07) (0.105, 348.95) (0.110, 352.14) (0.115, 353.32) (0.120, 358.53) (0.125, 360.66) (0.130, 365.29) (0.135, 365.02) (0.140, 368.28) (0.145, 368.67) (0.150, 373.98) (0.155, 375.42) (0.160, 377.82) (0.165, 379.35) (0.170, 381.89) (0.175, 382.59) (0.180, 386.02) (0.185, 385.10) (0.190, 384.26) (0.195, 387.67) (0.200, 386.86)}
\defdata{img512-cfgB-kl-rb-rmb-rgsb-rgm-aa-ff-ca-zi-eq-nls-wn-fwn-lrd-ref70kb-rc-rr-res-pne-nnd-np-rb0.3-lag-log-bc-rd-guid1.5-2147-is-hi}{(0.010, 245.20) (0.015, 275.85) (0.020, 284.31) (0.025, 291.77) (0.030, 297.41) (0.035, 301.52) (0.040, 305.99) (0.045, 309.31) (0.050, 313.37) (0.055, 316.22) (0.060, 320.91) (0.065, 323.99) (0.070, 327.68) (0.075, 331.57) (0.080, 335.42) (0.085, 339.57) (0.090, 342.63) (0.095, 342.89) (0.100, 348.89) (0.105, 351.27) (0.110, 355.15) (0.115, 354.70) (0.120, 360.86) (0.125, 364.17) (0.130, 366.16) (0.135, 367.41) (0.140, 370.31) (0.145, 374.59) (0.150, 377.04) (0.155, 378.33) (0.160, 378.73) (0.165, 382.43) (0.170, 383.35) (0.175, 385.62) (0.180, 388.92) (0.185, 386.95) (0.190, 389.09) (0.195, 390.05) (0.200, 390.20)}
\defdata{img512-cfgB-kl-rb-rmb-rgsb-rgm-aa-ff-ca-zi-eq-nls-wn-fwn-lrd-ref70kb-rc-rr-res-pne-nnd-np-rb0.3-lag-log-bc-rd-guid1.5-2147-recall3}{0.608}
\defdata{img512-cfgB-kl-rb-rmb-rgsb-rgm-aa-ff-ca-zi-eq-nls-wn-fwn-lrd-ref70kb-rc-rr-res-pne-nnd-np-rb0.3-lag-log-bc-rd-guid1.5-2147-recall-phema3}{0.145}
\defdata{img512-cfgB-kl-rb-rmb-rgsb-rgm-aa-ff-ca-zi-eq-nls-wn-fwn-lrd-ref70kb-rc-rr-res-pne-nnd-np-rb0.3-lag-log-bc-rd-guid1.5-2147-recall-lo}{(0.010, 0.5319) (0.015, 0.5712) (0.020, 0.5754) (0.025, 0.5809) (0.030, 0.5827) (0.035, 0.5844) (0.040, 0.5853) (0.045, 0.5860) (0.050, 0.5908) (0.055, 0.5888) (0.060, 0.5897) (0.065, 0.5922) (0.070, 0.5930) (0.075, 0.5940) (0.080, 0.5974) (0.085, 0.5975) (0.090, 0.5986) (0.095, 0.6001) (0.100, 0.5984) (0.105, 0.6005) (0.110, 0.6045) (0.115, 0.6022) (0.120, 0.6007) (0.125, 0.6002) (0.130, 0.6036) (0.135, 0.5991) (0.140, 0.6020) (0.145, 0.6041) (0.150, 0.6002) (0.155, 0.6005) (0.160, 0.6002) (0.165, 0.5963) (0.170, 0.5937) (0.175, 0.5890) (0.180, 0.5849) (0.185, 0.5813) (0.190, 0.5745) (0.195, 0.5712) (0.200, 0.5657)}
\defdata{img512-cfgB-kl-rb-rmb-rgsb-rgm-aa-ff-ca-zi-eq-nls-wn-fwn-lrd-ref70kb-rc-rr-res-pne-nnd-np-rb0.3-lag-log-bc-rd-guid1.5-2147-recall-hi}{(0.010, 0.5386) (0.015, 0.5741) (0.020, 0.5774) (0.025, 0.5851) (0.030, 0.5840) (0.035, 0.5880) (0.040, 0.5898) (0.045, 0.5910) (0.050, 0.5920) (0.055, 0.5945) (0.060, 0.5934) (0.065, 0.5947) (0.070, 0.5960) (0.075, 0.5973) (0.080, 0.6014) (0.085, 0.6011) (0.090, 0.6017) (0.095, 0.6042) (0.100, 0.6029) (0.105, 0.6013) (0.110, 0.6054) (0.115, 0.6028) (0.120, 0.6069) (0.125, 0.6058) (0.130, 0.6072) (0.135, 0.6047) (0.140, 0.6066) (0.145, 0.6076) (0.150, 0.6053) (0.155, 0.6047) (0.160, 0.6040) (0.165, 0.6000) (0.170, 0.5979) (0.175, 0.5968) (0.180, 0.5948) (0.185, 0.5838) (0.190, 0.5833) (0.195, 0.5738) (0.200, 0.5730)}

\defdata{img512-cfgB-kl-rb-rmb-rgsb-rgm-aa-ff-ca-zi-eq-nls-wn-fwn-lrd-ref70kb-rc-rr-res-pne-nnd-np-rb0.3-lag-log-bc-rd-guid1.9-2147-fid2}{3.66}
\defdata{img512-cfgB-kl-rb-rmb-rgsb-rgm-aa-ff-ca-zi-eq-nls-wn-fwn-lrd-ref70kb-rc-rr-res-pne-nnd-np-rb0.3-lag-log-bc-rd-guid1.9-2147-fid4}{3.6580}
\defdata{img512-cfgB-kl-rb-rmb-rgsb-rgm-aa-ff-ca-zi-eq-nls-wn-fwn-lrd-ref70kb-rc-rr-res-pne-nnd-np-rb0.3-lag-log-bc-rd-guid1.9-2147-phema3}{0.015}
\defdata{img512-cfgB-kl-rb-rmb-rgsb-rgm-aa-ff-ca-zi-eq-nls-wn-fwn-lrd-ref70kb-rc-rr-res-pne-nnd-np-rb0.3-lag-log-bc-rd-guid1.9-2147-lo}{(0.010, 4.3821) (0.015, 3.6580) (0.020, 3.7560) (0.025, 3.7554) (0.030, 3.9428) (0.035, 4.1559) (0.040, 4.2142) (0.045, 4.4061) (0.050, 4.5078) (0.055, 4.7772) (0.060, 4.9761) (0.065, 5.1342) (0.070, 5.3153) (0.075, 5.4317) (0.080, 5.6869) (0.085, 5.9052) (0.090, 6.0337) (0.095, 6.2699) (0.100, 6.4040) (0.105, 6.5760) (0.110, 6.9575) (0.115, 7.1276) (0.120, 7.3483) (0.125, 7.6135) (0.130, 7.8024) (0.135, 8.0306) (0.140, 8.2568) (0.145, 8.4846) (0.150, 8.9525) (0.155, 9.0940) (0.160, 9.2675) (0.165, 9.5706) (0.170, 9.8932) (0.175, 10.0811) (0.180, 10.1830) (0.185, 10.4096) (0.190, 10.7087) (0.195, 10.8914) (0.200, 11.0937)}
\defdata{img512-cfgB-kl-rb-rmb-rgsb-rgm-aa-ff-ca-zi-eq-nls-wn-fwn-lrd-ref70kb-rc-rr-res-pne-nnd-np-rb0.3-lag-log-bc-rd-guid1.9-2147-hi}{(0.010, 4.4569) (0.015, 3.7402) (0.020, 3.8251) (0.025, 3.8876) (0.030, 4.0200) (0.035, 4.2029) (0.040, 4.3143) (0.045, 4.5022) (0.050, 4.7082) (0.055, 4.7982) (0.060, 5.1314) (0.065, 5.1708) (0.070, 5.3591) (0.075, 5.5391) (0.080, 5.7159) (0.085, 6.0219) (0.090, 6.0800) (0.095, 6.4140) (0.100, 6.6314) (0.105, 6.8606) (0.110, 6.9967) (0.115, 7.2748) (0.120, 7.5804) (0.125, 7.7338) (0.130, 7.9304) (0.135, 8.1633) (0.140, 8.5163) (0.145, 8.6575) (0.150, 9.0306) (0.155, 9.2434) (0.160, 9.3963) (0.165, 9.6805) (0.170, 9.9979) (0.175, 10.1993) (0.180, 10.4078) (0.185, 10.6789) (0.190, 10.8460) (0.195, 11.0504) (0.200, 11.3081)}
\defdata{img512-cfgB-kl-rb-rmb-rgsb-rgm-aa-ff-ca-zi-eq-nls-wn-fwn-lrd-ref70kb-rc-rr-res-pne-nnd-np-rb0.3-lag-log-bc-rd-guid1.9-2147-pkl}{1533/52/phema/2147483-0.015.pkl}
\defdata{img512-cfgB-kl-rb-rmb-rgsb-rgm-aa-ff-ca-zi-eq-nls-wn-fwn-lrd-ref70kb-rc-rr-res-pne-nnd-np-rb0.3-lag-log-bc-rd-guid1.9-2147-is1}{414.7}
\defdata{img512-cfgB-kl-rb-rmb-rgsb-rgm-aa-ff-ca-zi-eq-nls-wn-fwn-lrd-ref70kb-rc-rr-res-pne-nnd-np-rb0.3-lag-log-bc-rd-guid1.9-2147-is-phema3}{0.185}
\defdata{img512-cfgB-kl-rb-rmb-rgsb-rgm-aa-ff-ca-zi-eq-nls-wn-fwn-lrd-ref70kb-rc-rr-res-pne-nnd-np-rb0.3-lag-log-bc-rd-guid1.9-2147-is-lo}{(0.010, 298.96) (0.015, 331.77) (0.020, 340.21) (0.025, 345.84) (0.030, 351.44) (0.035, 355.01) (0.040, 359.04) (0.045, 358.39) (0.050, 362.79) (0.055, 366.38) (0.060, 368.56) (0.065, 369.13) (0.070, 374.61) (0.075, 377.46) (0.080, 377.87) (0.085, 381.20) (0.090, 381.88) (0.095, 384.21) (0.100, 388.42) (0.105, 390.27) (0.110, 392.42) (0.115, 393.83) (0.120, 398.08) (0.125, 397.98) (0.130, 397.89) (0.135, 401.67) (0.140, 402.29) (0.145, 402.63) (0.150, 405.96) (0.155, 405.94) (0.160, 409.66) (0.165, 409.31) (0.170, 406.74) (0.175, 409.19) (0.180, 409.78) (0.185, 411.49) (0.190, 410.32) (0.195, 411.04) (0.200, 408.59)}
\defdata{img512-cfgB-kl-rb-rmb-rgsb-rgm-aa-ff-ca-zi-eq-nls-wn-fwn-lrd-ref70kb-rc-rr-res-pne-nnd-np-rb0.3-lag-log-bc-rd-guid1.9-2147-is-hi}{(0.010, 302.91) (0.015, 336.84) (0.020, 345.71) (0.025, 349.91) (0.030, 353.04) (0.035, 356.77) (0.040, 360.62) (0.045, 363.91) (0.050, 365.34) (0.055, 367.88) (0.060, 370.66) (0.065, 374.31) (0.070, 375.14) (0.075, 379.60) (0.080, 379.46) (0.085, 383.07) (0.090, 387.32) (0.095, 388.70) (0.100, 389.65) (0.105, 393.81) (0.110, 393.26) (0.115, 394.82) (0.120, 400.60) (0.125, 399.31) (0.130, 399.97) (0.135, 402.98) (0.140, 406.30) (0.145, 405.50) (0.150, 409.05) (0.155, 408.13) (0.160, 411.99) (0.165, 410.50) (0.170, 411.63) (0.175, 413.45) (0.180, 410.09) (0.185, 414.72) (0.190, 414.02) (0.195, 413.40) (0.200, 412.82)}
\defdata{img512-cfgB-kl-rb-rmb-rgsb-rgm-aa-ff-ca-zi-eq-nls-wn-fwn-lrd-ref70kb-rc-rr-res-pne-nnd-np-rb0.3-lag-log-bc-rd-guid1.9-2147-recall3}{0.597}
\defdata{img512-cfgB-kl-rb-rmb-rgsb-rgm-aa-ff-ca-zi-eq-nls-wn-fwn-lrd-ref70kb-rc-rr-res-pne-nnd-np-rb0.3-lag-log-bc-rd-guid1.9-2147-recall-phema3}{0.070}
\defdata{img512-cfgB-kl-rb-rmb-rgsb-rgm-aa-ff-ca-zi-eq-nls-wn-fwn-lrd-ref70kb-rc-rr-res-pne-nnd-np-rb0.3-lag-log-bc-rd-guid1.9-2147-recall-lo}{(0.010, 0.5289) (0.015, 0.5734) (0.020, 0.5776) (0.025, 0.5868) (0.030, 0.5871) (0.035, 0.5848) (0.040, 0.5864) (0.045, 0.5862) (0.050, 0.5917) (0.055, 0.5905) (0.060, 0.5904) (0.065, 0.5911) (0.070, 0.5931) (0.075, 0.5880) (0.080, 0.5937) (0.085, 0.5911) (0.090, 0.5937) (0.095, 0.5862) (0.100, 0.5886) (0.105, 0.5930) (0.110, 0.5875) (0.115, 0.5880) (0.120, 0.5837) (0.125, 0.5833) (0.130, 0.5849) (0.135, 0.5774) (0.140, 0.5784) (0.145, 0.5719) (0.150, 0.5732) (0.155, 0.5706) (0.160, 0.5648) (0.165, 0.5633) (0.170, 0.5580) (0.175, 0.5563) (0.180, 0.5456) (0.185, 0.5414) (0.190, 0.5383) (0.195, 0.5329) (0.200, 0.5234)}
\defdata{img512-cfgB-kl-rb-rmb-rgsb-rgm-aa-ff-ca-zi-eq-nls-wn-fwn-lrd-ref70kb-rc-rr-res-pne-nnd-np-rb0.3-lag-log-bc-rd-guid1.9-2147-recall-hi}{(0.010, 0.5338) (0.015, 0.5759) (0.020, 0.5867) (0.025, 0.5882) (0.030, 0.5916) (0.035, 0.5900) (0.040, 0.5905) (0.045, 0.5916) (0.050, 0.5947) (0.055, 0.5943) (0.060, 0.5918) (0.065, 0.5937) (0.070, 0.5969) (0.075, 0.5956) (0.080, 0.5951) (0.085, 0.5933) (0.090, 0.5950) (0.095, 0.5943) (0.100, 0.5915) (0.105, 0.5946) (0.110, 0.5936) (0.115, 0.5904) (0.120, 0.5889) (0.125, 0.5881) (0.130, 0.5871) (0.135, 0.5838) (0.140, 0.5859) (0.145, 0.5804) (0.150, 0.5786) (0.155, 0.5753) (0.160, 0.5729) (0.165, 0.5676) (0.170, 0.5697) (0.175, 0.5608) (0.180, 0.5572) (0.185, 0.5502) (0.190, 0.5425) (0.195, 0.5391) (0.200, 0.5310)}

\defdata{img512-b9-mc448-lr65-lim65-drop0.10-XS-guid1.2-940-fid2}{1.81}
\defdata{img512-b9-mc448-lr65-lim65-drop0.10-XS-guid1.2-940-fid4}{1.8096}
\defdata{img512-b9-mc448-lr65-lim65-drop0.10-XS-guid1.2-940-phema3}{0.015}
\defdata{img512-b9-mc448-lr65-lim65-drop0.10-XS-guid1.2-940-lo}{(0.010, 2.0420) (0.015, 1.8096) (0.020, 1.8198) (0.025, 1.8664) (0.030, 1.9199) (0.035, 2.0374) (0.040, 2.1702) (0.045, 2.2111) (0.050, 2.3845) (0.055, 2.4174) (0.060, 2.5831)}
\defdata{img512-b9-mc448-lr65-lim65-drop0.10-XS-guid1.2-940-hi}{(0.010, 2.0542) (0.015, 1.8366) (0.020, 1.8469) (0.025, 1.9326) (0.030, 2.0598) (0.035, 2.0952) (0.040, 2.2578) (0.045, 2.2950) (0.050, 2.4120) (0.055, 2.4648) (0.060, 2.6025)}
\defdata{img512-b9-mc448-lr65-lim65-drop0.10-XS-guid1.2-940-pkl}{1538/21/phema/939524-0.015.pkl}
\defdata{img512-b9-mc448-lr65-lim65-drop0.10-XS-guid1.2-940-is1}{310.8}
\defdata{img512-b9-mc448-lr65-lim65-drop0.10-XS-guid1.2-940-is-phema3}{0.060}
\defdata{img512-b9-mc448-lr65-lim65-drop0.10-XS-guid1.2-940-is-lo}{(0.010, 254.47) (0.015, 264.91) (0.020, 273.85) (0.025, 279.31) (0.030, 284.78) (0.035, 292.03) (0.040, 293.67) (0.045, 297.63) (0.050, 300.99) (0.055, 304.27) (0.060, 309.57)}
\defdata{img512-b9-mc448-lr65-lim65-drop0.10-XS-guid1.2-940-is-hi}{(0.010, 255.53) (0.015, 269.71) (0.020, 276.82) (0.025, 283.01) (0.030, 286.96) (0.035, 292.47) (0.040, 296.66) (0.045, 302.38) (0.050, 303.06) (0.055, 306.28) (0.060, 310.81)}
\defdata{img512-b9-mc448-lr65-lim65-drop0.10-XS-guid1.2-940-recall3}{0.635}
\defdata{img512-b9-mc448-lr65-lim65-drop0.10-XS-guid1.2-940-recall-phema3}{0.060}
\defdata{img512-b9-mc448-lr65-lim65-drop0.10-XS-guid1.2-940-recall-lo}{(0.010, 0.5902) (0.015, 0.6108) (0.020, 0.6168) (0.025, 0.6190) (0.030, 0.6210) (0.035, 0.6277) (0.040, 0.6292) (0.045, 0.6280) (0.050, 0.6312) (0.055, 0.6319) (0.060, 0.6293)}
\defdata{img512-b9-mc448-lr65-lim65-drop0.10-XS-guid1.2-940-recall-hi}{(0.010, 0.5915) (0.015, 0.6111) (0.020, 0.6206) (0.025, 0.6211) (0.030, 0.6260) (0.035, 0.6300) (0.040, 0.6311) (0.045, 0.6308) (0.050, 0.6331) (0.055, 0.6338) (0.060, 0.6349)}

\defdata{img512-b9-mc448-lr65-lim65-drop0.10-S-guid1.2-940-fid2}{1.80}
\defdata{img512-b9-mc448-lr65-lim65-drop0.10-S-guid1.2-940-fid4}{1.8048}
\defdata{img512-b9-mc448-lr65-lim65-drop0.10-S-guid1.2-940-phema3}{0.015}
\defdata{img512-b9-mc448-lr65-lim65-drop0.10-S-guid1.2-940-lo}{(0.010, 2.0223) (0.015, 1.8048) (0.020, 1.8217) (0.025, 1.9017) (0.030, 1.9621) (0.035, 2.0332) (0.040, 2.0904) (0.045, 2.1846) (0.050, 2.2504)}
\defdata{img512-b9-mc448-lr65-lim65-drop0.10-S-guid1.2-940-hi}{(0.010, 2.0583) (0.015, 1.8510) (0.020, 1.8586) (0.025, 1.9411) (0.030, 1.9843) (0.035, 2.0444) (0.040, 2.1797) (0.045, 2.2866) (0.050, 2.3336)}
\defdata{img512-b9-mc448-lr65-lim65-drop0.10-S-guid1.2-940-pkl}{1538/21/phema/939524-0.015.pkl}
\defdata{img512-b9-mc448-lr65-lim65-drop0.10-S-guid1.2-940-is1}{300.9}
\defdata{img512-b9-mc448-lr65-lim65-drop0.10-S-guid1.2-940-is-phema3}{0.050}
\defdata{img512-b9-mc448-lr65-lim65-drop0.10-S-guid1.2-940-is-lo}{(0.010, 249.07) (0.015, 261.35) (0.020, 272.13) (0.025, 275.33) (0.030, 284.24) (0.035, 288.51) (0.040, 291.12) (0.045, 293.80) (0.050, 299.60)}
\defdata{img512-b9-mc448-lr65-lim65-drop0.10-S-guid1.2-940-is-hi}{(0.010, 251.83) (0.015, 265.03) (0.020, 272.73) (0.025, 278.81) (0.030, 285.45) (0.035, 290.44) (0.040, 293.65) (0.045, 296.55) (0.050, 300.94)}
\defdata{img512-b9-mc448-lr65-lim65-drop0.10-S-guid1.2-940-recall3}{0.634}
\defdata{img512-b9-mc448-lr65-lim65-drop0.10-S-guid1.2-940-recall-phema3}{0.050}
\defdata{img512-b9-mc448-lr65-lim65-drop0.10-S-guid1.2-940-recall-lo}{(0.010, 0.5958) (0.015, 0.6106) (0.020, 0.6147) (0.025, 0.6175) (0.030, 0.6194) (0.035, 0.6228) (0.040, 0.6256) (0.045, 0.6274) (0.050, 0.6280)}
\defdata{img512-b9-mc448-lr65-lim65-drop0.10-S-guid1.2-940-recall-hi}{(0.010, 0.5991) (0.015, 0.6136) (0.020, 0.6169) (0.025, 0.6217) (0.030, 0.6227) (0.035, 0.6250) (0.040, 0.6273) (0.045, 0.6292) (0.050, 0.6338)}

\defdata{img512-b9-mc448-lr65-lim65-drop0.10-M-guid1.2-940-fid2}{1.80}
\defdata{img512-b9-mc448-lr65-lim65-drop0.10-M-guid1.2-940-fid4}{1.8013}
\defdata{img512-b9-mc448-lr65-lim65-drop0.10-M-guid1.2-940-phema3}{0.015}
\defdata{img512-b9-mc448-lr65-lim65-drop0.10-M-guid1.2-940-lo}{(0.010, 2.0978) (0.015, 1.8013) (0.020, 1.9004) (0.025, 1.8724) (0.030, 1.9863) (0.035, 2.0447) (0.040, 2.0875) (0.045, 2.1888) (0.050, 2.3408)}
\defdata{img512-b9-mc448-lr65-lim65-drop0.10-M-guid1.2-940-hi}{(0.010, 2.1481) (0.015, 1.9030) (0.020, 1.9141) (0.025, 1.9094) (0.030, 1.9991) (0.035, 2.1095) (0.040, 2.1065) (0.045, 2.2089) (0.050, 2.3695)}
\defdata{img512-b9-mc448-lr65-lim65-drop0.10-M-guid1.2-940-pkl}{1538/21/phema/939524-0.015.pkl}
\defdata{img512-b9-mc448-lr65-lim65-drop0.10-M-guid1.2-940-is1}{301.4}
\defdata{img512-b9-mc448-lr65-lim65-drop0.10-M-guid1.2-940-is-phema3}{0.050}
\defdata{img512-b9-mc448-lr65-lim65-drop0.10-M-guid1.2-940-is-lo}{(0.010, 249.29) (0.015, 263.51) (0.020, 270.99) (0.025, 274.62) (0.030, 283.23) (0.035, 285.02) (0.040, 287.82) (0.045, 294.84) (0.050, 298.68)}
\defdata{img512-b9-mc448-lr65-lim65-drop0.10-M-guid1.2-940-is-hi}{(0.010, 252.58) (0.015, 264.80) (0.020, 271.36) (0.025, 278.65) (0.030, 283.86) (0.035, 287.30) (0.040, 291.23) (0.045, 296.04) (0.050, 301.42)}
\defdata{img512-b9-mc448-lr65-lim65-drop0.10-M-guid1.2-940-recall3}{0.629}
\defdata{img512-b9-mc448-lr65-lim65-drop0.10-M-guid1.2-940-recall-phema3}{0.045}
\defdata{img512-b9-mc448-lr65-lim65-drop0.10-M-guid1.2-940-recall-lo}{(0.010, 0.5841) (0.015, 0.6043) (0.020, 0.6124) (0.025, 0.6148) (0.030, 0.6182) (0.035, 0.6204) (0.040, 0.6225) (0.045, 0.6237) (0.050, 0.6247)}
\defdata{img512-b9-mc448-lr65-lim65-drop0.10-M-guid1.2-940-recall-hi}{(0.010, 0.5874) (0.015, 0.6091) (0.020, 0.6164) (0.025, 0.6198) (0.030, 0.6213) (0.035, 0.6264) (0.040, 0.6245) (0.045, 0.6287) (0.050, 0.6278)}

\defdata{img512-b9-mc448-lr65-lim65-drop0.10-L-guid1.2-940-fid2}{1.86}
\defdata{img512-b9-mc448-lr65-lim65-drop0.10-L-guid1.2-940-fid4}{1.8564}
\defdata{img512-b9-mc448-lr65-lim65-drop0.10-L-guid1.2-940-phema3}{0.020}
\defdata{img512-b9-mc448-lr65-lim65-drop0.10-L-guid1.2-940-lo}{(0.010, 2.1199) (0.015, 1.8598) (0.020, 1.8564) (0.025, 1.8929) (0.030, 1.9048) (0.035, 2.0003) (0.040, 2.0578) (0.045, 2.2134) (0.050, 2.2230)}
\defdata{img512-b9-mc448-lr65-lim65-drop0.10-L-guid1.2-940-hi}{(0.010, 2.1886) (0.015, 1.8926) (0.020, 1.8736) (0.025, 1.9392) (0.030, 2.0047) (0.035, 2.0355) (0.040, 2.1744) (0.045, 2.2254) (0.050, 2.2778)}
\defdata{img512-b9-mc448-lr65-lim65-drop0.10-L-guid1.2-940-pkl}{1538/21/phema/939524-0.020.pkl}
\defdata{img512-b9-mc448-lr65-lim65-drop0.10-L-guid1.2-940-is1}{298.7}
\defdata{img512-b9-mc448-lr65-lim65-drop0.10-L-guid1.2-940-is-phema3}{0.050}
\defdata{img512-b9-mc448-lr65-lim65-drop0.10-L-guid1.2-940-is-lo}{(0.010, 247.54) (0.015, 260.02) (0.020, 269.12) (0.025, 274.10) (0.030, 281.26) (0.035, 284.25) (0.040, 288.04) (0.045, 294.22) (0.050, 294.04)}
\defdata{img512-b9-mc448-lr65-lim65-drop0.10-L-guid1.2-940-is-hi}{(0.010, 250.02) (0.015, 262.09) (0.020, 271.64) (0.025, 276.96) (0.030, 283.45) (0.035, 287.80) (0.040, 292.42) (0.045, 295.59) (0.050, 298.68)}
\defdata{img512-b9-mc448-lr65-lim65-drop0.10-L-guid1.2-940-recall3}{0.629}
\defdata{img512-b9-mc448-lr65-lim65-drop0.10-L-guid1.2-940-recall-phema3}{0.050}
\defdata{img512-b9-mc448-lr65-lim65-drop0.10-L-guid1.2-940-recall-lo}{(0.010, 0.5841) (0.015, 0.6027) (0.020, 0.6111) (0.025, 0.6128) (0.030, 0.6218) (0.035, 0.6202) (0.040, 0.6216) (0.045, 0.6227) (0.050, 0.6273)}
\defdata{img512-b9-mc448-lr65-lim65-drop0.10-L-guid1.2-940-recall-hi}{(0.010, 0.5860) (0.015, 0.6092) (0.020, 0.6144) (0.025, 0.6160) (0.030, 0.6243) (0.035, 0.6217) (0.040, 0.6274) (0.045, 0.6240) (0.050, 0.6293)}

\defdata{img512-b9-mc448-lr65-lim65-drop0.10-XL-guid1.2-940-fid2}{1.82}
\defdata{img512-b9-mc448-lr65-lim65-drop0.10-XL-guid1.2-940-fid4}{1.8160}
\defdata{img512-b9-mc448-lr65-lim65-drop0.10-XL-guid1.2-940-phema3}{0.020}
\defdata{img512-b9-mc448-lr65-lim65-drop0.10-XL-guid1.2-940-lo}{(0.010, 2.1530) (0.015, 1.8446) (0.020, 1.8160) (0.025, 1.8917) (0.030, 1.9094) (0.035, 1.9687) (0.040, 2.1035) (0.045, 2.1634) (0.050, 2.2106)}
\defdata{img512-b9-mc448-lr65-lim65-drop0.10-XL-guid1.2-940-hi}{(0.010, 2.1871) (0.015, 1.8818) (0.020, 1.8545) (0.025, 1.9492) (0.030, 1.9365) (0.035, 2.0441) (0.040, 2.1343) (0.045, 2.1886) (0.050, 2.2836)}
\defdata{img512-b9-mc448-lr65-lim65-drop0.10-XL-guid1.2-940-pkl}{1538/21/phema/939524-0.020.pkl}
\defdata{img512-b9-mc448-lr65-lim65-drop0.10-XL-guid1.2-940-is1}{299.0}
\defdata{img512-b9-mc448-lr65-lim65-drop0.10-XL-guid1.2-940-is-phema3}{0.050}
\defdata{img512-b9-mc448-lr65-lim65-drop0.10-XL-guid1.2-940-is-lo}{(0.010, 246.91) (0.015, 261.33) (0.020, 267.55) (0.025, 273.37) (0.030, 276.96) (0.035, 281.90) (0.040, 287.70) (0.045, 290.44) (0.050, 295.31)}
\defdata{img512-b9-mc448-lr65-lim65-drop0.10-XL-guid1.2-940-is-hi}{(0.010, 249.97) (0.015, 264.82) (0.020, 271.44) (0.025, 277.15) (0.030, 282.11) (0.035, 286.05) (0.040, 289.36) (0.045, 293.16) (0.050, 299.01)}
\defdata{img512-b9-mc448-lr65-lim65-drop0.10-XL-guid1.2-940-recall3}{0.628}
\defdata{img512-b9-mc448-lr65-lim65-drop0.10-XL-guid1.2-940-recall-phema3}{0.050}
\defdata{img512-b9-mc448-lr65-lim65-drop0.10-XL-guid1.2-940-recall-lo}{(0.010, 0.5825) (0.015, 0.6033) (0.020, 0.6114) (0.025, 0.6136) (0.030, 0.6158) (0.035, 0.6210) (0.040, 0.6205) (0.045, 0.6231) (0.050, 0.6226)}
\defdata{img512-b9-mc448-lr65-lim65-drop0.10-XL-guid1.2-940-recall-hi}{(0.010, 0.5874) (0.015, 0.6076) (0.020, 0.6138) (0.025, 0.6159) (0.030, 0.6202) (0.035, 0.6229) (0.040, 0.6248) (0.045, 0.6276) (0.050, 0.6277)}

\defdata{img512-b9-mc448-lr65-lim65-drop0.10-XXL-guid1.2-940-fid2}{1.85}
\defdata{img512-b9-mc448-lr65-lim65-drop0.10-XXL-guid1.2-940-fid4}{1.8542}
\defdata{img512-b9-mc448-lr65-lim65-drop0.10-XXL-guid1.2-940-phema3}{0.020}
\defdata{img512-b9-mc448-lr65-lim65-drop0.10-XXL-guid1.2-940-lo}{(0.010, 2.1530) (0.015, 1.8641) (0.020, 1.8542) (0.025, 1.9041) (0.030, 1.8812) (0.035, 1.9702) (0.040, 2.0681) (0.045, 2.1140) (0.050, 2.2482)}
\defdata{img512-b9-mc448-lr65-lim65-drop0.10-XXL-guid1.2-940-hi}{(0.010, 2.1867) (0.015, 1.9019) (0.020, 1.8807) (0.025, 1.9385) (0.030, 1.9259) (0.035, 2.0566) (0.040, 2.1369) (0.045, 2.1924) (0.050, 2.2798)}
\defdata{img512-b9-mc448-lr65-lim65-drop0.10-XXL-guid1.2-940-pkl}{1538/21/phema/939524-0.020.pkl}
\defdata{img512-b9-mc448-lr65-lim65-drop0.10-XXL-guid1.2-940-is1}{297.5}
\defdata{img512-b9-mc448-lr65-lim65-drop0.10-XXL-guid1.2-940-is-phema3}{0.050}
\defdata{img512-b9-mc448-lr65-lim65-drop0.10-XXL-guid1.2-940-is-lo}{(0.010, 248.78) (0.015, 259.29) (0.020, 268.87) (0.025, 273.52) (0.030, 278.81) (0.035, 285.53) (0.040, 287.40) (0.045, 290.21) (0.050, 295.17)}
\defdata{img512-b9-mc448-lr65-lim65-drop0.10-XXL-guid1.2-940-is-hi}{(0.010, 248.93) (0.015, 262.49) (0.020, 270.19) (0.025, 275.97) (0.030, 283.65) (0.035, 287.02) (0.040, 290.58) (0.045, 293.23) (0.050, 297.51)}
\defdata{img512-b9-mc448-lr65-lim65-drop0.10-XXL-guid1.2-940-recall3}{0.626}
\defdata{img512-b9-mc448-lr65-lim65-drop0.10-XXL-guid1.2-940-recall-phema3}{0.050}
\defdata{img512-b9-mc448-lr65-lim65-drop0.10-XXL-guid1.2-940-recall-lo}{(0.010, 0.5814) (0.015, 0.6020) (0.020, 0.6081) (0.025, 0.6137) (0.030, 0.6133) (0.035, 0.6194) (0.040, 0.6219) (0.045, 0.6199) (0.050, 0.6228)}
\defdata{img512-b9-mc448-lr65-lim65-drop0.10-XXL-guid1.2-940-recall-hi}{(0.010, 0.5826) (0.015, 0.6071) (0.020, 0.6116) (0.025, 0.6188) (0.030, 0.6181) (0.035, 0.6216) (0.040, 0.6251) (0.045, 0.6244) (0.050, 0.6259)}

\gdef\perTensorListAppendix{\begin{itemize}[leftmargin=2em]\item {\texttt{enc-64x64-block0-affine}}\item {\texttt{enc-64x64-block0-conv0}}\item {\texttt{enc-64x64-block0-conv1}}\item {\texttt{enc-64x64-conv}}\item {\texttt{enc-32x32-block0-conv0}}\item {\texttt{enc-32x32-block0-skip}}\item {\texttt{enc-16x16-block0-affine}}\item {\texttt{enc-16x16-block0-conv0}}\item {\texttt{enc-16x16-block2-conv0}}\item {\texttt{enc-8x8-block0-affine}}\item {\texttt{enc-8x8-block0-skip}}\item {\texttt{enc-8x8-block1-conv0}}\item {\texttt{enc-8x8-block2-conv0}}\item {\texttt{dec-8x8-block0-conv0}}\item {\texttt{dec-8x8-block2-skip}}\item {\texttt{dec-8x8-in0-affine}}\item {\texttt{dec-16x16-block0-affine}}\item {\texttt{dec-16x16-block0-conv1}}\item {\texttt{dec-16x16-block0-skip}}\item {\texttt{dec-32x32-block0-conv1}}\item {\texttt{dec-32x32-block0-skip}}\item {\texttt{dec-32x32-up-affine}}\item {\texttt{dec-64x64-block0-conv1}}\item {\texttt{dec-64x64-block0-skip}}\item {\texttt{dec-64x64-block3-skip}}\item {\texttt{dec-64x64-up-affine}}\item {\texttt{map-label}}\item {\texttt{map-layer0}}\end{itemize}}

\gdef\figPerTensorplotAA{\addplot[C1,opacity=0.3 ,forget plot,dash phase=49.7pt] coordinates {\rawdata{pertensor-phema-img512-cfgB-enc-64x64_conv}};}
\gdef\figPerTensorplotAB{\addplot[C1,opacity=0.3 ,forget plot,dash phase=49.8pt] coordinates {\rawdata{pertensor-phema-img512-cfgB-enc-64x64_block0-conv0}};}
\gdef\figPerTensorplotAC{\addplot[C1,opacity=0.3 ,forget plot,dash phase=0.4pt] coordinates {\rawdata{pertensor-phema-img512-cfgB-enc-64x64_block0-conv1}};}
\gdef\figPerTensorplotAD{\addplot[C1,opacity=0.3 ,forget plot,dash phase=48.0pt] coordinates {\rawdata{pertensor-phema-img512-cfgB-enc-32x32_block0-conv0}};}
\gdef\figPerTensorplotAE{\addplot[C1,opacity=0.3 ,forget plot,dash phase=1.7pt] coordinates {\rawdata{pertensor-phema-img512-cfgB-enc-16x16_block0-conv0}};}
\gdef\figPerTensorplotAF{\addplot[C1,opacity=0.3 ,forget plot,dash phase=25.3pt] coordinates {\rawdata{pertensor-phema-img512-cfgB-enc-16x16_block2-conv0}};}
\gdef\figPerTensorplotAG{\addplot[C1,opacity=0.3 ,forget plot,dash phase=7.0pt] coordinates {\rawdata{pertensor-phema-img512-cfgB-enc-8x8_block1-conv0}};}
\gdef\figPerTensorplotAH{\addplot[C1,opacity=0.3 ,forget plot,dash phase=37.4pt] coordinates {\rawdata{pertensor-phema-img512-cfgB-enc-8x8_block2-conv0}};}
\gdef\figPerTensorplotAI{\addplot[C1,opacity=0.3 ,forget plot,dash phase=40.0pt] coordinates {\rawdata{pertensor-phema-img512-cfgB-dec-8x8_block0-conv0}};}
\gdef\figPerTensorplotAJ{\addplot[C1,opacity=0.3 ,forget plot,dash phase=19.2pt] coordinates {\rawdata{pertensor-phema-img512-cfgB-dec-16x16_block0-conv1}};}
\gdef\figPerTensorplotAK{\addplot[C1,opacity=0.3 ,forget plot,dash phase=7.8pt] coordinates {\rawdata{pertensor-phema-img512-cfgB-dec-32x32_block0-conv1}};}
\gdef\figPerTensorplotAL{\addplot[C1,opacity=0.3 ,forget plot,dash phase=32.5pt] coordinates {\rawdata{pertensor-phema-img512-cfgB-dec-64x64_block0-conv1}};}
\gdef\figPerTensorplotAM{\addplot[C1,opacity=0.3 ,forget plot,dash phase=7.4pt] coordinates {\rawdata{pertensor-phema-img512-cfgB-map_layer0}};}
\gdef\figPerTensorplotAN{\addplot[C1,opacity=0.3 ,forget plot,dash phase=22.8pt] coordinates {\rawdata{pertensor-phema-img512-cfgB-map_label}};}
\gdef\figPerTensorplotAO{\addplot[C1,opacity=0.3 ,forget plot,dash phase=50.9pt] coordinates {\rawdata{pertensor-phema-img512-cfgB-enc-64x64_block0-affine}};}
\gdef\figPerTensorplotAP{\addplot[C1,opacity=0.3 ,forget plot,dash phase=22.9pt] coordinates {\rawdata{pertensor-phema-img512-cfgB-enc-16x16_block0-affine}};}
\gdef\figPerTensorplotAQ{\addplot[C1,opacity=0.3 ,forget plot,dash phase=32.1pt] coordinates {\rawdata{pertensor-phema-img512-cfgB-enc-8x8_block0-affine}};}
\gdef\figPerTensorplotAR{\addplot[C1,opacity=0.3 ,forget plot,dash phase=9.3pt] coordinates {\rawdata{pertensor-phema-img512-cfgB-dec-8x8_in0-affine}};}
\gdef\figPerTensorplotAS{\addplot[C1,opacity=0.3 ,forget plot,dash phase=7.3pt] coordinates {\rawdata{pertensor-phema-img512-cfgB-dec-16x16_block0-affine}};}
\gdef\figPerTensorplotAT{\addplot[C1,opacity=0.3 ,forget plot,dash phase=49.4pt] coordinates {\rawdata{pertensor-phema-img512-cfgB-dec-32x32_up-affine}};}
\gdef\figPerTensorplotAU{\addplot[C1,opacity=0.3 ,forget plot,dash phase=23.4pt] coordinates {\rawdata{pertensor-phema-img512-cfgB-dec-64x64_up-affine}};}
\gdef\figPerTensorplotAV{\addplot[C1,opacity=0.3 ,forget plot,dash phase=27.3pt] coordinates {\rawdata{pertensor-phema-img512-cfgB-enc-32x32_block0-skip}};}
\gdef\figPerTensorplotAW{\addplot[C1,opacity=0.3 ,forget plot,dash phase=36.8pt] coordinates {\rawdata{pertensor-phema-img512-cfgB-enc-8x8_block0-skip}};}
\gdef\figPerTensorplotAX{\addplot[C1,opacity=0.3 ,forget plot,dash phase=31.4pt] coordinates {\rawdata{pertensor-phema-img512-cfgB-dec-8x8_block2-skip}};}
\gdef\figPerTensorplotAY{\addplot[C1,opacity=0.3 ,forget plot,dash phase=28.5pt] coordinates {\rawdata{pertensor-phema-img512-cfgB-dec-16x16_block0-skip}};}
\gdef\figPerTensorplotAZ{\addplot[C1,opacity=0.3 ,forget plot,dash phase=10.7pt] coordinates {\rawdata{pertensor-phema-img512-cfgB-dec-32x32_block0-skip}};}
\gdef\figPerTensorplotBA{\addplot[C1,opacity=0.3 ,forget plot,dash phase=46.5pt] coordinates {\rawdata{pertensor-phema-img512-cfgB-dec-64x64_block0-skip}};}
\gdef\figPerTensorplotBB{\addplot[C1,opacity=0.3 ,forget plot,dash phase=49.1pt] coordinates {\rawdata{pertensor-phema-img512-cfgB-dec-64x64_block3-skip}};}
\gdef\figPerTensorplotBC{\addplot[C1!80!black,thick,forget plot] coordinates {\rawdata{pertensor-phema-img512-cfgB}};}
\gdef\figPerTensorplotBD{\addplot[C2,opacity=0.35,forget plot,dash phase=0.2pt] coordinates {\rawdata{pertensor-phema-img512-cfgG-enc-64x64_conv}};}
\gdef\figPerTensorplotBE{\addplot[C2,opacity=0.35,forget plot,dash phase=8.5pt] coordinates {\rawdata{pertensor-phema-img512-cfgG-enc-64x64_block0-conv0}};}
\gdef\figPerTensorplotBF{\addplot[C2,opacity=0.35,forget plot,dash phase=3.6pt] coordinates {\rawdata{pertensor-phema-img512-cfgG-enc-64x64_block0-conv1}};}
\gdef\figPerTensorplotBG{\addplot[C2,opacity=0.35,forget plot,dash phase=41.7pt] coordinates {\rawdata{pertensor-phema-img512-cfgG-enc-32x32_block0-conv0}};}
\gdef\figPerTensorplotBH{\addplot[C2,opacity=0.35,forget plot,dash phase=37.8pt] coordinates {\rawdata{pertensor-phema-img512-cfgG-enc-16x16_block0-conv0}};}
\gdef\figPerTensorplotBI{\addplot[C2,opacity=0.35,forget plot,dash phase=34.2pt] coordinates {\rawdata{pertensor-phema-img512-cfgG-enc-16x16_block2-conv0}};}
\gdef\figPerTensorplotBJ{\addplot[C2,opacity=0.35,forget plot,dash phase=17.0pt] coordinates {\rawdata{pertensor-phema-img512-cfgG-enc-8x8_block1-conv0}};}
\gdef\figPerTensorplotBK{\addplot[C2,opacity=0.35,forget plot,dash phase=12.5pt] coordinates {\rawdata{pertensor-phema-img512-cfgG-enc-8x8_block2-conv0}};}
\gdef\figPerTensorplotBL{\addplot[C2,opacity=0.35,forget plot,dash phase=36.6pt] coordinates {\rawdata{pertensor-phema-img512-cfgG-dec-8x8_block0-conv0}};}
\gdef\figPerTensorplotBM{\addplot[C2,opacity=0.35,forget plot,dash phase=13.3pt] coordinates {\rawdata{pertensor-phema-img512-cfgG-dec-16x16_block0-conv1}};}
\gdef\figPerTensorplotBN{\addplot[C2,opacity=0.35,forget plot,dash phase=4.1pt] coordinates {\rawdata{pertensor-phema-img512-cfgG-dec-32x32_block0-conv1}};}
\gdef\figPerTensorplotBO{\addplot[C2,opacity=0.35,forget plot,dash phase=46.2pt] coordinates {\rawdata{pertensor-phema-img512-cfgG-dec-64x64_block0-conv1}};}
\gdef\figPerTensorplotBP{\addplot[C2,opacity=0.35,forget plot,dash phase=36.7pt] coordinates {\rawdata{pertensor-phema-img512-cfgG-map_layer0}};}
\gdef\figPerTensorplotBQ{\addplot[C2,opacity=0.35,forget plot,dash phase=42.5pt] coordinates {\rawdata{pertensor-phema-img512-cfgG-map_label}};}
\gdef\figPerTensorplotBR{\addplot[C2,opacity=0.35,forget plot,dash phase=40.9pt] coordinates {\rawdata{pertensor-phema-img512-cfgG-enc-64x64_block0-affine}};}
\gdef\figPerTensorplotBS{\addplot[C2,opacity=0.35,forget plot,dash phase=48.4pt] coordinates {\rawdata{pertensor-phema-img512-cfgG-enc-16x16_block0-affine}};}
\gdef\figPerTensorplotBT{\addplot[C2,opacity=0.35,forget plot,dash phase=41.4pt] coordinates {\rawdata{pertensor-phema-img512-cfgG-enc-8x8_block0-affine}};}
\gdef\figPerTensorplotBU{\addplot[C2,opacity=0.35,forget plot,dash phase=17.6pt] coordinates {\rawdata{pertensor-phema-img512-cfgG-dec-8x8_in0-affine}};}
\gdef\figPerTensorplotBV{\addplot[C2,opacity=0.35,forget plot,dash phase=28.4pt] coordinates {\rawdata{pertensor-phema-img512-cfgG-dec-16x16_block0-affine}};}
\gdef\figPerTensorplotBW{\addplot[C2,opacity=0.35,forget plot,dash phase=27.0pt] coordinates {\rawdata{pertensor-phema-img512-cfgG-dec-32x32_up-affine}};}
\gdef\figPerTensorplotBX{\addplot[C2,opacity=0.35,forget plot,dash phase=36.9pt] coordinates {\rawdata{pertensor-phema-img512-cfgG-dec-64x64_up-affine}};}
\gdef\figPerTensorplotBY{\addplot[C2,opacity=0.35,forget plot,dash phase=48.4pt] coordinates {\rawdata{pertensor-phema-img512-cfgG-enc-32x32_block0-skip}};}
\gdef\figPerTensorplotBZ{\addplot[C2,opacity=0.35,forget plot,dash phase=5.0pt] coordinates {\rawdata{pertensor-phema-img512-cfgG-enc-8x8_block0-skip}};}
\gdef\figPerTensorplotCA{\addplot[C2,opacity=0.35,forget plot,dash phase=43.1pt] coordinates {\rawdata{pertensor-phema-img512-cfgG-dec-8x8_block2-skip}};}
\gdef\figPerTensorplotCB{\addplot[C2,opacity=0.35,forget plot,dash phase=13.0pt] coordinates {\rawdata{pertensor-phema-img512-cfgG-dec-16x16_block0-skip}};}
\gdef\figPerTensorplotCC{\addplot[C2,opacity=0.35,forget plot,dash phase=28.2pt] coordinates {\rawdata{pertensor-phema-img512-cfgG-dec-32x32_block0-skip}};}
\gdef\figPerTensorplotCD{\addplot[C2,opacity=0.35,forget plot,dash phase=37.9pt] coordinates {\rawdata{pertensor-phema-img512-cfgG-dec-64x64_block0-skip}};}
\gdef\figPerTensorplotCE{\addplot[C2,opacity=0.35,forget plot,dash phase=37.2pt] coordinates {\rawdata{pertensor-phema-img512-cfgG-dec-64x64_block3-skip}};}
\gdef\figPerTensorplotCF{\addplot[C2!80!black,thick,forget plot] coordinates {\rawdata{pertensor-phema-img512-cfgG}};}
\gdef\figPerTensorPlotCommands{\figPerTensorplotAA\figPerTensorplotAB\figPerTensorplotAC\figPerTensorplotAD\figPerTensorplotAE\figPerTensorplotAF\figPerTensorplotAG\figPerTensorplotAH\figPerTensorplotAI\figPerTensorplotAJ\figPerTensorplotAK\figPerTensorplotAL\figPerTensorplotAM\figPerTensorplotAN\figPerTensorplotAO\figPerTensorplotAP\figPerTensorplotAQ\figPerTensorplotAR\figPerTensorplotAS\figPerTensorplotAT\figPerTensorplotAU\figPerTensorplotAV\figPerTensorplotAW\figPerTensorplotAX\figPerTensorplotAY\figPerTensorplotAZ\figPerTensorplotBA\figPerTensorplotBB\figPerTensorplotBC\figPerTensorplotBD\figPerTensorplotBE\figPerTensorplotBF\figPerTensorplotBG\figPerTensorplotBH\figPerTensorplotBI\figPerTensorplotBJ\figPerTensorplotBK\figPerTensorplotBL\figPerTensorplotBM\figPerTensorplotBN\figPerTensorplotBO\figPerTensorplotBP\figPerTensorplotBQ\figPerTensorplotBR\figPerTensorplotBS\figPerTensorplotBT\figPerTensorplotBU\figPerTensorplotBV\figPerTensorplotBW\figPerTensorplotBX\figPerTensorplotBY\figPerTensorplotBZ\figPerTensorplotCA\figPerTensorplotCB\figPerTensorplotCC\figPerTensorplotCD\figPerTensorplotCE\figPerTensorplotCF}

\defdata{pertensor-phema-img512-cfgB-enc-64x64_conv}{%
(0.010, 11.8501)
(0.020, 10.2627)
(0.030, 9.4511)
(0.040, 8.8048)
(0.050, 8.2235)
(0.060, 7.9162)
(0.070, 7.7710)
(0.080, 7.4494)
(0.090, 7.4399)
(0.100, 7.5308)
(0.110, 7.4097)
(0.120, 8.0284)
(0.130, 8.4141)
(0.140, 8.8041)
(0.150, 9.5372)
(0.160, 10.4944)
(0.170, 11.4645)
(0.180, 13.0326)
(0.190, 14.8183)
(0.200, 16.7905)
(0.210, 19.7367)
(0.220, 23.4708)
(0.230, 28.2429)
(0.240, 34.2516)
(0.250, 41.7487)
}
\defdata{pertensor-phema-img512-cfgB-enc-64x64_block0-conv0}{%
(0.010, 9.9345)
(0.020, 9.1860)
(0.030, 8.8014)
(0.040, 8.2632)
(0.050, 7.8114)
(0.060, 7.7693)
(0.070, 7.5705)
(0.080, 7.4077)
(0.090, 7.4289)
(0.100, 7.4343)
(0.110, 7.5436)
(0.120, 7.4994)
(0.130, 7.5851)
(0.140, 7.8593)
(0.150, 7.8487)
(0.160, 7.8906)
(0.170, 8.1760)
(0.180, 8.1033)
(0.190, 8.2293)
(0.200, 8.3633)
(0.210, 8.5951)
(0.220, 8.9664)
(0.230, 9.3206)
(0.240, 9.3547)
(0.250, 9.8497)
}
\defdata{pertensor-phema-img512-cfgB-enc-64x64_block0-conv1}{%
(0.010, 8.1403)
(0.020, 7.9464)
(0.030, 8.0218)
(0.040, 7.9523)
(0.050, 7.7110)
(0.060, 7.5361)
(0.070, 7.6177)
(0.080, 7.3427)
(0.090, 7.4801)
(0.100, 7.3197)
(0.110, 7.2539)
(0.120, 7.0459)
(0.130, 6.9871)
(0.140, 6.9056)
(0.150, 6.9815)
(0.160, 6.8639)
(0.170, 6.8320)
(0.180, 6.7000)
(0.190, 6.7610)
(0.200, 6.6495)
(0.210, 6.5913)
(0.220, 6.6179)
(0.230, 6.5943)
(0.240, 6.6406)
(0.250, 6.8228)
}
\defdata{pertensor-phema-img512-cfgB-enc-32x32_block0-conv0}{%
(0.010, 7.3681)
(0.020, 7.3713)
(0.030, 7.4268)
(0.040, 7.3532)
(0.050, 7.4264)
(0.060, 7.4431)
(0.070, 7.3086)
(0.080, 7.3819)
(0.090, 7.4377)
(0.100, 7.3011)
(0.110, 7.4983)
(0.120, 7.3194)
(0.130, 7.2948)
(0.140, 7.4102)
(0.150, 7.4068)
(0.160, 7.3376)
(0.170, 7.5389)
(0.180, 7.4218)
(0.190, 7.3994)
(0.200, 7.4605)
(0.210, 7.4217)
(0.220, 7.4860)
(0.230, 7.4128)
(0.240, 7.6267)
(0.250, 7.6479)
}
\defdata{pertensor-phema-img512-cfgB-enc-16x16_block0-conv0}{%
(0.010, 7.4646)
(0.020, 7.5025)
(0.030, 7.3967)
(0.040, 7.3458)
(0.050, 7.4532)
(0.060, 7.4051)
(0.070, 7.3654)
(0.080, 7.4480)
(0.090, 7.4422)
(0.100, 7.4468)
(0.110, 7.3465)
(0.120, 7.3217)
(0.130, 7.3686)
(0.140, 7.5019)
(0.150, 7.4421)
(0.160, 7.4756)
(0.170, 7.4923)
(0.180, 7.5782)
(0.190, 7.5750)
(0.200, 7.5129)
(0.210, 7.3614)
(0.220, 7.3475)
(0.230, 7.3374)
(0.240, 7.5503)
(0.250, 7.4485)
}
\defdata{pertensor-phema-img512-cfgB-enc-16x16_block2-conv0}{%
(0.010, 7.3670)
(0.020, 7.4495)
(0.030, 7.5129)
(0.040, 7.3778)
(0.050, 7.3323)
(0.060, 7.4096)
(0.070, 7.3851)
(0.080, 7.3061)
(0.090, 7.4361)
(0.100, 7.4416)
(0.110, 7.5044)
(0.120, 7.3603)
(0.130, 7.6039)
(0.140, 7.4670)
(0.150, 7.4055)
(0.160, 7.3419)
(0.170, 7.3390)
(0.180, 7.4873)
(0.190, 7.3717)
(0.200, 7.5943)
(0.210, 7.4759)
(0.220, 7.4313)
(0.230, 7.5060)
(0.240, 7.5219)
(0.250, 7.4943)
}
\defdata{pertensor-phema-img512-cfgB-enc-8x8_block1-conv0}{%
(0.010, 7.5019)
(0.020, 7.4194)
(0.030, 7.4286)
(0.040, 7.4303)
(0.050, 7.4940)
(0.060, 7.4637)
(0.070, 7.3291)
(0.080, 7.4578)
(0.090, 7.3369)
(0.100, 7.4170)
(0.110, 7.4772)
(0.120, 7.4861)
(0.130, 7.3991)
(0.140, 7.3197)
(0.150, 7.1744)
(0.160, 7.3790)
(0.170, 7.3480)
(0.180, 7.4391)
(0.190, 7.4662)
(0.200, 7.1336)
(0.210, 7.3488)
(0.220, 7.2857)
(0.230, 7.2724)
(0.240, 7.3836)
(0.250, 7.3911)
}
\defdata{pertensor-phema-img512-cfgB-enc-8x8_block2-conv0}{%
(0.010, 7.4949)
(0.020, 7.2882)
(0.030, 7.4051)
(0.040, 7.4831)
(0.050, 7.4813)
(0.060, 7.3812)
(0.070, 7.3472)
(0.080, 7.4795)
(0.090, 7.5630)
(0.100, 7.4073)
(0.110, 7.4892)
(0.120, 7.4614)
(0.130, 7.4541)
(0.140, 7.5033)
(0.150, 7.3774)
(0.160, 7.4651)
(0.170, 7.3958)
(0.180, 7.3647)
(0.190, 7.5167)
(0.200, 7.3799)
(0.210, 7.3799)
(0.220, 7.3395)
(0.230, 7.4917)
(0.240, 7.3363)
(0.250, 7.2472)
}
\defdata{pertensor-phema-img512-cfgB-dec-8x8_block0-conv0}{%
(0.010, 7.3654)
(0.020, 7.4522)
(0.030, 7.3396)
(0.040, 7.4251)
(0.050, 7.5020)
(0.060, 7.3468)
(0.070, 7.2929)
(0.080, 7.3773)
(0.090, 7.3710)
(0.100, 7.3452)
(0.110, 7.5639)
(0.120, 7.5379)
(0.130, 7.3549)
(0.140, 7.3720)
(0.150, 7.3760)
(0.160, 7.5531)
(0.170, 7.4132)
(0.180, 7.2680)
(0.190, 7.3906)
(0.200, 7.3561)
(0.210, 7.2738)
(0.220, 7.2819)
(0.230, 7.1351)
(0.240, 7.2072)
(0.250, 7.1818)
}
\defdata{pertensor-phema-img512-cfgB-dec-16x16_block0-conv1}{%
(0.010, 6.4561)
(0.020, 6.5854)
(0.030, 6.6973)
(0.040, 6.7915)
(0.050, 6.8865)
(0.060, 7.1328)
(0.070, 7.2590)
(0.080, 7.1789)
(0.090, 7.4786)
(0.100, 7.6209)
(0.110, 7.7046)
(0.120, 7.8299)
(0.130, 8.0770)
(0.140, 8.2715)
(0.150, 8.3571)
(0.160, 8.7092)
(0.170, 8.7844)
(0.180, 8.9987)
(0.190, 9.4131)
(0.200, 9.7181)
(0.210, 10.2273)
(0.220, 10.5723)
(0.230, 11.0403)
(0.240, 11.4112)
(0.250, 12.2167)
}
\defdata{pertensor-phema-img512-cfgB-dec-32x32_block0-conv1}{%
(0.010, 7.1430)
(0.020, 6.8211)
(0.030, 6.8983)
(0.040, 7.0887)
(0.050, 7.0500)
(0.060, 7.2800)
(0.070, 7.3409)
(0.080, 7.2962)
(0.090, 7.4861)
(0.100, 7.5843)
(0.110, 7.6114)
(0.120, 7.6835)
(0.130, 7.7339)
(0.140, 8.0599)
(0.150, 8.0118)
(0.160, 7.9308)
(0.170, 8.1797)
(0.180, 8.4335)
(0.190, 8.4053)
(0.200, 8.7432)
(0.210, 8.9446)
(0.220, 9.3860)
(0.230, 9.4216)
(0.240, 9.6480)
(0.250, 9.8917)
}
\defdata{pertensor-phema-img512-cfgB-dec-64x64_block0-conv1}{%
(0.010, 7.0094)
(0.020, 6.9493)
(0.030, 7.0763)
(0.040, 7.2075)
(0.050, 7.1909)
(0.060, 7.1748)
(0.070, 7.3049)
(0.080, 7.4132)
(0.090, 7.3597)
(0.100, 7.5267)
(0.110, 7.6574)
(0.120, 7.6500)
(0.130, 7.7034)
(0.140, 7.8199)
(0.150, 7.7929)
(0.160, 7.9952)
(0.170, 8.0724)
(0.180, 8.2963)
(0.190, 8.2929)
(0.200, 8.5423)
(0.210, 8.5436)
(0.220, 8.8286)
(0.230, 8.7911)
(0.240, 9.2941)
(0.250, 9.4370)
}
\defdata{pertensor-phema-img512-cfgB-map_layer0}{%
(0.010, 8.1628)
(0.020, 7.8312)
(0.030, 7.5940)
(0.040, 7.6721)
(0.050, 7.6003)
(0.060, 7.5304)
(0.070, 7.4879)
(0.080, 7.4913)
(0.090, 7.3521)
(0.100, 7.4939)
(0.110, 7.4004)
(0.120, 7.4493)
(0.130, 7.6150)
(0.140, 7.5227)
(0.150, 7.5739)
(0.160, 7.4591)
(0.170, 7.6504)
(0.180, 7.8646)
(0.190, 7.8749)
(0.200, 8.1117)
(0.210, 8.4207)
(0.220, 8.8504)
(0.230, 9.5076)
(0.240, 10.0592)
(0.250, 11.1456)
}
\defdata{pertensor-phema-img512-cfgB-map_label}{%
(0.010, 9.1427)
(0.020, 8.8603)
(0.030, 8.7492)
(0.040, 8.3207)
(0.050, 8.2598)
(0.060, 7.9491)
(0.070, 7.7539)
(0.080, 7.7132)
(0.090, 7.4601)
(0.100, 7.2556)
(0.110, 7.2066)
(0.120, 6.9146)
(0.130, 6.8220)
(0.140, 6.7653)
(0.150, 6.7495)
(0.160, 6.6223)
(0.170, 6.7468)
(0.180, 6.7568)
(0.190, 6.7622)
(0.200, 7.0346)
(0.210, 7.1912)
(0.220, 7.7456)
(0.230, 8.1736)
(0.240, 8.8510)
(0.250, 10.1511)
}
\defdata{pertensor-phema-img512-cfgB-enc-64x64_block0-affine}{%
(0.010, 8.0239)
(0.020, 7.7927)
(0.030, 7.9100)
(0.040, 7.8605)
(0.050, 7.7076)
(0.060, 7.6277)
(0.070, 7.6169)
(0.080, 7.3701)
(0.090, 7.4336)
(0.100, 7.3394)
(0.110, 7.2685)
(0.120, 7.2399)
(0.130, 7.0404)
(0.140, 7.1312)
(0.150, 7.0225)
(0.160, 6.9522)
(0.170, 7.1822)
(0.180, 7.1017)
(0.190, 7.1665)
(0.200, 7.2593)
(0.210, 7.1991)
(0.220, 7.3905)
(0.230, 7.5706)
(0.240, 7.7832)
(0.250, 8.1154)
}
\defdata{pertensor-phema-img512-cfgB-enc-16x16_block0-affine}{%
(0.010, 7.4829)
(0.020, 7.4643)
(0.030, 7.3123)
(0.040, 7.3806)
(0.050, 7.4159)
(0.060, 7.4680)
(0.070, 7.1427)
(0.080, 7.2427)
(0.090, 7.3272)
(0.100, 7.4308)
(0.110, 7.4600)
(0.120, 7.5656)
(0.130, 7.3838)
(0.140, 7.2759)
(0.150, 7.4607)
(0.160, 7.4192)
(0.170, 7.3974)
(0.180, 7.3765)
(0.190, 7.4016)
(0.200, 7.4594)
(0.210, 7.5649)
(0.220, 7.5849)
(0.230, 7.7314)
(0.240, 7.6519)
(0.250, 7.7694)
}
\defdata{pertensor-phema-img512-cfgB-enc-8x8_block0-affine}{%
(0.010, 7.4781)
(0.020, 7.3605)
(0.030, 7.3596)
(0.040, 7.4388)
(0.050, 7.3365)
(0.060, 7.2751)
(0.070, 7.4154)
(0.080, 7.1932)
(0.090, 7.5553)
(0.100, 7.3296)
(0.110, 7.5884)
(0.120, 7.3497)
(0.130, 7.4495)
(0.140, 7.5860)
(0.150, 7.6564)
(0.160, 7.4803)
(0.170, 7.5804)
(0.180, 7.7854)
(0.190, 7.7281)
(0.200, 7.8240)
(0.210, 7.9075)
(0.220, 7.9403)
(0.230, 7.8053)
(0.240, 8.0558)
(0.250, 8.2708)
}
\defdata{pertensor-phema-img512-cfgB-dec-8x8_in0-affine}{%
(0.010, 7.2784)
(0.020, 7.3344)
(0.030, 7.4224)
(0.040, 7.3520)
(0.050, 7.5062)
(0.060, 7.4014)
(0.070, 7.4063)
(0.080, 7.5027)
(0.090, 7.3558)
(0.100, 7.4999)
(0.110, 7.4595)
(0.120, 7.5785)
(0.130, 7.3661)
(0.140, 7.6771)
(0.150, 7.3374)
(0.160, 7.4736)
(0.170, 7.5755)
(0.180, 7.5964)
(0.190, 7.6494)
(0.200, 7.7333)
(0.210, 7.7317)
(0.220, 7.7599)
(0.230, 7.7645)
(0.240, 7.7217)
(0.250, 7.7876)
}
\defdata{pertensor-phema-img512-cfgB-dec-16x16_block0-affine}{%
(0.010, 6.9733)
(0.020, 7.0886)
(0.030, 7.0909)
(0.040, 7.2407)
(0.050, 7.1737)
(0.060, 7.1189)
(0.070, 7.2210)
(0.080, 7.3483)
(0.090, 7.3744)
(0.100, 7.4534)
(0.110, 7.5848)
(0.120, 7.5869)
(0.130, 7.7514)
(0.140, 7.8179)
(0.150, 7.8627)
(0.160, 7.9449)
(0.170, 8.0546)
(0.180, 8.1598)
(0.190, 8.5034)
(0.200, 8.5811)
(0.210, 8.6492)
(0.220, 9.0520)
(0.230, 9.1451)
(0.240, 9.4945)
(0.250, 9.7304)
}
\defdata{pertensor-phema-img512-cfgB-dec-32x32_up-affine}{%
(0.010, 7.2447)
(0.020, 7.2314)
(0.030, 7.1884)
(0.040, 7.3445)
(0.050, 7.2710)
(0.060, 7.1369)
(0.070, 7.3602)
(0.080, 7.3450)
(0.090, 7.5082)
(0.100, 7.4246)
(0.110, 7.5330)
(0.120, 7.5050)
(0.130, 7.6661)
(0.140, 7.6014)
(0.150, 7.5913)
(0.160, 7.7439)
(0.170, 7.7938)
(0.180, 7.8390)
(0.190, 7.8678)
(0.200, 7.9540)
(0.210, 8.0600)
(0.220, 8.1170)
(0.230, 8.2894)
(0.240, 8.2934)
(0.250, 8.4824)
}
\defdata{pertensor-phema-img512-cfgB-dec-64x64_up-affine}{%
(0.010, 7.1026)
(0.020, 7.1443)
(0.030, 7.1998)
(0.040, 7.2374)
(0.050, 7.2300)
(0.060, 7.2740)
(0.070, 7.3880)
(0.080, 7.3564)
(0.090, 7.4063)
(0.100, 7.4567)
(0.110, 7.5264)
(0.120, 7.5774)
(0.130, 7.5932)
(0.140, 7.6950)
(0.150, 7.8984)
(0.160, 7.8678)
(0.170, 7.8986)
(0.180, 7.9972)
(0.190, 8.2252)
(0.200, 8.2258)
(0.210, 8.4986)
(0.220, 8.4935)
(0.230, 8.8155)
(0.240, 8.8161)
(0.250, 9.0275)
}
\defdata{pertensor-phema-img512-cfgB-enc-32x32_block0-skip}{%
(0.010, 7.5948)
(0.020, 7.4931)
(0.030, 7.4099)
(0.040, 7.4808)
(0.050, 7.4744)
(0.060, 7.3466)
(0.070, 7.3475)
(0.080, 7.4170)
(0.090, 7.3499)
(0.100, 7.4207)
(0.110, 7.4753)
(0.120, 7.4869)
(0.130, 7.5202)
(0.140, 7.6098)
(0.150, 7.3843)
(0.160, 7.5117)
(0.170, 7.4040)
(0.180, 7.5443)
(0.190, 7.5222)
(0.200, 7.6290)
(0.210, 7.5823)
(0.220, 7.6449)
(0.230, 7.6813)
(0.240, 7.8207)
(0.250, 7.9052)
}
\defdata{pertensor-phema-img512-cfgB-enc-8x8_block0-skip}{%
(0.010, 7.3735)
(0.020, 7.4207)
(0.030, 7.3910)
(0.040, 7.4468)
(0.050, 7.3064)
(0.060, 7.3320)
(0.070, 7.4867)
(0.080, 7.4385)
(0.090, 7.6012)
(0.100, 7.5424)
(0.110, 7.3520)
(0.120, 7.4920)
(0.130, 7.1600)
(0.140, 7.3122)
(0.150, 7.5250)
(0.160, 7.4632)
(0.170, 7.4149)
(0.180, 7.3189)
(0.190, 7.3399)
(0.200, 7.3413)
(0.210, 7.5267)
(0.220, 7.4307)
(0.230, 7.3783)
(0.240, 7.4388)
(0.250, 7.4832)
}
\defdata{pertensor-phema-img512-cfgB-dec-8x8_block2-skip}{%
(0.010, 7.6221)
(0.020, 7.3885)
(0.030, 7.2622)
(0.040, 7.4218)
(0.050, 7.4551)
(0.060, 7.4664)
(0.070, 7.6529)
(0.080, 7.3283)
(0.090, 7.3534)
(0.100, 7.4061)
(0.110, 7.2745)
(0.120, 7.3460)
(0.130, 7.4141)
(0.140, 7.3283)
(0.150, 7.1422)
(0.160, 7.0960)
(0.170, 7.3423)
(0.180, 7.4226)
(0.190, 7.2099)
(0.200, 7.1351)
(0.210, 7.1748)
(0.220, 7.1719)
(0.230, 7.1167)
(0.240, 7.0720)
(0.250, 7.1272)
}
\defdata{pertensor-phema-img512-cfgB-dec-16x16_block0-skip}{%
(0.010, 7.5498)
(0.020, 7.5044)
(0.030, 7.4741)
(0.040, 7.3722)
(0.050, 7.4164)
(0.060, 7.3175)
(0.070, 7.4034)
(0.080, 7.4747)
(0.090, 7.1751)
(0.100, 7.2427)
(0.110, 7.5634)
(0.120, 7.4445)
(0.130, 7.4270)
(0.140, 7.4488)
(0.150, 7.4660)
(0.160, 7.6081)
(0.170, 7.4136)
(0.180, 7.6090)
(0.190, 7.5578)
(0.200, 7.7689)
(0.210, 7.4829)
(0.220, 7.5674)
(0.230, 7.7179)
(0.240, 7.8342)
(0.250, 8.1049)
}
\defdata{pertensor-phema-img512-cfgB-dec-32x32_block0-skip}{%
(0.010, 7.1620)
(0.020, 7.0159)
(0.030, 7.0320)
(0.040, 7.0481)
(0.050, 7.2048)
(0.060, 7.2340)
(0.070, 7.3083)
(0.080, 7.3154)
(0.090, 7.4093)
(0.100, 7.4783)
(0.110, 7.4962)
(0.120, 7.7106)
(0.130, 7.8261)
(0.140, 7.7395)
(0.150, 8.0502)
(0.160, 8.1234)
(0.170, 8.2986)
(0.180, 8.3060)
(0.190, 8.5979)
(0.200, 8.7913)
(0.210, 9.0620)
(0.220, 9.5421)
(0.230, 9.9205)
(0.240, 10.4046)
(0.250, 11.0694)
}
\defdata{pertensor-phema-img512-cfgB-dec-64x64_block0-skip}{%
(0.010, 6.8645)
(0.020, 6.9885)
(0.030, 6.8897)
(0.040, 6.9791)
(0.050, 6.9818)
(0.060, 7.0422)
(0.070, 7.1298)
(0.080, 7.3187)
(0.090, 7.5178)
(0.100, 7.5530)
(0.110, 7.7087)
(0.120, 7.8390)
(0.130, 8.0404)
(0.140, 8.3942)
(0.150, 8.6009)
(0.160, 9.0264)
(0.170, 9.2755)
(0.180, 9.5972)
(0.190, 10.1765)
(0.200, 10.7890)
(0.210, 11.3985)
(0.220, 12.3359)
(0.230, 13.2364)
(0.240, 15.0070)
(0.250, 16.6514)
}
\defdata{pertensor-phema-img512-cfgB-dec-64x64_block3-skip}{%
(0.010, 9.2161)
(0.020, 9.0632)
(0.030, 8.7277)
(0.040, 8.2403)
(0.050, 7.9439)
(0.060, 7.8790)
(0.070, 7.7142)
(0.080, 7.7153)
(0.090, 7.3506)
(0.100, 7.4284)
(0.110, 7.1740)
(0.120, 7.2196)
(0.130, 7.3783)
(0.140, 7.3063)
(0.150, 7.3678)
(0.160, 7.4951)
(0.170, 7.5382)
(0.180, 7.8933)
(0.190, 7.9794)
(0.200, 8.6048)
(0.210, 8.9590)
(0.220, 9.8497)
(0.230, 10.6974)
(0.240, 12.0539)
(0.250, 14.0519)
}
\defdata{pertensor-phema-img512-cfgB}{%
(0.010, 15.7549)
(0.015, 9.1255)
(0.020, 8.0786)
(0.025, 7.7800)
(0.030, 7.6317)
(0.035, 7.6603)
(0.040, 7.5384)
(0.045, 7.5960)
(0.050, 7.5236)
(0.055, 7.4953)
(0.060, 7.4777)
(0.065, 7.4014)
(0.070, 7.4147)
(0.075, 7.3727)
(0.080, 7.3848)
(0.085, 7.4045)
(0.090, 7.2439)
(0.095, 7.3540)
(0.100, 7.3445)
(0.105, 7.3511)
(0.110, 7.2606)
(0.115, 7.3835)
(0.120, 7.4288)
(0.125, 7.4144)
(0.130, 7.5757)
(0.135, 7.6211)
(0.140, 7.6296)
(0.145, 7.8221)
(0.150, 7.8697)
(0.155, 7.8452)
(0.160, 7.9838)
(0.165, 8.3056)
(0.170, 8.4079)
(0.175, 8.7885)
(0.180, 8.7724)
(0.185, 9.1096)
(0.190, 9.4881)
(0.195, 9.7559)
(0.200, 10.0344)
}
\defdata{pertensor-phema-img512-cfgG-enc-64x64_conv}{%
(0.010, 3.5389)
(0.020, 3.4221)
(0.030, 3.3635)
(0.040, 3.1818)
(0.050, 3.1021)
(0.060, 3.0791)
(0.070, 2.9740)
(0.080, 2.8524)
(0.090, 2.6992)
(0.100, 2.7279)
(0.110, 2.6548)
(0.120, 2.6154)
(0.130, 2.6012)
(0.140, 2.5895)
(0.150, 2.5762)
(0.160, 2.6209)
(0.170, 2.6414)
(0.180, 2.7186)
(0.190, 2.9758)
(0.200, 3.0859)
(0.210, 3.4890)
(0.220, 3.9629)
(0.230, 4.6360)
(0.240, 5.7653)
(0.250, 7.4953)
}
\defdata{pertensor-phema-img512-cfgG-enc-64x64_block0-conv0}{%
(0.010, 2.5876)
(0.020, 2.5590)
(0.030, 2.6218)
(0.040, 2.5726)
(0.050, 2.5863)
(0.060, 2.5025)
(0.070, 2.5631)
(0.080, 2.5933)
(0.090, 2.6351)
(0.100, 2.5488)
(0.110, 2.5970)
(0.120, 2.6098)
(0.130, 2.5232)
(0.140, 2.6023)
(0.150, 2.5986)
(0.160, 2.6540)
(0.170, 2.5845)
(0.180, 2.5964)
(0.190, 2.6452)
(0.200, 2.5505)
(0.210, 2.6100)
(0.220, 2.6432)
(0.230, 2.6530)
(0.240, 2.6604)
(0.250, 2.6466)
}
\defdata{pertensor-phema-img512-cfgG-enc-64x64_block0-conv1}{%
(0.010, 2.7971)
(0.020, 2.6993)
(0.030, 2.6736)
(0.040, 2.7090)
(0.050, 2.6467)
(0.060, 2.6731)
(0.070, 2.5991)
(0.080, 2.6294)
(0.090, 2.5467)
(0.100, 2.5277)
(0.110, 2.5834)
(0.120, 2.6102)
(0.130, 2.5492)
(0.140, 2.5853)
(0.150, 2.6954)
(0.160, 2.6642)
(0.170, 2.7760)
(0.180, 2.8732)
(0.190, 3.0473)
(0.200, 3.2796)
(0.210, 3.5492)
(0.220, 3.8104)
(0.230, 4.2641)
(0.240, 4.7903)
(0.250, 5.5432)
}
\defdata{pertensor-phema-img512-cfgG-enc-32x32_block0-conv0}{%
(0.010, 2.5118)
(0.020, 2.5998)
(0.030, 2.6017)
(0.040, 2.5586)
(0.050, 2.5685)
(0.060, 2.5982)
(0.070, 2.5418)
(0.080, 2.6173)
(0.090, 2.5560)
(0.100, 2.5437)
(0.110, 2.5804)
(0.120, 2.5587)
(0.130, 2.5679)
(0.140, 2.5934)
(0.150, 2.5581)
(0.160, 2.6219)
(0.170, 2.6008)
(0.180, 2.5569)
(0.190, 2.5894)
(0.200, 2.6053)
(0.210, 2.6206)
(0.220, 2.5120)
(0.230, 2.5667)
(0.240, 2.5729)
(0.250, 2.6515)
}
\defdata{pertensor-phema-img512-cfgG-enc-16x16_block0-conv0}{%
(0.010, 2.5980)
(0.020, 2.6102)
(0.030, 2.5971)
(0.040, 2.5846)
(0.050, 2.6149)
(0.060, 2.5637)
(0.070, 2.5576)
(0.080, 2.5639)
(0.090, 2.5034)
(0.100, 2.5540)
(0.110, 2.6181)
(0.120, 2.5823)
(0.130, 2.5283)
(0.140, 2.6041)
(0.150, 2.5656)
(0.160, 2.5678)
(0.170, 2.5318)
(0.180, 2.6163)
(0.190, 2.6117)
(0.200, 2.5537)
(0.210, 2.5080)
(0.220, 2.5938)
(0.230, 2.4597)
(0.240, 2.5505)
(0.250, 2.5241)
}
\defdata{pertensor-phema-img512-cfgG-enc-16x16_block2-conv0}{%
(0.010, 2.5585)
(0.020, 2.5981)
(0.030, 2.5564)
(0.040, 2.5651)
(0.050, 2.5869)
(0.060, 2.5604)
(0.070, 2.5302)
(0.080, 2.5517)
(0.090, 2.5951)
(0.100, 2.5683)
(0.110, 2.5805)
(0.120, 2.5965)
(0.130, 2.5675)
(0.140, 2.6301)
(0.150, 2.5399)
(0.160, 2.5831)
(0.170, 2.5283)
(0.180, 2.6179)
(0.190, 2.5571)
(0.200, 2.5971)
(0.210, 2.5968)
(0.220, 2.5846)
(0.230, 2.5845)
(0.240, 2.6288)
(0.250, 2.5930)
}
\defdata{pertensor-phema-img512-cfgG-enc-8x8_block1-conv0}{%
(0.010, 2.5472)
(0.020, 2.5622)
(0.030, 2.5948)
(0.040, 2.5859)
(0.050, 2.5648)
(0.060, 2.5676)
(0.070, 2.5650)
(0.080, 2.6015)
(0.090, 2.6171)
(0.100, 2.6079)
(0.110, 2.5459)
(0.120, 2.6591)
(0.130, 2.5524)
(0.140, 2.5940)
(0.150, 2.5649)
(0.160, 2.5950)
(0.170, 2.5888)
(0.180, 2.6427)
(0.190, 2.5523)
(0.200, 2.5793)
(0.210, 2.5882)
(0.220, 2.5414)
(0.230, 2.5623)
(0.240, 2.6257)
(0.250, 2.5522)
}
\defdata{pertensor-phema-img512-cfgG-enc-8x8_block2-conv0}{%
(0.010, 2.6025)
(0.020, 2.5376)
(0.030, 2.6081)
(0.040, 2.5756)
(0.050, 2.5651)
(0.060, 2.5458)
(0.070, 2.5893)
(0.080, 2.5539)
(0.090, 2.5652)
(0.100, 2.5864)
(0.110, 2.5924)
(0.120, 2.5929)
(0.130, 2.5965)
(0.140, 2.5777)
(0.150, 2.5748)
(0.160, 2.5622)
(0.170, 2.5973)
(0.180, 2.5985)
(0.190, 2.6529)
(0.200, 2.6177)
(0.210, 2.5835)
(0.220, 2.6363)
(0.230, 2.5835)
(0.240, 2.5711)
(0.250, 2.5828)
}
\defdata{pertensor-phema-img512-cfgG-dec-8x8_block0-conv0}{%
(0.010, 2.5810)
(0.020, 2.6372)
(0.030, 2.5969)
(0.040, 2.5506)
(0.050, 2.5729)
(0.060, 2.5831)
(0.070, 2.6083)
(0.080, 2.6018)
(0.090, 2.5806)
(0.100, 2.5165)
(0.110, 2.6075)
(0.120, 2.6068)
(0.130, 2.6235)
(0.140, 2.5886)
(0.150, 2.6058)
(0.160, 2.5694)
(0.170, 2.5845)
(0.180, 2.6141)
(0.190, 2.5341)
(0.200, 2.5979)
(0.210, 2.6364)
(0.220, 2.6197)
(0.230, 2.6750)
(0.240, 2.6544)
(0.250, 2.6515)
}
\defdata{pertensor-phema-img512-cfgG-dec-16x16_block0-conv1}{%
(0.010, 2.5429)
(0.020, 2.5796)
(0.030, 2.5755)
(0.040, 2.6221)
(0.050, 2.5839)
(0.060, 2.5794)
(0.070, 2.5784)
(0.080, 2.6525)
(0.090, 2.5579)
(0.100, 2.5190)
(0.110, 2.6193)
(0.120, 2.5832)
(0.130, 2.6003)
(0.140, 2.6713)
(0.150, 2.6123)
(0.160, 2.5361)
(0.170, 2.5765)
(0.180, 2.5727)
(0.190, 2.6250)
(0.200, 2.6264)
(0.210, 2.5782)
(0.220, 2.6114)
(0.230, 2.6423)
(0.240, 2.5747)
(0.250, 2.6297)
}
\defdata{pertensor-phema-img512-cfgG-dec-32x32_block0-conv1}{%
(0.010, 2.5322)
(0.020, 2.6013)
(0.030, 2.5357)
(0.040, 2.5532)
(0.050, 2.5846)
(0.060, 2.5441)
(0.070, 2.5782)
(0.080, 2.6539)
(0.090, 2.6056)
(0.100, 2.5765)
(0.110, 2.5468)
(0.120, 2.6073)
(0.130, 2.5511)
(0.140, 2.6037)
(0.150, 2.6007)
(0.160, 2.5720)
(0.170, 2.5712)
(0.180, 2.5842)
(0.190, 2.6205)
(0.200, 2.6057)
(0.210, 2.6065)
(0.220, 2.6578)
(0.230, 2.6203)
(0.240, 2.6607)
(0.250, 2.6578)
}
\defdata{pertensor-phema-img512-cfgG-dec-64x64_block0-conv1}{%
(0.010, 2.5425)
(0.020, 2.5159)
(0.030, 2.6141)
(0.040, 2.5464)
(0.050, 2.5925)
(0.060, 2.6686)
(0.070, 2.5860)
(0.080, 2.5945)
(0.090, 2.5937)
(0.100, 2.5995)
(0.110, 2.5550)
(0.120, 2.6074)
(0.130, 2.6401)
(0.140, 2.5190)
(0.150, 2.6359)
(0.160, 2.5815)
(0.170, 2.6353)
(0.180, 2.5675)
(0.190, 2.5427)
(0.200, 2.5639)
(0.210, 2.5905)
(0.220, 2.5489)
(0.230, 2.5801)
(0.240, 2.6320)
(0.250, 2.6295)
}
\defdata{pertensor-phema-img512-cfgG-map_layer0}{%
(0.010, 2.9831)
(0.020, 2.8710)
(0.030, 2.8431)
(0.040, 2.7168)
(0.050, 2.6965)
(0.060, 2.6586)
(0.070, 2.5585)
(0.080, 2.4996)
(0.090, 2.5012)
(0.100, 2.5312)
(0.110, 2.5679)
(0.120, 2.6055)
(0.130, 2.6191)
(0.140, 2.6873)
(0.150, 2.7267)
(0.160, 2.8151)
(0.170, 3.1045)
(0.180, 3.3297)
(0.190, 3.6774)
(0.200, 4.1778)
(0.210, 4.9446)
(0.220, 5.9473)
(0.230, 7.6501)
(0.240, 9.9246)
(0.250, 13.5538)
}
\defdata{pertensor-phema-img512-cfgG-map_label}{%
(0.010, 2.5794)
(0.020, 2.5866)
(0.030, 2.5321)
(0.040, 2.5561)
(0.050, 2.5714)
(0.060, 2.5941)
(0.070, 2.5203)
(0.080, 2.6258)
(0.090, 2.6146)
(0.100, 2.6347)
(0.110, 2.5648)
(0.120, 2.5984)
(0.130, 2.5640)
(0.140, 2.5801)
(0.150, 2.6002)
(0.160, 2.5937)
(0.170, 2.5460)
(0.180, 2.5907)
(0.190, 2.6314)
(0.200, 2.6006)
(0.210, 2.6043)
(0.220, 2.7088)
(0.230, 2.6582)
(0.240, 2.6343)
(0.250, 2.7052)
}
\defdata{pertensor-phema-img512-cfgG-enc-64x64_block0-affine}{%
(0.010, 3.3815)
(0.020, 3.2196)
(0.030, 3.2494)
(0.040, 3.0724)
(0.050, 2.9645)
(0.060, 2.9137)
(0.070, 2.9247)
(0.080, 2.8035)
(0.090, 2.7224)
(0.100, 2.6494)
(0.110, 2.6463)
(0.120, 2.6675)
(0.130, 2.5434)
(0.140, 2.5710)
(0.150, 2.6143)
(0.160, 2.5759)
(0.170, 2.6584)
(0.180, 2.7290)
(0.190, 2.8157)
(0.200, 2.8998)
(0.210, 3.1682)
(0.220, 3.3791)
(0.230, 3.7589)
(0.240, 4.2803)
(0.250, 4.9961)
}
\defdata{pertensor-phema-img512-cfgG-enc-16x16_block0-affine}{%
(0.010, 2.6260)
(0.020, 2.5079)
(0.030, 2.5653)
(0.040, 2.5504)
(0.050, 2.5436)
(0.060, 2.5508)
(0.070, 2.6161)
(0.080, 2.6173)
(0.090, 2.6410)
(0.100, 2.5598)
(0.110, 2.5758)
(0.120, 2.5986)
(0.130, 2.5442)
(0.140, 2.5838)
(0.150, 2.5415)
(0.160, 2.5451)
(0.170, 2.6239)
(0.180, 2.5896)
(0.190, 2.5734)
(0.200, 2.5911)
(0.210, 2.5813)
(0.220, 2.5162)
(0.230, 2.5653)
(0.240, 2.5731)
(0.250, 2.5694)
}
\defdata{pertensor-phema-img512-cfgG-enc-8x8_block0-affine}{%
(0.010, 2.6408)
(0.020, 2.6639)
(0.030, 2.6279)
(0.040, 2.6128)
(0.050, 2.5764)
(0.060, 2.6604)
(0.070, 2.5993)
(0.080, 2.5876)
(0.090, 2.5919)
(0.100, 2.5858)
(0.110, 2.5909)
(0.120, 2.5402)
(0.130, 2.6245)
(0.140, 2.5070)
(0.150, 2.5661)
(0.160, 2.5889)
(0.170, 2.5112)
(0.180, 2.5644)
(0.190, 2.5447)
(0.200, 2.4844)
(0.210, 2.5371)
(0.220, 2.5538)
(0.230, 2.5464)
(0.240, 2.5378)
(0.250, 2.5154)
}
\defdata{pertensor-phema-img512-cfgG-dec-8x8_in0-affine}{%
(0.010, 2.5990)
(0.020, 2.5777)
(0.030, 2.5776)
(0.040, 2.6150)
(0.050, 2.5701)
(0.060, 2.5997)
(0.070, 2.5643)
(0.080, 2.5720)
(0.090, 2.5876)
(0.100, 2.5704)
(0.110, 2.5523)
(0.120, 2.5854)
(0.130, 2.6063)
(0.140, 2.5772)
(0.150, 2.5698)
(0.160, 2.6098)
(0.170, 2.5934)
(0.180, 2.6066)
(0.190, 2.6048)
(0.200, 2.5681)
(0.210, 2.5471)
(0.220, 2.5805)
(0.230, 2.6111)
(0.240, 2.6243)
(0.250, 2.6283)
}
\defdata{pertensor-phema-img512-cfgG-dec-16x16_block0-affine}{%
(0.010, 2.5928)
(0.020, 2.6298)
(0.030, 2.5818)
(0.040, 2.6016)
(0.050, 2.5876)
(0.060, 2.5811)
(0.070, 2.5124)
(0.080, 2.5992)
(0.090, 2.5766)
(0.100, 2.6164)
(0.110, 2.5765)
(0.120, 2.5174)
(0.130, 2.6020)
(0.140, 2.6010)
(0.150, 2.6060)
(0.160, 2.6423)
(0.170, 2.5755)
(0.180, 2.5643)
(0.190, 2.6277)
(0.200, 2.6179)
(0.210, 2.5953)
(0.220, 2.6077)
(0.230, 2.6357)
(0.240, 2.5976)
(0.250, 2.5838)
}
\defdata{pertensor-phema-img512-cfgG-dec-32x32_up-affine}{%
(0.010, 2.6349)
(0.020, 2.5292)
(0.030, 2.5062)
(0.040, 2.5591)
(0.050, 2.6182)
(0.060, 2.5531)
(0.070, 2.5692)
(0.080, 2.5567)
(0.090, 2.5740)
(0.100, 2.5891)
(0.110, 2.5685)
(0.120, 2.5043)
(0.130, 2.5973)
(0.140, 2.6029)
(0.150, 2.5811)
(0.160, 2.6332)
(0.170, 2.5897)
(0.180, 2.6603)
(0.190, 2.5712)
(0.200, 2.6409)
(0.210, 2.5846)
(0.220, 2.5555)
(0.230, 2.5730)
(0.240, 2.6152)
(0.250, 2.6294)
}
\defdata{pertensor-phema-img512-cfgG-dec-64x64_up-affine}{%
(0.010, 2.5607)
(0.020, 2.5501)
(0.030, 2.4972)
(0.040, 2.5805)
(0.050, 2.4953)
(0.060, 2.6192)
(0.070, 2.5694)
(0.080, 2.5840)
(0.090, 2.5395)
(0.100, 2.6270)
(0.110, 2.5535)
(0.120, 2.5425)
(0.130, 2.5966)
(0.140, 2.5828)
(0.150, 2.5788)
(0.160, 2.5479)
(0.170, 2.5080)
(0.180, 2.5601)
(0.190, 2.6338)
(0.200, 2.6071)
(0.210, 2.6115)
(0.220, 2.6643)
(0.230, 2.6374)
(0.240, 2.7160)
(0.250, 2.7518)
}
\defdata{pertensor-phema-img512-cfgG-enc-32x32_block0-skip}{%
(0.010, 2.8555)
(0.020, 2.8061)
(0.030, 2.7600)
(0.040, 2.7254)
(0.050, 2.7772)
(0.060, 2.6825)
(0.070, 2.6423)
(0.080, 2.6442)
(0.090, 2.6522)
(0.100, 2.5910)
(0.110, 2.5893)
(0.120, 2.5768)
(0.130, 2.5827)
(0.140, 2.5709)
(0.150, 2.5742)
(0.160, 2.6726)
(0.170, 2.6370)
(0.180, 2.6811)
(0.190, 2.6940)
(0.200, 2.7534)
(0.210, 2.8606)
(0.220, 2.8823)
(0.230, 3.0118)
(0.240, 3.1389)
(0.250, 3.3263)
}
\defdata{pertensor-phema-img512-cfgG-enc-8x8_block0-skip}{%
(0.010, 2.6015)
(0.020, 2.5855)
(0.030, 2.5682)
(0.040, 2.5456)
(0.050, 2.5823)
(0.060, 2.5150)
(0.070, 2.5974)
(0.080, 2.5808)
(0.090, 2.5855)
(0.100, 2.6489)
(0.110, 2.5900)
(0.120, 2.5806)
(0.130, 2.6003)
(0.140, 2.5937)
(0.150, 2.5192)
(0.160, 2.6214)
(0.170, 2.5462)
(0.180, 2.5980)
(0.190, 2.5815)
(0.200, 2.6794)
(0.210, 2.6465)
(0.220, 2.6276)
(0.230, 2.6555)
(0.240, 2.6077)
(0.250, 2.7845)
}
\defdata{pertensor-phema-img512-cfgG-dec-8x8_block2-skip}{%
(0.010, 2.5866)
(0.020, 2.5423)
(0.030, 2.6067)
(0.040, 2.6517)
(0.050, 2.4814)
(0.060, 2.5701)
(0.070, 2.5954)
(0.080, 2.5884)
(0.090, 2.5797)
(0.100, 2.5780)
(0.110, 2.5966)
(0.120, 2.5490)
(0.130, 2.5788)
(0.140, 2.5856)
(0.150, 2.5850)
(0.160, 2.6413)
(0.170, 2.5871)
(0.180, 2.5374)
(0.190, 2.5818)
(0.200, 2.6258)
(0.210, 2.5777)
(0.220, 2.6143)
(0.230, 2.6102)
(0.240, 2.6034)
(0.250, 2.6416)
}
\defdata{pertensor-phema-img512-cfgG-dec-16x16_block0-skip}{%
(0.010, 2.5855)
(0.020, 2.5639)
(0.030, 2.5720)
(0.040, 2.5980)
(0.050, 2.5320)
(0.060, 2.5697)
(0.070, 2.5596)
(0.080, 2.5770)
(0.090, 2.6007)
(0.100, 2.5862)
(0.110, 2.5271)
(0.120, 2.5996)
(0.130, 2.5705)
(0.140, 2.5780)
(0.150, 2.5997)
(0.160, 2.6115)
(0.170, 2.5879)
(0.180, 2.5208)
(0.190, 2.5684)
(0.200, 2.5540)
(0.210, 2.6074)
(0.220, 2.6220)
(0.230, 2.5718)
(0.240, 2.6798)
(0.250, 2.5519)
}
\defdata{pertensor-phema-img512-cfgG-dec-32x32_block0-skip}{%
(0.010, 2.6129)
(0.020, 2.6185)
(0.030, 2.6385)
(0.040, 2.6268)
(0.050, 2.6172)
(0.060, 2.6399)
(0.070, 2.5784)
(0.080, 2.6320)
(0.090, 2.5373)
(0.100, 2.6098)
(0.110, 2.5583)
(0.120, 2.6279)
(0.130, 2.5707)
(0.140, 2.6184)
(0.150, 2.5836)
(0.160, 2.5535)
(0.170, 2.6181)
(0.180, 2.5962)
(0.190, 2.5934)
(0.200, 2.6023)
(0.210, 2.6146)
(0.220, 2.5988)
(0.230, 2.6106)
(0.240, 2.6005)
(0.250, 2.6178)
}
\defdata{pertensor-phema-img512-cfgG-dec-64x64_block0-skip}{%
(0.010, 2.5682)
(0.020, 2.5802)
(0.030, 2.6442)
(0.040, 2.5831)
(0.050, 2.5648)
(0.060, 2.5879)
(0.070, 2.5302)
(0.080, 2.5741)
(0.090, 2.5847)
(0.100, 2.5872)
(0.110, 2.6648)
(0.120, 2.5248)
(0.130, 2.5837)
(0.140, 2.5446)
(0.150, 2.6293)
(0.160, 2.5835)
(0.170, 2.6535)
(0.180, 2.5736)
(0.190, 2.6501)
(0.200, 2.7209)
(0.210, 2.6909)
(0.220, 2.7015)
(0.230, 2.6879)
(0.240, 2.7558)
(0.250, 2.8379)
}
\defdata{pertensor-phema-img512-cfgG-dec-64x64_block3-skip}{%
(0.010, 2.6471)
(0.020, 2.6404)
(0.030, 2.5887)
(0.040, 2.5846)
(0.050, 2.6264)
(0.060, 2.5844)
(0.070, 2.5991)
(0.080, 2.5771)
(0.090, 2.5999)
(0.100, 2.5826)
(0.110, 2.5787)
(0.120, 2.5860)
(0.130, 2.5871)
(0.140, 2.5840)
(0.150, 2.5904)
(0.160, 2.5983)
(0.170, 2.6857)
(0.180, 2.6522)
(0.190, 2.6487)
(0.200, 2.6462)
(0.210, 2.7287)
(0.220, 2.7378)
(0.230, 2.8692)
(0.240, 2.9735)
(0.250, 2.9370)
}
\defdata{pertensor-phema-img512-cfgG}{%
(0.010, 11.8664)
(0.015, 8.6684)
(0.020, 7.3421)
(0.025, 6.7381)
(0.030, 6.3240)
(0.035, 5.9067)
(0.040, 5.6074)
(0.045, 5.1538)
(0.050, 4.9897)
(0.055, 4.6545)
(0.060, 4.4475)
(0.065, 4.0713)
(0.070, 3.8972)
(0.075, 3.6710)
(0.080, 3.4648)
(0.085, 3.2699)
(0.090, 3.1578)
(0.095, 3.0557)
(0.100, 2.9153)
(0.105, 2.7542)
(0.110, 2.7053)
(0.115, 2.6453)
(0.120, 2.5668)
(0.125, 2.5853)
(0.130, 2.5555)
(0.135, 2.6162)
(0.140, 2.6274)
(0.145, 2.7232)
(0.150, 2.8026)
(0.155, 2.9088)
(0.160, 3.0992)
(0.165, 3.2793)
(0.170, 3.5150)
(0.175, 3.7516)
(0.180, 3.9318)
(0.185, 4.3639)
(0.190, 4.6058)
(0.195, 4.9291)
(0.200, 5.3385)
(0.205, 5.6643)
(0.210, 6.0781)
(0.215, 6.5125)
(0.220, 6.8325)
(0.225, 7.3471)
(0.230, 7.9054)
(0.235, 8.2642)
(0.240, 8.7922)
(0.245, 9.2545)
(0.250, 9.7974)
}

\defdata{magnitudes-Config-B-1532-29-2147483-activation-norms-enc-64-max}{%
(0.064, 2.6238)
(0.128, 3.6149)
(0.192, 4.3193)
(0.256, 4.8725)
(0.320, 5.2535)
(0.384, 5.7754)
(0.448, 6.2239)
(0.512, 6.5830)
(0.576, 6.9185)
(0.640, 7.2277)
(0.704, 7.5324)
(0.768, 7.7925)
(0.832, 8.0556)
(0.896, 8.2989)
(0.960, 8.5685)
(1.024, 8.7971)
(1.088, 9.0037)
(1.152, 9.2187)
(1.216, 9.4087)
(1.280, 9.6162)
(1.344, 9.8642)
(1.408, 10.0336)
(1.472, 10.2585)
(1.536, 10.4192)
(1.600, 10.5720)
(1.664, 10.7971)
(1.728, 10.9467)
(1.792, 11.1118)
(1.856, 11.3243)
(1.920, 11.4830)
(1.984, 11.6577)
(2.048, 11.7881)
}
\defdata{magnitudes-Config-B-1532-29-2147483-activation-norms-enc-32-max}{%
(0.064, 7.9996)
(0.128, 12.4478)
(0.192, 15.8026)
(0.256, 18.4645)
(0.320, 20.4671)
(0.384, 22.2845)
(0.448, 24.2614)
(0.512, 25.8345)
(0.576, 27.1183)
(0.640, 28.1614)
(0.704, 29.1598)
(0.768, 30.0004)
(0.832, 30.6624)
(0.896, 31.5751)
(0.960, 32.3379)
(1.024, 32.9492)
(1.088, 33.5244)
(1.152, 34.0923)
(1.216, 34.5752)
(1.280, 35.1181)
(1.344, 35.7238)
(1.408, 36.3407)
(1.472, 36.9956)
(1.536, 37.5291)
(1.600, 38.0811)
(1.664, 38.7081)
(1.728, 39.1755)
(1.792, 39.7705)
(1.856, 40.3166)
(1.920, 40.7849)
(1.984, 41.3554)
(2.048, 41.7321)
}
\defdata{magnitudes-Config-B-1532-29-2147483-activation-norms-enc-16-max}{%
(0.064, 30.2633)
(0.128, 44.1051)
(0.192, 54.4404)
(0.256, 63.4612)
(0.320, 71.2901)
(0.384, 77.7852)
(0.448, 83.7989)
(0.512, 88.8494)
(0.576, 93.5010)
(0.640, 97.8293)
(0.704, 101.8668)
(0.768, 105.7461)
(0.832, 108.5533)
(0.896, 112.5738)
(0.960, 115.9452)
(1.024, 118.9357)
(1.088, 121.7016)
(1.152, 124.4809)
(1.216, 126.8542)
(1.280, 129.3677)
(1.344, 132.3517)
(1.408, 134.8420)
(1.472, 137.9089)
(1.536, 140.1559)
(1.600, 142.2679)
(1.664, 145.0741)
(1.728, 146.9466)
(1.792, 149.0184)
(1.856, 152.0192)
(1.920, 154.0483)
(1.984, 156.5401)
(2.048, 158.4613)
}
\defdata{magnitudes-Config-B-1532-29-2147483-activation-norms-enc-8-max}{%
(0.064, 42.2923)
(0.128, 57.3914)
(0.192, 68.5113)
(0.256, 80.1687)
(0.320, 91.9817)
(0.384, 103.2071)
(0.448, 113.3536)
(0.512, 121.8617)
(0.576, 129.7336)
(0.640, 138.0848)
(0.704, 145.9972)
(0.768, 153.5836)
(0.832, 161.0778)
(0.896, 168.2170)
(0.960, 175.4529)
(1.024, 182.1967)
(1.088, 188.1411)
(1.152, 194.8728)
(1.216, 200.3567)
(1.280, 205.9304)
(1.344, 211.1130)
(1.408, 216.4478)
(1.472, 221.9265)
(1.536, 226.7306)
(1.600, 231.4366)
(1.664, 236.4242)
(1.728, 240.6481)
(1.792, 244.6842)
(1.856, 249.1776)
(1.920, 253.2855)
(1.984, 257.4164)
(2.048, 261.2481)
}
\defdata{magnitudes-Config-B-1532-29-2147483-activation-norms-dec-8-max}{%
(0.064, 30.7014)
(0.128, 48.9009)
(0.192, 62.2459)
(0.256, 76.8851)
(0.320, 90.6003)
(0.384, 102.5315)
(0.448, 112.8493)
(0.512, 123.1216)
(0.576, 132.5556)
(0.640, 141.7701)
(0.704, 150.6124)
(0.768, 158.7491)
(0.832, 165.2753)
(0.896, 172.2803)
(0.960, 178.6461)
(1.024, 184.4591)
(1.088, 190.0904)
(1.152, 194.9342)
(1.216, 199.8417)
(1.280, 204.7536)
(1.344, 209.5713)
(1.408, 213.7605)
(1.472, 217.8651)
(1.536, 221.0556)
(1.600, 224.1404)
(1.664, 227.5752)
(1.728, 231.5474)
(1.792, 234.6622)
(1.856, 237.8743)
(1.920, 241.5450)
(1.984, 244.5644)
(2.048, 247.3598)
}
\defdata{magnitudes-Config-B-1532-29-2147483-activation-norms-dec-16-max}{%
(0.064, 29.2772)
(0.128, 39.0113)
(0.192, 47.2299)
(0.256, 55.9750)
(0.320, 63.3885)
(0.384, 69.9406)
(0.448, 75.6201)
(0.512, 80.5079)
(0.576, 84.4006)
(0.640, 87.9468)
(0.704, 91.5033)
(0.768, 94.6462)
(0.832, 96.4325)
(0.896, 99.7347)
(0.960, 102.5240)
(1.024, 104.7555)
(1.088, 106.7659)
(1.152, 108.8096)
(1.216, 110.5205)
(1.280, 112.7284)
(1.344, 114.6691)
(1.408, 116.4072)
(1.472, 118.4616)
(1.536, 119.7664)
(1.600, 121.4169)
(1.664, 123.1717)
(1.728, 124.7740)
(1.792, 126.7212)
(1.856, 128.3547)
(1.920, 129.9423)
(1.984, 131.6682)
(2.048, 133.1020)
}
\defdata{magnitudes-Config-B-1532-29-2147483-activation-norms-dec-32-max}{%
(0.064, 13.4837)
(0.128, 17.3570)
(0.192, 19.6925)
(0.256, 21.2716)
(0.320, 22.6289)
(0.384, 23.3754)
(0.448, 23.9678)
(0.512, 25.9930)
(0.576, 28.6397)
(0.640, 31.0448)
(0.704, 33.4434)
(0.768, 35.7809)
(0.832, 37.9592)
(0.896, 40.1917)
(0.960, 42.3504)
(1.024, 44.3448)
(1.088, 46.1918)
(1.152, 48.1189)
(1.216, 49.9815)
(1.280, 51.7921)
(1.344, 53.5102)
(1.408, 55.1819)
(1.472, 57.0700)
(1.536, 58.5586)
(1.600, 60.1995)
(1.664, 61.8529)
(1.728, 63.3861)
(1.792, 64.8912)
(1.856, 66.6332)
(1.920, 68.0833)
(1.984, 69.5959)
(2.048, 70.8926)
}
\defdata{magnitudes-Config-B-1532-29-2147483-activation-norms-dec-64-max}{%
(0.064, 5.1964)
(0.128, 7.0066)
(0.192, 8.3619)
(0.256, 9.4852)
(0.320, 11.5898)
(0.384, 13.9919)
(0.448, 16.5205)
(0.512, 19.0766)
(0.576, 21.3867)
(0.640, 23.8576)
(0.704, 26.5809)
(0.768, 29.2012)
(0.832, 31.1720)
(0.896, 34.5644)
(0.960, 37.3459)
(1.024, 40.1761)
(1.088, 42.6686)
(1.152, 45.3787)
(1.216, 47.8156)
(1.280, 50.6632)
(1.344, 54.1656)
(1.408, 56.6764)
(1.472, 60.7122)
(1.536, 62.9767)
(1.600, 65.3988)
(1.664, 69.0406)
(1.728, 71.2436)
(1.792, 73.9989)
(1.856, 77.5056)
(1.920, 80.3020)
(1.984, 83.3287)
(2.048, 85.4278)
}
\defdata{magnitudes-Config-B-1532-29-2147483-weight-norms-enc-64-max}{%
(0.064, 0.1292)
(0.128, 0.1480)
(0.192, 0.1648)
(0.256, 0.1789)
(0.320, 0.1927)
(0.384, 0.2059)
(0.448, 0.2180)
(0.512, 0.2293)
(0.576, 0.2407)
(0.640, 0.2526)
(0.704, 0.2638)
(0.768, 0.2744)
(0.832, 0.2844)
(0.896, 0.2940)
(0.960, 0.3031)
(1.024, 0.3119)
(1.088, 0.3204)
(1.152, 0.3290)
(1.216, 0.3372)
(1.280, 0.3452)
(1.344, 0.3530)
(1.408, 0.3606)
(1.472, 0.3676)
(1.536, 0.3743)
(1.600, 0.3808)
(1.664, 0.3871)
(1.728, 0.3933)
(1.792, 0.3999)
(1.856, 0.4068)
(1.920, 0.4136)
(1.984, 0.4202)
(2.048, 0.4265)
}
\defdata{magnitudes-Config-B-1532-29-2147483-weight-norms-enc-32-max}{%
(0.064, 0.0724)
(0.128, 0.0893)
(0.192, 0.1001)
(0.256, 0.1094)
(0.320, 0.1232)
(0.384, 0.1384)
(0.448, 0.1523)
(0.512, 0.1647)
(0.576, 0.1787)
(0.640, 0.1934)
(0.704, 0.2075)
(0.768, 0.2211)
(0.832, 0.2344)
(0.896, 0.2473)
(0.960, 0.2599)
(1.024, 0.2723)
(1.088, 0.2845)
(1.152, 0.2965)
(1.216, 0.3084)
(1.280, 0.3201)
(1.344, 0.3317)
(1.408, 0.3431)
(1.472, 0.3543)
(1.536, 0.3655)
(1.600, 0.3767)
(1.664, 0.3878)
(1.728, 0.3987)
(1.792, 0.4095)
(1.856, 0.4228)
(1.920, 0.4364)
(1.984, 0.4499)
(2.048, 0.4632)
}
\defdata{magnitudes-Config-B-1532-29-2147483-weight-norms-enc-16-max}{%
(0.064, 0.0543)
(0.128, 0.0819)
(0.192, 0.1034)
(0.256, 0.1239)
(0.320, 0.1429)
(0.384, 0.1597)
(0.448, 0.1750)
(0.512, 0.1890)
(0.576, 0.2019)
(0.640, 0.2148)
(0.704, 0.2269)
(0.768, 0.2404)
(0.832, 0.2533)
(0.896, 0.2653)
(0.960, 0.2771)
(1.024, 0.2886)
(1.088, 0.2998)
(1.152, 0.3108)
(1.216, 0.3215)
(1.280, 0.3319)
(1.344, 0.3420)
(1.408, 0.3520)
(1.472, 0.3619)
(1.536, 0.3712)
(1.600, 0.3804)
(1.664, 0.3894)
(1.728, 0.3983)
(1.792, 0.4072)
(1.856, 0.4183)
(1.920, 0.4289)
(1.984, 0.4392)
(2.048, 0.4490)
}
\defdata{magnitudes-Config-B-1532-29-2147483-weight-norms-enc-8-max}{%
(0.064, 0.0699)
(0.128, 0.0985)
(0.192, 0.1312)
(0.256, 0.1550)
(0.320, 0.1725)
(0.384, 0.1869)
(0.448, 0.1991)
(0.512, 0.2098)
(0.576, 0.2184)
(0.640, 0.2286)
(0.704, 0.2380)
(0.768, 0.2471)
(0.832, 0.2561)
(0.896, 0.2645)
(0.960, 0.2730)
(1.024, 0.2804)
(1.088, 0.2870)
(1.152, 0.2931)
(1.216, 0.3020)
(1.280, 0.3120)
(1.344, 0.3218)
(1.408, 0.3313)
(1.472, 0.3405)
(1.536, 0.3496)
(1.600, 0.3583)
(1.664, 0.3668)
(1.728, 0.3752)
(1.792, 0.3835)
(1.856, 0.3912)
(1.920, 0.3988)
(1.984, 0.4063)
(2.048, 0.4137)
}
\defdata{magnitudes-Config-B-1532-29-2147483-weight-norms-dec-8-max}{%
(0.064, 0.0684)
(0.128, 0.1075)
(0.192, 0.1288)
(0.256, 0.1479)
(0.320, 0.1653)
(0.384, 0.1811)
(0.448, 0.1941)
(0.512, 0.2058)
(0.576, 0.2163)
(0.640, 0.2254)
(0.704, 0.2338)
(0.768, 0.2444)
(0.832, 0.2561)
(0.896, 0.2681)
(0.960, 0.2797)
(1.024, 0.2900)
(1.088, 0.3001)
(1.152, 0.3098)
(1.216, 0.3192)
(1.280, 0.3286)
(1.344, 0.3380)
(1.408, 0.3467)
(1.472, 0.3589)
(1.536, 0.3719)
(1.600, 0.3847)
(1.664, 0.3977)
(1.728, 0.4105)
(1.792, 0.4233)
(1.856, 0.4359)
(1.920, 0.4485)
(1.984, 0.4611)
(2.048, 0.4736)
}
\defdata{magnitudes-Config-B-1532-29-2147483-weight-norms-dec-16-max}{%
(0.064, 0.0665)
(0.128, 0.1067)
(0.192, 0.1408)
(0.256, 0.1680)
(0.320, 0.1914)
(0.384, 0.2130)
(0.448, 0.2331)
(0.512, 0.2509)
(0.576, 0.2653)
(0.640, 0.2817)
(0.704, 0.2968)
(0.768, 0.3110)
(0.832, 0.3251)
(0.896, 0.3389)
(0.960, 0.3521)
(1.024, 0.3659)
(1.088, 0.3819)
(1.152, 0.3968)
(1.216, 0.4110)
(1.280, 0.4247)
(1.344, 0.4380)
(1.408, 0.4519)
(1.472, 0.4650)
(1.536, 0.4783)
(1.600, 0.4907)
(1.664, 0.5029)
(1.728, 0.5147)
(1.792, 0.5264)
(1.856, 0.5383)
(1.920, 0.5498)
(1.984, 0.5618)
(2.048, 0.5740)
}
\defdata{magnitudes-Config-B-1532-29-2147483-weight-norms-dec-32-max}{%
(0.064, 0.0703)
(0.128, 0.1097)
(0.192, 0.1748)
(0.256, 0.2217)
(0.320, 0.2579)
(0.384, 0.2882)
(0.448, 0.3137)
(0.512, 0.3368)
(0.576, 0.3591)
(0.640, 0.3795)
(0.704, 0.3981)
(0.768, 0.4168)
(0.832, 0.4352)
(0.896, 0.4529)
(0.960, 0.4705)
(1.024, 0.4877)
(1.088, 0.5037)
(1.152, 0.5181)
(1.216, 0.5328)
(1.280, 0.5469)
(1.344, 0.5610)
(1.408, 0.5749)
(1.472, 0.5887)
(1.536, 0.6020)
(1.600, 0.6146)
(1.664, 0.6273)
(1.728, 0.6397)
(1.792, 0.6516)
(1.856, 0.6637)
(1.920, 0.6757)
(1.984, 0.6874)
(2.048, 0.6991)
}
\defdata{magnitudes-Config-B-1532-29-2147483-weight-norms-dec-64-max}{%
(0.064, 0.0549)
(0.128, 0.0822)
(0.192, 0.1024)
(0.256, 0.1181)
(0.320, 0.1290)
(0.384, 0.1382)
(0.448, 0.1458)
(0.512, 0.1528)
(0.576, 0.1593)
(0.640, 0.1652)
(0.704, 0.1706)
(0.768, 0.1758)
(0.832, 0.1870)
(0.896, 0.2006)
(0.960, 0.2139)
(1.024, 0.2270)
(1.088, 0.2398)
(1.152, 0.2523)
(1.216, 0.2646)
(1.280, 0.2766)
(1.344, 0.2883)
(1.408, 0.2998)
(1.472, 0.3111)
(1.536, 0.3223)
(1.600, 0.3334)
(1.664, 0.3444)
(1.728, 0.3551)
(1.792, 0.3657)
(1.856, 0.3762)
(1.920, 0.3866)
(1.984, 0.3970)
(2.048, 0.4072)
}
\defdata{magnitudes-Config-C-1533-41-2147483-activation-norms-enc-64-max}{%
(0.064, 8.8362)
(0.128, 11.3444)
(0.192, 13.5393)
(0.256, 15.8259)
(0.320, 18.0350)
(0.384, 20.2494)
(0.448, 22.3836)
(0.512, 24.5552)
(0.576, 26.5796)
(0.640, 28.5677)
(0.704, 30.3833)
(0.768, 32.3227)
(0.832, 33.9274)
(0.896, 35.7195)
(0.960, 37.3157)
(1.024, 39.1493)
(1.088, 40.4659)
(1.152, 42.1005)
(1.216, 43.6148)
(1.280, 44.9671)
(1.344, 46.4367)
(1.408, 47.7603)
(1.472, 49.2244)
(1.536, 50.3651)
(1.600, 51.8049)
(1.664, 53.0331)
(1.728, 54.2353)
(1.792, 55.4110)
(1.856, 56.7472)
(1.920, 57.8908)
(1.984, 59.0377)
(2.048, 60.1162)
}
\defdata{magnitudes-Config-C-1533-41-2147483-activation-norms-enc-32-max}{%
(0.064, 10.0913)
(0.128, 15.9150)
(0.192, 21.5252)
(0.256, 26.8602)
(0.320, 31.8137)
(0.384, 36.9605)
(0.448, 41.4161)
(0.512, 46.0089)
(0.576, 51.3679)
(0.640, 55.2069)
(0.704, 59.8698)
(0.768, 64.4469)
(0.832, 68.2323)
(0.896, 73.0345)
(0.960, 76.7410)
(1.024, 81.3440)
(1.088, 84.9402)
(1.152, 89.9409)
(1.216, 93.3707)
(1.280, 96.7831)
(1.344, 100.2707)
(1.408, 103.9453)
(1.472, 108.0297)
(1.536, 110.8010)
(1.600, 114.7230)
(1.664, 117.9996)
(1.728, 122.0562)
(1.792, 124.0364)
(1.856, 128.4232)
(1.920, 132.7704)
(1.984, 136.6636)
(2.048, 139.3722)
}
\defdata{magnitudes-Config-C-1533-41-2147483-activation-norms-enc-16-max}{%
(0.064, 28.1811)
(0.128, 46.9772)
(0.192, 66.2492)
(0.256, 85.3043)
(0.320, 104.4869)
(0.384, 123.4342)
(0.448, 141.5495)
(0.512, 160.2588)
(0.576, 180.1085)
(0.640, 197.3262)
(0.704, 215.4576)
(0.768, 234.4328)
(0.832, 251.5568)
(0.896, 270.6798)
(0.960, 288.2711)
(1.024, 307.5160)
(1.088, 324.9291)
(1.152, 345.1654)
(1.216, 362.2166)
(1.280, 379.6525)
(1.344, 396.9313)
(1.408, 416.4696)
(1.472, 434.9132)
(1.536, 451.1871)
(1.600, 469.0977)
(1.664, 487.2361)
(1.728, 505.9838)
(1.792, 519.5947)
(1.856, 539.1262)
(1.920, 558.5109)
(1.984, 576.6994)
(2.048, 591.6521)
}
\defdata{magnitudes-Config-C-1533-41-2147483-activation-norms-enc-8-max}{%
(0.064, 42.3878)
(0.128, 69.9108)
(0.192, 92.0064)
(0.256, 111.9357)
(0.320, 130.7245)
(0.384, 147.4008)
(0.448, 162.3627)
(0.512, 178.3684)
(0.576, 194.5179)
(0.640, 207.2493)
(0.704, 221.8232)
(0.768, 235.5168)
(0.832, 247.7948)
(0.896, 261.3950)
(0.960, 273.7065)
(1.024, 286.3995)
(1.088, 297.6877)
(1.152, 311.5087)
(1.216, 322.4531)
(1.280, 334.1281)
(1.344, 345.3686)
(1.408, 357.1210)
(1.472, 368.7145)
(1.536, 378.8813)
(1.600, 390.0544)
(1.664, 401.7576)
(1.728, 412.3982)
(1.792, 421.5651)
(1.856, 432.2595)
(1.920, 443.7112)
(1.984, 454.4707)
(2.048, 463.3930)
}
\defdata{magnitudes-Config-C-1533-41-2147483-activation-norms-dec-8-max}{%
(0.064, 16.2623)
(0.128, 25.3507)
(0.192, 38.2424)
(0.256, 54.3553)
(0.320, 70.3663)
(0.384, 86.7785)
(0.448, 101.7345)
(0.512, 118.5703)
(0.576, 136.1099)
(0.640, 150.8267)
(0.704, 167.2219)
(0.768, 184.2175)
(0.832, 199.3783)
(0.896, 217.1494)
(0.960, 233.2065)
(1.024, 251.5673)
(1.088, 266.0502)
(1.152, 286.0660)
(1.216, 298.5317)
(1.280, 317.2978)
(1.344, 332.7401)
(1.408, 349.3141)
(1.472, 368.0420)
(1.536, 380.7721)
(1.600, 399.1501)
(1.664, 415.2498)
(1.728, 433.7037)
(1.792, 446.0404)
(1.856, 461.3250)
(1.920, 481.6169)
(1.984, 496.9229)
(2.048, 510.9465)
}
\defdata{magnitudes-Config-C-1533-41-2147483-activation-norms-dec-16-max}{%
(0.064, 40.1091)
(0.128, 58.5491)
(0.192, 75.0126)
(0.256, 92.2982)
(0.320, 108.0438)
(0.384, 123.5371)
(0.448, 137.1358)
(0.512, 151.7675)
(0.576, 166.8840)
(0.640, 179.1522)
(0.704, 191.6830)
(0.768, 204.8283)
(0.832, 216.2389)
(0.896, 230.1052)
(0.960, 240.9600)
(1.024, 254.3184)
(1.088, 265.1484)
(1.152, 279.2903)
(1.216, 288.6014)
(1.280, 299.6167)
(1.344, 310.1827)
(1.408, 321.7930)
(1.472, 332.9326)
(1.536, 341.3368)
(1.600, 353.0300)
(1.664, 364.5037)
(1.728, 375.7624)
(1.792, 382.7832)
(1.856, 392.5275)
(1.920, 404.8261)
(1.984, 415.5749)
(2.048, 424.2291)
}
\defdata{magnitudes-Config-C-1533-41-2147483-activation-norms-dec-32-max}{%
(0.064, 19.2071)
(0.128, 29.3036)
(0.192, 37.7274)
(0.256, 45.2225)
(0.320, 51.4970)
(0.384, 57.9591)
(0.448, 63.3152)
(0.512, 68.9909)
(0.576, 75.6281)
(0.640, 79.8894)
(0.704, 85.0056)
(0.768, 90.0050)
(0.832, 94.1485)
(0.896, 99.3968)
(0.960, 103.4593)
(1.024, 108.4680)
(1.088, 112.1751)
(1.152, 117.6599)
(1.216, 120.6673)
(1.280, 124.7047)
(1.344, 127.9011)
(1.408, 131.9350)
(1.472, 136.0018)
(1.536, 138.9466)
(1.600, 143.0410)
(1.664, 146.6157)
(1.728, 150.4303)
(1.792, 152.1700)
(1.856, 156.1481)
(1.920, 160.2232)
(1.984, 164.0005)
(2.048, 166.1838)
}
\defdata{magnitudes-Config-C-1533-41-2147483-activation-norms-dec-64-max}{%
(0.064, 9.5649)
(0.128, 11.4693)
(0.192, 12.9947)
(0.256, 14.3658)
(0.320, 15.4884)
(0.384, 16.6359)
(0.448, 17.3410)
(0.512, 18.1414)
(0.576, 19.1729)
(0.640, 21.1785)
(0.704, 23.3531)
(0.768, 25.5844)
(0.832, 27.6764)
(0.896, 29.9383)
(0.960, 32.1526)
(1.024, 34.4943)
(1.088, 36.6596)
(1.152, 39.1462)
(1.216, 41.3153)
(1.280, 43.6398)
(1.344, 45.9000)
(1.408, 48.2555)
(1.472, 50.9232)
(1.536, 52.9099)
(1.600, 55.3843)
(1.664, 57.7577)
(1.728, 60.2355)
(1.792, 62.3452)
(1.856, 64.9701)
(1.920, 67.6320)
(1.984, 70.2161)
(2.048, 72.4696)
}
\defdata{magnitudes-Config-C-1533-41-2147483-weight-norms-enc-64-max}{%
(0.064, 0.1100)
(0.128, 0.1243)
(0.192, 0.1404)
(0.256, 0.1549)
(0.320, 0.1682)
(0.384, 0.1805)
(0.448, 0.1919)
(0.512, 0.2023)
(0.576, 0.2119)
(0.640, 0.2207)
(0.704, 0.2336)
(0.768, 0.2495)
(0.832, 0.2650)
(0.896, 0.2799)
(0.960, 0.2943)
(1.024, 0.3087)
(1.088, 0.3229)
(1.152, 0.3370)
(1.216, 0.3510)
(1.280, 0.3649)
(1.344, 0.3783)
(1.408, 0.3914)
(1.472, 0.4045)
(1.536, 0.4174)
(1.600, 0.4301)
(1.664, 0.4426)
(1.728, 0.4552)
(1.792, 0.4676)
(1.856, 0.4799)
(1.920, 0.4922)
(1.984, 0.5044)
(2.048, 0.5165)
}
\defdata{magnitudes-Config-C-1533-41-2147483-weight-norms-enc-32-max}{%
(0.064, 0.0697)
(0.128, 0.1144)
(0.192, 0.1535)
(0.256, 0.1886)
(0.320, 0.2212)
(0.384, 0.2527)
(0.448, 0.2842)
(0.512, 0.3142)
(0.576, 0.3439)
(0.640, 0.3728)
(0.704, 0.4011)
(0.768, 0.4286)
(0.832, 0.4558)
(0.896, 0.4839)
(0.960, 0.5111)
(1.024, 0.5382)
(1.088, 0.5646)
(1.152, 0.5900)
(1.216, 0.6152)
(1.280, 0.6401)
(1.344, 0.6646)
(1.408, 0.6887)
(1.472, 0.7123)
(1.536, 0.7354)
(1.600, 0.7582)
(1.664, 0.7808)
(1.728, 0.8029)
(1.792, 0.8245)
(1.856, 0.8461)
(1.920, 0.8678)
(1.984, 0.8894)
(2.048, 0.9110)
}
\defdata{magnitudes-Config-C-1533-41-2147483-weight-norms-enc-16-max}{%
(0.064, 0.0965)
(0.128, 0.1485)
(0.192, 0.1903)
(0.256, 0.2327)
(0.320, 0.2789)
(0.384, 0.3228)
(0.448, 0.3650)
(0.512, 0.4058)
(0.576, 0.4462)
(0.640, 0.4859)
(0.704, 0.5246)
(0.768, 0.5633)
(0.832, 0.6016)
(0.896, 0.6395)
(0.960, 0.6770)
(1.024, 0.7129)
(1.088, 0.7492)
(1.152, 0.7855)
(1.216, 0.8212)
(1.280, 0.8561)
(1.344, 0.8913)
(1.408, 0.9265)
(1.472, 0.9612)
(1.536, 0.9948)
(1.600, 1.0280)
(1.664, 1.0617)
(1.728, 1.0948)
(1.792, 1.1276)
(1.856, 1.1602)
(1.920, 1.1930)
(1.984, 1.2250)
(2.048, 1.2570)
}
\defdata{magnitudes-Config-C-1533-41-2147483-weight-norms-enc-8-max}{%
(0.064, 0.0767)
(0.128, 0.1170)
(0.192, 0.1502)
(0.256, 0.1776)
(0.320, 0.2010)
(0.384, 0.2206)
(0.448, 0.2381)
(0.512, 0.2623)
(0.576, 0.2855)
(0.640, 0.3085)
(0.704, 0.3302)
(0.768, 0.3519)
(0.832, 0.3728)
(0.896, 0.3933)
(0.960, 0.4137)
(1.024, 0.4336)
(1.088, 0.4533)
(1.152, 0.4728)
(1.216, 0.4918)
(1.280, 0.5111)
(1.344, 0.5306)
(1.408, 0.5495)
(1.472, 0.5682)
(1.536, 0.5865)
(1.600, 0.6047)
(1.664, 0.6223)
(1.728, 0.6400)
(1.792, 0.6576)
(1.856, 0.6752)
(1.920, 0.6925)
(1.984, 0.7098)
(2.048, 0.7270)
}
\defdata{magnitudes-Config-C-1533-41-2147483-weight-norms-dec-8-max}{%
(0.064, 0.0705)
(0.128, 0.0950)
(0.192, 0.1109)
(0.256, 0.1343)
(0.320, 0.1569)
(0.384, 0.1765)
(0.448, 0.1939)
(0.512, 0.2100)
(0.576, 0.2244)
(0.640, 0.2381)
(0.704, 0.2513)
(0.768, 0.2636)
(0.832, 0.2753)
(0.896, 0.2865)
(0.960, 0.2968)
(1.024, 0.3072)
(1.088, 0.3168)
(1.152, 0.3259)
(1.216, 0.3366)
(1.280, 0.3482)
(1.344, 0.3596)
(1.408, 0.3709)
(1.472, 0.3822)
(1.536, 0.3933)
(1.600, 0.4040)
(1.664, 0.4146)
(1.728, 0.4259)
(1.792, 0.4373)
(1.856, 0.4489)
(1.920, 0.4605)
(1.984, 0.4718)
(2.048, 0.4829)
}
\defdata{magnitudes-Config-C-1533-41-2147483-weight-norms-dec-16-max}{%
(0.064, 0.0961)
(0.128, 0.1325)
(0.192, 0.1521)
(0.256, 0.1715)
(0.320, 0.1956)
(0.384, 0.2194)
(0.448, 0.2459)
(0.512, 0.2707)
(0.576, 0.2946)
(0.640, 0.3175)
(0.704, 0.3395)
(0.768, 0.3605)
(0.832, 0.3806)
(0.896, 0.4003)
(0.960, 0.4194)
(1.024, 0.4382)
(1.088, 0.4565)
(1.152, 0.4744)
(1.216, 0.4918)
(1.280, 0.5090)
(1.344, 0.5257)
(1.408, 0.5421)
(1.472, 0.5582)
(1.536, 0.5740)
(1.600, 0.5896)
(1.664, 0.6049)
(1.728, 0.6200)
(1.792, 0.6350)
(1.856, 0.6497)
(1.920, 0.6641)
(1.984, 0.6783)
(2.048, 0.6925)
}
\defdata{magnitudes-Config-C-1533-41-2147483-weight-norms-dec-32-max}{%
(0.064, 0.0640)
(0.128, 0.0924)
(0.192, 0.1060)
(0.256, 0.1219)
(0.320, 0.1487)
(0.384, 0.1651)
(0.448, 0.1774)
(0.512, 0.1957)
(0.576, 0.2130)
(0.640, 0.2302)
(0.704, 0.2467)
(0.768, 0.2623)
(0.832, 0.2777)
(0.896, 0.2930)
(0.960, 0.3074)
(1.024, 0.3215)
(1.088, 0.3351)
(1.152, 0.3478)
(1.216, 0.3601)
(1.280, 0.3720)
(1.344, 0.3832)
(1.408, 0.3938)
(1.472, 0.4038)
(1.536, 0.4135)
(1.600, 0.4233)
(1.664, 0.4328)
(1.728, 0.4418)
(1.792, 0.4504)
(1.856, 0.4591)
(1.920, 0.4677)
(1.984, 0.4762)
(2.048, 0.4844)
}
\defdata{magnitudes-Config-C-1533-41-2147483-weight-norms-dec-64-max}{%
(0.064, 0.0507)
(0.128, 0.0758)
(0.192, 0.0933)
(0.256, 0.1056)
(0.320, 0.1147)
(0.384, 0.1209)
(0.448, 0.1258)
(0.512, 0.1325)
(0.576, 0.1466)
(0.640, 0.1608)
(0.704, 0.1757)
(0.768, 0.1906)
(0.832, 0.2052)
(0.896, 0.2194)
(0.960, 0.2335)
(1.024, 0.2474)
(1.088, 0.2613)
(1.152, 0.2751)
(1.216, 0.2888)
(1.280, 0.3024)
(1.344, 0.3159)
(1.408, 0.3292)
(1.472, 0.3426)
(1.536, 0.3561)
(1.600, 0.3697)
(1.664, 0.3832)
(1.728, 0.3968)
(1.792, 0.4104)
(1.856, 0.4239)
(1.920, 0.4372)
(1.984, 0.4505)
(2.048, 0.4637)
}
\defdata{magnitudes-Config-D-1533-45-2147483-activation-norms-enc-64-max}{%
(0.064, 7.7214)
(0.128, 6.7114)
(0.192, 6.4944)
(0.256, 6.3175)
(0.320, 6.2099)
(0.384, 6.1284)
(0.448, 6.0490)
(0.512, 5.9843)
(0.576, 5.9226)
(0.640, 5.9026)
(0.704, 5.8897)
(0.768, 5.8800)
(0.832, 5.8780)
(0.896, 5.8828)
(0.960, 5.8840)
(1.024, 5.8858)
(1.088, 5.8845)
(1.152, 5.8897)
(1.216, 5.8860)
(1.280, 5.8878)
(1.344, 5.8757)
(1.408, 5.8724)
(1.472, 5.8685)
(1.536, 5.8570)
(1.600, 5.8547)
(1.664, 5.8456)
(1.728, 5.8353)
(1.792, 5.8246)
(1.856, 5.8247)
(1.920, 5.8074)
(1.984, 5.8018)
(2.048, 5.7936)
}
\defdata{magnitudes-Config-D-1533-45-2147483-activation-norms-enc-32-max}{%
(0.064, 7.1894)
(0.128, 7.6048)
(0.192, 7.7238)
(0.256, 7.7701)
(0.320, 7.7714)
(0.384, 7.8024)
(0.448, 7.7955)
(0.512, 7.8168)
(0.576, 7.8257)
(0.640, 7.8271)
(0.704, 7.8690)
(0.768, 7.8835)
(0.832, 7.8917)
(0.896, 7.9136)
(0.960, 7.9315)
(1.024, 7.9459)
(1.088, 7.9528)
(1.152, 7.9933)
(1.216, 7.9984)
(1.280, 8.0139)
(1.344, 8.0328)
(1.408, 8.0533)
(1.472, 8.0969)
(1.536, 8.0694)
(1.600, 8.1163)
(1.664, 8.1154)
(1.728, 8.1299)
(1.792, 8.1232)
(1.856, 8.1452)
(1.920, 8.1500)
(1.984, 8.1460)
(2.048, 8.1488)
}
\defdata{magnitudes-Config-D-1533-45-2147483-activation-norms-enc-16-max}{%
(0.064, 8.6286)
(0.128, 8.6772)
(0.192, 8.4890)
(0.256, 8.3068)
(0.320, 8.3621)
(0.384, 8.4661)
(0.448, 8.5424)
(0.512, 8.6699)
(0.576, 8.7748)
(0.640, 8.8701)
(0.704, 9.0318)
(0.768, 9.1551)
(0.832, 9.2588)
(0.896, 9.3529)
(0.960, 9.4490)
(1.024, 9.5461)
(1.088, 9.6669)
(1.152, 9.7480)
(1.216, 9.8810)
(1.280, 9.9096)
(1.344, 10.0090)
(1.408, 10.0764)
(1.472, 10.1552)
(1.536, 10.1831)
(1.600, 10.2576)
(1.664, 10.3254)
(1.728, 10.3837)
(1.792, 10.4418)
(1.856, 10.4946)
(1.920, 10.5417)
(1.984, 10.5682)
(2.048, 10.5892)
}
\defdata{magnitudes-Config-D-1533-45-2147483-activation-norms-enc-8-max}{%
(0.064, 10.1344)
(0.128, 9.3473)
(0.192, 8.7184)
(0.256, 8.2208)
(0.320, 7.9112)
(0.384, 7.6164)
(0.448, 7.3225)
(0.512, 7.0666)
(0.576, 6.8415)
(0.640, 6.6443)
(0.704, 6.5256)
(0.768, 6.4172)
(0.832, 6.2687)
(0.896, 6.0893)
(0.960, 5.9813)
(1.024, 5.8626)
(1.088, 5.7511)
(1.152, 5.6419)
(1.216, 5.5971)
(1.280, 5.4828)
(1.344, 5.4042)
(1.408, 5.3278)
(1.472, 5.2582)
(1.536, 5.1713)
(1.600, 5.1119)
(1.664, 5.0576)
(1.728, 5.0045)
(1.792, 4.9470)
(1.856, 4.8965)
(1.920, 4.8520)
(1.984, 4.7788)
(2.048, 4.7339)
}
\defdata{magnitudes-Config-D-1533-45-2147483-activation-norms-dec-8-max}{%
(0.064, 5.3935)
(0.128, 5.8919)
(0.192, 5.5649)
(0.256, 5.3994)
(0.320, 5.4437)
(0.384, 5.4238)
(0.448, 5.4383)
(0.512, 5.4459)
(0.576, 5.4577)
(0.640, 5.4567)
(0.704, 5.4660)
(0.768, 5.4708)
(0.832, 5.4453)
(0.896, 5.4422)
(0.960, 5.4591)
(1.024, 5.4536)
(1.088, 5.4361)
(1.152, 5.4177)
(1.216, 5.3992)
(1.280, 5.4003)
(1.344, 5.3846)
(1.408, 5.3210)
(1.472, 5.3230)
(1.536, 5.3068)
(1.600, 5.2946)
(1.664, 5.2639)
(1.728, 5.2502)
(1.792, 5.2615)
(1.856, 5.2243)
(1.920, 5.2244)
(1.984, 5.1861)
(2.048, 5.1834)
}
\defdata{magnitudes-Config-D-1533-45-2147483-activation-norms-dec-16-max}{%
(0.064, 10.5403)
(0.128, 11.4926)
(0.192, 11.6519)
(0.256, 11.6602)
(0.320, 11.6947)
(0.384, 11.6604)
(0.448, 11.7115)
(0.512, 11.7382)
(0.576, 11.7610)
(0.640, 11.7591)
(0.704, 11.8206)
(0.768, 11.8421)
(0.832, 11.8307)
(0.896, 11.8721)
(0.960, 11.9148)
(1.024, 11.9812)
(1.088, 12.0135)
(1.152, 12.0336)
(1.216, 12.1135)
(1.280, 12.1122)
(1.344, 12.1703)
(1.408, 12.1790)
(1.472, 12.2302)
(1.536, 12.2029)
(1.600, 12.2604)
(1.664, 12.2828)
(1.728, 12.3203)
(1.792, 12.3802)
(1.856, 12.3944)
(1.920, 12.4276)
(1.984, 12.4177)
(2.048, 12.4303)
}
\defdata{magnitudes-Config-D-1533-45-2147483-activation-norms-dec-32-max}{%
(0.064, 9.3560)
(0.128, 9.9061)
(0.192, 9.7695)
(0.256, 9.6349)
(0.320, 9.4810)
(0.384, 9.2949)
(0.448, 9.0782)
(0.512, 8.9190)
(0.576, 8.7677)
(0.640, 8.6028)
(0.704, 8.4721)
(0.768, 8.3417)
(0.832, 8.1792)
(0.896, 8.0194)
(0.960, 7.9080)
(1.024, 7.7713)
(1.088, 7.6182)
(1.152, 7.5104)
(1.216, 7.4439)
(1.280, 7.3662)
(1.344, 7.3038)
(1.408, 7.2142)
(1.472, 7.1610)
(1.536, 7.0766)
(1.600, 7.0263)
(1.664, 6.9540)
(1.728, 6.8859)
(1.792, 6.8377)
(1.856, 6.7728)
(1.920, 6.7304)
(1.984, 6.6542)
(2.048, 6.6170)
}
\defdata{magnitudes-Config-D-1533-45-2147483-activation-norms-dec-64-max}{%
(0.064, 9.4735)
(0.128, 10.1868)
(0.192, 10.2698)
(0.256, 10.1734)
(0.320, 10.0785)
(0.384, 9.9273)
(0.448, 9.7843)
(0.512, 9.6052)
(0.576, 9.4954)
(0.640, 9.3292)
(0.704, 9.2343)
(0.768, 9.1236)
(0.832, 8.9313)
(0.896, 8.8125)
(0.960, 8.7002)
(1.024, 8.5693)
(1.088, 8.4624)
(1.152, 8.3478)
(1.216, 8.4579)
(1.280, 8.3765)
(1.344, 8.3370)
(1.408, 8.1341)
(1.472, 8.2816)
(1.536, 8.0637)
(1.600, 8.1740)
(1.664, 8.0788)
(1.728, 7.9093)
(1.792, 8.2221)
(1.856, 7.9891)
(1.920, 8.2344)
(1.984, 7.9868)
(2.048, 7.7778)
}
\defdata{magnitudes-Config-D-1533-45-2147483-weight-norms-enc-64-max}{%
(0.064, 1.6272)
(0.128, 2.4477)
(0.192, 3.2019)
(0.256, 3.9668)
(0.320, 4.6529)
(0.384, 5.3675)
(0.448, 6.1075)
(0.512, 6.8075)
(0.576, 7.4368)
(0.640, 8.0370)
(0.704, 8.5993)
(0.768, 9.1273)
(0.832, 9.6192)
(0.896, 10.0913)
(0.960, 10.5523)
(1.024, 10.9907)
(1.088, 11.4361)
(1.152, 11.9198)
(1.216, 12.3963)
(1.280, 12.8641)
(1.344, 13.3234)
(1.408, 13.7622)
(1.472, 14.2081)
(1.536, 14.6446)
(1.600, 15.0698)
(1.664, 15.5195)
(1.728, 16.1764)
(1.792, 16.8337)
(1.856, 17.4887)
(1.920, 18.1522)
(1.984, 18.8085)
(2.048, 19.4687)
}
\defdata{magnitudes-Config-D-1533-45-2147483-weight-norms-enc-32-max}{%
(0.064, 1.7720)
(0.128, 2.7254)
(0.192, 3.5546)
(0.256, 4.3212)
(0.320, 5.2143)
(0.384, 6.0184)
(0.448, 6.7749)
(0.512, 7.5183)
(0.576, 8.2721)
(0.640, 8.9734)
(0.704, 9.6729)
(0.768, 10.3570)
(0.832, 11.2468)
(0.896, 12.1753)
(0.960, 13.0997)
(1.024, 14.0583)
(1.088, 15.0108)
(1.152, 15.9515)
(1.216, 16.8867)
(1.280, 17.8452)
(1.344, 18.8270)
(1.408, 19.8050)
(1.472, 20.7668)
(1.536, 21.7067)
(1.600, 22.6523)
(1.664, 23.5828)
(1.728, 24.5160)
(1.792, 25.4652)
(1.856, 26.4364)
(1.920, 27.4168)
(1.984, 28.4116)
(2.048, 29.3995)
}
\defdata{magnitudes-Config-D-1533-45-2147483-weight-norms-enc-16-max}{%
(0.064, 1.8866)
(0.128, 2.7709)
(0.192, 3.7919)
(0.256, 4.7188)
(0.320, 5.6380)
(0.384, 6.5752)
(0.448, 7.4548)
(0.512, 8.3378)
(0.576, 9.2076)
(0.640, 10.0759)
(0.704, 10.9114)
(0.768, 11.7495)
(0.832, 12.5724)
(0.896, 13.4602)
(0.960, 14.4530)
(1.024, 15.4478)
(1.088, 16.4637)
(1.152, 17.5119)
(1.216, 18.5578)
(1.280, 19.5912)
(1.344, 20.6161)
(1.408, 21.6309)
(1.472, 22.6407)
(1.536, 23.6592)
(1.600, 24.6972)
(1.664, 25.7120)
(1.728, 26.7268)
(1.792, 27.7223)
(1.856, 28.7190)
(1.920, 29.7142)
(1.984, 30.7131)
(2.048, 31.7161)
}
\defdata{magnitudes-Config-D-1533-45-2147483-weight-norms-enc-8-max}{%
(0.064, 2.0538)
(0.128, 2.9920)
(0.192, 3.8063)
(0.256, 4.5947)
(0.320, 5.4660)
(0.384, 6.3097)
(0.448, 7.1741)
(0.512, 8.1034)
(0.576, 9.0013)
(0.640, 9.8475)
(0.704, 10.7107)
(0.768, 11.5838)
(0.832, 12.4481)
(0.896, 13.3176)
(0.960, 14.1656)
(1.024, 15.0134)
(1.088, 15.8501)
(1.152, 16.6831)
(1.216, 17.4941)
(1.280, 18.2958)
(1.344, 19.1091)
(1.408, 19.9173)
(1.472, 20.7290)
(1.536, 21.5084)
(1.600, 22.3037)
(1.664, 23.0838)
(1.728, 23.8475)
(1.792, 24.6156)
(1.856, 25.3677)
(1.920, 26.1281)
(1.984, 26.8815)
(2.048, 27.6285)
}
\defdata{magnitudes-Config-D-1533-45-2147483-weight-norms-dec-8-max}{%
(0.064, 2.2074)
(0.128, 3.3358)
(0.192, 4.2970)
(0.256, 5.3933)
(0.320, 6.4934)
(0.384, 7.5645)
(0.448, 8.6921)
(0.512, 9.7446)
(0.576, 10.7842)
(0.640, 11.8128)
(0.704, 12.8227)
(0.768, 13.8279)
(0.832, 14.8210)
(0.896, 15.8319)
(0.960, 16.8058)
(1.024, 17.7903)
(1.088, 18.7761)
(1.152, 19.7781)
(1.216, 20.7522)
(1.280, 21.6960)
(1.344, 22.6258)
(1.408, 23.5693)
(1.472, 24.4940)
(1.536, 25.3947)
(1.600, 26.2763)
(1.664, 27.1345)
(1.728, 27.9960)
(1.792, 28.8468)
(1.856, 29.7076)
(1.920, 30.5708)
(1.984, 31.3929)
(2.048, 32.2123)
}
\defdata{magnitudes-Config-D-1533-45-2147483-weight-norms-dec-16-max}{%
(0.064, 1.9113)
(0.128, 2.7180)
(0.192, 3.4269)
(0.256, 4.1169)
(0.320, 4.7361)
(0.384, 5.3246)
(0.448, 5.9255)
(0.512, 6.4853)
(0.576, 7.0244)
(0.640, 7.6431)
(0.704, 8.2094)
(0.768, 8.7704)
(0.832, 9.3101)
(0.896, 9.8390)
(0.960, 10.3491)
(1.024, 10.8450)
(1.088, 11.3410)
(1.152, 11.8028)
(1.216, 12.2326)
(1.280, 12.6764)
(1.344, 13.0891)
(1.408, 13.5295)
(1.472, 13.9505)
(1.536, 14.3581)
(1.600, 14.7508)
(1.664, 15.1371)
(1.728, 15.5425)
(1.792, 15.9406)
(1.856, 16.3122)
(1.920, 16.6804)
(1.984, 17.0363)
(2.048, 17.3990)
}
\defdata{magnitudes-Config-D-1533-45-2147483-weight-norms-dec-32-max}{%
(0.064, 1.8390)
(0.128, 2.9157)
(0.192, 4.0294)
(0.256, 4.9500)
(0.320, 5.7323)
(0.384, 6.4666)
(0.448, 7.2431)
(0.512, 7.9669)
(0.576, 8.6535)
(0.640, 9.3168)
(0.704, 9.9533)
(0.768, 10.5299)
(0.832, 11.0916)
(0.896, 11.6255)
(0.960, 12.1334)
(1.024, 12.6282)
(1.088, 13.1058)
(1.152, 13.7663)
(1.216, 14.4085)
(1.280, 15.0623)
(1.344, 15.6836)
(1.408, 16.2964)
(1.472, 16.9133)
(1.536, 17.5294)
(1.600, 18.1367)
(1.664, 18.7343)
(1.728, 19.3138)
(1.792, 19.8660)
(1.856, 20.4230)
(1.920, 20.9695)
(1.984, 21.4996)
(2.048, 22.0106)
}
\defdata{magnitudes-Config-D-1533-45-2147483-weight-norms-dec-64-max}{%
(0.064, 1.7650)
(0.128, 2.8889)
(0.192, 3.8719)
(0.256, 5.1158)
(0.320, 6.4235)
(0.384, 7.3826)
(0.448, 8.1247)
(0.512, 8.7647)
(0.576, 9.3730)
(0.640, 9.9334)
(0.704, 10.4702)
(0.768, 11.1896)
(0.832, 11.9613)
(0.896, 12.6977)
(0.960, 13.4531)
(1.024, 14.2304)
(1.088, 14.9953)
(1.152, 15.7498)
(1.216, 16.4690)
(1.280, 17.1671)
(1.344, 17.8488)
(1.408, 18.5178)
(1.472, 19.1783)
(1.536, 19.8206)
(1.600, 20.4401)
(1.664, 21.0253)
(1.728, 21.6035)
(1.792, 22.1533)
(1.856, 22.6843)
(1.920, 23.2118)
(1.984, 23.7209)
(2.048, 24.2303)
}
\defdata{magnitudes-Config-E-1533-46-2147483-activation-norms-enc-64-max}{%
(0.064, 7.0939)
(0.128, 6.1428)
(0.192, 5.9285)
(0.256, 5.8107)
(0.320, 5.7241)
(0.384, 5.6786)
(0.448, 5.6626)
(0.512, 5.6774)
(0.576, 5.6806)
(0.640, 5.6873)
(0.704, 5.7157)
(0.768, 5.7277)
(0.832, 5.7446)
(0.896, 5.7527)
(0.960, 5.7609)
(1.024, 5.7912)
(1.088, 5.8054)
(1.152, 5.8080)
(1.216, 5.8197)
(1.280, 5.8431)
(1.344, 5.8477)
(1.408, 5.8631)
(1.472, 5.8691)
(1.536, 5.8696)
(1.600, 5.8873)
(1.664, 5.9013)
(1.728, 5.9036)
(1.792, 5.9027)
(1.856, 5.9005)
(1.920, 5.9210)
(1.984, 5.9316)
(2.048, 5.9428)
}
\defdata{magnitudes-Config-E-1533-46-2147483-activation-norms-enc-32-max}{%
(0.064, 7.0491)
(0.128, 7.5292)
(0.192, 7.8008)
(0.256, 7.8999)
(0.320, 7.9059)
(0.384, 7.8929)
(0.448, 7.8217)
(0.512, 7.7797)
(0.576, 7.7015)
(0.640, 7.6318)
(0.704, 7.6025)
(0.768, 7.5748)
(0.832, 7.5983)
(0.896, 7.5921)
(0.960, 7.5739)
(1.024, 7.5582)
(1.088, 7.5283)
(1.152, 7.5179)
(1.216, 7.5136)
(1.280, 7.4924)
(1.344, 7.4738)
(1.408, 7.5045)
(1.472, 7.4752)
(1.536, 7.4665)
(1.600, 7.4732)
(1.664, 7.4715)
(1.728, 7.4414)
(1.792, 7.4347)
(1.856, 7.4277)
(1.920, 7.4261)
(1.984, 7.4553)
(2.048, 7.4525)
}
\defdata{magnitudes-Config-E-1533-46-2147483-activation-norms-enc-16-max}{%
(0.064, 8.6652)
(0.128, 8.3103)
(0.192, 8.6356)
(0.256, 8.8680)
(0.320, 8.9111)
(0.384, 8.8716)
(0.448, 9.1875)
(0.512, 9.4743)
(0.576, 9.6496)
(0.640, 9.7930)
(0.704, 9.9259)
(0.768, 9.9962)
(0.832, 10.1077)
(0.896, 10.2112)
(0.960, 10.2961)
(1.024, 10.3258)
(1.088, 10.3730)
(1.152, 10.4642)
(1.216, 10.5465)
(1.280, 10.6118)
(1.344, 10.6695)
(1.408, 10.7582)
(1.472, 10.7882)
(1.536, 10.8741)
(1.600, 10.9081)
(1.664, 10.9664)
(1.728, 10.9863)
(1.792, 11.0208)
(1.856, 11.0756)
(1.920, 11.1185)
(1.984, 11.1694)
(2.048, 11.2138)
}
\defdata{magnitudes-Config-E-1533-46-2147483-activation-norms-enc-8-max}{%
(0.064, 11.1015)
(0.128, 10.8144)
(0.192, 10.4339)
(0.256, 10.1141)
(0.320, 9.8756)
(0.384, 9.5549)
(0.448, 9.2994)
(0.512, 8.9898)
(0.576, 8.7071)
(0.640, 8.4316)
(0.704, 8.2107)
(0.768, 7.9680)
(0.832, 7.7427)
(0.896, 7.6001)
(0.960, 7.3865)
(1.024, 7.2014)
(1.088, 7.0157)
(1.152, 6.9083)
(1.216, 6.8833)
(1.280, 6.8826)
(1.344, 6.8673)
(1.408, 6.8755)
(1.472, 6.8798)
(1.536, 6.8856)
(1.600, 6.8577)
(1.664, 6.8441)
(1.728, 6.8408)
(1.792, 6.8180)
(1.856, 6.8110)
(1.920, 6.7880)
(1.984, 6.7799)
(2.048, 6.7598)
}
\defdata{magnitudes-Config-E-1533-46-2147483-activation-norms-dec-8-max}{%
(0.064, 6.5224)
(0.128, 7.5836)
(0.192, 8.1528)
(0.256, 8.6255)
(0.320, 8.7101)
(0.384, 8.5064)
(0.448, 8.4590)
(0.512, 8.4789)
(0.576, 8.5473)
(0.640, 8.6024)
(0.704, 8.5755)
(0.768, 8.6227)
(0.832, 8.5872)
(0.896, 8.6018)
(0.960, 8.5571)
(1.024, 8.5583)
(1.088, 8.5683)
(1.152, 8.4967)
(1.216, 8.4569)
(1.280, 8.4715)
(1.344, 8.4748)
(1.408, 8.4280)
(1.472, 8.3843)
(1.536, 8.3381)
(1.600, 8.3787)
(1.664, 8.3172)
(1.728, 8.2893)
(1.792, 8.2901)
(1.856, 8.2271)
(1.920, 8.2307)
(1.984, 8.1802)
(2.048, 8.1699)
}
\defdata{magnitudes-Config-E-1533-46-2147483-activation-norms-dec-16-max}{%
(0.064, 10.6703)
(0.128, 10.4907)
(0.192, 10.6032)
(0.256, 10.6177)
(0.320, 10.6749)
(0.384, 10.7216)
(0.448, 10.8077)
(0.512, 10.8682)
(0.576, 10.9485)
(0.640, 11.0188)
(0.704, 11.1350)
(0.768, 11.2923)
(0.832, 11.4135)
(0.896, 11.5350)
(0.960, 11.6488)
(1.024, 11.7757)
(1.088, 11.8539)
(1.152, 11.9460)
(1.216, 12.0658)
(1.280, 12.1772)
(1.344, 12.2809)
(1.408, 12.3957)
(1.472, 12.4483)
(1.536, 12.4970)
(1.600, 12.6022)
(1.664, 12.6652)
(1.728, 12.7178)
(1.792, 12.7421)
(1.856, 12.7662)
(1.920, 12.8447)
(1.984, 12.8894)
(2.048, 12.9307)
}
\defdata{magnitudes-Config-E-1533-46-2147483-activation-norms-dec-32-max}{%
(0.064, 9.2301)
(0.128, 9.0365)
(0.192, 8.9796)
(0.256, 8.9679)
(0.320, 8.9075)
(0.384, 8.8329)
(0.448, 8.7246)
(0.512, 8.5961)
(0.576, 8.4326)
(0.640, 8.2057)
(0.704, 8.1140)
(0.768, 8.0792)
(0.832, 8.0751)
(0.896, 8.0402)
(0.960, 8.0172)
(1.024, 7.9694)
(1.088, 7.9180)
(1.152, 7.8837)
(1.216, 7.8680)
(1.280, 7.8209)
(1.344, 7.7722)
(1.408, 7.7640)
(1.472, 7.6818)
(1.536, 7.6425)
(1.600, 7.6070)
(1.664, 7.5549)
(1.728, 7.4868)
(1.792, 7.3951)
(1.856, 7.3298)
(1.920, 7.2546)
(1.984, 7.1942)
(2.048, 7.1158)
}
\defdata{magnitudes-Config-E-1533-46-2147483-activation-norms-dec-64-max}{%
(0.064, 9.5641)
(0.128, 9.9920)
(0.192, 9.8633)
(0.256, 9.6400)
(0.320, 9.3164)
(0.384, 9.4983)
(0.448, 9.9522)
(0.512, 9.7128)
(0.576, 9.8481)
(0.640, 9.5855)
(0.704, 9.7761)
(0.768, 9.6887)
(0.832, 9.5948)
(0.896, 9.4050)
(0.960, 9.3862)
(1.024, 9.2732)
(1.088, 9.1797)
(1.152, 9.1452)
(1.216, 9.3677)
(1.280, 9.3715)
(1.344, 9.2133)
(1.408, 9.4936)
(1.472, 9.2386)
(1.536, 8.8790)
(1.600, 9.3440)
(1.664, 9.2424)
(1.728, 9.2994)
(1.792, 9.0240)
(1.856, 9.0434)
(1.920, 9.0950)
(1.984, 9.1739)
(2.048, 9.1342)
}
\defdata{magnitudes-Config-E-1533-46-2147483-weight-norms-enc-64-max}{%
(0.064, 0.9997)
(0.128, 0.9998)
(0.192, 0.9998)
(0.256, 0.9998)
(0.320, 0.9999)
(0.384, 0.9999)
(0.448, 0.9999)
(0.512, 0.9999)
(0.576, 1.0000)
(0.640, 1.0000)
(0.704, 1.0000)
(0.768, 1.0001)
(0.832, 1.0002)
(0.896, 1.0001)
(0.960, 1.0005)
(1.024, 1.0004)
(1.088, 1.0003)
(1.152, 1.0005)
(1.216, 1.0004)
(1.280, 1.0001)
(1.344, 1.0003)
(1.408, 1.0005)
(1.472, 1.0002)
(1.536, 1.0002)
(1.600, 1.0002)
(1.664, 1.0003)
(1.728, 1.0006)
(1.792, 1.0008)
(1.856, 1.0010)
(1.920, 1.0011)
(1.984, 1.0010)
(2.048, 1.0008)
}
\defdata{magnitudes-Config-E-1533-46-2147483-weight-norms-enc-32-max}{%
(0.064, 0.9989)
(0.128, 0.9975)
(0.192, 0.9972)
(0.256, 0.9977)
(0.320, 0.9981)
(0.384, 0.9981)
(0.448, 0.9985)
(0.512, 0.9987)
(0.576, 0.9988)
(0.640, 0.9988)
(0.704, 0.9988)
(0.768, 0.9990)
(0.832, 0.9990)
(0.896, 0.9991)
(0.960, 0.9990)
(1.024, 0.9990)
(1.088, 0.9992)
(1.152, 0.9990)
(1.216, 0.9991)
(1.280, 0.9993)
(1.344, 0.9996)
(1.408, 0.9993)
(1.472, 0.9998)
(1.536, 0.9997)
(1.600, 0.9998)
(1.664, 1.0000)
(1.728, 0.9995)
(1.792, 0.9995)
(1.856, 0.9993)
(1.920, 0.9993)
(1.984, 0.9995)
(2.048, 0.9994)
}
\defdata{magnitudes-Config-E-1533-46-2147483-weight-norms-enc-16-max}{%
(0.064, 0.9985)
(0.128, 0.9964)
(0.192, 0.9949)
(0.256, 0.9960)
(0.320, 0.9968)
(0.384, 0.9973)
(0.448, 0.9972)
(0.512, 0.9976)
(0.576, 0.9978)
(0.640, 0.9977)
(0.704, 0.9979)
(0.768, 0.9977)
(0.832, 0.9978)
(0.896, 0.9977)
(0.960, 0.9978)
(1.024, 0.9981)
(1.088, 0.9980)
(1.152, 0.9981)
(1.216, 0.9979)
(1.280, 0.9980)
(1.344, 0.9982)
(1.408, 0.9980)
(1.472, 0.9985)
(1.536, 0.9984)
(1.600, 0.9987)
(1.664, 0.9988)
(1.728, 0.9985)
(1.792, 0.9986)
(1.856, 0.9985)
(1.920, 0.9983)
(1.984, 0.9982)
(2.048, 0.9984)
}
\defdata{magnitudes-Config-E-1533-46-2147483-weight-norms-enc-8-max}{%
(0.064, 0.9917)
(0.128, 0.9883)
(0.192, 0.9873)
(0.256, 0.9898)
(0.320, 0.9910)
(0.384, 0.9913)
(0.448, 0.9919)
(0.512, 0.9927)
(0.576, 0.9930)
(0.640, 0.9929)
(0.704, 0.9928)
(0.768, 0.9929)
(0.832, 0.9930)
(0.896, 0.9926)
(0.960, 0.9928)
(1.024, 0.9929)
(1.088, 0.9927)
(1.152, 0.9927)
(1.216, 0.9930)
(1.280, 0.9931)
(1.344, 0.9933)
(1.408, 0.9934)
(1.472, 0.9934)
(1.536, 0.9935)
(1.600, 0.9939)
(1.664, 0.9941)
(1.728, 0.9946)
(1.792, 0.9949)
(1.856, 0.9954)
(1.920, 0.9950)
(1.984, 0.9947)
(2.048, 0.9944)
}
\defdata{magnitudes-Config-E-1533-46-2147483-weight-norms-dec-8-max}{%
(0.064, 0.9946)
(0.128, 0.9897)
(0.192, 0.9898)
(0.256, 0.9908)
(0.320, 0.9915)
(0.384, 0.9921)
(0.448, 0.9924)
(0.512, 0.9927)
(0.576, 0.9929)
(0.640, 0.9932)
(0.704, 0.9931)
(0.768, 0.9932)
(0.832, 0.9932)
(0.896, 0.9935)
(0.960, 0.9940)
(1.024, 0.9946)
(1.088, 0.9947)
(1.152, 0.9950)
(1.216, 0.9952)
(1.280, 0.9952)
(1.344, 0.9954)
(1.408, 0.9956)
(1.472, 0.9960)
(1.536, 0.9961)
(1.600, 0.9961)
(1.664, 0.9959)
(1.728, 0.9958)
(1.792, 0.9958)
(1.856, 0.9960)
(1.920, 0.9961)
(1.984, 0.9960)
(2.048, 0.9966)
}
\defdata{magnitudes-Config-E-1533-46-2147483-weight-norms-dec-16-max}{%
(0.064, 0.9991)
(0.128, 0.9977)
(0.192, 0.9968)
(0.256, 0.9969)
(0.320, 0.9970)
(0.384, 0.9972)
(0.448, 0.9974)
(0.512, 0.9975)
(0.576, 0.9975)
(0.640, 0.9978)
(0.704, 0.9977)
(0.768, 0.9978)
(0.832, 0.9978)
(0.896, 0.9979)
(0.960, 0.9979)
(1.024, 0.9980)
(1.088, 0.9980)
(1.152, 0.9981)
(1.216, 0.9981)
(1.280, 0.9983)
(1.344, 0.9981)
(1.408, 0.9981)
(1.472, 0.9981)
(1.536, 0.9981)
(1.600, 0.9981)
(1.664, 0.9981)
(1.728, 0.9982)
(1.792, 0.9983)
(1.856, 0.9986)
(1.920, 0.9989)
(1.984, 0.9990)
(2.048, 0.9990)
}
\defdata{magnitudes-Config-E-1533-46-2147483-weight-norms-dec-32-max}{%
(0.064, 0.9990)
(0.128, 0.9980)
(0.192, 0.9978)
(0.256, 0.9977)
(0.320, 0.9979)
(0.384, 0.9981)
(0.448, 0.9982)
(0.512, 0.9983)
(0.576, 0.9983)
(0.640, 0.9983)
(0.704, 0.9984)
(0.768, 0.9984)
(0.832, 0.9984)
(0.896, 0.9983)
(0.960, 0.9982)
(1.024, 0.9983)
(1.088, 0.9984)
(1.152, 0.9984)
(1.216, 0.9984)
(1.280, 0.9984)
(1.344, 0.9983)
(1.408, 0.9982)
(1.472, 0.9983)
(1.536, 0.9983)
(1.600, 0.9984)
(1.664, 0.9984)
(1.728, 0.9983)
(1.792, 0.9983)
(1.856, 0.9984)
(1.920, 0.9984)
(1.984, 0.9983)
(2.048, 0.9983)
}
\defdata{magnitudes-Config-E-1533-46-2147483-weight-norms-dec-64-max}{%
(0.064, 0.9993)
(0.128, 0.9988)
(0.192, 0.9983)
(0.256, 0.9985)
(0.320, 0.9987)
(0.384, 0.9988)
(0.448, 0.9988)
(0.512, 0.9988)
(0.576, 0.9989)
(0.640, 0.9989)
(0.704, 0.9989)
(0.768, 0.9992)
(0.832, 0.9994)
(0.896, 0.9994)
(0.960, 0.9995)
(1.024, 0.9995)
(1.088, 0.9993)
(1.152, 0.9990)
(1.216, 0.9987)
(1.280, 0.9987)
(1.344, 0.9988)
(1.408, 0.9988)
(1.472, 0.9988)
(1.536, 0.9987)
(1.600, 0.9988)
(1.664, 0.9988)
(1.728, 0.9992)
(1.792, 0.9990)
(1.856, 0.9989)
(1.920, 0.9987)
(1.984, 0.9990)
(2.048, 0.9989)
}
\defdata{magnitudes-Config-F-1533-72-2147483-activation-norms-enc-64-max}{%
(0.064, 2.9391)
(0.128, 2.9373)
(0.192, 2.9388)
(0.256, 2.9369)
(0.320, 3.1929)
(0.384, 3.4349)
(0.448, 3.6208)
(0.512, 3.8006)
(0.576, 3.9488)
(0.640, 4.0682)
(0.704, 4.1581)
(0.768, 4.2592)
(0.832, 4.3601)
(0.896, 4.4215)
(0.960, 4.4953)
(1.024, 4.5793)
(1.088, 4.6419)
(1.152, 4.7171)
(1.216, 4.7778)
(1.280, 4.8200)
(1.344, 4.8871)
(1.408, 4.9315)
(1.472, 4.9916)
(1.536, 5.0197)
(1.600, 5.0701)
(1.664, 5.0825)
(1.728, 5.1394)
(1.792, 5.1783)
(1.856, 5.2242)
(1.920, 5.2450)
(1.984, 5.2808)
(2.048, 5.3125)
}
\defdata{magnitudes-Config-F-1533-72-2147483-activation-norms-enc-32-max}{%
(0.064, 8.4731)
(0.128, 9.2730)
(0.192, 9.3831)
(0.256, 9.3229)
(0.320, 9.2486)
(0.384, 9.1473)
(0.448, 9.0807)
(0.512, 9.0494)
(0.576, 9.0123)
(0.640, 9.0149)
(0.704, 9.0194)
(0.768, 9.0426)
(0.832, 9.0368)
(0.896, 9.0658)
(0.960, 9.0712)
(1.024, 9.0880)
(1.088, 9.1037)
(1.152, 9.1213)
(1.216, 9.1468)
(1.280, 9.1618)
(1.344, 9.1730)
(1.408, 9.1727)
(1.472, 9.1886)
(1.536, 9.2166)
(1.600, 9.2182)
(1.664, 9.2495)
(1.728, 9.2497)
(1.792, 9.2660)
(1.856, 9.2615)
(1.920, 9.2782)
(1.984, 9.2921)
(2.048, 9.2835)
}
\defdata{magnitudes-Config-F-1533-72-2147483-activation-norms-enc-16-max}{%
(0.064, 10.4583)
(0.128, 12.2367)
(0.192, 12.9365)
(0.256, 13.3144)
(0.320, 13.5837)
(0.384, 13.7398)
(0.448, 13.8536)
(0.512, 13.9613)
(0.576, 14.0295)
(0.640, 14.1407)
(0.704, 14.2571)
(0.768, 14.3525)
(0.832, 14.3927)
(0.896, 14.4978)
(0.960, 14.5553)
(1.024, 14.5680)
(1.088, 14.6376)
(1.152, 14.6758)
(1.216, 14.7155)
(1.280, 14.7856)
(1.344, 14.8291)
(1.408, 14.8240)
(1.472, 14.8843)
(1.536, 14.9255)
(1.600, 14.9553)
(1.664, 15.0259)
(1.728, 15.0475)
(1.792, 15.0593)
(1.856, 15.0823)
(1.920, 15.1200)
(1.984, 15.1485)
(2.048, 15.1535)
}
\defdata{magnitudes-Config-F-1533-72-2147483-activation-norms-enc-8-max}{%
(0.064, 17.7430)
(0.128, 18.7000)
(0.192, 20.1235)
(0.256, 20.5335)
(0.320, 20.5260)
(0.384, 20.4668)
(0.448, 20.2945)
(0.512, 20.1062)
(0.576, 19.9014)
(0.640, 19.7080)
(0.704, 19.5574)
(0.768, 19.3776)
(0.832, 19.4402)
(0.896, 19.5134)
(0.960, 19.5819)
(1.024, 19.6413)
(1.088, 19.7316)
(1.152, 19.8110)
(1.216, 19.9016)
(1.280, 19.9995)
(1.344, 20.0941)
(1.408, 20.1731)
(1.472, 20.2716)
(1.536, 20.3615)
(1.600, 20.4489)
(1.664, 20.5525)
(1.728, 20.6363)
(1.792, 20.7208)
(1.856, 20.7942)
(1.920, 20.8921)
(1.984, 20.9815)
(2.048, 21.0435)
}
\defdata{magnitudes-Config-F-1533-72-2147483-activation-norms-dec-8-max}{%
(0.064, 8.5846)
(0.128, 11.5861)
(0.192, 13.0032)
(0.256, 13.3206)
(0.320, 14.1908)
(0.384, 14.1897)
(0.448, 13.9844)
(0.512, 13.7366)
(0.576, 13.4751)
(0.640, 13.2228)
(0.704, 12.9933)
(0.768, 12.7822)
(0.832, 12.5964)
(0.896, 12.4483)
(0.960, 12.2782)
(1.024, 12.1262)
(1.088, 11.9797)
(1.152, 11.8712)
(1.216, 11.7629)
(1.280, 11.6552)
(1.344, 11.5414)
(1.408, 11.4351)
(1.472, 11.3358)
(1.536, 11.2526)
(1.600, 11.1769)
(1.664, 11.0636)
(1.728, 10.9880)
(1.792, 10.9199)
(1.856, 10.8347)
(1.920, 10.7826)
(1.984, 10.6912)
(2.048, 10.6157)
}
\defdata{magnitudes-Config-F-1533-72-2147483-activation-norms-dec-16-max}{%
(0.064, 3.6606)
(0.128, 4.0931)
(0.192, 4.5147)
(0.256, 4.9459)
(0.320, 5.1521)
(0.384, 5.6933)
(0.448, 5.9276)
(0.512, 6.2663)
(0.576, 6.6217)
(0.640, 6.8905)
(0.704, 7.1585)
(0.768, 7.3760)
(0.832, 7.5533)
(0.896, 7.7329)
(0.960, 7.8860)
(1.024, 8.0324)
(1.088, 8.1710)
(1.152, 8.2665)
(1.216, 8.3867)
(1.280, 8.4802)
(1.344, 8.5723)
(1.408, 8.6268)
(1.472, 8.6855)
(1.536, 8.7630)
(1.600, 8.7826)
(1.664, 8.8566)
(1.728, 8.8827)
(1.792, 8.9073)
(1.856, 8.9459)
(1.920, 8.9758)
(1.984, 8.9897)
(2.048, 8.9805)
}
\defdata{magnitudes-Config-F-1533-72-2147483-activation-norms-dec-32-max}{%
(0.064, 4.9507)
(0.128, 4.8096)
(0.192, 4.5990)
(0.256, 4.3650)
(0.320, 4.2105)
(0.384, 4.0714)
(0.448, 3.9255)
(0.512, 3.8923)
(0.576, 3.8357)
(0.640, 3.8004)
(0.704, 3.7625)
(0.768, 3.7809)
(0.832, 3.7702)
(0.896, 3.7447)
(0.960, 3.7214)
(1.024, 3.6753)
(1.088, 3.6657)
(1.152, 3.6556)
(1.216, 3.6223)
(1.280, 3.6083)
(1.344, 3.5552)
(1.408, 3.5178)
(1.472, 3.5179)
(1.536, 3.5070)
(1.600, 3.4585)
(1.664, 3.4374)
(1.728, 3.4288)
(1.792, 3.4155)
(1.856, 3.4085)
(1.920, 3.3841)
(1.984, 3.3475)
(2.048, 3.3495)
}
\defdata{magnitudes-Config-F-1533-72-2147483-activation-norms-dec-64-max}{%
(0.064, 2.3379)
(0.128, 2.3413)
(0.192, 2.8787)
(0.256, 3.3658)
(0.320, 3.8637)
(0.384, 4.2913)
(0.448, 4.5992)
(0.512, 4.9137)
(0.576, 5.2564)
(0.640, 5.4987)
(0.704, 5.7490)
(0.768, 5.9818)
(0.832, 6.2201)
(0.896, 6.4071)
(0.960, 6.6195)
(1.024, 6.8729)
(1.088, 7.0033)
(1.152, 7.2118)
(1.216, 7.3805)
(1.280, 7.5426)
(1.344, 7.6698)
(1.408, 7.8940)
(1.472, 8.0013)
(1.536, 8.1293)
(1.600, 8.2606)
(1.664, 8.3500)
(1.728, 8.4694)
(1.792, 8.5932)
(1.856, 8.7204)
(1.920, 8.8111)
(1.984, 8.8672)
(2.048, 9.0499)
}
\defdata{magnitudes-Config-F-1533-72-2147483-weight-norms-enc-64-max}{%
(0.064, 0.9998)
(0.128, 1.0000)
(0.192, 0.9999)
(0.256, 0.9999)
(0.320, 0.9999)
(0.384, 0.9999)
(0.448, 1.0000)
(0.512, 1.0000)
(0.576, 1.0001)
(0.640, 1.0000)
(0.704, 1.0001)
(0.768, 1.0002)
(0.832, 1.0002)
(0.896, 1.0002)
(0.960, 1.0003)
(1.024, 1.0002)
(1.088, 1.0004)
(1.152, 1.0003)
(1.216, 1.0005)
(1.280, 1.0003)
(1.344, 1.0007)
(1.408, 1.0007)
(1.472, 1.0008)
(1.536, 1.0007)
(1.600, 1.0011)
(1.664, 1.0010)
(1.728, 1.0010)
(1.792, 1.0009)
(1.856, 1.0011)
(1.920, 1.0014)
(1.984, 1.0014)
(2.048, 1.0012)
}
\defdata{magnitudes-Config-F-1533-72-2147483-weight-norms-enc-32-max}{%
(0.064, 0.9983)
(0.128, 0.9973)
(0.192, 0.9975)
(0.256, 0.9981)
(0.320, 0.9985)
(0.384, 0.9986)
(0.448, 0.9986)
(0.512, 0.9988)
(0.576, 0.9987)
(0.640, 0.9985)
(0.704, 0.9986)
(0.768, 0.9987)
(0.832, 0.9990)
(0.896, 0.9992)
(0.960, 0.9991)
(1.024, 0.9989)
(1.088, 0.9986)
(1.152, 0.9987)
(1.216, 0.9988)
(1.280, 0.9988)
(1.344, 0.9988)
(1.408, 0.9988)
(1.472, 0.9991)
(1.536, 0.9994)
(1.600, 0.9998)
(1.664, 0.9997)
(1.728, 1.0000)
(1.792, 1.0000)
(1.856, 0.9999)
(1.920, 1.0001)
(1.984, 1.0000)
(2.048, 0.9998)
}
\defdata{magnitudes-Config-F-1533-72-2147483-weight-norms-enc-16-max}{%
(0.064, 0.9987)
(0.128, 0.9984)
(0.192, 0.9980)
(0.256, 0.9979)
(0.320, 0.9987)
(0.384, 0.9989)
(0.448, 0.9987)
(0.512, 0.9989)
(0.576, 0.9987)
(0.640, 0.9983)
(0.704, 0.9983)
(0.768, 0.9991)
(0.832, 0.9991)
(0.896, 0.9985)
(0.960, 0.9981)
(1.024, 0.9982)
(1.088, 0.9984)
(1.152, 0.9985)
(1.216, 0.9983)
(1.280, 0.9984)
(1.344, 0.9989)
(1.408, 0.9999)
(1.472, 1.0002)
(1.536, 0.9992)
(1.600, 0.9992)
(1.664, 0.9986)
(1.728, 0.9987)
(1.792, 0.9989)
(1.856, 0.9986)
(1.920, 0.9986)
(1.984, 0.9987)
(2.048, 0.9989)
}
\defdata{magnitudes-Config-F-1533-72-2147483-weight-norms-enc-8-max}{%
(0.064, 0.9929)
(0.128, 0.9926)
(0.192, 0.9921)
(0.256, 0.9937)
(0.320, 0.9945)
(0.384, 0.9946)
(0.448, 0.9952)
(0.512, 0.9957)
(0.576, 0.9959)
(0.640, 0.9960)
(0.704, 0.9959)
(0.768, 0.9962)
(0.832, 0.9960)
(0.896, 0.9967)
(0.960, 0.9964)
(1.024, 0.9968)
(1.088, 0.9968)
(1.152, 0.9973)
(1.216, 0.9975)
(1.280, 0.9977)
(1.344, 0.9979)
(1.408, 0.9979)
(1.472, 0.9982)
(1.536, 0.9984)
(1.600, 0.9987)
(1.664, 0.9989)
(1.728, 0.9986)
(1.792, 0.9992)
(1.856, 0.9993)
(1.920, 0.9994)
(1.984, 0.9991)
(2.048, 0.9988)
}
\defdata{magnitudes-Config-F-1533-72-2147483-weight-norms-dec-8-max}{%
(0.064, 0.9934)
(0.128, 0.9895)
(0.192, 0.9887)
(0.256, 0.9898)
(0.320, 0.9902)
(0.384, 0.9905)
(0.448, 0.9910)
(0.512, 0.9914)
(0.576, 0.9919)
(0.640, 0.9925)
(0.704, 0.9929)
(0.768, 0.9933)
(0.832, 0.9936)
(0.896, 0.9939)
(0.960, 0.9941)
(1.024, 0.9943)
(1.088, 0.9946)
(1.152, 0.9948)
(1.216, 0.9950)
(1.280, 0.9951)
(1.344, 0.9952)
(1.408, 0.9953)
(1.472, 0.9954)
(1.536, 0.9955)
(1.600, 0.9956)
(1.664, 0.9957)
(1.728, 0.9959)
(1.792, 0.9960)
(1.856, 0.9960)
(1.920, 0.9962)
(1.984, 0.9962)
(2.048, 0.9963)
}
\defdata{magnitudes-Config-F-1533-72-2147483-weight-norms-dec-16-max}{%
(0.064, 0.9989)
(0.128, 0.9975)
(0.192, 0.9969)
(0.256, 0.9969)
(0.320, 0.9971)
(0.384, 0.9973)
(0.448, 0.9974)
(0.512, 0.9976)
(0.576, 0.9977)
(0.640, 0.9978)
(0.704, 0.9978)
(0.768, 0.9979)
(0.832, 0.9979)
(0.896, 0.9979)
(0.960, 0.9980)
(1.024, 0.9980)
(1.088, 0.9981)
(1.152, 0.9982)
(1.216, 0.9981)
(1.280, 0.9981)
(1.344, 0.9982)
(1.408, 0.9982)
(1.472, 0.9983)
(1.536, 0.9983)
(1.600, 0.9985)
(1.664, 0.9984)
(1.728, 0.9984)
(1.792, 0.9985)
(1.856, 0.9984)
(1.920, 0.9985)
(1.984, 0.9986)
(2.048, 0.9986)
}
\defdata{magnitudes-Config-F-1533-72-2147483-weight-norms-dec-32-max}{%
(0.064, 0.9990)
(0.128, 0.9979)
(0.192, 0.9976)
(0.256, 0.9976)
(0.320, 0.9978)
(0.384, 0.9977)
(0.448, 0.9979)
(0.512, 0.9980)
(0.576, 0.9982)
(0.640, 0.9981)
(0.704, 0.9982)
(0.768, 0.9983)
(0.832, 0.9983)
(0.896, 0.9983)
(0.960, 0.9983)
(1.024, 0.9985)
(1.088, 0.9986)
(1.152, 0.9986)
(1.216, 0.9985)
(1.280, 0.9984)
(1.344, 0.9985)
(1.408, 0.9984)
(1.472, 0.9984)
(1.536, 0.9987)
(1.600, 0.9988)
(1.664, 0.9987)
(1.728, 0.9989)
(1.792, 0.9992)
(1.856, 0.9995)
(1.920, 0.9998)
(1.984, 0.9998)
(2.048, 0.9994)
}
\defdata{magnitudes-Config-F-1533-72-2147483-weight-norms-dec-64-max}{%
(0.064, 0.9996)
(0.128, 0.9989)
(0.192, 0.9985)
(0.256, 0.9985)
(0.320, 0.9987)
(0.384, 0.9986)
(0.448, 0.9987)
(0.512, 0.9986)
(0.576, 0.9987)
(0.640, 0.9987)
(0.704, 0.9987)
(0.768, 0.9987)
(0.832, 0.9987)
(0.896, 0.9988)
(0.960, 0.9988)
(1.024, 0.9990)
(1.088, 0.9996)
(1.152, 1.0004)
(1.216, 1.0009)
(1.280, 1.0015)
(1.344, 1.0013)
(1.408, 1.0015)
(1.472, 1.0018)
(1.536, 1.0019)
(1.600, 1.0023)
(1.664, 1.0028)
(1.728, 1.0030)
(1.792, 1.0032)
(1.856, 1.0031)
(1.920, 1.0030)
(1.984, 1.0030)
(2.048, 1.0030)
}
\defdata{magnitudes-Config-G-1533-52-2147483-activation-norms-enc-64-max}{%
(0.064, 6.1130)
(0.128, 7.4729)
(0.192, 8.8977)
(0.256, 10.1806)
(0.320, 11.2040)
(0.384, 12.0082)
(0.448, 12.5630)
(0.512, 13.0529)
(0.576, 13.4211)
(0.640, 13.8299)
(0.704, 14.1287)
(0.768, 14.2798)
(0.832, 14.5611)
(0.896, 14.7540)
(0.960, 14.9330)
(1.024, 14.9413)
(1.088, 15.1309)
(1.152, 15.1691)
(1.216, 15.3115)
(1.280, 15.5751)
(1.344, 15.8096)
(1.408, 15.8696)
(1.472, 16.1929)
(1.536, 16.5355)
(1.600, 16.6478)
(1.664, 16.9194)
(1.728, 17.0519)
(1.792, 17.2058)
(1.856, 17.5113)
(1.920, 17.6945)
(1.984, 17.9109)
(2.048, 18.1239)
}
\defdata{magnitudes-Config-G-1533-52-2147483-activation-norms-enc-32-max}{%
(0.064, 8.2467)
(0.128, 8.7016)
(0.192, 8.6476)
(0.256, 8.7410)
(0.320, 8.8488)
(0.384, 9.9002)
(0.448, 11.1160)
(0.512, 11.9907)
(0.576, 12.6764)
(0.640, 13.3184)
(0.704, 13.8254)
(0.768, 14.1875)
(0.832, 14.5048)
(0.896, 14.8355)
(0.960, 15.0745)
(1.024, 15.2819)
(1.088, 15.5609)
(1.152, 15.7055)
(1.216, 15.9111)
(1.280, 15.9828)
(1.344, 16.0020)
(1.408, 16.1757)
(1.472, 16.2576)
(1.536, 16.3115)
(1.600, 16.3436)
(1.664, 16.3685)
(1.728, 16.3197)
(1.792, 16.3698)
(1.856, 16.4072)
(1.920, 16.4241)
(1.984, 16.5138)
(2.048, 16.6139)
}
\defdata{magnitudes-Config-G-1533-52-2147483-activation-norms-enc-16-max}{%
(0.064, 12.8119)
(0.128, 13.9705)
(0.192, 14.6957)
(0.256, 15.2877)
(0.320, 15.6840)
(0.384, 15.8795)
(0.448, 16.0496)
(0.512, 16.1753)
(0.576, 16.2685)
(0.640, 16.3646)
(0.704, 16.4236)
(0.768, 16.4731)
(0.832, 16.5295)
(0.896, 17.3268)
(0.960, 18.7712)
(1.024, 20.0183)
(1.088, 21.0943)
(1.152, 21.8601)
(1.216, 22.4462)
(1.280, 22.9262)
(1.344, 23.2730)
(1.408, 23.5066)
(1.472, 23.7466)
(1.536, 23.9939)
(1.600, 24.1039)
(1.664, 24.2565)
(1.728, 24.3511)
(1.792, 24.4439)
(1.856, 24.5707)
(1.920, 24.6441)
(1.984, 24.7215)
(2.048, 24.7597)
}
\defdata{magnitudes-Config-G-1533-52-2147483-activation-norms-enc-8-max}{%
(0.064, 17.4644)
(0.128, 19.8220)
(0.192, 20.6350)
(0.256, 20.9266)
(0.320, 21.0143)
(0.384, 21.0222)
(0.448, 21.1222)
(0.512, 21.1579)
(0.576, 21.1443)
(0.640, 21.1190)
(0.704, 21.1003)
(0.768, 21.1200)
(0.832, 21.1583)
(0.896, 21.1400)
(0.960, 21.1434)
(1.024, 21.1108)
(1.088, 21.0766)
(1.152, 21.0424)
(1.216, 20.9946)
(1.280, 21.0690)
(1.344, 21.1494)
(1.408, 21.2051)
(1.472, 21.3060)
(1.536, 21.3976)
(1.600, 21.4445)
(1.664, 21.5455)
(1.728, 21.6878)
(1.792, 22.0117)
(1.856, 22.3795)
(1.920, 22.5778)
(1.984, 22.8174)
(2.048, 23.0310)
}
\defdata{magnitudes-Config-G-1533-52-2147483-activation-norms-dec-8-max}{%
(0.064, 6.5142)
(0.128, 8.4137)
(0.192, 10.5353)
(0.256, 12.0495)
(0.320, 13.2381)
(0.384, 14.4814)
(0.448, 16.2208)
(0.512, 17.7144)
(0.576, 19.0885)
(0.640, 20.2633)
(0.704, 21.2937)
(0.768, 22.0708)
(0.832, 22.8872)
(0.896, 23.6336)
(0.960, 24.3405)
(1.024, 25.0192)
(1.088, 25.7689)
(1.152, 26.4091)
(1.216, 27.1439)
(1.280, 27.9859)
(1.344, 28.6133)
(1.408, 29.0760)
(1.472, 29.6887)
(1.536, 30.1082)
(1.600, 30.5189)
(1.664, 30.8681)
(1.728, 31.1583)
(1.792, 31.6213)
(1.856, 32.2182)
(1.920, 32.6341)
(1.984, 33.1902)
(2.048, 33.7309)
}
\defdata{magnitudes-Config-G-1533-52-2147483-activation-norms-dec-16-max}{%
(0.064, 4.9820)
(0.128, 6.7008)
(0.192, 8.3556)
(0.256, 9.6642)
(0.320, 10.4308)
(0.384, 11.2609)
(0.448, 12.0221)
(0.512, 12.8873)
(0.576, 13.5441)
(0.640, 14.2628)
(0.704, 14.9555)
(0.768, 15.5662)
(0.832, 16.2292)
(0.896, 16.8865)
(0.960, 17.4524)
(1.024, 17.9793)
(1.088, 18.6440)
(1.152, 19.1403)
(1.216, 19.7543)
(1.280, 20.4066)
(1.344, 20.9302)
(1.408, 21.3819)
(1.472, 21.9500)
(1.536, 22.4616)
(1.600, 22.8509)
(1.664, 23.2777)
(1.728, 23.6334)
(1.792, 24.0476)
(1.856, 24.4740)
(1.920, 24.8590)
(1.984, 25.2723)
(2.048, 25.6499)
}
\defdata{magnitudes-Config-G-1533-52-2147483-activation-norms-dec-32-max}{%
(0.064, 3.3567)
(0.128, 4.1181)
(0.192, 5.1725)
(0.256, 5.9634)
(0.320, 6.5174)
(0.384, 7.1224)
(0.448, 7.5309)
(0.512, 7.9478)
(0.576, 8.2887)
(0.640, 8.5530)
(0.704, 8.8694)
(0.768, 9.0896)
(0.832, 9.2690)
(0.896, 9.4633)
(0.960, 9.7064)
(1.024, 9.8952)
(1.088, 10.0842)
(1.152, 10.2226)
(1.216, 10.4289)
(1.280, 10.6280)
(1.344, 10.8036)
(1.408, 10.9402)
(1.472, 11.0693)
(1.536, 11.1700)
(1.600, 11.3111)
(1.664, 11.3738)
(1.728, 11.4892)
(1.792, 11.5773)
(1.856, 11.6646)
(1.920, 11.7362)
(1.984, 11.8080)
(2.048, 11.8935)
}
\defdata{magnitudes-Config-G-1533-52-2147483-activation-norms-dec-64-max}{%
(0.064, 4.0519)
(0.128, 4.5180)
(0.192, 5.0805)
(0.256, 6.0543)
(0.320, 8.2943)
(0.384, 9.6312)
(0.448, 10.1148)
(0.512, 10.5792)
(0.576, 10.8500)
(0.640, 10.9638)
(0.704, 11.0646)
(0.768, 11.0044)
(0.832, 10.9284)
(0.896, 10.8326)
(0.960, 10.7904)
(1.024, 11.0680)
(1.088, 11.3427)
(1.152, 11.5661)
(1.216, 11.8299)
(1.280, 12.0670)
(1.344, 12.2945)
(1.408, 12.4830)
(1.472, 12.7018)
(1.536, 12.8912)
(1.600, 13.0790)
(1.664, 13.2216)
(1.728, 13.3767)
(1.792, 13.4956)
(1.856, 13.6215)
(1.920, 13.7511)
(1.984, 13.8421)
(2.048, 13.9501)
}
\defdata{magnitudes-Config-G-1533-52-2147483-weight-norms-enc-64-max}{%
(0.064, 1.0000)
(0.128, 1.0000)
(0.192, 1.0001)
(0.256, 0.9999)
(0.320, 1.0000)
(0.384, 0.9999)
(0.448, 1.0000)
(0.512, 1.0000)
(0.576, 1.0000)
(0.640, 1.0001)
(0.704, 1.0002)
(0.768, 1.0002)
(0.832, 1.0003)
(0.896, 1.0003)
(0.960, 1.0002)
(1.024, 1.0004)
(1.088, 1.0006)
(1.152, 1.0010)
(1.216, 1.0008)
(1.280, 1.0010)
(1.344, 1.0011)
(1.408, 1.0012)
(1.472, 1.0014)
(1.536, 1.0014)
(1.600, 1.0014)
(1.664, 1.0013)
(1.728, 1.0014)
(1.792, 1.0014)
(1.856, 1.0014)
(1.920, 1.0018)
(1.984, 1.0020)
(2.048, 1.0021)
}
\defdata{magnitudes-Config-G-1533-52-2147483-weight-norms-enc-32-max}{%
(0.064, 0.9983)
(0.128, 0.9974)
(0.192, 0.9989)
(0.256, 0.9992)
(0.320, 0.9994)
(0.384, 0.9994)
(0.448, 0.9994)
(0.512, 0.9995)
(0.576, 0.9996)
(0.640, 0.9996)
(0.704, 0.9996)
(0.768, 0.9999)
(0.832, 1.0000)
(0.896, 0.9996)
(0.960, 0.9992)
(1.024, 0.9992)
(1.088, 0.9992)
(1.152, 0.9992)
(1.216, 0.9999)
(1.280, 0.9998)
(1.344, 1.0002)
(1.408, 1.0010)
(1.472, 1.0001)
(1.536, 0.9998)
(1.600, 1.0000)
(1.664, 1.0005)
(1.728, 1.0000)
(1.792, 1.0003)
(1.856, 1.0004)
(1.920, 1.0006)
(1.984, 1.0007)
(2.048, 1.0001)
}
\defdata{magnitudes-Config-G-1533-52-2147483-weight-norms-enc-16-max}{%
(0.064, 0.9989)
(0.128, 0.9981)
(0.192, 0.9976)
(0.256, 0.9981)
(0.320, 0.9983)
(0.384, 0.9990)
(0.448, 0.9987)
(0.512, 0.9987)
(0.576, 0.9990)
(0.640, 0.9988)
(0.704, 0.9990)
(0.768, 0.9984)
(0.832, 0.9980)
(0.896, 0.9984)
(0.960, 0.9996)
(1.024, 0.9990)
(1.088, 0.9986)
(1.152, 0.9984)
(1.216, 0.9986)
(1.280, 0.9983)
(1.344, 0.9987)
(1.408, 0.9988)
(1.472, 0.9987)
(1.536, 0.9988)
(1.600, 0.9983)
(1.664, 0.9987)
(1.728, 0.9985)
(1.792, 0.9986)
(1.856, 0.9985)
(1.920, 0.9982)
(1.984, 0.9982)
(2.048, 0.9985)
}
\defdata{magnitudes-Config-G-1533-52-2147483-weight-norms-enc-8-max}{%
(0.064, 0.9935)
(0.128, 0.9948)
(0.192, 0.9972)
(0.256, 0.9972)
(0.320, 0.9972)
(0.384, 0.9974)
(0.448, 0.9974)
(0.512, 0.9978)
(0.576, 0.9975)
(0.640, 0.9979)
(0.704, 0.9975)
(0.768, 0.9973)
(0.832, 0.9971)
(0.896, 0.9970)
(0.960, 0.9971)
(1.024, 0.9971)
(1.088, 0.9972)
(1.152, 0.9972)
(1.216, 0.9973)
(1.280, 0.9973)
(1.344, 0.9974)
(1.408, 0.9974)
(1.472, 0.9974)
(1.536, 0.9974)
(1.600, 0.9974)
(1.664, 0.9974)
(1.728, 0.9973)
(1.792, 0.9973)
(1.856, 0.9972)
(1.920, 0.9972)
(1.984, 0.9971)
(2.048, 0.9971)
}
\defdata{magnitudes-Config-G-1533-52-2147483-weight-norms-dec-8-max}{%
(0.064, 0.9947)
(0.128, 0.9899)
(0.192, 0.9893)
(0.256, 0.9903)
(0.320, 0.9908)
(0.384, 0.9912)
(0.448, 0.9920)
(0.512, 0.9925)
(0.576, 0.9929)
(0.640, 0.9932)
(0.704, 0.9935)
(0.768, 0.9939)
(0.832, 0.9940)
(0.896, 0.9943)
(0.960, 0.9944)
(1.024, 0.9945)
(1.088, 0.9947)
(1.152, 0.9948)
(1.216, 0.9950)
(1.280, 0.9952)
(1.344, 0.9952)
(1.408, 0.9953)
(1.472, 0.9954)
(1.536, 0.9955)
(1.600, 0.9956)
(1.664, 0.9957)
(1.728, 0.9958)
(1.792, 0.9958)
(1.856, 0.9959)
(1.920, 0.9960)
(1.984, 0.9961)
(2.048, 0.9961)
}
\defdata{magnitudes-Config-G-1533-52-2147483-weight-norms-dec-16-max}{%
(0.064, 0.9987)
(0.128, 0.9977)
(0.192, 0.9971)
(0.256, 0.9972)
(0.320, 0.9975)
(0.384, 0.9976)
(0.448, 0.9977)
(0.512, 0.9977)
(0.576, 0.9977)
(0.640, 0.9977)
(0.704, 0.9979)
(0.768, 0.9978)
(0.832, 0.9980)
(0.896, 0.9980)
(0.960, 0.9981)
(1.024, 0.9981)
(1.088, 0.9982)
(1.152, 0.9982)
(1.216, 0.9982)
(1.280, 0.9982)
(1.344, 0.9982)
(1.408, 0.9983)
(1.472, 0.9982)
(1.536, 0.9983)
(1.600, 0.9983)
(1.664, 0.9985)
(1.728, 0.9984)
(1.792, 0.9983)
(1.856, 0.9983)
(1.920, 0.9984)
(1.984, 0.9984)
(2.048, 0.9985)
}
\defdata{magnitudes-Config-G-1533-52-2147483-weight-norms-dec-32-max}{%
(0.064, 0.9987)
(0.128, 0.9975)
(0.192, 0.9971)
(0.256, 0.9975)
(0.320, 0.9976)
(0.384, 0.9980)
(0.448, 0.9981)
(0.512, 0.9983)
(0.576, 0.9983)
(0.640, 0.9983)
(0.704, 0.9984)
(0.768, 0.9985)
(0.832, 0.9986)
(0.896, 0.9989)
(0.960, 0.9987)
(1.024, 0.9986)
(1.088, 0.9989)
(1.152, 0.9988)
(1.216, 0.9986)
(1.280, 0.9986)
(1.344, 0.9986)
(1.408, 0.9988)
(1.472, 0.9987)
(1.536, 0.9987)
(1.600, 0.9990)
(1.664, 0.9992)
(1.728, 0.9992)
(1.792, 0.9994)
(1.856, 0.9991)
(1.920, 0.9992)
(1.984, 0.9991)
(2.048, 0.9988)
}
\defdata{magnitudes-Config-G-1533-52-2147483-weight-norms-dec-64-max}{%
(0.064, 0.9996)
(0.128, 0.9991)
(0.192, 0.9988)
(0.256, 0.9989)
(0.320, 0.9990)
(0.384, 0.9990)
(0.448, 0.9991)
(0.512, 0.9991)
(0.576, 0.9992)
(0.640, 0.9992)
(0.704, 0.9992)
(0.768, 0.9992)
(0.832, 0.9992)
(0.896, 0.9993)
(0.960, 0.9992)
(1.024, 0.9993)
(1.088, 0.9992)
(1.152, 0.9992)
(1.216, 0.9993)
(1.280, 0.9993)
(1.344, 0.9992)
(1.408, 0.9992)
(1.472, 0.9993)
(1.536, 0.9999)
(1.600, 0.9996)
(1.664, 0.9994)
(1.728, 0.9997)
(1.792, 0.9996)
(1.856, 0.9993)
(1.920, 0.9993)
(1.984, 0.9996)
(2.048, 0.9997)
}
\defdata{magnitudes-Config-B-1532-29-2147483-activation-norms-enc-64-mean}{%
(0.064, 0.7045)
(0.128, 0.8405)
(0.192, 0.9445)
(0.256, 1.0349)
(0.320, 1.1115)
(0.384, 1.1789)
(0.448, 1.2399)
(0.512, 1.2944)
(0.576, 1.3429)
(0.640, 1.3886)
(0.704, 1.4328)
(0.768, 1.4736)
(0.832, 1.5080)
(0.896, 1.5486)
(0.960, 1.5843)
(1.024, 1.6181)
(1.088, 1.6483)
(1.152, 1.6792)
(1.216, 1.7073)
(1.280, 1.7368)
(1.344, 1.7670)
(1.408, 1.7928)
(1.472, 1.8252)
(1.536, 1.8484)
(1.600, 1.8726)
(1.664, 1.9009)
(1.728, 1.9224)
(1.792, 1.9468)
(1.856, 1.9743)
(1.920, 1.9970)
(1.984, 2.0221)
(2.048, 2.0435)
}
\defdata{magnitudes-Config-B-1532-29-2147483-activation-norms-enc-32-mean}{%
(0.064, 0.9318)
(0.128, 1.2356)
(0.192, 1.4603)
(0.256, 1.6529)
(0.320, 1.8093)
(0.384, 1.9450)
(0.448, 2.0641)
(0.512, 2.1682)
(0.576, 2.2583)
(0.640, 2.3422)
(0.704, 2.4198)
(0.768, 2.4920)
(0.832, 2.5499)
(0.896, 2.6206)
(0.960, 2.6821)
(1.024, 2.7406)
(1.088, 2.7935)
(1.152, 2.8450)
(1.216, 2.8911)
(1.280, 2.9401)
(1.344, 2.9939)
(1.408, 3.0374)
(1.472, 3.0931)
(1.536, 3.1321)
(1.600, 3.1720)
(1.664, 3.2166)
(1.728, 3.2550)
(1.792, 3.2944)
(1.856, 3.3407)
(1.920, 3.3793)
(1.984, 3.4207)
(2.048, 3.4577)
}
\defdata{magnitudes-Config-B-1532-29-2147483-activation-norms-enc-16-mean}{%
(0.064, 1.3805)
(0.128, 1.7210)
(0.192, 1.9478)
(0.256, 2.1323)
(0.320, 2.2764)
(0.384, 2.3992)
(0.448, 2.5023)
(0.512, 2.5936)
(0.576, 2.6686)
(0.640, 2.7401)
(0.704, 2.8099)
(0.768, 2.8737)
(0.832, 2.9192)
(0.896, 2.9837)
(0.960, 3.0364)
(1.024, 3.0873)
(1.088, 3.1331)
(1.152, 3.1754)
(1.216, 3.2161)
(1.280, 3.2588)
(1.344, 3.3113)
(1.408, 3.3495)
(1.472, 3.4010)
(1.536, 3.4334)
(1.600, 3.4668)
(1.664, 3.5099)
(1.728, 3.5422)
(1.792, 3.5770)
(1.856, 3.6201)
(1.920, 3.6557)
(1.984, 3.6932)
(2.048, 3.7255)
}
\defdata{magnitudes-Config-B-1532-29-2147483-activation-norms-enc-8-mean}{%
(0.064, 1.9094)
(0.128, 2.3488)
(0.192, 2.6671)
(0.256, 2.9508)
(0.320, 3.2041)
(0.384, 3.4377)
(0.448, 3.6438)
(0.512, 3.8389)
(0.576, 4.0081)
(0.640, 4.1734)
(0.704, 4.3262)
(0.768, 4.4654)
(0.832, 4.5974)
(0.896, 4.7195)
(0.960, 4.8369)
(1.024, 4.9461)
(1.088, 5.0518)
(1.152, 5.1484)
(1.216, 5.2404)
(1.280, 5.3316)
(1.344, 5.4187)
(1.408, 5.5028)
(1.472, 5.5793)
(1.536, 5.6546)
(1.600, 5.7267)
(1.664, 5.8015)
(1.728, 5.8684)
(1.792, 5.9366)
(1.856, 5.9944)
(1.920, 6.0609)
(1.984, 6.1188)
(2.048, 6.1804)
}
\defdata{magnitudes-Config-B-1532-29-2147483-activation-norms-dec-8-mean}{%
(0.064, 2.2715)
(0.128, 2.8241)
(0.192, 3.2300)
(0.256, 3.5881)
(0.320, 3.9110)
(0.384, 4.1985)
(0.448, 4.4597)
(0.512, 4.7033)
(0.576, 4.9301)
(0.640, 5.1515)
(0.704, 5.3515)
(0.768, 5.5443)
(0.832, 5.7358)
(0.896, 5.9040)
(0.960, 6.0726)
(1.024, 6.2400)
(1.088, 6.3988)
(1.152, 6.5519)
(1.216, 6.7059)
(1.280, 6.8506)
(1.344, 6.9910)
(1.408, 7.1286)
(1.472, 7.2528)
(1.536, 7.3911)
(1.600, 7.5147)
(1.664, 7.6454)
(1.728, 7.7673)
(1.792, 7.8835)
(1.856, 8.0058)
(1.920, 8.1207)
(1.984, 8.2313)
(2.048, 8.3381)
}
\defdata{magnitudes-Config-B-1532-29-2147483-activation-norms-dec-16-mean}{%
(0.064, 1.8804)
(0.128, 2.4429)
(0.192, 2.8565)
(0.256, 3.2062)
(0.320, 3.5052)
(0.384, 3.7695)
(0.448, 4.0024)
(0.512, 4.2180)
(0.576, 4.4179)
(0.640, 4.6111)
(0.704, 4.7903)
(0.768, 4.9599)
(0.832, 5.1308)
(0.896, 5.2880)
(0.960, 5.4377)
(1.024, 5.5933)
(1.088, 5.7387)
(1.152, 5.8808)
(1.216, 6.0225)
(1.280, 6.1594)
(1.344, 6.3068)
(1.408, 6.4332)
(1.472, 6.5665)
(1.536, 6.6978)
(1.600, 6.8206)
(1.664, 6.9535)
(1.728, 7.0768)
(1.792, 7.1922)
(1.856, 7.3173)
(1.920, 7.4384)
(1.984, 7.5546)
(2.048, 7.6731)
}
\defdata{magnitudes-Config-B-1532-29-2147483-activation-norms-dec-32-mean}{%
(0.064, 1.4186)
(0.128, 1.8953)
(0.192, 2.2443)
(0.256, 2.5331)
(0.320, 2.7762)
(0.384, 2.9950)
(0.448, 3.1895)
(0.512, 3.3734)
(0.576, 3.5462)
(0.640, 3.7197)
(0.704, 3.8869)
(0.768, 4.0456)
(0.832, 4.2007)
(0.896, 4.3551)
(0.960, 4.4980)
(1.024, 4.6533)
(1.088, 4.7916)
(1.152, 4.9327)
(1.216, 5.0711)
(1.280, 5.2087)
(1.344, 5.3461)
(1.408, 5.4792)
(1.472, 5.6125)
(1.536, 5.7423)
(1.600, 5.8655)
(1.664, 5.9918)
(1.728, 6.1124)
(1.792, 6.2213)
(1.856, 6.3553)
(1.920, 6.4609)
(1.984, 6.5829)
(2.048, 6.6923)
}
\defdata{magnitudes-Config-B-1532-29-2147483-activation-norms-dec-64-mean}{%
(0.064, 0.8173)
(0.128, 1.0800)
(0.192, 1.2966)
(0.256, 1.4925)
(0.320, 1.6697)
(0.384, 1.8370)
(0.448, 1.9955)
(0.512, 2.1512)
(0.576, 2.2974)
(0.640, 2.4471)
(0.704, 2.5871)
(0.768, 2.7242)
(0.832, 2.8574)
(0.896, 2.9916)
(0.960, 3.1238)
(1.024, 3.2552)
(1.088, 3.3841)
(1.152, 3.5112)
(1.216, 3.6415)
(1.280, 3.7680)
(1.344, 3.8828)
(1.408, 4.0147)
(1.472, 4.1302)
(1.536, 4.2576)
(1.600, 4.3753)
(1.664, 4.4902)
(1.728, 4.6056)
(1.792, 4.7202)
(1.856, 4.8426)
(1.920, 4.9444)
(1.984, 5.0640)
(2.048, 5.1741)
}
\defdata{magnitudes-Config-B-1532-29-2147483-weight-norms-enc-64-mean}{%
(0.064, 0.0240)
(0.128, 0.0280)
(0.192, 0.0315)
(0.256, 0.0346)
(0.320, 0.0373)
(0.384, 0.0399)
(0.448, 0.0424)
(0.512, 0.0447)
(0.576, 0.0469)
(0.640, 0.0491)
(0.704, 0.0512)
(0.768, 0.0532)
(0.832, 0.0552)
(0.896, 0.0571)
(0.960, 0.0590)
(1.024, 0.0609)
(1.088, 0.0627)
(1.152, 0.0645)
(1.216, 0.0662)
(1.280, 0.0679)
(1.344, 0.0696)
(1.408, 0.0713)
(1.472, 0.0729)
(1.536, 0.0745)
(1.600, 0.0761)
(1.664, 0.0776)
(1.728, 0.0791)
(1.792, 0.0807)
(1.856, 0.0821)
(1.920, 0.0836)
(1.984, 0.0851)
(2.048, 0.0865)
}
\defdata{magnitudes-Config-B-1532-29-2147483-weight-norms-enc-32-mean}{%
(0.064, 0.0215)
(0.128, 0.0275)
(0.192, 0.0326)
(0.256, 0.0370)
(0.320, 0.0408)
(0.384, 0.0443)
(0.448, 0.0476)
(0.512, 0.0506)
(0.576, 0.0534)
(0.640, 0.0562)
(0.704, 0.0587)
(0.768, 0.0612)
(0.832, 0.0636)
(0.896, 0.0659)
(0.960, 0.0681)
(1.024, 0.0703)
(1.088, 0.0724)
(1.152, 0.0744)
(1.216, 0.0764)
(1.280, 0.0783)
(1.344, 0.0802)
(1.408, 0.0820)
(1.472, 0.0838)
(1.536, 0.0856)
(1.600, 0.0873)
(1.664, 0.0890)
(1.728, 0.0906)
(1.792, 0.0923)
(1.856, 0.0939)
(1.920, 0.0955)
(1.984, 0.0970)
(2.048, 0.0986)
}
\defdata{magnitudes-Config-B-1532-29-2147483-weight-norms-enc-16-mean}{%
(0.064, 0.0285)
(0.128, 0.0377)
(0.192, 0.0451)
(0.256, 0.0515)
(0.320, 0.0573)
(0.384, 0.0625)
(0.448, 0.0674)
(0.512, 0.0719)
(0.576, 0.0762)
(0.640, 0.0803)
(0.704, 0.0842)
(0.768, 0.0880)
(0.832, 0.0916)
(0.896, 0.0951)
(0.960, 0.0984)
(1.024, 0.1017)
(1.088, 0.1048)
(1.152, 0.1079)
(1.216, 0.1109)
(1.280, 0.1138)
(1.344, 0.1167)
(1.408, 0.1195)
(1.472, 0.1222)
(1.536, 0.1248)
(1.600, 0.1275)
(1.664, 0.1300)
(1.728, 0.1325)
(1.792, 0.1350)
(1.856, 0.1374)
(1.920, 0.1398)
(1.984, 0.1422)
(2.048, 0.1445)
}
\defdata{magnitudes-Config-B-1532-29-2147483-weight-norms-enc-8-mean}{%
(0.064, 0.0310)
(0.128, 0.0419)
(0.192, 0.0509)
(0.256, 0.0587)
(0.320, 0.0656)
(0.384, 0.0719)
(0.448, 0.0777)
(0.512, 0.0832)
(0.576, 0.0884)
(0.640, 0.0933)
(0.704, 0.0980)
(0.768, 0.1024)
(0.832, 0.1067)
(0.896, 0.1109)
(0.960, 0.1149)
(1.024, 0.1187)
(1.088, 0.1225)
(1.152, 0.1262)
(1.216, 0.1297)
(1.280, 0.1332)
(1.344, 0.1366)
(1.408, 0.1399)
(1.472, 0.1431)
(1.536, 0.1463)
(1.600, 0.1494)
(1.664, 0.1524)
(1.728, 0.1554)
(1.792, 0.1583)
(1.856, 0.1612)
(1.920, 0.1640)
(1.984, 0.1668)
(2.048, 0.1695)
}
\defdata{magnitudes-Config-B-1532-29-2147483-weight-norms-dec-8-mean}{%
(0.064, 0.0318)
(0.128, 0.0434)
(0.192, 0.0527)
(0.256, 0.0608)
(0.320, 0.0679)
(0.384, 0.0745)
(0.448, 0.0806)
(0.512, 0.0863)
(0.576, 0.0916)
(0.640, 0.0967)
(0.704, 0.1016)
(0.768, 0.1063)
(0.832, 0.1108)
(0.896, 0.1151)
(0.960, 0.1193)
(1.024, 0.1234)
(1.088, 0.1273)
(1.152, 0.1312)
(1.216, 0.1349)
(1.280, 0.1385)
(1.344, 0.1421)
(1.408, 0.1456)
(1.472, 0.1490)
(1.536, 0.1523)
(1.600, 0.1555)
(1.664, 0.1587)
(1.728, 0.1619)
(1.792, 0.1650)
(1.856, 0.1680)
(1.920, 0.1710)
(1.984, 0.1739)
(2.048, 0.1768)
}
\defdata{magnitudes-Config-B-1532-29-2147483-weight-norms-dec-16-mean}{%
(0.064, 0.0290)
(0.128, 0.0388)
(0.192, 0.0466)
(0.256, 0.0534)
(0.320, 0.0595)
(0.384, 0.0651)
(0.448, 0.0703)
(0.512, 0.0752)
(0.576, 0.0799)
(0.640, 0.0843)
(0.704, 0.0886)
(0.768, 0.0927)
(0.832, 0.0967)
(0.896, 0.1006)
(0.960, 0.1043)
(1.024, 0.1080)
(1.088, 0.1115)
(1.152, 0.1150)
(1.216, 0.1183)
(1.280, 0.1216)
(1.344, 0.1248)
(1.408, 0.1280)
(1.472, 0.1311)
(1.536, 0.1341)
(1.600, 0.1371)
(1.664, 0.1400)
(1.728, 0.1429)
(1.792, 0.1457)
(1.856, 0.1485)
(1.920, 0.1512)
(1.984, 0.1539)
(2.048, 0.1566)
}
\defdata{magnitudes-Config-B-1532-29-2147483-weight-norms-dec-32-mean}{%
(0.064, 0.0214)
(0.128, 0.0283)
(0.192, 0.0340)
(0.256, 0.0390)
(0.320, 0.0434)
(0.384, 0.0475)
(0.448, 0.0513)
(0.512, 0.0549)
(0.576, 0.0583)
(0.640, 0.0616)
(0.704, 0.0647)
(0.768, 0.0678)
(0.832, 0.0707)
(0.896, 0.0735)
(0.960, 0.0763)
(1.024, 0.0790)
(1.088, 0.0816)
(1.152, 0.0842)
(1.216, 0.0867)
(1.280, 0.0891)
(1.344, 0.0915)
(1.408, 0.0939)
(1.472, 0.0962)
(1.536, 0.0984)
(1.600, 0.1007)
(1.664, 0.1028)
(1.728, 0.1050)
(1.792, 0.1071)
(1.856, 0.1092)
(1.920, 0.1113)
(1.984, 0.1133)
(2.048, 0.1153)
}
\defdata{magnitudes-Config-B-1532-29-2147483-weight-norms-dec-64-mean}{%
(0.064, 0.0199)
(0.128, 0.0250)
(0.192, 0.0295)
(0.256, 0.0335)
(0.320, 0.0372)
(0.384, 0.0406)
(0.448, 0.0438)
(0.512, 0.0469)
(0.576, 0.0498)
(0.640, 0.0527)
(0.704, 0.0555)
(0.768, 0.0582)
(0.832, 0.0608)
(0.896, 0.0634)
(0.960, 0.0659)
(1.024, 0.0684)
(1.088, 0.0708)
(1.152, 0.0732)
(1.216, 0.0756)
(1.280, 0.0779)
(1.344, 0.0802)
(1.408, 0.0825)
(1.472, 0.0847)
(1.536, 0.0869)
(1.600, 0.0891)
(1.664, 0.0912)
(1.728, 0.0934)
(1.792, 0.0955)
(1.856, 0.0976)
(1.920, 0.0996)
(1.984, 0.1017)
(2.048, 0.1037)
}
\defdata{magnitudes-Config-C-1533-41-2147483-activation-norms-enc-64-mean}{%
(0.064, 0.6703)
(0.128, 0.7588)
(0.192, 0.8340)
(0.256, 0.9051)
(0.320, 0.9680)
(0.384, 1.0311)
(0.448, 1.0868)
(0.512, 1.1422)
(0.576, 1.2004)
(0.640, 1.2472)
(0.704, 1.2968)
(0.768, 1.3447)
(0.832, 1.3871)
(0.896, 1.4328)
(0.960, 1.4737)
(1.024, 1.5181)
(1.088, 1.5551)
(1.152, 1.5993)
(1.216, 1.6329)
(1.280, 1.6704)
(1.344, 1.7039)
(1.408, 1.7408)
(1.472, 1.7832)
(1.536, 1.8065)
(1.600, 1.8422)
(1.664, 1.8737)
(1.728, 1.9084)
(1.792, 1.9329)
(1.856, 1.9658)
(1.920, 1.9996)
(1.984, 2.0313)
(2.048, 2.0560)
}
\defdata{magnitudes-Config-C-1533-41-2147483-activation-norms-enc-32-mean}{%
(0.064, 0.7136)
(0.128, 0.9498)
(0.192, 1.1452)
(0.256, 1.3204)
(0.320, 1.4698)
(0.384, 1.6183)
(0.448, 1.7446)
(0.512, 1.8695)
(0.576, 2.0041)
(0.640, 2.1061)
(0.704, 2.2178)
(0.768, 2.3262)
(0.832, 2.4170)
(0.896, 2.5210)
(0.960, 2.6107)
(1.024, 2.7141)
(1.088, 2.7957)
(1.152, 2.9021)
(1.216, 2.9724)
(1.280, 3.0568)
(1.344, 3.1311)
(1.408, 3.2149)
(1.472, 3.3033)
(1.536, 3.3592)
(1.600, 3.4438)
(1.664, 3.5160)
(1.728, 3.5944)
(1.792, 3.6401)
(1.856, 3.7205)
(1.920, 3.8028)
(1.984, 3.8756)
(2.048, 3.9260)
}
\defdata{magnitudes-Config-C-1533-41-2147483-activation-norms-enc-16-mean}{%
(0.064, 1.2818)
(0.128, 1.8771)
(0.192, 2.3746)
(0.256, 2.8299)
(0.320, 3.2414)
(0.384, 3.6381)
(0.448, 4.0110)
(0.512, 4.3692)
(0.576, 4.7353)
(0.640, 5.0627)
(0.704, 5.3995)
(0.768, 5.7263)
(0.832, 6.0366)
(0.896, 6.3441)
(0.960, 6.6415)
(1.024, 6.9683)
(1.088, 7.2377)
(1.152, 7.5465)
(1.216, 7.8145)
(1.280, 8.0992)
(1.344, 8.3657)
(1.408, 8.6457)
(1.472, 9.0138)
(1.536, 9.1752)
(1.600, 9.4529)
(1.664, 9.7220)
(1.728, 9.9685)
(1.792, 10.1968)
(1.856, 10.4637)
(1.920, 10.7254)
(1.984, 10.9856)
(2.048, 11.2021)
}
\defdata{magnitudes-Config-C-1533-41-2147483-activation-norms-enc-8-mean}{%
(0.064, 2.2669)
(0.128, 3.3582)
(0.192, 4.2234)
(0.256, 4.9850)
(0.320, 5.6911)
(0.384, 6.3140)
(0.448, 6.9532)
(0.512, 7.5043)
(0.576, 8.0622)
(0.640, 8.6177)
(0.704, 9.1553)
(0.768, 9.6753)
(0.832, 10.2049)
(0.896, 10.6475)
(0.960, 11.1224)
(1.024, 11.6312)
(1.088, 12.0695)
(1.152, 12.5355)
(1.216, 12.9761)
(1.280, 13.4596)
(1.344, 13.8396)
(1.408, 14.2923)
(1.472, 14.8927)
(1.536, 15.1843)
(1.600, 15.6689)
(1.664, 16.0953)
(1.728, 16.4781)
(1.792, 16.8995)
(1.856, 17.3067)
(1.920, 17.6927)
(1.984, 18.0739)
(2.048, 18.5127)
}
\defdata{magnitudes-Config-C-1533-41-2147483-activation-norms-dec-8-mean}{%
(0.064, 2.9309)
(0.128, 4.3454)
(0.192, 5.5070)
(0.256, 6.5788)
(0.320, 7.5946)
(0.384, 8.5289)
(0.448, 9.5102)
(0.512, 10.3919)
(0.576, 11.2633)
(0.640, 12.1829)
(0.704, 13.0861)
(0.768, 13.9495)
(0.832, 14.8582)
(0.896, 15.6547)
(0.960, 16.5265)
(1.024, 17.3720)
(1.088, 18.2288)
(1.152, 19.0563)
(1.216, 19.9465)
(1.280, 20.8054)
(1.344, 21.6249)
(1.408, 22.4890)
(1.472, 23.4807)
(1.536, 24.2302)
(1.600, 25.1503)
(1.664, 26.0125)
(1.728, 26.7924)
(1.792, 27.7039)
(1.856, 28.5592)
(1.920, 29.3404)
(1.984, 30.1104)
(2.048, 30.9938)
}
\defdata{magnitudes-Config-C-1533-41-2147483-activation-norms-dec-16-mean}{%
(0.064, 1.7547)
(0.128, 2.5631)
(0.192, 3.2660)
(0.256, 3.9242)
(0.320, 4.5521)
(0.384, 5.1501)
(0.448, 5.7602)
(0.512, 6.3301)
(0.576, 6.8907)
(0.640, 7.4632)
(0.704, 8.0279)
(0.768, 8.5804)
(0.832, 9.1370)
(0.896, 9.6639)
(0.960, 10.2047)
(1.024, 10.7495)
(1.088, 11.2802)
(1.152, 11.7952)
(1.216, 12.3377)
(1.280, 12.8744)
(1.344, 13.3983)
(1.408, 13.9148)
(1.472, 14.5361)
(1.536, 14.9709)
(1.600, 15.5126)
(1.664, 16.0176)
(1.728, 16.5204)
(1.792, 17.0487)
(1.856, 17.5982)
(1.920, 18.0805)
(1.984, 18.6051)
(2.048, 19.0931)
}
\defdata{magnitudes-Config-C-1533-41-2147483-activation-norms-dec-32-mean}{%
(0.064, 1.2212)
(0.128, 1.6671)
(0.192, 2.0266)
(0.256, 2.3389)
(0.320, 2.6152)
(0.384, 2.8700)
(0.448, 3.1071)
(0.512, 3.3272)
(0.576, 3.5377)
(0.640, 3.7360)
(0.704, 3.9256)
(0.768, 4.1122)
(0.832, 4.2850)
(0.896, 4.4527)
(0.960, 4.6150)
(1.024, 4.7793)
(1.088, 4.9293)
(1.152, 5.0814)
(1.216, 5.2250)
(1.280, 5.3707)
(1.344, 5.5060)
(1.408, 5.6488)
(1.472, 5.8009)
(1.536, 5.9115)
(1.600, 6.0420)
(1.664, 6.1629)
(1.728, 6.2873)
(1.792, 6.4024)
(1.856, 6.5306)
(1.920, 6.6477)
(1.984, 6.7679)
(2.048, 6.8776)
}
\defdata{magnitudes-Config-C-1533-41-2147483-activation-norms-dec-64-mean}{%
(0.064, 0.8669)
(0.128, 1.1026)
(0.192, 1.3060)
(0.256, 1.4932)
(0.320, 1.6614)
(0.384, 1.8247)
(0.448, 1.9794)
(0.512, 2.1259)
(0.576, 2.2725)
(0.640, 2.4083)
(0.704, 2.5415)
(0.768, 2.6742)
(0.832, 2.8021)
(0.896, 2.9263)
(0.960, 3.0489)
(1.024, 3.1746)
(1.088, 3.2897)
(1.152, 3.4159)
(1.216, 3.5228)
(1.280, 3.6370)
(1.344, 3.7435)
(1.408, 3.8575)
(1.472, 3.9769)
(1.536, 4.0637)
(1.600, 4.1741)
(1.664, 4.2769)
(1.728, 4.3821)
(1.792, 4.4791)
(1.856, 4.5812)
(1.920, 4.6826)
(1.984, 4.7777)
(2.048, 4.8722)
}
\defdata{magnitudes-Config-C-1533-41-2147483-weight-norms-enc-64-mean}{%
(0.064, 0.0264)
(0.128, 0.0307)
(0.192, 0.0345)
(0.256, 0.0380)
(0.320, 0.0413)
(0.384, 0.0443)
(0.448, 0.0472)
(0.512, 0.0500)
(0.576, 0.0526)
(0.640, 0.0552)
(0.704, 0.0577)
(0.768, 0.0600)
(0.832, 0.0623)
(0.896, 0.0646)
(0.960, 0.0668)
(1.024, 0.0689)
(1.088, 0.0710)
(1.152, 0.0730)
(1.216, 0.0750)
(1.280, 0.0769)
(1.344, 0.0788)
(1.408, 0.0807)
(1.472, 0.0825)
(1.536, 0.0843)
(1.600, 0.0861)
(1.664, 0.0878)
(1.728, 0.0895)
(1.792, 0.0912)
(1.856, 0.0929)
(1.920, 0.0945)
(1.984, 0.0961)
(2.048, 0.0977)
}
\defdata{magnitudes-Config-C-1533-41-2147483-weight-norms-enc-32-mean}{%
(0.064, 0.0234)
(0.128, 0.0303)
(0.192, 0.0359)
(0.256, 0.0408)
(0.320, 0.0452)
(0.384, 0.0492)
(0.448, 0.0529)
(0.512, 0.0564)
(0.576, 0.0596)
(0.640, 0.0627)
(0.704, 0.0656)
(0.768, 0.0684)
(0.832, 0.0711)
(0.896, 0.0737)
(0.960, 0.0762)
(1.024, 0.0786)
(1.088, 0.0810)
(1.152, 0.0832)
(1.216, 0.0855)
(1.280, 0.0876)
(1.344, 0.0897)
(1.408, 0.0918)
(1.472, 0.0938)
(1.536, 0.0958)
(1.600, 0.0977)
(1.664, 0.0996)
(1.728, 0.1015)
(1.792, 0.1033)
(1.856, 0.1051)
(1.920, 0.1069)
(1.984, 0.1086)
(2.048, 0.1103)
}
\defdata{magnitudes-Config-C-1533-41-2147483-weight-norms-enc-16-mean}{%
(0.064, 0.0306)
(0.128, 0.0408)
(0.192, 0.0493)
(0.256, 0.0566)
(0.320, 0.0632)
(0.384, 0.0692)
(0.448, 0.0747)
(0.512, 0.0799)
(0.576, 0.0848)
(0.640, 0.0894)
(0.704, 0.0939)
(0.768, 0.0981)
(0.832, 0.1022)
(0.896, 0.1061)
(0.960, 0.1099)
(1.024, 0.1135)
(1.088, 0.1171)
(1.152, 0.1205)
(1.216, 0.1239)
(1.280, 0.1272)
(1.344, 0.1304)
(1.408, 0.1335)
(1.472, 0.1365)
(1.536, 0.1395)
(1.600, 0.1424)
(1.664, 0.1453)
(1.728, 0.1481)
(1.792, 0.1509)
(1.856, 0.1536)
(1.920, 0.1563)
(1.984, 0.1589)
(2.048, 0.1615)
}
\defdata{magnitudes-Config-C-1533-41-2147483-weight-norms-enc-8-mean}{%
(0.064, 0.0333)
(0.128, 0.0452)
(0.192, 0.0545)
(0.256, 0.0625)
(0.320, 0.0696)
(0.384, 0.0760)
(0.448, 0.0819)
(0.512, 0.0875)
(0.576, 0.0927)
(0.640, 0.0976)
(0.704, 0.1023)
(0.768, 0.1069)
(0.832, 0.1112)
(0.896, 0.1154)
(0.960, 0.1194)
(1.024, 0.1233)
(1.088, 0.1271)
(1.152, 0.1308)
(1.216, 0.1344)
(1.280, 0.1379)
(1.344, 0.1413)
(1.408, 0.1447)
(1.472, 0.1480)
(1.536, 0.1512)
(1.600, 0.1543)
(1.664, 0.1574)
(1.728, 0.1604)
(1.792, 0.1634)
(1.856, 0.1663)
(1.920, 0.1691)
(1.984, 0.1720)
(2.048, 0.1747)
}
\defdata{magnitudes-Config-C-1533-41-2147483-weight-norms-dec-8-mean}{%
(0.064, 0.0333)
(0.128, 0.0456)
(0.192, 0.0553)
(0.256, 0.0635)
(0.320, 0.0709)
(0.384, 0.0776)
(0.448, 0.0838)
(0.512, 0.0896)
(0.576, 0.0951)
(0.640, 0.1004)
(0.704, 0.1053)
(0.768, 0.1101)
(0.832, 0.1147)
(0.896, 0.1192)
(0.960, 0.1235)
(1.024, 0.1276)
(1.088, 0.1317)
(1.152, 0.1356)
(1.216, 0.1395)
(1.280, 0.1432)
(1.344, 0.1469)
(1.408, 0.1504)
(1.472, 0.1539)
(1.536, 0.1574)
(1.600, 0.1608)
(1.664, 0.1641)
(1.728, 0.1673)
(1.792, 0.1705)
(1.856, 0.1736)
(1.920, 0.1767)
(1.984, 0.1798)
(2.048, 0.1828)
}
\defdata{magnitudes-Config-C-1533-41-2147483-weight-norms-dec-16-mean}{%
(0.064, 0.0303)
(0.128, 0.0410)
(0.192, 0.0499)
(0.256, 0.0577)
(0.320, 0.0646)
(0.384, 0.0711)
(0.448, 0.0770)
(0.512, 0.0826)
(0.576, 0.0880)
(0.640, 0.0930)
(0.704, 0.0979)
(0.768, 0.1025)
(0.832, 0.1070)
(0.896, 0.1114)
(0.960, 0.1156)
(1.024, 0.1197)
(1.088, 0.1236)
(1.152, 0.1275)
(1.216, 0.1313)
(1.280, 0.1350)
(1.344, 0.1386)
(1.408, 0.1421)
(1.472, 0.1456)
(1.536, 0.1490)
(1.600, 0.1523)
(1.664, 0.1556)
(1.728, 0.1588)
(1.792, 0.1620)
(1.856, 0.1651)
(1.920, 0.1682)
(1.984, 0.1712)
(2.048, 0.1742)
}
\defdata{magnitudes-Config-C-1533-41-2147483-weight-norms-dec-32-mean}{%
(0.064, 0.0224)
(0.128, 0.0296)
(0.192, 0.0353)
(0.256, 0.0403)
(0.320, 0.0447)
(0.384, 0.0487)
(0.448, 0.0525)
(0.512, 0.0560)
(0.576, 0.0593)
(0.640, 0.0624)
(0.704, 0.0655)
(0.768, 0.0684)
(0.832, 0.0711)
(0.896, 0.0738)
(0.960, 0.0764)
(1.024, 0.0790)
(1.088, 0.0814)
(1.152, 0.0838)
(1.216, 0.0861)
(1.280, 0.0884)
(1.344, 0.0907)
(1.408, 0.0929)
(1.472, 0.0950)
(1.536, 0.0971)
(1.600, 0.0992)
(1.664, 0.1012)
(1.728, 0.1032)
(1.792, 0.1052)
(1.856, 0.1071)
(1.920, 0.1090)
(1.984, 0.1109)
(2.048, 0.1127)
}
\defdata{magnitudes-Config-C-1533-41-2147483-weight-norms-dec-64-mean}{%
(0.064, 0.0219)
(0.128, 0.0272)
(0.192, 0.0319)
(0.256, 0.0360)
(0.320, 0.0399)
(0.384, 0.0435)
(0.448, 0.0470)
(0.512, 0.0503)
(0.576, 0.0534)
(0.640, 0.0565)
(0.704, 0.0595)
(0.768, 0.0624)
(0.832, 0.0652)
(0.896, 0.0680)
(0.960, 0.0707)
(1.024, 0.0733)
(1.088, 0.0759)
(1.152, 0.0785)
(1.216, 0.0810)
(1.280, 0.0834)
(1.344, 0.0859)
(1.408, 0.0883)
(1.472, 0.0907)
(1.536, 0.0930)
(1.600, 0.0953)
(1.664, 0.0976)
(1.728, 0.0998)
(1.792, 0.1020)
(1.856, 0.1043)
(1.920, 0.1064)
(1.984, 0.1086)
(2.048, 0.1107)
}
\defdata{magnitudes-Config-D-1533-45-2147483-activation-norms-enc-64-mean}{%
(0.064, 0.7860)
(0.128, 0.7197)
(0.192, 0.6800)
(0.256, 0.6524)
(0.320, 0.6319)
(0.384, 0.6154)
(0.448, 0.6025)
(0.512, 0.5904)
(0.576, 0.5813)
(0.640, 0.5721)
(0.704, 0.5654)
(0.768, 0.5585)
(0.832, 0.5509)
(0.896, 0.5456)
(0.960, 0.5402)
(1.024, 0.5350)
(1.088, 0.5305)
(1.152, 0.5260)
(1.216, 0.5227)
(1.280, 0.5179)
(1.344, 0.5141)
(1.408, 0.5100)
(1.472, 0.5071)
(1.536, 0.5030)
(1.600, 0.5006)
(1.664, 0.4975)
(1.728, 0.4943)
(1.792, 0.4923)
(1.856, 0.4895)
(1.920, 0.4872)
(1.984, 0.4841)
(2.048, 0.4812)
}
\defdata{magnitudes-Config-D-1533-45-2147483-activation-norms-enc-32-mean}{%
(0.064, 0.6882)
(0.128, 0.6461)
(0.192, 0.6152)
(0.256, 0.5946)
(0.320, 0.5796)
(0.384, 0.5678)
(0.448, 0.5588)
(0.512, 0.5500)
(0.576, 0.5436)
(0.640, 0.5367)
(0.704, 0.5323)
(0.768, 0.5276)
(0.832, 0.5210)
(0.896, 0.5176)
(0.960, 0.5133)
(1.024, 0.5094)
(1.088, 0.5061)
(1.152, 0.5024)
(1.216, 0.5015)
(1.280, 0.4971)
(1.344, 0.4939)
(1.408, 0.4913)
(1.472, 0.4894)
(1.536, 0.4847)
(1.600, 0.4836)
(1.664, 0.4820)
(1.728, 0.4795)
(1.792, 0.4782)
(1.856, 0.4767)
(1.920, 0.4741)
(1.984, 0.4710)
(2.048, 0.4692)
}
\defdata{magnitudes-Config-D-1533-45-2147483-activation-norms-enc-16-mean}{%
(0.064, 1.0397)
(0.128, 0.9260)
(0.192, 0.8546)
(0.256, 0.8063)
(0.320, 0.7710)
(0.384, 0.7439)
(0.448, 0.7231)
(0.512, 0.7032)
(0.576, 0.6877)
(0.640, 0.6734)
(0.704, 0.6616)
(0.768, 0.6510)
(0.832, 0.6392)
(0.896, 0.6314)
(0.960, 0.6223)
(1.024, 0.6144)
(1.088, 0.6074)
(1.152, 0.6004)
(1.216, 0.5954)
(1.280, 0.5881)
(1.344, 0.5821)
(1.408, 0.5770)
(1.472, 0.5722)
(1.536, 0.5652)
(1.600, 0.5617)
(1.664, 0.5580)
(1.728, 0.5535)
(1.792, 0.5497)
(1.856, 0.5465)
(1.920, 0.5418)
(1.984, 0.5376)
(2.048, 0.5337)
}
\defdata{magnitudes-Config-D-1533-45-2147483-activation-norms-enc-8-mean}{%
(0.064, 1.2726)
(0.128, 1.1451)
(0.192, 1.0679)
(0.256, 1.0137)
(0.320, 0.9727)
(0.384, 0.9420)
(0.448, 0.9170)
(0.512, 0.8953)
(0.576, 0.8758)
(0.640, 0.8605)
(0.704, 0.8465)
(0.768, 0.8329)
(0.832, 0.8224)
(0.896, 0.8133)
(0.960, 0.8030)
(1.024, 0.7946)
(1.088, 0.7864)
(1.152, 0.7792)
(1.216, 0.7711)
(1.280, 0.7654)
(1.344, 0.7595)
(1.408, 0.7538)
(1.472, 0.7484)
(1.536, 0.7428)
(1.600, 0.7382)
(1.664, 0.7339)
(1.728, 0.7288)
(1.792, 0.7243)
(1.856, 0.7212)
(1.920, 0.7166)
(1.984, 0.7135)
(2.048, 0.7098)
}
\defdata{magnitudes-Config-D-1533-45-2147483-activation-norms-dec-8-mean}{%
(0.064, 1.5546)
(0.128, 1.3606)
(0.192, 1.2442)
(0.256, 1.1668)
(0.320, 1.1082)
(0.384, 1.0670)
(0.448, 1.0316)
(0.512, 1.0032)
(0.576, 0.9776)
(0.640, 0.9582)
(0.704, 0.9398)
(0.768, 0.9219)
(0.832, 0.9091)
(0.896, 0.8981)
(0.960, 0.8855)
(1.024, 0.8747)
(1.088, 0.8649)
(1.152, 0.8571)
(1.216, 0.8482)
(1.280, 0.8414)
(1.344, 0.8344)
(1.408, 0.8296)
(1.472, 0.8223)
(1.536, 0.8167)
(1.600, 0.8116)
(1.664, 0.8068)
(1.728, 0.8013)
(1.792, 0.7958)
(1.856, 0.7929)
(1.920, 0.7878)
(1.984, 0.7850)
(2.048, 0.7808)
}
\defdata{magnitudes-Config-D-1533-45-2147483-activation-norms-dec-16-mean}{%
(0.064, 1.3660)
(0.128, 1.1643)
(0.192, 1.0575)
(0.256, 0.9881)
(0.320, 0.9385)
(0.384, 0.9026)
(0.448, 0.8727)
(0.512, 0.8481)
(0.576, 0.8270)
(0.640, 0.8092)
(0.704, 0.7931)
(0.768, 0.7788)
(0.832, 0.7668)
(0.896, 0.7569)
(0.960, 0.7453)
(1.024, 0.7353)
(1.088, 0.7269)
(1.152, 0.7190)
(1.216, 0.7116)
(1.280, 0.7039)
(1.344, 0.6970)
(1.408, 0.6917)
(1.472, 0.6852)
(1.536, 0.6786)
(1.600, 0.6735)
(1.664, 0.6692)
(1.728, 0.6635)
(1.792, 0.6587)
(1.856, 0.6552)
(1.920, 0.6498)
(1.984, 0.6454)
(2.048, 0.6416)
}
\defdata{magnitudes-Config-D-1533-45-2147483-activation-norms-dec-32-mean}{%
(0.064, 0.9863)
(0.128, 0.9277)
(0.192, 0.8856)
(0.256, 0.8538)
(0.320, 0.8288)
(0.384, 0.8092)
(0.448, 0.7927)
(0.512, 0.7772)
(0.576, 0.7651)
(0.640, 0.7540)
(0.704, 0.7433)
(0.768, 0.7347)
(0.832, 0.7259)
(0.896, 0.7192)
(0.960, 0.7115)
(1.024, 0.7047)
(1.088, 0.6989)
(1.152, 0.6935)
(1.216, 0.6883)
(1.280, 0.6826)
(1.344, 0.6775)
(1.408, 0.6740)
(1.472, 0.6689)
(1.536, 0.6643)
(1.600, 0.6603)
(1.664, 0.6577)
(1.728, 0.6538)
(1.792, 0.6501)
(1.856, 0.6474)
(1.920, 0.6434)
(1.984, 0.6400)
(2.048, 0.6372)
}
\defdata{magnitudes-Config-D-1533-45-2147483-activation-norms-dec-64-mean}{%
(0.064, 1.0522)
(0.128, 0.9721)
(0.192, 0.9267)
(0.256, 0.8913)
(0.320, 0.8658)
(0.384, 0.8428)
(0.448, 0.8258)
(0.512, 0.8088)
(0.576, 0.7953)
(0.640, 0.7820)
(0.704, 0.7708)
(0.768, 0.7598)
(0.832, 0.7488)
(0.896, 0.7411)
(0.960, 0.7315)
(1.024, 0.7236)
(1.088, 0.7172)
(1.152, 0.7099)
(1.216, 0.7034)
(1.280, 0.6971)
(1.344, 0.6912)
(1.408, 0.6849)
(1.472, 0.6804)
(1.536, 0.6749)
(1.600, 0.6704)
(1.664, 0.6664)
(1.728, 0.6612)
(1.792, 0.6585)
(1.856, 0.6543)
(1.920, 0.6507)
(1.984, 0.6461)
(2.048, 0.6424)
}
\defdata{magnitudes-Config-D-1533-45-2147483-weight-norms-enc-64-mean}{%
(0.064, 1.2447)
(0.128, 1.5326)
(0.192, 1.8081)
(0.256, 2.0692)
(0.320, 2.3177)
(0.384, 2.5555)
(0.448, 2.7839)
(0.512, 3.0039)
(0.576, 3.2168)
(0.640, 3.4235)
(0.704, 3.6245)
(0.768, 3.8204)
(0.832, 4.0116)
(0.896, 4.1983)
(0.960, 4.3812)
(1.024, 4.5604)
(1.088, 4.7363)
(1.152, 4.9090)
(1.216, 5.0789)
(1.280, 5.2459)
(1.344, 5.4104)
(1.408, 5.5722)
(1.472, 5.7318)
(1.536, 5.8891)
(1.600, 6.0443)
(1.664, 6.1973)
(1.728, 6.3484)
(1.792, 6.4977)
(1.856, 6.6454)
(1.920, 6.7914)
(1.984, 6.9357)
(2.048, 7.0784)
}
\defdata{magnitudes-Config-D-1533-45-2147483-weight-norms-enc-32-mean}{%
(0.064, 1.3541)
(0.128, 1.7992)
(0.192, 2.2185)
(0.256, 2.6079)
(0.320, 2.9722)
(0.384, 3.3162)
(0.448, 3.6431)
(0.512, 3.9556)
(0.576, 4.2560)
(0.640, 4.5455)
(0.704, 4.8255)
(0.768, 5.0972)
(0.832, 5.3616)
(0.896, 5.6190)
(0.960, 5.8704)
(1.024, 6.1163)
(1.088, 6.3568)
(1.152, 6.5925)
(1.216, 6.8238)
(1.280, 7.0507)
(1.344, 7.2738)
(1.408, 7.4931)
(1.472, 7.7088)
(1.536, 7.9212)
(1.600, 8.1304)
(1.664, 8.3367)
(1.728, 8.5401)
(1.792, 8.7409)
(1.856, 8.9391)
(1.920, 9.1350)
(1.984, 9.3286)
(2.048, 9.5200)
}
\defdata{magnitudes-Config-D-1533-45-2147483-weight-norms-enc-16-mean}{%
(0.064, 1.5490)
(0.128, 2.1239)
(0.192, 2.6141)
(0.256, 3.0479)
(0.320, 3.4421)
(0.384, 3.8065)
(0.448, 4.1474)
(0.512, 4.4693)
(0.576, 4.7754)
(0.640, 5.0679)
(0.704, 5.3486)
(0.768, 5.6190)
(0.832, 5.8804)
(0.896, 6.1335)
(0.960, 6.3794)
(1.024, 6.6185)
(1.088, 6.8515)
(1.152, 7.0790)
(1.216, 7.3014)
(1.280, 7.5189)
(1.344, 7.7320)
(1.408, 7.9410)
(1.472, 8.1462)
(1.536, 8.3476)
(1.600, 8.5457)
(1.664, 8.7405)
(1.728, 8.9323)
(1.792, 9.1212)
(1.856, 9.3074)
(1.920, 9.4909)
(1.984, 9.6718)
(2.048, 9.8504)
}
\defdata{magnitudes-Config-D-1533-45-2147483-weight-norms-enc-8-mean}{%
(0.064, 1.7599)
(0.128, 2.3994)
(0.192, 2.9122)
(0.256, 3.3521)
(0.320, 3.7436)
(0.384, 4.1000)
(0.448, 4.4294)
(0.512, 4.7372)
(0.576, 5.0273)
(0.640, 5.3023)
(0.704, 5.5644)
(0.768, 5.8154)
(0.832, 6.0566)
(0.896, 6.2890)
(0.960, 6.5137)
(1.024, 6.7312)
(1.088, 6.9422)
(1.152, 7.1474)
(1.216, 7.3472)
(1.280, 7.5420)
(1.344, 7.7321)
(1.408, 7.9180)
(1.472, 8.0998)
(1.536, 8.2779)
(1.600, 8.4524)
(1.664, 8.6235)
(1.728, 8.7915)
(1.792, 8.9565)
(1.856, 9.1187)
(1.920, 9.2782)
(1.984, 9.4352)
(2.048, 9.5897)
}
\defdata{magnitudes-Config-D-1533-45-2147483-weight-norms-dec-8-mean}{%
(0.064, 1.7310)
(0.128, 2.3518)
(0.192, 2.8492)
(0.256, 3.2756)
(0.320, 3.6549)
(0.384, 4.0001)
(0.448, 4.3191)
(0.512, 4.6173)
(0.576, 4.8983)
(0.640, 5.1648)
(0.704, 5.4190)
(0.768, 5.6624)
(0.832, 5.8963)
(0.896, 6.1219)
(0.960, 6.3398)
(1.024, 6.5510)
(1.088, 6.7559)
(1.152, 6.9552)
(1.216, 7.1492)
(1.280, 7.3383)
(1.344, 7.5230)
(1.408, 7.7034)
(1.472, 7.8800)
(1.536, 8.0529)
(1.600, 8.2224)
(1.664, 8.3886)
(1.728, 8.5518)
(1.792, 8.7120)
(1.856, 8.8696)
(1.920, 9.0245)
(1.984, 9.1769)
(2.048, 9.3269)
}
\defdata{magnitudes-Config-D-1533-45-2147483-weight-norms-dec-16-mean}{%
(0.064, 1.5001)
(0.128, 2.0110)
(0.192, 2.4467)
(0.256, 2.8324)
(0.320, 3.1827)
(0.384, 3.5063)
(0.448, 3.8086)
(0.512, 4.0937)
(0.576, 4.3645)
(0.640, 4.6230)
(0.704, 4.8708)
(0.768, 5.1093)
(0.832, 5.3396)
(0.896, 5.5624)
(0.960, 5.7785)
(1.024, 5.9886)
(1.088, 6.1931)
(1.152, 6.3925)
(1.216, 6.5872)
(1.280, 6.7776)
(1.344, 6.9638)
(1.408, 7.1462)
(1.472, 7.3251)
(1.536, 7.5006)
(1.600, 7.6730)
(1.664, 7.8424)
(1.728, 8.0090)
(1.792, 8.1729)
(1.856, 8.3343)
(1.920, 8.4933)
(1.984, 8.6500)
(2.048, 8.8044)
}
\defdata{magnitudes-Config-D-1533-45-2147483-weight-norms-dec-32-mean}{%
(0.064, 1.3655)
(0.128, 1.7699)
(0.192, 2.1344)
(0.256, 2.4680)
(0.320, 2.7786)
(0.384, 3.0714)
(0.448, 3.3492)
(0.512, 3.6145)
(0.576, 3.8692)
(0.640, 4.1147)
(0.704, 4.3519)
(0.768, 4.5817)
(0.832, 4.8049)
(0.896, 5.0220)
(0.960, 5.2337)
(1.024, 5.4403)
(1.088, 5.6423)
(1.152, 5.8400)
(1.216, 6.0335)
(1.280, 6.2232)
(1.344, 6.4093)
(1.408, 6.5920)
(1.472, 6.7715)
(1.536, 6.9481)
(1.600, 7.1219)
(1.664, 7.2929)
(1.728, 7.4613)
(1.792, 7.6273)
(1.856, 7.7910)
(1.920, 7.9524)
(1.984, 8.1116)
(2.048, 8.2688)
}
\defdata{magnitudes-Config-D-1533-45-2147483-weight-norms-dec-64-mean}{%
(0.064, 1.2850)
(0.128, 1.5740)
(0.192, 1.8416)
(0.256, 2.0926)
(0.320, 2.3310)
(0.384, 2.5586)
(0.448, 2.7773)
(0.512, 2.9887)
(0.576, 3.1935)
(0.640, 3.3926)
(0.704, 3.5867)
(0.768, 3.7758)
(0.832, 3.9604)
(0.896, 4.1408)
(0.960, 4.3173)
(1.024, 4.4904)
(1.088, 4.6601)
(1.152, 4.8267)
(1.216, 4.9903)
(1.280, 5.1510)
(1.344, 5.3091)
(1.408, 5.4645)
(1.472, 5.6176)
(1.536, 5.7684)
(1.600, 5.9171)
(1.664, 6.0636)
(1.728, 6.2082)
(1.792, 6.3510)
(1.856, 6.4919)
(1.920, 6.6311)
(1.984, 6.7687)
(2.048, 6.9046)
}
\defdata{magnitudes-Config-E-1533-46-2147483-activation-norms-enc-64-mean}{%
(0.064, 0.7738)
(0.128, 0.7072)
(0.192, 0.6715)
(0.256, 0.6451)
(0.320, 0.6239)
(0.384, 0.6078)
(0.448, 0.5963)
(0.512, 0.5847)
(0.576, 0.5765)
(0.640, 0.5682)
(0.704, 0.5626)
(0.768, 0.5570)
(0.832, 0.5517)
(0.896, 0.5467)
(0.960, 0.5422)
(1.024, 0.5376)
(1.088, 0.5339)
(1.152, 0.5299)
(1.216, 0.5271)
(1.280, 0.5244)
(1.344, 0.5208)
(1.408, 0.5185)
(1.472, 0.5152)
(1.536, 0.5118)
(1.600, 0.5109)
(1.664, 0.5080)
(1.728, 0.5058)
(1.792, 0.5035)
(1.856, 0.5012)
(1.920, 0.4994)
(1.984, 0.4974)
(2.048, 0.4957)
}
\defdata{magnitudes-Config-E-1533-46-2147483-activation-norms-enc-32-mean}{%
(0.064, 0.6872)
(0.128, 0.6442)
(0.192, 0.6200)
(0.256, 0.6021)
(0.320, 0.5863)
(0.384, 0.5755)
(0.448, 0.5684)
(0.512, 0.5604)
(0.576, 0.5551)
(0.640, 0.5484)
(0.704, 0.5448)
(0.768, 0.5409)
(0.832, 0.5365)
(0.896, 0.5331)
(0.960, 0.5299)
(1.024, 0.5264)
(1.088, 0.5230)
(1.152, 0.5207)
(1.216, 0.5197)
(1.280, 0.5182)
(1.344, 0.5154)
(1.408, 0.5142)
(1.472, 0.5108)
(1.536, 0.5089)
(1.600, 0.5092)
(1.664, 0.5074)
(1.728, 0.5052)
(1.792, 0.5048)
(1.856, 0.5024)
(1.920, 0.5015)
(1.984, 0.5002)
(2.048, 0.4999)
}
\defdata{magnitudes-Config-E-1533-46-2147483-activation-norms-enc-16-mean}{%
(0.064, 1.0827)
(0.128, 0.9734)
(0.192, 0.9017)
(0.256, 0.8517)
(0.320, 0.8140)
(0.384, 0.7873)
(0.448, 0.7672)
(0.512, 0.7489)
(0.576, 0.7346)
(0.640, 0.7205)
(0.704, 0.7095)
(0.768, 0.6992)
(0.832, 0.6887)
(0.896, 0.6800)
(0.960, 0.6724)
(1.024, 0.6642)
(1.088, 0.6567)
(1.152, 0.6504)
(1.216, 0.6449)
(1.280, 0.6395)
(1.344, 0.6335)
(1.408, 0.6279)
(1.472, 0.6216)
(1.536, 0.6163)
(1.600, 0.6128)
(1.664, 0.6083)
(1.728, 0.6033)
(1.792, 0.6002)
(1.856, 0.5952)
(1.920, 0.5916)
(1.984, 0.5877)
(2.048, 0.5857)
}
\defdata{magnitudes-Config-E-1533-46-2147483-activation-norms-enc-8-mean}{%
(0.064, 1.4200)
(0.128, 1.3373)
(0.192, 1.2551)
(0.256, 1.1749)
(0.320, 1.1127)
(0.384, 1.0650)
(0.448, 1.0267)
(0.512, 0.9961)
(0.576, 0.9687)
(0.640, 0.9476)
(0.704, 0.9272)
(0.768, 0.9100)
(0.832, 0.8941)
(0.896, 0.8806)
(0.960, 0.8683)
(1.024, 0.8568)
(1.088, 0.8478)
(1.152, 0.8387)
(1.216, 0.8283)
(1.280, 0.8204)
(1.344, 0.8129)
(1.408, 0.8051)
(1.472, 0.7988)
(1.536, 0.7920)
(1.600, 0.7855)
(1.664, 0.7791)
(1.728, 0.7728)
(1.792, 0.7673)
(1.856, 0.7621)
(1.920, 0.7558)
(1.984, 0.7497)
(2.048, 0.7441)
}
\defdata{magnitudes-Config-E-1533-46-2147483-activation-norms-dec-8-mean}{%
(0.064, 1.7517)
(0.128, 1.6326)
(0.192, 1.5203)
(0.256, 1.4105)
(0.320, 1.3324)
(0.384, 1.2716)
(0.448, 1.2217)
(0.512, 1.1824)
(0.576, 1.1474)
(0.640, 1.1183)
(0.704, 1.0906)
(0.768, 1.0686)
(0.832, 1.0479)
(0.896, 1.0306)
(0.960, 1.0133)
(1.024, 0.9970)
(1.088, 0.9857)
(1.152, 0.9743)
(1.216, 0.9607)
(1.280, 0.9500)
(1.344, 0.9385)
(1.408, 0.9303)
(1.472, 0.9216)
(1.536, 0.9147)
(1.600, 0.9050)
(1.664, 0.8976)
(1.728, 0.8908)
(1.792, 0.8843)
(1.856, 0.8785)
(1.920, 0.8707)
(1.984, 0.8646)
(2.048, 0.8584)
}
\defdata{magnitudes-Config-E-1533-46-2147483-activation-norms-dec-16-mean}{%
(0.064, 1.4293)
(0.128, 1.2463)
(0.192, 1.1388)
(0.256, 1.0618)
(0.320, 1.0103)
(0.384, 0.9722)
(0.448, 0.9414)
(0.512, 0.9176)
(0.576, 0.8966)
(0.640, 0.8784)
(0.704, 0.8619)
(0.768, 0.8477)
(0.832, 0.8352)
(0.896, 0.8238)
(0.960, 0.8142)
(1.024, 0.8031)
(1.088, 0.7944)
(1.152, 0.7879)
(1.216, 0.7793)
(1.280, 0.7715)
(1.344, 0.7635)
(1.408, 0.7570)
(1.472, 0.7512)
(1.536, 0.7462)
(1.600, 0.7387)
(1.664, 0.7343)
(1.728, 0.7287)
(1.792, 0.7243)
(1.856, 0.7200)
(1.920, 0.7146)
(1.984, 0.7105)
(2.048, 0.7065)
}
\defdata{magnitudes-Config-E-1533-46-2147483-activation-norms-dec-32-mean}{%
(0.064, 1.0041)
(0.128, 0.9472)
(0.192, 0.9015)
(0.256, 0.8658)
(0.320, 0.8391)
(0.384, 0.8190)
(0.448, 0.8023)
(0.512, 0.7886)
(0.576, 0.7764)
(0.640, 0.7653)
(0.704, 0.7560)
(0.768, 0.7475)
(0.832, 0.7396)
(0.896, 0.7330)
(0.960, 0.7269)
(1.024, 0.7208)
(1.088, 0.7146)
(1.152, 0.7108)
(1.216, 0.7065)
(1.280, 0.7016)
(1.344, 0.6970)
(1.408, 0.6929)
(1.472, 0.6894)
(1.536, 0.6860)
(1.600, 0.6821)
(1.664, 0.6797)
(1.728, 0.6762)
(1.792, 0.6740)
(1.856, 0.6712)
(1.920, 0.6684)
(1.984, 0.6661)
(2.048, 0.6641)
}
\defdata{magnitudes-Config-E-1533-46-2147483-activation-norms-dec-64-mean}{%
(0.064, 1.0603)
(0.128, 0.9935)
(0.192, 0.9457)
(0.256, 0.9110)
(0.320, 0.8819)
(0.384, 0.8598)
(0.448, 0.8424)
(0.512, 0.8242)
(0.576, 0.8111)
(0.640, 0.7987)
(0.704, 0.7870)
(0.768, 0.7779)
(0.832, 0.7680)
(0.896, 0.7607)
(0.960, 0.7527)
(1.024, 0.7454)
(1.088, 0.7396)
(1.152, 0.7331)
(1.216, 0.7280)
(1.280, 0.7222)
(1.344, 0.7169)
(1.408, 0.7116)
(1.472, 0.7070)
(1.536, 0.7011)
(1.600, 0.6985)
(1.664, 0.6942)
(1.728, 0.6900)
(1.792, 0.6866)
(1.856, 0.6826)
(1.920, 0.6797)
(1.984, 0.6761)
(2.048, 0.6733)
}
\defdata{magnitudes-Config-E-1533-46-2147483-weight-norms-enc-64-mean}{%
(0.064, 0.9943)
(0.128, 0.9911)
(0.192, 0.9901)
(0.256, 0.9907)
(0.320, 0.9911)
(0.384, 0.9915)
(0.448, 0.9918)
(0.512, 0.9921)
(0.576, 0.9923)
(0.640, 0.9923)
(0.704, 0.9924)
(0.768, 0.9925)
(0.832, 0.9926)
(0.896, 0.9927)
(0.960, 0.9928)
(1.024, 0.9929)
(1.088, 0.9929)
(1.152, 0.9930)
(1.216, 0.9930)
(1.280, 0.9930)
(1.344, 0.9930)
(1.408, 0.9931)
(1.472, 0.9931)
(1.536, 0.9931)
(1.600, 0.9931)
(1.664, 0.9931)
(1.728, 0.9930)
(1.792, 0.9930)
(1.856, 0.9930)
(1.920, 0.9930)
(1.984, 0.9930)
(2.048, 0.9930)
}
\defdata{magnitudes-Config-E-1533-46-2147483-weight-norms-enc-32-mean}{%
(0.064, 0.9903)
(0.128, 0.9821)
(0.192, 0.9796)
(0.256, 0.9807)
(0.320, 0.9816)
(0.384, 0.9824)
(0.448, 0.9831)
(0.512, 0.9837)
(0.576, 0.9842)
(0.640, 0.9845)
(0.704, 0.9848)
(0.768, 0.9851)
(0.832, 0.9854)
(0.896, 0.9857)
(0.960, 0.9859)
(1.024, 0.9862)
(1.088, 0.9864)
(1.152, 0.9867)
(1.216, 0.9868)
(1.280, 0.9870)
(1.344, 0.9872)
(1.408, 0.9874)
(1.472, 0.9875)
(1.536, 0.9876)
(1.600, 0.9877)
(1.664, 0.9878)
(1.728, 0.9879)
(1.792, 0.9880)
(1.856, 0.9880)
(1.920, 0.9881)
(1.984, 0.9882)
(2.048, 0.9882)
}
\defdata{magnitudes-Config-E-1533-46-2147483-weight-norms-enc-16-mean}{%
(0.064, 0.9839)
(0.128, 0.9720)
(0.192, 0.9683)
(0.256, 0.9697)
(0.320, 0.9711)
(0.384, 0.9722)
(0.448, 0.9732)
(0.512, 0.9741)
(0.576, 0.9748)
(0.640, 0.9755)
(0.704, 0.9760)
(0.768, 0.9765)
(0.832, 0.9769)
(0.896, 0.9773)
(0.960, 0.9776)
(1.024, 0.9780)
(1.088, 0.9783)
(1.152, 0.9787)
(1.216, 0.9790)
(1.280, 0.9792)
(1.344, 0.9795)
(1.408, 0.9798)
(1.472, 0.9800)
(1.536, 0.9802)
(1.600, 0.9804)
(1.664, 0.9807)
(1.728, 0.9809)
(1.792, 0.9810)
(1.856, 0.9812)
(1.920, 0.9814)
(1.984, 0.9815)
(2.048, 0.9817)
}
\defdata{magnitudes-Config-E-1533-46-2147483-weight-norms-enc-8-mean}{%
(0.064, 0.9682)
(0.128, 0.9463)
(0.192, 0.9389)
(0.256, 0.9407)
(0.320, 0.9428)
(0.384, 0.9446)
(0.448, 0.9462)
(0.512, 0.9476)
(0.576, 0.9489)
(0.640, 0.9501)
(0.704, 0.9512)
(0.768, 0.9522)
(0.832, 0.9531)
(0.896, 0.9540)
(0.960, 0.9548)
(1.024, 0.9556)
(1.088, 0.9564)
(1.152, 0.9571)
(1.216, 0.9577)
(1.280, 0.9584)
(1.344, 0.9590)
(1.408, 0.9596)
(1.472, 0.9602)
(1.536, 0.9607)
(1.600, 0.9612)
(1.664, 0.9617)
(1.728, 0.9622)
(1.792, 0.9626)
(1.856, 0.9631)
(1.920, 0.9635)
(1.984, 0.9639)
(2.048, 0.9643)
}
\defdata{magnitudes-Config-E-1533-46-2147483-weight-norms-dec-8-mean}{%
(0.064, 0.9708)
(0.128, 0.9540)
(0.192, 0.9489)
(0.256, 0.9508)
(0.320, 0.9527)
(0.384, 0.9544)
(0.448, 0.9559)
(0.512, 0.9573)
(0.576, 0.9584)
(0.640, 0.9595)
(0.704, 0.9605)
(0.768, 0.9613)
(0.832, 0.9622)
(0.896, 0.9629)
(0.960, 0.9636)
(1.024, 0.9643)
(1.088, 0.9649)
(1.152, 0.9655)
(1.216, 0.9660)
(1.280, 0.9665)
(1.344, 0.9670)
(1.408, 0.9675)
(1.472, 0.9679)
(1.536, 0.9683)
(1.600, 0.9687)
(1.664, 0.9691)
(1.728, 0.9695)
(1.792, 0.9698)
(1.856, 0.9702)
(1.920, 0.9705)
(1.984, 0.9708)
(2.048, 0.9711)
}
\defdata{magnitudes-Config-E-1533-46-2147483-weight-norms-dec-16-mean}{%
(0.064, 0.9858)
(0.128, 0.9783)
(0.192, 0.9765)
(0.256, 0.9780)
(0.320, 0.9792)
(0.384, 0.9803)
(0.448, 0.9811)
(0.512, 0.9818)
(0.576, 0.9824)
(0.640, 0.9829)
(0.704, 0.9834)
(0.768, 0.9838)
(0.832, 0.9842)
(0.896, 0.9845)
(0.960, 0.9847)
(1.024, 0.9850)
(1.088, 0.9852)
(1.152, 0.9855)
(1.216, 0.9857)
(1.280, 0.9859)
(1.344, 0.9861)
(1.408, 0.9863)
(1.472, 0.9864)
(1.536, 0.9866)
(1.600, 0.9867)
(1.664, 0.9869)
(1.728, 0.9870)
(1.792, 0.9871)
(1.856, 0.9872)
(1.920, 0.9873)
(1.984, 0.9874)
(2.048, 0.9875)
}
\defdata{magnitudes-Config-E-1533-46-2147483-weight-norms-dec-32-mean}{%
(0.064, 0.9900)
(0.128, 0.9834)
(0.192, 0.9813)
(0.256, 0.9821)
(0.320, 0.9830)
(0.384, 0.9837)
(0.448, 0.9842)
(0.512, 0.9847)
(0.576, 0.9852)
(0.640, 0.9855)
(0.704, 0.9858)
(0.768, 0.9861)
(0.832, 0.9864)
(0.896, 0.9866)
(0.960, 0.9869)
(1.024, 0.9871)
(1.088, 0.9873)
(1.152, 0.9875)
(1.216, 0.9876)
(1.280, 0.9878)
(1.344, 0.9879)
(1.408, 0.9881)
(1.472, 0.9882)
(1.536, 0.9883)
(1.600, 0.9884)
(1.664, 0.9885)
(1.728, 0.9886)
(1.792, 0.9886)
(1.856, 0.9887)
(1.920, 0.9888)
(1.984, 0.9888)
(2.048, 0.9889)
}
\defdata{magnitudes-Config-E-1533-46-2147483-weight-norms-dec-64-mean}{%
(0.064, 0.9928)
(0.128, 0.9894)
(0.192, 0.9883)
(0.256, 0.9890)
(0.320, 0.9896)
(0.384, 0.9901)
(0.448, 0.9905)
(0.512, 0.9907)
(0.576, 0.9909)
(0.640, 0.9911)
(0.704, 0.9912)
(0.768, 0.9914)
(0.832, 0.9916)
(0.896, 0.9918)
(0.960, 0.9920)
(1.024, 0.9921)
(1.088, 0.9922)
(1.152, 0.9922)
(1.216, 0.9922)
(1.280, 0.9923)
(1.344, 0.9923)
(1.408, 0.9923)
(1.472, 0.9924)
(1.536, 0.9924)
(1.600, 0.9924)
(1.664, 0.9924)
(1.728, 0.9924)
(1.792, 0.9924)
(1.856, 0.9924)
(1.920, 0.9924)
(1.984, 0.9924)
(2.048, 0.9924)
}
\defdata{magnitudes-Config-F-1533-72-2147483-activation-norms-enc-64-mean}{%
(0.064, 0.4579)
(0.128, 0.4365)
(0.192, 0.4211)
(0.256, 0.4087)
(0.320, 0.3993)
(0.384, 0.3922)
(0.448, 0.3860)
(0.512, 0.3813)
(0.576, 0.3783)
(0.640, 0.3744)
(0.704, 0.3714)
(0.768, 0.3683)
(0.832, 0.3662)
(0.896, 0.3637)
(0.960, 0.3620)
(1.024, 0.3612)
(1.088, 0.3586)
(1.152, 0.3574)
(1.216, 0.3560)
(1.280, 0.3545)
(1.344, 0.3525)
(1.408, 0.3525)
(1.472, 0.3507)
(1.536, 0.3492)
(1.600, 0.3480)
(1.664, 0.3463)
(1.728, 0.3452)
(1.792, 0.3444)
(1.856, 0.3436)
(1.920, 0.3423)
(1.984, 0.3410)
(2.048, 0.3411)
}
\defdata{magnitudes-Config-F-1533-72-2147483-activation-norms-enc-32-mean}{%
(0.064, 0.4146)
(0.128, 0.4025)
(0.192, 0.3924)
(0.256, 0.3816)
(0.320, 0.3738)
(0.384, 0.3668)
(0.448, 0.3624)
(0.512, 0.3582)
(0.576, 0.3551)
(0.640, 0.3517)
(0.704, 0.3476)
(0.768, 0.3448)
(0.832, 0.3430)
(0.896, 0.3395)
(0.960, 0.3369)
(1.024, 0.3361)
(1.088, 0.3340)
(1.152, 0.3324)
(1.216, 0.3307)
(1.280, 0.3283)
(1.344, 0.3265)
(1.408, 0.3265)
(1.472, 0.3243)
(1.536, 0.3226)
(1.600, 0.3217)
(1.664, 0.3191)
(1.728, 0.3182)
(1.792, 0.3175)
(1.856, 0.3166)
(1.920, 0.3149)
(1.984, 0.3137)
(2.048, 0.3136)
}
\defdata{magnitudes-Config-F-1533-72-2147483-activation-norms-enc-16-mean}{%
(0.064, 0.9049)
(0.128, 0.8102)
(0.192, 0.7440)
(0.256, 0.6956)
(0.320, 0.6622)
(0.384, 0.6374)
(0.448, 0.6192)
(0.512, 0.6047)
(0.576, 0.5931)
(0.640, 0.5824)
(0.704, 0.5726)
(0.768, 0.5644)
(0.832, 0.5582)
(0.896, 0.5508)
(0.960, 0.5448)
(1.024, 0.5408)
(1.088, 0.5355)
(1.152, 0.5314)
(1.216, 0.5265)
(1.280, 0.5220)
(1.344, 0.5184)
(1.408, 0.5162)
(1.472, 0.5124)
(1.536, 0.5091)
(1.600, 0.5065)
(1.664, 0.5027)
(1.728, 0.5007)
(1.792, 0.4986)
(1.856, 0.4969)
(1.920, 0.4941)
(1.984, 0.4923)
(2.048, 0.4913)
}
\defdata{magnitudes-Config-F-1533-72-2147483-activation-norms-enc-8-mean}{%
(0.064, 1.3027)
(0.128, 1.2255)
(0.192, 1.1511)
(0.256, 1.0804)
(0.320, 1.0273)
(0.384, 0.9846)
(0.448, 0.9528)
(0.512, 0.9258)
(0.576, 0.9013)
(0.640, 0.8810)
(0.704, 0.8640)
(0.768, 0.8484)
(0.832, 0.8353)
(0.896, 0.8239)
(0.960, 0.8128)
(1.024, 0.8027)
(1.088, 0.7943)
(1.152, 0.7863)
(1.216, 0.7776)
(1.280, 0.7704)
(1.344, 0.7641)
(1.408, 0.7571)
(1.472, 0.7514)
(1.536, 0.7457)
(1.600, 0.7407)
(1.664, 0.7356)
(1.728, 0.7315)
(1.792, 0.7271)
(1.856, 0.7235)
(1.920, 0.7177)
(1.984, 0.7154)
(2.048, 0.7113)
}
\defdata{magnitudes-Config-F-1533-72-2147483-activation-norms-dec-8-mean}{%
(0.064, 1.6833)
(0.128, 1.6321)
(0.192, 1.5631)
(0.256, 1.4657)
(0.320, 1.3937)
(0.384, 1.3370)
(0.448, 1.2928)
(0.512, 1.2526)
(0.576, 1.2176)
(0.640, 1.1890)
(0.704, 1.1646)
(0.768, 1.1420)
(0.832, 1.1209)
(0.896, 1.1040)
(0.960, 1.0867)
(1.024, 1.0705)
(1.088, 1.0549)
(1.152, 1.0391)
(1.216, 1.0282)
(1.280, 1.0178)
(1.344, 1.0080)
(1.408, 0.9949)
(1.472, 0.9851)
(1.536, 0.9740)
(1.600, 0.9658)
(1.664, 0.9571)
(1.728, 0.9520)
(1.792, 0.9446)
(1.856, 0.9368)
(1.920, 0.9261)
(1.984, 0.9212)
(2.048, 0.9166)
}
\defdata{magnitudes-Config-F-1533-72-2147483-activation-norms-dec-16-mean}{%
(0.064, 1.2283)
(0.128, 1.0928)
(0.192, 1.0079)
(0.256, 0.9407)
(0.320, 0.8932)
(0.384, 0.8574)
(0.448, 0.8315)
(0.512, 0.8093)
(0.576, 0.7889)
(0.640, 0.7734)
(0.704, 0.7597)
(0.768, 0.7462)
(0.832, 0.7361)
(0.896, 0.7244)
(0.960, 0.7157)
(1.024, 0.7064)
(1.088, 0.6984)
(1.152, 0.6891)
(1.216, 0.6825)
(1.280, 0.6764)
(1.344, 0.6699)
(1.408, 0.6631)
(1.472, 0.6573)
(1.536, 0.6504)
(1.600, 0.6458)
(1.664, 0.6407)
(1.728, 0.6374)
(1.792, 0.6329)
(1.856, 0.6290)
(1.920, 0.6220)
(1.984, 0.6196)
(2.048, 0.6166)
}
\defdata{magnitudes-Config-F-1533-72-2147483-activation-norms-dec-32-mean}{%
(0.064, 0.7213)
(0.128, 0.6888)
(0.192, 0.6597)
(0.256, 0.6332)
(0.320, 0.6108)
(0.384, 0.5918)
(0.448, 0.5794)
(0.512, 0.5686)
(0.576, 0.5579)
(0.640, 0.5495)
(0.704, 0.5410)
(0.768, 0.5340)
(0.832, 0.5293)
(0.896, 0.5222)
(0.960, 0.5160)
(1.024, 0.5119)
(1.088, 0.5079)
(1.152, 0.5034)
(1.216, 0.4993)
(1.280, 0.4948)
(1.344, 0.4910)
(1.408, 0.4884)
(1.472, 0.4843)
(1.536, 0.4807)
(1.600, 0.4785)
(1.664, 0.4752)
(1.728, 0.4729)
(1.792, 0.4707)
(1.856, 0.4682)
(1.920, 0.4649)
(1.984, 0.4636)
(2.048, 0.4623)
}
\defdata{magnitudes-Config-F-1533-72-2147483-activation-norms-dec-64-mean}{%
(0.064, 0.6395)
(0.128, 0.5716)
(0.192, 0.5337)
(0.256, 0.5063)
(0.320, 0.4854)
(0.384, 0.4683)
(0.448, 0.4569)
(0.512, 0.4471)
(0.576, 0.4395)
(0.640, 0.4320)
(0.704, 0.4249)
(0.768, 0.4182)
(0.832, 0.4148)
(0.896, 0.4091)
(0.960, 0.4039)
(1.024, 0.4018)
(1.088, 0.3981)
(1.152, 0.3953)
(1.216, 0.3924)
(1.280, 0.3888)
(1.344, 0.3857)
(1.408, 0.3845)
(1.472, 0.3815)
(1.536, 0.3787)
(1.600, 0.3770)
(1.664, 0.3738)
(1.728, 0.3720)
(1.792, 0.3708)
(1.856, 0.3691)
(1.920, 0.3668)
(1.984, 0.3652)
(2.048, 0.3645)
}
\defdata{magnitudes-Config-F-1533-72-2147483-weight-norms-enc-64-mean}{%
(0.064, 0.9949)
(0.128, 0.9915)
(0.192, 0.9904)
(0.256, 0.9907)
(0.320, 0.9910)
(0.384, 0.9912)
(0.448, 0.9915)
(0.512, 0.9917)
(0.576, 0.9918)
(0.640, 0.9920)
(0.704, 0.9921)
(0.768, 0.9922)
(0.832, 0.9923)
(0.896, 0.9924)
(0.960, 0.9925)
(1.024, 0.9925)
(1.088, 0.9926)
(1.152, 0.9926)
(1.216, 0.9927)
(1.280, 0.9927)
(1.344, 0.9928)
(1.408, 0.9928)
(1.472, 0.9928)
(1.536, 0.9929)
(1.600, 0.9929)
(1.664, 0.9929)
(1.728, 0.9930)
(1.792, 0.9930)
(1.856, 0.9930)
(1.920, 0.9931)
(1.984, 0.9931)
(2.048, 0.9932)
}
\defdata{magnitudes-Config-F-1533-72-2147483-weight-norms-enc-32-mean}{%
(0.064, 0.9912)
(0.128, 0.9834)
(0.192, 0.9804)
(0.256, 0.9804)
(0.320, 0.9808)
(0.384, 0.9813)
(0.448, 0.9818)
(0.512, 0.9822)
(0.576, 0.9827)
(0.640, 0.9831)
(0.704, 0.9835)
(0.768, 0.9838)
(0.832, 0.9841)
(0.896, 0.9844)
(0.960, 0.9847)
(1.024, 0.9849)
(1.088, 0.9852)
(1.152, 0.9854)
(1.216, 0.9856)
(1.280, 0.9858)
(1.344, 0.9860)
(1.408, 0.9862)
(1.472, 0.9864)
(1.536, 0.9866)
(1.600, 0.9867)
(1.664, 0.9869)
(1.728, 0.9870)
(1.792, 0.9872)
(1.856, 0.9873)
(1.920, 0.9875)
(1.984, 0.9876)
(2.048, 0.9877)
}
\defdata{magnitudes-Config-F-1533-72-2147483-weight-norms-enc-16-mean}{%
(0.064, 0.9847)
(0.128, 0.9735)
(0.192, 0.9699)
(0.256, 0.9710)
(0.320, 0.9722)
(0.384, 0.9732)
(0.448, 0.9741)
(0.512, 0.9749)
(0.576, 0.9757)
(0.640, 0.9763)
(0.704, 0.9769)
(0.768, 0.9774)
(0.832, 0.9779)
(0.896, 0.9783)
(0.960, 0.9787)
(1.024, 0.9791)
(1.088, 0.9794)
(1.152, 0.9798)
(1.216, 0.9801)
(1.280, 0.9804)
(1.344, 0.9807)
(1.408, 0.9809)
(1.472, 0.9812)
(1.536, 0.9814)
(1.600, 0.9817)
(1.664, 0.9819)
(1.728, 0.9821)
(1.792, 0.9823)
(1.856, 0.9825)
(1.920, 0.9827)
(1.984, 0.9829)
(2.048, 0.9831)
}
\defdata{magnitudes-Config-F-1533-72-2147483-weight-norms-enc-8-mean}{%
(0.064, 0.9674)
(0.128, 0.9449)
(0.192, 0.9373)
(0.256, 0.9392)
(0.320, 0.9412)
(0.384, 0.9431)
(0.448, 0.9447)
(0.512, 0.9461)
(0.576, 0.9474)
(0.640, 0.9486)
(0.704, 0.9497)
(0.768, 0.9508)
(0.832, 0.9517)
(0.896, 0.9526)
(0.960, 0.9535)
(1.024, 0.9543)
(1.088, 0.9551)
(1.152, 0.9559)
(1.216, 0.9566)
(1.280, 0.9573)
(1.344, 0.9580)
(1.408, 0.9586)
(1.472, 0.9592)
(1.536, 0.9599)
(1.600, 0.9604)
(1.664, 0.9610)
(1.728, 0.9616)
(1.792, 0.9621)
(1.856, 0.9626)
(1.920, 0.9631)
(1.984, 0.9636)
(2.048, 0.9641)
}
\defdata{magnitudes-Config-F-1533-72-2147483-weight-norms-dec-8-mean}{%
(0.064, 0.9676)
(0.128, 0.9486)
(0.192, 0.9432)
(0.256, 0.9456)
(0.320, 0.9480)
(0.384, 0.9500)
(0.448, 0.9517)
(0.512, 0.9532)
(0.576, 0.9546)
(0.640, 0.9558)
(0.704, 0.9569)
(0.768, 0.9579)
(0.832, 0.9589)
(0.896, 0.9597)
(0.960, 0.9606)
(1.024, 0.9613)
(1.088, 0.9621)
(1.152, 0.9627)
(1.216, 0.9634)
(1.280, 0.9640)
(1.344, 0.9646)
(1.408, 0.9652)
(1.472, 0.9657)
(1.536, 0.9662)
(1.600, 0.9667)
(1.664, 0.9672)
(1.728, 0.9676)
(1.792, 0.9681)
(1.856, 0.9685)
(1.920, 0.9689)
(1.984, 0.9693)
(2.048, 0.9697)
}
\defdata{magnitudes-Config-F-1533-72-2147483-weight-norms-dec-16-mean}{%
(0.064, 0.9853)
(0.128, 0.9776)
(0.192, 0.9756)
(0.256, 0.9770)
(0.320, 0.9782)
(0.384, 0.9792)
(0.448, 0.9801)
(0.512, 0.9809)
(0.576, 0.9816)
(0.640, 0.9821)
(0.704, 0.9826)
(0.768, 0.9831)
(0.832, 0.9835)
(0.896, 0.9838)
(0.960, 0.9842)
(1.024, 0.9845)
(1.088, 0.9847)
(1.152, 0.9850)
(1.216, 0.9852)
(1.280, 0.9854)
(1.344, 0.9856)
(1.408, 0.9858)
(1.472, 0.9860)
(1.536, 0.9861)
(1.600, 0.9863)
(1.664, 0.9865)
(1.728, 0.9866)
(1.792, 0.9868)
(1.856, 0.9869)
(1.920, 0.9870)
(1.984, 0.9871)
(2.048, 0.9873)
}
\defdata{magnitudes-Config-F-1533-72-2147483-weight-norms-dec-32-mean}{%
(0.064, 0.9902)
(0.128, 0.9838)
(0.192, 0.9816)
(0.256, 0.9822)
(0.320, 0.9827)
(0.384, 0.9832)
(0.448, 0.9836)
(0.512, 0.9841)
(0.576, 0.9845)
(0.640, 0.9848)
(0.704, 0.9852)
(0.768, 0.9855)
(0.832, 0.9858)
(0.896, 0.9860)
(0.960, 0.9863)
(1.024, 0.9865)
(1.088, 0.9868)
(1.152, 0.9870)
(1.216, 0.9872)
(1.280, 0.9874)
(1.344, 0.9875)
(1.408, 0.9877)
(1.472, 0.9879)
(1.536, 0.9880)
(1.600, 0.9881)
(1.664, 0.9883)
(1.728, 0.9884)
(1.792, 0.9885)
(1.856, 0.9886)
(1.920, 0.9887)
(1.984, 0.9888)
(2.048, 0.9889)
}
\defdata{magnitudes-Config-F-1533-72-2147483-weight-norms-dec-64-mean}{%
(0.064, 0.9936)
(0.128, 0.9902)
(0.192, 0.9886)
(0.256, 0.9888)
(0.320, 0.9890)
(0.384, 0.9893)
(0.448, 0.9897)
(0.512, 0.9900)
(0.576, 0.9902)
(0.640, 0.9905)
(0.704, 0.9906)
(0.768, 0.9908)
(0.832, 0.9910)
(0.896, 0.9911)
(0.960, 0.9912)
(1.024, 0.9914)
(1.088, 0.9915)
(1.152, 0.9916)
(1.216, 0.9917)
(1.280, 0.9918)
(1.344, 0.9918)
(1.408, 0.9919)
(1.472, 0.9920)
(1.536, 0.9921)
(1.600, 0.9922)
(1.664, 0.9922)
(1.728, 0.9923)
(1.792, 0.9924)
(1.856, 0.9924)
(1.920, 0.9925)
(1.984, 0.9926)
(2.048, 0.9926)
}
\defdata{magnitudes-Config-G-1533-52-2147483-activation-norms-enc-64-mean}{%
(0.064, 0.7556)
(0.128, 0.7700)
(0.192, 0.7932)
(0.256, 0.8147)
(0.320, 0.8306)
(0.384, 0.8522)
(0.448, 0.8631)
(0.512, 0.8781)
(0.576, 0.8915)
(0.640, 0.8979)
(0.704, 0.9098)
(0.768, 0.9198)
(0.832, 0.9234)
(0.896, 0.9294)
(0.960, 0.9400)
(1.024, 0.9509)
(1.088, 0.9550)
(1.152, 0.9613)
(1.216, 0.9694)
(1.280, 0.9741)
(1.344, 0.9805)
(1.408, 0.9896)
(1.472, 0.9928)
(1.536, 0.9947)
(1.600, 1.0047)
(1.664, 1.0057)
(1.728, 1.0134)
(1.792, 1.0194)
(1.856, 1.0220)
(1.920, 1.0260)
(1.984, 1.0294)
(2.048, 1.0330)
}
\defdata{magnitudes-Config-G-1533-52-2147483-activation-norms-enc-32-mean}{%
(0.064, 0.6366)
(0.128, 0.6721)
(0.192, 0.6974)
(0.256, 0.7140)
(0.320, 0.7265)
(0.384, 0.7424)
(0.448, 0.7510)
(0.512, 0.7609)
(0.576, 0.7709)
(0.640, 0.7782)
(0.704, 0.7874)
(0.768, 0.7929)
(0.832, 0.7987)
(0.896, 0.8038)
(0.960, 0.8113)
(1.024, 0.8165)
(1.088, 0.8224)
(1.152, 0.8263)
(1.216, 0.8323)
(1.280, 0.8362)
(1.344, 0.8406)
(1.408, 0.8454)
(1.472, 0.8497)
(1.536, 0.8544)
(1.600, 0.8591)
(1.664, 0.8620)
(1.728, 0.8646)
(1.792, 0.8684)
(1.856, 0.8723)
(1.920, 0.8750)
(1.984, 0.8782)
(2.048, 0.8826)
}
\defdata{magnitudes-Config-G-1533-52-2147483-activation-norms-enc-16-mean}{%
(0.064, 0.8050)
(0.128, 0.7311)
(0.192, 0.6910)
(0.256, 0.6638)
(0.320, 0.6465)
(0.384, 0.6374)
(0.448, 0.6292)
(0.512, 0.6239)
(0.576, 0.6208)
(0.640, 0.6180)
(0.704, 0.6167)
(0.768, 0.6155)
(0.832, 0.6144)
(0.896, 0.6137)
(0.960, 0.6146)
(1.024, 0.6149)
(1.088, 0.6152)
(1.152, 0.6157)
(1.216, 0.6164)
(1.280, 0.6169)
(1.344, 0.6176)
(1.408, 0.6184)
(1.472, 0.6196)
(1.536, 0.6204)
(1.600, 0.6220)
(1.664, 0.6225)
(1.728, 0.6237)
(1.792, 0.6253)
(1.856, 0.6263)
(1.920, 0.6270)
(1.984, 0.6277)
(2.048, 0.6293)
}
\defdata{magnitudes-Config-G-1533-52-2147483-activation-norms-enc-8-mean}{%
(0.064, 1.0925)
(0.128, 1.0408)
(0.192, 0.9991)
(0.256, 0.9588)
(0.320, 0.9290)
(0.384, 0.9052)
(0.448, 0.8853)
(0.512, 0.8690)
(0.576, 0.8554)
(0.640, 0.8456)
(0.704, 0.8361)
(0.768, 0.8275)
(0.832, 0.8214)
(0.896, 0.8163)
(0.960, 0.8114)
(1.024, 0.8054)
(1.088, 0.8025)
(1.152, 0.7991)
(1.216, 0.7958)
(1.280, 0.7934)
(1.344, 0.7909)
(1.408, 0.7874)
(1.472, 0.7869)
(1.536, 0.7861)
(1.600, 0.7841)
(1.664, 0.7828)
(1.728, 0.7815)
(1.792, 0.7790)
(1.856, 0.7797)
(1.920, 0.7785)
(1.984, 0.7774)
(2.048, 0.7778)
}
\defdata{magnitudes-Config-G-1533-52-2147483-activation-norms-dec-8-mean}{%
(0.064, 1.3863)
(0.128, 1.4100)
(0.192, 1.3927)
(0.256, 1.3580)
(0.320, 1.3409)
(0.384, 1.3180)
(0.448, 1.3065)
(0.512, 1.2977)
(0.576, 1.2879)
(0.640, 1.2891)
(0.704, 1.2839)
(0.768, 1.2788)
(0.832, 1.2856)
(0.896, 1.2884)
(0.960, 1.2877)
(1.024, 1.2808)
(1.088, 1.2873)
(1.152, 1.2845)
(1.216, 1.2857)
(1.280, 1.2874)
(1.344, 1.2899)
(1.408, 1.2829)
(1.472, 1.2903)
(1.536, 1.2954)
(1.600, 1.2942)
(1.664, 1.2967)
(1.728, 1.2954)
(1.792, 1.2921)
(1.856, 1.2946)
(1.920, 1.2998)
(1.984, 1.2990)
(2.048, 1.3016)
}
\defdata{magnitudes-Config-G-1533-52-2147483-activation-norms-dec-16-mean}{%
(0.064, 0.9744)
(0.128, 0.8921)
(0.192, 0.8423)
(0.256, 0.8047)
(0.320, 0.7841)
(0.384, 0.7669)
(0.448, 0.7568)
(0.512, 0.7482)
(0.576, 0.7393)
(0.640, 0.7367)
(0.704, 0.7321)
(0.768, 0.7277)
(0.832, 0.7279)
(0.896, 0.7276)
(0.960, 0.7250)
(1.024, 0.7214)
(1.088, 0.7225)
(1.152, 0.7212)
(1.216, 0.7193)
(1.280, 0.7204)
(1.344, 0.7202)
(1.408, 0.7164)
(1.472, 0.7199)
(1.536, 0.7216)
(1.600, 0.7213)
(1.664, 0.7208)
(1.728, 0.7206)
(1.792, 0.7194)
(1.856, 0.7198)
(1.920, 0.7229)
(1.984, 0.7217)
(2.048, 0.7228)
}
\defdata{magnitudes-Config-G-1533-52-2147483-activation-norms-dec-32-mean}{%
(0.064, 0.7168)
(0.128, 0.7162)
(0.192, 0.7192)
(0.256, 0.7218)
(0.320, 0.7265)
(0.384, 0.7338)
(0.448, 0.7384)
(0.512, 0.7431)
(0.576, 0.7476)
(0.640, 0.7533)
(0.704, 0.7586)
(0.768, 0.7615)
(0.832, 0.7657)
(0.896, 0.7700)
(0.960, 0.7741)
(1.024, 0.7773)
(1.088, 0.7819)
(1.152, 0.7842)
(1.216, 0.7882)
(1.280, 0.7919)
(1.344, 0.7951)
(1.408, 0.7970)
(1.472, 0.8013)
(1.536, 0.8057)
(1.600, 0.8086)
(1.664, 0.8110)
(1.728, 0.8135)
(1.792, 0.8149)
(1.856, 0.8183)
(1.920, 0.8210)
(1.984, 0.8225)
(2.048, 0.8262)
}
\defdata{magnitudes-Config-G-1533-52-2147483-activation-norms-dec-64-mean}{%
(0.064, 0.8591)
(0.128, 0.8625)
(0.192, 0.8772)
(0.256, 0.8930)
(0.320, 0.9069)
(0.384, 0.9301)
(0.448, 0.9424)
(0.512, 0.9577)
(0.576, 0.9731)
(0.640, 0.9846)
(0.704, 0.9999)
(0.768, 1.0129)
(0.832, 1.0200)
(0.896, 1.0298)
(0.960, 1.0418)
(1.024, 1.0546)
(1.088, 1.0625)
(1.152, 1.0705)
(1.216, 1.0811)
(1.280, 1.0880)
(1.344, 1.0948)
(1.408, 1.1058)
(1.472, 1.1109)
(1.536, 1.1154)
(1.600, 1.1260)
(1.664, 1.1297)
(1.728, 1.1375)
(1.792, 1.1443)
(1.856, 1.1486)
(1.920, 1.1536)
(1.984, 1.1579)
(2.048, 1.1648)
}
\defdata{magnitudes-Config-G-1533-52-2147483-weight-norms-enc-64-mean}{%
(0.064, 0.9946)
(0.128, 0.9913)
(0.192, 0.9903)
(0.256, 0.9909)
(0.320, 0.9915)
(0.384, 0.9919)
(0.448, 0.9922)
(0.512, 0.9924)
(0.576, 0.9927)
(0.640, 0.9929)
(0.704, 0.9931)
(0.768, 0.9932)
(0.832, 0.9933)
(0.896, 0.9935)
(0.960, 0.9936)
(1.024, 0.9937)
(1.088, 0.9937)
(1.152, 0.9938)
(1.216, 0.9939)
(1.280, 0.9939)
(1.344, 0.9940)
(1.408, 0.9941)
(1.472, 0.9941)
(1.536, 0.9942)
(1.600, 0.9942)
(1.664, 0.9943)
(1.728, 0.9943)
(1.792, 0.9943)
(1.856, 0.9944)
(1.920, 0.9944)
(1.984, 0.9944)
(2.048, 0.9945)
}
\defdata{magnitudes-Config-G-1533-52-2147483-weight-norms-enc-32-mean}{%
(0.064, 0.9890)
(0.128, 0.9814)
(0.192, 0.9793)
(0.256, 0.9805)
(0.320, 0.9816)
(0.384, 0.9824)
(0.448, 0.9831)
(0.512, 0.9837)
(0.576, 0.9842)
(0.640, 0.9847)
(0.704, 0.9851)
(0.768, 0.9855)
(0.832, 0.9858)
(0.896, 0.9861)
(0.960, 0.9864)
(1.024, 0.9867)
(1.088, 0.9869)
(1.152, 0.9871)
(1.216, 0.9873)
(1.280, 0.9875)
(1.344, 0.9877)
(1.408, 0.9879)
(1.472, 0.9880)
(1.536, 0.9882)
(1.600, 0.9883)
(1.664, 0.9885)
(1.728, 0.9886)
(1.792, 0.9887)
(1.856, 0.9888)
(1.920, 0.9889)
(1.984, 0.9891)
(2.048, 0.9892)
}
\defdata{magnitudes-Config-G-1533-52-2147483-weight-norms-enc-16-mean}{%
(0.064, 0.9829)
(0.128, 0.9719)
(0.192, 0.9690)
(0.256, 0.9705)
(0.320, 0.9719)
(0.384, 0.9731)
(0.448, 0.9741)
(0.512, 0.9750)
(0.576, 0.9757)
(0.640, 0.9764)
(0.704, 0.9770)
(0.768, 0.9775)
(0.832, 0.9780)
(0.896, 0.9784)
(0.960, 0.9788)
(1.024, 0.9792)
(1.088, 0.9795)
(1.152, 0.9798)
(1.216, 0.9802)
(1.280, 0.9804)
(1.344, 0.9807)
(1.408, 0.9809)
(1.472, 0.9812)
(1.536, 0.9814)
(1.600, 0.9816)
(1.664, 0.9818)
(1.728, 0.9820)
(1.792, 0.9822)
(1.856, 0.9823)
(1.920, 0.9825)
(1.984, 0.9826)
(2.048, 0.9828)
}
\defdata{magnitudes-Config-G-1533-52-2147483-weight-norms-enc-8-mean}{%
(0.064, 0.9668)
(0.128, 0.9443)
(0.192, 0.9370)
(0.256, 0.9390)
(0.320, 0.9412)
(0.384, 0.9430)
(0.448, 0.9446)
(0.512, 0.9460)
(0.576, 0.9472)
(0.640, 0.9484)
(0.704, 0.9494)
(0.768, 0.9504)
(0.832, 0.9513)
(0.896, 0.9522)
(0.960, 0.9531)
(1.024, 0.9539)
(1.088, 0.9547)
(1.152, 0.9554)
(1.216, 0.9562)
(1.280, 0.9569)
(1.344, 0.9576)
(1.408, 0.9583)
(1.472, 0.9590)
(1.536, 0.9597)
(1.600, 0.9604)
(1.664, 0.9610)
(1.728, 0.9617)
(1.792, 0.9623)
(1.856, 0.9629)
(1.920, 0.9635)
(1.984, 0.9641)
(2.048, 0.9647)
}
\defdata{magnitudes-Config-G-1533-52-2147483-weight-norms-dec-8-mean}{%
(0.064, 0.9691)
(0.128, 0.9510)
(0.192, 0.9458)
(0.256, 0.9481)
(0.320, 0.9504)
(0.384, 0.9523)
(0.448, 0.9539)
(0.512, 0.9553)
(0.576, 0.9565)
(0.640, 0.9576)
(0.704, 0.9587)
(0.768, 0.9596)
(0.832, 0.9604)
(0.896, 0.9612)
(0.960, 0.9620)
(1.024, 0.9627)
(1.088, 0.9633)
(1.152, 0.9640)
(1.216, 0.9646)
(1.280, 0.9651)
(1.344, 0.9657)
(1.408, 0.9662)
(1.472, 0.9667)
(1.536, 0.9672)
(1.600, 0.9677)
(1.664, 0.9682)
(1.728, 0.9686)
(1.792, 0.9691)
(1.856, 0.9695)
(1.920, 0.9700)
(1.984, 0.9704)
(2.048, 0.9708)
}
\defdata{magnitudes-Config-G-1533-52-2147483-weight-norms-dec-16-mean}{%
(0.064, 0.9853)
(0.128, 0.9783)
(0.192, 0.9769)
(0.256, 0.9786)
(0.320, 0.9800)
(0.384, 0.9812)
(0.448, 0.9821)
(0.512, 0.9829)
(0.576, 0.9835)
(0.640, 0.9841)
(0.704, 0.9845)
(0.768, 0.9850)
(0.832, 0.9854)
(0.896, 0.9857)
(0.960, 0.9860)
(1.024, 0.9863)
(1.088, 0.9866)
(1.152, 0.9868)
(1.216, 0.9870)
(1.280, 0.9872)
(1.344, 0.9874)
(1.408, 0.9876)
(1.472, 0.9878)
(1.536, 0.9879)
(1.600, 0.9881)
(1.664, 0.9882)
(1.728, 0.9883)
(1.792, 0.9884)
(1.856, 0.9885)
(1.920, 0.9886)
(1.984, 0.9888)
(2.048, 0.9888)
}
\defdata{magnitudes-Config-G-1533-52-2147483-weight-norms-dec-32-mean}{%
(0.064, 0.9888)
(0.128, 0.9830)
(0.192, 0.9817)
(0.256, 0.9829)
(0.320, 0.9840)
(0.384, 0.9848)
(0.448, 0.9855)
(0.512, 0.9859)
(0.576, 0.9863)
(0.640, 0.9867)
(0.704, 0.9871)
(0.768, 0.9874)
(0.832, 0.9876)
(0.896, 0.9879)
(0.960, 0.9882)
(1.024, 0.9884)
(1.088, 0.9886)
(1.152, 0.9888)
(1.216, 0.9889)
(1.280, 0.9891)
(1.344, 0.9893)
(1.408, 0.9894)
(1.472, 0.9895)
(1.536, 0.9897)
(1.600, 0.9898)
(1.664, 0.9899)
(1.728, 0.9900)
(1.792, 0.9901)
(1.856, 0.9902)
(1.920, 0.9903)
(1.984, 0.9904)
(2.048, 0.9905)
}
\defdata{magnitudes-Config-G-1533-52-2147483-weight-norms-dec-64-mean}{%
(0.064, 0.9931)
(0.128, 0.9905)
(0.192, 0.9900)
(0.256, 0.9909)
(0.320, 0.9916)
(0.384, 0.9921)
(0.448, 0.9924)
(0.512, 0.9928)
(0.576, 0.9930)
(0.640, 0.9932)
(0.704, 0.9934)
(0.768, 0.9936)
(0.832, 0.9937)
(0.896, 0.9938)
(0.960, 0.9940)
(1.024, 0.9941)
(1.088, 0.9942)
(1.152, 0.9942)
(1.216, 0.9943)
(1.280, 0.9944)
(1.344, 0.9945)
(1.408, 0.9945)
(1.472, 0.9946)
(1.536, 0.9946)
(1.600, 0.9947)
(1.664, 0.9948)
(1.728, 0.9948)
(1.792, 0.9948)
(1.856, 0.9949)
(1.920, 0.9949)
(1.984, 0.9950)
(2.048, 0.9950)
}

\ifappendix
  \newcommand{\refappResults}{\autoref{app:results}}
  \newcommand{\refappArchitecture}{\autoref{app:architecture}}
  \newcommand{\refappPhema}{\autoref{app:phema}}
  \newcommand{\refappImplementation}{\autoref{app:implementation}}
  \newcommand{\refappConfigB}{\autoref{app:configB}}
  \newcommand{\refappConfigD}{\autoref{app:configD}}
\else
  \newcommand{\refappResults}{Appendix~A}
  \newcommand{\refappArchitecture}{Appendix~B}
  \newcommand{\refappPhema}{Appendix~C}
  \newcommand{\refappImplementation}{Appendix~D}
  \newcommand{\refappConfigB}{Appendix~B.2}
  \newcommand{\refappConfigD}{Appendix~B.4}
\fi

\title{Analyzing and Improving the Training Dynamics of Diffusion Models}

\newcommand{\authorbox}[1]{\makebox[40mm][c]{#1}}
\ifincludeemails
  \newcommand{\makeauthor}[3]{\authorbox{#1}\\\authorbox{#2}\\\authorbox{\tt\small#3}}
\else
  \newcommand{\makeauthor}[3]{\authorbox{#1}\\\authorbox{#2}}
\fi

\author{%
    \makeauthor{Tero Karras}{NVIDIA}{tkarras@nvidia.com}%
\and\makeauthor{Miika Aittala}{NVIDIA}{maittala@nvidia.com}%
\and\makeauthor{Jaakko Lehtinen}{NVIDIA, Aalto University}{jlehtinen@nvidia.com}%
\and\makeauthor{Janne Hellsten}{NVIDIA}{jhellsten@nvidia.com}%
\and\makeauthor{Timo Aila}{NVIDIA}{taila@nvidia.com}%
\and\makeauthor{Samuli Laine}{NVIDIA}{slaine@nvidia.com}%
}

\begin{document}

\ifforceanon\togglefalse{cvprfinal}\fi
\maketitle
\ifforceanon\toggletrue{cvprfinal}\fi

\begin{abstract}
Diffusion models currently dominate the field of data-driven image synthesis with their unparalleled scaling
  to large datasets.
In this paper, we identify and rectify several causes for uneven and ineffective training in the popular ADM diffusion model architecture, without altering its high-level structure.
Observing uncontrolled magnitude changes and imbalances in both the network activations and weights over the course of training, we redesign the network layers to
  preserve activation, weight, and update magnitudes on expectation.
We find that systematic application of this philosophy eliminates the observed drifts and imbalances, resulting in considerably better networks at equal computational complexity.
Our modifications improve the previous record FID of \data{img512-StyleGAN-XL-cfg2} in \mbox{ImageNet-512} synthesis to \data{img512-\cidXXLD-cfg2},
  achieved using fast deterministic sampling.

As an independent contribution, we present a method for setting the 
  exponential moving average (EMA) parameters post-hoc,
  i.e., after completing the training run.
This allows precise tuning of EMA length without the cost of performing several training runs,
  and reveals its surprising interactions with network architecture, training time, and guidance.
\end{abstract}

\section{Introduction}

High-quality image synthesis based on text prompts, example images, or other forms of input has become
  widely popular thanks to advances in denoising diffusion models \cite{SohlDickstein2015,Song2019gradients,Song2020ddim,Song2021sde,Ho2020,Nichol2021a,Vincent11}.
Diffusion-based approaches produce high-quality images while offering versatile controls \cite{Ho2021classifierfree,Brooks2022,Hertz2022,Zhang2023,Mokady2023}
  and convenient ways to introduce novel subjects~\cite{Ruiz2023,Gal2023},
  and they also extend to other modalities such as audio~\cite{Popov2021,Kong2021}, video~\cite{Ho2022,Blattmann2023,imagenvideo}, and 3D~shapes \cite{Poole2023,Lin2023,Raj2023,Shue2023}.
A recent survey of methods and applications is given by \citet{Yang2023}.

On a high level, diffusion models convert an image of pure noise to a novel generated image through repeated application of image denoising.
Mathematically, each denoising step can be understood through the lens of score matching~\cite{Hyvarinen05}, and it is typically implemented using a \mbox{U-Net} \cite{Ronneberger2015,Ho2020} equipped with self-attention~\cite{Vaswani2017} layers.
Since we do not contribute to the theory behind diffusion models, we refer the interested reader to the seminal works of
  \citet{SohlDickstein2015}, \citet{Song2019gradients}, and \citet{Ho2020}, as well as to \citet{Karras2022elucidating},
  who frame various mathematical frameworks in a common context.

\figQualityComputeScatter

Despite the seemingly frictionless scaling to very large datasets and models, the training dynamics of diffusion models remain challenging due to the highly stochastic loss function.
The final image quality is dictated by faint
  image details predicted throughout the sampling chain,
  and small mistakes at intermediate steps can have snowball effects in subsequent iterations.
The network must accurately estimate the average clean image across a vast range of noise levels, Gaussian noise realizations, and conditioning inputs. Learning to do so is difficult given the chaotic training signal that is randomized over all of these aspects.

To learn efficiently in such a noisy training environment, the network should ideally have a
  predictable and even response to parameter updates.
We argue that this ideal is not met in current state-of-the-art designs,
hurting the quality of the models and making it difficult to improve them due to complex interactions between hyperparameters, network design, and training setups.

Our overarching goal is to understand the sometimes subtle ways in which
  the training dynamics of the score network can become imbalanced by unintended phenomena,
  and to remove these effects one by one.
At the heart of our approach are the \emph{expected magnitudes} of weights, activations, gradients, and weight updates,
  all of which have been identified as important factors in previous work
  (e.g., \cite{Salimans2016,Arpit2016,Laarhoven2017,Cho2017,You2017,Klambauer2017,Zhang2019,Bernstein2020,Li2020,Brock2021a,Brock2021b,Kosson2023}).
  Our approach is, roughly speaking, to standardize \emph{all} magnitudes through a clean set of design choices that address their interdependencies in a unified manner.

Concretely, we present a series of modifications to the
  ADM~\cite{Dhariwal2021} \mbox{U-Net} architecture without changing its overall structure,
  and show considerable quality improvement along the way
  (\autoref{sec:archImprovements}).
The final network is a drop-in replacement for ADM.
It sets new record FIDs of \data{img512-\cidXXLD-cfg2} and \data{img512-\cidXXLD-fid2} for ImageNet-512 image synthesis
  with and without guidance, respectively,
  where the previous state-of-the-art FIDs were \data{img512-StyleGAN-XL-cfg2} and \data{img512-VDMpp-fid2}.
It performs particularly well with respect to model complexity
  (\autoref{figQualityComputeScatter}),
and achieves these results using fast deterministic sampling instead of
  the much slower stochastic sampling used in previous methods.

As an independent contribution, we present a method for setting the exponential moving average (EMA)
  parameters \emph{post hoc}, i.e., after the training run has completed (\autoref{sec:phema}).
Model averaging~\cite{Ruppert1988,Polyak1992,Tarvainen2017,Izmailov2018,Yazici2019} is an indispensable technique in all high-quality image synthesis methods
  \cite{Song2021sde,Dhariwal2021,Karras2019stylegan,Sauer2022,Balaji2022,Ho2021cascaded,Kang2023,Karras2022elucidating,Nichol2021a,Peebles2022,Rombach2021highresolution,Song2020ddim}.
Unfortunately, the EMA decay constant is a cumbersome hyperparameter to tune
  because the effects of small changes become apparent only when the training is nearly converged.
Our \mbox{\emph{post-hoc~EMA}} allows accurate and efficient reconstruction of networks with
  arbitrary EMA profiles based on pre-integrated weight snapshots stored during training.
It also enables many kinds of exploration that have not been computationally feasible before (\autoref{sec:phemaanalysis}).

Our implementation and pre-trained models are available at {\small\url{https://github.com/NVlabs/edm2}}

\section{Improving the training dynamics}
\label{sec:archImprovements}

\tabBridgeMain
\figArchitecture

Let us now proceed to study and eliminate effects related to various imbalances
  in the training dynamics of a score network.
As our baseline, we take the ADM~\cite{Dhariwal2021} network
  as implemented in the EDM~\cite{Karras2022elucidating} framework.
The architecture combines a
  U-Net~\cite{Ronneberger2015} with self-attention~\cite{Vaswani2017} layers
  (\autoref{figArchitecture}\mbox{a,b}), and its variants have been widely adopted in large-scale diffusion models, including Imagen \cite{imagen}, Stable Diffusion \cite{Rombach2021highresolution}, eDiff-I \cite{Balaji2022}, DALL-E~2 \cite{dalle2,glide}, and \mbox{DALL-E~3} \cite{dalle3}.
Our training and sampling setups are based on the EDM formulation with constant learning rate
  and~32 deterministic~2\textsuperscript{nd} order sampling steps.

We use the class-conditional ImageNet~\cite{Deng2009imagenet} 512$\times$512 dataset for evaluation,
  and, like most high-resolution diffusion models,
  operate in the latent space of a pre-trained decoder \cite{Rombach2021highresolution} that performs~$8\times$ spatial upsampling. Thus, our
  output is 64$\times$64$\times$4 prior to decoding.
During exploration, we use a modestly sized network configuration with approx.~300M trainable parameters,
  with results for scaled-up networks presented later in \autoref{sec:results}.
The training is done for \data{img512-\cidG-mimg0}M ($= 2^{31}$) images in batches of~2048,
  which is sufficient for these models to reach their optimal FID.

We will build our improved architecture and training procedure in several steps.
Our exposition focuses on fundamental principles and the associated changes to the network.
For comprehensive details of each architectural step, along with the related equations,
  see \refappArchitecture{}.

\vspace{-2mm}%
\paragraph{Baseline (\bfconfig{a}).}
As the original EDM configuration is targeted for RGB images,
  we increase the output channel count to~4 and replace the training dataset with
  64$\times$64$\times$4 latent representations of ImageNet-512 images,
  standardized globally to zero mean and standard deviation $\sdata = 0.5$.
In this setup, we obtain a baseline FID of~\data{img512-\cidA-fid2} (see \autoref{tabBridgeMain}).

\subsection{Preliminary changes}

\paragraph{Improved baseline (\bfconfig{b}).}
We first tune the hyperparameters (learning rate, EMA length, training noise level distribution, etc.) to optimize the performance of the baseline model.
We also disable self-attention at 32$\times$32 resolution, similar to many prior works~\cite{Hoogeboom2023b,Ho2020,Nichol2021a}.

We then address a shortcoming in the original EDM training setup:
While the loss weighting in EDM standardizes loss magnitude to $1.0$ for all noise levels at initialization,
  this situation no longer holds as the training progresses.
The magnitude of the gradient feedback then varies between noise levels,
re-weighting their relative contribution in an uncontrolled manner.

To counteract this effect,
  we adopt a continuous generalization of the multi-task loss proposed by \citet{Kendall2018}.
Effectively, we track the raw loss value as a function of the noise level,
  and scale the training loss by its reciprocal.
See \refappConfigB{} for further details and reasoning.
Together, these changes decrease the FID from \data{img512-\cidA-fid2} to \data{img512-\cidB-fid2}.

\vspace{-1mm}%
\paragraph{Architectural streamlining (\bfconfig{c}).}
To facilitate the analysis of training dynamics, we proceed to streamline and stabilize the architecture.
To avoid having to deal with multiple different types of trainable parameters,
  we remove the additive biases from all convolutional and linear layers, as well as from the conditioning pathway.
To restore the capability of the network to offset the data, we concatenate an additional channel of constant~$1$ to the network's input.
We further unify the initialization of all weights using He's uniform init~\cite{He2015},
  switch from ADM's original positional encoding scheme to the more standard
  Fourier features~\cite{Tancik2020fourier},
  and simplify the group normalization layers by removing their mean subtraction and learned scaling.

Finally, we observe that the attention maps often end up in a brittle and spiky configuration
  due to magnitude growth of the key and query vectors over the course of training.
We rectify this by switching to \emph{cosine attention}~\cite{Gidaris2018,Luo2018,Nguyen2023}
  that normalizes the vectors prior to computing the dot products.
As a practical benefit, this allows using 16-bit floating point math
  throughout the network, improving efficiency.
Together, these changes reduce the FID from \data{img512-\cidB-fid2} to \data{img512-\cidC-fid2}.

\subsection{Standardizing activation magnitudes}

With the architecture simplified, we now turn to fixing the first problem in training dynamics: activation magnitudes.
As illustrated in the first row of \autoref{figMagnitudes}, the activation magnitudes grow
  uncontrollably in \config{c} as training progresses, despite the use of group normalizations within each block.
Notably, the growth shows no signs of tapering off or stabilizing towards the end of the training run.

Looking at the architecture in \autoref{figArchitecture}b, the growth is perhaps not too surprising:
  Due to the residual structure of encoder, decoder, and self-attention blocks,
  ADM networks contain long signal paths without any normalizations.
These paths accumulate contributions from residual branches and can amplify their
  activations through repeated convolutions.
We hypothesize that this unabated growth of activation magnitudes is detrimental to training
  by keeping the network in a perpetually unconverged and unoptimal state.

We tried introducing group normalization layers to the main path as well, but this caused a significant deterioration of result quality.
This may be related to previous findings regarding StyleGAN~\cite{Karras2019stylegan}, where the network's capabilities were impaired by excessive normalization, to the extent that the layers learned to bypass it via contrived image artifacts.
Inspired by the solutions adopted in StyleGAN2~\cite{Karras2020cvpr} and other works that have sought alternatives to explicit normalization~\cite{Arpit2016,Klambauer2017,Brock2021a}, we choose to modify the network so that individual layers and
  pathways preserve the activation magnitudes \emph{on expectation}, with the goal of removing or at least reducing
  the need for data-dependent normalization.

\vspace{-1mm}%
\paragraph{Magnitude-preserving learned layers (\bfconfig{d}).}

To preserve expected activation magnitudes, we divide the output of each layer by the
  expected scaling of activation magnitudes caused by that layer without looking at the activations themselves.
We first apply this to all learned layers (convolutions and fully-connected)
  in every part of the model.

Given that we seek a scheme that is agnostic to the actual content of the incoming activations, we have to make some statistical assumptions about them.
For simplicity, we will assume that the pixels and feature maps are
  mutually uncorrelated and of equal standard deviation $\sigma_\text{act}$.
Both fully connected and convolutional layers can be thought of as consisting of stacked units,
  one per output channel.
Each unit effectively applies a dot product of a weight vector
  $\boldwi \in \RR^n$ on some subset of the input activations to produce each output element.
Under our assumptions, the standard deviation of the output features of the $i$th channel
  becomes $\norm{\boldwi}_2 \sigma_\text{act}$.
To restore the input activation magnitude, we thus divide by $\norm{\boldwi}_2$ channel-wise.%
\footnote{The primary goal is to sever the direct link from weight to activation magnitude; for this, the statistical assumptions do not need to hold exactly.}

We can equally well think of the scalar division as applying to $\boldwi$ itself.
As long as gradients are propagated through the computation of the norm,
  this scheme is equivalent to \emph{weight normalization}~\cite{Salimans2016}
  without the learned output scale;
  we will use this term hereafter.
As the overall weight magnitudes no longer have an effect on activations, we 
  initialize all weights by drawing from the unit Gaussian distribution.

This modification removes any direct means the network has for learning to change the overall activation magnitudes,
  and as shown in \autoref{figMagnitudes} (\config{d}), the magnitude drift is successfully eliminated.
The FID also improves significantly, from \data{img512-\cidC-fid2} to \data{img512-\cidD-fid2}.

\subsection{Standardizing weights and updates}

With activations standardized, we turn our attention to network weights and learning rate.
As seen in \autoref{figMagnitudes}, there is a clear tendency of network weights to grow in \config{d},
  even more so than in \config{c}.
The mechanism causing this is well known~\cite{Salimans2016}:
  Normalization of weights before use forces loss gradients to be perpendicular to the weight vector,
  and taking a step along this direction always lands on a point further away from the origin.
Even with gradient magnitudes standardized by the Adam optimizer, the net effect is that the \emph{effective learning rate}, i.e., the relative size of the update to network weights, decays as the training progresses.

While it has been suggested that this decay of effective learning rate is a desirable
  effect~\cite{Salimans2016}, we argue for explicit control over it
  rather than having it drift uncontrollably and unequally between layers.
Hence, we treat this as another imbalance in training dynamics that we seek to remedy.
Note that initializing all weights to unit Gaussian ensures uniform effective learning rate at initialization,
  but not afterwards.

\figMagnitudes

\vspace{-1mm}%
\paragraph{Controlling effective learning rate (\bfconfig{e}).}

We propose to address the weight growth with \emph{forced weight normalization}, where
  we explicitly normalize every weight vector $\boldwi$ to unit variance before each training step.
Importantly, we still apply the ``standard'' weight normalization on top of this during training,
  i.e., normalize the weight vectors upon use.
This has the effect of projecting the training gradients onto the tangent plane of the now unit-magnitude hypersphere
  where $\boldwi$ lies (see \refappConfigD{} for a derivation).
This ensures that Adam's variance estimates are computed for the actual tangent plane steps
  and are not corrupted by the to-be erased normal component of the gradient vector.
With both weight and gradient magnitudes now equalized across the network,
  we have unified the effective learning rate as well.
Assuming no correlation between weights and gradients, each Adam step now replaces an approximately fixed proportion of the weights with the gradients.
Some optimizers~\cite{You2017,Bernstein2020,Kosson2023} explicitly implement a similar effect by data-dependent re-scaling of the gradient.

We now have direct control over the effective learning rate.
A constant learning rate no longer induces convergence, and
  thus we introduce an inverse square root learning rate decay schedule as advocated by \citet{Kingma2015}.
Concretely, we define
  \mbox{$\alpha(t) = \alpha_\text{ref} / \!\sqrt{ \max(t / t_\text{ref}, 1) }$},
  where $t$ is the current training iteration and
  $\alpha_\text{ref}$ and $t_\text{ref}$ are hyperparameters (see \refappImplementation{} for implementation details).
As shown in \autoref{figMagnitudes}, the resulting \config{e} successfully
  preserves both activation and weight magnitudes throughout the training.
As a result, the FID improves from \data{img512-\cidD-fid2} to \data{img512-\cidE-fid2}.

\subsection{%
\texorpdfstring{Removing group normalizations (\bfconfig{f})}%
               {Removing group normalizations (Config F)}%
}

With activation, weight, and update magnitudes under control,
  we are now ready to remove the data-dependent group normalization layers that
  operate across pixels with potentially detrimental results~\cite{Karras2020cvpr}.
Although the network trains successfully without any normalization layers,
  we find that there is still a small benefit from introducing much weaker
  \emph{pixel normalization}~\cite{Karras2018} layers to the encoder main path.
Our hypothesis is that pixel normalization helps by counteracting correlations that
  violate the statistical assumptions behind our standardization efforts in \config{d}.
We thus remove all group normalization layers and replace them with
  \mbox{1/4} as many pixel normalization layers.
We also remove the second linear layer from the embedding network and the nonlinearity from the network output,
  and combine the resampling operations in the residual blocks onto the main path.
The FID improves \mbox{from \data{img512-\cidE-fid2} to~\data{img512-\cidF-fid2}}.

\vspace{-.5mm}%
\subsection{%
\texorpdfstring{Magnitude-preserving fixed-function layers\\\hspace*{-.45em}(\bfconfig{g})}%
               {Magnitude-preserving fixed-function layers (Config G)}%
}
\vspace{-.5mm}%

For the sake of completeness, we note that the network still has layers that do not preserve activation
  magnitudes.
First, the sine and cosine functions of the Fourier features do not have unit variance,
  which we rectify by scaling them up by $\smash{\sqrt{2}}$.
Second, the SiLU~\cite{Hendrycks2016} nonlinearities attenuate the expected unit-variance
  distribution of activations unless this is compensated for.
Accordingly, we modify them to divide the output by
  \mbox{$\smash{\EE_{x\sim\NN(0,1)}[\,\silu(x)^2\,]^{1/2}} \approx 0.596$}.%

Third, we consider instances where two network branches join, either through addition or concatenation. In previous configurations, the contribution from each branch to the output depended on uncontrolled activation magnitudes. By now we can expect these to be standardized, and thus the balance between the branches is exposed as a meaningfully controllable parameter \cite{Brock2021b}.
We switch the addition operations to weighted sums, and observe experimentally that a fixed residual path weight of 30\% worked best in encoder and decoder blocks, and 50\% in the embedding. We divide the output by the expected \mbox{standard deviation of this weighted sum.}

The concatenation of the \mbox{U-Net} skips in the decoder is already magnitude-preserving, as we can expect similar magnitudes from both branches.
However, the relative contribution of the two inputs in subsequent layers is proportional to their respective
  channel counts, which we consider to be an unwanted and unintuitive dependence between encoder and decoder hyperparameters.
We remove this dependency by scaling the inputs such that the overall magnitude of the concatenated result
  remains unchanged, but the contributions of the inputs become equal.

With the standardization completed,
  we identify two specific places where it is still necessary to scale activations by a learned amount.
First, we add a learned, zero-initialized scalar gain (i.e., scaling) at the very end of the network,
  as we cannot expect the desired output to always have unit variance.
Second, we apply a similar learned gain to the conditioning signal within each residual block,
  so that the conditioning is disabled at initialization and its strength in each encoder/decoder block
  becomes a learned parameter.
At this point we can disable dropout~\cite{Hinton2012,Srivastava2014} during training with no ill effects,
  which has not been previously possible.

\autoref{figArchitecture}c illustrates our final design that
  is significantly simpler and easier to reason about than the baseline.
The resulting FID of \data{img512-\cidG-fid2} is highly competitive with the current state of the art,
  especially considering the modest computational complexity of our exploration architecture.

\section{Post-hoc EMA}
\label{sec:phema}

It is well known that exponential moving average (EMA) of model weights plays an important role in generative image synthesis~\cite{Song2021sde,Nichol2021a}, and that the choice of its decay parameter has a significant impact on results \cite{Nichol2021a,Kang2023}.

Despite its known importance, little is known about the relationships between the decay parameter
  and other aspects of training and sampling.
To analyze these questions, we develop a method for choosing the EMA profile \emph{post hoc}, i.e.,
  without the need to specify it before the training.
This allows us to sample the length of EMA densely and plot its effect on quality,
  revealing interesting interactions with network architecture, training time, and classifier-free guidance.

Further details, derivations, and discussion on the equations and methods in this section
  are included in \refappPhema{}.

\subsection{Power function EMA profile}
\label{sec:powerEMA}

Traditional EMA maintains a running weighted average $\hat{\theta}_\beta$ of the network parameters
  alongside the parameters $\theta$ that are being trained.
At each training step, the average is updated by
  \mbox{$\hat{\theta}_\beta(t) = \beta \,\hat{\theta}_\beta(t\!-\!1) + (1\!-\!\beta) \,\theta(t)$},
  where $t$ indicates the current training step, yielding an exponential decay profile in the contributions of earlier training steps.
The rate of decay is determined by the constant $\beta$ that is typically close to one.

For two reasons, we propose using a slightly altered averaging profile based on power functions instead of exponential decay.
First,
  our architectural modifications tend to favor longer averages;
  yet, very long exponential EMA puts non-negligible weight on initial stages of training
  where network parameters are mostly random.
Second, we have observed a clear trend that longer training runs benefit from longer EMA decay,
  and thus the averaging profile should ideally scale automatically with training time.

Both of the above requirements are fulfilled by power functions.
We define the averaged parameters at time $t$ as
\begin{equation}
  \hat{\theta}_\gamma(t)
  ~=~ \frac{\int_0^t \tau^\gamma \theta(\tau)\,\diff\tau}{\int_0^t \tau^\gamma\,\diff\tau}
  ~=~ \frac{\gamma+1}{t^{\gamma+1}}\int_0^t \tau^\gamma \theta(\tau)\,\diff\tau \label{eq:powerema}
  \text{,}\vspace{-1mm}%
\end{equation}
where the constant $\gamma$ controls the sharpness of the profile.
With this formulation, the weight of $\theta_{t=0}$ is always zero.
  This is desirable, as the random initialization should have no effect in the average.
The resulting averaging profile is also scale-independent:
  doubling the training time automatically stretches the profile by the same factor.

To compute $\hat{\theta}_\gamma(t)$ in practice,
  we perform an incremental update after each training step as follows:
\vspace{-0.5mm}%
\begin{equation}
  \begin{aligned}
    \hat{\theta}_\gamma(t) &~=~ \beta_\gamma(t) \,\hat{\theta}_\gamma(t-1) + \big(1-\beta_\gamma(t)\big) \,\theta(t) \\
    \text{where\ \ } \beta_\gamma(t) &~=~ (1 - 1/t)^{\gamma+1} \label{eq:powerupdate}
    \text{.}
  \end{aligned}\vspace{-0.5mm}%
\end{equation}
The update is thus similar to traditional EMA,
  but with the exception that $\beta$ depends on the current training time.%
  \footnote{%
  Technically, calling this an ``EMA profile'' is a misnomer, as the weight decay is not exponential.
  However, given that it serves the same purpose as traditional EMA, we feel that coining a new term here would be misleading.
}

Finally, while parameter $\gamma$ is mathematically straightforward,
  it has a somewhat unintuitive effect on the shape of the averaging profile.
Therefore, we prefer to parameterize the profile via its relative standard deviation $\srel$,
  i.e., the ``width'' of its peak relative to training time:
  \mbox{$\srel = (\gamma + 1)^{1/2}(\gamma + 2)^{-1}(\gamma + 3)^{-1/2}$}.
Thus, when reporting, say, EMA length of 10\%, we refer to a profile with $\srel=0.10$
  (equiv.~$\gamma \approx 6.94$).

\vspace{1mm}%
\subsection{Synthesizing novel EMA profiles after training}
\label{sec:novelEMA}

Our goal is to allow choosing $\gamma$, or equivalently $\srel$, freely after training.
To achieve this, we maintain two averaged parameter vectors $\hat{\theta}_{\gamma_1}$
  and $\hat{\theta}_{\gamma_2}$ during training, with constants \mbox{$\gamma_1=16.97$} and \mbox{$\gamma_2=6.94$},
  corresponding to $\srel$ of $0.05$ and $0.10$, respectively.
These averaged parameter vectors are stored periodically in snapshots saved during the training run.
In all our experiments, we store a snapshot once every $\sim$8 million training images,
  i.e., once every 4096 training steps with batch size of 2048.

To reconstruct an approximate $\hat{\theta}$ corresponding to an arbitrary EMA profile at any point during or after training,
  we find the least-squares optimal fit between the EMA profiles of the stored $\hat{\theta}_{\gamma_i}$ and the desired EMA profile,
  and take the corresponding linear combination of the \mbox{stored $\smash{\hat{\theta}_{\gamma_i}}$}.
See \autoref{figEmaSnapshots} for an illustration.

\figEmaSnapshots

We note that post-hoc EMA reconstruction is not limited to power function averaging profiles,
  or to using the same types of profiles for snapshots and the reconstruction.
Furthermore,
  it can be done even from a single stored $\hat{\theta}$ per snapshot,
  albeit with much lower accuracy than with two stored $\hat{\theta}$.
This opens the possibility of revisiting previous training runs that were not run with post-hoc EMA in mind,
  and experimenting with novel averaging profiles,
  as long as a sufficient number of training snapshots are available.

\vspace{1mm}%
\subsection{Analysis}
\label{sec:phemaanalysis}

\figEmaTriplet

Armed with the post-hoc EMA technique, we now analyze the effect of different EMA lengths in various setups.

\autoref{figEmaTriplet}a shows how FID varies based on EMA length in configurations \textsc{b}--\textsc{g} of \autoref{tabBridgeMain}.
We can see that the optimal EMA length differs considerably between the configs.
Moreover, the optimum becomes narrower as we approach the final config~\textsc{g},
  which might initially seem alarming.

However, as illustrated in \autoref{figEmaTriplet}b, the narrowness of the optimum seems to be explained by the model becoming more
  uniform in terms of which EMA length is ``preferred'' by each weight tensor.
In this test, we first select a subset of weight tensors from different parts of the network.
Then, separately for each chosen tensor, we perform a sweep where only the chosen tensor's EMA is changed, while all others remain at the global optimum.
The results, shown as one line per tensor, reveal surprisingly large effects on FID.
Interestingly, while it seems obvious that one weight tensor being out-of-sync with the others can be harmful, we observe that in \config{b}, FID can \emph{improve} as much as 10\%, from 7.24 to $\sim$6.5.
In one instance, this is achieved using a very short per-tensor EMA, and in another, a very long one.
We hypothesize that these different preferences mean that any global choice is an uneasy compromise.
For our final \config{g}, this effect disappears and the optimum is sharper: no significant improvement in FID can be seen, and the tensors now agree about the optimal EMA.
While post-hoc EMA allows choosing the EMA length on a per-tensor basis, we have not explored this opportunity outside this experiment.

Finally, \autoref{figEmaTriplet}c illustrates the evolution of the optimal EMA length over the course of training.
Even though our definition of EMA length is already relative to the length of training,
  we observe that the optimum slowly shifts towards relatively longer EMA as the training progresses.

\vspace{-0.5mm}
\section{Results}
\vspace{-0.5mm}
\label{sec:results}

We use ImageNet~\cite{Deng2009imagenet} in 512$\times$512 resolution as our main dataset.
\autoref{tabResultsFiveTwelve} summarizes FIDs for various model sizes using our method,
  as well as several earlier techniques.

Let us first consider FID without guidance~\cite{Ho2021classifierfree},
  where the best previous method is VDM++~\cite{Kingma2023} with FID of \data{img512-VDMpp-fid2}.
Even our small model \mbox{\OURS-S} that was used for the architecture exploration
  in \autoref{sec:archImprovements} beats this with FID of \data{img512-\cidS-fid2}.
Scaling our model up further improves FID to \data{img512-\cidXXLD-fid2},
  surpassing the previous record by a considerable margin.
As shown in \autoref{figQualityComputeScatter}, our results are even more significant
  in terms of model complexity.

We have found that enabling dropout~\cite{Hinton2012,Srivastava2014} improves our results in cases that exhibit overfitting, i.e.,
  when the training loss continues to decrease but validation loss and
  FID start increasing.
We thus enable dropout in our larger configurations (M--XXL) that show signs of overfitting,
  while disabling it in the smaller configurations (XS, S) where it is harmful.

Additional quantitative results, example images,
  and detailed comparisons for this section are given in \refappResults{}.

\tabResultsFiveTwelve

\vspace{-1mm}%
\paragraph{Guidance.}
It is interesting to note that several earlier methods~\cite{Dhariwal2021,Peebles2022}
  report competitive results only when classifier-free guidance~\cite{Ho2021classifierfree} is used.
While guidance remains an invaluable tool for controlling the balance between the perceptual quality of individual result images and the coverage of the generated distribution, it should not be necessary when the goal is to simply match image distributions.

\autoref{figEmaGuidance} plots the FID for our small model (\mbox{\OURS-S}) using a variety of guidance strengths as a function of EMA length.
The surprising takeaway is that the optimal EMA length depends very strongly on the guidance strength.
These kinds of studies are extremely expensive without post-hoc EMA, and 
  we therefore postulate that the large discrepancy between vanilla and guidance results in some prior art
  may be partially an artifact of using non-optimal EMA parameters.
With our largest model, a modest amount of guidance (1.2) further improves the ImageNet-512 FID
  from \data{img512-\cidXXLD-fid2} to \data{img512-\cidXXLD-cfg2}, setting a new record for this dataset.

\figEmaGuidance

\vspace{-1mm}%
\paragraph{Low-cost guidance.}
The standard way of implementing classifier-free guidance is to train a single model to support both conditional and unconditional generation \cite{Ho2021classifierfree}.
While conceptually simple, this makes the implicit assumption that a similarly complex model is needed for both tasks.
However, this does not seem to be the case:
  In our tests, the smallest (XS) unconditional model was found to be sufficient for
  guiding even the largest (XXL) conditional model\,---\,%
  using a larger unconditional model did not improve the results at all.

Our results in \autoref{tabResultsFiveTwelve} are computed using an XS-sized unconditional model
  for all of our configurations.
Using a small unconditional model can greatly reduce the typical 2$\times$ computational overhead of guidance.

\vspace{-1mm}%
\paragraph{ImageNet-64.}
To demonstrate that our method is not limited to latent diffusion, we provide results for RGB-space diffusion in ImageNet-64.
\autoref{tabResultsSixtyFour} shows that our results are superior to earlier methods that use deterministic sampling.
The previous record FID of \data{img64-EDM-EDM-det-fid2} set by EDM~\cite{Karras2022elucidating} improves
  to \data{img64-\cidSfS-fid2} at similar model complexity, and further to \data{img64-\cidSfXLD-fid2} via scaling.
The L-sized model is able to saturate this dataset.

This result is close to the record FID of \data{img64-RIN-stoch-fid2} achieved by RIN using stochastic sampling.
Stochastic sampling can correct for the inaccuracies of the denoising network, but
this comes at a considerable tuning effort and computational cost (e.g.,~1000 vs.~63 NFE),
  making stochastic sampling unattractive for large-scale systems.
It is likely that our results could be improved further using stochastic sampling,
   but we leave that as future work.

\tabResultsSixtyFour

\vspace{-1mm}%
\paragraph{Post-hoc EMA observations.}
Besides the interactions discussed in preceding sections, we have made two preliminary findings related to EMA length.
We present them here as anecdotal, and leave a detailed study for future work.

First, we observed that the optimal EMA length goes down when learning rate is increased, and vice versa, roughly according to
  \mbox{$\smash{\srel \propto 1 / ( \alpha_\text{ref}^2 ~t_\text{ref} )}$}.
The resulting FID also remains relatively stable over a perhaps 2$\times$ range of $t_\text{ref}$.
In practice, setting $\alpha_\text{ref}$ and $t_\text{ref}$ within the right ballpark thus seems to be sufficient, which reduces the need to tune these hyperparameters carefully.

Second, we observed that the optimal EMA length tends to go down when the model capacity is increased, and also when the complexity of the dataset is decreased.
This seems to imply that simpler problems warrant a shorter EMA.

\section{Discussion and future work}

Our improved denoiser architecture was designed to be a drop-in replacement for the widely used ADM network, and thus we hope it will find widespread use in large-scale image generators.
With various aspects of the training now much less entangled, it becomes easier to make local modifications to the architecture without something breaking elsewhere.
This should allow further studies to the structure and balance of the U-Net, among other things.

An interesting question is whether similar methodology would be equally beneficial for other diffusion architectures such as RIN~\cite{Jabri2023} and DiT~\cite{Peebles2022}, as well as other application areas besides diffusion models.
It would seem this sort of magnitude-focusing work has attracted relatively little attention outside the specific topic of \mbox{ImageNet classifiers \cite{Brock2021a,Brock2021b}.}

We believe that post-hoc EMA will enable a range of interesting studies that have been infeasible before.
Some of our plots would have taken a thousand GPU-years to produce without it; they now took only a GPU-month instead.
We hope that the cheap-to-produce EMA data will enable new breakthroughs in understanding the precise role of EMA in diffusion models and finding principled ways to set the EMA length\,---\,possibly on a per-layer or per-parameter basis.

\ifacks
\vspace{-2mm}%
\paragraph{Acknowledgments.}
We thank
  Eric Chan,
  Qinsheng Zhang,
  Erik H\"{a}rk\"{o}nen,
  Tuomas Kynk\"{a}\"{a}nniemi,
  Arash Vahdat,
  \mbox{Ming-Yu} Liu,
  and David Luebke
  for discussions and comments,
  and Tero Kuosmanen and Samuel Klenberg for maintaining our compute infrastructure.
\fi

{\small
  \bibliographystyle{ieeenat_fullname}
  \bibliography{paper}
}

\ifappendix
  \clearpage
  \appendix
  \newcommand{\refpaper}[1]{\autoref{#1}}
  \newcommand{\figEmaComparison}{%
\begin{figure}[t]%
\centering\footnotesize%
\def\pct{\makebox[0mm][l]{\raisebox{0.15mm}{\scalebox{0.75}{\%}}}}
\def\CA{C1}\def\CB{C2}\def\CC{C4}\def\CD{C6}

\begin{tikzpicture}%
\begin{axis}[
  width={1.05\columnwidth}, height={50mm}, grid={major},
  xmin={0.01}, xmax={0.25}, xmode={linear}, xtick={0.02, 0.04, 0.06, 0.08, 0.10, 0.12, 0.14, 0.16, 0.18, 0.20, 0.22, 0.24}, xticklabels={},
  ymin={1.6}, ymax={8}, ymode={linear}, ytick={2, 3, 4, 5, 6, 7, 8, 9}, yticklabels={$2$, $3$, $4$, $5$, $6$, $7$, \raisebox{-1.5ex}[0ex][0ex]{\hspace*{5.55mm}\bf FID}},
  legend pos={south east}, legend cell align={left},
]

\gdef\pcurve##1##2{\addplot[##1, forget plot] coordinates {\rawdata{##2-lo}}; \addlegendimage{##1, mark=*}}
\pcurve{\CA}{img512-\cidB-phema-2147}
\pcurve{\CB}{img512-\cidG-guid1.0-2147}
\pcurve{\CC}{img512-\cidG-guid1.4-2147}
\pcurve{\CD}{img512-\cidG-guid1.9-2147}

\gdef\pdot##1##2##3##4{\addplot[##1, mark=*, forget plot, nodes near coords align={##3}, nodes near coords=\datalabel{\hs{##4}$\data{##2-fid2}$}] coordinates {(\rawdata{##2-phema3}, \rawdata{##2-fid4})};}
\pdot{\CA}{img512-\cidB-phema-2147}{north}{0}
\pdot{\CB}{img512-\cidG-guid1.0-2147}{north}{0}
\pdot{\CC}{img512-\cidG-guid1.4-2147}{north}{0}
\pdot{\CD}{img512-\cidG-guid1.9-2147}{north}{4}

\end{axis}
\end{tikzpicture}%
\vspace{-1mm}

\begin{tikzpicture}%
\begin{axis}[
  width={1.05\columnwidth}, height={50mm}, grid={major},
  xmin={0.01}, xmax={0.25}, xmode={linear}, xtick={0.02, 0.04, 0.06, 0.08, 0.10, 0.12, 0.14, 0.16, 0.18, 0.20, 0.22, 0.24}, xticklabels={\makebox[0mm][r]{EMA $=$ }$2\pct$, $4\pct$, $6\pct$, $8\pct$, $10\pct$, $12\pct$, $14\pct$, $16\pct$, $18\pct$, $20\pct$, $22\pct$, $24\pct$},
  ymin={33}, ymax={225}, ymode={linear}, ytick={50, 100, 150, 200, 225}, yticklabels={$50$, $100$, $150$, $200$, \raisebox{-1.5ex}[0ex][0ex]{\bf FD\textsubscript{DINOv2}}},
  legend pos={north east}, legend cell align={left},
]

\addplot[\CA, forget plot] coordinates {(0.010, 255.5) (0.015, 220.9) (0.020, 214.1) (0.025, 212.0) (0.030, 210.6) (0.035, 209.3) (0.040, 208.5) (0.045, 207.8) (0.050, 207.2) (0.055, 206.2) (0.060, 206.2) (0.065, 205.4) (0.070, 204.6) (0.075, 204.4) (0.080, 204.3) (0.085, 204.1) (0.090, 204.4) (0.095, 204.3) (0.100, 204.2) (0.105, 204.1) (0.110, 204.3) (0.115, 204.7) (0.120, 205.2) (0.125, 205.7) (0.130, 206.4) (0.135, 207.4) (0.140, 208.3) (0.145, 209.5) (0.150, 210.8) (0.155, 212.2) (0.160, 214.0) (0.165, 216.5) (0.170, 218.8) (0.175, 221.6) (0.180, 224.9) (0.185, 228.3) (0.190, 232.1) (0.195, 236.9) (0.200, 241.5) (0.205, 247.4) (0.210, 253.9) (0.215, 261.9) (0.220, 270.3) (0.225, 280.3) (0.230, 292.1) (0.235, 305.2) (0.240, 320.7) (0.245, 338.7) (0.250, 359.2)};
\addlegendimage{\CA, mark=*}

\addplot[\CB, forget plot] coordinates {(0.010, 245.67734895458477) (0.015, 209.59427679320163) (0.020, 194.7800449) (0.025, 184.9512744) (0.030, 178.2848995) (0.035, 171.5736682) (0.040, 165.937242) (0.045, 159.9709675) (0.050, 155.0322713) (0.055, 150.254791) (0.060, 145.9646027) (0.065, 141.3422053) (0.070, 136.1168106) (0.075, 132.0395664) (0.080, 127.1365723) (0.085, 122.4184389) (0.090, 119.0798621) (0.095, 114.8968355) (0.100, 110.9264593) (0.105, 106.7236166) (0.110, 102.9236033) (0.115, 99.33433975) (0.120, 95.71377467) (0.125, 92.35991844) (0.130, 89.32306645) (0.135, 86.44217565) (0.140, 83.32399346) (0.145, 80.78309833) (0.150, 78.31921125) (0.155, 76.36830928) (0.160, 74.36582442) (0.165, 72.44087775) (0.170, 71.09620372) (0.175, 69.92437514) (0.180, 69.29365131) (0.185, 68.71862621) (0.190, 68.64122764) (0.195, 69.09198635) (0.200, 69.72389894) (0.205, 71.23311762) (0.210, 72.61970087) (0.215, 74.97277588) (0.220, 77.79321161) (0.225, 81.25335583) (0.230, 85.49695641) (0.235, 89.84939198) (0.240, 95.76166316) (0.245, 102.3414038) (0.250, 109.5955508)};
\addlegendimage{\CB, mark=*}

\addplot[\CC, forget plot] coordinates {(0.010, 122.0) (0.015, 99.44) (0.020, 92.44) (0.025, 88.15) (0.030, 85.55919673) (0.035, 82.53414396) (0.040, 79.96909925) (0.045, 77.29403836) (0.050, 75.33801337) (0.055, 73.44424448) (0.060, 71.78741993) (0.065, 69.79247268) (0.070, 67.99259105) (0.075, 66.67480927) (0.080, 64.89783057) (0.085, 63.4390123) (0.090, 62.40139864) (0.095, 61.24180894) (0.100, 60.24486052) (0.105, 59.30653585) (0.110, 58.45526846) (0.115, 57.56855758) (0.120, 56.87985362) (0.125, 56.20583891) (0.130, 55.74018351) (0.135, 55.4581534) (0.140, 55.23115415) (0.145, 55.3788582) (0.150, 55.44348667) (0.155, 55.6530641) (0.160, 56.14783333) (0.165, 56.73398842) (0.170, 57.67330897) (0.175, 58.65162435) (0.180, 60.16896098) (0.185, 61.65999506) (0.190, 63.53509679) (0.195, 65.49626341) (0.200, 67.90631503) (0.205, 70.64366733) (0.210, 73.4470073) (0.215, 76.93102845) (0.220, 80.68950306) (0.225, 85.00984277) (0.230, 89.79996725) (0.235, 94.68154387) (0.240, 101.1354689) (0.245, 107.7111047) (0.250, 115.356833)};
\addlegendimage{\CC, mark=*}

\addplot[\CD, forget plot] coordinates {(0.010, 79.42) (0.015, 63.04) (0.020, 59.77) (0.025, 57.95) (0.030, 57.05) (0.035, 56.09) (0.040, 55.42) (0.045, 54.63) (0.050, 54.21) (0.055, 53.81) (0.060, 53.54) (0.065, 53.17) (0.070, 52.82) (0.075, 52.72) (0.080, 52.43) (0.085, 52.32) (0.090, 52.36) (0.095, 52.46) (0.100, 52.66) (0.105, 52.85) (0.110, 53.2) (0.115, 53.6) (0.120, 54.12) (0.125, 54.79) (0.130, 55.3) (0.135, 56.28) (0.140, 56.99) (0.145, 58.06) (0.150, 59.23) (0.155, 60.38) (0.160, 61.66) (0.165, 63.43) (0.170, 65.09) (0.175, 66.72) (0.180, 68.72) (0.185, 70.94) (0.190, 73.2) (0.195, 75.67) (0.200, 78.27) (0.205, 81.59) (0.210, 84.54) (0.215, 88.34) (0.220, 92.12) (0.225, 96.69) (0.230, 101.6) (0.235, 106.5) (0.240, 113.1) (0.245, 119.7) (0.250, 127.3)};
\addlegendimage{\CD, mark=*}

\addplot[\CA, mark=*, forget plot, nodes near coords align={north}, nodes near coords=\datalabel{$204.1$}] coordinates {(0.085, 204.1)};
\addplot[\CB, mark=*, forget plot, nodes near coords align={north}, nodes near coords=\raisebox{-2mm}{\datalabel{$68.64$}}] coordinates {(0.190, 68.64)};
\addplot[\CC, mark=*, forget plot, nodes near coords align={north}, nodes near coords=\datalabel{$55.23$}] coordinates {(0.140, 55.23)};
\addplot[\CD, mark=*, forget plot, nodes near coords align={north}, nodes near coords=\datalabel{$52.32$}] coordinates {(0.085, 52.32)};

\legend{\textsc{Config B}, \textsc{Config G}, + guidance 1.4, + guidance 1.9}
\end{axis}
\end{tikzpicture}%
\vspace{-1mm}%

\caption{\label{figEmaComparison}%
FID and FD\textsubscript{DINOv2} as a function of EMA length using S-sized models on ImageNet-512. 
\textsc{Configs B} and \textsc{G} illustrate the improvement from our changes.
We also show two guidance strengths: FID's optimum (1.4) and FD\textsubscript{DINOv2}'s optimum (1.9).
}%
\vspace{-1mm}%
\end{figure}
}%

\newcommand{\tabDinoFiveTwelve}{%
\begin{table}[t]%
\centering\footnotesize%
\resizebox{\linewidth}{!}{%
\begin{tabu}{|l@{\hs{1.5}}l|r@{\hs{3.5}}c|r@{\hs{4.5}}r@{\hs{3.1}}r@{\hs{2.4}}|}
\tabucline{-}
\multicolumn{2}{|l|}{\multirow{2}{*}{\bf ImageNet-512}} & \multicolumn{2}{c|}{\tstrut \bf FD\textsubscript{DINOv2} $\downarrow$} & \multicolumn{3}{c|}{\bf Model size} \\
&& \multicolumn{1}{c}{no CFG} & \multicolumn{1}{@{}c|}{w/CFG} & \multicolumn{1}{c}{Mparams} & \multicolumn{1}{@{}c}{Gflops} & \multicolumn{1}{@{}c|}{NFE} \\
\tabucline{-}\tstrut%
\gdef\cid{\cidXS}{\OURS-XS}    & {} & {103.39}  & {79.94} & {\data{\did-\cid-mparams0}} & {\data{\did-\cid-gflops0}} & {\data{\did-\cid-nfe}} \\
\gdef\cid{\cidS}{\OURS-S}      & {} & {68.64}   & {52.32} & {\data{\did-\cid-mparams0}} & {\data{\did-\cid-gflops0}} & {\data{\did-\cid-nfe}} \\
\gdef\cid{\cidMD}{\OURS-M}     & {} & {58.44}   & {41.98} & {\data{\did-\cid-mparams0}} & {\data{\did-\cid-gflops0}} & {\data{\did-\cid-nfe}} \\
\gdef\cid{\cidLD}{\OURS-L}     & {} & {52.25}   & {38.20} & {\data{\did-\cid-mparams0}} & {\data{\did-\cid-gflops0}} & {\data{\did-\cid-nfe}} \\
\gdef\cid{\cidXLD}{\OURS-XL}   & {} & {45.96}   & {35.67} & {\data{\did-\cid-mparams0}} & {\data{\did-\cid-gflops0}} & {\data{\did-\cid-nfe}} \\
\gdef\cid{\cidXXLD}{\OURS-XXL} & {} & {42.84}   & {33.09} & {\data{\did-\cid-mparams0}} & {\data{\did-\cid-gflops0}} & {\data{\did-\cid-nfe}} \\
\tabucline{-}
\end{tabu}%
}%
\caption{\label{tabDinoFiveTwelve}%
Version of \refpaper{tabResultsFiveTwelve} using FD\textsubscript{DINOv2} instead of FID on ImageNet-512.
The ``w/CFG'' and ``no CFG'' columns show the lowest FID obtained with and without classifier-free guidance, respectively.
NFE tells how many times the score function is evaluated when generating an image.
}%
\end{table}
}%

\newcommand{\tabUncondSweep}{%
\begin{table}[t]%
\centering\footnotesize%
\resizebox{\linewidth}{!}{%
\begin{tabu}{|l|@{\hs{1.2}}c@{\hs{2.5}}c@{\hs{2.5}}c@{\hs{1.5}}|@{\hs{1.2}}c@{\hs{1.2}}|}
\tabucline{-}\tstrut%
\bf Unconditional & FID $\downarrow$ & Total capacity & Sampling cost & EMA length \\
\bf model         &                  & (Gparams)      & (Tflops)      &            \\
\tabucline{-}\tstrut%
\gdef\did{img512}\gdef\cid{\cidXXLD}\gdef\rid{guid1.2-940}%
\gdef\trow##1##2{%
  ##1 &
  \data{\did-\cid-##1-\rid-fid2} &
  \eval[2]{(\rawdata{\did-\cid-mparams1} + \rawdata{\did-##2-mparams1}) / 1e3} &
  \eval[1]{(\rawdata{\did-\cid-nfe} * (\rawdata{\did-\cid-gflops1} + \rawdata{\did-##2-gflops1}) + \rawdata{\did-VAE-gflops1}) / 1e3} &
  \eval[1]{\rawdata{\did-\cid-##1-\rid-phema3} * 100}\%%
}%
\trow{XS}{\cidXS} \\
\trow{S}{\cidS} \\
\trow{M}{\cidM} \\
\trow{L}{\cidL} \\
\trow{XL}{\cidXL} \\
\trow{XXL}{\cidXXL} \\
\tabucline{-}
\end{tabu}%
}%
\caption{\label{tabUncondSweep}%
Effect of the unconditional model's size in guiding our XXL-sized ImageNet-512 model.
The total capacity and sampling cost refer to the combined cost of the XXL-sized conditional model and the chosen unconditional model.
Guidance strength of 1.2 was used in this test.
}%
\end{table}
}%

\newcommand{\figQualitySamplingScatter}{%
\begin{figure}[t]%
\centering\footnotesize%
\gdef\did{img512}%
\begin{tikzpicture}%
\begin{axis}[
  width={1.11\linewidth}, height={85mm}, grid={major},
  xmin={3}, xmax={1700}, xmode={log}, xtick={5, 10, 20, 50, 100, 200, 500, 1000}, xticklabels={$5$, $10$, $20$, $50$, $100$, $200$, $500$, $1000$},
  ymin={1.7}, ymax={30}, y coord trafo/.code=\pgfmathparse{ln(ln(##1))}, ytick={2, 3, 5, 10, 20, 30}, yticklabels={$2$, $3$, $5$, $10$, $20$, \raisebox{-1.5ex}[0ex][0ex]{FID}},
  xlabel={Sampling cost (teraflops per image), ImageNet-512}, x label style={at={(axis description cs:0.5,-0.06)}, anchor=north},
  legend pos={north west}, legend cell align={left},
]
\gdef\pdotSize{1.5pt}

\gdef\pmx##1{\addplot[black, opacity=0.08, thin, forget plot] coordinates {(##1,1.7) (##1,30)};}
\gdef\pmy##1{\addplot[black, opacity=0.08, thin, forget plot] coordinates {(3,##1) (1700,##1)};}
\pmx{4}\pmx{6}\pmx{7}\pmx{8}\pmx{9}\pmx{20}\pmx{30}\pmx{40}\pmx{50}\pmx{60}\pmx{70}\pmx{80}\pmx{90}\pmx{200}\pmx{300}\pmx{400}\pmx{600}\pmx{700}\pmx{800}\pmx{900}
\pmy{1.8}\pmy{1.9}\pmy{2.1}\pmy{2.2}\pmy{2.3}\pmy{2.4}\pmy{2.5}\pmy{2.6}\pmy{2.7}\pmy{2.8}\pmy{2.9}\pmy{4}\pmy{6}\pmy{7}\pmy{8}\pmy{9}

\gdef\pcNoGuid##1{(\rawdata{\did-##1-nfe} * \rawdata{\did-##1-gflops0} / 1e3, \rawdata{\did-##1-fid2})}
\gdef\pcPrevGuid##1{(\rawdata{\did-##1-nfe} * \rawdata{\did-##1-gflops0} * 2 / 1e3, \rawdata{\did-##1-cfg2})}
\gdef\pcNoGuidVAE##1{((\rawdata{\did-##1-nfe} * \rawdata{\did-##1-gflops0} + \rawdata{\did-VAE-gflops1}) / 1e3, \rawdata{\did-##1-fid2})}
\gdef\pcPrevGuidVAE##1{((\rawdata{\did-##1-nfe} * \rawdata{\did-##1-gflops0} * 2 + \rawdata{\did-VAE-gflops1}) / 1e3, \rawdata{\did-##1-cfg2})}
\gdef\pcOurGuidVAE##1{((\rawdata{\did-##1-nfe} * (\rawdata{\did-##1-gflops0} + \rawdata{\did-\cidXS-gflops0}) + \rawdata{\did-VAE-gflops1}) / 1e3, \rawdata{\did-##1-cfg2})}

\addplot[C2, forget plot] coordinates {\pcNoGuidVAE{\cidXS}\pcNoGuidVAE{\cidS}\pcNoGuidVAE{\cidMD}\pcNoGuidVAE{\cidLD}\pcNoGuidVAE{\cidXLD}\pcNoGuidVAE{\cidXXLD}};
\addplot[C4, forget plot] coordinates {\pcOurGuidVAE{\cidXS}\pcOurGuidVAE{\cidS}\pcOurGuidVAE{\cidMD}\pcOurGuidVAE{\cidLD}\pcOurGuidVAE{\cidXLD}\pcOurGuidVAE{\cidXXLD}};

\gdef\pdot##1##2##3##4{\addplot[##1, mark=*, mark size=\pdotSize, forget plot, nodes near coords align={##2}, nodes near coords=\textlabel{##3}] coordinates {##4};}
\pdot{C0}{east}{ADM}{\pcNoGuid{ADM}}
\pdot{C1}{east}{ADM}{\pcPrevGuid{ADM}}
\pdot{C0}{east}{ADM-U}{\pcNoGuid{ADM-U}}
\pdot{C1}{east}{ADM-U}{\pcPrevGuid{ADM-U}}
\pdot{C0}{west}{DiT-XL/2}{\pcNoGuidVAE{DiT-XL}}
\pdot{C1}{west}{DiT-XL/2}{\pcPrevGuidVAE{DiT-XL}}
\pdot{C0}{east}{RIN}{\pcNoGuid{RIN}}
\pdot{C0}{east}{U-ViT, L}{\pcNoGuid{U-ViT-L}}
\pdot{C0}{east}{VDM++}{\pcNoGuid{VDMpp}}
\pdot{C1}{east}{VDM++}{\pcPrevGuid{VDMpp}}
\pdot{C2}{south}{XS}{\pcNoGuidVAE{\cidXS}}
\pdot{C2}{north}{S}{\pcNoGuidVAE{\cidS}}
\pdot{C2}{south}{M}{\pcNoGuidVAE{\cidMD}}
\pdot{C2}{south}{L}{\pcNoGuidVAE{\cidLD}}
\pdot{C2}{south}{XL}{\pcNoGuidVAE{\cidXLD}}
\pdot{C2}{south}{\hs{2}XXL}{\pcNoGuidVAE{\cidXXLD}}
\pdot{C4}{south}{XS}{\pcOurGuidVAE{\cidXS}}
\pdot{C4}{north}{S}{\pcOurGuidVAE{\cidS}}
\pdot{C4}{north}{M}{\pcOurGuidVAE{\cidMD}}
\pdot{C4}{north}{L}{\pcOurGuidVAE{\cidLD}}
\pdot{C4}{north}{XL}{\pcOurGuidVAE{\cidXLD}}
\pdot{C4}{north}{\hs{2}XXL}{\pcOurGuidVAE{\cidXXLD}}

\addlegendimage{C0, only marks, mark size=\pdotSize}\addlegendentry{Previous, no guidance}
\addlegendimage{C1, only marks, mark size=\pdotSize}\addlegendentry{Previous, with guidance}
\addlegendimage{C2, mark=*, mark size=\pdotSize}\addlegendentry{Ours, no guidance}
\addlegendimage{C4, mark=*, mark size=\pdotSize}\addlegendentry{Ours, with guidance}

\end{axis}
\end{tikzpicture}%
\caption{\label{figQualitySamplingScatter}%
FID vs. sampling cost on ImageNet-512.
For latent diffusion models (DiT-XL/2 and ours), we include the cost of running the VAE decoder at the end (\data{\did-VAE-gflops1} gigaflops per image).
}%
\end{figure}
}%

\newcommand{\figQualityMparamsScatter}{%
\begin{figure}[t]%
\centering\footnotesize%
\begin{tikzpicture}%
\begin{axis}[
  width={1.11\linewidth}, height={85mm}, grid={major},
  xmin={110}, xmax={3600}, xmode={log}, xtick={200, 500, 1000, 2000}, xticklabels={$200$, $500$, $1000$, $2000$},
  ymin={1.7}, ymax={30}, y coord trafo/.code=\pgfmathparse{ln(ln(##1))}, ytick={2, 3, 5, 10, 20, 30}, yticklabels={$2$, $3$, $5$, $10$, $20$, \raisebox{-1.5ex}[0ex][0ex]{FID}},
  xlabel={Model capacity (millions of trainable parameters), ImageNet-512}, x label style={at={(axis description cs:0.5,-0.06)}, anchor=north},
  legend pos={north west}, legend cell align={left},
]
\gdef\did{img512}
\gdef\pdotSize{1.5pt}

\gdef\pmx##1{\addplot[black, opacity=0.08, thin, forget plot] coordinates {(##1,1.7) (##1,30)};}
\gdef\pmy##1{\addplot[black, opacity=0.08, thin, forget plot] coordinates {(110,##1) (3600,##1)};}
\pmx{300}\pmx{400}\pmx{600}\pmx{700}\pmx{800}\pmx{900}\pmx{3000}
\pmy{1.8}\pmy{1.9}\pmy{2.1}\pmy{2.2}\pmy{2.3}\pmy{2.4}\pmy{2.5}\pmy{2.6}\pmy{2.7}\pmy{2.8}\pmy{2.9}\pmy{4}\pmy{6}\pmy{7}\pmy{8}\pmy{9}

\gdef\pcNoGuid##1{(\rawdata{\did-##1-mparams0}, \rawdata{\did-##1-fid2})}
\gdef\pcPrevGuid##1{(\rawdata{\did-##1-mparams0}, \rawdata{\did-##1-cfg2})}
\gdef\pcOurGuid##1{(\rawdata{\did-##1-mparams0} + \rawdata{\did-\cidXS-mparams0}, \rawdata{\did-##1-cfg2})}

\addplot[C2, forget plot] coordinates {\pcNoGuid{\cidXS}\pcNoGuid{\cidS}\pcNoGuid{\cidMD}\pcNoGuid{\cidLD}\pcNoGuid{\cidXLD}\pcNoGuid{\cidXXLD}};
\addplot[C4, forget plot] coordinates {\pcOurGuid{\cidXS}\pcOurGuid{\cidS}\pcOurGuid{\cidMD}\pcOurGuid{\cidLD}\pcOurGuid{\cidXLD}\pcOurGuid{\cidXXLD}};

\gdef\pdot##1##2##3##4{\addplot[##1, mark=*, mark size=\pdotSize, forget plot, nodes near coords align={##2}, nodes near coords=\textlabel{##3}] coordinates {##4};}
\pdot{C0}{west}{ADM}{\pcNoGuid{ADM}}
\pdot{C1}{west}{ADM}{\pcPrevGuid{ADM}}
\pdot{C0}{west}{ADM-U}{\pcNoGuid{ADM-U}}
\pdot{C1}{west}{ADM-U}{\pcPrevGuid{ADM-U}}
\pdot{C0}{west}{DiT-XL/2}{\pcNoGuid{DiT-XL}}
\pdot{C1}{west}{DiT-XL/2}{\pcPrevGuid{DiT-XL}}
\pdot{C0}{west}{RIN}{\pcNoGuid{RIN}}
\pdot{C0}{east}{U-ViT, L}{\pcNoGuid{U-ViT-L}}
\pdot{C0}{east}{VDM++}{\pcNoGuid{VDMpp}}
\pdot{C1}{east}{VDM++}{\pcPrevGuid{VDMpp}}
\pdot{C1}{north}{StyleGAN-XL}{\pcPrevGuid{StyleGAN-XL}}
\pdot{C2}{south}{XS}{\pcNoGuid{\cidXS}}
\pdot{C2}{north}{S}{\pcNoGuid{\cidS}}
\pdot{C2}{south}{M}{\pcNoGuid{\cidMD}}
\pdot{C2}{south}{L}{\pcNoGuid{\cidLD}}
\pdot{C2}{south}{XL}{\pcNoGuid{\cidXLD}}
\pdot{C2}{south}{XXL}{\pcNoGuid{\cidXXLD}}
\pdot{C4}{south}{XS}{\pcOurGuid{\cidXS}}
\pdot{C4}{north}{S}{\pcOurGuid{\cidS}}
\pdot{C4}{north}{M}{\pcOurGuid{\cidMD}}
\pdot{C4}{north}{L}{\pcOurGuid{\cidLD}}
\pdot{C4}{north}{XL}{\pcOurGuid{\cidXLD}}
\pdot{C4}{north}{XXL}{\pcOurGuid{\cidXXLD}}

\addlegendimage{C0, only marks, mark size=\pdotSize}\addlegendentry{Previous, no guidance}
\addlegendimage{C1, only marks, mark size=\pdotSize}\addlegendentry{Previous, with guidance}
\addlegendimage{C2, mark=*, mark size=\pdotSize}\addlegendentry{Ours, no guidance}
\addlegendimage{C4, mark=*, mark size=\pdotSize}\addlegendentry{Ours, with guidance}

\end{axis}
\end{tikzpicture}%
\caption{\label{figQualityMparamsScatter}%
FID vs. model capacity on ImageNet-512.
For our method with guidance, we account for the number of parameters in the XS-sized unconditional model.
}%
\end{figure}
}%

\newcommand{\figQualityTrainingScatter}{%
\begin{figure}[t]%
\centering\footnotesize%
\begin{tikzpicture}%
\def\CA{C0}\def\CB{C2!70!white}\def\CC{C2!70!black}
\begin{axis}[
  width={1.11\linewidth}, height={85mm}, grid={major},
  xmin={0}, xmax={3.1}, xmode={linear}, xtick={0.0, 0.5, 1.0, 1.5, 2.0, 2.5, 3.0}, xticklabels={$0.0$, $0.5$, $1.0$, $1.5$, $2.0$, $2.5$, $3.0$},
  ymin={1.8}, ymax={27}, y coord trafo/.code=\pgfmathparse{ln(ln(##1))}, ytick={2, 3, 5, 10, 20, 27}, yticklabels={$2$, $3$, $5$, $10$, $20$, \raisebox{-1.5ex}[0ex][0ex]{FID}},
  xlabel={Training cost (zettaflops per model), ImageNet-512}, x label style={at={(axis description cs:0.5,-0.06)}, anchor=north},
  legend pos={north west}, legend cell align={left},
]
\gdef\did{img512}
\gdef\pdotSize{1.5pt}

\gdef\pmy##1{\addplot[black, opacity=0.08, thin, forget plot] coordinates {(0,##1) (3.1,##1)};}
\pmy{1.9}\pmy{2.1}\pmy{2.2}\pmy{2.3}\pmy{2.4}\pmy{2.5}\pmy{2.6}\pmy{2.7}\pmy{2.8}\pmy{2.9}\pmy{4}\pmy{6}\pmy{7}\pmy{8}\pmy{9}

\gdef\pcFinal##1{(\rawdata{\did-##1-mimg0} * \rawdata{\did-##1-gflops0} * 3 / 1e6, \rawdata{\did-##1-fid2})}
\gdef\pcProg##1##2{(##2 * \rawdata{\did-##1-gflops0} * 3 / 1e6, \rawdata{\did-##1-phema-##2-fid2})}

\addplot[\CB, forget plot] coordinates {\pcProg{\cidXS}{134}\pcProg{\cidXS}{268}\pcProg{\cidXS}{537}\pcProg{\cidXS}{1074}\pcProg{\cidXS}{1611}\pcProg{\cidXS}{2147}};
\addplot[\CB, forget plot] coordinates {\pcProg{\cidS}{134}\pcProg{\cidS}{268}\pcProg{\cidS}{537}\pcProg{\cidS}{805}\pcProg{\cidS}{1074}\pcProg{\cidS}{1342}\pcProg{\cidS}{1611}\pcProg{\cidS}{1879}\pcProg{\cidS}{2147}};
\addplot[\CB, forget plot] coordinates {\pcProg{\cidM}{134}\pcProg{\cidM}{268}\pcProg{\cidM}{537}\pcProg{\cidM}{805}\pcProg{\cidM}{1074}\pcProg{\cidM}{1342}\pcProg{\cidM}{1611}\pcProg{\cidM}{1879}};
\addplot[\CB, forget plot] coordinates {\pcProg{\cidL}{134}\pcProg{\cidL}{268}\pcProg{\cidL}{537}\pcProg{\cidL}{805}\pcProg{\cidL}{1074}};
\addplot[\CB, forget plot] coordinates {\pcProg{\cidXL}{134}\pcProg{\cidXL}{268}\pcProg{\cidXL}{537}\pcProg{\cidXL}{671}\pcProg{\cidXL}{805}};
\addplot[\CB, forget plot] coordinates {\pcProg{\cidXXL}{134}\pcProg{\cidXXL}{268}\pcProg{\cidXXL}{403}\pcProg{\cidXXL}{537}};

\addplot[\CC, forget plot] coordinates {\pcProg{\cidXSD}{134}\pcProg{\cidXSD}{268}\pcProg{\cidXSD}{537}\pcProg{\cidXSD}{1074}\pcProg{\cidXSD}{1611}\pcProg{\cidXSD}{2147}};
\addplot[\CC, forget plot] coordinates {\pcProg{\cidSD}{134}\pcProg{\cidSD}{268}\pcProg{\cidSD}{537}\pcProg{\cidSD}{805}\pcProg{\cidSD}{1074}\pcProg{\cidSD}{1342}\pcProg{\cidSD}{1611}\pcProg{\cidSD}{1879}\pcProg{\cidSD}{2147}};
\addplot[\CC, forget plot] coordinates {\pcProg{\cidMD}{134}\pcProg{\cidMD}{268}\pcProg{\cidMD}{537}\pcProg{\cidMD}{805}\pcProg{\cidMD}{1074}\pcProg{\cidMD}{1342}\pcProg{\cidMD}{1611}\pcProg{\cidMD}{1879}\pcProg{\cidMD}{2147}};
\addplot[\CC, forget plot] coordinates {\pcProg{\cidLD}{134}\pcProg{\cidLD}{268}\pcProg{\cidLD}{537}\pcProg{\cidLD}{805}\pcProg{\cidLD}{1074}\pcProg{\cidLD}{1342}\pcProg{\cidLD}{1611}\pcProg{\cidLD}{1879}};
\addplot[\CC, forget plot] coordinates {\pcProg{\cidXLD}{134}\pcProg{\cidXLD}{268}\pcProg{\cidXLD}{403}\pcProg{\cidXLD}{537}\pcProg{\cidXLD}{671}\pcProg{\cidXLD}{805}\pcProg{\cidXLD}{940}\pcProg{\cidXLD}{1074}\pcProg{\cidXLD}{1208}\pcProg{\cidXLD}{1342}};
\addplot[\CC, forget plot] coordinates {\pcProg{\cidXXLD}{134}\pcProg{\cidXXLD}{268}\pcProg{\cidXXLD}{403}\pcProg{\cidXXLD}{537}\pcProg{\cidXXLD}{671}\pcProg{\cidXXLD}{805}\pcProg{\cidXXLD}{940}};

\gdef\pdot##1##2##3##4{\addplot[##1, mark=*, mark size=\pdotSize, forget plot, nodes near coords align={##2}, nodes near coords=\textlabel{##3}] coordinates {##4};}
\gdef\pdott##1##2##3##4{\addplot[opacity=0.8, forget plot, nodes near coords align={##2}, nodes near coords=\textlabel{##3}] coordinates {##4};}
\pdot{\CA}{east}{ADM}{\pcFinal{ADM}}
\pdot{\CA}{east}{ADM-U}{\pcFinal{ADM-U}}
\pdot{\CA}{west}{DiT-XL/2}{\pcFinal{DiT-XL}}
\pdot{\CA}{west}{RIN}{\pcFinal{RIN}}
\pdot{\CA}{west}{U-ViT, L}{\pcFinal{U-ViT-L}}
\pdot{\CA}{west}{VDM++}{\pcFinal{VDMpp}}
\pdot{\CB}{west}{XS}{\pcFinal{\cidXS}}
\pdot{\CB}{west}{S}{\pcFinal{\cidS}}
\pdot{\CB}{west}{M}{\pcFinal{\cidM}}
\pdot{\CB}{west}{\raisebox{-2mm}{L}}{\pcFinal{\cidL}}
\pdott{\CB}{west}{\raisebox{-2mm}{\colorbox{white}{\makebox[1.6mm]{}}}}{\pcFinal{\cidXL}}
\pdot{\CB}{west}{\hs{0.25}\raisebox{-2mm}{XL}}{\pcFinal{\cidXL}}
\pdott{\CB}{west}{\hs{-0.5}\raisebox{-5.5mm}{\colorbox{white}{\makebox[3.2mm]{}}}}{\pcFinal{\cidXXL}}
\pdot{\CB}{west}{\hs{-0.25}\raisebox{-5.5mm}{XXL}}{ \pcFinal{\cidXXL}}
\pdot{\CC}{west}{XS}{\pcFinal{\cidXSD}}
\pdot{\CC}{west}{S}{\pcFinal{\cidSD}}
\pdot{\CC}{west}{M}{\pcFinal{\cidMD}}
\pdot{\CC}{west}{L}{\pcFinal{\cidLD}}
\pdot{\CC}{west}{XL}{\pcFinal{\cidXLD}}
\pdot{\CC}{north}{XXL}{\pcFinal{\cidXXLD}}

\addlegendimage{\CA, only marks, mark size=\pdotSize}\addlegendentry{Previous}
\addlegendimage{\CB, mark=*, mark size=\pdotSize}\addlegendentry{Ours, no dropout}
\addlegendimage{\CC, mark=*, mark size=\pdotSize}\addlegendentry{Ours, 10\% dropout}

\end{axis}
\end{tikzpicture}%
\caption{\label{figQualityTrainingScatter}%
FID vs. training cost on ImageNet-512 without guidance.
Note that one zettaflop $= 10^{21}$ flops $= 10^{12}$ gigaflops.
We assume that one training iteration is three times as expensive as evaluating the model (i.e., forward pass, backprop to inputs, backprop to weights).
}%
\end{figure}
}%

\newcommand{\plotEmaEvolutionXXL}{%
\centering\footnotesize%
\begin{tikzpicture}%
\def\pct{\makebox[0mm][l]{\raisebox{0.15mm}{\scalebox{0.75}{\%}}}}
\def\CA{C2!40!C10}\def\CB{C2!55!C10}\def\CC{C2!75!C10}\def\CD{C2!99!C11}\def\CE{C2!75!C11}\def\CF{C2!60!C11}
\begin{axis}[
  width={1.05\columnwidth}, height={60mm}, grid={major},
  xmin={0.01}, xmax={0.18}, xmode={linear}, xtick={0.02, 0.04, 0.06, 0.08, 0.10, 0.12, 0.14, 0.16}, xticklabels={\makebox[0mm][r]{EMA $=$ }$2\pct$, $4\pct$, $6\pct$, $8\pct$, $10\pct$, $12\pct$, $14\pct$, $16\pct$},
  ymin={1.7}, ymax={3.7}, ymode={linear}, ytick={2.0, 2.5, 3.0, 3.5, 3.7}, yticklabels={$2.0$, $2.5$, $3.0$, $3.5$, \raisebox{-1.5ex}[0ex][0ex]{FID}},
  legend pos={south east}, legend cell align={left},
]
\gdef\did{img512}\gdef\cid{\cidXXL}

\gdef\pcurve##1##2{\addplot[##1, forget plot] coordinates {\rawdata{\did-\cid-phema-##2-lo}}; \addlegendimage{##1, mark=*}}
\pcurve{\CA}{268}\pcurve{\CB}{403}\pcurve{\CC}{537}\pcurve{\CD}{671}\pcurve{\CE}{805}\pcurve{\CF}{940}

\gdef\pdot##1##2##3##4{\addplot[##1, mark=*, forget plot, nodes near coords align={##3}, nodes near coords=\datalabel{##4}] coordinates {(\rawdata{\did-\cid-phema-##2-phema3}, \rawdata{\did-\cid-phema-##2-fid4})};}
\pdot{\CA}{268}{south}{\hs{-2}$\data{\did-\cid-phema-268-fid2}$}
\pdot{\CB}{403}{south}{\hs{1}$\data{\did-\cid-phema-403-fid2}$}
\pdot{\CC}{537}{north}{\hs{-1}$\data{\did-\cid-phema-537-fid2}$}
\pdot{\CD}{671}{north}{\hs{2}$\data{\did-\cid-phema-671-fid2}$}
\pdot{\CE}{805}{north}{$\data{\did-\cid-phema-805-fid2}$}
\pdot{\CF}{940}{north}{\hs{7}$\data{\did-\cid-phema-940-fid2}$}

\legend{268M, 403M, 537M, 671M, 805M, 940M}
\end{axis}
\end{tikzpicture}%
}%

\newcommand{\plotEmaEvolutionXXLD}{%
\centering\footnotesize%
\begin{tikzpicture}%
\def\pct{\makebox[0mm][l]{\raisebox{0.15mm}{\scalebox{0.75}{\%}}}}
\def\CA{C2!40!C10}\def\CB{C2!55!C10}\def\CC{C2!75!C10}\def\CD{C2!99!C11}\def\CE{C2!75!C11}\def\CF{C2!60!C11}
\begin{axis}[
  width={1.05\columnwidth}, height={60mm}, grid={major},
  xmin={0.01}, xmax={0.18}, xmode={linear}, xtick={0.02, 0.04, 0.06, 0.08, 0.10, 0.12, 0.14, 0.16}, xticklabels={\makebox[0mm][r]{EMA $=$ }$2\pct$, $4\pct$, $6\pct$, $8\pct$, $10\pct$, $12\pct$, $14\pct$, $16\pct$},
  ymin={1.7}, ymax={3.7}, ymode={linear}, ytick={2.0, 2.5, 3.0, 3.5, 3.7}, yticklabels={$2.0$, $2.5$, $3.0$, $3.5$, \raisebox{-1.5ex}[0ex][0ex]{FID}},
  legend pos={south east}, legend cell align={left},
]
\gdef\did{img512}\gdef\cid{\cidXXLD}

\gdef\pcurve##1##2{\addplot[##1, forget plot] coordinates {\rawdata{\did-\cid-phema-##2-lo}}; \addlegendimage{##1, mark=*}}
\pcurve{\CA}{268}\pcurve{\CB}{403}\pcurve{\CC}{537}\pcurve{\CD}{671}\pcurve{\CE}{805}\pcurve{\CF}{940}

\gdef\pdot##1##2##3##4{\addplot[##1, mark=*, forget plot, nodes near coords align={##3}, nodes near coords=\datalabel{##4}] coordinates {(\rawdata{\did-\cid-phema-##2-phema3}, \rawdata{\did-\cid-phema-##2-fid4})};}
\pdot{\CA}{268}{south}{\hs{1}$\data{\did-\cid-phema-268-fid2}$}
\pdot{\CB}{403}{south}{\hs{-1}$\data{\did-\cid-phema-403-fid2}$}
\pdot{\CC}{537}{south}{\hs{3}$\data{\did-\cid-phema-537-fid2}$}
\pdot{\CD}{671}{north}{\hs{-3}$\data{\did-\cid-phema-671-fid2}$}
\pdot{\CE}{805}{north}{\hs{1.5}$\data{\did-\cid-phema-805-fid2}$}
\pdot{\CF}{940}{north}{\hs{2}$\data{\did-\cid-phema-940-fid2}$}

\legend{268M, 403M, 537M, 671M, 805M, 940M}
\end{axis}
\end{tikzpicture}%
}%

\newcommand{\figOverfitting}{%
\begin{figure*}[t]%
\begin{subfigure}{0.5\linewidth}\plotEmaEvolutionXXL\vspace{1mm}\caption{\small \OURS-XXL, no dropout}\end{subfigure}\hfill%
\begin{subfigure}{0.5\linewidth}\plotEmaEvolutionXXLD\vspace{1mm}\caption{\small \OURS-XXL, 10\% dropout}\end{subfigure}%
\caption{\label{figOverfitting}%
Effect of dropout on the training of \OURS-XXL in ImageNet-512.
\textbf{(a)}
Without dropout, the training starts to overfit after 537 million training images.
\textbf{(b)}
With dropout, the training starts off slightly slower, but it makes forward progress for much longer.
}%
\end{figure*}
}%

\newcommand{\figEmaLR}{%
\begin{figure}[t]%
\centering\footnotesize%
\begin{tikzpicture}%
\def\pct{\makebox[0mm][l]{\raisebox{0.15mm}{\scalebox{0.75}{\%}}}}
\def\CA{C0!70!white}\def\CB{C0}\def\CC{C0!80!black}\def\CD{C2}\def\CE{C1!80!black}\def\CF{C1}\def\CG{C1!70!white}
\begin{axis}[
  width={1.12\columnwidth}, height={60mm}, grid={major},
  xmin={0.02}, xmax={0.22}, xmode={linear}, xtick={0.04, 0.06, 0.08, 0.10, 0.12, 0.14, 0.16, 0.18, 0.20}, xticklabels={\makebox[0mm][r]{EMA $=$ }$4\pct$, $6\pct$, $8\pct$, $10\pct$, $12\pct$, $14\pct$, $16\pct$, $18\pct$, $20\pct$},
  ymin={2.3}, ymax={5}, ymode={linear}, ytick={2.5, 3.0, 3.5, 4.0, 4.5, 5.0}, yticklabels={$2.5$, $3.0$, $3.5$, $4.0$, $4.5$, \raisebox{-1.5ex}[0ex][0ex]{FID}},
  legend pos={north west}, legend cell align={left}, legend columns={7},
]
\gdef\did{img512}\gdef\rid{phema-2147}

\gdef\pshade##1##2{\fillbetween[##1, opacity=0.2, forget plot]{coordinates {\rawdata{\did-##2-\rid-lo}}}{coordinates {\rawdata{\did-##2-\rid-hi}}};}
\pshade{\CA}{b9-ref30k}\pshade{\CB}{b9-ref40k}\pshade{\CC}{b9-ref50k}\pshade{\CD}{\cidG}\pshade{\CE}{b9-ref100k}\pshade{\CF}{b9-ref120k}\pshade{\CG}{b9-ref160k}

\gdef\pcurve##1##2{\addplot[##1, forget plot] coordinates {\rawdata{\did-##2-\rid-lo}}; \addlegendimage{##1, mark=*}}
\pcurve{\CA}{b9-ref30k}\pcurve{\CB}{b9-ref40k}\pcurve{\CC}{b9-ref50k}\pcurve{\CD}{\cidG}\pcurve{\CE}{b9-ref100k}\pcurve{\CF}{b9-ref120k}\pcurve{\CG}{b9-ref160k}

\gdef\pdot##1##2##3##4{\addplot[##1, mark=*, forget plot, nodes near coords align={##3}, nodes near coords=\datalabel{##4}] coordinates {(\rawdata{\did-##2-\rid-phema3}, \rawdata{\did-##2-\rid-fid4})};}
\pdot{\CA}{b9-ref30k}{north}{$\data{\did-b9-ref30k-\rid-fid2}$}
\pdot{\CB}{b9-ref40k}{north}{\hs{2}$\data{\did-b9-ref40k-\rid-fid2}$}
\pdot{\CC}{b9-ref50k}{north}{$\data{\did-b9-ref50k-\rid-fid2}$}
\pdot{\CD}{\cidG}{north}{$\data{\did-\cidG-\rid-fid2}$}
\pdot{\CE}{b9-ref100k}{north}{$\data{\did-b9-ref100k-\rid-fid2}$}
\pdot{\CF}{b9-ref120k}{north}{$\data{\did-b9-ref120k-\rid-fid2}$}
\pdot{\CG}{b9-ref160k}{north}{\hs{-2}$\data{\did-b9-ref160k-\rid-fid2}$}

\legend{30k, 40k, 50k, 70k, 100k, 120k, 160k}
\end{axis}
\end{tikzpicture}%
\vspace{-1mm}%
\caption{\label{figEmaLR}%
Interaction between EMA length and learning rate decay ($t_\text{ref}$, different colors) using \OURS-S on ImageNet-512.
}%
\end{figure}\vspace{-1mm}
}%

\define@cmdkeys{figConfig}[]{flabel, cletter, archpage, cid, batch, mixedp, dropout, aref, tref, pmean, pstd, abeta, lscale, attres, attblk, cdesc}
\newcommand{\figConfig}[1]{\setkeys{figConfig}{#1}%
\begin{figure*}[t]%
\centering\footnotesize%
\includegraphics[width=\linewidth, trim={23.3 9 25.6 8}, clip, page=\archpage]{architecture/architecture.pdf}\\[2mm]%
\resizebox{\linewidth}{!}{%
\begin{tabu}{|p{27mm}p{9mm}|p{36mm}p{9mm}|p{19mm}p{10mm}|p{30mm}p{7mm}|}
\tabucline{-}\tstrut%
\gdef\did{img512}%
Number of GPUs                            & $32$ &
Learning rate max ($\alpha_\text{ref}$)   & $\aref$ &
Adam $\beta_1$                            & $0.9$ &
FID                                       & $\data{\did-\cid-fid2}$ \\
Minibatch size                            & $\batch$ &
Learning rate decay ($t_\text{ref}$)      & $\tref$ &
Adam $\beta_2$                            & $\abeta$ &
EMA length ($\sigma_\text{rel}$)          & $\data{\did-\cid-phema-2147-phema3}$ \\
Duration (Mimg)                           & $\data{\did-\cid-mimg1}$ &
Learning rate rampup (Mimg)               & $10$ &
Loss scaling                              & $\lscale$ &
Model capacity (Mparams)                  & $\data{\did-\cid-mparams1}$ \\
Mixed-precision (FP16)                    & \mixedp &
Noise distribution mean ($P_\text{mean}$) & $\pmean$ &
Attention res.                            & $\attres$ &
Model complexity (Gflops)                 & $\data{\did-\cid-gflops1}$ \\
Dropout probability                       & $\dropout\%$ &
Noise distribution std. ($P_\text{std}$)  & $\pstd$ &
Attention blocks                          & $\attblk$ &
Sampling cost (Tflops)                    & $\eval[2]{(\rawdata{\did-\cid-nfe} * \rawdata{\did-\cid-gflops1} + \rawdata{\did-VAE-gflops1}) / 1e3}$ \\
\tabucline{-}
\end{tabu}%
}%
\vspace{-1.5mm}%
\caption{\label{\flabel}%
Full architecture diagram and hyperparameters for \vconfig{\cletter} (\cdesc).
}%
\vspace{-2.5mm}%
\end{figure*}
}%

\newcommand{\figForcedWN}{%
\begin{figure}[t]%
\centering\footnotesize%
\includegraphics[width=1.0\linewidth]{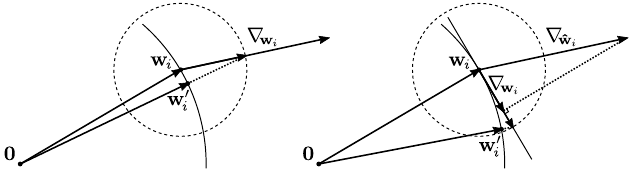}\\%
\makebox[0.55\linewidth][c]{(a) Forced WN only}%
\makebox[0.40\linewidth][c]{(b) Forced + standard WN}%
\caption{\label{figForcedWN}%
Illustration of the importance of performing ``standard'' weight normalization in addition to
  forcing the weights to a predefined norm.
The dashed circle illustrates Adam's target variance for updates\,---\,%
  the proportions are greatly exaggerated and the effects of momentum are ignored.
\textbf{(a)}
Forced weight normalization without the standard weight normalization.
The raw weight vector $\boldwi$ is updated by adding the gradient $\smash{\nnabla_{\boldwi}}$
  after being scaled by Adam, after which the result is normalized back to the hypersphere (solid arc)
  yielding new weight vector $\boldwi'$.
Adam's variance estimate includes the non-tangent component of the gradient,
  and the resulting weight update is significantly smaller than intended.
\textbf{(b)}
With standard weight normalization, the gradient $\nnabla_{\boldwi}$ is obtained by
  projecting the raw gradient $\smash{\nnabla_{\boldwhati}}$ onto the tangent plane perpendicular to $\boldwi$.
Adam's variance estimate now considers this projected gradient,
  resulting in the correct step size;
  the effect of normalization after update is close to negligible
  from a single step's perspective.
}%
\end{figure}
}%

\newcommand{\figEmaLength}{%
\begin{figure}[t]%
\centering\footnotesize%
\begin{tikzpicture}%
\begin{axis}[
  width={1.09\linewidth}, height={60mm}, grid={major},
  xmin={0}, xmax={1}, xmode={linear}, xtick={0.0, 0.2, 0.4, 0.6, 0.8, 1.0}, xticklabels={\makebox[0mm][r]{t $=$ }$0$, $0.2$, $0.4$, $0.6$, $0.8$, $1$},
  ymin={0}, ymax={1}, ymode={linear}, ytick={0.2, 0.4, 0.6, 0.8, 1.0}, yticklabels={$0.2$, $0.4$, $0.6$, $0.8$, \raisebox{-1.5ex}[0ex][0ex]{$f(t)$}},
  legend pos={north west}, legend cell align={left},
]
\addplot[C0] coordinates {\rawdata{emaprofile-exp0}}; \addlegendentry{$\srel = 0.25,\ \gamma=\s0.72$}
\addplot[C1] coordinates {\rawdata{emaprofile-exp1}}; \addlegendentry{$\srel = 0.20,\ \gamma=\s1.83$}
\addplot[C2] coordinates {\rawdata{emaprofile-exp2}}; \addlegendentry{$\srel = 0.15,\ \gamma=\s3.56$}
\addplot[C3] coordinates {\rawdata{emaprofile-exp3}}; \addlegendentry{$\srel = 0.10,\ \gamma=\s6.94$}
\addplot[C4] coordinates {\rawdata{emaprofile-exp4}}; \addlegendentry{$\srel = 0.05,\ \gamma= 16.97$}
\end{axis}
\end{tikzpicture}%
\caption{\label{figEmaLength}%
Examples of the canonical response function of our power function EMA profile (\autoref{eq:powerEma}).
Each curve corresponds to a particular choice for the relative standard deviation $\srel$; the corresponding exponent $\gamma$ is calculated using \autoref{algGamma}.
}%
\end{figure}
}%

\newcommand{\tabPickles}{%
\begin{table}[t]%
\centering\footnotesize%
\resizebox{\linewidth}{!}{%
\begin{tabu}{|l|l|l|}
\tabucline{-}\tstrut%
{\bf Cfg} & {\bf Pre-EMA pickle} & {\bf Post-EMA pickle} \\
\tabucline{-}
\multicolumn{3}{|l|}{\tstrut ImageNet-512 bridge}\gdef\did{img512}\\
\tabucline{-}\tstrut%
\gdef\prepkl##1{\texttt{\data{##1-pkl}}}%
\gdef\postpkl##1{\texttt{\data{##1-phema-\ldata{TBD}{##1-mimg0}-pkl}}}%
\gdef\cid{\cidA}{A} & \prepkl{\did-\cid} & \postpkl{\did-\cid} \\
\gdef\cid{\cidB}{B} & \prepkl{\did-\cid} & \postpkl{\did-\cid} \\
\gdef\cid{\cidC}{C} & \prepkl{\did-\cid} & \postpkl{\did-\cid} \\
\gdef\cid{\cidD}{D} & \prepkl{\did-\cid} & \postpkl{\did-\cid} \\
\gdef\cid{\cidE}{E} & \prepkl{\did-\cid} & \postpkl{\did-\cid} \\
\gdef\cid{\cidF}{F} & \prepkl{\did-\cid} & \postpkl{\did-\cid} \\
\gdef\cid{\cidG}{G} & \prepkl{\did-\cid} & \postpkl{\did-\cid} \\
\tabucline{-}
\multicolumn{3}{|l|}{\tstrut ImageNet-512 scaling, no dropout}\gdef\did{img512}\\
\tabucline{-}\tstrut%
\gdef\cid{\cidXS}{XS} & \prepkl{\did-\cid} & \postpkl{\did-\cid} \\
\gdef\cid{\cidS}{S} & \prepkl{\did-\cid} & \postpkl{\did-\cid} \\
\gdef\cid{\cidM}{M} & \prepkl{\did-\cid} & \postpkl{\did-\cid} \\
\gdef\cid{\cidL}{L} & \prepkl{\did-\cid} & \postpkl{\did-\cid} \\
\gdef\cid{\cidXL}{XL} & \prepkl{\did-\cid} & \postpkl{\did-\cid} \\
\gdef\cid{\cidXXL}{XXL} & \prepkl{\did-\cid} & \postpkl{\did-\cid} \\
\tabucline{-}
\multicolumn{3}{|l|}{\tstrut ImageNet-512 scaling, with dropout}\gdef\did{img512}\\
\tabucline{-}\tstrut%
\gdef\cid{\cidXSD}{XSD} & \prepkl{\did-\cid} & \postpkl{\did-\cid} \\
\gdef\cid{\cidSD}{SD} & \prepkl{\did-\cid} & \postpkl{\did-\cid} \\
\gdef\cid{\cidMD}{MD} & \prepkl{\did-\cid} & \postpkl{\did-\cid} \\
\gdef\cid{\cidLD}{LD} & \prepkl{\did-\cid} & \postpkl{\did-\cid} \\
\gdef\cid{\cidXLD}{XLD} & \prepkl{\did-\cid} & \postpkl{\did-\cid} \\
\gdef\cid{\cidXXLD}{XXLD} & \prepkl{\did-\cid} & \postpkl{\did-\cid} \\
\tabucline{-}
\multicolumn{3}{|l|}{\tstrut ImageNet-64 scaling, no dropout}\gdef\did{img64}\\
\tabucline{-}\tstrut%
\gdef\cid{\cidSfS}{S} & \prepkl{\did-\cid} & \postpkl{\did-\cid} \\
\gdef\cid{\cidSfM}{M} & \prepkl{\did-\cid} & \postpkl{\did-\cid} \\
\gdef\cid{\cidSfL}{L} & \prepkl{\did-\cid} & \postpkl{\did-\cid} \\
\gdef\cid{\cidSfXL}{XL} & \prepkl{\did-\cid} & \postpkl{\did-\cid} \\
\tabucline{-}
\multicolumn{3}{|l|}{\tstrut ImageNet-64 scaling, with dropout}\gdef\did{img64}\\
\tabucline{-}\tstrut%
\gdef\cid{\cidSfSD}{SD} & \prepkl{\did-\cid} & \postpkl{\did-\cid} \\
\gdef\cid{\cidSfMD}{MD} & \prepkl{\did-\cid} & \postpkl{\did-\cid} \\
\gdef\cid{\cidSfLD}{LD} & \prepkl{\did-\cid} & \postpkl{\did-\cid} \\
\gdef\cid{\cidSfXLD}{XLD} & \prepkl{\did-\cid} & \postpkl{\did-\cid} \\
\tabucline{-}
\end{tabu}%
}%
\caption{\label{tabPickles}%
Pickles corresponding to each config.
}%
\end{table}
}%

\newcommand{\figCurated}{%
\begin{figure*}[p]%
\centering\footnotesize
\includegraphics[width=.99\linewidth]{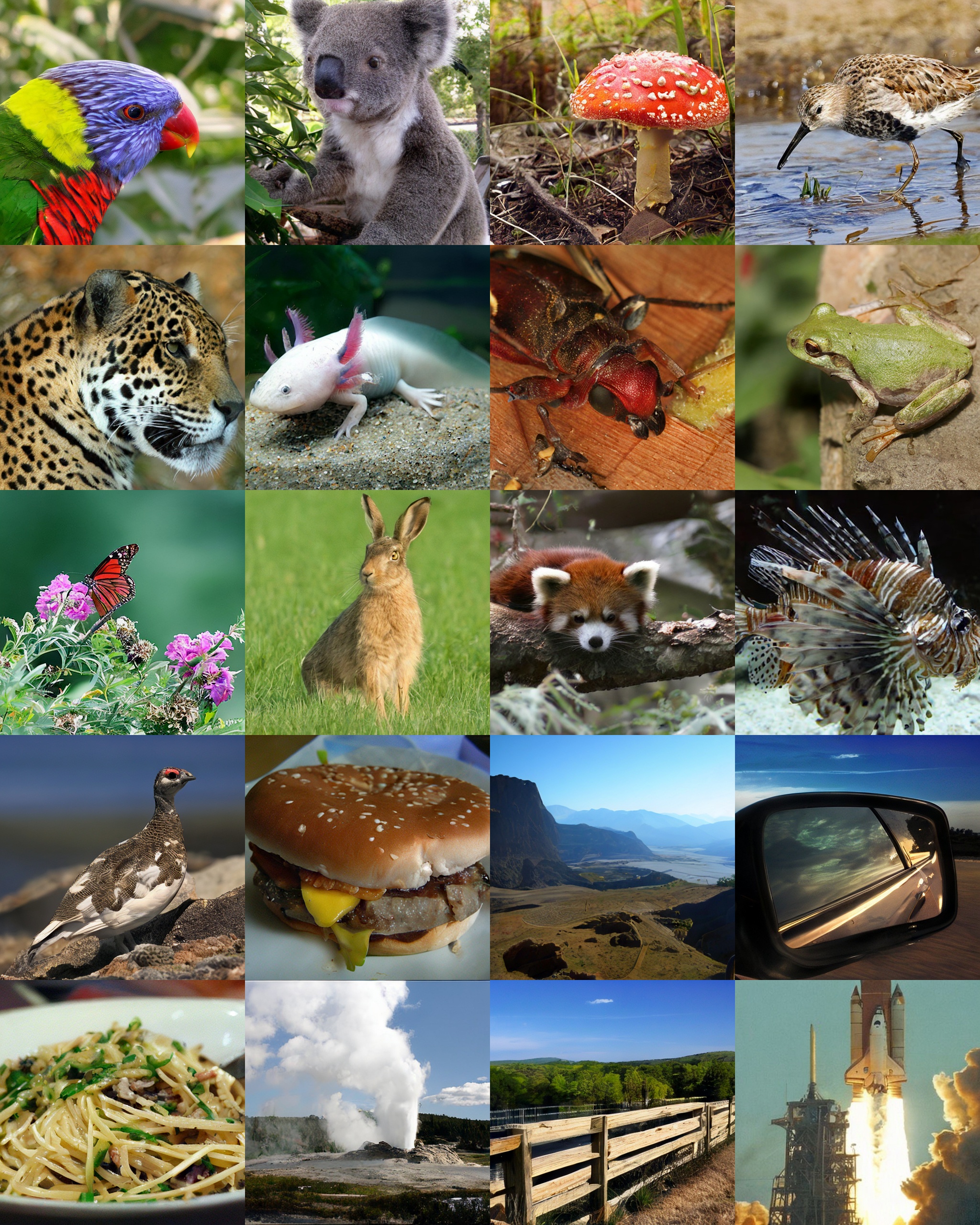}%
\vspace*{1mm}%
\caption{\label{figCurated}%
Selected images generated using our largest (XXL) ImageNet-512 model without guidance.
}%
\end{figure*}
}%

\newcommand{\cgrow}[3]{%
\rotatebox{90}{\makebox[63mm][c]{\normalsize Class #1 (#3), guidance #2}}%
&\includegraphics[width=168mm]{classgrid/grid-#1-#2.jpg}\vspace*{4mm}\\%
}
\newcommand{\cgtemplate}[2]{%
\begin{figure*}[p]%
\centering\footnotesize%
\begin{tabular}{@{}c@{\ \ }c@{}}
#1
\end{tabular}\vspace*{-2mm}%
\caption{\label{#2}%
Uncurated images generated using our largest (XXL) ImageNet-512 model.
}\end{figure*}}

\newcommand{\figClassGridA}{%
\cgtemplate{%
\cgrow{88}{2.0}{macaw}%
\cgrow{29}{1.0}{axolotl}%
\cgrow{127}{2.0}{white stork}%
}{figClassGridA}%
}%

\newcommand{\figClassGridB}{%
\cgtemplate{%
\cgrow{89}{3.0}{cockatoo}%
\cgrow{980}{1.2}{volcano}%
\cgrow{33}{2.0}{loggerhead}%
}{figClassGridB}%
}%

\newcommand{\figClassGridC}{%
\cgtemplate{%
\cgrow{15}{1.0}{robin}%
\cgrow{975}{2.0}{lakeside}%
\cgrow{279}{2.0}{arctic fox}%
}{figClassGridC}%
}%

\defdata{img512-\cidXS-mspeval3}{0.874}
\defdata{img512-\cidS-mspeval3}{1.578}
\defdata{img512-\cidMD-mspeval3}{2.387}
\defdata{img512-\cidLD-mspeval3}{3.431}
\defdata{img512-\cidXLD-mspeval3}{4.537}
\defdata{img512-\cidXXLD-mspeval3}{5.975}
\defdata{img64-\cidSfS-mspeval3}{\rawdata{img512-\cidS-mspeval3}}
\defdata{img64-\cidSfMD-mspeval3}{\rawdata{img512-\cidMD-mspeval3}}
\defdata{img64-\cidSfLD-mspeval3}{\rawdata{img512-\cidLD-mspeval3}}
\defdata{img64-\cidSfXLD-mspeval3}{\rawdata{img512-\cidXLD-mspeval3}}
\defdata{img512-VAE-mspeval3}{56.736}
\gdef\wattsPerNode{4300}

\newcommand{\tabHyperparams}{%
\begin{table*}[t]%
\centering\footnotesize%
\begin{tabu}{|l|cccccc|cccc|}
\tabucline{-}
\multirow{2}{*}{\bf Model details} & \multicolumn{6}{c|}{\tstrut ImageNet-512} & \multicolumn{4}{c|}{\tstrut ImageNet-64} \\
& XS & S & M & L & XL & XXL & S & M & L & XL \\
\tabucline{-}\tstrut%
\gdef\largeconfigs{\callback{img512-\cidXS} & \callback{img512-\cidS} & \callback{img512-\cidMD} & \callback{img512-\cidLD} & \callback{img512-\cidXLD} & \callback{img512-\cidXXLD}}%
\gdef\smallconfigs{\callback{img64-\cidSfS} & \callback{img64-\cidSfMD} & \callback{img64-\cidSfLD} & \callback{img64-\cidSfXLD}}%
\gdef\allconfigs{\largeconfigs & \smallconfigs}%
Number of GPUs                              & \gdef\callback##1{$32$}\allconfigs \\
Minibatch size                              & \gdef\callback##1{$2048$}\allconfigs \\
Duration (Mimg)                             & \gdef\callback##1{$\data{##1-mimg1}$}\allconfigs \\
Channel multiplier                          & $128$    & $192$    & $256$    & $320$    & $384$    & $448$    & $192$    & $256$    & $320$    & $384$    \\
Dropout probability                         & $0\%$    & $0\%$    & $10\%$   & $10\%$   & $10\%$   & $10\%$   & $0\%$    & $10\%$   & $10\%$   & $10\%$   \\
Learning rate max ($\alpha_\text{ref}$)     & $0.0120$ & $0.0100$ & $0.0090$ & $0.0080$ & $0.0070$ & $0.0065$ & $0.0100$ & $0.0090$ & $0.0080$ & $0.0070$ \\
Learning rate decay ($t_\text{ref}$)        & \gdef\callback##1{$70000$}\largeconfigs & \gdef\callback##1{$35000$}\smallconfigs \\
Noise distribution mean ($P_\text{mean}$)   & \gdef\callback##1{$-0.4$}\largeconfigs  & \gdef\callback##1{$-0.8$}\smallconfigs \\
Noise distribution std. ($P_\text{std}$)    & \gdef\callback##1{$1.0$}\largeconfigs   & \gdef\callback##1{$1.6$}\smallconfigs \\
\tabucline{-}
\multicolumn{11}{|l|}{\bf\tstrut Model size and training cost}\\
\tabucline{-}\tstrut%
Model capacity (Mparams)                    & \gdef\callback##1{$\data{##1-mparams1}$}\allconfigs \\
Model complexity (gigaflops)                & \gdef\callback##1{$\data{##1-gflops1}$}\allconfigs \\
Training cost (zettaflops)                  & \gdef\callback##1{$\eval[2]{\rawdata{##1-mimg1} * \rawdata{##1-gflops1} * 3 / 1e6}$}\allconfigs \\
Training speed (images$/$sec)               & \gdef\callback##1{$\eval[0]{1 / \rawdata{##1-spkimg3} * 1e3}$}\allconfigs \\
Training time (days)                        & \gdef\callback##1{$\eval[1]{\rawdata{##1-mimg1} * \rawdata{##1-spkimg3} / 3600 / 24 * 1e3}$}\allconfigs \\
Training energy (MWh)                       & \gdef\callback##1{$\eval[1]{\rawdata{##1-mimg1} * \rawdata{##1-spkimg3} / 3600 * 4 * \wattsPerNode * 1e-3}$}\allconfigs \\
\tabucline{-}
\multicolumn{11}{|l|}{\bf\tstrut Sampling without guidance, FID}\\
\tabucline{-}\tstrut%
FID                                         & \gdef\callback##1{$\data{##1-fid2}$}\allconfigs \\
EMA length ($\sigma_\text{rel}$)            & \gdef\callback##1{$\data{##1-phema-\ldata{TBD}{##1-mimg0}-phema3}$}\allconfigs \\
Sampling cost (teraflops)                   & \gdef\callback##1{$\eval[2]{(\rawdata{##1-nfe} * \rawdata{##1-gflops1} + \rawdata{img512-VAE-gflops1}) / 1e3}$}\largeconfigs & \gdef\callback##1{$\eval[2]{\rawdata{##1-nfe} * \rawdata{##1-gflops1} / 1e3}$}\smallconfigs \\
Sampling speed (images$/$sec$/$GPU)         & \gdef\callback##1{$\eval[1]{1e3 / (\rawdata{##1-nfe} * \rawdata{##1-mspeval3} + \rawdata{img512-VAE-mspeval3})}$}\largeconfigs & \gdef\callback##1{$\eval[1]{1e3 / (\rawdata{##1-nfe} * \rawdata{##1-mspeval3})}$}\smallconfigs \\
Sampling energy (mWh$/$image)               & \gdef\callback##1{$\eval[0]{(\rawdata{##1-nfe} * \rawdata{##1-mspeval3} + \rawdata{img512-VAE-mspeval3}) * \wattsPerNode / 8 / 3600}$}\largeconfigs & \gdef\callback##1{$\eval[0]{(\rawdata{##1-nfe} * \rawdata{##1-mspeval3}) * \wattsPerNode / 8 / 3600}$}\smallconfigs \\
\tabucline{-}
\multicolumn{11}{|l|}{\bf\tstrut Sampling with guidance, FID}\\
\tabucline{-}\tstrut%
FID                                         & \gdef\callback##1{$\data{##1-cfg2}$}\largeconfigs & \gdef\callback##1{--}\smallconfigs \\
EMA length ($\sigma_\text{rel}$)            & $0.045$ & $0.025$ & $0.030$ & $0.015$ & $0.020$ & $0.015$ & \gdef\callback##1{--}\smallconfigs\\
Guidance strength                           & $1.4$   & $1.4$   & $1.2$   & $1.2$   & $1.2$   & $1.2$   & \gdef\callback##1{--}\smallconfigs \\
Sampling cost (teraflops)                   & \gdef\callback##1{$\eval[2]{(\rawdata{##1-nfe} * (\rawdata{##1-gflops1} + \rawdata{img512-\cidXS-gflops1}) + \rawdata{img512-VAE-gflops1}) / 1e3}$}\largeconfigs & \gdef\callback##1{--}\smallconfigs \\
Sampling speed (images$/$sec$/$GPU)         & \gdef\callback##1{$\eval[1]{1e3 / (\rawdata{##1-nfe} * (\rawdata{##1-mspeval3} + \rawdata{img512-\cidXS-mspeval3}) + \rawdata{img512-VAE-mspeval3})}$}\largeconfigs & \gdef\callback##1{--}\smallconfigs \\
Sampling energy (mWh$/$image)               & \gdef\callback##1{$\eval[0]{(\rawdata{##1-nfe} * (\rawdata{##1-mspeval3} + \rawdata{img512-\cidXS-mspeval3}) + \rawdata{img512-VAE-mspeval3}) * \wattsPerNode / 8 / 3600}$}\largeconfigs & \gdef\callback##1{--}\smallconfigs \\
\tabucline{-}
\multicolumn{11}{|l|}{\bf\tstrut Sampling without guidance, FD\textsubscript{DINOv2}}\\
\tabucline{-}\tstrut%
FD\textsubscript{DINOv2}                    & $103.39$ & $68.64$ & $58.44$ & $52.25$ & $45.96$ & $42.84$ & \gdef\callback##1{--}\smallconfigs \\
EMA length ($\sigma_\text{rel}$)            & $0.200$  & $0.190$ & $0.155$ & $0.155$ & $0.155$ & $0.150$ & \gdef\callback##1{--}\smallconfigs\\
\tabucline{-}
\multicolumn{11}{|l|}{\bf\tstrut Sampling with guidance, FD\textsubscript{DINOv2}}\\
\tabucline{-}\tstrut%
FD\textsubscript{DINOv2}                    & $79.94$ & $52.32$ & $41.98$ & $38.20$ & $35.67$ & $33.09$ & \gdef\callback##1{--}\smallconfigs \\
EMA length ($\sigma_\text{rel}$)            & $0.150$ & $0.085$ & $0.015$ & $0.035$ & $0.030$ & $0.015$ & \gdef\callback##1{--}\smallconfigs\\
Guidance strength                           & $1.7$   & $1.9$   & $2.0$   & $1.7$   & $1.7$   & $1.7$   & \gdef\callback##1{--}\smallconfigs \\
\tabucline{-}
\end{tabu}%
\caption{\label{tabHyperparams}%
Details of all models discussed in \refpaper{sec:results}.
For ImageNet-512, \OURS-S is the same as \vconfig{g} in \autoref{figConfigG}.
}%
\end{table*}
}%

\newcommand{\figMagnitudesAll}{%
\begin{figure*}[t]%
\centering\footnotesize%
\gdef\figMagnitudeAxisWidth{0.44\columnwidth}
\gdef\figMagnitudeAxisHeight{0.22\columnwidth}
\gdef\figMagnitudesTicksAxisMin{0.07}
\gdef\figMagnitudesTicksAxisMax{2.0}
\gdef\figMagnitudesTicks{0.5,1.0,1.5}
\gdef\colorEncA{blue}\gdef\colorEncB{white}
\gdef\colorDecA{red}\gdef\colorDecB{white}
\gdef\colorEncOne{\colorEncA!100!\colorEncB}\gdef\colorEncTwo{\colorEncA!80!\colorEncB}\gdef\colorEncThree{\colorEncA!60!\colorEncB}\gdef\colorEncFour{\colorEncA!40!\colorEncB}
\gdef\colorDecOne{\colorDecA!100!\colorDecB}\gdef\colorDecTwo{\colorDecA!80!\colorDecB}\gdef\colorDecThree{\colorDecA!60!\colorDecB}\gdef\colorDecFour{\colorDecA!40!\colorDecB}
\gdef\insertMagnitudePlots[##1][##2]{%
\addplot[\colorEncOne] coordinates {\rawdata{magnitudes-##1-enc-64-##2}};
\addplot[\colorDecOne] coordinates {\rawdata{magnitudes-##1-dec-64-##2}};
\addplot[\colorEncTwo] coordinates {\rawdata{magnitudes-##1-enc-32-##2}};
\addplot[\colorDecTwo] coordinates {\rawdata{magnitudes-##1-dec-32-##2}};
\addplot[\colorEncThree] coordinates {\rawdata{magnitudes-##1-enc-16-##2}};
\addplot[\colorDecThree] coordinates {\rawdata{magnitudes-##1-dec-16-##2}};
\addplot[\colorEncFour] coordinates {\rawdata{magnitudes-##1-enc-8-##2}};
\addplot[\colorDecFour] coordinates {\rawdata{magnitudes-##1-dec-8-##2}};}
\begin{tabular}{@{}c@{\hs{0.1}}r@{\hs{0.6}}r@{\hs{0.6}}r@{\hs{0.6}}r@{}}%
\rotatebox{90}{\makebox[\figMagnitudeAxisHeight]{\vconfig{b}}}&
\begin{tikzpicture}[baseline]
\begin{axis}[
  title={Activations (max)}, title style={at={(0.5,1.15)}, anchor=north},
  xmin=\figMagnitudesTicksAxisMin,xmax=\figMagnitudesTicksAxisMax,ymin=-80, ymax=680,
  xtick={\figMagnitudesTicks}, xticklabels=\empty, ytick={0,200,400,600}, yticklabels={$0$,$200$,$400$,$600$},
  width=\figMagnitudeAxisWidth,height=\figMagnitudeAxisHeight,scale only axis,grid=major,
]
\insertMagnitudePlots[Config-B-1532-29-2147483-activation-norms][max]
\end{axis}
\end{tikzpicture}&
\begin{tikzpicture}[baseline]
\begin{axis}[
  title={Weights (max)}, title style={at={(0.5,1.15)}, anchor=north},
  xmin=\figMagnitudesTicksAxisMin,xmax=\figMagnitudesTicksAxisMax,ymin=-0.2, ymax=1.7,
  xtick={\figMagnitudesTicks}, xticklabels=\empty, ytick={0.0,0.5,1.0,1.5}, yticklabels={$0$,$0.5$,$1.0$,$1.5$},
  width=\figMagnitudeAxisWidth,height=\figMagnitudeAxisHeight,scale only axis,grid=major,
]
\insertMagnitudePlots[Config-B-1532-29-2147483-weight-norms][max]
\end{axis}
\end{tikzpicture}&
\begin{tikzpicture}[baseline]
  \begin{axis}[
    title={Activations (mean)}, title style={at={(0.5,1.15)}, anchor=north},
    xmin=\figMagnitudesTicksAxisMin,xmax=\figMagnitudesTicksAxisMax,ymin=-4, ymax=34,
    xtick={\figMagnitudesTicks}, xticklabels=\empty, ytick={0,10,20,30}, yticklabels={$0$,$10$,$20$,$30$},
    width=\figMagnitudeAxisWidth,height=\figMagnitudeAxisHeight,scale only axis,grid=major,
  ]
  \insertMagnitudePlots[Config-B-1532-29-2147483-activation-norms][mean]
  \end{axis}
  \end{tikzpicture}&
  \begin{tikzpicture}[baseline]
  \begin{axis}[
    title={Weights (mean)}, title style={at={(0.5,1.15)}, anchor=north},
    xmin=\figMagnitudesTicksAxisMin,xmax=\figMagnitudesTicksAxisMax,ymin=-0.04, ymax=0.34,
    xtick={\figMagnitudesTicks}, xticklabels=\empty, ytick={0.0,0.1,0.2,0.3}, yticklabels={$0$,$0.1$,$0.2$,$0.3$},
    width=\figMagnitudeAxisWidth,height=\figMagnitudeAxisHeight,scale only axis,grid=major,
  ]
  \insertMagnitudePlots[Config-B-1532-29-2147483-weight-norms][mean]
  \end{axis}
  \end{tikzpicture}\\
\rotatebox{90}{\makebox[\figMagnitudeAxisHeight]{\vconfig{c}}}&
\begin{tikzpicture}[baseline]
\begin{axis}[
  xmin=\figMagnitudesTicksAxisMin,xmax=\figMagnitudesTicksAxisMax,ymin=-80, ymax=680,
  xtick={\figMagnitudesTicks}, xticklabels=\empty, ytick={0,200,400,600}, yticklabels={$0$,$200$,$400$,$600$},
  width=\figMagnitudeAxisWidth,height=\figMagnitudeAxisHeight,scale only axis,grid=major,
]
\insertMagnitudePlots[Config-C-1533-41-2147483-activation-norms][max]
\end{axis}
\end{tikzpicture}&
\begin{tikzpicture}[baseline]
\begin{axis}[
  xmin=\figMagnitudesTicksAxisMin,xmax=\figMagnitudesTicksAxisMax,ymin=-0.2, ymax=1.7,
  xtick={\figMagnitudesTicks}, xticklabels=\empty, ytick={0.0,0.5,1.0,1.5}, yticklabels={$0$,$0.5$,$1.0$,$1.5$},
  width=\figMagnitudeAxisWidth,height=\figMagnitudeAxisHeight,scale only axis,grid=major,
]
\insertMagnitudePlots[Config-C-1533-41-2147483-weight-norms][max]
\end{axis}
\end{tikzpicture}&
\begin{tikzpicture}[baseline]
\begin{axis}[
  xmin=\figMagnitudesTicksAxisMin,xmax=\figMagnitudesTicksAxisMax,ymin=-4, ymax=34,
  xtick={\figMagnitudesTicks}, xticklabels=\empty, ytick={0,10,20,30}, yticklabels={$0$,$10$,$20$,$30$},
  width=\figMagnitudeAxisWidth,height=\figMagnitudeAxisHeight,scale only axis,grid=major,
]
\insertMagnitudePlots[Config-C-1533-41-2147483-activation-norms][mean]
\end{axis}
\end{tikzpicture}&
\begin{tikzpicture}[baseline]
\begin{axis}[
  xmin=\figMagnitudesTicksAxisMin,xmax=\figMagnitudesTicksAxisMax,ymin=-0.04, ymax=0.34,
  xtick={\figMagnitudesTicks}, xticklabels=\empty, ytick={0.0,0.1,0.2,0.3}, yticklabels={$0$,$0.1$,$0.2$,$0.3$},
  width=\figMagnitudeAxisWidth,height=\figMagnitudeAxisHeight,scale only axis,grid=major,
]
\insertMagnitudePlots[Config-C-1533-41-2147483-weight-norms][mean]
\end{axis}
\end{tikzpicture}\\
\rotatebox{90}{\makebox[\figMagnitudeAxisHeight]{\vconfig{d}}}&
\begin{tikzpicture}[baseline]
\begin{axis}[
  xmin=\figMagnitudesTicksAxisMin,xmax=\figMagnitudesTicksAxisMax,ymin=-4, ymax=34,
  xtick={\figMagnitudesTicks}, xticklabels=\empty, ytick={0,10,20,30}, yticklabels={$0$,$10$,$20$,$30$},
  width=\figMagnitudeAxisWidth,height=\figMagnitudeAxisHeight,scale only axis,grid=major,
]
\insertMagnitudePlots[Config-D-1533-45-2147483-activation-norms][max]
\end{axis}
\end{tikzpicture}&
\begin{tikzpicture}[baseline]
\begin{axis}[
  xmin=\figMagnitudesTicksAxisMin,xmax=\figMagnitudesTicksAxisMax,ymin=-4, ymax=34,
  xtick={\figMagnitudesTicks}, xticklabels=\empty, ytick={0,10,20,30}, yticklabels={$0$,$10$,$20$,$30$},
  width=\figMagnitudeAxisWidth,height=\figMagnitudeAxisHeight,scale only axis,grid=major,
]
\insertMagnitudePlots[Config-D-1533-45-2147483-weight-norms][max]
\end{axis}
\end{tikzpicture}&
\begin{tikzpicture}[baseline]
\begin{axis}[
  xmin=\figMagnitudesTicksAxisMin,xmax=\figMagnitudesTicksAxisMax,ymin=-0.2, ymax=1.7,
  xtick={\figMagnitudesTicks}, xticklabels=\empty, ytick={0,0.5,1,1.5}, yticklabels={$0$,$0.5$,$1.0$,$1.5$},
  width=\figMagnitudeAxisWidth,height=\figMagnitudeAxisHeight,scale only axis,grid=major,
]
\insertMagnitudePlots[Config-D-1533-45-2147483-activation-norms][mean]
\end{axis}
\end{tikzpicture}&
\begin{tikzpicture}[baseline]
\begin{axis}[
  xmin=\figMagnitudesTicksAxisMin,xmax=\figMagnitudesTicksAxisMax,ymin=-2, ymax=17,
  xtick={\figMagnitudesTicks}, xticklabels=\empty, ytick={0,5,10,15}, yticklabels={$0$,$5$,$10$,$15$},
  width=\figMagnitudeAxisWidth,height=\figMagnitudeAxisHeight,scale only axis,grid=major,
]
\insertMagnitudePlots[Config-D-1533-45-2147483-weight-norms][mean]
\end{axis}
\end{tikzpicture}\\
\rotatebox{90}{\makebox[\figMagnitudeAxisHeight]{\vconfig{e}}}&
\begin{tikzpicture}[baseline]
\begin{axis}[
  xmin=\figMagnitudesTicksAxisMin,xmax=\figMagnitudesTicksAxisMax,ymin=-4, ymax=34,
  xtick={\figMagnitudesTicks}, xticklabels=\empty, ytick={0,10,20,30}, yticklabels={$0$,$10$,$20$,$30$},
  width=\figMagnitudeAxisWidth,height=\figMagnitudeAxisHeight,scale only axis,grid=major,
]
\insertMagnitudePlots[Config-E-1533-46-2147483-activation-norms][max]
\end{axis}
\end{tikzpicture}&
\begin{tikzpicture}[baseline]
\begin{axis}[
  xmin=\figMagnitudesTicksAxisMin,xmax=\figMagnitudesTicksAxisMax,ymin=-0.2, ymax=1.7,
  xtick={\figMagnitudesTicks}, xticklabels=\empty, ytick={0.0,0.5,1.0,1.5}, yticklabels={$0$,$0.5$,$1.0$,$1.5$},
  width=\figMagnitudeAxisWidth,height=\figMagnitudeAxisHeight,scale only axis,grid=major,
]
\insertMagnitudePlots[Config-E-1533-46-2147483-weight-norms][max]
\end{axis}
\end{tikzpicture}&
\begin{tikzpicture}[baseline]
\begin{axis}[
  xmin=\figMagnitudesTicksAxisMin,xmax=\figMagnitudesTicksAxisMax,ymin=-0.2, ymax=1.7,
  xtick={\figMagnitudesTicks}, xticklabels=\empty, ytick={0,0.5,1,1.5}, yticklabels={$0$,$0.5$,$1.0$,$1.5$},
  width=\figMagnitudeAxisWidth,height=\figMagnitudeAxisHeight,scale only axis,grid=major,
]
\insertMagnitudePlots[Config-E-1533-46-2147483-activation-norms][mean]
\end{axis}
\end{tikzpicture}&
\begin{tikzpicture}[baseline]
\begin{axis}[
  xmin=\figMagnitudesTicksAxisMin,xmax=\figMagnitudesTicksAxisMax,ymin=-0.2, ymax=1.7,
  xtick={\figMagnitudesTicks}, xticklabels=\empty, ytick={0.0,0.5,1.0,1.5}, yticklabels={$0$,$0.5$,$1.0$,$1.5$},
  width=\figMagnitudeAxisWidth,height=\figMagnitudeAxisHeight,scale only axis,grid=major,
]
\insertMagnitudePlots[Config-E-1533-46-2147483-weight-norms][mean]
\end{axis}
\end{tikzpicture}\\
\rotatebox{90}{\makebox[\figMagnitudeAxisHeight]{\vconfig{f}}}&
\begin{tikzpicture}[baseline]
\begin{axis}[
  xmin=\figMagnitudesTicksAxisMin,xmax=\figMagnitudesTicksAxisMax,ymin=-4, ymax=34,
  xtick={\figMagnitudesTicks}, xticklabels=\empty, ytick={0,10,20,30}, yticklabels={$0$,$10$,$20$,$30$},
  width=\figMagnitudeAxisWidth,height=\figMagnitudeAxisHeight,scale only axis,grid=major,
]
\insertMagnitudePlots[Config-F-1533-72-2147483-activation-norms][max]
\end{axis}
\end{tikzpicture}&
\begin{tikzpicture}[baseline]
\begin{axis}[
  xmin=\figMagnitudesTicksAxisMin,xmax=\figMagnitudesTicksAxisMax,ymin=-0.2, ymax=1.7,
  xtick={\figMagnitudesTicks}, xticklabels=\empty, ytick={0.0,0.5,1.0,1.5}, yticklabels={$0$,$0.5$,$1.0$,$1.5$},
  width=\figMagnitudeAxisWidth,height=\figMagnitudeAxisHeight,scale only axis,grid=major,
]
\insertMagnitudePlots[Config-F-1533-72-2147483-weight-norms][max]
\end{axis}
\end{tikzpicture}&
\begin{tikzpicture}[baseline]
\begin{axis}[
  xmin=\figMagnitudesTicksAxisMin,xmax=\figMagnitudesTicksAxisMax,ymin=-0.2, ymax=1.7,
  xtick={\figMagnitudesTicks}, xticklabels=\empty, ytick={0,0.5,1,1.5}, yticklabels={$0$,$0.5$,$1.0$,$1.5$},
  width=\figMagnitudeAxisWidth,height=\figMagnitudeAxisHeight,scale only axis,grid=major,
]
\insertMagnitudePlots[Config-F-1533-72-2147483-activation-norms][mean]
\end{axis}
\end{tikzpicture}&
\begin{tikzpicture}[baseline]
\begin{axis}[
  xmin=\figMagnitudesTicksAxisMin,xmax=\figMagnitudesTicksAxisMax,ymin=-0.2, ymax=1.7,
  xtick={\figMagnitudesTicks}, xticklabels=\empty, ytick={0.0,0.5,1.0,1.5}, yticklabels={$0$,$0.5$,$1.0$,$1.5$},
  width=\figMagnitudeAxisWidth,height=\figMagnitudeAxisHeight,scale only axis,grid=major,
]
\insertMagnitudePlots[Config-F-1533-72-2147483-weight-norms][mean]
\end{axis}
\end{tikzpicture}\\
\rotatebox{90}{\makebox[\figMagnitudeAxisHeight]{\vconfig{g}}}&
\begin{tikzpicture}[baseline]
\begin{axis}[
  xmin=\figMagnitudesTicksAxisMin,xmax=\figMagnitudesTicksAxisMax,ymin=-4, ymax=34,
  xtick={0.5,1.0,1.5}, xticklabels={\makebox[0mm][r]{Gimg $=$ }$0.5$,$1.0$,$1.5$}, ytick={0,10,20,30}, yticklabels={$0$,$10$,$20$,$30$},
  width=\figMagnitudeAxisWidth,height=\figMagnitudeAxisHeight,scale only axis,grid=major,
]
\insertMagnitudePlots[Config-G-1533-52-2147483-activation-norms][max]
\end{axis}
\end{tikzpicture}&
\begin{tikzpicture}[baseline]
\begin{axis}[
  xmin=\figMagnitudesTicksAxisMin,xmax=\figMagnitudesTicksAxisMax,ymin=-0.2, ymax=1.7,
  xtick={\figMagnitudesTicks}, xticklabels={\makebox[0mm][r]{Gimg $=$ }$0.5$,$1.0$,$1.5$}, ytick={0.0,0.5,1.0,1.5}, yticklabels={$0$,$0.5$,$1.0$,$1.5$},
  width=\figMagnitudeAxisWidth,height=\figMagnitudeAxisHeight,scale only axis,grid=major,
]
\insertMagnitudePlots[Config-G-1533-52-2147483-weight-norms][max]
\end{axis}
\end{tikzpicture}&
\begin{tikzpicture}[baseline]
\begin{axis}[
  xmin=\figMagnitudesTicksAxisMin,xmax=\figMagnitudesTicksAxisMax,ymin=-0.2, ymax=1.7,
  xtick={0.5,1.0,1.5}, xticklabels={\makebox[0mm][r]{Gimg $=$ }$0.5$,$1.0$,$1.5$}, ytick={0,0.5,1,1.5}, yticklabels={$0$,$0.5$,$1.0$,$1.5$},
  width=\figMagnitudeAxisWidth,height=\figMagnitudeAxisHeight,scale only axis,grid=major,
]
\insertMagnitudePlots[Config-G-1533-52-2147483-activation-norms][mean]
\end{axis}
\end{tikzpicture}&
\begin{tikzpicture}[baseline]
\begin{axis}[
  xmin=\figMagnitudesTicksAxisMin,xmax=\figMagnitudesTicksAxisMax,ymin=-0.2, ymax=1.7,
  xtick={\figMagnitudesTicks}, xticklabels={\makebox[0mm][r]{Gimg $=$ }$0.5$,$1.0$,$1.5$}, ytick={0.0,0.5,1.0,1.5}, yticklabels={$0$,$0.5$,$1.0$,$1.5$},
  width=\figMagnitudeAxisWidth,height=\figMagnitudeAxisHeight,scale only axis,grid=major,
  legend pos={south east}, legend cell align={center}, legend columns={2}, legend cell align={left},
  legend style={column sep=.5mm,font=\scriptsize,inner xsep=1pt,inner ysep=1pt,nodes={inner sep=.5pt,text depth=0.15em,scale=0.75}}
]
\insertMagnitudePlots[Config-G-1533-52-2147483-weight-norms][mean]
\legend{Enc-64x64~~~,Dec-64x64,Enc-32x32,Dec-32x32,Enc-16x16,Dec-16x16,Enc-8x8,Dec-8x8}
\end{axis}
\end{tikzpicture}\\
\end{tabular}%
\caption{\label{figMagnitudesAll}%
Training-time evolution of the maximum and mean dimension-weighted
  $L_2$ norms of activations and weights over different depths of the
  the EMA-averaged score network.
  As discussed in \protect\refpaper{sec:archImprovements},
  our architectural modifications aim to standardize the activation magnitudes in \vconfig{D} and
  weight magnitudes in \vconfig{E}.
  Details of the computation are discussed in \protect\autoref{app:awmag}.
}%
\vspace{-1mm}%
\end{figure*}
}%

\definecolor{deepblue}{rgb}{0,0,0.5}
\definecolor{deepred}{rgb}{0.6,0,0}
\definecolor{deepgreen}{rgb}{0,0.5,0}
\DeclareFixedFont{\ttb}{T1}{txtt}{bx}{n}{6.6}
\DeclareFixedFont{\ttm}{T1}{txtt}{m}{n}{6.6}

\lstdefinestyle{styleForcedWN}{
  language=Python,
  basicstyle=\ttm,
  keywordstyle=\ttb\color{deepblue},
  stringstyle=\color{deepgreen},
  showstringspaces=false,
  morekeywords={self, torch, nn, functional, linalg, np, with},
  emph={normalize, forward, Conv2d, vector_norm, numel, add, __init__, randn, Module, Parameter, copy_, conv2d, no_grad, sqrt},
  emphstyle=\ttb\color{deepred},
}

\newsavebox{\codeForcedWN}
\begin{lrbox}{\codeForcedWN}%
\footnotesize%
\begin{lstlisting}[style=styleForcedWN]
def normalize(x, eps=1e-4):
    dim = list(range(1, x.ndim))
    n = torch.linalg.vector_norm(x, dim=dim, keepdim=True)
    alpha = np.sqrt(n.numel() / x.numel())
    return x / torch.add(eps, n, alpha=alpha)

class Conv2d(torch.nn.Module):
    def __init__(self, C_in, C_out, k):
        super().__init__()
        w = torch.randn(C_out, C_in, k, k)
        self.weight = torch.nn.Parameter(w)

    def forward(self, x):
        if self.training:
            with torch.no_grad():
                self.weight.copy_(normalize(self.weight))
        fan_in = self.weight[0].numel()
        w = normalize(self.weight) / np.sqrt(fan_in)
        x = torch.nn.functional.conv2d(x, w, padding='same')
        return x
\end{lstlisting}%
\end{lrbox}

\newcommand{\algForcedWN}{%
\begin{algorithm}[t]%
\usebox{\codeForcedWN}
\caption{\label{algForcedWN}%
\ \scalebox{0.92}{PyTorch code for forced weight normalization.}
}%
\end{algorithm}
}%

\lstdefinestyle{stylePhema}{
  language=Python,
  basicstyle=\ttm,
  keywordstyle=\ttb\color{deepblue},
  stringstyle=\color{deepgreen},
  showstringspaces=false,
  morekeywords={np, linalg},
  emph={p_dot_p, solve_weights, rv, cv, where, maximum, float64, reshape, solve},
  emphstyle=\ttb\color{deepred},
}

\newsavebox{\codePhema}
\begin{lrbox}{\codePhema}%
\footnotesize%
\begin{lstlisting}[style=stylePhema]
def p_dot_p(t_a, gamma_a, t_b, gamma_b):
    t_ratio = t_a / t_b
    t_exp = np.where(t_a < t_b, gamma_b, -gamma_a)
    t_max = np.maximum(t_a, t_b)
    num = (gamma_a + 1) * (gamma_b + 1) * t_ratio ** t_exp
    den = (gamma_a + gamma_b + 1) * t_max
    return num / den

def solve_weights(t_i, gamma_i, t_r, gamma_r):
    rv = lambda x: np.float64(x).reshape(-1, 1)
    cv = lambda x: np.float64(x).reshape(1, -1)
    A = p_dot_p(rv(t_i), rv(gamma_i), cv(t_i), cv(gamma_i))
    B = p_dot_p(rv(t_i), rv(gamma_i), cv(t_r), cv(gamma_r))
    X = np.linalg.solve(A, B)
    return X
\end{lstlisting}%
\end{lrbox}

\newcommand{\algPhema}{%
\begin{algorithm}[t]%
\usebox{\codePhema}
\caption{\label{algPhema}%
\ \scalebox{0.92}{NumPy code for solving post-hoc EMA weights.}
}%
\end{algorithm}
}%

\lstdefinestyle{styleGamma}{
  language=Python,
  basicstyle=\ttm,
  keywordstyle=\ttb\color{deepblue},
  stringstyle=\color{deepgreen},
  showstringspaces=false,
  morekeywords={np},
  emph={sigma_rel_to_gamma, roots},
  emphstyle=\ttb\color{deepred},
}

\newsavebox{\codeGamma}
\begin{lrbox}{\codeGamma}%
\footnotesize%
\begin{lstlisting}[style=styleGamma]
def sigma_rel_to_gamma(sigma_rel):
    t = sigma_rel ** -2
    gamma = np.roots([1, 7, 16 - t, 12 - t]).real.max()
    return gamma
\end{lstlisting}%
\end{lrbox}

\newcommand{\algGamma}{%
\begin{algorithm}[t]%
\usebox{\codeGamma}
\caption{\label{algGamma}%
\ \scalebox{0.92}{NumPy code for converting $\srel$ to $\gamma$.}
}%
\end{algorithm}
}%

\newcommand{\figTrainingTeaser}{%
\begin{figure}[t]%
\centering\footnotesize%
\begin{tikzpicture}%
\begin{axis}[
  width={1.12\linewidth}, height={70mm}, grid={major},
  xmin={0}, xmax={3.3}, xmode={linear}, xtick={0.0, 0.5, 1.0, 1.5, 2.0, 2.5, 3.0}, xticklabels={$0.0$, $0.5$, $1.0$, $1.5$, $2.0$, $2.5$, $3.0$},
  ymin={1.7}, ymax={10}, y coord trafo/.code=\pgfmathparse{ln(ln(##1))}, ytick={2, 3, 5, 10}, yticklabels={$2$, $3$, $5$, \raisebox{-1.5ex}[0ex][0ex]{FID}},
  xlabel={Training cost (zettaflops per model)}, x label style={at={(axis description cs:0.5,-0.06)}, anchor=north},
  legend pos={north west}, legend cell align={left},
]
\gdef\did{img512}
\gdef\pdotSize{1.5pt}

\gdef\pmy##1{\addplot[black, opacity=0.08, thin, forget plot] coordinates {(0,##1) (3.3,##1)};}
\pmy{1.8}\pmy{1.9}\pmy{2.1}\pmy{2.2}\pmy{2.3}\pmy{2.4}\pmy{2.5}\pmy{2.6}\pmy{2.7}\pmy{2.8}\pmy{2.9}\pmy{4}\pmy{6}\pmy{7}\pmy{8}\pmy{9}

\gdef\pcPrev##1{(\rawdata{\did-##1-mimg0} * \rawdata{\did-##1-gflops0} * 3 / 1e6, \rawdata{\did-##1-fid2})}
\gdef\pcPrevGuid##1{(\rawdata{\did-##1-mimg0} * \rawdata{\did-##1-gflops0} * 3 / 1e6, \rawdata{\did-##1-cfg2})}
\gdef\pcOurGuid##1{(\rawdata{\did-##1-mimg0} * \rawdata{\did-##1-gflops0} * 3 / 1e6, \rawdata{\did-##1-cfg2})}

\addplot[C2, forget plot] coordinates {\pcOurGuid{\cidXS}\pcOurGuid{\cidS}\pcOurGuid{\cidMD}\pcOurGuid{\cidXXLD}};

\gdef\pdot##1##2##3##4{\addplot[##1, mark=*, mark size=\pdotSize, forget plot, nodes near coords align={##2}, nodes near coords=\textlabel{##3}] coordinates {##4};}
\pdot{C0}{south}{ADM}{\pcPrevGuid{ADM}}
\pdot{C0}{south}{ADM-U}{\pcPrevGuid{ADM-U}}
\pdot{C0}{south}{DiT-XL/2}{\pcPrevGuid{DiT-XL}}
\pdot{C0}{south}{RIN}{\pcPrev{RIN}}
\pdot{C0}{south}{U-ViT, L}{\pcPrevGuid{U-ViT-L}}
\pdot{C0}{south}{VDM++}{\pcPrevGuid{VDMpp}}
\pdot{C2}{north}{\hs{-2}XS}{\pcOurGuid{\cidXS}}
\pdot{C2}{north}{\hs{-1}S}{\pcOurGuid{\cidS}}
\pdot{C2}{north}{\hs{-2}M}{\pcOurGuid{\cidMD}}
\pdot{C2}{north}{XXL}{\pcOurGuid{\cidXXLD}}

\addlegendimage{C0, only marks, mark size=\pdotSize}\addlegendentry{Previous work}
\addlegendimage{C2, mark=*, mark size=\pdotSize}\addlegendentry{EDM2}

\end{axis}
\end{tikzpicture}%
\end{figure}
}%

\newcommand{\figSamplingTeaser}{%
\begin{figure}[t]%
\centering\footnotesize%
\gdef\did{img512}%
\begin{tikzpicture}%
\begin{axis}[
  width={1.12\linewidth}, height={70mm}, grid={major},
  xmin={5}, xmax={2000}, xmode={log}, xtick={10, 20, 50, 100, 200, 500, 1000}, xticklabels={$10$, $20$, $50$, $100$, $200$, $500$, $1000$},
  ymin={1.7}, ymax={10}, y coord trafo/.code=\pgfmathparse{ln(ln(##1))}, ytick={2, 3, 5, 10}, yticklabels={$2$, $3$, $5$, \raisebox{-1.5ex}[0ex][0ex]{FID}},
  xlabel={Sampling cost (teraflops per image)}, x label style={at={(axis description cs:0.5,-0.06)}, anchor=north},
  legend pos={north west}, legend cell align={left},
]
\gdef\pdotSize{1.5pt}

\gdef\pmx##1{\addplot[black, opacity=0.08, thin, forget plot] coordinates {(##1,1.7) (##1,30)};}
\gdef\pmy##1{\addplot[black, opacity=0.08, thin, forget plot] coordinates {(5,##1) (2000,##1)};}
\pmx{6}\pmx{7}\pmx{8}\pmx{9}\pmx{20}\pmx{30}\pmx{40}\pmx{50}\pmx{60}\pmx{70}\pmx{80}\pmx{90}\pmx{200}\pmx{300}\pmx{400}\pmx{600}\pmx{700}\pmx{800}\pmx{900}
\pmy{1.8}\pmy{1.9}\pmy{2.1}\pmy{2.2}\pmy{2.3}\pmy{2.4}\pmy{2.5}\pmy{2.6}\pmy{2.7}\pmy{2.8}\pmy{2.9}\pmy{4}\pmy{6}\pmy{7}\pmy{8}\pmy{9}

\gdef\pcNoGuid##1{(\rawdata{\did-##1-nfe} * \rawdata{\did-##1-gflops0} / 1e3, \rawdata{\did-##1-fid2})}
\gdef\pcPrevGuid##1{(\rawdata{\did-##1-nfe} * \rawdata{\did-##1-gflops0} * 2 / 1e3, \rawdata{\did-##1-cfg2})}
\gdef\pcNoGuidVAE##1{((\rawdata{\did-##1-nfe} * \rawdata{\did-##1-gflops0} + \rawdata{\did-VAE-gflops1}) / 1e3, \rawdata{\did-##1-fid2})}
\gdef\pcPrevGuidVAE##1{((\rawdata{\did-##1-nfe} * \rawdata{\did-##1-gflops0} * 2 + \rawdata{\did-VAE-gflops1}) / 1e3, \rawdata{\did-##1-cfg2})}
\gdef\pcOurGuidVAE##1{((\rawdata{\did-##1-nfe} * (\rawdata{\did-##1-gflops0} + \rawdata{\did-\cidXS-gflops0}) + \rawdata{\did-VAE-gflops1}) / 1e3, \rawdata{\did-##1-cfg2})}

\addplot[C2, forget plot] coordinates {\pcOurGuidVAE{\cidXS}\pcOurGuidVAE{\cidS}\pcOurGuidVAE{\cidMD}\pcOurGuidVAE{\cidXXLD}};

\gdef\pdot##1##2##3##4{\addplot[##1, mark=*, mark size=\pdotSize, forget plot, nodes near coords align={##2}, nodes near coords=\textlabel{##3}] coordinates {##4};}
\pdot{C0}{south}{ADM}{\pcPrevGuid{ADM}}
\pdot{C0}{south}{\hs{-1}ADM-U}{\pcPrevGuid{ADM-U}}
\pdot{C0}{south}{DiT-XL/2}{\pcPrevGuidVAE{DiT-XL}}
\pdot{C0}{south}{RIN}{\pcNoGuid{RIN}}
\pdot{C0}{south}{VDM++}{\pcPrevGuid{VDMpp}}
\pdot{C2}{north}{\hs{-2}XS}{\pcOurGuidVAE{\cidXS}}
\pdot{C2}{north}{\hs{-1}S}{\pcOurGuidVAE{\cidS}}
\pdot{C2}{north}{\hs{-1}M}{\pcOurGuidVAE{\cidMD}}
\pdot{C2}{north}{\hs{2}XXL}{\pcOurGuidVAE{\cidXXLD}}

\addlegendimage{C0, only marks, mark size=\pdotSize}\addlegendentry{Previous work}
\addlegendimage{C2, mark=*, mark size=\pdotSize}\addlegendentry{EDM2}

\end{axis}
\end{tikzpicture}%
\end{figure}
}%

\newcommand{\plotExtraConvergence}{%
\centering\footnotesize%
\begin{tikzpicture}%
\begin{axis}[
  width={1.19\columnwidth}, height={60mm}, grid={major},
  xmin={0}, xmax={7.8}, xmode={linear}, xtick={0, 1, 2, 3, 4, 5, 6, 7}, xticklabels={$0$, $1$, $2$, $3$, $4$, $5$, $6$, $7$},
  ymin={1.71}, ymax={12.5}, ymode={log}, ytick={2, 3, 5, 10}, yticklabels={$2$, $3$, $5$, $10$},
  legend pos={south east}, legend cell align={left}, legend columns={7}, legend style={/tikz/every even column/.append style={column sep=0.43mm}},
]
\gdef\did{img512}

\gdef\pmy##1{\addplot[black, opacity=0.08, thin, forget plot] coordinates {(0,##1) (7.8,##1)};}
\pmy{4}\pmy{6}\pmy{7}\pmy{8}\pmy{9}

\gdef\pcFinal##1{(\rawdata{\did-##1-mimg0} * \rawdata{\did-##1-spkimg3} / 3600 / 24 * 1e3, \rawdata{\did-##1-fid2})}
\gdef\pcProg##1##2{(##2 * \rawdata{\did-##1-spkimg3} / 3600 / 24 * 1e3, \rawdata{\did-##1-phema-##2-fid2})}

\gdef\pcurve##1##2{\addplot[##1, forget plot] coordinates {\pcProg{##2}{134}\pcProg{##2}{268}\pcProg{##2}{537}\pcProg{##2}{805}\pcProg{##2}{1074}\pcProg{##2}{1342}\pcProg{##2}{1611}\pcProg{##2}{1879}\pcProg{##2}{2147}};}
\pcurve{C0}{\cidA}
\pcurve{C1}{\cidB}
\pcurve{C3}{\cidC}
\pcurve{C4}{\cidD}
\pcurve{C5}{\cidE}
\pcurve{C6}{\cidF}
\pcurve{C2}{\cidG}

\gdef\pdot##1##2##3##4{\addplot[##1, mark=*, forget plot, nodes near coords align={##2}, nodes near coords=\datalabel{##4}] coordinates {\pcFinal{##3}}; \addlegendimage{##1, mark=*}}
\pdot{C0}{south}{\cidA}{\hs{-3.6}$\data{\did-\cidA-fid2}$}
\pdot{C1}{west}{\cidB}{\raisebox{1.5mm}{$\data{\did-\cidB-fid2}$}}
\pdot{C3}{west}{\cidC}{\raisebox{-2.5mm}{$\data{\did-\cidC-fid2}$}}
\pdot{C4}{west}{\cidD}{$\data{\did-\cidD-fid2}$}
\pdot{C5}{west}{\cidE}{$\data{\did-\cidE-fid2}$}
\pdot{C6}{west}{\cidF}{\raisebox{1.5mm}{$\data{\did-\cidF-fid2}$}}
\pdot{C2}{west}{\cidG}{\raisebox{-1.5mm}{$\data{\did-\cidG-fid2}$}}

\legend{A, B, C, D, E, F, G}
\end{axis}
\end{tikzpicture}%
}%

\newcommand{\plotExtraIS}{%
\centering\footnotesize%
\begin{tikzpicture}%
\def\pct{\makebox[0mm][l]{\raisebox{0.15mm}{\scalebox{0.75}{\%}}}}
\begin{axis}[
  width={1.17\columnwidth}, height={60mm}, grid={major},
  xmin={0.02}, xmax={0.295}, xmode={linear}, xtick={0.05, 0.10, 0.15, 0.20, 0.25}, xticklabels={$5\pct$, $10\pct$, $15\pct$, $20\pct$, $25\pct$},
  ymin={75}, ymax={335}, ymode={linear}, ytick={100, 150, 200, 250, 300}, yticklabels={$100$, $150$, $200$, $250$, $300$},
]
\gdef\did{img512}\gdef\rid{phema-2147}

\gdef\pshade##1##2{\fillbetween[##1, opacity=0.2, forget plot]{coordinates {\rawdata{\did-##2-\rid-is-lo}}}{coordinates {\rawdata{\did-##2-\rid-is-hi}}};}
\pshade{C1}{\cidB}\pshade{C3}{\cidC}\pshade{C4}{\cidD}\pshade{C5}{\cidE}\pshade{C6}{\cidF}\pshade{C2}{\cidG}

\gdef\pcurve##1##2{\addplot[##1, forget plot] coordinates {\rawdata{\did-##2-\rid-is-hi}};}
\pcurve{C1}{\cidB}\pcurve{C3}{\cidC}\pcurve{C4}{\cidD}\pcurve{C5}{\cidE}\pcurve{C6}{\cidF}\pcurve{C2}{\cidG}

\gdef\pdot##1##2##3##4{\addplot[##1, mark=*, forget plot, nodes near coords align={##3}, nodes near coords=\datalabel{##4}] coordinates {(\rawdata{\did-##2-\rid-is-phema3}, \rawdata{\did-##2-\rid-is1})};}
\pdot{C0}{\cidA}{north}{\hs{2}$\data{\did-\cidA-\rid-is1}$}
\pdot{C1}{\cidB}{north}{$\data{\did-\cidB-\rid-is1}$}
\pdot{C3}{\cidC}{south}{\hs{1}$\data{\did-\cidC-\rid-is1}$}
\pdot{C4}{\cidD}{south}{\hs{-4}$\data{\did-\cidD-\rid-is1}$}
\pdot{C5}{\cidE}{south}{\hs{-1}$\data{\did-\cidE-\rid-is1}$}
\pdot{C6}{\cidF}{south}{$\data{\did-\cidF-\rid-is1}$}
\pdot{C2}{\cidG}{south}{$\data{\did-\cidG-\rid-is1}$}

\end{axis}
\end{tikzpicture}%
}%

\newcommand{\plotExtraRecall}{%
\centering\footnotesize%
\begin{tikzpicture}%
\def\pct{\makebox[0mm][l]{\raisebox{0.15mm}{\scalebox{0.75}{\%}}}}
\begin{axis}[
  width={1.15\columnwidth}, height={60mm}, grid={major},
  xmin={0.015}, xmax={0.265}, xmode={linear}, xtick={0.05, 0.10, 0.15, 0.20, 0.25}, xticklabels={$5\pct$, $10\pct$, $15\pct$, $20\pct$, $25\pct$},
  ymin={0.46}, ymax={0.64}, ymode={linear}, ytick={0.50, 0.55, 0.60}, yticklabels={$0.50$, $0.55$, $0.60$},
]
\gdef\did{img512}\gdef\rid{phema-2147}

\gdef\pshade##1##2{\fillbetween[##1, opacity=0.2, forget plot]{coordinates {\rawdata{\did-##2-\rid-recall-lo}}}{coordinates {\rawdata{\did-##2-\rid-recall-hi}}};}
\pshade{C1}{\cidB}\pshade{C3}{\cidC}\pshade{C4}{\cidD}\pshade{C5}{\cidE}\pshade{C6}{\cidF}\pshade{C2}{\cidG}

\gdef\pcurve##1##2{\addplot[##1, forget plot] coordinates {\rawdata{\did-##2-\rid-recall-hi}};}
\pcurve{C1}{\cidB}\pcurve{C3}{\cidC}\pcurve{C4}{\cidD}\pcurve{C5}{\cidE}\pcurve{C6}{\cidF}\pcurve{C2}{\cidG}

\gdef\pdot##1##2##3##4{\addplot[##1, mark=*, forget plot, nodes near coords align={##3}, nodes near coords=\datalabel{##4}] coordinates {(\rawdata{\did-##2-\rid-recall-phema3}, \rawdata{\did-##2-\rid-recall3})};}
\pdot{C0}{\cidA}{west}{\hs{1}$\data{\did-\cidA-\rid-recall3}$}
\pdot{C1}{\cidB}{north}{\raisebox{-4.5mm}{$\data{\did-\cidB-\rid-recall3}$}}
\pdot{C3}{\cidC}{north}{\raisebox{-5.7mm}{$\data{\did-\cidC-\rid-recall3}$}}
\pdot{C4}{\cidD}{south}{$\data{\did-\cidD-\rid-recall3}$}
\pdot{C5}{\cidE}{south}{$\data{\did-\cidE-\rid-recall3}$}
\pdot{C6}{\cidF}{north}{\raisebox{-3mm}{$\data{\did-\cidF-\rid-recall3}$}}
\pdot{C2}{\cidG}{south}{$\data{\did-\cidG-\rid-recall3}$}

\end{axis}
\end{tikzpicture}%
}%

\newcommand{\figExtraTriplet}{%
\begin{figure*}[t]%
\begin{subfigure}{0.325\linewidth}\plotExtraConvergence\vspace{-0.6mm}\caption{\small FID $\downarrow$ vs. training time (days)}\end{subfigure}\hfill%
\begin{subfigure}{0.325\linewidth}\plotExtraIS\vspace{-0.6mm}\caption{\small Inception score $\uparrow$ vs. EMA length}\end{subfigure}\hfill%
\begin{subfigure}{0.325\linewidth}\plotExtraRecall\vspace{-0.6mm}\caption{\small Recall $\uparrow$ vs. EMA length}\end{subfigure}%
\caption{\label{figExtraTriplet}%
\config{A--G} on ImageNet-512.
\textbf{(a)} Convergence in wall-clock time for equal-length training runs.
\textbf{(b,\,c)} Additional metrics.
}%
\end{figure*}%
}%

  \setcounter{figure}{6}
\setcounter{equation}{2}
\setcounter{table}{3}
\setcounter{algorithm}{0}
\setcounter{footnote}{1}
\setcounter{linenumber}{818}
\setcounter{topnumber}{1}
\setcounter{dbltopnumber}{1}
\allowdisplaybreaks

\section{Additional results}
\label{app:results}

\subsection{Generated images}

\figCurated

\autoref{figCurated} shows hand-selected images generated using our largest (XXL) ImageNet-512 model without
  classifier-free guidance.
Figures~\ref{figClassGridA}--\ref{figClassGridC} show uncurated images from the same model
  for various ImageNet classes, with guidance strength selected per class.

\subsection{Quality vs.~compute}

\figQualityMparamsScatter
\figQualitySamplingScatter
\figQualityTrainingScatter
\figExtraTriplet

\refpaper{figQualityComputeScatter} in the main paper quantifies the model's cost using gigaflops per evaluation, but this is just one possible option. We could equally well consider several alternative definitions for the model's cost.

\autoref{figQualityMparamsScatter} shows that the  efficiency improvement observed in \refpaper{figQualityComputeScatter} is retained when the model's cost is quantified using the number of trainable parameters instead.
\autoref{figQualitySamplingScatter} plots the same with respect to the sampling cost per image, demonstrating even greater improvements due to our low number of score function evaluations (NFE).
Finally, \autoref{figQualityTrainingScatter} plots the training cost of the model.
According to all of these metrics, our model reaches the same quality much quicker, and proceeds to improve the achievable result quality significantly.

\autoref{figExtraTriplet}a shows the convergence of our different configurations as a function of wall clock time;
  the early cleanup done in \config{B} improves both convergence and execution speed in addition to providing a cleaner starting point for experimentation.
During the project we tracked various quality metrics in addition to FID, including Inception score~\cite{IS}, KID~\cite{KID} and Recall~\cite{Tuomas2019}.
We standardized to FID because of its popularity and because the other metrics were largely consistent with it\,---\,see~\autoref{figExtraTriplet}b,c compared to~\refpaper{figEmaTriplet}a.

\figOverfitting

\autoref{figOverfitting} shows post-hoc EMA sweeps for a set of snapshots for our XXL-sized ImageNet-512 model with and without dropout. We observe that in this large model, overfitting starts to compromise the results without dropout, while a~10\% dropout allows steady convergence.
\autoref{figQualityTrainingScatter} further shows the convergence of different model sizes as a function of training cost with and without dropout. For the smaller models (XS, S) dropout is detrimental, but for the larger models it clearly helps, albeit at a cost of slightly slower initial convergence.

\subsection{Guidance vs.~unconditional model capacity}

\tabUncondSweep

\autoref{tabUncondSweep} shows quantitatively that using a large unconditional model is not useful in classifier-free guidance.
Using a very small unconditional model for guiding the conditional model reduces the computational cost of guided diffusion by almost~50\%.
The EMA lengths in the table apply to both conditional and unconditional model;
  it is typical that very short EMAs yield best results when sampling with guidance.

\subsection{Learning rate vs.~EMA length}

\figEmaLR

\autoref{figEmaLR} visualizes the interaction between EMA length and learning rate. While a sweet spot for the learning rate decay parameter still exists (\mbox{$t_\text{ref}=70\textrm{k}$} in this case), the possibility of sweeping over the EMA lengths post hoc drastically reduces the importance of this exact choice.
A wide bracket of learning rate decays \mbox{$t_\text{ref} \in [30\textrm{k},160\textrm{k}]$} yields FIDs within~10\% of the optimum using post-hoc EMA.

In contrast, if the EMA length was fixed at 13\%, varying $t_\text{ref}$ would increase FID much more, at worst by~72\% in the tested range.

\subsection{Fr\'echet distances using DINOv2}

\tabDinoFiveTwelve
\figEmaComparison

The DINOv2 feature space \cite{oquab2023dinov2} has been observed to align much better with human preferences compared to the widely used InceptionV3 feature space \cite{stein2023exposing}.
We provide a version of \refpaper{tabResultsFiveTwelve} using the Fr\'echet distance computed in the DINOv2 space (FD\textsubscript{DINOv2}) in \autoref{tabDinoFiveTwelve} to facilitate future comparisons.

We use the publicly available implementation\footnote{\url{https://github.com/layer6ai-labs/dgm-eval}} by \citet{stein2023exposing} for computing FD\textsubscript{DINOv2}.
We use 50,000 generated images and all 1,281,167 available real images, following the established best practices in FID computation.  
Class labels for the 50k generated samples are drawn from a uniform distribution.
We evaluate FD only once per 50k sample as we observe little random variation
between runs.

\autoref{figEmaComparison} compares FID and FD\textsubscript{DINOv2} as a function of EMA length. 
We can make three interesting observations. 
First, without guidance, the optima of the two \textsc{Config G} curves (green) are in a clearly different place, with FD\textsubscript{DINOv2} preferring longer EMA.
The disagreement between the two metrics is quite significant: FID considers FD\textsubscript{DINOv2}'s optimum (19\%) to be a poor choice, and vice versa.

Second, with guidance strength 1.4 (the optimal choice for FID according to \refpaper{figEmaGuidance}) the curves are astonishingly different. 
While both metrics agree that a modest amount of guidance is helpful, their preferred EMA lengths are totally different (2\% vs 14\%).
FID considers FD\textsubscript{DINOv2}'s optimum (14\%) to be a terrible choice and vice versa.
Based on a cursory assessment of the generated images, it seems that FD\textsubscript{DINOv2} prefers images with better global coherency, which often maps to higher perceptual quality, corroborating the conclusions of \citet{stein2023exposing}.
The significant differences in the optimal EMA length highlight the importance of searching the optimum specifically for the chosen quality metric.

Third, FD\textsubscript{DINOv2} prefers higher guidance strength than FID (1.9 vs 1.4). FID considers 1.9 clearly excessive.

The figure furthermore shows that our changes (\textsc{Config B} vs \textsc{G}) yield an improvement in FD\textsubscript{DINOv2} that is at least as significant as the drop we observed using FID.

\figMagnitudesAll

\subsection{Activation and weight magnitudes}
\label{app:awmag}

\autoref{figMagnitudesAll} shows an extended version of 
  \refpaper{figMagnitudes}, including
  activation and weight magnitude plots for
  \mbox{\config{B--G}} measured using both max and mean aggregation over
  each resolution bucket.
The details of the \mbox{computation are as follows.}

We first identify all trainable weight tensors within the
  \mbox{U-Net} encoder/decoder blocks of each resolution,
  including those in the associated self-attention layers.
This yields a set of tensors for each of the eight resolution buckets identified
  in the legend, i.e., \mbox{\{Enc, Dec\}$\times$\{8$\times$8, $\hdots$, 64$\times$64\}}.
The analyzed activations are the immediate outputs of the operations involving
  these tensors before any nonlinearity, and the analyzed weights are the
  tensors themselves.

In \config{b}, we do not include trainable biases in the weight analysis,
  but the activations are measured after applying the biases.
In \config{g}, we exclude the learned scalar gains from the weight analysis,
  but measure the activations after the gains have been applied.

\vparagraph{Activations.}
The activation magnitudes are computed as an expectation
  over~4096 training samples.
Ignoring the minibatch axis for clarity,
  most activations are shaped \mbox{$N\!\times\!H\!\times\!W$}
  where $N$ is the number of feature maps
  and $H$ and $W$ are the spatial dimensions.
For the purposes of analysis, we reshape these to 
  \mbox{$N\!\times\!M$} where \mbox{$M=H W$}.
The outputs of the linear transformation of the class embedding
  vector are considered to have shape \mbox{$N\!\times\!1$}.

Given a potentially reshaped activation tensor $\smash{\boldh \in \RR^{N\!\times\!M}}$,
  we compute the magnitudes of the individual features $\boldh_i$ as
\begin{equation}
  \MM[\boldh_i] ~=~ \sqrt{ \frac{1}{M} \sum_{j=1}^{M} \boldh_{i,j}^2 } \label{eq:magnitude}
  \text{.}
\end{equation}
The result contains the per-feature $L_2$ norms of the activations in tensor $\boldh$,
  scaled such that unit-normal distributed activations yield an expected magnitude
  of $1$ regardless of their dimensions.

All of these per-feature scalar magnitudes within a resolution bucket are 
  aggregated into a single number by taking either their maximum or their mean.
Taking the maximum magnitude
  (\refpaper{figMagnitudes} and \autoref{figMagnitudesAll}, left half)
  ensures that potential extreme behavior is not missed, whereas
  the mean magnitude
  (\autoref{figMagnitudesAll}, right half)
  is a closer indicator of average behavior.
Regardless of the choice, the qualitative behavior is similar.

\vparagraph{Weights.}

All weight tensors under analysis are of shape \mbox{$N\!\times\cdots$} where
  $N$ is the number of output features.
We thus reshape them all into \mbox{$N\!\times\!M$} 
  and compute the per-output-feature magnitudes using \autoref{eq:magnitude}.
Similar to activations, this ensures that unit-normal distributed weights have
  an expected magnitude of $1$ regardless of degree or dimensions of the weight tensor.
We again aggregate all magnitudes within a resolution
  bucket into a single number by taking either the maximum or the mean.
\refpaper{figMagnitudes} displays maximum magnitudes, whereas
  the extended version in \autoref{figMagnitudesAll} shows both maximum and mean magnitudes.

\section{Architecture details}
\label{app:architecture}

\figConfig{flabel=figConfigA, cletter=A, archpage=2, cid=\cidA, batch=4096, mixedp=partial, dropout=10, aref=0.0001, tref=\infty, pmean=-1.2, pstd=1.2, abeta=0.999, lscale=100, attres={32,16,8}, attblk=22, cdesc={EDM baseline}}
\figConfig{flabel=figConfigB, cletter=B, archpage=3, cid=\cidB, batch=2048, mixedp=partial, dropout=10, aref=0.0002, tref=\infty, pmean=-0.4, pstd=1.0, abeta=0.99,  lscale=100, attres={16,8},    attblk=15, cdesc={Minor improvements}}
\figConfig{flabel=figConfigC, cletter=C, archpage=4, cid=\cidC, batch=2048, mixedp=full,    dropout=10, aref=0.0002, tref=\infty, pmean=-0.4, pstd=1.0, abeta=0.99,  lscale=100, attres={16,8},    attblk=15, cdesc={Architectural streamlining}}
\figConfig{flabel=figConfigD, cletter=D, archpage=5, cid=\cidD, batch=2048, mixedp=full,    dropout=10, aref=0.0100, tref=\infty, pmean=-0.4, pstd=1.0, abeta=0.99,  lscale=1,   attres={16,8},    attblk=15, cdesc={Magnitude-preserving learned layers}}
\figConfig{flabel=figConfigE, cletter=E, archpage=6, cid=\cidE, batch=2048, mixedp=full,    dropout=10, aref=0.0100, tref=70000,  pmean=-0.4, pstd=1.0, abeta=0.99,  lscale=1,   attres={16,8},    attblk=15, cdesc={Control effective learning rate}}
\figConfig{flabel=figConfigF, cletter=F, archpage=7, cid=\cidF, batch=2048, mixedp=full,    dropout=10, aref=0.0100, tref=70000,  pmean=-0.4, pstd=1.0, abeta=0.99,  lscale=1,   attres={16,8},    attblk=15, cdesc={Remove group normalizations}}
\figConfig{flabel=figConfigG, cletter=G, archpage=8, cid=\cidG, batch=2048, mixedp=full,    dropout=0,  aref=0.0100, tref=70000,  pmean=-0.4, pstd=1.0, abeta=0.99,  lscale=1,   attres={16,8},    attblk=15, cdesc={Magnitude-preserving fixed-function layers}}

In this section, we present comprehensive details for the architectural changes introduced in \refpaper{sec:archImprovements}.
Figures~\mbox{\ref{figConfigA}--\ref{figConfigG}} illustrate the architecture diagram corresponding to each configuration, along with the associated hyperparameters.
In order to observe the individual changes, we invite the reader to flip through the figures in digital form.

\subsection{%
\texorpdfstring{EDM baseline (\bfconfig{a})}%
               {EDM baseline (Config A)}%
}
\label{app:configA}

Our baseline corresponds to the ADM~\cite{Dhariwal2021} network as implemented in the EDM~\cite{Karras2022elucidating} framework,
  operating in the latent space of a pre-trained variational autoencoder (VAE)~\cite{Rombach2021highresolution}.
We train the network for $2^{19}$ training iterations with batch size~4096, i.e.,~2147.5 million images,
  using the same hyperparameter choices that were previously used for \mbox{ImageNet-64} by \citet{Karras2022elucidating}.
In this configuration, we use traditional EMA with a half-life of~50M images, i.e.,~12k training iterations,
  which translates to $\srel \approx 0.034$ at the end of the training.
The architecture and hyperparameters as summarized in \autoref{figConfigA}.

\vparagraph{Preconditioning.}
Following the EDM framework, the network implements denoiser $\hat{\yy} = D_\theta(\xx; \sigma, \cc)$,
  where $\xx$ is a noisy input image, $\sigma$ is the corresponding noise level, $\cc$ is a one-hot class label, and $\hat{\yy}$ is the resulting denoised image;
  in the following, we will omit $\cc$ for conciseness.
The framework further breaks down the denoiser as
\begin{align}
\!\!\!  D_\theta(\xx; \sigma) &\,=\, \smash{\cskip(\sigma) \xx + \cout(\sigma) F_\theta\big( \cin(\sigma) \xx; \cnoise(\sigma) \big)} \label{eq:edmPrecond} \\[.5ex]
\!\!\!  \cskip(\sigma)        &\,=\, \smash{\sdata^2 \,/\, \big( \sigma^2 + \sdata^2 \big)} \label{eq:cskip} \\[.5ex]
\!\!\!  \cout(\sigma)         &\,=\, \smash{(\sigma \cdot \sdata) \,\big/\, \sqrt{\sigma^2 + \smash{\sdata^2}}} \label{eq:cout} \\[.5ex]
\!\!\!  \cin(\sigma)          &\,=\, \smash{1 \,\big/\, \sqrt{\sigma^2 + \smash{\sdata^2}}} \label{eq:cin} \\[.5ex]
\!\!\!  \cnoise(\sigma)       &\,=\, \smash{\tfrac{1}{4} \ln(\sigma)} \label{eq:cnoise}
  \text{,}
\end{align}
where the inputs and outputs of the raw network layers $F_\theta$ are preconditioned according to $\cin$, $\cout$, $\cskip$, and $\cnoise$.
$\sdata$ is the expected standard deviation of the training data.
The preconditioning is reflected in \autoref{figConfigA} by the blue boxes around the main inputs and outputs.

\vparagraph{Latent diffusion.}
For ImageNet-512, we follow \citet{Peebles2022}
  by preprocessing each 512$\times$512$\times$3 image in the dataset with a pre-trained off-the-shelf VAE encoder from Stable Diffusion%
  \footnote{\url{https://huggingface.co/stabilityai/sd-vae-ft-mse}}
  and postprocessing each generated image with the correspoding decoder.
For a given input image, the encoder produces a 4-channel latent at 8$\times$8 times lower resolution than the original, yielding a dimension of 64$\times$64$\times$4 for $\xx$ and $\hat{\yy}$.
The mapping between images and latents is not strictly bijective:
  The encoder turns a given image into a distribution of latents, where each channel $c$ of each pixel $(x, y)$ is drawn independently from $\NN(\mu_{x,y,c}, \sigma_{x,y,c}^2)$.
When preprocessing the dataset, we store the values of $\mu_{x,y,c}$ and $\sigma_{x,y,c}$ as 32-bit floating point, and draw a novel sample each time we encounter a given image during training.

The EDM formulation in \autoref{eq:edmPrecond} makes relatively strong assumptions about the mean and standard deviation of the training data.
We choose to normalize the training data globally to satisfy these assumptions\,---\,as opposed to, e.g., changing the value of $\sdata$, which might have far-reaching consequences in terms of the other hyperparameters.
We thus keep $\sdata$ at its default value $0.5$,
  subtract $[5.81, 3.25, 0.12, -2.15]$ from the latents during dataset preprocessing to make them zero mean,
  and multiply them by $0.5 ~/~ [4.17, 4.62, 3.71, 3.28]$ to make their standard deviation agree with $\sdata$.
When generating images, we undo this normalization before running the VAE decoder.

\vparagraph{Architecture walkthrough.}
The ADM~\cite{Dhariwal2021} network starts by feeding the noisy input image, multiplied by $\cnoise$, through an input block (``In'') to expand it to 192 channels.
It then processes the resulting activation tensor through a series of encoder and decoder blocks,
  organized as a U-Net structure~\cite{Ronneberger2015} and connected to each other via skip connections (faint curved arrows).
At the end, the activation tensor is contracted back to 4 channels by an output block (``Out''),
  and the final denoised image is obtained using $\cout$ and $\cskip$ as defined by \autoref{eq:edmPrecond}.
The encoder gradually decreases the resolution from 64$\times$64 to 32$\times$32, 16$\times$16, and 8$\times$8 by a set of downsampling blocks (``EncD''),
  and the channel count is simultaneously increased from 192 to 384, 576, and 768.
The decoder implements the same progression in reverse using corresponding upsampling blocks (``DecU'').

The operation of the encoder and decoder blocks is conditioned by a 768-dimensional embedding vector, obtained by feeding the noise level $\sigma$ and class label $\cc$ through a separate embedding network (``Embedding'').
The value of $\cnoise(\sigma)$ is fed through a sinusoidal timestep embedding layer%
  \footnote{%
  \href{https://github.com/openai/guided-diffusion/blob/22e0df8183507e13a7813f8d38d51b072ca1e67c/guided_diffusion/nn.py\#L103}{\texttt{https://github.com/openai/guided-diffusion/blob/22e0}}
  \href{https://github.com/openai/guided-diffusion/blob/22e0df8183507e13a7813f8d38d51b072ca1e67c/guided_diffusion/nn.py\#L103}{\hspace*{.1em}\texttt{df8183507e13a7813f8d38d51b072ca1e67c/guided\_diffusion/n}}
  \href{https://github.com/openai/guided-diffusion/blob/22e0df8183507e13a7813f8d38d51b072ca1e67c/guided_diffusion/nn.py\#L103}{\hspace*{.1em}\texttt{n.py\#L103}}%
  }%
  \textsuperscript{,}%
  \footnote{%
  \href{https://github.com/NVlabs/edm/blob/62072d2612c7da05165d6233d13d17d71f213fee/training/networks.py\#L193}{\texttt{https://github.com/NVlabs/edm/blob/62072d2612c7da051}}
  \href{https://github.com/NVlabs/edm/blob/62072d2612c7da05165d6233d13d17d71f213fee/training/networks.py\#L193}{\hspace*{.1em}\texttt{65d6233d13d17d71f213fee/training/networks.py\#L193}}%
  }%
  (``PosEmb'') to turn it into a 192-dimensional feature vector.
The result is then processed by two fully-connected layers with SiLU nonlinearity~\cite{Hendrycks2016}, defined as
\begin{equation}
  \silu(x) ~=~ \frac{x}{1 + e^{-x}} \label{eq:silu}
  \text{,}
\end{equation}
adding in a learned per-class embedding before the second nonlinearity.

The encoder and decoder blocks follow the standard pre-activation design of ResNets~\cite{He2016b}.
The main path (bold line) undergoes minimal processing:
It includes an optional 2$\times$2 upsampling or downsampling using box filter if the resolution changes,
  and an 1$\times$1 convolution if the number of channels changes.
The residual path employs two 3$\times$3 convolutions, preceded by group normalization and SiLU nonlinearity.
The group normalization computes empirical statistics for each group of 32 channels, normalizes them to zero mean and unit variance, and then applies learned per-group scaling and bias.
Between the convolutions, each channel is further scaled and biased based on the value of the embedding vector, processed by a per-block fully-connected layer.
The ADM architecture further employs dropout before the second convolution, setting individual elements of the activation tensor to zero with 10\% probability during training.
The U-Net skip connections originate from the outputs of the encoder blocks, and they are concatenated to the inputs of the corresponding decoder blocks.

Most of the encoder and decoder blocks operating at 32$\times$32 resolution and below (``EncA'' and ``DecA'') further employ self-attention after the residual branch.
The implementation follows the standard multi-head scaled dot product attention~\cite{Vaswani2017}, where each pixel of the incoming activation tensor is treated as a separate token.
For a single attention head, the operation is defined as
\begin{align}
  \boldA &~=~ \softmax(\boldW) \boldV \label{eq:attention} \\
  \boldW &~=~ \frac{1}{\sqrt{N_c}}\,\boldQ \boldK^\top \label{eq:attentionWeightMatrix}
  \text{,}
\end{align}
where $\boldQ = [\boldq_1, \ldots]^\top$, $\boldK = [\boldk_1, \ldots]^\top$, and $\boldV = [\boldv_1, \ldots]^\top$
  are matrices containing the query, key, and value vectors for each token, derived from the incoming activations using a 1$\times$1 convolution.
The dimensionality of the query and key vectors is denoted by $N_c$.

The elements of the weight matrix in \autoref{eq:attentionWeightMatrix} can equivalently be expressed as dot products between the individual query and key vectors:
\begin{equation}
  w_{i,j} ~=~ \frac{1}{\sqrt{N_c}}\,\big\langle \boldq_i, \boldk_j \big\rangle \label{eq:dotProductAttention}
  \text{.}
\end{equation}

\autoref{eq:attention} is repeated for each attention head, after which the resulting tokens $\boldA$ are concatenated, transformed by a 1$\times$1 convolution, and added back to the main path.
The number of heads $N_h$ is determined by the incoming channel count so that there is one head for each set of 64 channels.
The dot product and softmax operations are executed using 32-bit floating point to avoid overflows, even though the rest of the network uses 16-bit floating point.

The weights of almost every convolution and fully-connected layer are initialized using He's uniform init~\cite{He2015}, and the corresponding biases are also drawn from the same distribution.
There are two exceptions, however:
The per-class embedding vectors are initialized to $\NN(0, 1)$, and the weights and biases of the last convolution of the residual blocks,
  self-attention blocks, and the final output block are initialized to zero (dark green).
This has the effect that \mbox{$D_\theta(\xx, \sigma) = \cskip(\sigma)\,\xx$} after initialization.

\vparagraph{Training loss.}
Following EDM~\cite{Karras2022elucidating}, the denoising score matching loss for denoiser $D_\theta$ on noise level $\sigma$ is given by
\begin{equation}
  \LL(D_\theta; \sigma) ~=~ \EE_{\signal, \noise} \Big[ {\big\lVert D_\theta(\signal + \noise; \sigma) - \signal \big\rVert}^2_2 \Big] \label{eq:lossPerNoiseLevel}
  \text{,}
\end{equation}
where $\signal \sim \pdata$ is a clean image sampled from the training set and
$\noise \sim \NN\big( \boldzero, \sigma^2 \boldI \big)$ is i.i.d.~Gaussian noise.

The overall training loss is defined~\cite{Karras2022elucidating} as a weighted expectation of $\LL(D_\theta; \sigma)$ over the noise levels:
\begin{align}
  \LL(D_\theta) &~=~ \EE_{\sigma} \big[ \lambda(\sigma) \LL(D_\theta; \sigma) \big] \label{eq:edmLoss} \\
  \lambda(\sigma) &~=~ \big( \sigma^2 + \sdata^2 \big) \,/\, (\sigma \cdot \sdata)^2 \label{eq:edmLambda} \\
  \ln(\sigma) &~\sim~ \NN\big( P_\stext{mean}, P_\stext{std}^2 \big) \label{eq:edmLogNormal}
  \text{,}
\end{align}
where the distribution of noise levels is controlled by hyperparameters $P_\stext{mean}$ and $P_\stext{std}$.
The weighting function $\lambda(\sigma)$ in \autoref{eq:edmLambda} ensures that $\lambda(\sigma) \LL(D_\theta; \sigma) = 1$ at the beginning of the training,
effectively equalizing the contribution of each noise level with respect to $\nnabla_\theta \LL(D_\theta)$.

\subsection{%
\texorpdfstring{Minor improvements (\bfconfig{b})}%
               {Minor improvements (Config B)}%
}
\label{app:configB}

Since the baseline configuration (\config{a}) was not originally targeted for latent diffusion,
  we re-examined the hyperparameter choices to obtain an improved baseline (\config{b}).
Our new hyperparameters are summarized in \autoref{figConfigB}.

In order to speed up convergence, we found it beneficial
  to halve the batch size (2048 instead of 4096) while doubling the learning rate (\mbox{$\alpha_\text{ref} = 0.0002$} instead of $0.0001$),
  and to significantly reduce Adam's response time to changes in gradient magnitudes (\mbox{$\beta_2 = 0.99$} instead of $0.999$).
These changes had the largest impact towards the beginning of the training, where the network reconfigures itself for the task at hand, but they also helped somewhat towards the end.
Furthermore, we found the self-attention layers at 32$\times$32 resolution to be somewhat harmful; removing them improved the overall stability while also speeding up the training.
In \config{b}, we also switch from traditional EMA to our power function averaging profile (\refpaper{sec:powerEMA}),
  with two averages stored per snapshot for high-quality post-hoc reconstruction (\refpaper{sec:novelEMA}).

\vparagraph{Loss weighting.}
With the EDM training loss (\autoref{eq:edmLoss}),
  the quality of the resulting distribution tends to be quite sensitive to the choice of $P_\text{mean}$, $P_\text{std}$, and $\lambda(\sigma)$.
The role of $P_\text{mean}$ and $P_\text{std}$ is to focus the training effort on the most important noise levels,
  whereas $\lambda(\sigma)$ aims to ensure that the gradients originating from each noise level are roughly of the same magnitude.
Referring to Figure~5a of \citet{Karras2022elucidating}, the value of $\LL(D_\theta; \sigma)$ behaves somewhat unevenly over the course of training:
  It remains largely unchanged for the lowest and highest noise levels, but drops quickly for the ones in between.
\citet{Karras2022elucidating} suggest setting $P_\text{mean}$ and $P_\text{std}$
  so that the resulting log-normal distribution (\autoref{eq:edmLogNormal}) roughly matches the location of this in-between region.
When operating with VAE latents, we have observed that the in-between region has shifted considerably toward higher noise levels compared to RGB images.
We thus set \mbox{$P_\text{mean} = -0.4$} and \mbox{$P_\text{std} = 1.0$} instead of $-1.2$ and $1.2$, respectively, to roughly match its location.

While the choice of $\lambda(\sigma)$ defined by \autoref{eq:edmLambda} is enough to ensure that the gradient magnitudes are balanced at initialization,
  this is no longer true as the training progresses.
To compensate for the changes in $\LL(D_\theta; \sigma)$ that happen over time,
  no static choice of $\lambda(\sigma)$ is sufficient\,---\,the weighting function must be able to adapt its shape dynamically.
To achieve this, we treat the integration over noise levels in $\LL(D_\theta)$ as a form of multi-task learning.
In the following, we will first summarize the uncertainty-based weighting approach proposed by \citet{Kendall2018}, defined over a finite number of tasks,
  and then generalize it over a continuous set of tasks to replace \autoref{eq:edmLoss}.

\vparagraph{Uncertainty-based multi-task learning.}
In a traditional multi-task setting, the model is simultaneously being trained to perform multiple tasks corresponding to loss terms $\{\LL_1, \LL_2, \ldots\}$.
The naive way to define the overall loss is to take a weighted sum over these individual losses, i.e., $\LL = \sum_i w_i \LL_i$.
The outcome of the training, however, tends to be very sensitive to the choice of weights $w_i$.
This choice can become particularly challenging if the balance between the loss terms changes considerably over time.
\citet{Kendall2018} propose a principled approach for choosing the weights dynamically, based on the idea of treating the model outputs as probability distributions and maximizing the resulting likelihood.
For isotropic Gaussians, this boils down to associating each loss term $\LL_i$ with an additional trainable parameter \mbox{$\sigma_i > 0$}, i.e., homoscedastic uncertainty, and defining the overall loss as
\begin{align}
  \LL &~=~ \sum_i \bigg[ \frac{1}{2 \sigma_i^2} \LL_i + \ln \sigma_i \bigg] \\
  &~=~ \frac{1}{2} \sum_i \bigg[ \frac{\LL_i}{\sigma_i^2} + \ln \sigma_i^2 \bigg] \label{eq:kendallOrig}
  \text{.}
\end{align}
Intuitively, the contribution of $\LL_i$ is weighted down if the model is uncertain about task $i$, i.e., if $\sigma_i$ is high.
At the same time, the model is penalized for this uncertainty, encouraging $\sigma_i$ to be as low as possible.

In practice, it can be quite challenging for typical optimizers\,---\,such as Adam\,---\,to handle $\sigma_i$ directly due to the logarithm and the requirement that $\sigma_i > 0$.
A more convenient formula~\cite{Kendall2018} is obtained by rewriting \autoref{eq:kendallOrig} in terms of log variance $u_i = \ln \sigma_i^2$:
\begin{align}
  \LL &~=~ \frac{1}{2} \sum_i \bigg[ \frac{\LL_i}{e^{u_i}} + u_i \bigg] \\
  &~\propto~ \sum_i \bigg[ \frac{\LL_i}{e^{u_i}} + u_i \bigg] \label{eq:kendallSimple}
  \text{,}
\end{align}
where we have dropped the constant multiplier $1/2$, as it has no effect on the optimum.

\vparagraph{Continuous generalization.}
For the purpose of applying \autoref{eq:kendallSimple} to the EDM loss in \autoref{eq:edmLoss}, we consider each noise level $\sigma$ to represent a different task.
This means that instead of a discrete number of tasks, we are faced with an infinite continuum of tasks \mbox{$0 < \sigma < \infty$}.
In accordance to \autoref{eq:edmLoss}, we consider the loss corresponding to task $\sigma$ to be $\lambda(\sigma) \LL(D_\theta; \sigma)$,
  leading to the following overall loss:
\begin{equation}
  \LL(D_\theta, u) ~=~ \EE_{\sigma} \bigg[ \frac{\lambda(\sigma)}{e^{u(\sigma)}} \LL(D_\theta; \sigma) + u(\sigma) \bigg] \label{eq:kendallOurs}
  \text{,}
\end{equation}
where we employ a continuous uncertainty function $u(\sigma)$ instead of a discrete set of scalars $\{u_i\}$.

In practice, we implement $u(\sigma)$ as a simple one-layer MLP (not shown in \autoref{figConfigB})
  that is trained alongside the main denoiser network and discarded afterwards.
The MLP evaluates $\cnoise(\sigma)$ as defined by \autoref{eq:cnoise},
  computes Fourier features for the resulting scalar (see \autoref{app:configC}),
  and feeds the resulting feature vector through a fully-connected layer that outputs one scalar.
All practical details of the MLP, including initialization, magnitude-preserving scaling, and forced weight normalization,
  follow the choices made in our training configurations (Appendices~\ref{app:configB}--\ref{app:configG}).

\vparagraph{Intuitive interpretation.}
To gain further insight onto the meaning of \autoref{eq:kendallOurs}, let us solve for the minimum of $\LL(D_\theta, u)$ by setting its derivative to zero with respect to $u(\sigma)$:
\begin{align}
  0 &~=~ \frac{\diff \LL(D_\theta, u)}{\diff u(\sigma)} \\
  &~=~ \frac{\diff}{\diff u(\sigma)} \bigg[ \frac{\lambda(\sigma)}{e^{u(\sigma)}} \LL(D_\theta; \sigma) + u(\sigma) \bigg] \\
  &~=~ -\frac{\lambda(\sigma)}{e^{u(\sigma)}} \LL(D_\theta; \sigma) + 1
  \text{,}
\end{align}
which leads to
\begin{align}
  e^{u(\sigma)} &~=~ \lambda(\sigma) \LL(D_\theta; \sigma) \label{eq:kendallLogvar} \\[2mm]
  u(\sigma) &~=~ \ln \LL(D_\theta; \sigma) + \ln \lambda(\sigma)
  \text{.}
\end{align}
In other words, $u(\sigma)$ effectively keeps track of how $\LL(D_\theta; \sigma)$ evolves over time.
Plugging \autoref{eq:kendallLogvar} back into \autoref{eq:kendallOurs}, we arrive at an alternative interpretation of the overall training loss:
\begin{align}
  \LL(D_\theta, u) &~=~ \EE_{\sigma} \bigg( \frac{ \lambda(\sigma) \LL(D_\theta; \sigma) }{\big[ \lambda(\sigma) \LL(D_\theta; \sigma) \big]} + \big[ u(\sigma) \big] \bigg) \\
  &~=~ \EE_{\sigma} \frac{\LL(D_\theta; \sigma)}{\big[\LL(D_\theta; \sigma)\big]} + \big[ \EE_{\sigma} u(\sigma) \big] \label{eq:kendallEffective}
  \text{,}
\end{align}
where the bracketed expressions are treated as constants when computing $\nnabla_\theta \LL(D_\theta, u)$.
In other words, \autoref{eq:kendallOurs} effectively scales the gradients originating from noise level $\sigma$ by the reciprocal of $\LL(D_\theta; \sigma)$,
  equalizing their contribution between noise levels and over time.

Note that the optimum of Equations~\ref{eq:kendallOurs} and~\ref{eq:kendallEffective} with respect to $\theta$ does not depend on the choice of $\lambda(\sigma)$.
In theory, we could thus drop $\lambda(\sigma)$ altogether, i.e., set \mbox{$\lambda(\sigma) = 1$}.
We have tested this in practice and found virtually no impact on the resulting FID or convergence speed.
However, we choose to keep $\lambda(\sigma)$ defined according to \autoref{eq:edmLambda} as a practical safety precaution;
  \autoref{eq:kendallEffective} only becomes effective once $u(\sigma)$ has converged reasonably close to the optimum,
  so the choice of $\lambda(\sigma)$ is still relevant at the beginning of the training.

\subsection{%
\texorpdfstring{Architectural streamlining (\bfconfig{c})}%
               {Architectural streamlining (Config C)}%
}
\label{app:configC}

The network architecture of \config{b} contains several different types of trainable parameters that each behave in a slightly different way:
  weights and biases of three kinds (uniform-initialized, zero-initialized, and self-attention) as well as group normalization scaling parameters and class embeddings.
Our goal in \config{c} is eliminate these differences and make all the remaining parameters behave more or less identically.
To this end, we make several changes to the architecture that can be seen by comparing Figures~\ref{figConfigB} and~\ref{figConfigC}.

\vparagraph{Biases and group normalizations.}
We have found that we can simply remove all biases with no ill effects.
We do this for all convolutions, fully-connected layers, and group normalization layers in the denoiser network
  as well as in the loss weighting MLP (\autoref{eq:kendallOurs}).
In theory, this could potentially lead to reduced expressive power of the network,
  especially when sensing the overall scale of the input values.
Even though we have not seen this to be an issue in practice, we mitigate the danger
  by concatenating an additional channel of constant $1$ to the incoming noisy image in the input block (``In'').

Furthermore, we remove all other bias-like constructs for consistency;
  namely, the dynamic conditioning offset derived from the embedding vector in the encoder and decoder blocks
  and the subtraction of the empirical mean in group normalization.
We further simplify the group normalization layers by removing their learned scale parameter.
After these changes, the operation becomes
\begin{equation}
  b_{x,y,c,g} ~=~ \frac{a_{x,y,c,g}}{\sqrt{ \tfrac{1}{N_x N_y N_c} {\sum}_{i, j, k} a_{i,j,k,g}^2 } + \epsilon} \label{eq:simpleGroupNorm}
  \text{,}
\end{equation}
where $a_{x,y,c,g}$ and $b_{x,y,c,g}$ denote the incoming and outgoing activations, respectively, for pixel $(x, y)$, channel $c$, and group $g$,
and $N_x$, $N_y$, and $N_c$ indicate their corresponding dimensions.
We set $\epsilon = 10^{-4}$.

\vparagraph{Cosine attention.}
The 1$\times$1 convolutions responsible for producing the query and key vectors for self-attention behave somewhat differently compared to the other convolutions.
This is because the resulting values of $w_{i,j}$ (\autoref{eq:dotProductAttention}) scale quadratically with respect to the overall magnitude of the convolution weights, as opposed to linear scaling in other convolutions.
We eliminate this discrepancy by utilizing cosine attention~\cite{Gidaris2018,Luo2018,Nguyen2023}.
In practice, we do this by replacing the group normalization, executed right before the convolution, with pixelwise feature vector normalization~\cite{Karras2018} (``PixelNorm''), executed right after it.
This operation is defined as
\begin{equation}
  b_{x,y,c} ~=~ \frac{a_{x,y,c}}{\sqrt{ \tfrac{1}{N_c} {\sum}_i a_{x,y,i}^2 } + \epsilon} \label{eq:pixelNormA}
  \text{,}
\end{equation}
where we use $\epsilon = 10^{-4}$, similar to \autoref{eq:simpleGroupNorm}.

To gain further insight regarding the effect of this normalization,
  we note that, ignoring $\epsilon$, \autoref{eq:pixelNormA} can be equivalently written as
\begin{equation}
  \boldb_{x,y} ~=~ \sqrt{N_c}\,\frac{\bolda_{x,y}}{\norm{\bolda_{x,y}}_2} \label{eq:pixelNormB}
  \text{.}
\end{equation}

Let us denote the normalized query and key vectors by $\boldqhat_i$ and $\boldkhat_j$, respectively.
Substituting \autoref{eq:pixelNormB} into \autoref{eq:dotProductAttention} gives
\begin{align}
  w_{i,j} &~=~ \frac{1}{\sqrt{N_c}}\,\big\langle \boldqhat_i,\boldkhat_j \big\rangle \\
  &~=~ \frac{1}{\sqrt{N_c}}\,\bigg\langle \sqrt{N_c}\,\frac{\boldq_i}{\norm{\boldq_i}_2}\,,\,\sqrt{N_c}\,\frac{\boldk_j}{\norm{\boldk_j}_2} \bigg\rangle \\
  &~=~ \sqrt{N_c}\,\cos(\phi_{i,j})
  ~\text{,}
\end{align}
where $\phi_{i,j}$ denotes the angle between $\boldq_i$ and $\boldk_j$.
In other words, the attention weights are now determined exclusively by the \emph{directions} of the query and key vectors,
  and their lengths no longer have any effect.
This curbs the uncontrolled growth of $w_{i,j}$ during training and enables using 16-bit floating point throughout the entire self-attention block.

\vparagraph{Other changes.}
To unify the behavior of the remaining trainable parameters,
  we change the zero-initialized layers (dark green) and the class embeddings to use the same uniform initialization as the rest of the layers.
In order to retain the same overall magnitude after the class embedding layer, we scale the incoming one-hot class labels by $\sqrt{N}$
  so that the result is of unit variance, i.e., \mbox{$\tfrac{1}{N} \sum_i a_i^2 = 1$}.

Finally, we replace ADM's original timestep embedding layer with the more standard Fourier features~\cite{Tancik2020fourier}.
Concretely, we compute feature vector $\boldb$ based on the incoming scalar $a = \cnoise(\sigma)$ as
\begin{equation}
  \boldb ~=~ \left[ \begin{array}{c}
    \!\!\cos \big( 2\pi (f_1\, a + \varphi_1) \big) \\
    \!\!\cos \big( 2\pi (f_2\, a + \varphi_2) \big) \\
    \vdots \\
    \,\cos \big( 2\pi (f_N a + \varphi_N) \big)
  \end{array} \right] \label{eq:fourierFeatures}
  \text{,}
\end{equation}
\begin{equation}
  \text{where}~~
  f_i \sim \NN(0, 1)
  ~~\text{and}~~
  \varphi_i \sim\,\UU(0, 1) \label{eq:fourierPhases}
  \text{.}
\end{equation}

After initialization, we treat the frequencies $\{f_i\}$ and phases $\{\varphi_i\}$ as constants.

\subsection{%
\texorpdfstring{Magnitude-preserving learned layers\\\hspace*{-.45em}(\bfconfig{d})}%
               {Magnitude-preserving learned layers (Config D)}%
}
\label{app:configD}

In \config{d}, we modify all learned layers according to our magnitude-preserving design principle as shown in \autoref{figConfigD}.
Let us consider a fully-connected layer with input activations $\bolda = [a_j]^\top$ and output activations $\boldb = [b_i]^\top$.
The operation of the layer is
\begin{equation}
  \boldb ~=~ \boldW{}\bolda
  \,\text{,}
\end{equation}
where \mbox{$\boldW = [\boldwi]$} is a trainable weight matrix.
We can equivalently write this in terms of a single output element:
\begin{equation}
  b_i ~=~ \boldwi\,\bolda \label{eq:fullyConnectedLayer}
  \text{.}
\end{equation}

The same definition extends to convolutional layers by applying \autoref{eq:fullyConnectedLayer} independently to each output element.
In this case, the elements of $\bolda$ correspond to the activations of each input pixel within the support for the convolution kernel,
  i.e., \mbox{$\dim(\bolda) = N_j = N_c\,k^2$}, where $N_c$ is the number of input channels and $k$ is the size of the convolution kernel.

Our goal is to modify \autoref{eq:fullyConnectedLayer} so that it preserves the overall magnitude of the input activations, without looking at their actual contents.
Let us start by calculating the standard deviation of $b_i$, assuming that $\{a_i\}$ are mutually uncorrelated and of equal standard deviation $\sigma_a$:
\begin{align}
  \sigma_{b_i} &~=~ \sqrt{\Var[b_i]} \\[1mm]
  &~=~ \sqrt{\Var[ \boldwi\,\bolda ]} \\[1mm]
  &~=~ \sqrt{{\sum}_j w_{ij}^2 \Var[a_j]} \\
  &~=~ \sqrt{{\sum}_j w_{ij}^2 \, \sigma_a^2} \\
  &~=~ \norm{\boldwi}_2 \sigma_a \label{eq:layerOutputStd}
  \text{.}
\end{align}

To make \autoref{eq:fullyConnectedLayer} magnitude-preserving, we scale its output so that it has the same standard deviation as the input:
\begin{align}
  \hat{b}_i &~=~ \frac{\sigma_a}{\sigma_{b_i}} \,b_i \\
  &~=~ \frac{\sigma_a}{\norm{\boldwi}_2 \sigma_a} \,\boldwi \,\bolda \\
  &~=~ \underbrace{ \frac{\boldwi}{\norm{\boldwi}_2} }_{\eqqcolon\,\boldwhati}\,\bolda \label{eq:ourWeightNorm}
  \text{.}
\end{align}
In other words, we simply normalize each $\boldwi$ to unit length before use.
In practice, we introduce \mbox{$\epsilon = 10^{-4}$} to avoid numerical issues, similar to Equations~\ref{eq:simpleGroupNorm} and~\ref{eq:pixelNormA}:
\begin{equation}
  \boldwhati ~=~ \frac{\boldwi}{\norm{\boldwi}_2 + \epsilon}
  \text{.}
\end{equation}

Given that $\hat{b}_i$ is now agnostic to the scale of $\boldwi$, we initialize \mbox{$w_{i,j} \sim \NN(0, 1)$} so that the weights of all layers are roughly of the same magnitude.
This implies that in the early stages of training, when the weights remain close to their initial magnitude,
  the updates performed by Adam~\cite{Kingma2015} will also have roughly equal impact across the entire model, similar to the concept of \emph{equalized learning rate}~\cite{Karras2018}.
Since the weights are now larger in magnitude, we have to increase the learning rate as well.
We therefore set \mbox{$\alpha_\text{ref} = 0.0100$} instead of $0.0002$.

\vparagraph{Comparison to previous work.}
Our approach is closely related to \emph{weight normalization}~\cite{Salimans2016} and \emph{weight standardization}~\cite{Qiao2019}.
Reusing the notation from \autoref{eq:ourWeightNorm}, \citet{Salimans2016} define weight normalization as
\begin{equation}
  \boldwhati ~=~ \frac{g_i}{\norm{\boldwi}_2} \, \boldwi \label{eq:salimansWeightNorm}
  \text{,}
\end{equation}
where $g_i$ is a learned per-channel scaling factor that is initialized to one.
The original motivation of \autoref{eq:salimansWeightNorm} is to reparameterize the weight tensor in order to speed up convergence, without affecting its expressive power.
As such, the value of $g_i$ is free to drift over the course of training, potentially leading to imbalances in the overall activation magnitudes.
Our motivation, on the other hand, is to explicitly avoid such imbalances by removing any direct means for the optimization to change the magnitude of $\hat{b}_i$.

\citet{Qiao2019}, on the other hand, define weight standardization as
\begin{align}
  \boldwhati &~=~ \frac{\boldwi - \mu_i}{\raisebox{.1ex}{$\sigma_i$}}\text{~~, where} \label{eq:qiaoWeightStd} \\
  \mu_i &~=~ \frac{\raisebox{-.2ex}{$1$}}{\raisebox{-.1ex}{$N$}} {\sum}_j w_{i,j} \\
  \sigma_i &~=~ \sqrt{ \frac{\raisebox{-.2ex}{$1$}}{\raisebox{-.1ex}{$N$}} {\sum}_j w_{i,j}^2 - \mu_i^2 + \epsilon }
  \,\text{,}
\end{align}
intended to serve as a replacement for batch normalization in the context of micro-batch training.
In practice, we suspect that \autoref{eq:qiaoWeightStd} would probably work just as well as \autoref{eq:ourWeightNorm} for our purposes.
However, we prefer to keep the formula as simple as possible with no unnecessary moving parts.

\vparagraph{Effect on the gradients.}
One particularly useful property of \autoref{eq:ourWeightNorm} is that it projects the gradient of $\boldwi$ to be perpedicular to $\boldwi$ itself.
Let us derive a formula for the gradient of loss $\LL$ with respect to $\boldwi$:
\begin{align}
  \nnabla_{\boldwi} \LL &~=~ \nnabla_{\boldwi} \boldwhati \cdot \nnabla_{\boldwhati} \LL \\
  &~=~ \nnabla_{\boldwi}\!\bigg[ \frac{\boldwi}{\norm{\boldwi}_2} \bigg] \nnabla_{\boldwhati} \LL \label{eq:beforeQuotientRule}
  \text{.}
\end{align}
We will proceed using the quotient rule
\begin{equation}
  \bigg[\frac{f}{g}\bigg]' ~=~ \frac{f'g - fg'}{g^2}
  \text{,}
\end{equation}
where
\begin{alignat}{6}
  f &~=~ \boldwi, & ~~f' &~=~ \nnabla_{\boldwi} \boldwi &&~=~ \boldI \\
  g &~=~ \norm{\boldwi}_2, & ~~g' &~=~ \nnabla_{\boldwi} \norm{\boldwi}_2 &&~=~ \frac{\boldwi^\top}{\norm{\boldwi}_2}
  \text{.}
\end{alignat}
Plugging this back into \autoref{eq:beforeQuotientRule} gives us
\begin{align}
  \nnabla_{\boldwi} \LL &~=~ \bigg[ \frac{f'g - fg'}{g^2} \bigg] \nnabla_{\boldwhati} \LL \\
  &~=~ \left[ \frac{\boldI \norm{\boldwi}_2 - \boldwi \frac{\boldwi^\top}{\norm{\boldwi}_2}}{\norm{\boldwi}_2^2} \right] \!\nnabla_{\boldwhati} \LL \\
  &~=~ \frac{1}{\norm{\boldwi}_2} \Bigg[\,\boldI - \frac{\boldwi \boldwi^\top}{\norm{\boldwi}_2^2}\,\Bigg] \nnabla_{\boldwhati} \LL \label{eq:weightNormGrad}
  \text{.}
\end{align}

The bracketed expression in \autoref{eq:weightNormGrad} corresponds to a projection matrix that keeps the incoming vector otherwise unchanged,
  but forces it to be perpendicular to $\boldwi$, i.e., $\big\langle \boldwi, \nnabla_{\boldwi} \LL \big\rangle = 0$.
In other words, gradient descent optimization will not attempt to modify the length of $\boldwi$ directly.
However, the length of $\boldwi$ can still change due to discretization errors resulting from finite step size.

\subsection{%
\texorpdfstring{Controlling effective learning rate (\bfconfig{e})}%
               {Controlling effective learning rate (Config E)}%
}
\label{app:configE}

In \config{d}, we have effectively constrained all weight vectors of our model to lie on the unit hypersphere,
  i.e., $\norm{\boldwhati}_2 = 1$, as far as evaluating $D_\theta(\xx; \sigma)$ is concerned.
However, the magnitudes of the raw weight vectors, i.e., $\norm{\boldwi}_2$, are still relevant during training
  due to their effect on $\nnabla_{\boldwi} \LL$ (\autoref{eq:weightNormGrad}).
Even though we have initialized $\boldwi$ so that these magnitudes are initially balanced across the layers,
  there is nothing to prevent them from drifting away from this ideal over the course of training.
This is problematic since the relative impact of optimizer updates, i.e.,
  the \emph{effective learning rate}, can vary uncontrollably across the layers and over time.
In \config{e}, we eliminate this drift through \emph{forced weight normalization} as shown in \autoref{figConfigE},
  and gain explicit control over the effective learning rate.

\vparagraph{Growth of weight magnitudes.}
As noted by \citet{Salimans2016}, Equations~\ref{eq:ourWeightNorm} and~\ref{eq:weightNormGrad} have the side effect
  that they cause the norm of $\boldwi$ to increase monotonically after each training step.
As an example, let us consider standard gradient descent with learning rate $\alpha$.
The update rule is defined as
\begin{align}
  \boldwi' &~=~ \boldwi - \alpha \nnabla_{\boldwi} \LL \label{eq:sgd} \\
  \boldwi &\,\leftarrow~ \boldwi' \label{eq:optimizerUpdate}
  \text{.}
\end{align}

We can use the Pythagorean theorem to calculate the norm of the updated weight vector $\boldwi'$:
\begin{align}
  {\big\lVert \boldwi' \big\rVert}_2^2 &~=~ {\big\lVert \boldwi - \alpha \nnabla_{\boldwi} \LL \big\rVert}_2^2 \\[1mm]
  &~=~ {\big\lVert \boldwi \big\rVert}_2^2 + \alpha^2 {\big\lVert \nnabla_{\boldwi} \LL \big\rVert}_2^2 - 2 \alpha \underbrace{\big\langle \boldwi,\!\nnabla_{\boldwi} \LL \big\rangle}_{=\,0} \\[-2ex]
  &~=~ {\big\lVert \boldwi \big\rVert}_2^2 + \alpha^2 {\big\lVert \nnabla_{\boldwi} \LL \big\rVert}_2^2 \\[1mm]
  &~\ge~ {\big\lVert \boldwi \big\rVert}_2^2 \label{eq:weightGrowth}
  \text{.}
\end{align}

In other words, the norm of $\boldwi$ will necessarily increase at each step unless \mbox{$\nnabla_{\boldwi} \LL = \boldzero$}.
A similar phenomenon has been observed with optimizers like Adam~\cite{Kingma2015},
  whose updates do not maintain strict orthogonality,
  as well as in numerous scenarios that do not obey \autoref{eq:ourWeightNorm} exactly.
The effect is apparent in our \config{c} (\refpaper{figMagnitudes}) as well.

\vparagraph{Forced weight normalization.}
Given that the normalization and initialization discussed in \autoref{app:configD} are already geared towards constraining the weight vectors to a hypersphere,
  we take this idea to its logical conclusion and perform the entire optimization strictly under such a constraint.

Concretely, we require $\norm{\boldwi}_2 = \sqrt{N_j}$ to be true for each layer after each training step, where $N_j$ is the dimension of $\boldwi$, i.e., the fan-in.
\autoref{eq:weightNormGrad} already constrains $\nnabla_{\boldwi} \LL$ to lie on the tangent plane with respect to this constraint;
  the only missing piece is to guarantee that the constraint itself is satisfied by \autoref{eq:optimizerUpdate}.
To do this, we modify the formula to forcefully re-normalize $\boldwi'$ before assigning it back to $\boldwi$:
\begin{equation}
  \boldwi ~\leftarrow~ \sqrt{N_j} ~\frac{\boldwi'}{\norm{\boldwi'}_2} \label{eq:forcedWN}
  \text{.}
\end{equation}

Note that \autoref{eq:forcedWN} is agnostic to the exact definition of $\boldwi'$, so it is readily compatible with most of the commonly used optimizers.
In theory, it makes no difference whether the normalization is done before or after the actual training step.
In practice, however, the former leads to a very simple and concise PyTorch implementation, shown in \autoref{algForcedWN}.

\algForcedWN

\vparagraph{Learning rate decay.}
Let us step back and consider \config{d} again for a moment,
  focusing on the overall effect that $\norm{\boldwi}_2$ had on the training dynamics.
Networks where magnitudes of weights have no effect on activations
  have previously been studied by, e.g., \citet{Laarhoven2017}.
In these networks, the only meaningful progress is made in the \emph{angular} direction of weight vectors.
This has two consequences for training dynamics:
  First, the gradients seen by the optimizer are inversely proportional to the weight magnitude.
  Second, the loss changes slower at larger magnitudes, as more distance needs to be covered for the same angular change.
Effectively, both of these phenomena can be interpreted as downscaling the effective learning rate as a function of the weight magnitude.
Adam~\cite{Kingma2015} counteracts the first effect by approximately normalizing the gradient magnitudes,
  but it does not address the second.

From this perspective, we can consider \config{d} to have effectively employed an implicit learning rate decay:
  The larger the weights have grown (\refpaper{figMagnitudes}), the smaller the effective learning rate.
In general, learning rate decay is considered desirable in the sense that
  it enables the training to converge closer and closer to the optimum despite the stochastic nature of the gradients \cite{Salimans2016,Kingma2015}.
However, we argue that the implicit form of learning rate decay imposed by \autoref{eq:weightGrowth} is not ideal,
  because it can lead to uncontrollable and unequal drift between layers.

With forced weight normalization in \config{e} and onwards,
  the drift is eliminated and the effective learning rate is directly proportional to the specified learning rate $\alpha$.
Thus, in order to have the learning rate decay, we have to explicitly modify the value of $\alpha$ over time.
We choose to use the commonly advocated inverse square root decay schedule~\cite{Kingma2015}:
\begin{equation}
  \alpha(t) ~=~ \frac{\alpha_\text{ref}}{\sqrt{\max(t / t_\text{ref}, 1)}} \label{eq:lrDecay}
  \text{,}
\end{equation}
where the learning rate initially stays at $\alpha_\text{ref}$ and then starts decaying after $t_\text{ref}$ training iterations.
The constant learning rate schedule of \configs{a}{d} can be seen as a special case of \autoref{eq:lrDecay} with \mbox{$t_\text{ref} = \infty$}.

In the context of \refpaper{tabBridgeMain}, we use \mbox{$\alpha_\text{ref} = 0.0100$} and \mbox{$t_\text{ref} = 70000$}.
We have, however, found that the optimal choices depend heavily on the capacity of the network as well as the dataset (see \autoref{tabHyperparams}).

\vparagraph{Discussion.}
It is worth noticing that we normalize the weight vectors \emph{twice} during each training step:
  first to obtain $\boldwhati$ in \autoref{eq:ourWeightNorm} and then to constrain $\boldwi'$ in \autoref{eq:forcedWN}.
This is also reflected by the two calls to \texttt{normalize()} in \autoref{algForcedWN}.

\figForcedWN

The reason why \autoref{eq:ourWeightNorm} is still necessary despite \autoref{eq:forcedWN}
  is that it ensures that Adam's variance estimates are computed for the actual tangent plane steps.
In other words, \autoref{eq:ourWeightNorm} lets Adam ``know'' that it is supposed to operate under the fixed-magnitude constraint.
If we used \autoref{eq:forcedWN} alone, without \autoref{eq:ourWeightNorm},
  the variance estimates would be corrupted by the to-be erased normal component of the raw gradient vectors,
  leading to considerably smaller updates of an uncontrolled magnitude.
See \autoref{figForcedWN} for an illustration.

Furthermore, we intentionally force the raw weights $\boldwi$ to have the norm \smash{$\sqrt{N_j}$},
  while weight normalization further scales them to norm $1$.
The reason for this subtle but important difference is, again, compatibility with the Adam optimizer.
Adam approximately normalizes the gradient updates so that they are proportional to \smash{$\sqrt{N_j}$}.
We normalize the weights to the same scale, so that the \emph{relative} magnitude of the update becomes independent of $N_j$.
This eliminates an implicit dependence between the learning rate and the layer size.
Optimizers like LARS~\cite{You2017} and Fromage~\cite{Bernstein2020} build on a similar motivation,
  and explicitly scale the norm of the gradient updates to a fixed fraction of the weight norm.

Finally, \autoref{eq:ourWeightNorm} is also quite convenient due to its positive interaction with EMA.
Even though the raw values of $\boldwi$ are normalized at each training step by \autoref{eq:forcedWN}, their weighted averages are not.
To correctly account for our fixed-magnitude constraint, the averaging must also happen along the surface of the corresponding hypersphere.
However, we do not actually need to change the averaging itself in any way, because this is already taken care of by \autoref{eq:ourWeightNorm}:
Even if the magnitudes of the weight vectors change considerably as a result of averaging, they are automatically re-normalized upon use.

\vparagraph{Previous work.}
Several previous works have analyzed the consequences of weight magnitude growth under different settings and proposed various remedies.
Weight decay has often been identified as a solution for keeping the magnitudes in check, and its interplay with different normalization schemes and optimizers has been studied extensively~\cite{Laarhoven2017,Hoffer2018,Zhang2019,Li2020,Li2020b,Wan2021,Kunin2021,Roburin2022}. 
\citet{Cho2017} and \citet{Laarhoven2017} consider more direct approaches
  where the weights are directly constrained to remain in the unit norm hypersphere, eliminating the growth altogether.
\citet{Arpit2016} also normalize the weights directly, motivated by a desire to reduce the parameter space.
Various optimizers~\cite{You2017,You2020,Bernstein2020,Bernstein2020b,Liu2021} also aim for similar effects through weight-relative scaling of the gradient updates.

As highlighted by the above discussion, the success of these approaches can depend heavily on various small but important nuances
  that may not be immediately evident.
As such, we leave a detailed comparison of these approaches as future work.

\subsection{%
\texorpdfstring{Removing group normalizations (\bfconfig{f})}%
               {Removing group normalizations (Config F)}%
}
\label{app:configF}

In \config{f}, our goal is to remove the group normalization layers that may negatively impact the results
  due to the fact that they operate across the entire image.
We also make a few minor simplifications to the architecture.
These changes can be seen by comparing Figures~\ref{figConfigE} and~\ref{figConfigF}.

\vparagraph{Dangers of global normalization.}
As has been previously noted \cite{Karras2020cvpr,Karras2021alias},
  global normalization that operates across the entire image should be used cautiously.
It is firmly at odds with the desire for the model to behave consistently across geometric
  transformations~\cite{Karras2021alias,Vasconcelos2021} or when synthesizing objects in different contexts.
Such consistency is easiest to achieve if the internal representations of the image contents are capable of being as
  localized as they need to be,
  but global normalization entangles the representations of every part of the image by
  eliminating the first-order statistics across the image.
Notably, while attention \emph{allows} the representations to communicate with each other
  in a way that best fits the task,
  global normalization \emph{forces} communication to occur, with no way for individual features to avoid it.

This phenomenon has been linked to concrete image artifacts in the context of GANs.
\citet{Karras2020cvpr} found that the AdaIN operation used in StyleGAN was destroying vital information,
  namely the relative scales of different feature maps,
  which the model counteracted by creating strong localized spikes in the activations.
These spikes manifested as artifacts, and were successfully eliminated by removing global normalization operations.
In a different context, \citet{Brock2021b} show that normalization is not necessary for obtaining
  high-quality results in image classification.
We see no reason why it should be necessary or even beneficial in diffusion models, either.

\vparagraph{Our approach.}
Having removed the drift in activation magnitudes, we find that we can simply remove all group normalization layers with no obvious downsides.
In particular, doing this for the decoder improves the FID considerably,
  which we suspect to be related to the fact that the absolute scale of the individual output pixels is quite important for the training loss (\autoref{eq:lossPerNoiseLevel}).
The network has to start preparing the correct scales towards the end of the U-Net, and explicit normalization is likely to make this more challenging.

Even though explicit normalization is no longer strictly necessary, we have found that we can further improve the results slightly through pixelwise feature vector normalization (\autoref{eq:pixelNormA}).
Our hypothesis is that a small amount of normalization helps by counteracting correlations that would otherwise violate the independence assumption behind \autoref{eq:layerOutputStd}.
We find that the best results are obtained by normalizing the incoming activations at the beginning of each encoder block.
This guarantees that the magnitudes on the main path remain standardized despite the series of cumulative adjustments made by the residual and self-attention blocks.
Furthermore, this also appears to help in terms of standardizing the magnitudes of the decoder\,---\,presumably due to the presence of the U-Net skip connections.

\vparagraph{Architectural simplifications.}
In addition to reworking the normalizations, we make four minor simplifications to other parts of the architecture:

\begin{enumerate}
  \vspace{2mm}\item Unify the upsampling and downsampling operations of the encoder and decoder blocks by placing them onto the main path.
  \vspace{2mm}\item Slightly increase the expressive power of the encoder blocks by moving the 1$\times$1 convolution to the beginning of the main path.
  \vspace{2mm}\item Remove the SiLU activation in the final output block.
  \vspace{2mm}\item Remove the second fully-connected layer in the embedding network.
  \vspace{2mm}
\end{enumerate}

\noindent These changes are more or less neutral in terms of the FID, but we find it valuable to keep the network as simple as possible considering future work.

\subsection{%
\texorpdfstring{Magnitude-preserving fixed-function layers\\\hspace*{-.45em}(\bfconfig{g})}%
               {Magnitude-preserving fixed-function layers (Config G)}%
}
\label{app:configG}

In \config{g}, we complete the effort that we started in \config{d}
  by extending our magnitude-preserving design principle to cover the remaining fixed-function layers in addition to the learned ones.
The exact set of changes can be seen by comparing Figures~\ref{figConfigF} and~\ref{figConfigG}.

We will build upon the concept of \emph{expected magnitude}
  that we define by generalizing \autoref{eq:magnitude} for multivariate random variable $\bolda$:
\begin{equation}
  \MM[\bolda] ~=~ \sqrt{ \frac{1}{N_a} \sum_{i=1}^{N_a} \EE\big[a_i^2\big] } \label{eq:expectedMagnitude}
  \text{.}
\end{equation}
If the elements of $\bolda$ have zero mean and equal variance, we have \mbox{$\MM[\bolda]^2 = \Var[a_i]$}.
If $\bolda$ is non-random, \autoref{eq:expectedMagnitude} simplifies to \mbox{$\MM[\bolda] = \norm{\bolda}_2 / \sqrt{N_a}$}.
We say that $\bolda$ is \mbox{\emph{standardized}} iff \mbox{$\MM[\bolda] = 1$}.

Concretely, we aim to achieve two things:
  First, every input to the network should be standardized,
  and second, every operation in the network should be such that if its input is standardized, the output is standardized as well.
If these two requirements are met, it follows that all activations throughout the entire network are standardized.

Similar to \autoref{app:configD}, we wish to avoid having to look at the actual values of activations,
  which necessitates making certain simplifying statistical assumptions about them.
Even though these assumptions are not strictly true in practice,
  we find that the end result is surprisingly close to our ideal,
  as can be seen in the ``Activations (mean)'' plot for \config{g} in \autoref{figMagnitudesAll}.

\vparagraph{Fourier features.}
Considering the inputs to our network, the noisy image and the class label are already standardized
  by virtue of having been scaled by $\cin(\sigma)$ (\autoref{eq:cin}) and $\sqrt{N}$ (\autoref{app:configC}), respectively.
The Fourier features (\autoref{app:configC}), however, are not.
Let us compute the expected magnitude of $\boldb$ (\autoref{eq:fourierFeatures}) with respect to the frequencies and phases (\autoref{eq:fourierPhases}):
\begin{align}
  \MM[\boldb]^2 &~=~ \frac{1}{N_b} \sum_{i=1}^{N_b} \EE\bigg[ \Big( \!\cos\big( 2\pi (f_i a + \varphi_i) \big) \!\Big)^2 \bigg] \\
  &~=~ \EE\bigg[ \Big( \!\cos\big( 2\pi (f_1 a + \varphi_1) \big) \!\Big)^2 \bigg] \\
  &~=~ \EE\Big[ \big( \!\cos(2\pi \varphi_1) \big)^2 \Big] \\
  &~=~ \EE\Big[ \tfrac{1}{2} \big( 1 + \cos(4\pi \varphi_1) \big) \Big] \\
  &~=~ \tfrac{1}{2} + \tfrac{1}{2}\, \smash{\underbrace{\EE\big[\!\cos(4\pi \varphi_1)\big]}_{=\,0}} \\[2mm]
  &~=~ \tfrac{1}{2}
  \text{.}
\end{align}

\noindent To standardize the output, we thus scale \autoref{eq:fourierFeatures} by $1 / \MM[\boldb] = \sqrt{2}$:
\begin{equation}
  \mathrm{MP\text{-}Fourier}(a) ~=~ \left[ \begin{array}{c}
    \!\!\sqrt{2} \cos \big( 2\pi (f_1\, a + \varphi_1) \big) \\
    \!\!\sqrt{2} \cos \big( 2\pi (f_2\, a + \varphi_2) \big) \\
    \vdots \\
    \,\sqrt{2} \cos \big( 2\pi (f_N a + \varphi_N) \big)
  \end{array} \right]
  \text{.}
\end{equation}

\vparagraph{SiLU.}
Similar reasoning applies to the SiLU nonlinearity (\autoref{eq:silu}) as well, used throughout the network.
Assuming that $\bolda \sim \NN(\boldzero, \boldI)$:
\begin{align}
  \MM\big[ \!\silu(\bolda) \big]^2 &~=~ \frac{1}{N_a} \sum_{i=1}^{N_a} \EE\Big[ \big( \!\silu(a_i) \big)^2\Big] \\
  &~=~ \EE\Bigg[ \bigg( \frac{a_1}{1 + e^{-a_1}} \bigg)^2 \Bigg] \\
  &~=~ \int_{-\infty}^\infty \frac{\NN(x; 0, 1) \,x^2}{(1 + e^{-x})^2} ~\diff x \\
  &~\approx~ 0.3558 \label{eq:siluMagnitudeSqr} \\
  \MM\big[ \!\silu(\bolda) \big] &~\approx~ \sqrt{0.3558} ~\approx~ 0.596 \label{eq:siluMagnitude}
  \text{.}
\end{align}
Dividing the output accordingly, we obtain
\begin{equation}
  \mathrm{MP\text{-}SiLU}(\bolda) ~=~ \frac{\silu(\bolda)}{0.596} ~=~ \bigg[ \frac{a_i}{0.596 \cdot (1 + e^{-a_i})} \bigg] \label{eq:mpSiLU}
  \text{.}
\end{equation}

\vparagraph{Sum.}
Let us consider the weighted sum of two random vectors, i.e., $\boldc = w_a \bolda + w_b \boldb$.
We assume that the elements within each vector have equal expected magnitude and that $\EE[a_i b_i] = 0$ for every $i$.
Now,
\begin{align}
  & \MM[\boldc]^2 ~=~ \frac{1}{N_c} \sum_{i=1}^{N_c} \EE\big[ (w_a a_i + w_b b_i)^2 \big] \\
  &=~ \frac{1}{N_c} \sum_{i=1}^{N_c} \EE\big[ w_a^2 a_i^2 + w_b^2 b_i^2 + 2 w_a w_b a_i b_i \big] \\
  &=~ \frac{1}{N_c} \sum_{i=1}^{N_c} \Big[
    w_a^2 \!\!\underbrace{\EE\big[a_i^2\big]}_{\!=\,\MM[\bolda]^2}\!\!\! {+}
    w_b^2 \!\!\underbrace{\EE\big[b_i^2\big]}_{\!=\,\MM[\boldb]^2}\!\!\! {+}
    2 w_a w_b \underbrace{\EE\big[a_i b_i\big]}_{=\,0} \Big] \\[1mm]
  &=~ \frac{1}{N_c} \sum_{i=1}^{N_c} \big[ w_a^2 \MM[\bolda]^2 + w_b^2 \MM[\boldb]^2 \big] \\[1mm]
  &=~ w_a^2 \MM[\bolda]^2 + w_b^2 \MM[\boldb]^2 \label{eq:sumMagnitudeSqr}
  \text{.}
\end{align}

If the inputs are standardized, \autoref{eq:sumMagnitudeSqr} further simplifies to \mbox{$\MM[\boldc] = \sqrt{w_a^2 + w_b^2}$}.
A standardized version of $\boldc$ is then given by
\begin{equation}
  \boldchat ~=~ \frac{\boldc}{\MM[\boldc]} ~=~ \frac{w_a \bolda + w_b \boldb}{\sqrt{w_a^2 + w_b^2}} \label{eq:standardizedSum}
  \text{.}
\end{equation}

Note that \autoref{eq:standardizedSum} is agnostic to the scale of $w_a$ and $w_b$.
Thus, we can conveniently define them in terms of blend factor $t \in [0, 1]$ that can be adjusted on a case-by-case basis.
Setting $w_a = (1-t)$ and $w_b = t$, we arrive at our final definition:
\begin{equation}
  \mathrm{MP\text{-}Sum}(\bolda, \boldb, t) ~=~ \frac{(1-t) \,\bolda + t \,\boldb}{\sqrt{(1-t)^2 + t^2}} \label{eq:mpSum}
  \text{.}
\end{equation}

We have found that the best results are obtained by setting \mbox{$t = 0.3$} in the encoder, decoder, and self-attention blocks,
  so that the residual path contributes 30\% to the result while the main path contributes 70\%.
In the embedding network \mbox{$t = 0.5$} seems to work well,
  leading to equal contribution between the noise level and the class label.

\vparagraph{Concatenation.}
Next, let us consider the concatenation of two random vectors $\bolda$ and $\boldb$, scaled by constants $w_a$ and $w_b$, respectively.
The result is given by \mbox{$\boldc = w_a \bolda \concat w_b \boldb$}, which implies that
\begin{align}
  \MM[\boldc]^2 &~=~ \frac{ \sum_{i=1}^{N_c} \EE\big[c_i^2\big] }{N_c} \\
  &~=~ \frac{ \sum_{i=1}^{N_a} \EE\big[w_a^2 a_i^2\big] + \sum_{i=1}^{N_b} \EE\big[w_b^2 b_i^2\big] }{N_a + N_b} \\
  &~=~ \frac{ w_a^2 N_a \MM[\bolda]^2 + w_b^2 N_b \MM[\boldb]^2 }{N_a + N_b} \label{eq:concatMagnitudeSqr}
  \text{.}
\end{align}

Note that the contribution of $\bolda$ and $\boldb$ in \autoref{eq:concatMagnitudeSqr} is proportional to $N_a$ and $N_b$, respectively.
If $N_a \gg N_b$, for example, the result will be dominated by $\bolda$ while the contribution of $\boldb$ is largely ignored.
In our architecture (\autoref{figConfigG}), this situation can arise at the beginning of the decoder blocks when the U-Net skip connection is concatenated into the main path.
We argue that the balance between the two branches should be treated as an independent hyperparameter,
  as opposed to being tied to their respective channel counts.

We first consider the case where we require the two inputs to contribute equally, i.e.,
\begin{equation}
  w_a^2 N_a \MM[\bolda]^2 ~=~ w_b^2 N_b \MM[\boldb]^2 ~=~ C^2
  \text{,}
\end{equation}
where $C$ is an arbitrary constant.
Solving for $w_a$ and $w_b$:
\begin{align}
  w_a &~=~ \frac{C}{\MM[\bolda]} \cdot \frac{1}{\sqrt{N_a}} \\
  w_b &~=~ \frac{C}{\MM[\boldb]} \cdot \frac{1}{\sqrt{N_b}}
\end{align}
Next, we introduce blend factor $t \in [0,1]$ to allow adjusting the balance between $\bolda$ and $\boldb$ on a case-by-case basis, similar to \autoref{eq:mpSum}:
\begin{alignat}{4}
  \hat{w}_a &~=~ w_a \,(1-t) &&~=~ \frac{C}{\MM[\bolda]} \cdot \frac{1-t}{\sqrt{N_a}} \label{eq:concatWa} \\
  \hat{w}_b &~=~ w_b \,t &&~=~ \frac{C}{\MM[\boldb]} \cdot \frac{t}{\sqrt{N_b}} \label{eq:concatWb}
  \text{.}
\end{alignat}

If the inputs are standardized, i.e., $\MM[\bolda] = \MM[\boldb] = 1$,
  we can solve for the value of $C$ that leads to the output being standardized as well:
\begin{align}
  1 &~=~ \MM[\boldc]^2 \\[1mm]
  &~=~ \frac{ \hat{w}_a^2 N_a \MM[\bolda]^2 + \hat{w}_b^2 N_b \MM[\boldb]^2 }{N_a + N_b} \\[1mm]
  &~=~ \frac{ \hat{w}_a^2 N_a + \hat{w}_b^2 N_b }{N_a + N_b} \\[1mm]
  &~=~ \frac{ \Big[ C^2 \frac{(1-t)^2}{N_a} \Big] N_a + \Big[ C^2 \frac{t^2}{N_b} \Big] N_b }{N_a + N_b} \\[1mm]
  &~=~ C^2 \,\frac{ (1-t)^2 + t^2 }{N_a + N_b}
  \text{,}
\end{align}
which yields
\begin{equation}
  C ~=~ \sqrt{\frac{N_a + N_b}{(1-t)^2 + t^2}} \label{eq:concatC}
  \text{.}
\end{equation}

Combining \autoref{eq:concatC} with Equations~\ref{eq:concatWa} and~\ref{eq:concatWb}, we arrive at our final definition:
\begin{equation}
  \mathrm{MP\text{-}Cat}(\bolda, \boldb, t) = \sqrt{\frac{N_a + N_b}{(1-t)^2 + t^2}} \cdot \Bigg[ \frac{1-t}{\sqrt{N_a}} \,\bolda \concat \frac{t}{\sqrt{N_b}} \,\boldb \Bigg] \label{eq:mpCat}
  \text{.}
\end{equation}

In practice, we have found that the behavior of the model is quite sensitive to the choice of $t$ and that the best results are obtained using \mbox{$t = 0.5$}.
We hope that the flexibility offered by \autoref{eq:mpCat} may prove useful in the future,
  especially in terms of exploring alternative network architectures.

\vparagraph{Learned gain.}
While our goal of standardizing activations throughout the network is beneficial for the training dynamics,
  it can also be harmful in cases where it is \emph{necessary} to have \mbox{$\MM[\bolda] \ne 1$} in order to satisfy the training loss.

We identify two such instances in our network:
  the raw pixels ($F_\theta$) produced by the final output block (``Out''),
  and the learned per-channel scaling in the encoder and decoder blocks.
In order to allow $\MM[\bolda]$ to deviate from $1$, we introduce a simple learned scaling layer at these points:
\begin{equation}
  \mathrm{Gain}(\bolda) ~=~ g \, \bolda
  \text{,}
\end{equation}
where $g$ is a learned scalar that is initialized to $0$.
We have not found it necessary to introduce multiple scaling factors on a per-channel, per-noise-level, or per-class basis.
Note that \mbox{$g = 0$} implies \mbox{$F_\theta(\xx; \sigma) = \boldzero$},
  meaning that \mbox{$D_\theta(\xx; \sigma) = \xx$} at initialization,
  similar to \configs{a}{b} (see \autoref{app:configA}).

\section{Post-hoc EMA details}
\label{app:phema}

As discussed in \refpaper{sec:phema}, our goal is to be able to select the EMA length, or more
  generally, the model averaging profile, after a training run has completed.
This is achieved by storing a number of pre-averaged models during training, after which
  these pre-averaged models can be linearly combined to obtain a model whose averaging profile
  has the desired shape and length.

As a related contribution, we present the power function EMA profile that automatically
  scales according to training time and has zero contribution at \mbox{$t=0$}.

In this section, we first derive the formulae related to the traditional exponential EMA from first
  principles, after which we do the same for the power function EMA.
We then discuss how to determine the appropriate linear combination of pre-averaged models
  stored in training snapshots in order to match a given averaging profile, and specifically,
  to match the power function EMA with a given length.

\subsection{Definitions}

Let us denote the weights of the network as a function of training time by $\theta(t)$,
  so that $\theta(0)$ corresponds to the initial state and $\theta(\tcur)$ corresponds to the most recent state.
$\tcur$ indicates the current training time in arbitrary units, e.g., number of training iterations.
As always, the training itself is performed using $\theta(\tcur)$,
  but evaluation and other downstream tasks use a weighted average instead, denoted by $\hat{\theta}(\tcur)$.
This average is typically defined as a sum over the training iterations:
\begin{equation}
  \hat{\theta}(\tcur) ~=~ \sum_{t=0}^{\tcur} \ptc(t) \,\theta(t) \label{eq:emaDiscrete}
  \text{,}
\end{equation}
where $\ptc$ is a time-dependent \emph{response function} that sums to one, i.e., $\sum_t \ptc(t) = 1$.

Instead of operating with discretized time steps,
  we simplify the derivation by treating $\theta$, $\hat{\theta}$, and $\ptc$ as continuous functions 
  defined over $t \in \RR_{\ge0}$.
A convenient way to generalize \autoref{eq:emaDiscrete} to this case
  is to interpret $\ptc$ as a continuous probability distribution
  and define $\hat{\theta}(\tcur)$ as the expectation of $\theta(\tcur)$ with respect to that distribution:
\begin{equation}
  \hat{\theta}(\tcur) ~=~ \EE_{t \sim \ptc(t)} \!\big[ \theta(t) \big] \label{eq:emaContinuous}
  \text{.}
\end{equation}

Considering the definition of $\ptc(t)$,
  we can express a large class of practically relevant response functions
  in terms of a \emph{canonical response function} $f(t)$:
\begin{equation}
  \ptc(t) ~=~
  \begin{cases}
    f(t) \,/\, g(\tcur) & \text{if } 0 \le t \le \tcur \\
    0 & \text{otherwise}
  \end{cases} \label{eq:emaCanonical}
  \text{,}
\end{equation}
\begin{equation}
  \text{where}~~~
  g(\tcur) ~=~ \int_0^{\tcur}\!\!f(t)\,\diff t
  \text{.}
\end{equation}

To characterize the properties, e.g., length, of a given response function,
  we consider its standard distribution statistics:
\begin{equation}
  \mu_{\tcur} = \EE[t]
  ~~~\text{and}~~~
  \sigma_{\tcur} = \sqrt{\Var[t]}
  ~~~\text{for}~~~
  t \sim \ptc(t)
  \text{.}
\end{equation}
These two quantities have intuitive interpretations:
  $\mu_{\tcur}$ indicates the average delay imposed by the response function,
  while $\sigma_{\tcur}$ correlates with the length of the time period that is averaged over.

\subsection{Traditional EMA profile}

The standard choice for the response function is the exponential moving average (EMA)
  where $\ptc$ decays exponentially as $t$ moves farther away from $\tcur$ into the past,
  often parameterized by EMA half-life $\lambda$.
In the context of \autoref{eq:emaCanonical}, we can express such exponential decay as
\mbox{$\ptc(t) = f(t) / g(\tcur)$}, where
\begin{align}
  f(t) &~=~
  \begin{cases}
    2^{t / \lambda} & \text{if } t > 0 \\
    \frac{\lambda}{\ln 2}\,\delta(t) & \text{otherwise}
  \end{cases} \label{eq:emaExponential} \\[.5ex]
  g(\tcur) &~=~ \frac{\lambda\,2^{\tcur / \lambda}}{\ln 2} \label{eq:emaExponentialG}
  \text{,}
\end{align}
and $\delta(t)$ is the Dirac delta function.

The second row of \autoref{eq:emaExponential} highlights an inconvenient aspect about the traditional EMA.
The exponential response function is infinite in the sense that it expects to be able to consult
  historical values of $\theta$ infinitely far in the past,
  even though the training starts at \mbox{$t=0$}.
Consistent with previous work, we thus deposit the probability mass
  that would otherwise appear at \mbox{$t<0$} onto \mbox{$t=0$} instead,
  corresponding to the standard practice of initializing the accumulated EMA weights
  to network's initial weights.

This implies that unless \mbox{$\lambda \ll \tcur$}, the averaged weights
  $\hat{\theta}(\tcur)$ end up receiving a considerable contribution from the initial
  state $\theta(0)$ that is, by definition, not
  meaningful for the task that the model is being trained for.

\subsection{Tracking the averaged weights during training}

In practice, the value of $\hat{\theta}(\tcur)$ is computed during training as follows.
Suppose that we are currently at time $\tcur$ and know the current $\hat{\theta}(\tcur)$.
We then run one training iteration to arrive at $\tnext = \tcur + \Delta t$
  so that the updated weights are given by $\theta(\tnext)$.
Here $\Delta t$ denotes the length of the training step in whatever units are being used for $t$.

To define $\theta(t)$ for all values of $t$, we consider it to be a piecewise constant function
  so that \mbox{$\theta(t) = \theta(\tnext)$} for every \mbox{$\tcur < t \le \tnext$}.
Let us now write the formula for \smash{$\hat{\theta}(\tnext)$} in terms of
  Equations~\ref{eq:emaContinuous} and~\ref{eq:emaCanonical}:
\begin{align}
  \!\!\hat{\theta}(\tnext) &= \EE_{t \sim \ptn(t)}\!\big[ \theta(t) \big] \\
  &= \int_{-\infty}^\infty \ptn(t) \,\theta(t) \, \diff t \\
  &= \int_0^{\tnext} \frac{f(t)}{g(\tnext)} \theta(t) \,\diff t \\
  &= \int_0^{\tcur}\hs{-1} \frac{f(t)}{g(\tnext)} \theta(t) \,\diff t + \int_{\tcur}^{\tnext}\hs{-1} \frac{f(t)}{g(\tnext)} \theta(t) \,\diff t \\[.5ex]
  &= \underbrace{\frac{g(\tcur)}{g(\tnext)}}_{\eqqcolon\,\beta(\tnext)}
     \underbrace{\int_0^{\tcur}\hs{-2} \frac{f(t)}{g(\tcur)} \theta(t) \diff t}_{=\,\hat{\theta}(\tcur)} +
     \frac{\theta(\tnext)}{g(\tnext)}\hs{-1}
     \underbrace{\int_{\tcur}^{\tnext}\hs{-2} f(t) \diff t}_{\!\!= \,g(\tnext) - g(\tcur)\!\!} \\[1mm]
  &= \beta(\tnext) \,\hat{\theta}(\tcur) + \frac{\theta(\tnext)}{g(\tnext)} \big( g(\tnext) - g(\tcur) \big) \\[1mm]
  &= \beta(\tnext) \,\hat{\theta}(\tcur) + \bigg[ 1 - \underbrace{\frac{g(\tcur)}{g(\tnext)}}_{\!\!=\,\beta(\tnext)\!\!} \bigg] \theta(\tnext) \\[1mm]
  &= \beta(\tnext) \,\hat{\theta}(\tcur) + \big( 1 - \beta(\tnext) \big) \,\theta(\tnext) \label{eq:emaUpdate}
  \text{.}
\end{align}
Thus, after each training iteration, we must linearly
  interpolate $\hat{\theta}$ toward $\theta$ by $\beta(\tnext)$.
In the case of exponential EMA, $\beta(\tnext)$ is constant and, consulting \autoref{eq:emaExponentialG}, given by
\begin{equation}
  \beta(\tnext) ~=~ \frac{g(\tcur)}{g(\tnext)}
  ~=~ \frac{2^{\tcur / \lambda}}{2^{\tnext / \lambda}}
  ~=~ 2^{-\Delta t / \lambda}
  \text{.}
\end{equation}

\subsection{Power function EMA profile}

In \refpaper{sec:archImprovements}, we make two observations that highlight the problematic
  aspects of the exponential EMA profile.
First, it is generally beneficial to employ unconventionally long averages,
  to the point where \mbox{$\lambda \ll \tcur$} is no longer true.
Second, the length of the response function should increase over the course of training proportional to $\tcur$.
As such, the definition of $f(t)$ in \autoref{eq:emaExponential} is not optimal for our purposes.

\figEmaLength

The most natural requirement for $f(t)$ is that it should be self-similar over different timescales, i.e., $f(c\,t) \propto f(t)$ for any positive stretching factor $c$.
This implies that the response functions for different values of $\tcur$ will also be stretched versions of each other; if $\tcur$ doubles, so does $\sigma_{\tcur}$.
Furthermore, we also require that $f(0)=0$ to avoid meaningless contribution from $\theta(0)$.
These requirements are uniquely satisfied, up to constant scaling, by the family of power functions
\mbox{$\ptc(t) = f(t) / g(\tcur)$}, where
\begin{equation}
  f(t) ~=~ t^\gamma
  ~~~~\text{and}~~~~
  g(\tcur) ~=~ \frac{\tcur^{\gamma+1}}{\gamma + 1}
  \text{.}
  \label{eq:powerEma}
\end{equation}
The constant $\gamma > 0$ controls the overall amount of averaging as illustrated in \autoref{figEmaLength}.

Considering the distribution statistics of our response function, we notice that $\ptc$ is equal to the
  beta distribution with \mbox{$\alpha=\gamma+1$} and \mbox{$\beta=1$}, stretched along the $t$-axis by $\tcur$.
The relative mean and standard deviation with respect to $\tcur$ are thus given by
\begin{alignat}{4}
  \mu_\text{rel} &~=~ \frac{\mu_{\tcur}}{\tcur} &&~=~ \frac{\gamma + 1}{\gamma + 2} \\
  \srel &~=~ \frac{\sigma_{\tcur}}{\tcur} &&~=~ \sqrt{\frac{\gamma + 1}{(\gamma + 2)^2 \,(\gamma + 3)}} \label{eq:powerEmaStd}
  \text{.}
\end{alignat}

In our experiments, we choose to use $\srel$ as the primary way of defining and reporting the amount of averaging,
including the EDM baseline (\config{a}) that employs the traditional EMA (\autoref{eq:emaExponential}).
Given $\srel$, we can obtain the value of $\gamma$ to be used with \autoref{eq:powerEma} by solving a 3\textsuperscript{rd} order polynomial equation
  and taking the unique positive root
\begin{align}
  \frac{\gamma + 1}{(\gamma + 2)^2 \,(\gamma + 3)} &~=~ \srel^2 \\
  (\gamma + 2)^2 \,(\gamma + 3) - (\gamma + 1) \,\srel^{-2} &~=~ 0 \\
  \gamma^3 + 7\gamma^2 + \big(16 - \srel^{-2}\big)\gamma + \big(12 - \srel^{-2}\big) &~=~ 0 \label{eq:powerEmaGamma}
  \text{,}
\end{align}
which can be done using NumPy as shown in \autoref{algGamma}.
The requirement $\gamma > 0$ implies that $\srel < 12^{-0.5} \approx 0.2886$, setting an upper bound
  for the relative standard deviation.

\algGamma

Finally, to compute $\hat{\theta}$ efficiently during training, we note that the derivation
  of \autoref{eq:emaUpdate} does not depend on any particular properties of functions $f$ or $g$.
Thus, the update formula remains the same, and we only need to determine $\beta(\tnext)$ corresponding
  to our response function (\autoref{eq:powerEma}):
\begin{equation}
  \beta(\tnext) ~=~ \frac{g(\tcur)}{g(\tnext)} ~=~ \bigg( \frac{\tcur}{\tnext} \bigg)^{\!\!\gamma+1} \!=~ \bigg( 1 - \frac{\Delta t}{\tnext} \bigg)^{\!\!\gamma+1}
  \!\text{.}
\end{equation}
The only practical difference to traditional EMA is thus that $\beta(\tnext)$ is no longer constant 
  but depends on $\tnext$.

\subsection{Synthesizing novel EMA profiles after training}

Using \autoref{eq:emaUpdate}, it is possible to track the averaged weights for an
  arbitrary set of pre-defined EMA profiles during training.
However, the number of EMA profiles that can be handled this way is limited in practice by the associated
  memory and storage costs.
Furthermore, it can be challenging to select the correct profiles beforehand,
  given how much the optimal EMA length tends to vary between different
  configurations; see \refpaper{figEmaTriplet}a, for example.
To overcome these challenges, we will now describe a way to synthesize novel EMA profiles \emph{after} the training.

\vparagraph{Problem definition.}
Suppose that we have stored a number of snapshots \mbox{$\hat{\Theta} = \{\hat{\theta}_1, \hat{\theta}_2, \ldots, \hat{\theta}_N\}$} during training,
  each of them corresponding to a different response function $p_i(t)$.
We can do this, for example, by tracking $\smash{\hat{\theta}}$ for a couple of different choices of $\gamma$ (\autoref{eq:powerEma}) and saving them at regular intervals.
In this case, each snapshot $\smash{\hat{\theta}_i}$ will correspond to a pair $(t_i, \gamma_i)$ so that $p_i(t) = p_{t_i, \gamma_i}(t)$.

Let $p_r(t)$ denote a novel response function that we wish to synthesize.
The corresponding averaged weights are given by \autoref{eq:emaContinuous}:
\begin{equation}
  \hat{\theta}_r ~=~ \EE_{t \sim p_r(t)}\!\big[ \theta(t) \big] \label{eq:phemaTarget}
  \text{.}
\end{equation}

However, we cannot hope to calculate the precise value of $\smash{\hat{\theta}_r}$ based on $\smash{\hat{\Theta}}$ alone.
Instead, we will approximate it by $\smash{\hat{\theta}^\ast_r}$ that we define as a weighted average over the snapshots:
\begin{align}
  \hat{\theta}^\ast_r &~=~ {\sum}_i x_i \,\hat{\theta}_i \\
  &~=~ {\sum}_i x_i \,\EE_{t \sim p_i(t)} \!\big[ \theta(t) \big] \\
  &~=~ {\sum}_i x_i \int_{-\infty}^\infty p_i(t) \,\theta(t) \,\diff t \\
  &~=~ \int_{-\infty}^\infty \theta(t) \underbrace{{\sum}_i p_i(t) \,x_i}_{\eqqcolon\,p_r^\ast(t)} \diff t
  \text{,}
\end{align}
where the contribution of each $\smash{\hat{\theta}_i}$ is weighted by $x_i \in \RR$,
  resulting in the corresponding approximate response function $p_r^\ast(t)$.
Our goal is to select $\{x_i\}$ so that $p_r^\ast(t)$ matches the desired response function $p_r(t)$ as closely as possible.

For notational convenience, we will denote weights by column vector $\boldx = [x_1, x_2, \ldots, x_N]^\top \in \RR^N$
  and the snapshot response functions by $\boldp = [p_1, p_2, \ldots, p_N]$
  so that $\boldp(t)$ maps to the row vector $[p_1(t), p_2(t), \ldots, p_N(t)] \in \RR^N$.
This allows us to express the approximate response function as an inner product:
\begin{equation}
  p_r^\ast(t) ~=~ \boldp(t) \,\boldx
  \text{.}
\end{equation}

\vparagraph{Least-squares solution.}
To find the value of $\boldx$, we choose to minimize the $L_2$ distance between $p_r^\ast(t)$ and $p_r(t)$:
\begin{equation}
  \LL(\boldx) ~=~ {\big\lVert p_r^\ast(t) - p_r(t) \big\rVert}_2^2 ~=~ \int_{-\infty}^\infty\! \big( p_r^\ast(t) - p_r(t) \big)^2 \,\diff t \label{eq:phemaLoss}
  \text{.}
\end{equation}
Let us solve for the minimum of $\LL(\boldx)$ by setting its gradient with respect to $\boldx$ to zero:
\begin{align}
  \boldzero~&=~ \nnabla_{\boldx} \LL(\boldx) \\
  &=~ \nnabla_{\boldx} \bigg[ \int_{-\infty}^\infty\! \big( \boldp(t) \,\boldx - p_r(t) \big)^2 ~\diff t \bigg]  \\
  &=~ \int_{-\infty}^\infty\! \nnabla_{\boldx} \Big[ \big( \boldp(t) \,\boldx - p_r(t) \big)^2 \Big] \,\diff t \\
  &=~ \int_{-\infty}^\infty\! \big( \boldp(t) \boldx - p_r(t) \big) \nnabla_{\boldx} \Big[ \boldp(t) \boldx - p_r(t) \Big] \diff t \\
  &=~ \int_{-\infty}^\infty\! \big( \boldp(t) \,\boldx - p_r(t) \big) \,\boldp(t)^\top \,\diff t \\
  &=~ \int_{-\infty}^\infty\! \big( \boldp(t)^\top \boldp(t) \,\boldx - \boldp(t)^\top p_r(t) \big) \,\diff t \\
  &=~ \underbrace{\int_{-\infty}^\infty\! \boldp(t)^\top \boldp(t) \,\diff t}_{\eqqcolon\,\boldA} ~\boldx - \underbrace{\int_{-\infty}^\infty\! \boldp(t)^\top p_r(t) \,\diff t}_{\eqqcolon\,\boldb} \label{eq:phemaOptimum}
\end{align}
where we denote the values of the two integrals by matrix $\boldA \in \RR^{N \times N}$ and column vector $\boldb \in \RR^N$, respectively.
We are thus faced with a standard matrix equation $\boldA \boldx - \boldb = \boldzero$, from which we obtain the solution $\boldx = \boldA^{-1} \,\boldb$.

Based on \autoref{eq:phemaOptimum}, we can express the individual elements of $\boldA$ and $\boldb$
  as inner products between their corresponding response functions:
\begin{alignat}{4}
  \boldA &~=~ [a_{ij}], & ~~~a_{ij} &~=~ \big\langle p_i, p_j \big\rangle \\
  \boldb &~=~ [b_i]^\top, & ~~~b_i &~=~ \big\langle p_i, p_r \big\rangle
  \text{,}
\end{alignat}
\begin{equation}
  \text{where}~~
  \big\langle f, g \big\rangle ~=\, \int_{-\infty}^\infty f(x) \,g(x)\,\diff x \label{eq:functionInnerProduct}
  \text{.}
\end{equation}
In practice, these inner products can be computed for arbitrary EMA profiles using standard numerical methods, such as Monte Carlo integration.

\vparagraph{Analytical formulas for power function EMA profile.}

If we assume that $\{p_i\}$ and $p_r$ are all defined according to our power function EMA profile (\autoref{eq:powerEma}),
  we can derive an accurate analytical formula for the inner products (\autoref{eq:functionInnerProduct}).
Compared to Monte Carlo integration, this leads to a considerably faster and more accurate implementation.
In this case, each response function is uniquely defined by its associated $(t, \gamma)$.
In other words, $p_i(t) = p_{t_i, \gamma_i\hspace*{-.1em}}(t)$ and $p_r(t) = p_{t_r, \gamma_r\hspace*{-.1em}}(t)$.

Let us consider the inner product between two such response functions, i.e., \mbox{$\big\langle p_{t_a, \gamma_a}, p_{t_b, \gamma_b} \big\rangle$}.
Without loss of generality, we will assume that $t_a \le t_b$.
If this is not the case, we can simply flip their definitions, i.e., \mbox{$(t_a, \gamma_a) \leftrightarrow (t_b, \gamma_b)$}.
Now,
\begin{align}
  & \big\langle p_{t_a, \gamma_a}, p_{t_b, \gamma_b} \big\rangle \\[1mm]
  &~~=~ \int_{-\infty}^\infty p_{t_a, \gamma_a\!}(t) \,p_{t_b, \gamma_b\hspace*{-.1em}}(t) \,\diff t \\
  &~~=~ \int_0^{t_a} \frac{f_{\gamma_a\!}(t)}{g_{\gamma_a\!}(t_a)} \cdot \frac{f_{\gamma_b\hspace*{-.1em}}(t)}{g_{\gamma_b\hspace*{-.1em}}(t_b)} \,\diff t \\
  &~~=~ \frac{ 1 }{ g_{\gamma_a\!}(t_a) \,g_{\gamma_b\hspace*{-.1em}}(t_b) } \int_0^{t_a} \!f_{\gamma_a\!}(t) \,f_{\gamma_b\hspace*{-.1em}}(t) \,\diff t \\[1mm]
  &~~=~ \frac{ (\gamma_a+1) \,(\gamma_b+1) }{ t_a^{\gamma_a+1} \,t_b^{\gamma_b+1} } \int_0^{t_a} \!t^{\gamma_a+\gamma_b} \,\diff t \\[1mm]
  &~~=~ \frac{ (\gamma_a+1) \,(\gamma_b+1) }{ t_a^{\gamma_a+1} \,t_b^{\gamma_b+1} } \cdot \frac{ t_a^{\gamma_a+\gamma_b+1} }{ \gamma_a+\gamma_b+1 } \label{eq:phemaAnalyticA} \\[1mm]
  &~~=~ \frac{ (\gamma_a+1) \,(\gamma_b+1) \,(t_a / t_b)^{\gamma_b} }{ (\gamma_a+\gamma_b+1) \,t_b } \label{eq:phemaAnalyticB}
  \text{.}
\end{align}

Note that \autoref{eq:phemaAnalyticB} is numerically robust
  because the exponentiation by $\gamma_b$ is done for the ratio $t_a / t_b$ instead of being done directly for either $t_a$ or $t_b$.
If we used \autoref{eq:phemaAnalyticA} instead, we would risk floating point overflows even with 64-bit floating point numbers.

\algPhema

Solving the weights $\{x_i\}$ thus boils down to first populating the elements of $\boldA$ and $\boldb$
  using \autoref{eq:phemaAnalyticB} and then solving the matrix equation $\boldA \boldx = \boldb$.
\autoref{algPhema} illustrates doing this simultaneously for multiple target response functions using NumPy.
It accepts a list of $\{t_i\}$ and $\{\gamma_i\}$, corresponding to the input snapshots,
  as well as a list of $\{t_r\}$ and $\{\gamma_r\}$, corresponding to the desired target responses.
The return value is a matrix whose columns represent the targets while the rows represent the snapshots.

\vparagraph{Practical considerations.}
In all of our training runs, we track two weighted averages $\hat{\theta}_1$ and $\hat{\theta}_2$
  that correspond to \mbox{$\srel = 0.05$} and \mbox{$\srel = 0.10$}, respectively.
We take a snapshot of each average once every 8 million training images, i.e.,
  between 4096 training iterations with batch size 2048,
  and store it using 16-bit floating point to conserve disk space.
The duration of our training runs ranges between 671--2147 million training images,
  and thus the number of pre-averaged models stored in the snapshots ranges between 160--512.
We find that these choices lead to nearly perfect reconstruction in the range \mbox{$\srel \in [0.015, 0.250]$}.
Detailed study of the associated cost vs.~accuracy tradeoffs is left as future work.

\section{Implementation details}
\label{app:implementation}

\tabHyperparams

We implemented our techniques on top of the publicly available EDM~\cite{Karras2022elucidating} codebase.%
  \footnote{\url{https://github.com/NVlabs/edm}}
We performed our experiments on NVIDIA A100-SXM4-80GB GPUs using Python 3.9.16, PyTorch 2.0.0, CUDA 11.8, and CuDNN 8.9.4.
We used 32 GPUs (4 DGX A100 nodes) for each training run, and 8 GPUs (1 node) for each evaluation run.

\autoref{tabHyperparams} lists the full details of our main models featured in \refpaper{tabResultsFiveTwelve} and \refpaper{tabResultsSixtyFour}.
Our implementation and pre-trained models are available at {\small\url{https://github.com/NVlabs/edm2}}

\subsection{Sampling}

We used the 2\textsuperscript{nd} order deterministic sampler from EDM
  (i.e., Algorithm~1 in \cite{Karras2022elucidating})
  in all experiments with \mbox{$\sigma(t) = t$} and \mbox{$s(t) = 1$}.
We used the default settings \mbox{$\sigma_\text{min} = 0.002$}, \mbox{$\sigma_\text{max} = 80$}, and \mbox{$\rho = 7$}.
While we did not perform extensive sweeps over the number of
  sampling steps $N$,
  we found \mbox{$N = 32$} to yield sufficiently high-quality results for both ImageNet-512 and ImageNet-64.

In terms of guidance, we follow the convention used by Imagen~\cite{imagen}.
Concretely, we define a new denoiser $\hat{D}$ based on the primary conditional model $D_\theta$ and a secondary unconditional model $D_u$:
\begin{equation}
  \hat{D}(\xx; \sigma, \boldc) ~=~ w \,D_\theta(\xx; \sigma, \boldc) + (1-w) \,D_u(\xx; \sigma)
  \text{,}
\end{equation}
where $w$ is the guidance weight.
Setting $w=1$ disables guidance, i.e., $\hat{D} = D_\theta$, while increasing $w>1$ strengthens the effect.
The corresponding ODE is then given by
\begin{equation}
  \diff\xx ~=~ \frac{\xx - \hat{D}(\xx; \sigma, \boldc)}{\sigma} ~\diff\sigma
  \text{.}
\end{equation}

In \refpaper{tabResultsFiveTwelve} and \refpaper{tabResultsSixtyFour},
  we define NFE as the total number of times that $\hat{D}$ is evaluated during sampling.
In other words, we do not consider the number of model evaluations to be affected by the choice of $w$.

\subsection{Mixed-precision training}

In order to utilize the high-performance tensor cores available in NVIDIA Ampere GPUs,
  we use mixed-precision training in all of our training runs.
Concretely, we store all trainable parameters as 32-bit floating point (FP32) but temporarily cast them to 16-bit floating point (FP16) before evaluating the model.
We store and process all activation tensors as FP16, except for the embedding network and the associated per-block linear layers, where we opt for FP32 due to the low computational cost.
In \configs{a}{b}, our baseline architecture uses FP32 in the self-attention blocks as well, as explained in \autoref{app:configA}.

We have found that our models train with FP16 just as well as with FP32, as long as the loss function is scaled with an appropriate constant
  (see ``Loss scaling'' in \mbox{Figures~\ref{figConfigA}--\ref{figConfigG}}).
In some rare cases, however, we have encountered occasional FP16 overflows that can lead to a collapse in the training dynamics unless they are properly dealt with.
As a safety measure, we force the gradients computed in each training iteration to be finite by replacing \texttt{NaN} and \texttt{Inf} values with $0$.
We also clamp the activation tensors to range $[-256, +256]$ at the end of each encoder and decoder block.
This range is large enough to contain all practically relevant variation (see \autoref{figMagnitudesAll}).

\subsection{Training data}

We preprocess the ImageNet dataset exactly as in the ADM implementation%
  \footnote{%
  \href{https://github.com/openai/guided-diffusion/blob/22e0df8183507e13a7813f8d38d51b072ca1e67c/guided_diffusion/image_datasets.py\#L126}{\texttt{https://github.com/openai/guided-diffusion/blob/22e0}}
  \href{https://github.com/openai/guided-diffusion/blob/22e0df8183507e13a7813f8d38d51b072ca1e67c/guided_diffusion/image_datasets.py\#L126}{\hspace*{.1em}\texttt{df8183507e13a7813f8d38d51b072ca1e67c/guided\_diffusion/i}}
  \href{https://github.com/openai/guided-diffusion/blob/22e0df8183507e13a7813f8d38d51b072ca1e67c/guided_diffusion/image_datasets.py\#L126}{\hspace*{.1em}\texttt{mage\_datasets.py\#L126}}%
  }
  by \citet{Dhariwal2021} to ensure a fair comparison.
The training images are mostly non-square at varying resolutions. To obtain image data in square
aspect ratio at the desired training resolution, the raw images are processed as follows:

\begin{enumerate}
  \item Resize the shorter edge to the desired training resolution using bicubic interpolation.
  \item Center crop.
\end{enumerate}

\noindent During training, we do not use horizontal flips or any other kinds of data augmentation.

\subsection{FID calculation}

We calculate FID~\cite{Heusel2017} following the protocol used in EDM~\cite{Karras2022elucidating}:
We use 50,000 generated images and all available real images, without any augmentation such as horizontal flips.
To reduce the impact of random variation, typically in the order of $\pm$2\%,
  we compute FID three times in each experiment and report the minimum.
The shaded regions in FID plots show the range of variation among the three evaluations.

We use the pre-trained Inception-v3 model%
\footnote{%
  \href{https://api.ngc.nvidia.com/v2/models/nvidia/research/stylegan3/versions/1/files/metrics/inception-2015-12-05.pkl}{\texttt{https://api.ngc.nvidia.com/v2/models/nvidia/research}}
  \href{https://api.ngc.nvidia.com/v2/models/nvidia/research/stylegan3/versions/1/files/metrics/inception-2015-12-05.pkl}{\hspace*{.1em}\texttt{/stylegan3/versions/1/files/metrics/inception-2015-12-0}}
  \href{https://api.ngc.nvidia.com/v2/models/nvidia/research/stylegan3/versions/1/files/metrics/inception-2015-12-05.pkl}{\hspace*{.1em}\texttt{5.pkl}}
  }
provided with StyleGAN3~\cite{Karras2021alias}, which is a direct PyTorch translation of the original TensorFlow-based model.%
\footnote{%
  \href{http://download.tensorflow.org/models/image/imagenet/inception-2015-12-05.tgz}{\texttt{http://download.tensorflow.org/models/image/imagenet}}
  \href{http://download.tensorflow.org/models/image/imagenet/inception-2015-12-05.tgz}{\hspace*{.1em}\texttt{/inception-2015-12-05.tgz}}
}

\subsection{Model complexity estimation}

Model complexity (Gflops) was estimated using a PyTorch script that runs the model through \texttt{torch.jit.trace} to collect
the exact tensor operations used in model evaluation.  This list of \texttt{aten::*} ops and tensor
input and output sizes was run through an estimator that outputs the number of floating point operations required
for a single evaluation of the model.

In practice, a small set of operations dominate the cost of evaluating a model.  In the case of our largest (XXL) ImageNet-512 model, 
the topmost gigaflops producing ops are distributed as follows:

\newcommand{\atenbox}[1]{\makebox[13em][l]{\texttt{#1}}}
\begin{itemize}[leftmargin=3em]
  \item \atenbox{aten::\_convolution} 545.50 Gflops
  \item \atenbox{aten::mul} \s\s1.68 Gflops
  \item \atenbox{aten::div} \s\s1.62 Gflops
  \item \atenbox{aten::linalg\_vector\_norm} \s\s1.54 Gflops
  \item \atenbox{aten::matmul} \s\s1.43 Gflops
\end{itemize}

Where available, results for previous work listed in \refpaper{tabResultsFiveTwelve} were obtained from
their respective publications.  In cases where model complexity was not publicly available, we used our PyTorch estimator
to compute a best effort estimate.  We believe our estimations are accurate to within 10\% accuracy.

\subsection{Per-layer sensitivity to EMA length}

List of layers included in the sweeps of \refpaper{figPerTensor} in the main paper are listed below. The analysis only includes  weight tensors\,---\,not biases, group norm scale factors, or affine layers' learned gains.
\\[-.5\baselineskip]

{\footnotesize\perTensorListAppendix}

\section{Negative societal impacts}

Large-scale image generators such as DALL$\cdot$E 3, Stable Diffusion XL, or MidJourney can have various negative societal effects, including types of disinformation or emphasizing sterotypes and harmful biases~\cite{Mishkin2022risks}.
Our advances to the result quality can potentially further amplify some of these issues. Even with our efficiency improvements, the training and sampling of diffusion models continue to require a lot of electricity, potentially contributing to wider issues such as climate change.

\figClassGridA{}
\figClassGridB{}
\figClassGridC{}

\fi

\end{document}